\PassOptionsToPackage{table}{xcolor}
\documentclass[acmtog,nonacm]{acmart}

\usepackage{booktabs} %

\usepackage{tabularray}
\usepackage{multirow}
\usepackage{graphicx}
\usepackage{subfig}
\usepackage{enumitem}

\definecolor{lightgreen}{RGB}{245, 255, 245}
\definecolor{lightyellow}{RGB}{255, 255, 245}
\definecolor{lightred}{RGB}{255, 245, 245}

\usepackage{bm}
\newcommand{\mb}[1]{\mathbf{#1}}

\copyrightyear{2025}
\acmYear{2025}
\setcopyright{acmlicensed}\acmConference[SIGGRAPH Conference Papers '25]{Special Interest Group on Computer Graphics and Interactive Techniques Conference Conference Papers }{August 10--14, 2025}{Vancouver, BC, Canada}
\acmBooktitle{Special Interest Group on Computer Graphics and Interactive Techniques Conference Conference Papers (SIGGRAPH Conference Papers '25), August 10--14, 2025, Vancouver, BC, Canada}
\acmDOI{10.1145/3721238.3730668}
\acmISBN{979-8-4007-1540-2/2025/08}

\begin{document}
\title{Reenact Anything: Semantic Video Motion Transfer Using Motion-Textual Inversion}

\author{Manuel Kansy}
\orcid{0000-0002-4238-7560}
\affiliation{%
	\institution{ETH Zürich}
	\city{Zürich}
	\country{Switzerland}}
\affiliation{%
	\institution{DisneyResearch|Studios}
	\city{Zürich}
	\country{Switzerland}}
\email{manuel.kansy@disneyresearch.com}
\author{Jacek Naruniec}
\orcid{0000-0002-2315-5166}
\affiliation{%
	\institution{DisneyResearch|Studios}
	\city{Zürich}
	\country{Switzerland}
}
\email{jacek.naruniec@disneyresearch.com}
\author{Christopher Schroers}
\orcid{0000-0003-1473-1878}
\affiliation{%
	\institution{DisneyResearch|Studios}
	\city{Zürich}
	\country{Switzerland}
}
\email{christopher.schroers@disneyresearch.com}
\author{Markus Gross}
\orcid{0009-0003-9324-779X}
\affiliation{%
	\institution{ETH Zürich}
	\city{Zürich}
	\country{Switzerland}}
\affiliation{%
	\institution{DisneyResearch|Studios}
	\city{Zürich}
	\country{Switzerland}}
\email{grossm@inf.ethz.ch}
\author{Romann M. Weber}
\orcid{0000-0003-1196-5425}
\affiliation{%
	\institution{DisneyResearch|Studios}
	\city{Zürich}
	\country{Switzerland}
}
\email{romann.weber@disneyresearch.com}

\begin{abstract}

Recent years have seen a tremendous improvement in the quality of video generation and editing approaches. While several techniques focus on editing appearance, few address motion. Current approaches using text, trajectories, or bounding boxes are limited to simple motions, so we specify motions with a single motion reference video instead. We further propose to use a pre-trained \underline{image}-to-video model rather than a \underline{text}-to-video model. This approach allows us to preserve the exact appearance and position of a target object or scene and helps disentangle appearance from motion.

Our method, called \emph{motion-textual inversion}, leverages our observation that image-to-video models extract appearance mainly from the (latent) image input, while the text/image embedding injected via cross-attention predominantly controls motion. We thus represent motion using text/image embedding tokens. By operating on an inflated motion-text embedding containing multiple text/image embedding tokens per frame, we achieve a high temporal motion granularity. Once optimized on the motion reference video, this embedding can be applied to various target images to generate videos with semantically similar motions.

Our approach does not require spatial alignment between the motion reference video and target image, generalizes across various domains, and can be applied to various tasks such as full-body and face reenactment, as well as controlling the motion of inanimate objects and the camera. We empirically demonstrate the effectiveness of our method in the semantic video motion transfer task, significantly outperforming existing methods in this context.

Project website: \textbf{\url{https://mkansy.github.io/reenact-anything/}}

\end{abstract}

\begin{teaserfigure}
    \centering
    \includegraphics[trim = 5mm 39mm 98mm 74mm, width=\textwidth, clip]{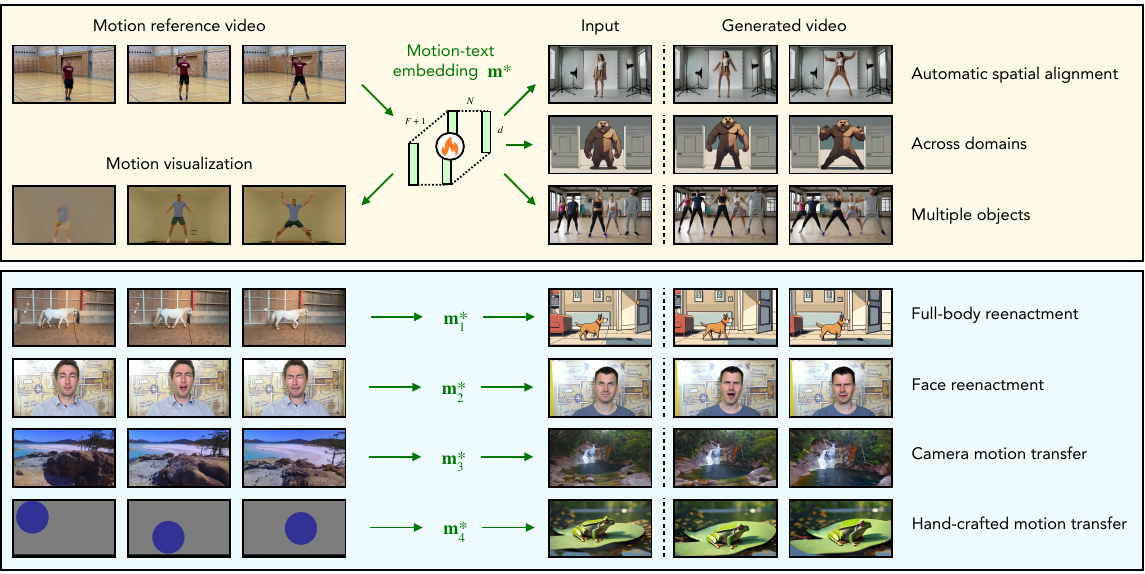}%
    \caption{We encode the motion of a reference video into a novel motion-text embedding using a frozen, pre-trained image-to-video diffusion model. This optimized motion-text embedding can then be applied to different starting images to generate videos with semantically similar motions. The general nature of our motion representation allows for successful motion transfer even when objects are not spatially aligned, across various domains, and for multiple objects. Additionally, our method supports multiple types of motions, including full-body, face, camera, and even hand-crafted motions. \\
    Please refer to \url{https://mkansy.github.io/reenact-anything/} for corresponding videos for all figures of this paper.
    }
    \Description{A figure made up of two blocks. In the top block, the rough model architecture is shown. A motion-text embedding is extracted from a motion reference of a person doing jumping jacks, and this motion is then applied to three different input images and to an unconditional appearance (motion visualization). The generated videos all show a jumping jack motion even if the objects are not spatially aligned, from different domains and for multiple objects. In the bottom block, a number of motion input and generated output videos are shown for full-body reenactment, face reenactment, hand-crafted motions, and camera motions.}
    \label{fig:teaser}
\end{teaserfigure}

\maketitle

\section{Introduction} \label{sec:introduction}

The ability to generate and edit videos has rapidly advanced thanks to diffusion models, enabling applications in filmmaking, marketing, and beyond. However, controlling \textit{how} objects move in generated videos---the semantics of motion---remains challenging and largely underexplored. Many existing methods excel at editing \textit{appearance} but struggle to intuitively control \textit{motion}.
For example, even state-of-the-art image-to-video models like Stable Video Diffusion~\cite{svd} offer little control over motion, i.e., only by modifying the random seed or adjusting micro-conditioning inputs like frame rate, neither of which is easily interpretable.

To make motion control more intuitive, we propose a new task: semantic video motion transfer from a reference video to a target image. Specifically, we aim to generate a video that replicates the semantic motion of a motion reference video while preserving the appearance and spatial layout of a target image. Crucially, we do not aim to copy pixel-wise trajectories but rather to transfer the meaning of the motion, even when objects are misaligned — for instance, producing a subject performing jumping jacks on the left side of the frame even if the motion reference was centered.

We identify two key challenges for this task: appearance leakage from the motion reference video and object misalignment. To tackle appearance leakage, we employ an \underline{image}-to-video rather than a \underline{text}-to-video model and do not fine-tune the model. To the best of our knowledge, we are the first to use an image-to-video model for general motion transfer. To address object misalignments between the motion reference video and the target image, we introduce a novel motion representation that eliminates the need for spatial alignment by not having a spatial dimension in the first place.

Our motion representation is based on our observation that image-to-video models extract the appearance predominantly from the image (latent) input, whereas the text/image embedding injected via cross-attention mostly controls the motion. We therefore propose to represent motion with several text/image embedding tokens, together referred to as \emph{motion-text embedding}, that we optimize on a given motion reference video. Thereby, our inflated motion-text embedding enables us to preserve the timing of the motion video very precisely, which is crucial for applications such as visual dubbing. Our approach, named \emph{motion-textual inversion}, is general in nature and works for various types of motions and objects.\footnote{Independently, a concurrent work, LEAD~\cite{lead}, introduced the term \emph{motion textual inversion} to describe their approach of applying textual inversion~\cite{textual_inversion} to a text-to-motion model. While the names are similar, the underlying methods differ significantly.} Perhaps surprising at first, it turns out that while words are not ideal for describing motions, their embeddings can describe motions exceptionally well. Fig.~\ref{fig:teaser} shows exemplary results of our method, including motion transfers to multiple (misaligned) objects.

To summarize, our contributions are:
\begin{enumerate}
    \item We introduce the semantic video motion transfer task in an image-to-video setting.
    \item We observe that text/image embeddings of image-to-video diffusion models store and affect motion and leverage them as a general and compact motion representation.
    \item We propose \textit{motion-textual inversion}, a novel method that optimizes multiple text/image embedding tokens on a motion reference video and transfers the learned motion to target images.
    \item We demonstrate superior performance over existing motion transfer approaches.
\end{enumerate}

\section{Related Work} \label{sec:related_work}

Our goal is to develop a general reenactment method that requires no large-scale domain-specific training. Given the impressive cross-domain translation capabilities of diffusion models~\cite{prompt_to_prompt, plug_and_play, pix2pix-zero} and the rise of video generation models~\cite{svd, lumiere, sora, cogvideox, hunyuanvideo, videojam}, we employ a diffusion-based video model for our general task to capitalize on its broad and general priors. In contrast, the most related non-diffusion methods, JOKR~\cite{jokr} and AnaMoDiff~\cite{anamodiff}, operate under more constrained conditions, typically requiring a target video, assuming mostly planar 2D motions, and lacking support for natural backgrounds.

In the following sections, we focus on video motion editing approaches based on video diffusion models. In Section~\ref{sec:related_work_extra}, we discuss additional related works on domain-specific reenactment~\cite{fsgan, wang2021one, hsu2022dual, megaportraits, li2023one, liveportrait, transmomo, everybody_dance_now, follow_your_pose, dreampose, motioneditor, champ, edit-your-motion, vividpose, motionfollower}, keypoint-based motion transfer~\cite{ni2023cross, fomm, mraa, tpsmm, hedlin2023unsupervised, dift, diffusion-hyperfeatures, sd-complements-dino, telling-left-from-right}, image and video generation~\cite{dalle2, imagen, animatediff, modelscope}, and the inversion-then-generation framework~\cite{pix2video, zeroshot, rerender_a_video, video_p2p, tokenflow, make-a-protagonist, genvideo, motionflow, null_text_inversion, ReNoise, MOFT, DiTFlow}.

\subsection{Video Motion Editing with Explicit Motions}

Existing methods for controlling motion with sparse control signals like text~\cite{dreamix, moca, animateanything, animateyourmotion}, boxes~\cite{boximator, animateyourmotion, trailblazer, motion_zero, peekaboo}, trajectories~\cite{MCDiff, dragnuwa, draganything, ReVideo, MOFA-video, FreeTraj, image_conductor, puppet-master, trackgo, motionprompting}, keypoints~\cite{videoswap, anamodiff, MOFA-video}, or camera motions~\cite{cameractrl, vd3d, CamTrol, CamCo, motionmaster, cami2v, cheong2024boosting, motionctrl, direct_a_video, motionbooth, puppet-master} are limited to simple motions in most practical scenarios and may require manual prompting. On the other hand, dense motion trajectories~\cite{videocomposer, diffusion_as_shader, go_with_the_flow, controlvideo, control_a_video} may leak the motion reference video's spatial structure, thus often failing in unaligned scenarios.

\subsection{Video Motion Editing with Implicit Motions}

In contrast to the methods discussed above, the methods in this section use less interpretable motion representations. Specifically, fine-tuning approaches encode motions in model weights, and inversion-then-generation approaches extract motions from model features or attention maps.

\subsubsection{Fine-Tuning} \label{sec:related_work-fine-tuning}

Approaches based on fine-tuning~\cite{tune_a_video, materzynska2024newmove, motiondirector, dreamvideo, motioncrafter, ren2024customize, CustomTTT, VMC} involve fine-tuning a model on one or several motion reference videos, similar to DreamBooth~\cite{dreambooth}. The methods primarily differ in the parts of the model they fine-tune and the techniques they use, such as LoRA~\cite{lora}, to train only the components responsible for motion. However, in practice, they often inadvertently learn the reference video's appearance as well, which can hinder generalization to new target object appearances. We make a similar observation to \citet{lamp}, namely that conditioning the diffusion model on the image helps the model concentrate on learning motion.

\subsubsection{Inversion-then-Generation} \label{sec:related_work-inversion-then-generation}

Approaches based on the inversion-then-generation paradigm~\cite{uniedit, space_time_diffusion, motionclone} extract model features such as attention maps from the motion reference video (e.g., via DDIM inversion~\cite{ddim}), which are then incorporated into the diffusion process of the generated video. This helps replicate the reference video's structure in the output. However, these approaches struggle when there are significant differences between the locations and geometries of the reference and target objects, leading to misaligned semantic features being injected or enforced.

\subsubsection{With Different Spatial Layout} \label{sec:related_work-spatial-variations}

Most of the one-shot reference-based methods produce videos with motions that are mostly spatially aligned with the motion reference video, i.e., they follow the layout as well as the subject scale and position of the reference video. We thus argue that many of these works~\cite{motioncrafter, VMC, space_time_diffusion} can be considered as an advanced form of appearance transfer rather than motion transfer. We focus on the general case where layouts may not align, a less explored scenario. Unlike existing methods~\cite{lamp, motiondirector, dreamvideo, materzynska2024newmove}, which use multiple motion videos to avoid overfitting to a single layout, we transfer motion from a single reference video with precise temporal alignment. Also, instead of relying on text to loosely define the subject's appearance~\cite{materzynska2024newmove, wang2024motion, ren2024customize, motrans}, we aim to generate videos that seamlessly continue from a given target image. Concurrently, \citet{wang2024motion} propose an approach that also learns a motion embedding while keeping the model frozen, but they do not incorporate a target image and appear to overfit to the reference video's layout.

\section{Method}

We propose to transfer the semantic motion of a motion reference video to a given target image by \emph{motion-textual inversion}. We thereby optimize a set of text/image embedding tokens, which we refer to as \emph{motion-text embedding}, for the motion reference video using a pre-trained image-to-video diffusion model. 

\subsection{Preliminaries} \label{sec:method_preliminaries}

\subsubsection{Diffusion} 

Diffusion models~\cite{ddpm1, song2020score} consist of two processes. In the \textit{forward process}, Gaussian noise is iteratively added to a clean data sample $\mb{x}_0$ until it is approximately pure noise. In the \textit{reverse process}, starting with pure noise $\mb{x}_T$, a learnable denoiser $D_{\bm{\theta}}$ iteratively removes noise to obtain a sample that matches the original data distribution $p_\text{data}$. We follow the continuous-time framework~\cite{song2020score, edm}, where the denoiser is trained via \textit{denoising score matching}: 
\begin{equation} \label{eq:standard-diffusion}
    \mathbb{E}_{(\mb{x}_0, \mb{c}) \sim p_\text{data}(\mb{x}_0, \mb{c}),(\sigma, \mb{n}) \sim p(\sigma, \mb{n})}[\lambda_\sigma ||D_{\bm{\theta}}(\mb{x}_0 + \mb{n}; \sigma, \mb{c}) - \mb{x}_0||^2_2],
\end{equation}
where $\mb{x}_0$ is a clean data sample and $\mb{c}$ an arbitrary conditioning signal from the original data distribution $p_\text{data}$; $p(\sigma, \mb{n}) = p(\sigma)\mathcal{N}(\mb{n}; \mb{0}, \sigma^2)$, where $p(\sigma)$ is a probability distribution over noise levels $\sigma$, and $\mb{n}$ is noise; and $\lambda_\sigma: \mathbb{R}_+ \rightarrow \mathbb{R}_+$ is a weighting function. The denoiser $D_{\bm{\theta}}$ is parameterized as
\begin{equation} \label{eq:edm-parametrization}
    D_{\bm{\theta}}(\mb{x}; \sigma) = c_\text{skip}(\sigma)\mb{x} + c_\text{out}(\sigma)F_{\bm{\theta}}(c_\text{in}(\sigma)\mb{x}; c_\text{noise}(\sigma)),
\end{equation}
where $F_{\bm{\theta}}$ is the neural network to be trained; $c_\text{skip}(\sigma)$ modulates the skip connection; $c_\text{out}(\sigma)$ and $c_\text{in}(\sigma)$ scale the output and input magnitudes respectively; and $c_\text{noise}(\sigma)$ maps noise level $\sigma$ into a conditioning input for $F_{\bm{\theta}}$. For more details, please refer to EDM~\cite{edm}.

\subsubsection{Latent Diffusion}

Latent diffusion models~\cite{ldm} operate in the latent space rather than in pixel space to reduce computation and thus enable higher resolutions. First, an encoder $\mathcal{E}$ produces a compressed latent $\mb{z} = \mathcal{E}(\mb{x})$. Then, we perform the diffusion process over $\mb{z}$. Lastly, a decoder $\mathcal{D}$ reconstructs the latent features back into pixel space.\footnote{To maintain consistency in notation, we use $\mb{x}$ for the diagrams and method description, even though the diffusion process actually occurs in latent space.}

\begin{figure*}
	\centering
	\begin{tblr}{
			cell{1}{1} = {c=2}{c},
			cell{1}{3} = {c=2}{c},
			vline{2} = {2-2}{dashed},
			vline{3} = {1-2}{},
			vline{4} = {2-2}{dashed},
		}
		Text input: ``A \underline{white} horse walking.'' & & Text input: ``A \underline{pink} horse walking.'' & \\
		\includegraphics[width=0.09\textwidth]{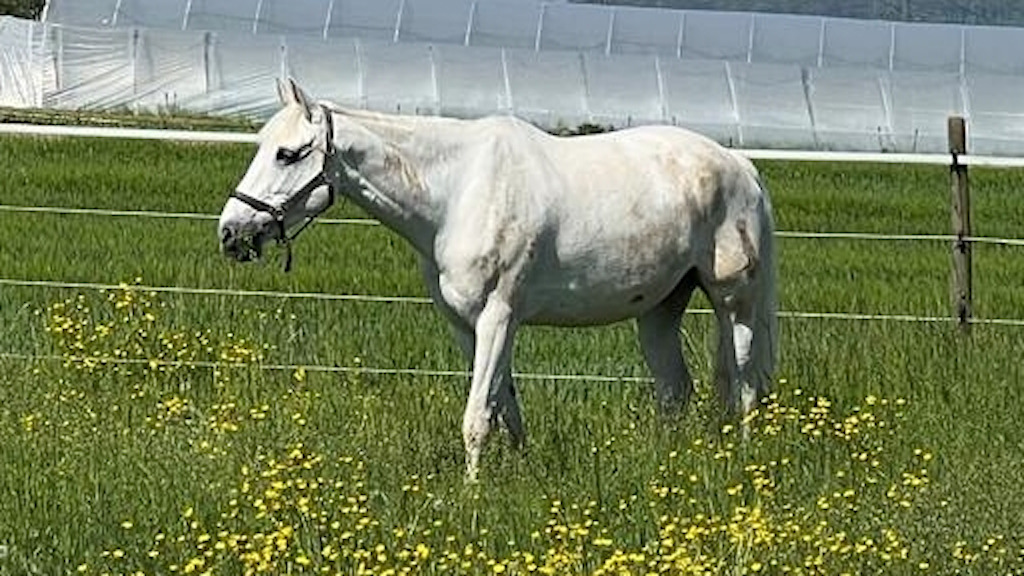} & \includegraphics[width=0.09\textwidth]{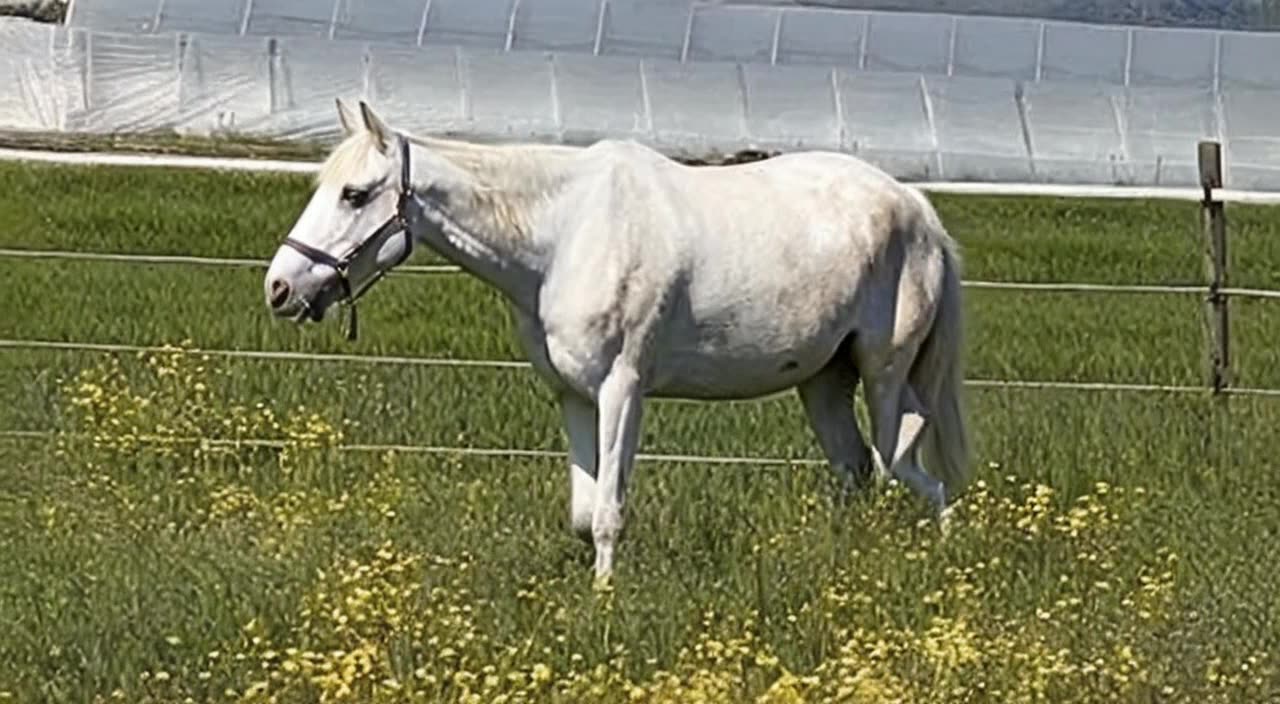} \includegraphics[width=0.09\textwidth]{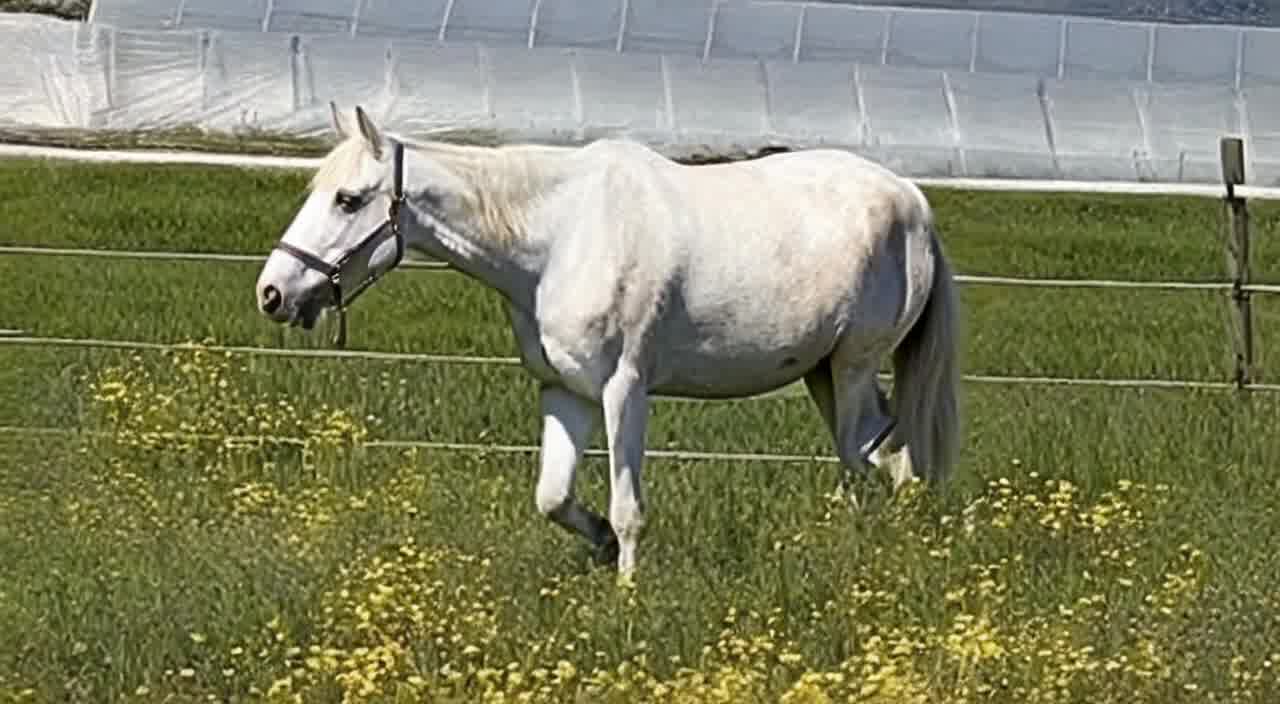} \includegraphics[width=0.09\textwidth]{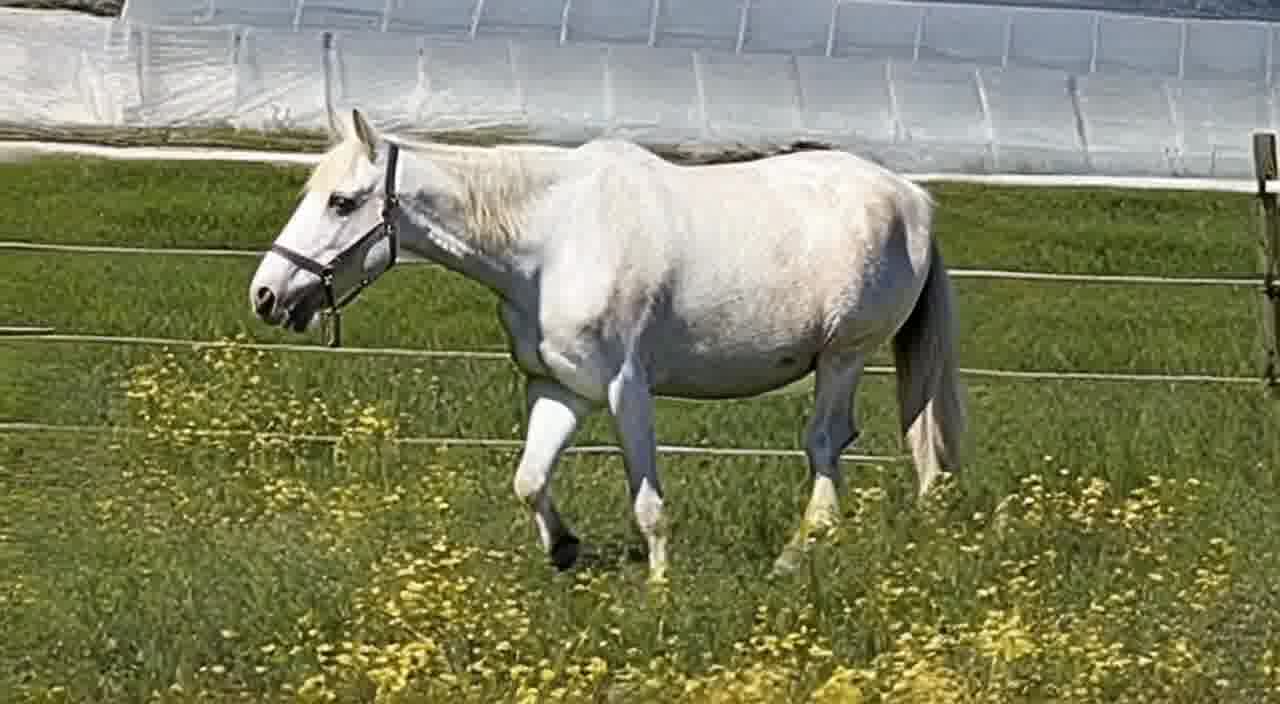} & \includegraphics[width=0.09\textwidth]{figures/i2v_ignores_text/white_horse.jpg} & \includegraphics[width=0.09\textwidth]{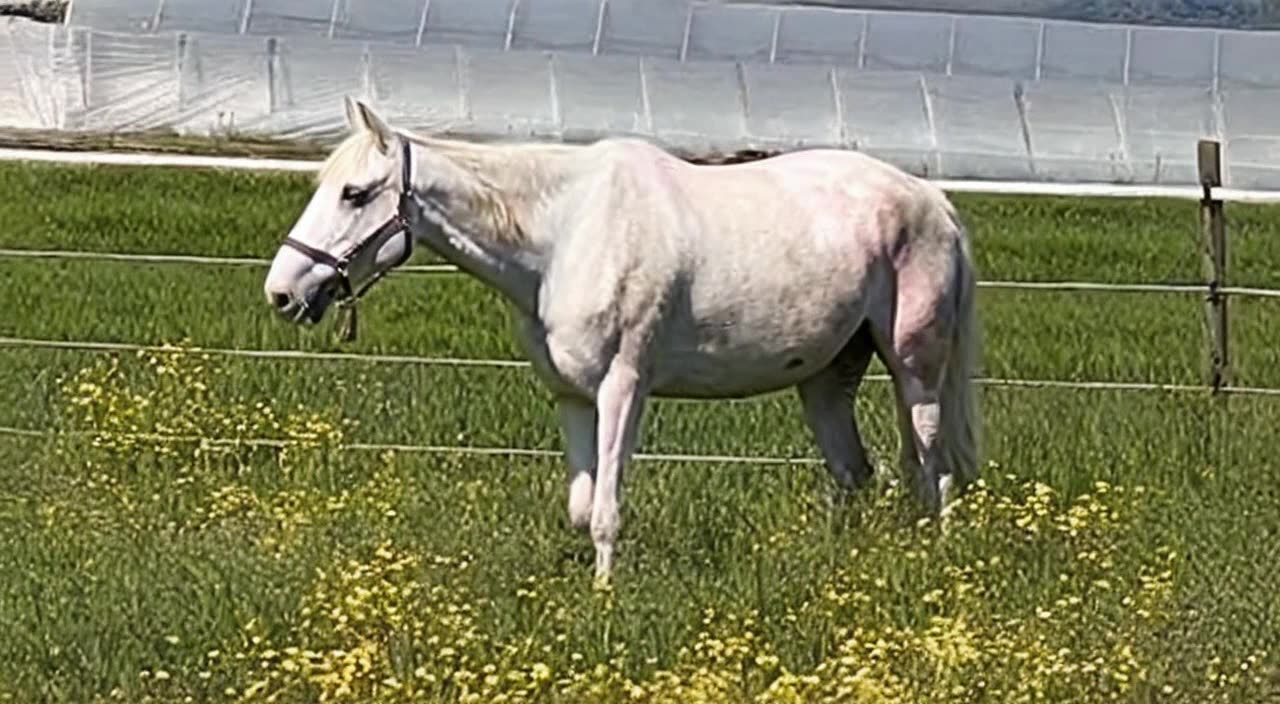} \includegraphics[width=0.09\textwidth]{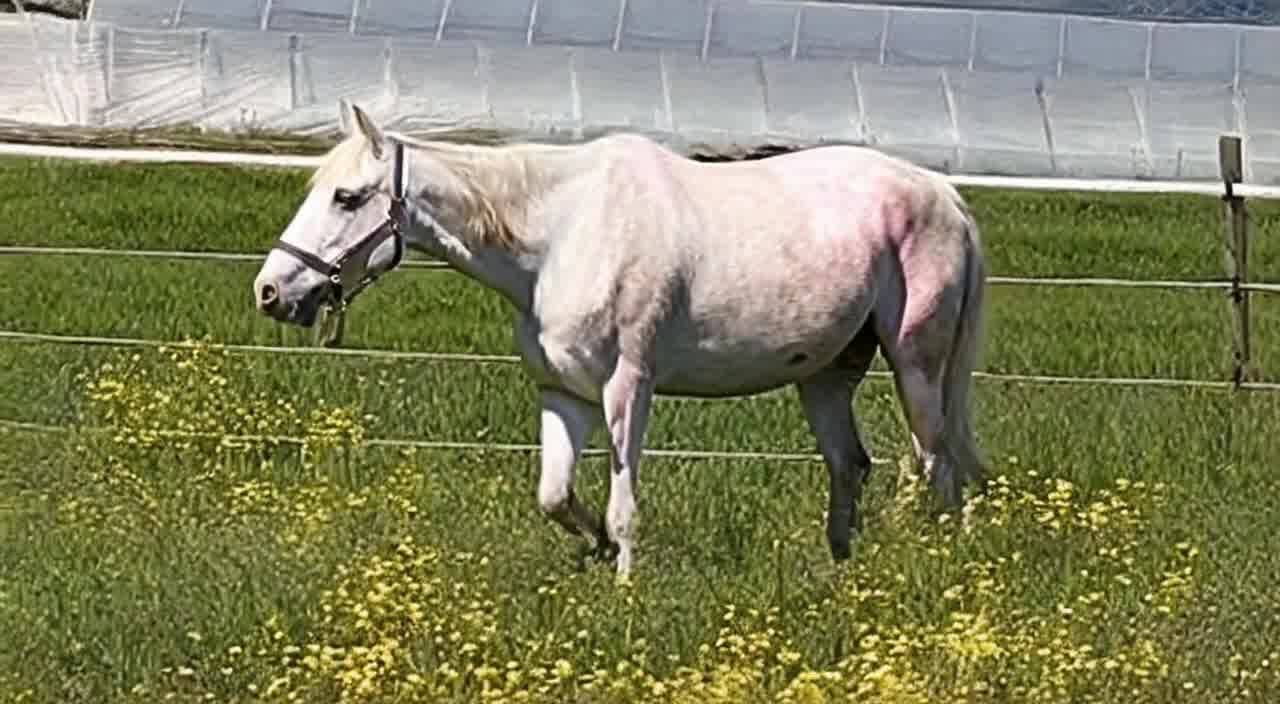} \includegraphics[width=0.09\textwidth]{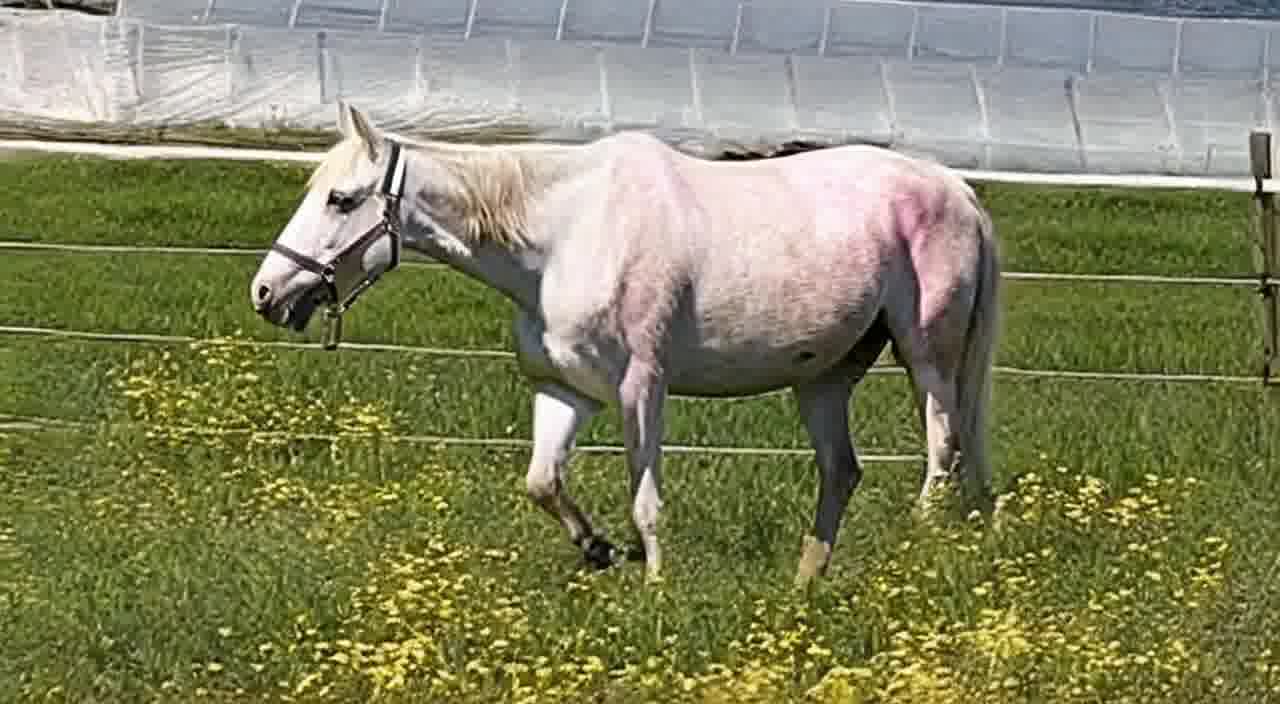}
	\end{tblr}
	\caption{Observation 1. In image-to-video models, the image input primarily dictates the appearance of the generated videos. For example, I2VGen-XL~\cite{i2vgen_xl} generates a video of a predominantly white horse from a white horse image, even when the input text specifies the horse's color as ``pink.''}
	\Description{Two subfigures. The left subfigure shows the input image of a white horse followed by three frames of a video where this white horse is walking. The right subfigure shows the same input image of a white horse followed by three frames of a video where the horse looks almost the same as in the left subfigure, so still mostly white but with a tiny hint of pink.}
	\label{fig:i2v_ignores_text}
\end{figure*}

\begin{figure*}
	\centering
	\begin{tblr}{
			cell{1}{3} = {c=2}{c},
			cell{1}{5} = {c=2}{c},
			cell{2}{1} = {r=2}{c},
			cell{2}{2} = {c},
			cell{3}{2} = {c},
			vline{4} = {2-3}{dashed},
			vline{5} = {1-3}{},
			vline{6} = {2-3}{dashed},
			hline{3} = {2-6}{},
		}
		&      & CLIP image embedding: Real horse &  & CLIP image embedding: Toy horse & \\
		{Image\\ latent} & {Real\\ horse} & \raisebox{-.5\height}{\includegraphics[width=0.09\textwidth]{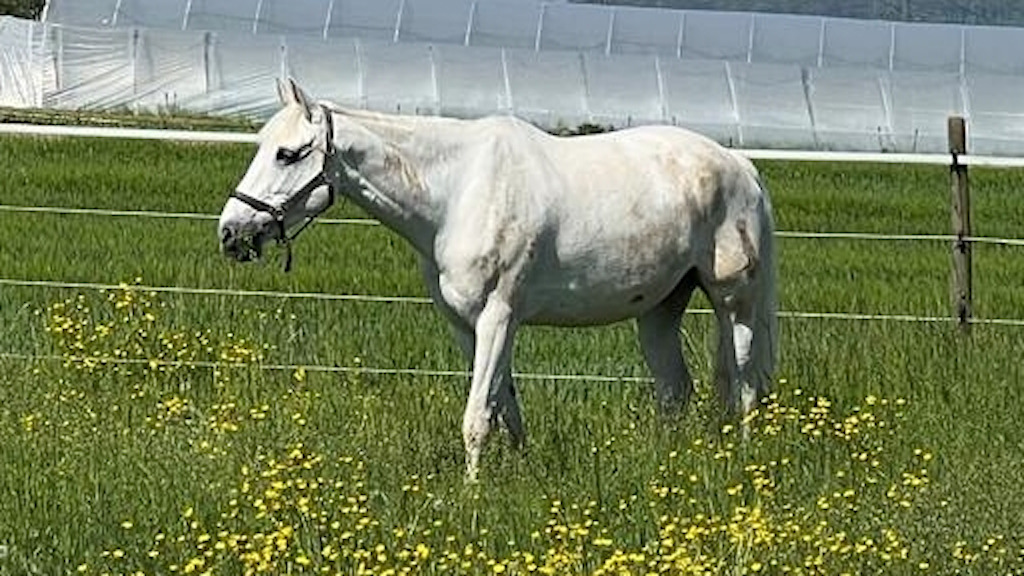}} & \raisebox{-.5\height}{\includegraphics[width=0.09\textwidth]{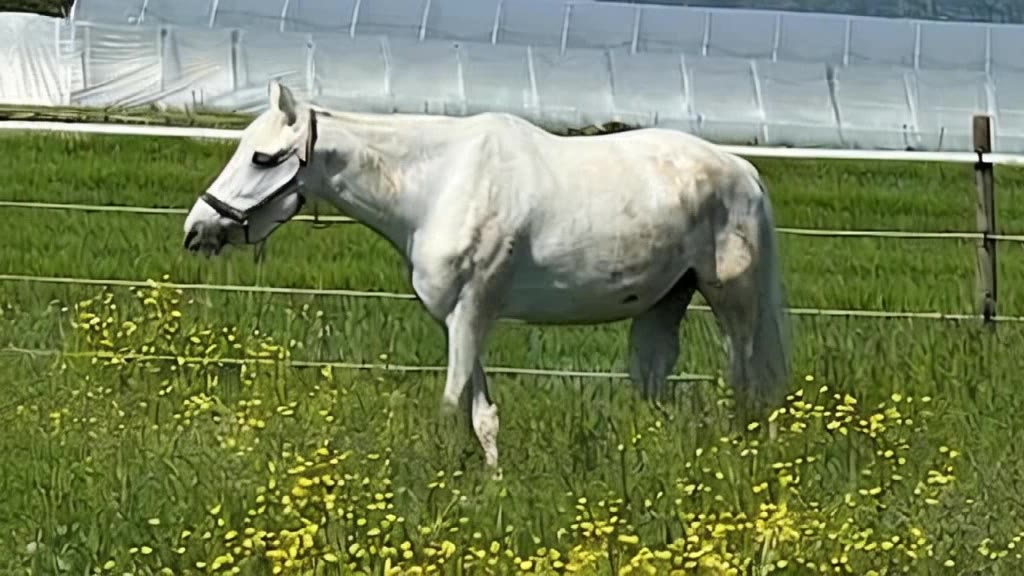}} \raisebox{-.5\height}{\includegraphics[width=0.09\textwidth]{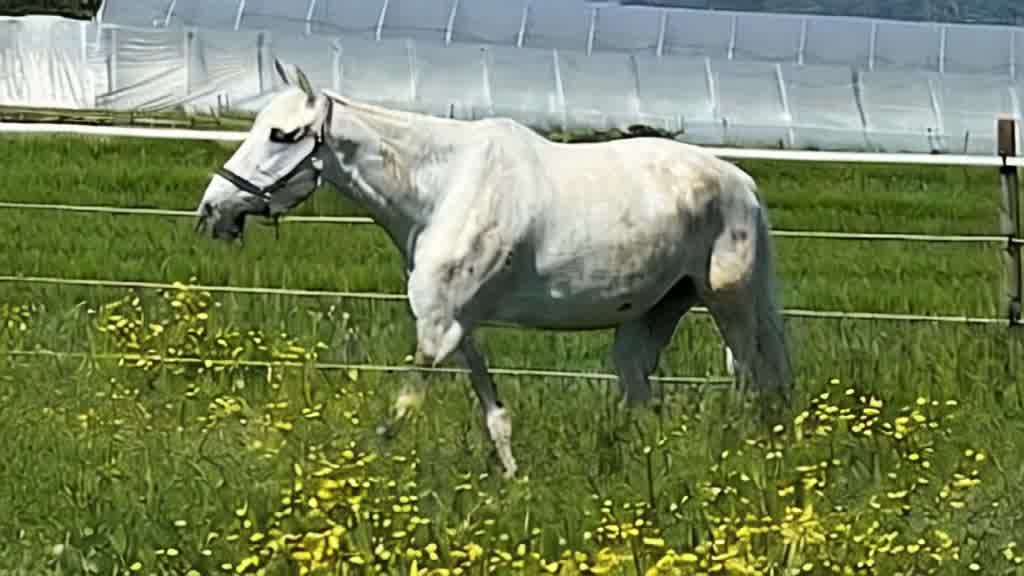}} \raisebox{-.5\height}{\includegraphics[width=0.09\textwidth]{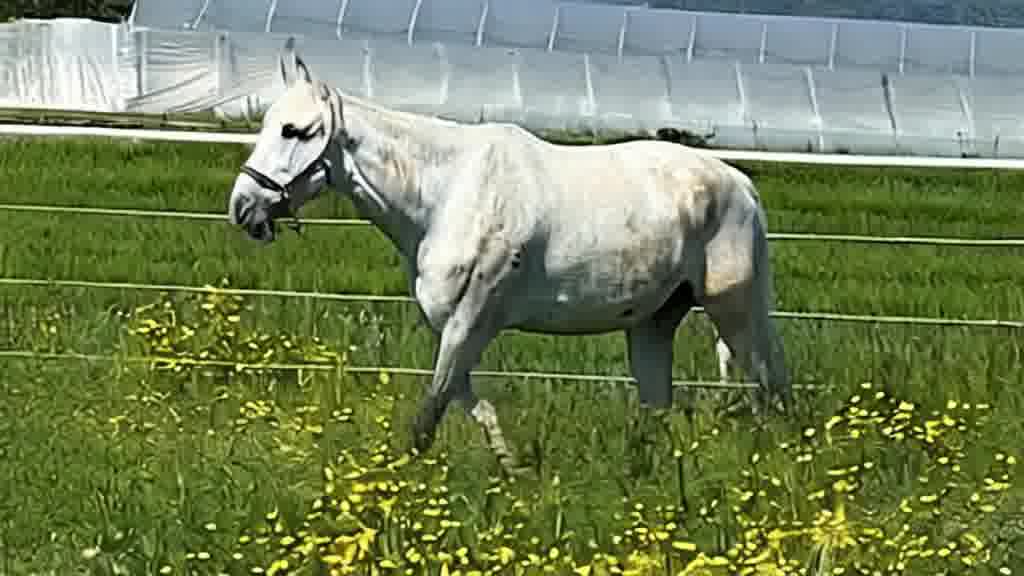}} & \raisebox{-.5\height}{\includegraphics[width=0.09\textwidth]{figures/horse_clip_swap_1024_576/white_horse.jpg}} & \raisebox{-.5\height}{\includegraphics[width=0.09\textwidth]{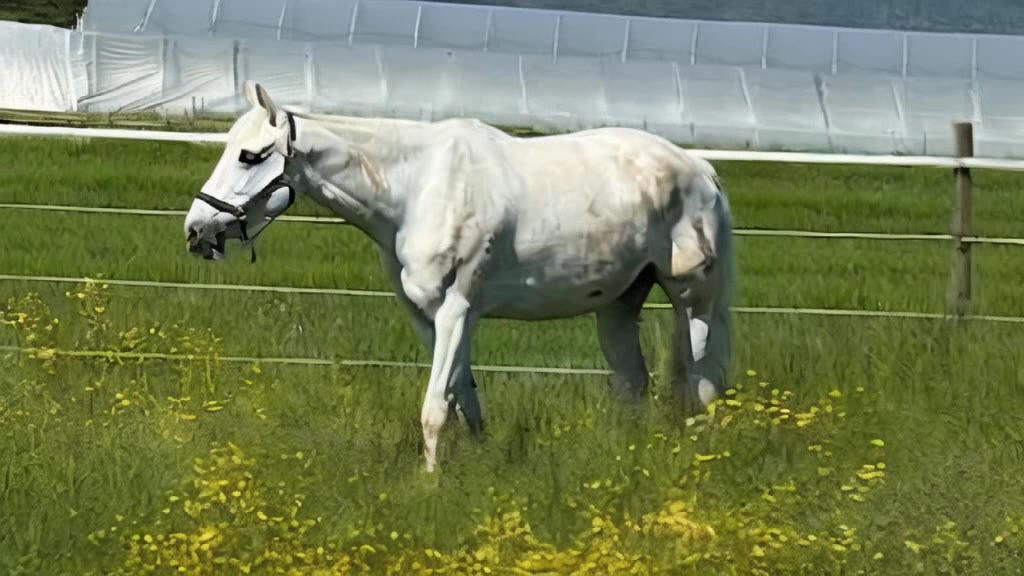}} \raisebox{-.5\height}{\includegraphics[width=0.09\textwidth]{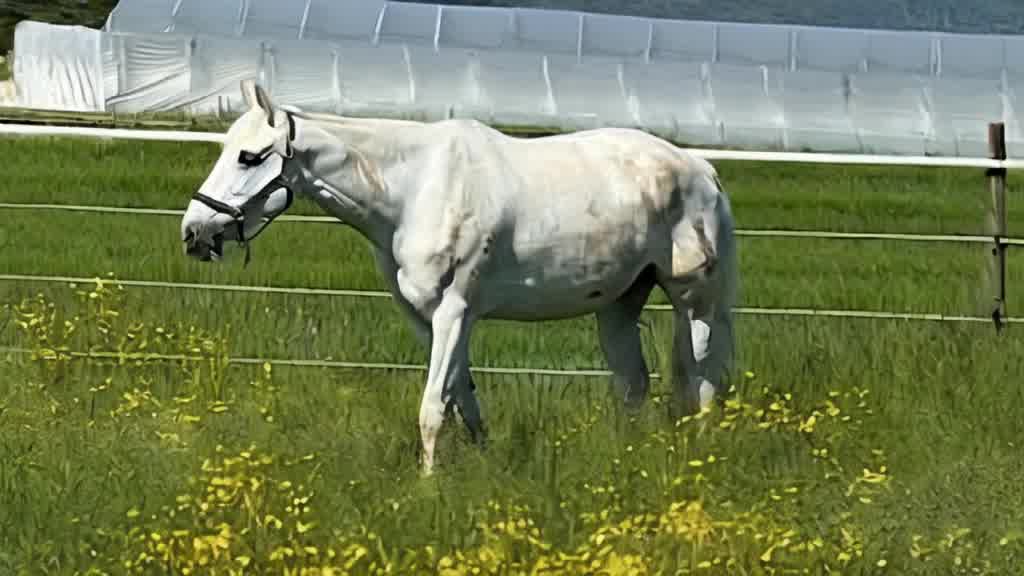}} \raisebox{-.5\height}{\includegraphics[width=0.09\textwidth]{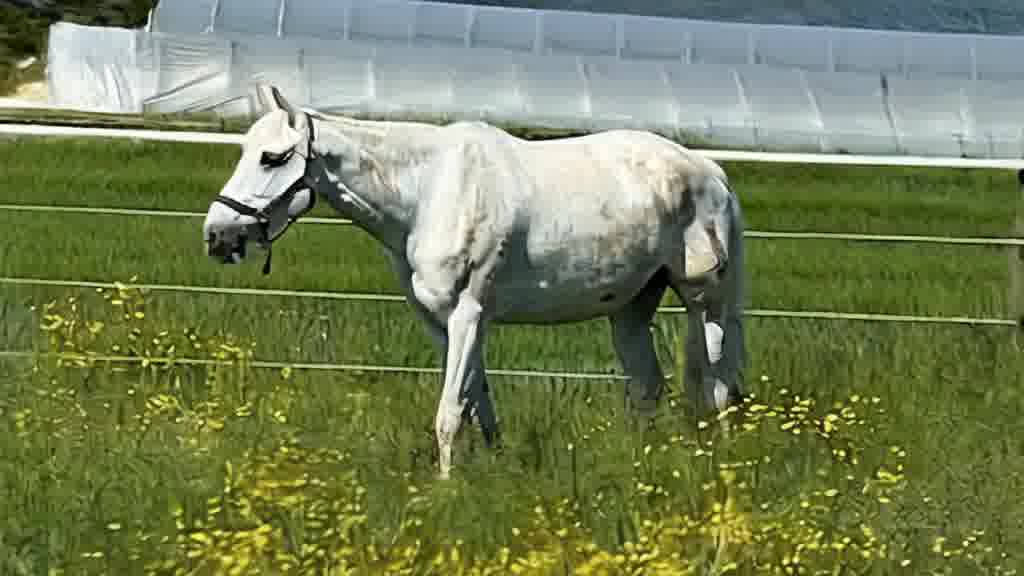}} \\
		& {Toy\\ horse}  & \raisebox{-.5\height}{\includegraphics[width=0.09\textwidth]{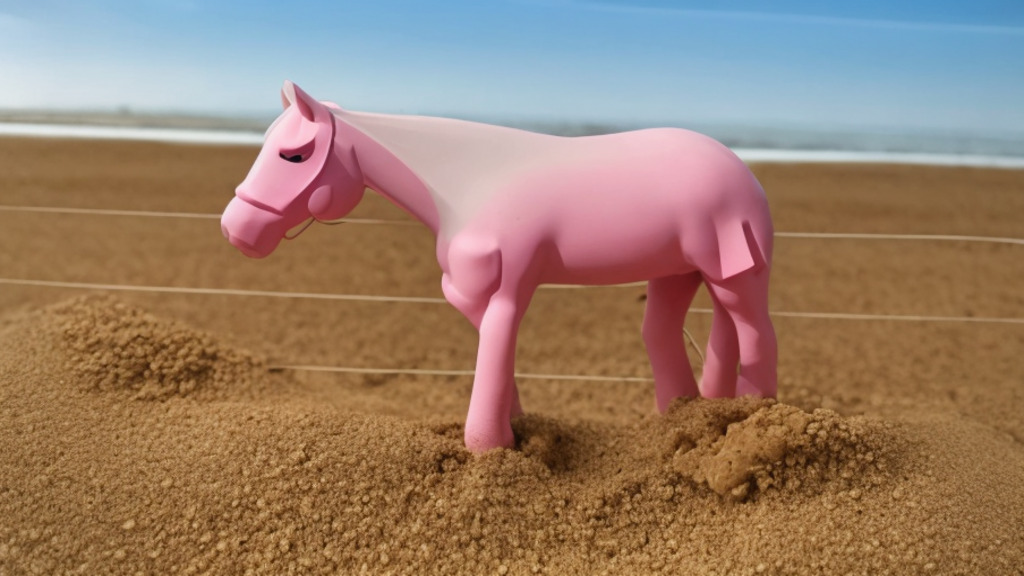}} & \raisebox{-.5\height}{\includegraphics[width=0.09\textwidth]{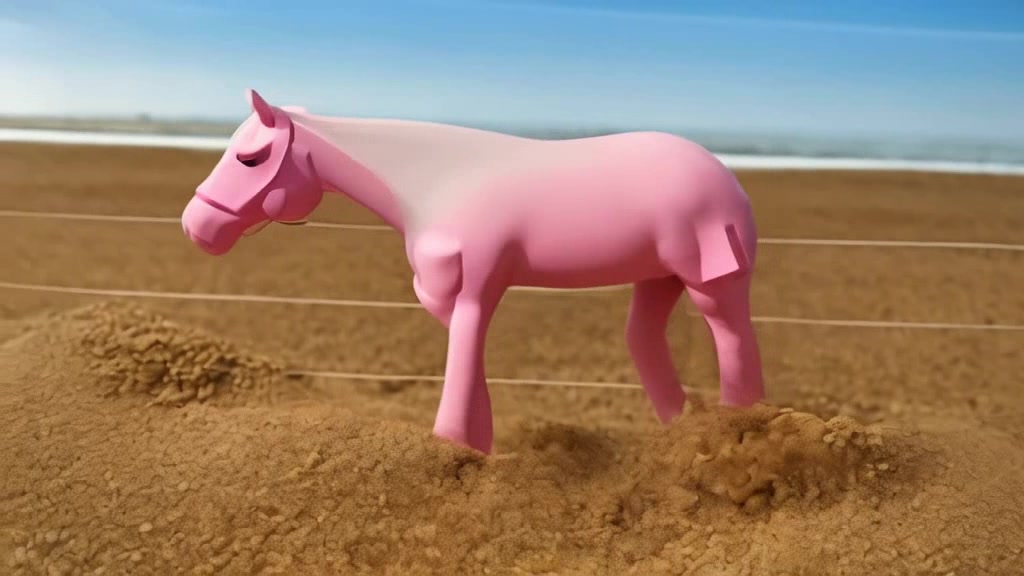}} \raisebox{-.5\height}{\includegraphics[width=0.09\textwidth]{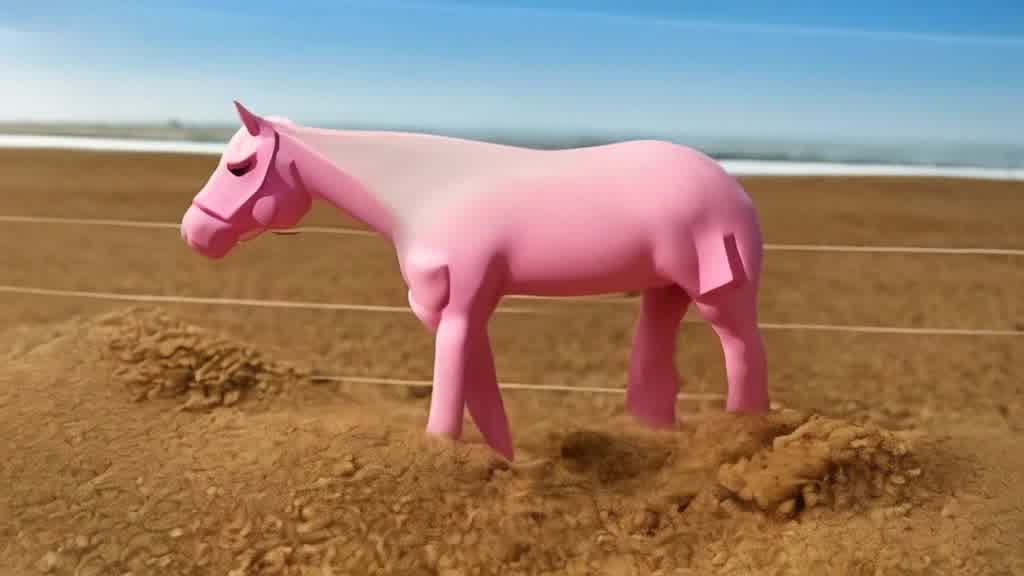}} \raisebox{-.5\height}{\includegraphics[width=0.09\textwidth]{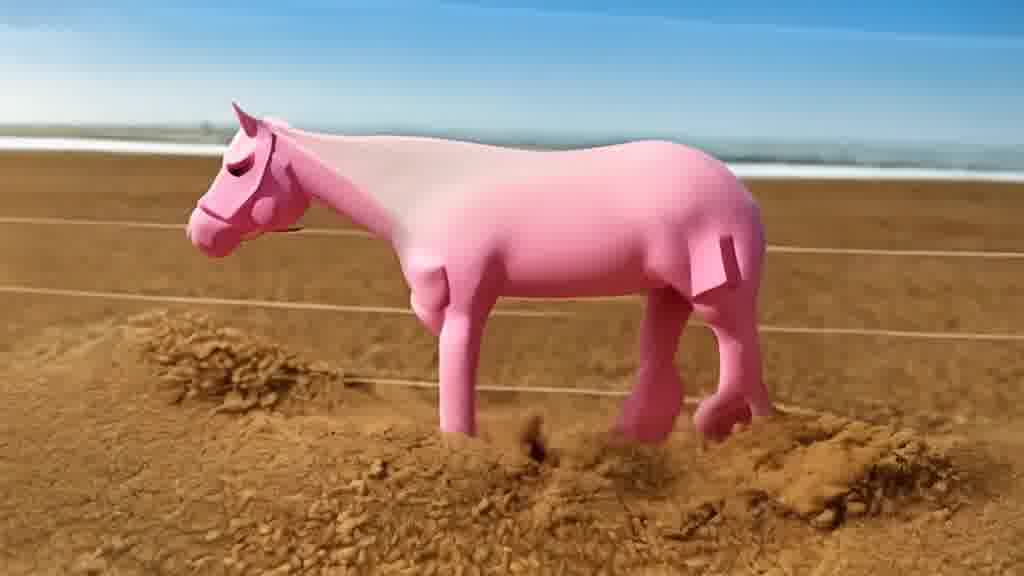}} & \raisebox{-.5\height}{\includegraphics[width=0.09\textwidth]{figures/horse_clip_swap_1024_576/pink_horse.jpg}} & \raisebox{-.5\height}{\includegraphics[width=0.09\textwidth]{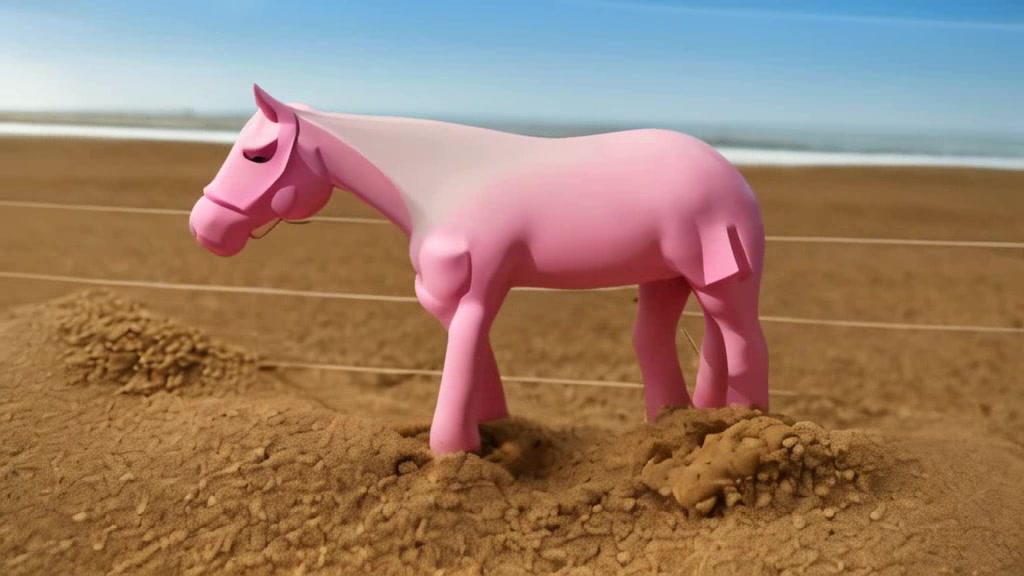}} \raisebox{-.5\height}{\includegraphics[width=0.09\textwidth]{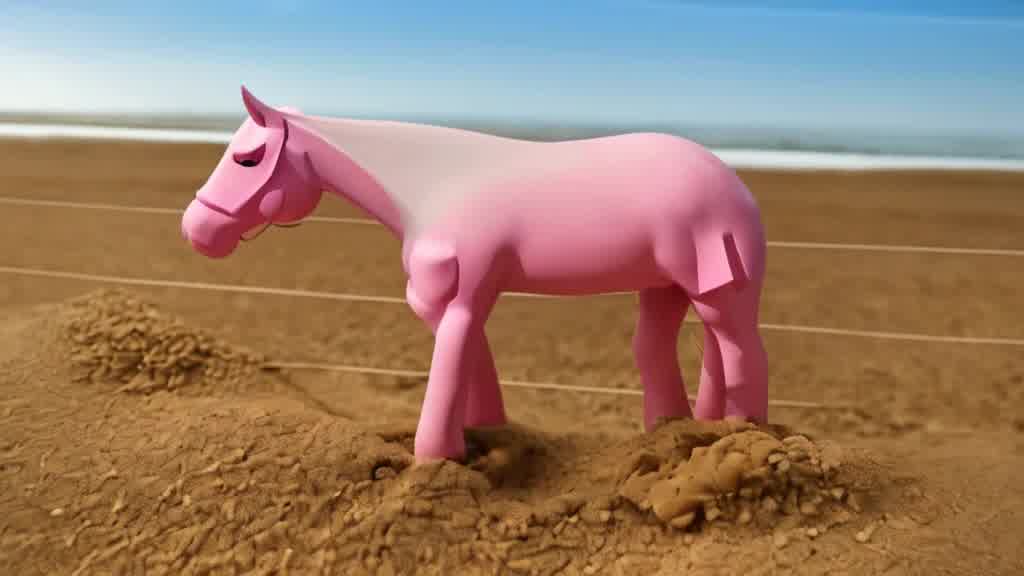}} \raisebox{-.5\height}{\includegraphics[width=0.09\textwidth]{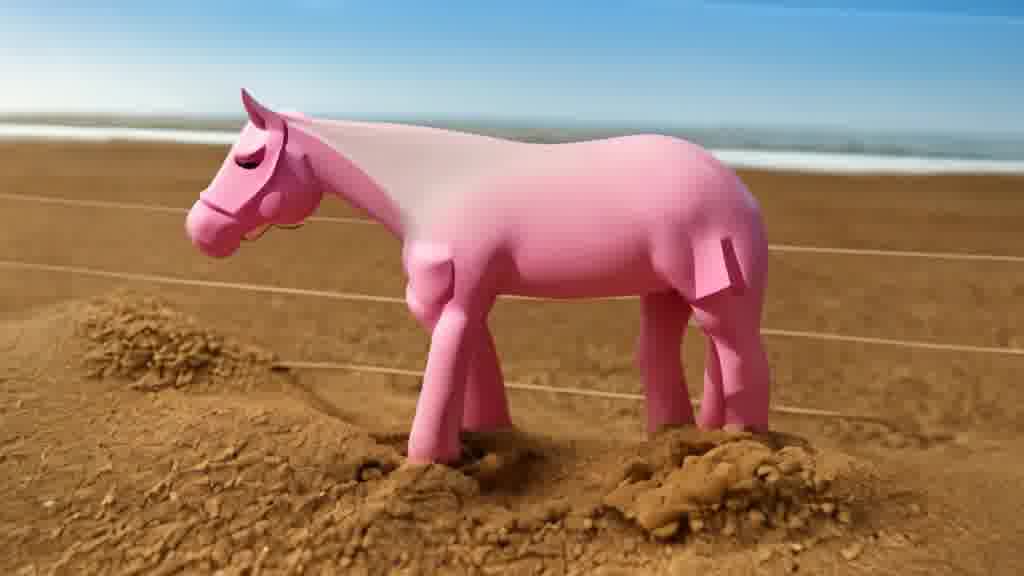}}
	\end{tblr}
	\caption{Observation 2. In image-to-video models, text/image embeddings significantly influence the generated motions. Swapping the CLIP~\cite{clip} image embeddings of a real horse and a toy horse in Stable Video Diffusion~\cite{svd} results in a swap of the motions in the output videos. This suggests that the real horse's embedding encodes a walking motion, while the toy horse's embedding encodes camera motion without object movement.}
	\Description{A 2x2 grid showing videos generated in four different scenarios. Row 1 uses the image latent of the real horse, and the generated results share the appearance of the real horse (white). Row 2 uses the image latent of the toy horse, and the generated results share the appearance of the toy horse (pink). Column 1 uses the CLIP image embedding of the real horse, and the generated results share the motion of the real horse (walking). Column 2 uses the CLIP image embedding of the toy horse, and the generated results share the motion of the toy horse (standing, camera motion).}
	\label{fig:observation}
\end{figure*}

\subsubsection{Baseline}

Stable Video Diffusion (SVD)~\cite{svd} is a video latent diffusion model trained in three stages: 1. A text-to-image model~\cite{ldm} is trained or fine-tuned on (image, text) pairs. 2. The diffusion model is inflated by inserting temporal convolution and attention layers following \citet{align_your_latents} and then trained on (video, text) pairs. 3. The diffusion model is refined on a smaller subset of high-quality videos with exact model adaptations and inputs depending on the task (text-to-video, image-to-video, frame interpolation, multi-view generation). For image-to-video generation, the task is to produce a video given its starting frame. The starting frame is supplied to the model in two places: as a CLIP~\cite{clip} image embedding via cross-attention (replacing the CLIP text embedding from the text-to-video pre-training) and as a latent repeated across frames and concatenated channel-wise to the video input. Additionally, the model is micro-conditioned on the frame rate, motion amount, and strength of the noise augmentation (applied to first frame latent).

\subsection{Motivation} \label{sec:method_motivation}

Transferring the motion of a reference video to a given target poses two key challenges, which our design solves quite naturally.

\subsubsection{Challenge 1: Appearance Leakage} 

Fine-tuning a \underline{text}-to-video model on a single reference video to learn its motion risks overfitting to its appearance, hindering the generation of correct target appearances during inference. We demonstrate that using a frozen \underline{image}-to-video model can preserve the target appearance without any of the special mechanisms from the literature.

By design, image-to-video models generate videos from a starting frame, naturally preserving the input appearance. We observe that image-to-video models primarily derive the appearance from the image (latent) input, even with an additional text input, as shown in Fig.~\ref{fig:i2v_ignores_text}. This is likely because the model can directly copy (latent) pixels from the first frame instead of hallucinating them from the sparse text input. This strong reliance on the image input reduces the chance of the reference video's appearance leaking through. To further minimize the risk of appearance leakage, we keep the model's weights frozen, so they cannot possibly store the reference video appearance. This also helps retain the rich video understanding and generalization capabilities of the pre-trained model.

\subsubsection{Challenge 2: Handling Object Misalignment.} \label{sec:alignment}

Our goal is to generate videos where subjects perform the same semantic actions, even if they are in different spatial locations or orientations. Handling misaligned objects is especially important when using image-to-video models because the subject's position is determined by the input image, which typically does not match the position in the motion reference video. 

As discussed in Section~\ref{sec:related_work-inversion-then-generation}, existing methods using the inversion-then-generation framework inject features from the motion reference video into the generated video, making it closely follow the reference structure. Arguably, these methods do not copy the motion at its origin but rather the \emph{per-frame structure} that results from a motion (e.g., rough object positions). For the general, unaligned case, these features would first need to be aligned spatially to avoid injecting the structure in the wrong place. This alignment is challenging since the final positions in the generated video are unknown during the diffusion process as they depend on the motion.

\begin{figure*}
	\includegraphics[trim = 21mm 76mm 39mm 78mm, width=\textwidth, clip]{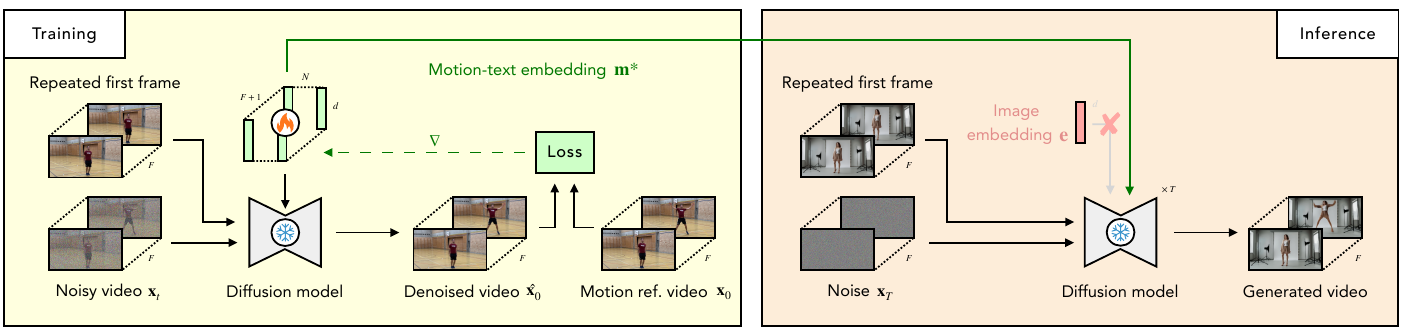}
	\caption{Method overview. The baseline image-to-video diffusion model, Stable Video Diffusion~\cite{svd} in our case, inputs the first frame in two places: as image (latent) concatenated with the noisy video and as image embedding (some other image-to-video diffusion models may input text embeddings here instead). We propose to replace the image embedding $\mb{e}$ (shown in red in the inference block) with a learned motion-text embedding $\mb{m}^*$ (green). The motion-text embedding is optimized directly with a regular diffusion model loss on one given motion reference video $\mb{x}_0$ while keeping the diffusion model frozen. For best results, the motion-text embedding is inflated prior to optimization to $(F+1) \times N$ tokens, where $F$ is the number of frames and $N$ is a hyperparameter, while keeping the embedding dimension $d$ the same to stay compatible with the pre-trained diffusion model. Note that the diffusion process operates in latent space in practice, and other conditionings and model parameterizations~\cite{edm} are omitted for clarity.}
	\Description{Method overview with a yellow box on the left for the training and an orange box on the right for inference. In the training box, the repeated first frame of a video of a person doing jumping jacks, a noisy version of the motion reference video, and the green motion-text embedding are passed as input to a diffusion model, which produces a denoised version of the video. This denoised video is input to the loss along with the motion reference video. There is an arrow from the loss to the motion-text embedding to show that the embedding is updated according to the gradient of the loss. In the inference box, the repeated first frame of a different target subject, a purely noise video, and the green motion-text embedding are passed as input to a diffusion model. The diffusion model runs for $T$ steps and then generates a video of the target subject doing jumping jacks.}
	\label{fig:architecture}
\end{figure*}

We forgo the alignment problem by representing motions with text or image embedding tokens that do not have a spatial dimension in the first place. Our novel motion representation was motivated by the observation shown in Fig.~\ref{fig:observation}. While SVD generated walking motions for an image of a real horse, it generated no object but mostly camera motion for an image of a pink toy horse, perhaps because the model learned that toys do not move.\footnote{Image was generated using the method by \citet{plug_and_play}.} Recall that SVD has the first frame as input in two places: as image latent and as CLIP~\cite{clip} image embedding. When using the image latent of the real horse but the CLIP embedding of the toy horse, the horse in the generated video does not move. Inversely, the toy horse starts walking when using the CLIP embedding of the real horse, implying that the CLIP embedding affects the motion. We believe that these embeddings are \emph{not just affecting} the motion but are actually the main \emph{origin} of the motion. 

Our intuition for why the text/image embeddings determine the motion (which may be surprising at first) is as follows: Videos can be divided into appearance and motion. Appearance is tied to the spatial arrangement of pixels, making it easier to extract it from spatial inputs like image latents. Motion depends on how pixels change over time, requiring a more global, semantic understanding. Thus, it is more effective to modify motion using image embeddings, which contain more semantic information, have no spatial dimension, and are injected in multiple places of the model. Furthermore, SVD was initially trained as a text-to-video model, with CLIP text embeddings describing motions like ``standing,'' ``walking,'' or ``running,'' incentivizing the model to control motion through cross-attention inputs to effectively denoise training videos.

\subsection{Motion-Textual Inversion} \label{sec:method_inversion}

While using embeddings from different images can alter the generated motion, it does not transfer the motion robustly. Moreover, selecting a specific frame to define a desired motion is difficult since motion is rarely captured by a single frame. To address this, we propose optimizing the embedding based on a given motion reference video, which bears some resemblance to textual inversion~\cite{textual_inversion}. In analogy to textual inversion, we name our method \textit{motion-textual inversion}.\footnote{In our implementation, it is actually an image embedding, but we refer to it as ``motion-\emph{textual} inversion'' since SVD's image and text embeddings share the same CLIP space, and other I2V methods use text embeddings instead. Also, it feels more intuitive to represent motions as text rather than an image.} Note, however, that our method has a completely different goal: using embeddings to encode video motion rather than image appearance.

Fig.~\ref{fig:architecture} shows a high-level overview of our method. Given a single motion reference video $\mb{x}_0$ containing $F$ frames, we optimize the motion-text embedding $\mb{m}$ directly by minimizing the diffusion model loss from Equation~\ref{eq:standard-diffusion}, keeping the diffusion model frozen:
\begin{equation} \label{eq:motion-textual-inversion-loss}
\begin{split}
    \mb{m}^* = \underset{\mb{m}}{\text{arg\,min}} \, & \mathbb{E}_{(\mb{x}_0, \mb{c}) \sim p_\text{data}(\mb{x}_0, \mb{c}),(\sigma, \mb{n}) \sim p(\sigma, \mb{n})} \\ 
    & [\lambda_\sigma ||D_{\bm{\theta}}(\mb{x}_0 + \mb{n}; \sigma, \mb{m}, \mb{c}) - \mb{x}_0||^2_2],
\end{split}
\end{equation}
where $\mb{c}$ encompasses all remaining conditionings of SVD (e.g., first frame latent, time/noise step, and micro-conditionings). All other symbols are defined in Equations~\ref{eq:standard-diffusion} and \ref{eq:edm-parametrization}.

The optimized motion-text embedding can be visualized with an unconditional appearance as seen in Fig.~\ref{fig:teaser} and further described in Section~\ref{sec:embedding-analysis}.

\subsection{Motion-Text Embedding and Cross-Attention Inflation} \label{sec:inflation}

Cross-attention allows the model to dynamically attend to different tokens ($\sim$ words in text-to-image and text-to-video) depending on the current features or context. It is computed as follows:
\begin{equation}
\begin{gathered}
    \text{Attention}(Q, K, V) = M V = \text{softmax}(\frac{QK^T}{\sqrt{d_a}}) V, \\ 
    Q = \varphi_i(\mb{z}_t) W_{Q,i}, \, K = \mb{m} W_{K,i}, \, V = \mb{m} W_{V,i},
\end{gathered}
\end{equation}
where $Q$, $K$, $V$ are the queries, keys, and values respectively; $M$ is the attention map; $d_a$ is the dimension used in the attention operation; $\varphi_i(\mb{z}_t)$ is an intermediate representation of the level $i$ features with $C_i$ channels; $\mb{m}$ is the motion-text embedding (or text/image embedding $\mb{e}$ in case of baseline SVD) with embedding dimension $d$; and $W_{Q,i} \in \mathbb{R}^{C_i \times d_a}$, $W_{K,i} \in \mathbb{R}^{d \times d_a}$, and $W_{V,i} \in \mathbb{R}^{d \times d_a}$ are learned weight matrices for queries, keys, and values respectively.

\begin{figure*}
	\centering
	\subfloat[a][Default SVD: Since the image embedding $\mb{e}$ has only one token, every spatial and temporal location attends $100\%$ to that single token. The cross-attention operation thus degenerates to a simple addition of a single broadcasted vector to the feature tensor.\label{fig:inflation-default}]{\includegraphics[trim = 187mm 127mm 30mm 33mm, width=0.45\textwidth, clip]{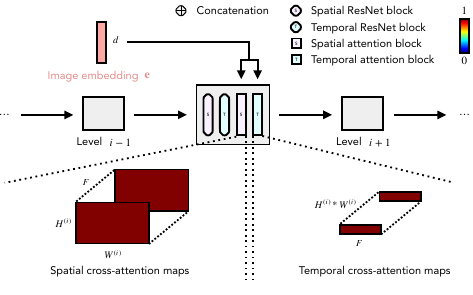}} \hspace{5mm}
	\subfloat[b][Inflated SVD (Ours): By introducing more tokens in the token dimension ($N$), every spatial and temporal location can dynamically attend to different tokens, e.g., different tokens for the foreground vs. background. For the spatial cross-attention, we use different tokens per frame, resulting in different keys and values per frame. This enables a higher temporal granularity of the motion.\label{fig:inflation-ours}]{\includegraphics[trim = 187mm 41mm 30mm 114mm, width=0.45\textwidth, clip]{figures/inflation.pdf}}
	\caption{High-level visualization of our motion-text embedding and cross-attention inflation. The SVD~\cite{svd} UNet is composed of several levels of blocks, shown in gray, that have similar structure. We visualize the sub-blocks of level $i$ and their cross-attention maps in more detail. Our inflated motion-text embedding produces more meaningful cross-attention maps, resulting in improved motion learning. The cross-attention maps were extracted from the example of the woman doing jumping jacks in Fig.~\ref{fig:architecture}.}
	\Description{Two visualizations of the model architecture and cross-attention maps, where (a) is for the default SVD~\cite{svd} and (b) is our proposed inflated version. Each visualization shows three gray blocks of the diffusion model. For the center block (level $i$), the sub-blocks are shown: spatial ResNet block, temporal ResNet block, spatial attention block, and temporal attention block. Visualization (a) shows the image embedding $\mb{e}$ with a single token on top, which is passed as input to the spatial and temporal attention blocks. Below the blocks, two spatial and two temporal cross-attention maps are visualized. All of them are the same color, i.e., all have the value 1. Visualization (b) shows our motion-text embedding $\mb{m}$ with $F+1 \times N$ tokens which is split into two parts. The part with $F \times N$ tokens is passed as input to the spatial attention block whereas the other part with $N$ tokens is passed as input to the temporal attention block. There are $N$ times as many cross-attention maps now compared to (a), i.e., $2 \times 2$ instead of $2$ maps of each type, and their values differ depending on the spatial or temporal location. For the spatial cross-attention maps referring to the first frame, one of the maps highlights the background whereas the other one highlights the foreground. The maps referring to the other frame have quite different values/colors to show that each frame has a different set of tokens. For the temporal cross-attention maps, the values/colors differ slightly for each temporal location.}
	\label{fig:inflation}
\end{figure*}

SVD's image embedding only has one token. This leads to a degenerate cross-attention where all entries of the attention map $M$ are $1$, as shown in Fig.~\ref{fig:inflation-default}. The model thus attends $100\%$ to that single token and applies its value to all spatial and temporal locations. 

\subsubsection{Multiple Tokens}

To enable richer motion control, we replace the single token with $N$ tokens, recovering the scenario from the text-to-image or text-to-video pre-training. This allows the model to dynamically attend to different tokens depending on the features, e.g., using different values for the background and foreground as seen in the spatial cross-attention maps in Fig.~\ref{fig:inflation-ours}.

\subsubsection{Different Tokens per Frame} \label{sec:inflation-different-tokens-per-frame}

For spatial cross-attention, SVD broadcasts the image embedding \textit{across all frames}. Instead, we use a different set of tokens per frame, i.e., $F \times N$ tokens, to obtain a higher temporal motion granularity.\footnote{Note that we always use the same $F$ frames of the motion reference video when optimizing the motion-text embedding.} This yields distinct keys and values for each frame: different keys enable attention to different spatial regions over time (e.g., arm vs. leg), while different values allow frame-specific feature modifications (e.g., shifting pixels in different directions). This is visualized in Fig.~\ref{fig:inflation-ours}, where the spatial cross-attention maps differ greatly between frames because they use different tokens. 

For temporal cross-attention, SVD broadcasts the image embedding \textit{across all spatial locations}. Inflating this analogously to the spatial case would require learning distinct tokens per spatial location, which is nontrivial due to resolution- and level-dependent spatial dimensions and may cause alignment issues (see Section~\ref{sec:alignment}). Furthermore, temporal cross-attention impacted motion less than spatial cross-attention empirically. We thus keep $N$ tokens for the temporal motion-text embedding but learn them independently from the $F \times N$ tokens of the spatial motion-text embedding, yielding a total of $(F+1) \times N$ tokens per reference video. See Section~\ref{sec:inflation-details} for an intuitive analogy and detailed tensor shapes.

\section{Experiments}

\subsection{Implementation Details}

Our method builds on the $14$-frame version of Stable Video Diffusion~(SVD)~\cite{svd, diffusers} but can be applied to other image-to-video models with a text/image embedding input. Per default, we use $N=5$ different tokens for each of the $F=14$ frames, so a total of $(14+1) \times 5 = 75$ tokens for the motion-text embedding. We further use the Adam optimizer~\cite{adam} and SVD's default guidance scale~\cite{cfg} (except for motion visualization). For our qualitative results, we use internal data sets and target images generated with SDXL~\cite{sdxl}. See Section~\ref{sec:implementation-details} for further details.

\subsection{Compared Methods}

As baseline, we use SVD~\cite{svd} without adaptations. Since it lacks motion conditioning, it rarely follows the correct motion but serves as a reference for typical SVD output quality and dynamics.
Our method is the first to tackle general motion transfer in the image-to-video setting. As no direct competitors exist, we apply the most closely related approaches from literature to our task and show issues inherent to the whole class of methodology. Specifically, we compare to VideoComposer~\cite{videocomposer}, an image-to-video method with an explicit, dense motion representation (motion vectors); the image-to-video setting of MotionClone~\cite{motionclone} which has an implicit motion representation (sparse temporal attention weights); and MotionDirector~\cite{motiondirector}, a text-to-video method with an implicit motion representation (learned model weights). 
We only compare to general methods that place no constraints on motion types and target images. Domain-specific methods rely on strong assumptions and typically fail when these are not met. For example, a face reenactment method cannot control transfer the motion of a horse to a boat. As domain-specific methods address a different task, a fair comparison is not possible. See Section~\ref{sec:additional-information-compared-methods} for further details.

\begin{table*}[htbp]
	\centering
	\caption{Quantitative evaluation. We compare our method to Stable Video Diffusion~\cite{svd} (baseline, no motion input), VideoComposer~\cite{videocomposer}, MotionClone~\cite{motionclone}, and MotionDirector~\cite{motiondirector}. The best performing method per column is marked in bold.}
	\label{table:quant_eval}
	\small{
		\begin{tabular}{llccclcccclc}
			\toprule
			\multirow{2}{*}{Method} & & \multicolumn{3}{c}{Image Appearance Preservation} & & \multicolumn{4}{c}{Video Motion Fidelity} & & \multicolumn{1}{c}{Overall} \\
			\cmidrule{3-5} \cmidrule{7-10} \cmidrule{12-12}
			& & {CLIP-Avg $\uparrow$} & {CLIP-1st $\uparrow$} & {User rank $\downarrow$} & & {Acc-Top-1 $\uparrow$} & {Acc-Top-5 $\uparrow$} & {Cos-Sim $\uparrow$} & {User rank $\downarrow$} & & {User rank $\downarrow$} \\
			\midrule
			Stable Video Diffusion & & \textbf{0.843} & 0.850 & \textbf{1.296} & & 3\% & 5\% & 0.370 & 4.211 & & 2.822 \\
			VideoComposer & & 0.719 & 0.857 & 3.785 & & 44\% & 62\% & 0.497 & 3.030 & & 3.552 \\
			MotionClone & & 0.637 & \textbf{0.885} & 4.585 & &37\% & 62\% & 0.523 & 3.137 & & 4.200 \\
			MotionDirector & & 0.750 & 0.763 & 3.522 & & 31\% & 58\% & 0.523 & 2.900 & & 3.059 \\
			\midrule
			Ours & & 0.779 & 0.884 & 1.811 & & \textbf{54\%} & \textbf{76\%} & \textbf{0.696} & \textbf{1.722} & & \textbf{1.367} \\
			\bottomrule
			\\
		\end{tabular}
	}
\end{table*}

\begin{figure*}
	\centering
	\begin{tblr}{
			cell{1}{2} = {c=3}{c},
			cell{1}{5} = {c=3}{c},
			cell{1}{8} = {c=3}{c},
			columns = {c},
			column{3} = {rightsep=1pt},
			column{4} = {leftsep=1pt},
			column{6} = {rightsep=1pt},
			column{7} = {leftsep=1pt},
			column{9} = {rightsep=1pt},
			column{10} = {leftsep=1pt},
			vline{3} = {3-7}{dashed},
			vline{5} = {1-7}{},
			vline{6} = {3-7}{dashed},
			vline{8} = {1-7}{},
			vline{9} = {3-7}{dashed},
			hline{3} = {1-10}{},
			hline{4} = {1-10}{},
			hline{5} = {1-10}{},
			hline{6} = {1-10}{},
			hline{7} = {1-10}{},
		}
		& Jumping jacks & & & Head nodding & & & Camera flying forwards & \\
		{Ref.} & \raisebox{-.5\height}{\includegraphics[width=0.08\textwidth]{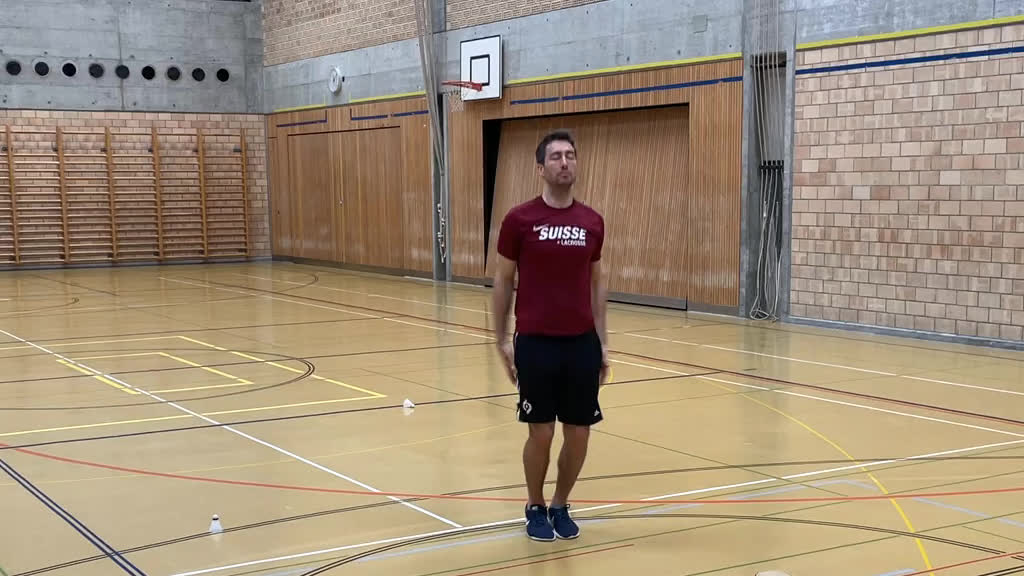}} & \raisebox{-.5\height}{\includegraphics[width=0.08\textwidth]{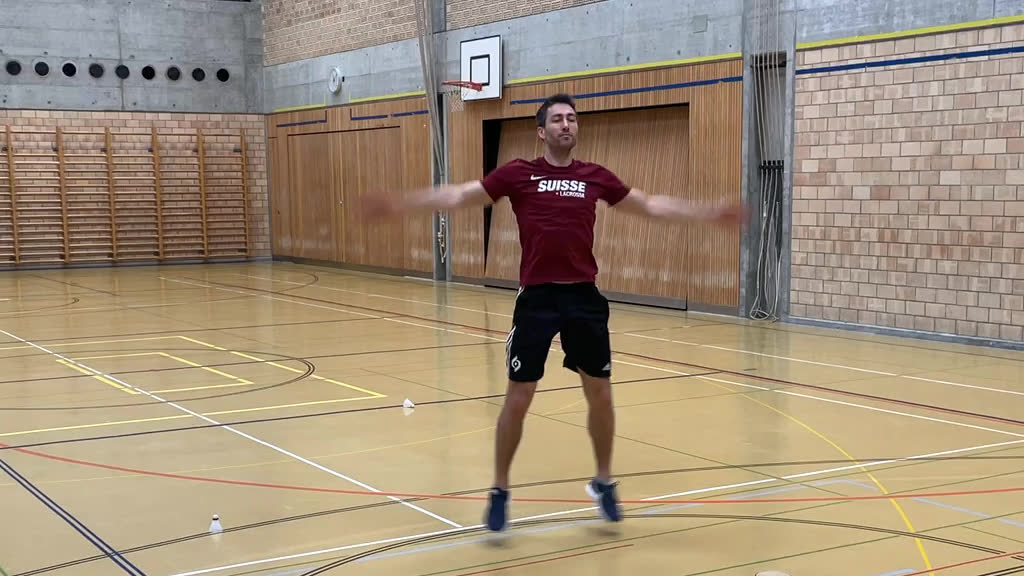}} & \raisebox{-.5\height}{\includegraphics[width=0.08\textwidth]{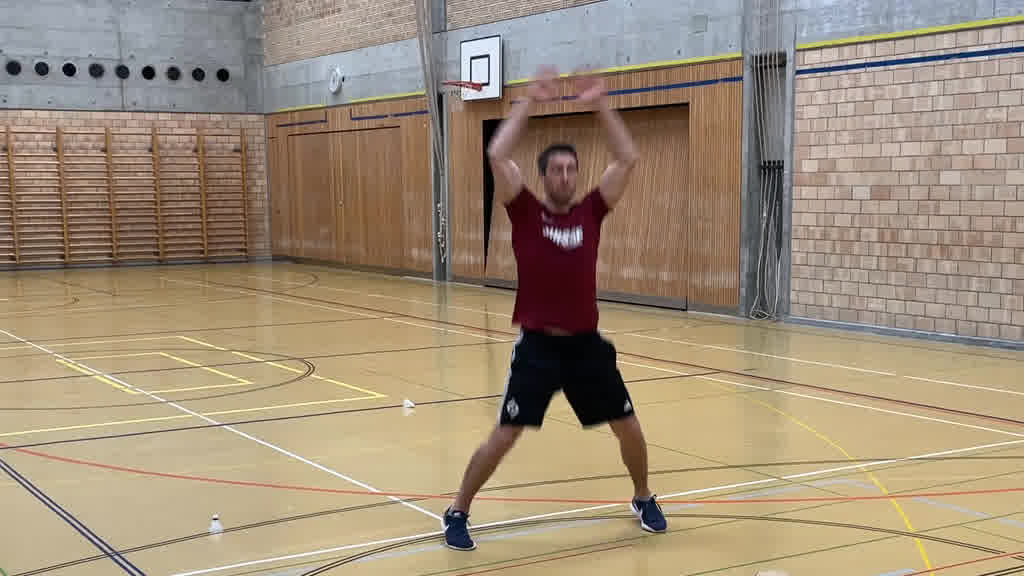}} & \raisebox{-.5\height}{\includegraphics[width=0.08\textwidth]{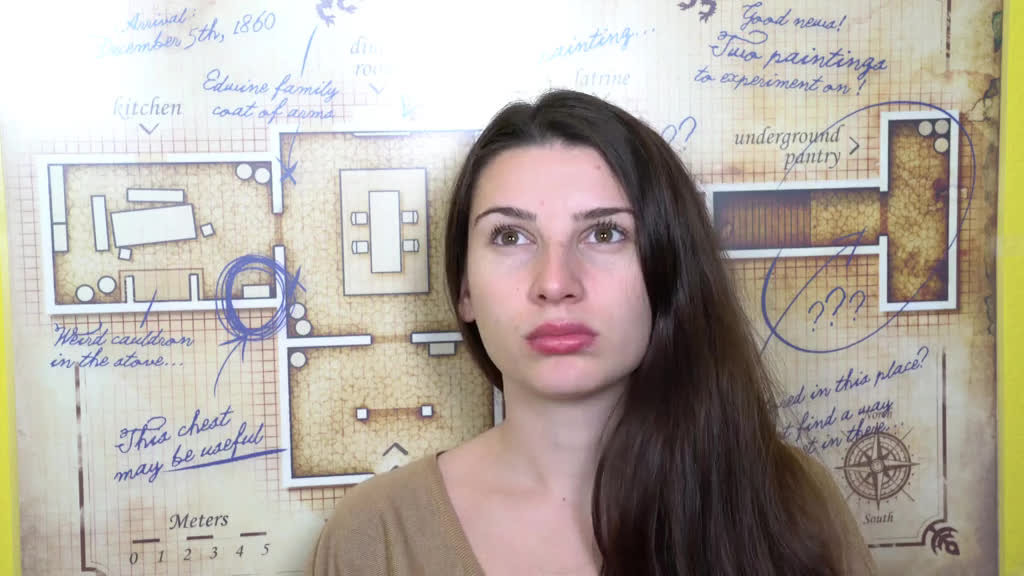}} & \raisebox{-.5\height}{\includegraphics[width=0.08\textwidth]{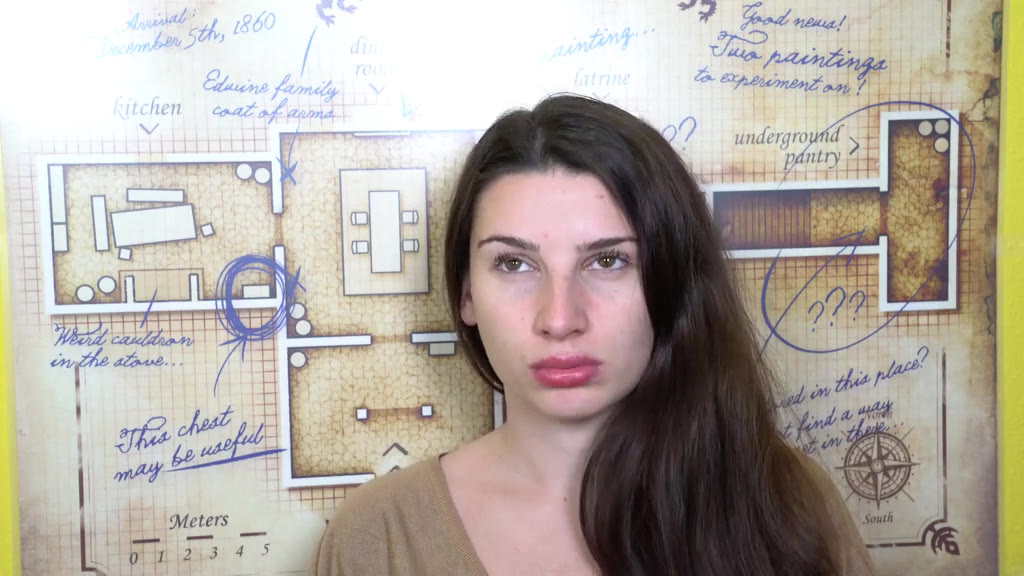}} & \raisebox{-.5\height}{\includegraphics[width=0.08\textwidth]{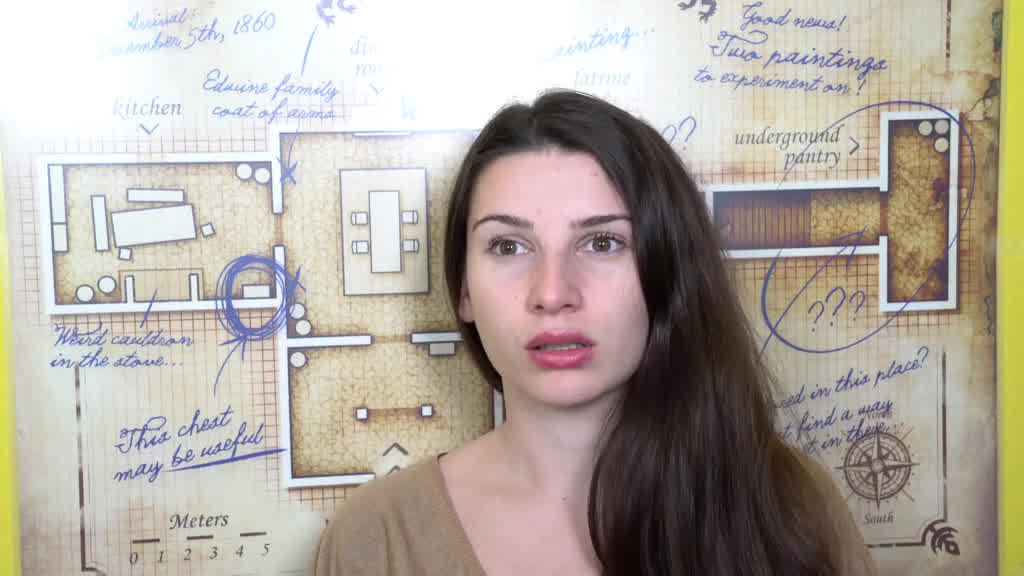}} & \raisebox{-.5\height}{\includegraphics[width=0.08\textwidth]{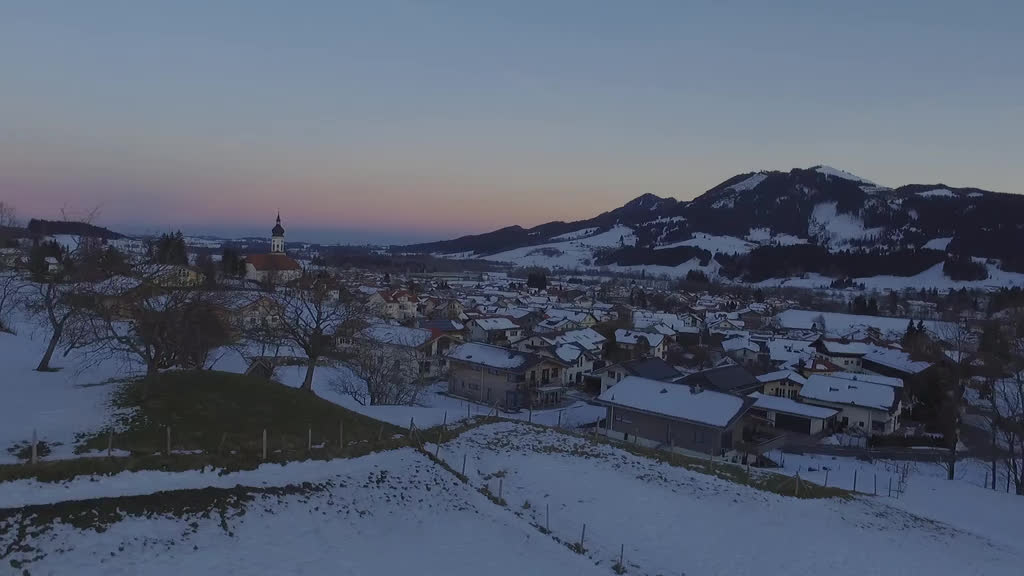}} & \raisebox{-.5\height}{\includegraphics[width=0.08\textwidth]{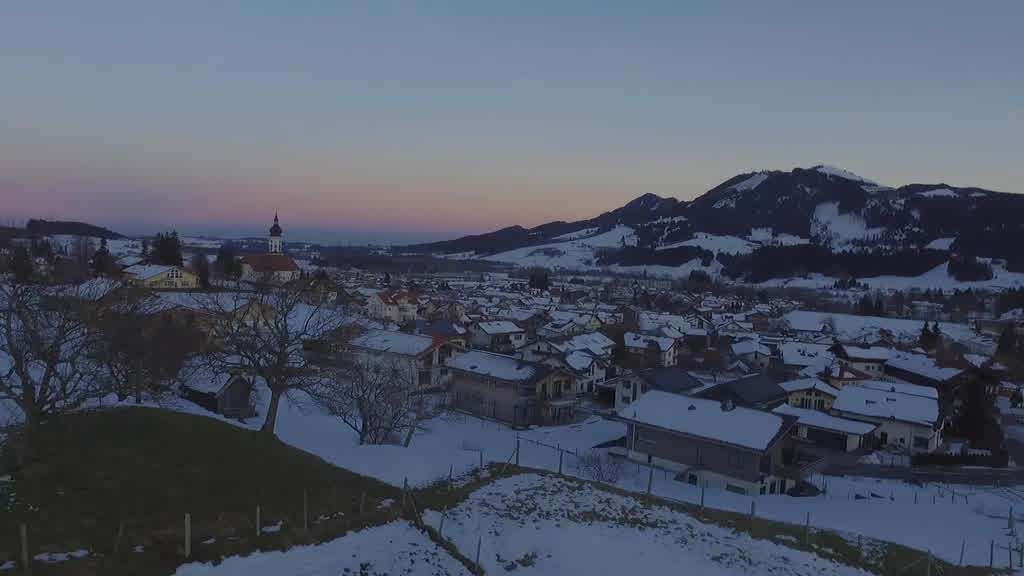}} & \raisebox{-.5\height}{\includegraphics[width=0.08\textwidth]{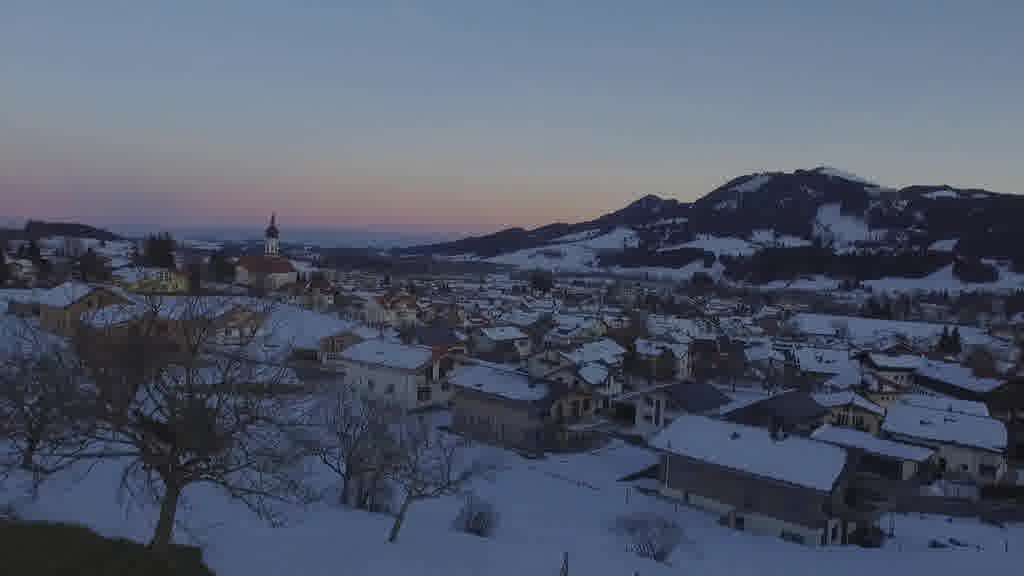}} \\
		{SVD}  & \raisebox{-.5\height}{\includegraphics[width=0.08\textwidth]{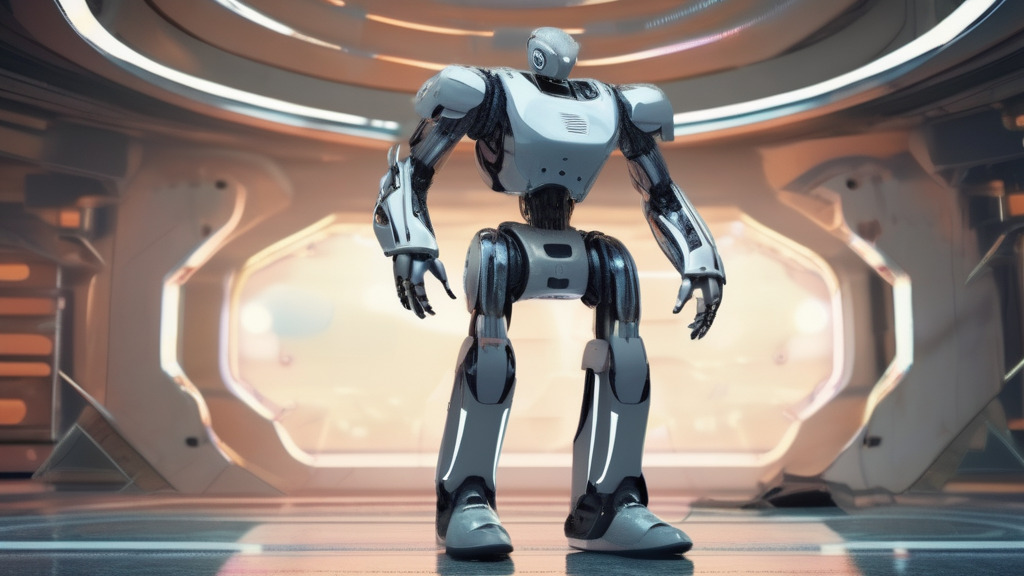}} & \raisebox{-.5\height}{\includegraphics[width=0.08\textwidth]{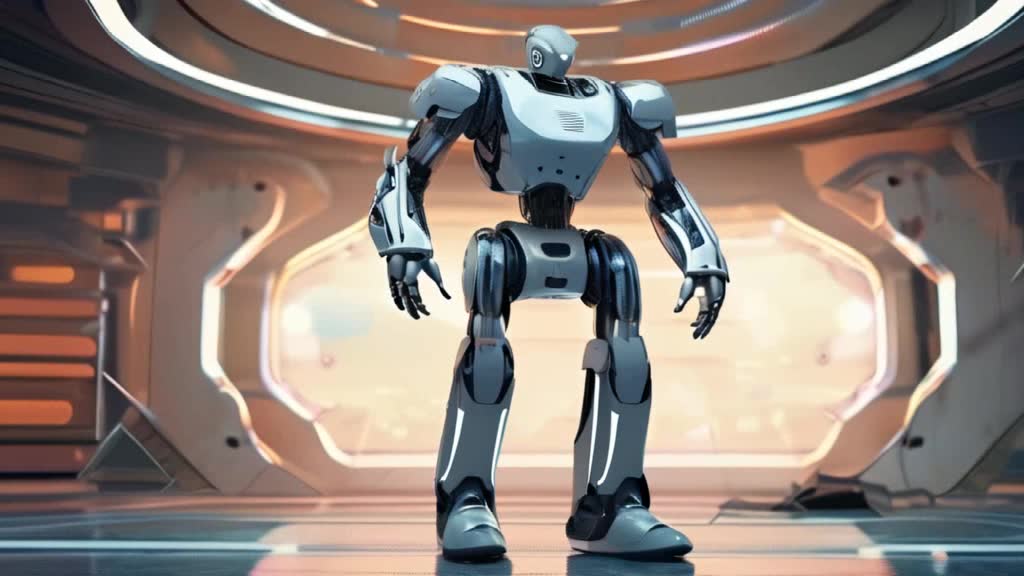}} & \raisebox{-.5\height}{\includegraphics[width=0.08\textwidth]{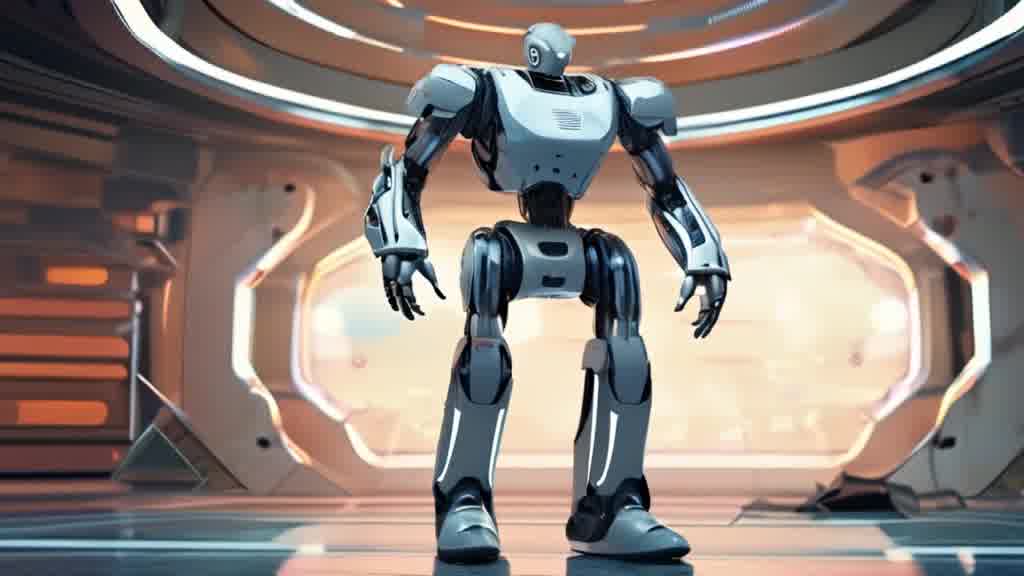}} & \raisebox{-.5\height}{\includegraphics[width=0.08\textwidth]{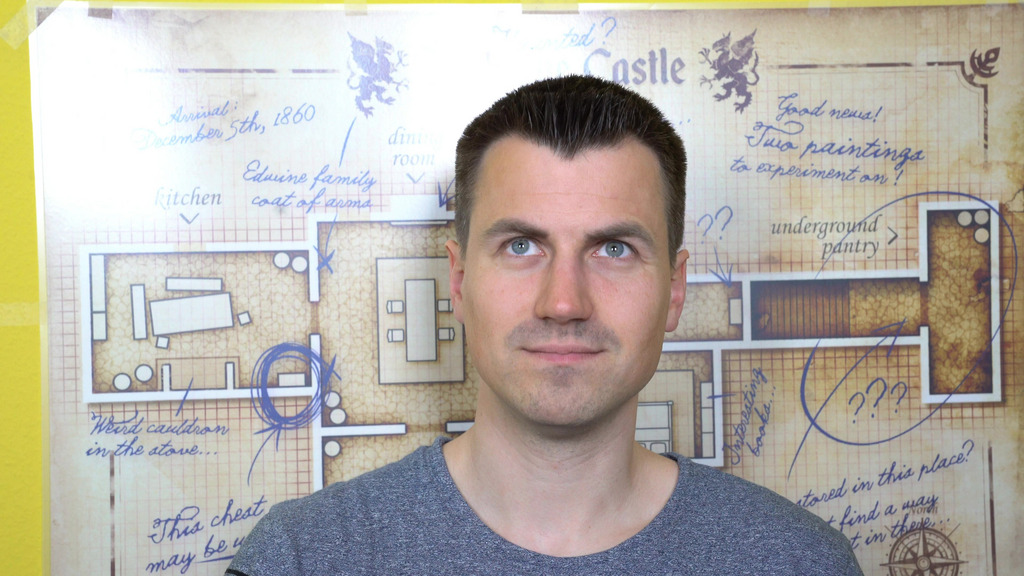}} & \raisebox{-.5\height}{\includegraphics[width=0.08\textwidth]{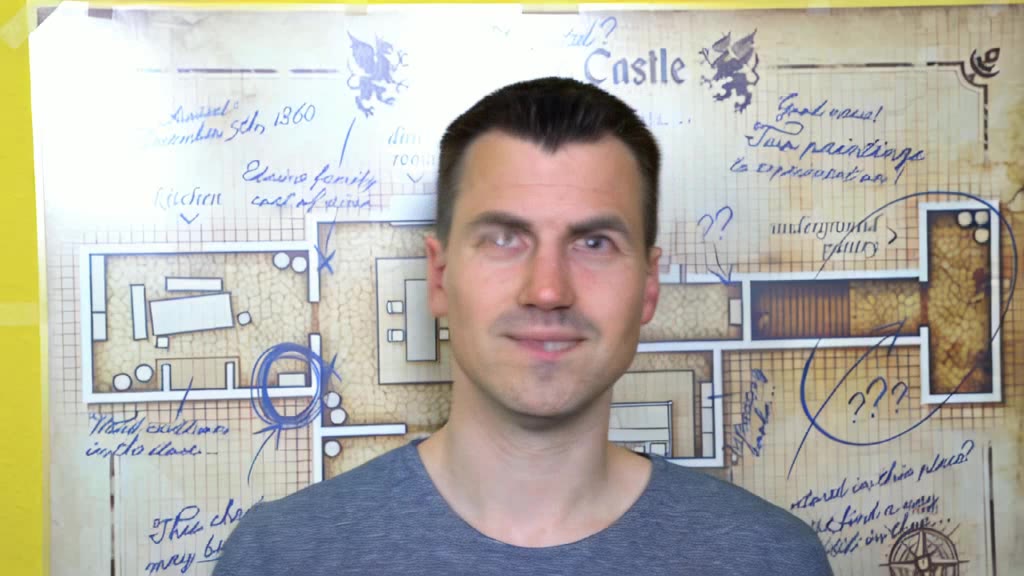}} & \raisebox{-.5\height}{\includegraphics[width=0.08\textwidth]{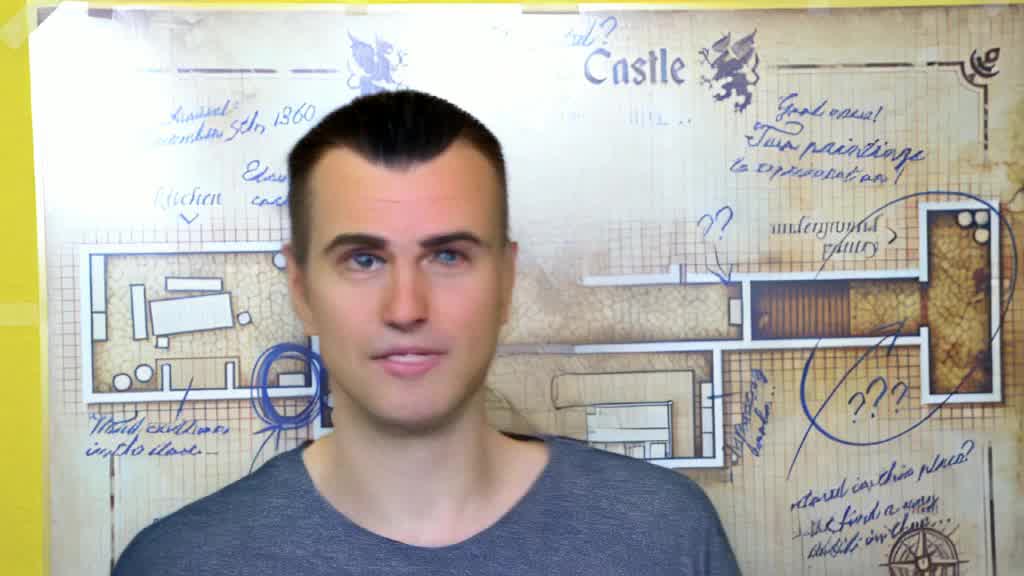}} & \raisebox{-.5\height}{\includegraphics[width=0.08\textwidth]{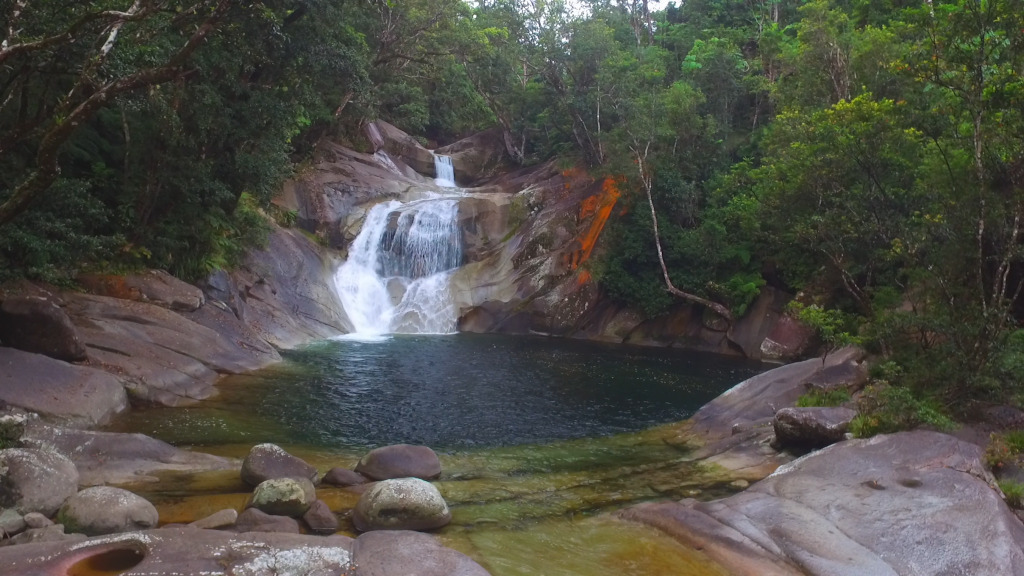}} & \raisebox{-.5\height}{\includegraphics[width=0.08\textwidth]{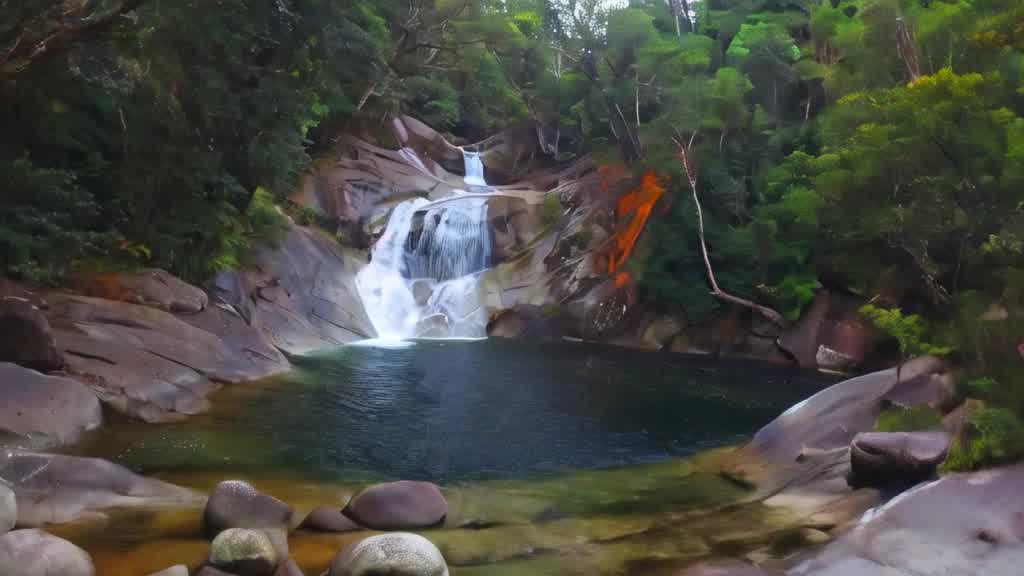}} & \raisebox{-.5\height}{\includegraphics[width=0.08\textwidth]{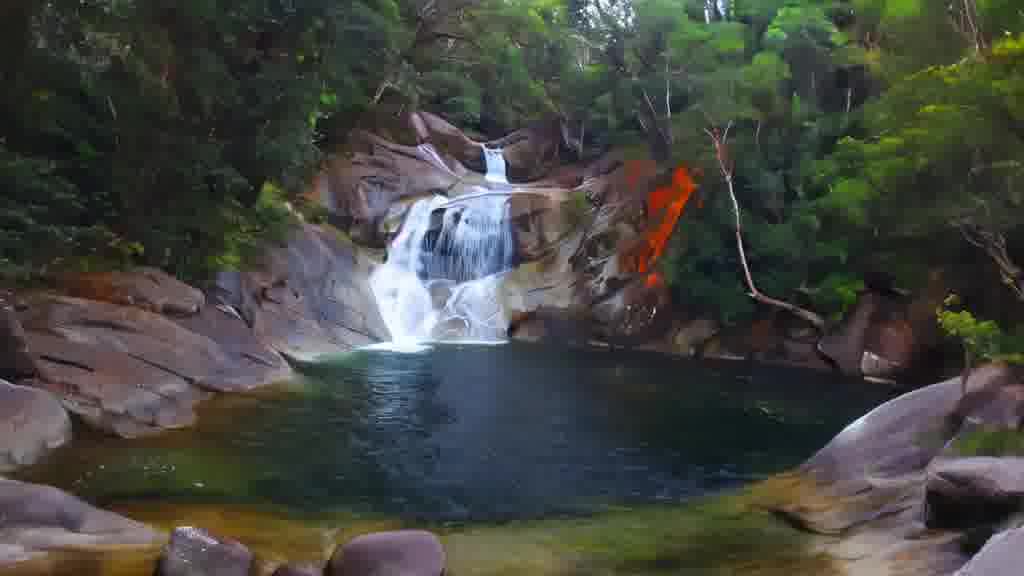}} \\
		{VC} & \raisebox{-.5\height}{\includegraphics[width=0.08\textwidth]{figures/qual_eval/fullbody_jumping_jacks/first_frame.jpg}} & \raisebox{-.5\height}{\includegraphics[width=0.045\textwidth]{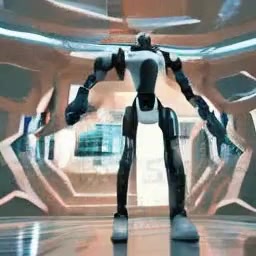}} & \raisebox{-.5\height}{\includegraphics[width=0.045\textwidth]{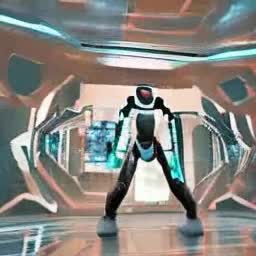}} & \raisebox{-.5\height}{\includegraphics[width=0.08\textwidth]{figures/qual_eval/face_nodding/first_frame.jpg}} & \raisebox{-.5\height}{\includegraphics[width=0.045\textwidth]{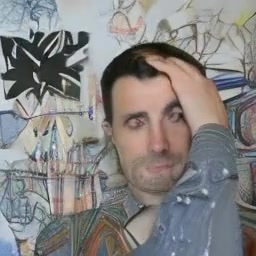}} & \raisebox{-.5\height}{\includegraphics[width=0.045\textwidth]{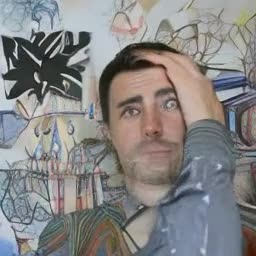}} & \raisebox{-.5\height}{\includegraphics[width=0.08\textwidth]{figures/qual_eval/camera_medium/first_frame.jpg}} & \raisebox{-.5\height}{\includegraphics[width=0.045\textwidth]{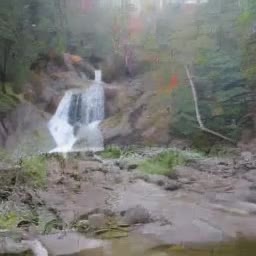}} & \raisebox{-.5\height}{\includegraphics[width=0.045\textwidth]{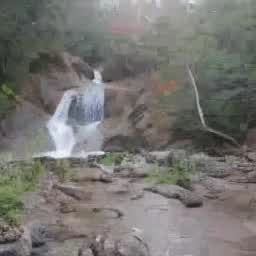}} \\
		{MC} & \raisebox{-.5\height}{\includegraphics[width=0.08\textwidth]{figures/qual_eval/fullbody_jumping_jacks/first_frame.jpg}} & \raisebox{-.5\height}{\includegraphics[width=0.045\textwidth]{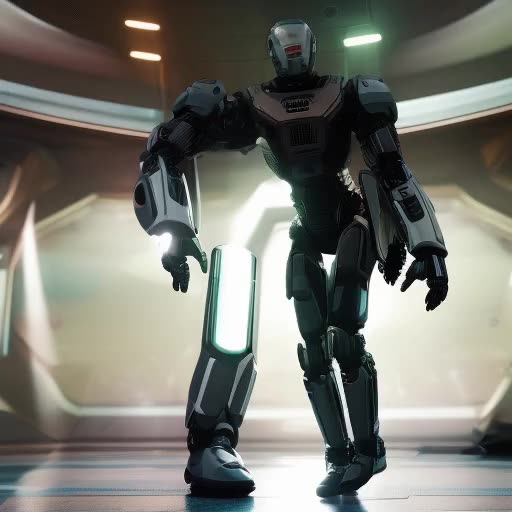}} & \raisebox{-.5\height}{\includegraphics[width=0.045\textwidth]{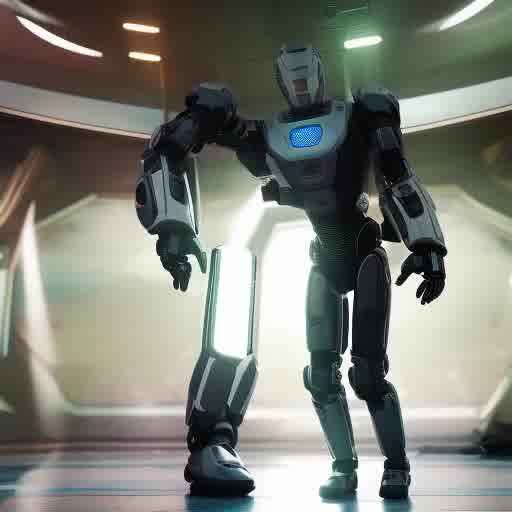}} & \raisebox{-.5\height}{\includegraphics[width=0.08\textwidth]{figures/qual_eval/face_nodding/first_frame.jpg}} & \raisebox{-.5\height}{\includegraphics[width=0.045\textwidth]{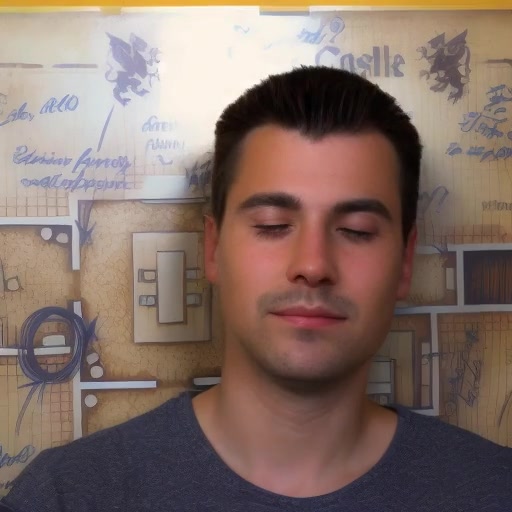}} & \raisebox{-.5\height}{\includegraphics[width=0.045\textwidth]{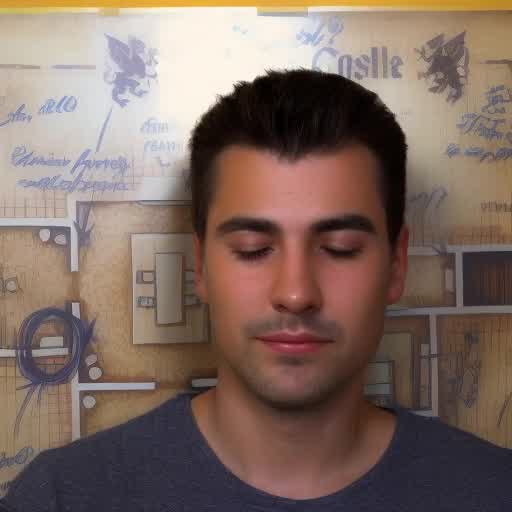}} & \raisebox{-.5\height}{\includegraphics[width=0.08\textwidth]{figures/qual_eval/camera_medium/first_frame.jpg}} & \raisebox{-.5\height}{\includegraphics[width=0.045\textwidth]{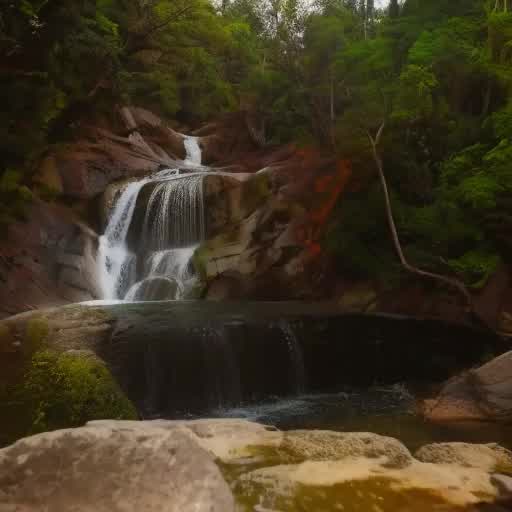}} & \raisebox{-.5\height}{\includegraphics[width=0.045\textwidth]{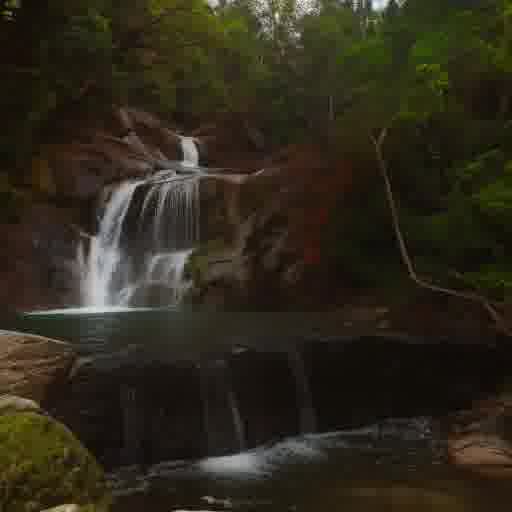}} \\
		{MD} & \raisebox{-.5\height}{\includegraphics[width=0.08\textwidth]{figures/qual_eval/fullbody_jumping_jacks/first_frame.jpg}} & \raisebox{-.5\height}{\includegraphics[width=0.045\textwidth]{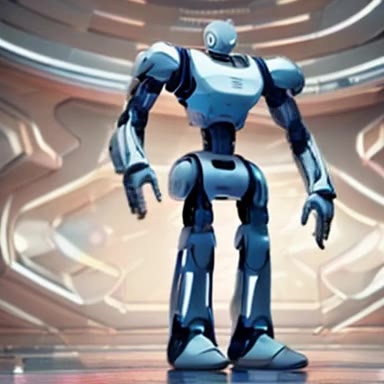}} & \raisebox{-.5\height}{\includegraphics[width=0.045\textwidth]{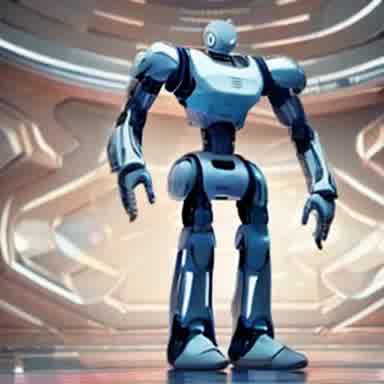}} & \raisebox{-.5\height}{\includegraphics[width=0.08\textwidth]{figures/qual_eval/face_nodding/first_frame.jpg}} & \raisebox{-.5\height}{\includegraphics[width=0.045\textwidth]{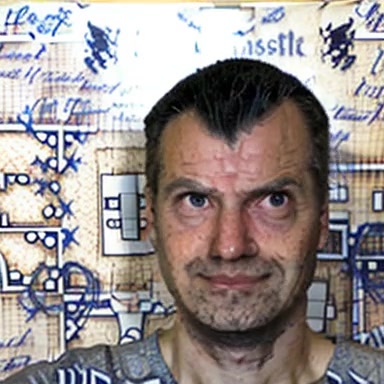}} & \raisebox{-.5\height}{\includegraphics[width=0.045\textwidth]{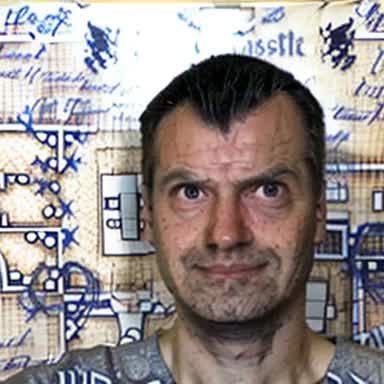}} & \raisebox{-.5\height}{\includegraphics[width=0.08\textwidth]{figures/qual_eval/camera_medium/first_frame.jpg}} & \raisebox{-.5\height}{\includegraphics[width=0.045\textwidth]{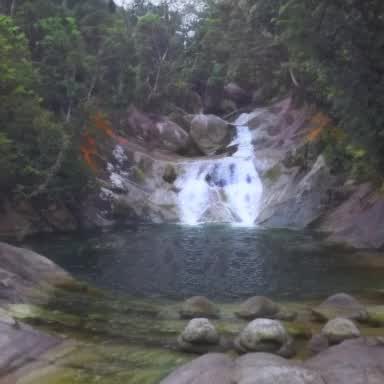}} & \raisebox{-.5\height}{\includegraphics[width=0.045\textwidth]{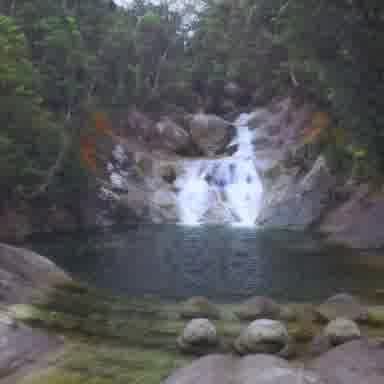}} \\
		{\textbf{Ours}} & \raisebox{-.5\height}{\includegraphics[width=0.08\textwidth]{figures/qual_eval/fullbody_jumping_jacks/first_frame.jpg}} & \raisebox{-.5\height}{\includegraphics[width=0.08\textwidth]{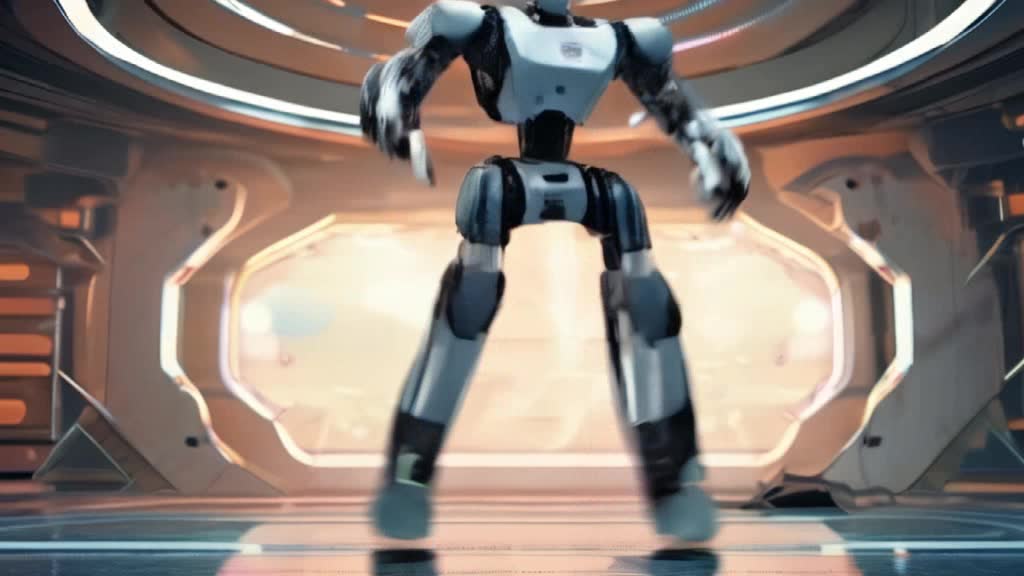}} & \raisebox{-.5\height}{\includegraphics[width=0.08\textwidth]{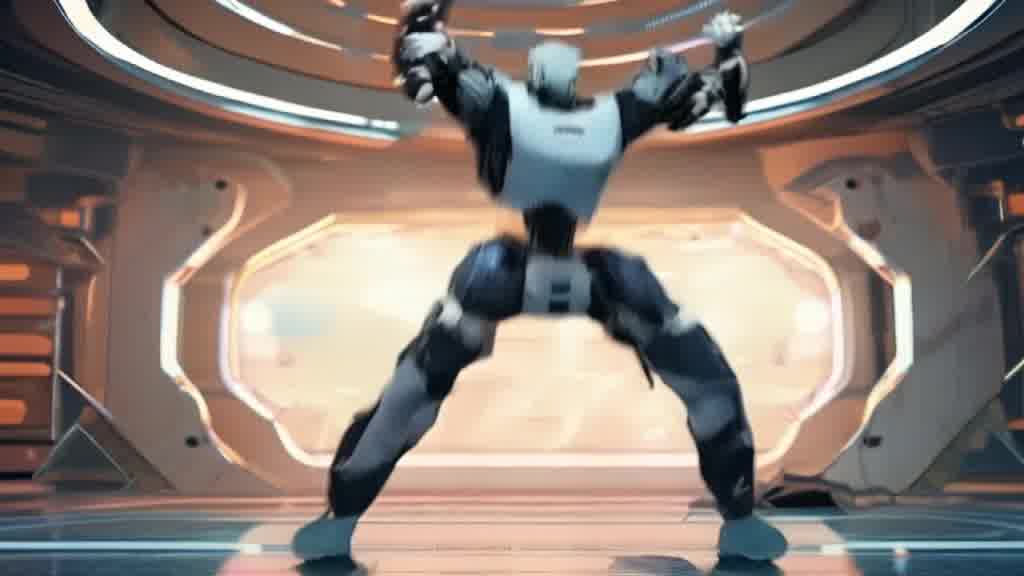}} & \raisebox{-.5\height}{\includegraphics[width=0.08\textwidth]{figures/qual_eval/face_nodding/first_frame.jpg}} & \raisebox{-.5\height}{\includegraphics[width=0.08\textwidth]{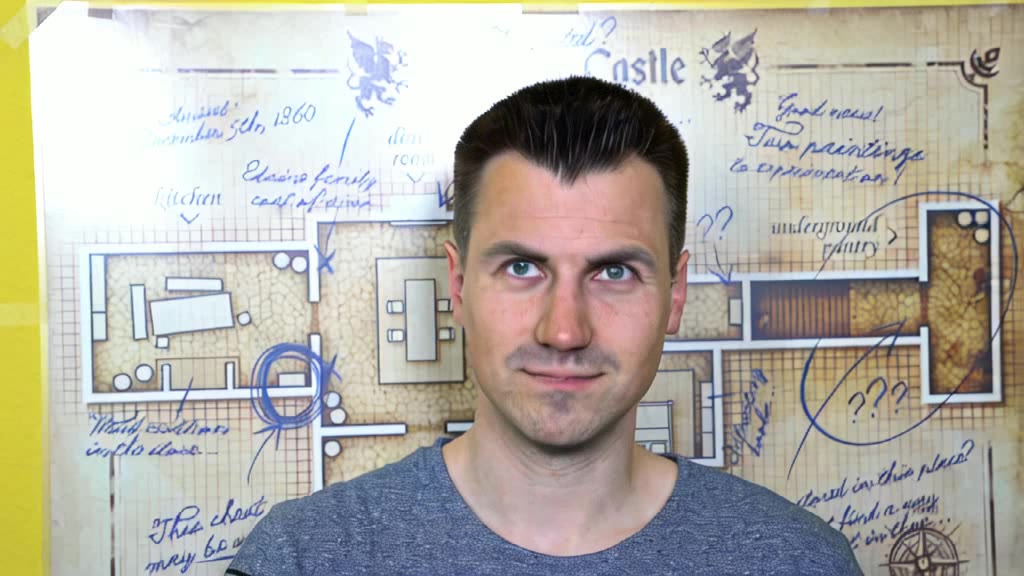}} &  \raisebox{-.5\height}{\includegraphics[width=0.08\textwidth]{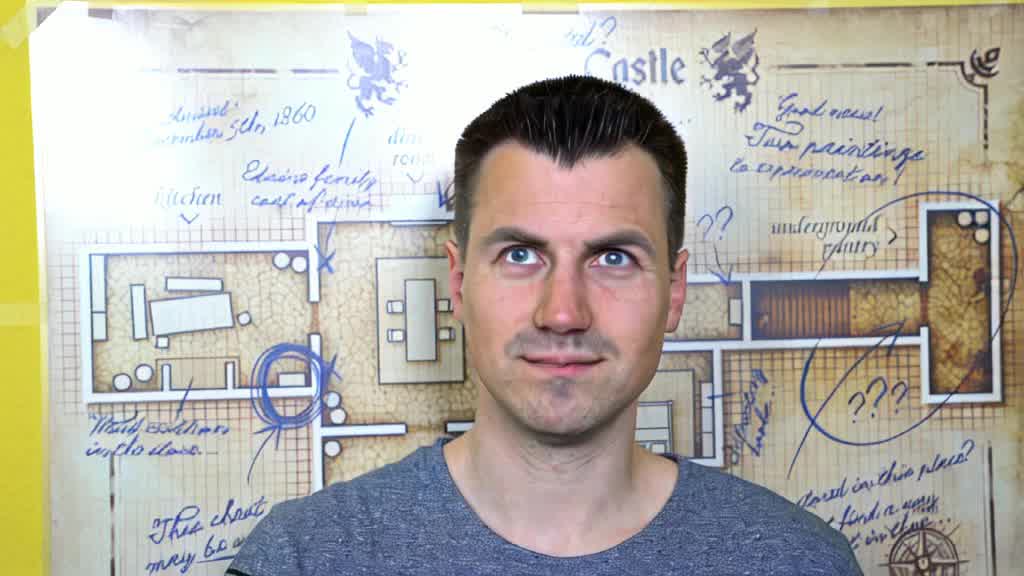}} & \raisebox{-.5\height}{\includegraphics[width=0.08\textwidth]{figures/qual_eval/camera_medium/first_frame.jpg}} & \raisebox{-.5\height}{\includegraphics[width=0.08\textwidth]{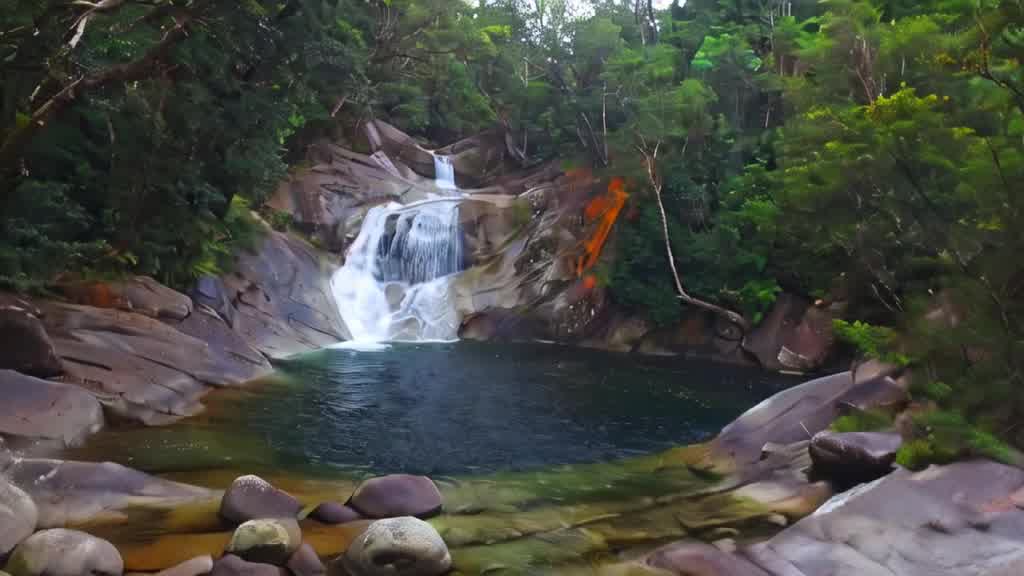}} & \raisebox{-.5\height}{\includegraphics[width=0.08\textwidth]{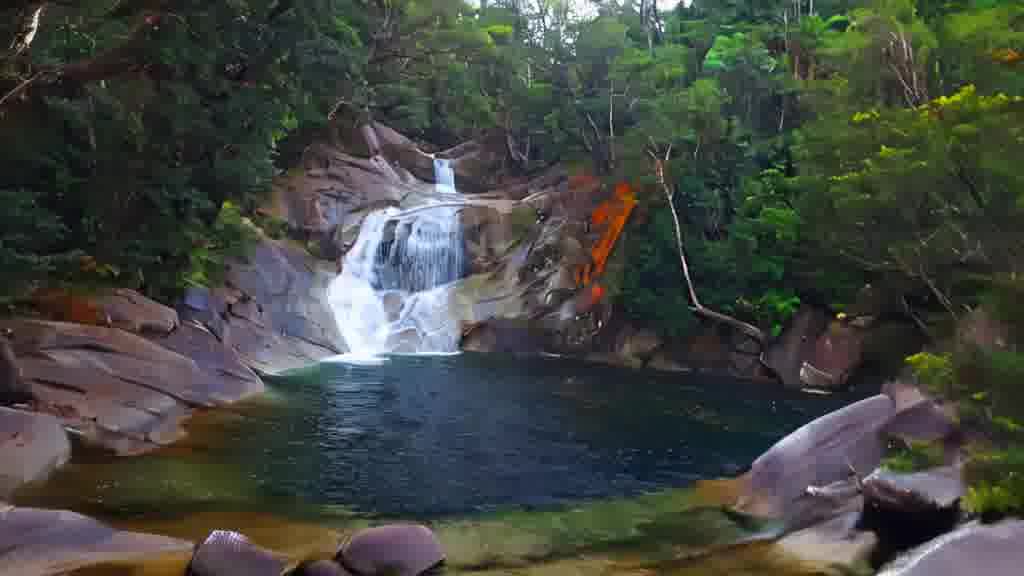}}
	\end{tblr}
	\caption{Qualitative evaluation. We compare our method to SVD = Stable Video Diffusion~\cite{svd} (baseline, no motion input), VC = VideoComposer~\cite{videocomposer}, MC = MotionClone~\cite{motionclone}, and MD = MotionDirector~\cite{motiondirector} for three different motions and target images: full-body reenactment, face reenactment, and camera motion. 
	}
	\Description{Grid with different methods in the rows and three different motion reference videos and starting frames in the columns. From left to right: jumping jack motion from human to robot, head nodding motion from human to human, camera flying forwards.}
	\label{fig:qual_eval}
\end{figure*}

\subsection{Qualitative Evaluation}

Fig.~\ref{fig:qual_eval} shows motion transfer results for three motions. As expected, the SVD baseline typically produces mismatched motions. For certain videos, like the face video, SVD produces significant artifacts and alters the subject identity. Due to its dense motion input, VideoComposer replicates motion in the spatial location of the reference video, leading to incorrect semantic motion and artifacts when structures misalign. MotionClone faces similar issues but handles minor structural differences better in the nodding example and has more high-level artifacts due to its higher-level motion representation. Since MotionDirector is based on a text-to-video model, it must learn the appearance and thus cannot continue naturally from the target image by design. Additionally, the motion is only transferred correctly for the head nodding example. Our method is the only one that preserves the input image's appearance and layout while successfully transferring the semantic motion of the video. Sections~\ref{sec:additional-qual-comp-baseline} and \ref{sec:additional-qual-comp-sota} provide additional qualitative comparisons, including an in-depth comparison with SVD and its embeddings.

\subsection{Quantitative Evaluation and User Study}

We evaluate our method on the Something-Something V2 data set~\cite{something_something}, selecting 10 classes from the validation set (5 with camera movements, 5 with object movements). For each class, one video serves as the motion reference, and 10 other videos' first frames act as target images, totaling 100 generated videos per method. This data set provides a challenging benchmark, as videos within each class have the same semantic action but vastly different spatial layouts. See Section~\ref{sec:additional-info-quant-eval} for further details.

For image appearance preservation, we calculate the mean cosine similarity between the CLIP~\cite{clip} image embeddings of the target image and the generated video, where \textbf{CLIP-Avg} is the average across all frames and \textbf{CLIP-1st} refers to the first frame. For video motion fidelity, we avoid metrics like optical flow or Motion-Fidelity-Score~\cite{space_time_diffusion}, which emphasize spatial over semantic motion. Instead, similar to MoTrans~\cite{motrans}, we use an action recognition network~\cite{videomae} trained on Something-Something V2 (174 classes). \textbf{Acc-Top-1} is the percentage of videos correctly classified, and \textbf{Acc-Top-5} the percentage with the correct class in the top 5 predictions. \textbf{Cos-Sim} is the cosine similarity between the logits of the generated and reference videos.

The results in Table~\ref{table:quant_eval} reflect our qualitative findings. SVD preserves the target image but fails to capture the motion. MotionDirector struggles with image preservation in the first frame, whereas image-to-video methods generally excel in this aspect by design. For motion fidelity, all competitor methods (except SVD) perform similarly, while our method outperforms them significantly.

Additionally, we conducted a user study with 27 users on a random subset of the evaluation data (one target image per motion video). For each of the 10 video sets, users ranked the methods from best (1) to worst (5) based on (a) \textbf{image appearance preservation}, (b) \textbf{video motion fidelity}, and (c) \textbf{overall} task fulfillment. The rankings align with the metrics but show an even stronger preference for our method. As seen in Table~\ref{table:quant_eval}, our method has the best average rank for motion fidelity and overall task fulfillment, voted best 75\% and 78\% of times respectively. It also performs well on appearance preservation, landing closely behind SVD. Note that this metric is biased towards methods that produce little motion, so it should only be regarded in combination with the motion fidelity.

\subsection{Ablation Study} \label{sec:ablation-study}

\begin{figure}
	\centering
	\begin{tblr}{
			vline{3} = {2-6}{dashed},
		}
		Reference & \raisebox{-.5\height}{\includegraphics[width=0.07\textwidth]{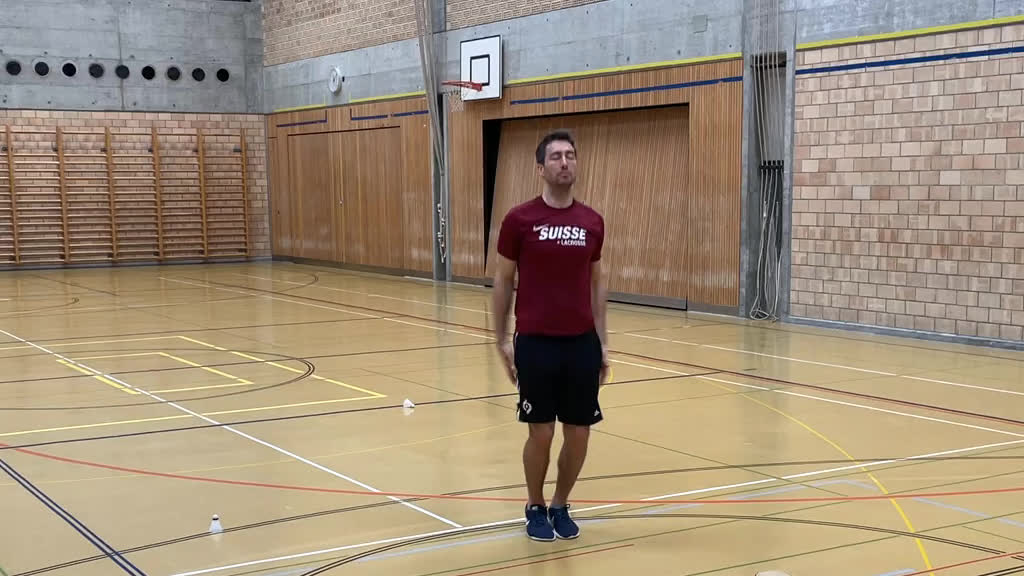}} & \raisebox{-.5\height}{\includegraphics[width=0.07\textwidth]{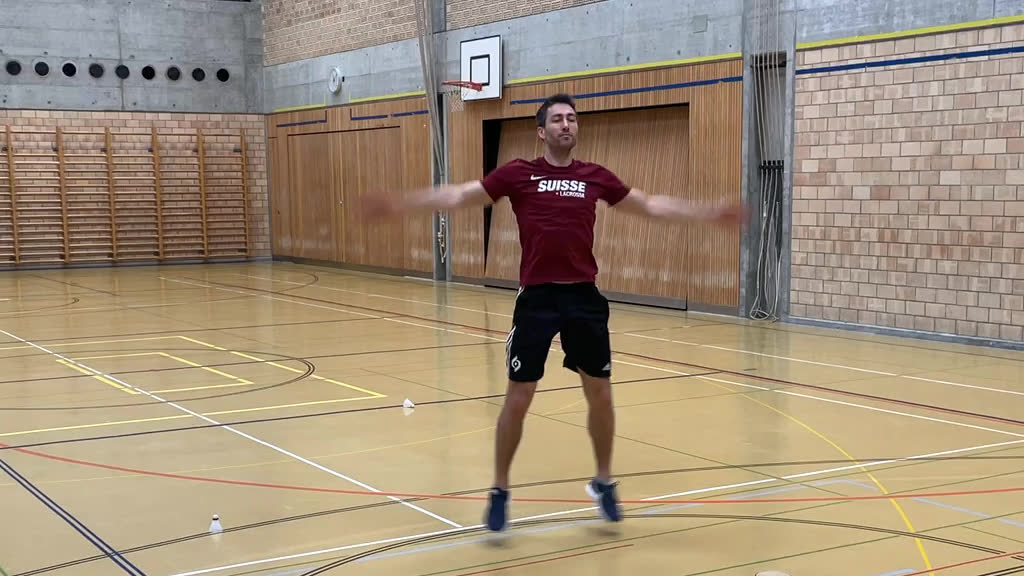}} \raisebox{-.5\height}{\includegraphics[width=0.07\textwidth]{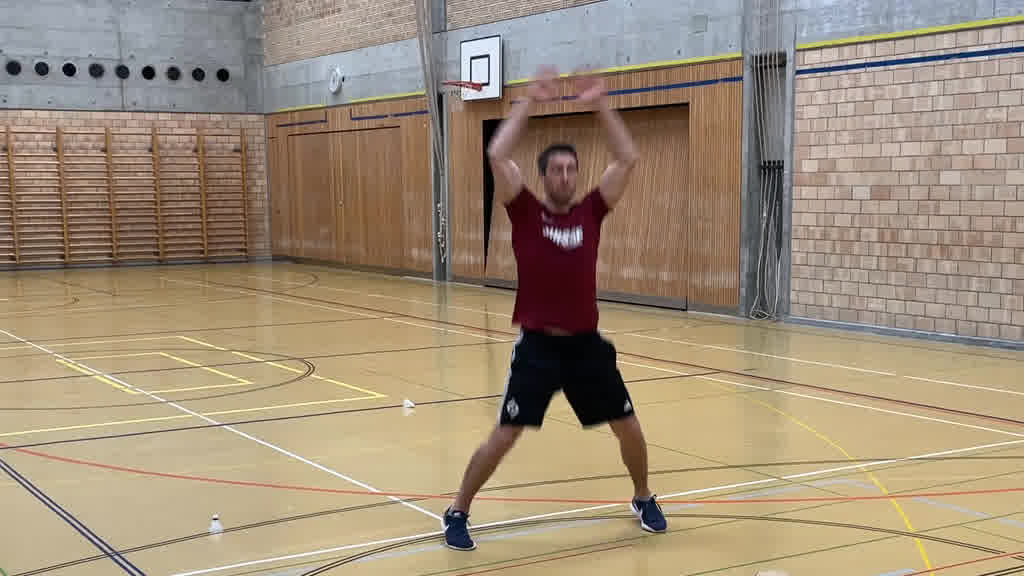}} \raisebox{-.5\height}{\includegraphics[width=0.07\textwidth]{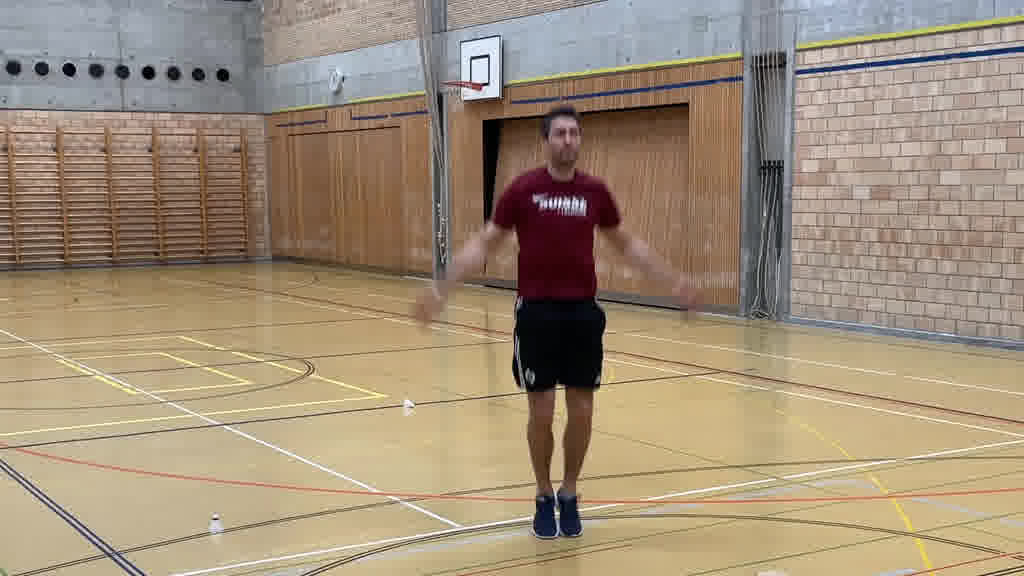}} \\
		$F'=1, N=1$ & \raisebox{-.5\height}{\includegraphics[width=0.07\textwidth]{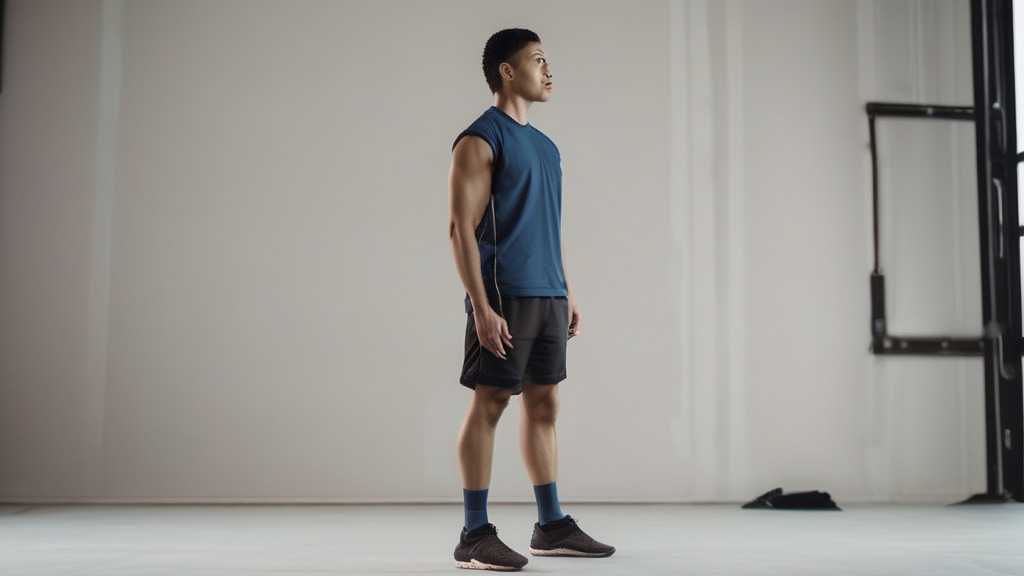}} & \raisebox{-.5\height}{\includegraphics[width=0.07\textwidth]{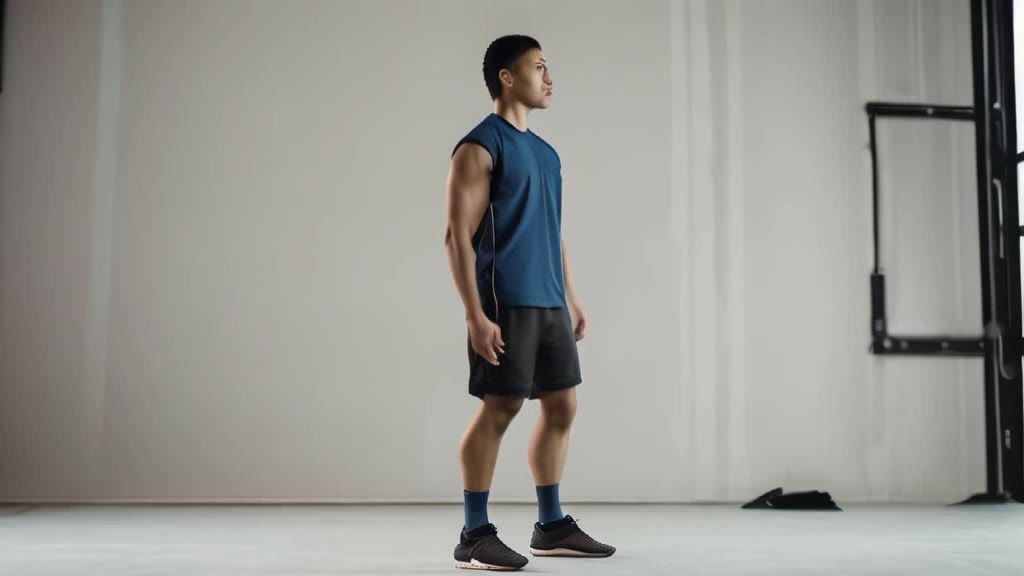}} \raisebox{-.5\height}{\includegraphics[width=0.07\textwidth]{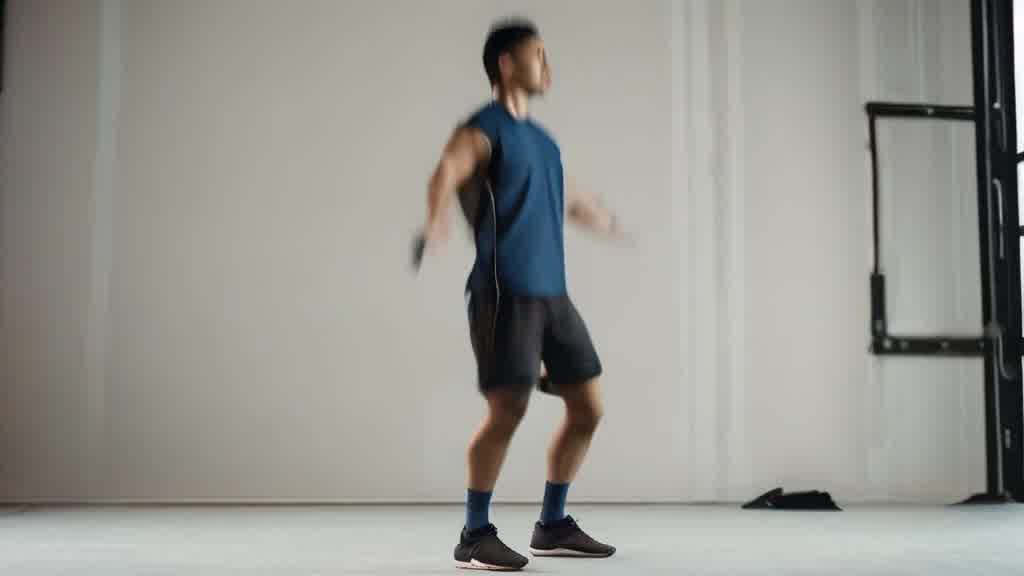}} \raisebox{-.5\height}{\includegraphics[width=0.07\textwidth]{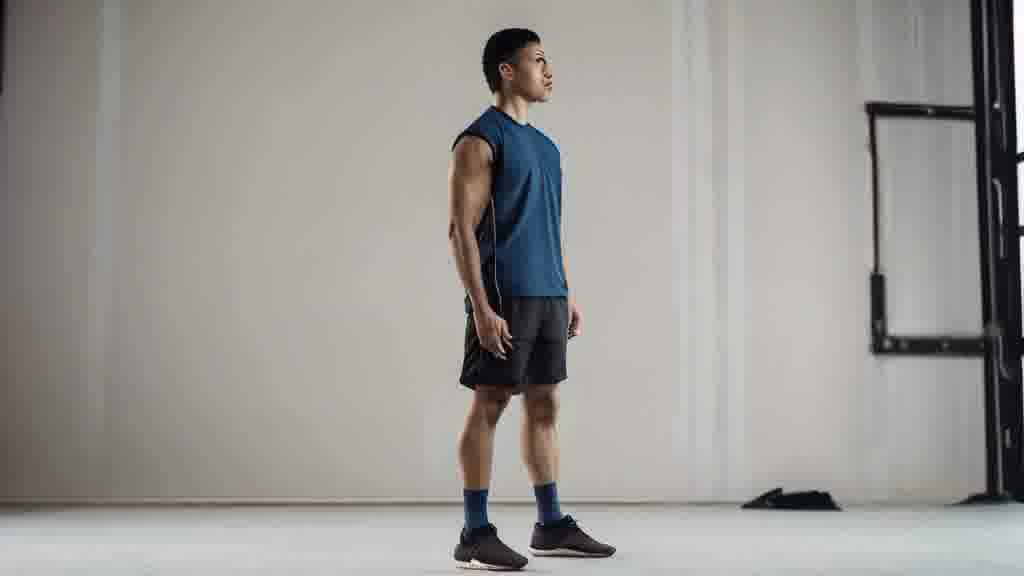}} \\
		$F'=1, N=15$ & \raisebox{-.5\height}{\includegraphics[width=0.07\textwidth]{figures/ablation_study/01.jpg}} & \raisebox{-.5\height}{\includegraphics[width=0.07\textwidth]{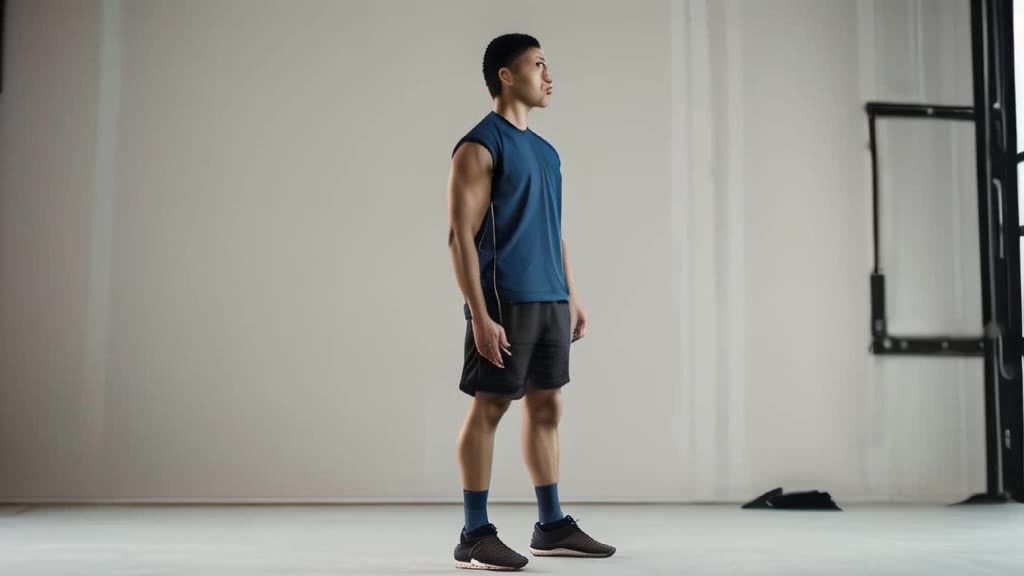}} \raisebox{-.5\height}{\includegraphics[width=0.07\textwidth]{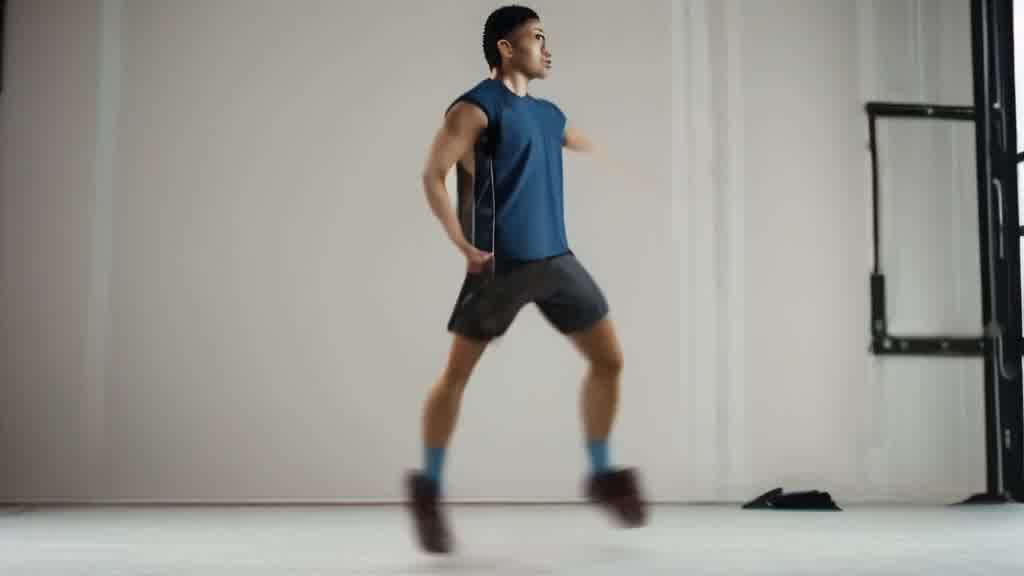}} \raisebox{-.5\height}{\includegraphics[width=0.07\textwidth]{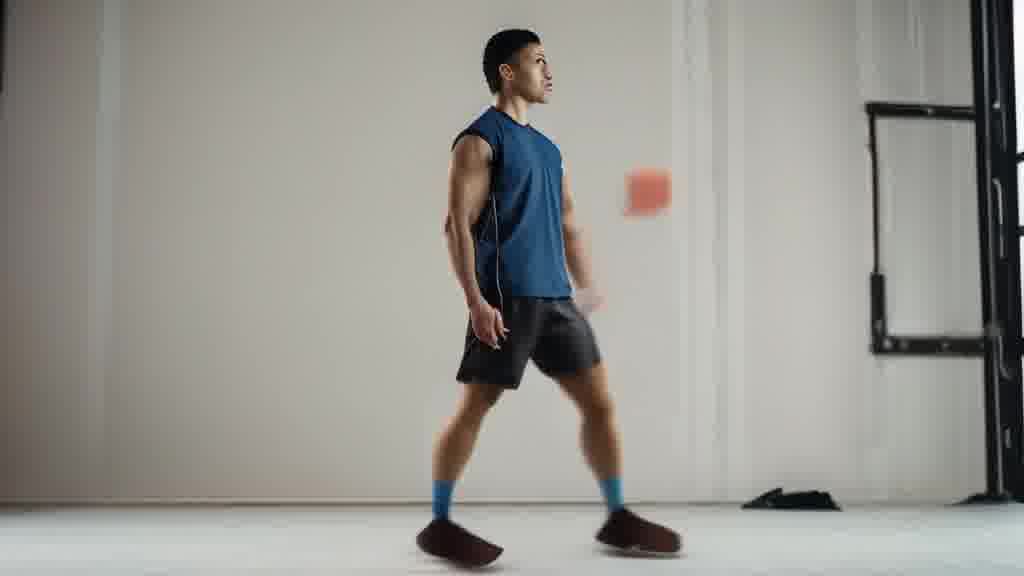}} \\
		$F'=15, N=1$ & \raisebox{-.5\height}{\includegraphics[width=0.07\textwidth]{figures/ablation_study/01.jpg}} & \raisebox{-.5\height}{\includegraphics[width=0.07\textwidth]{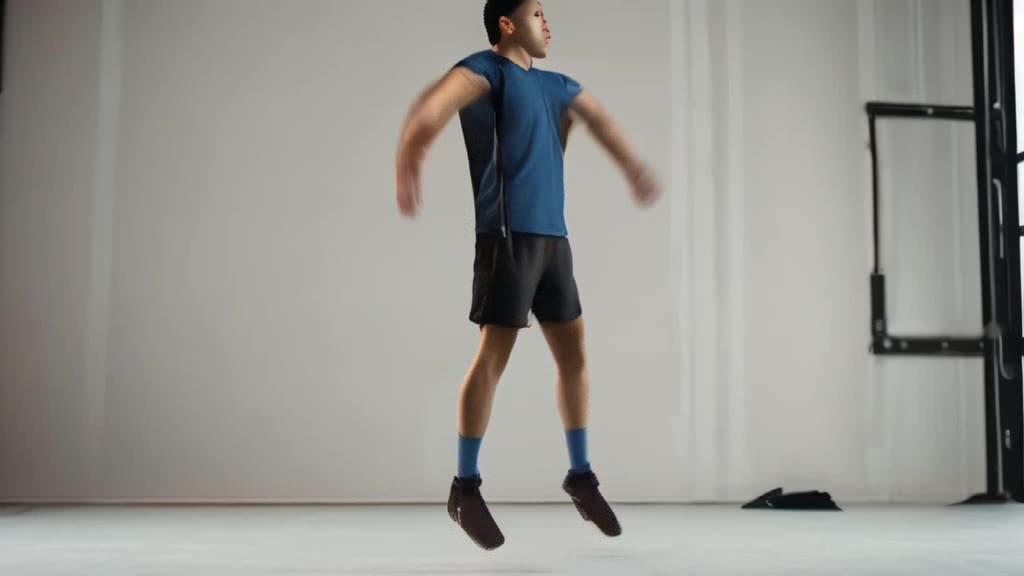}} \raisebox{-.5\height}{\includegraphics[width=0.07\textwidth]{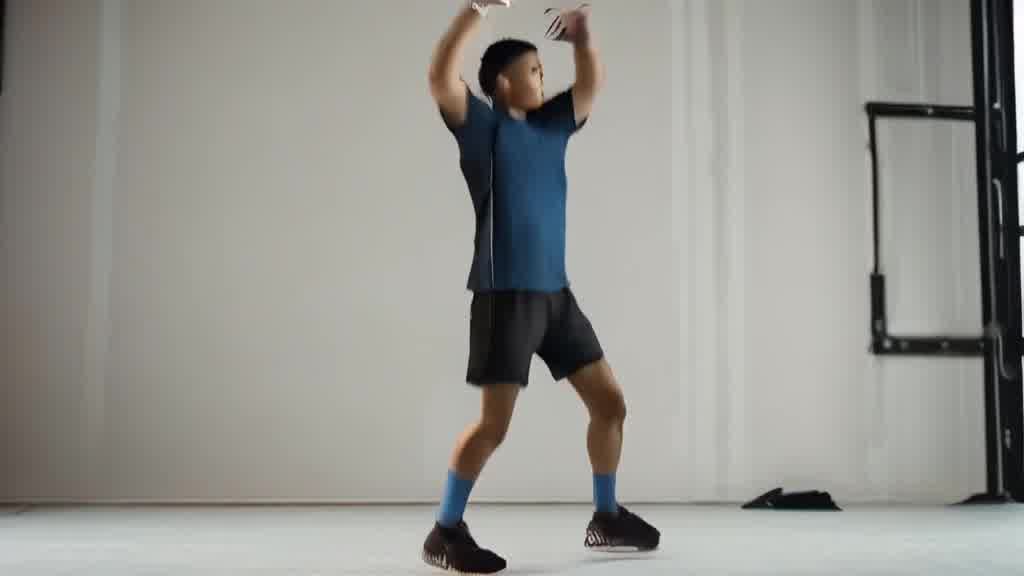}} \raisebox{-.5\height}{\includegraphics[width=0.07\textwidth]{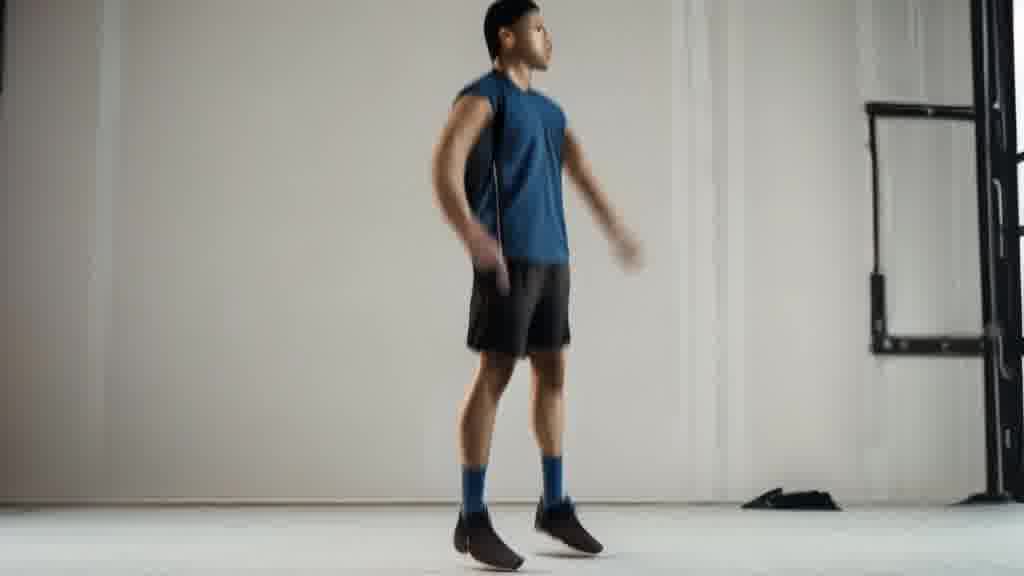}} \\
		{$F'=15, N=5$ \\ (Default)} & \raisebox{-.5\height}{\includegraphics[width=0.07\textwidth]{figures/ablation_study/01.jpg}} & \raisebox{-.5\height}{\includegraphics[width=0.07\textwidth]{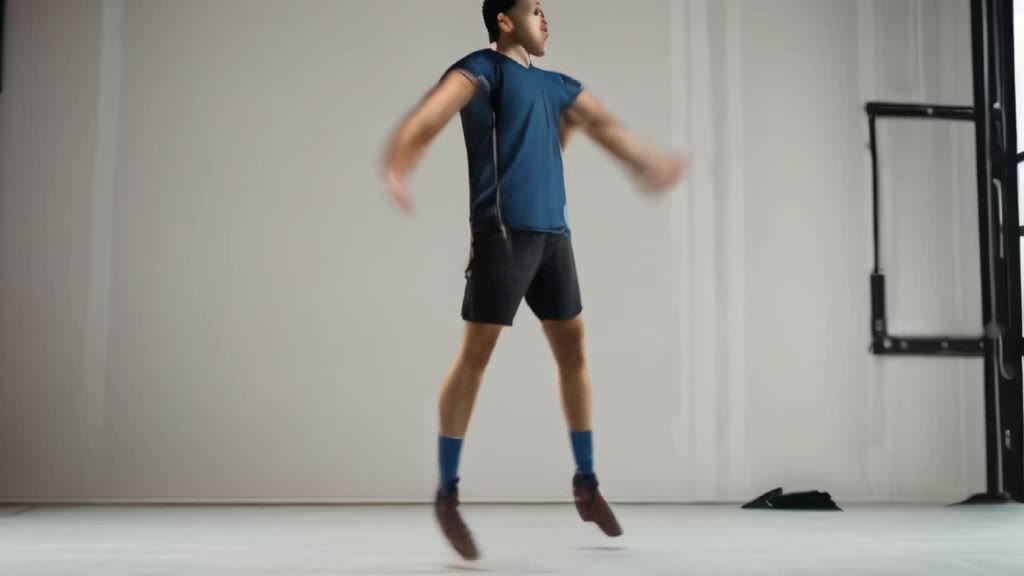}} \raisebox{-.5\height}{\includegraphics[width=0.07\textwidth]{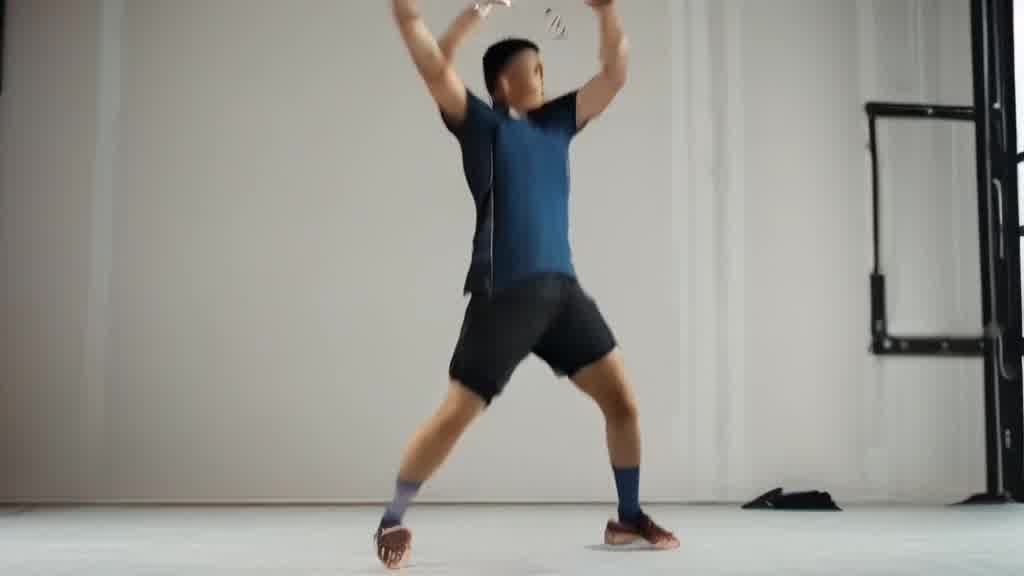}} \raisebox{-.5\height}{\includegraphics[width=0.07\textwidth]{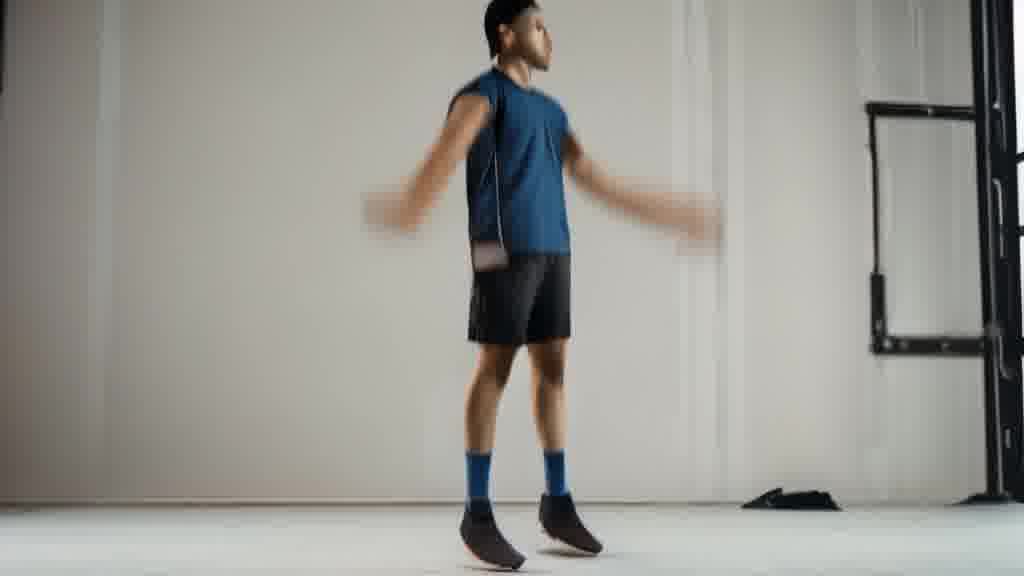}} \\
		$F'=15, N=15$ & \raisebox{-.5\height}{\includegraphics[width=0.07\textwidth]{figures/ablation_study/01.jpg}} & \raisebox{-.5\height}{\includegraphics[width=0.07\textwidth]{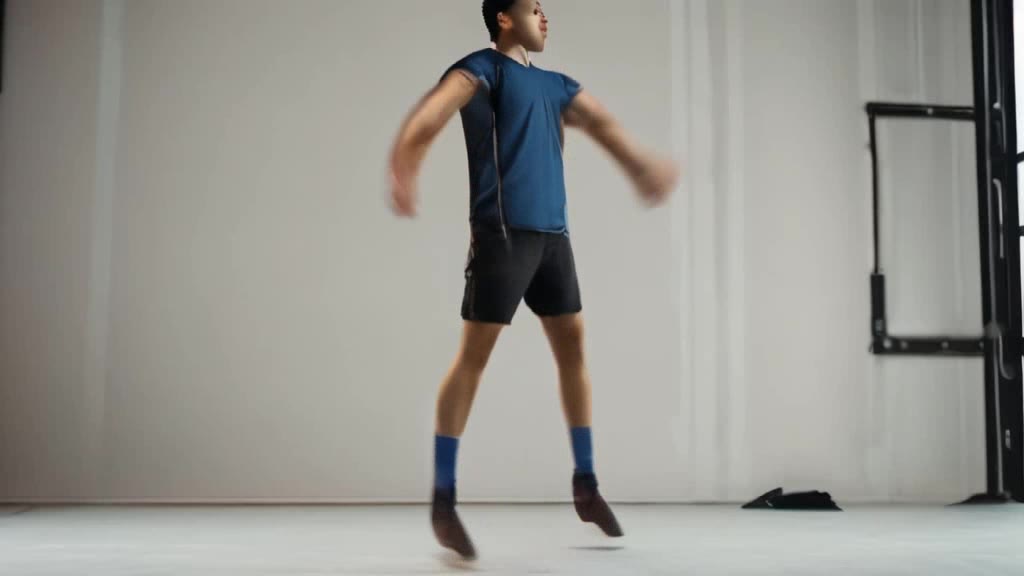}} \raisebox{-.5\height}{\includegraphics[width=0.07\textwidth]{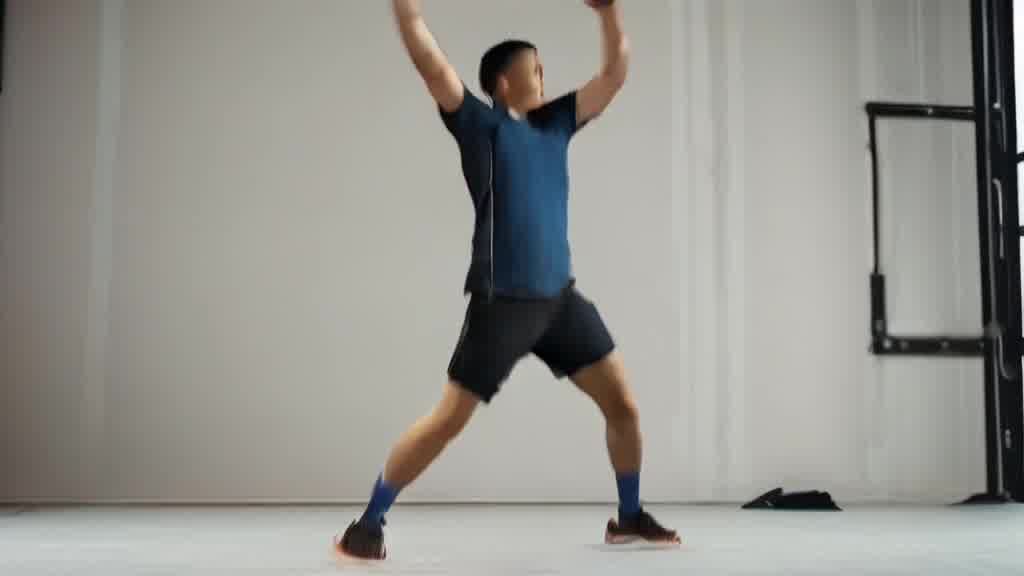}} \raisebox{-.5\height}{\includegraphics[width=0.07\textwidth]{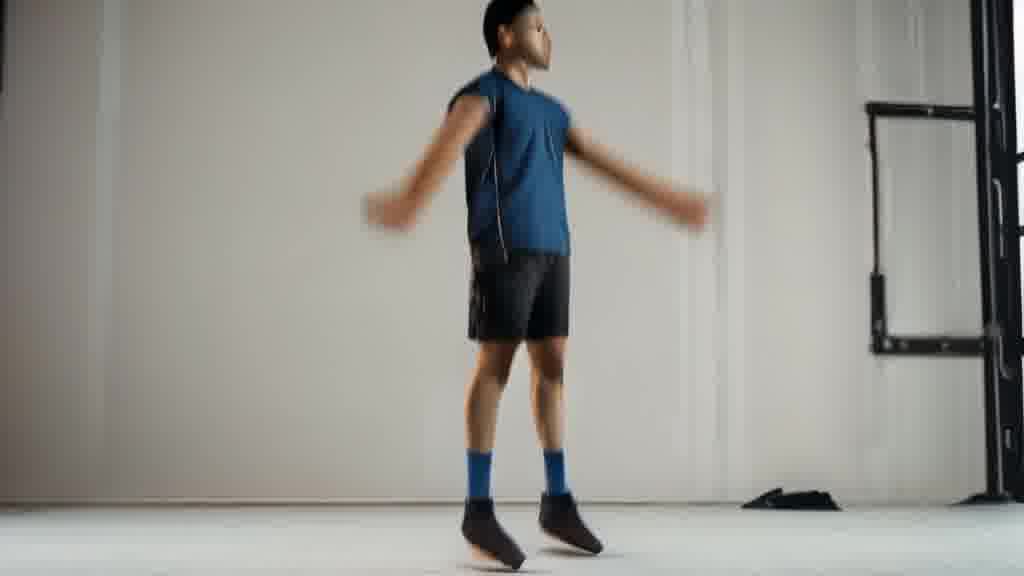}}
	\end{tblr}
	\caption{Ablation. Our proposed motion-text embedding inflation is crucial for successful motion transfer. While adding more tokens (increasing $N$) improves the results already, the biggest gain comes from having different tokens for each frame (where $F' = F+1 = 15$).}
	\Description{The first row shows frames of the motion reference video of a person doing jumping jacks. Below, each row shows one setting for the motion-text embedding. The lower the row, the more tokens are added, and the performance improves dramatically (especially once having different tokens per frame) but eventually saturates.}
	\label{fig:ablation}
\end{figure}

Our motion-text embedding inflation is key to high-quality motion transfer. Fig.~\ref{fig:ablation} shows different embedding configurations. A single token captures only limited motion. Adding more tokens shared across frames helps, but the crucial factor is \emph{having different tokens per frame}. Rows 2 and 3 both use 15 tokens, but allowing the embedding to adapt frame-wise performs significantly better, especially for complex motions. Increasing tokens per frame further improves results slightly before saturating, so we default to $N=5$. Section~\ref{sec:additional-ablation} provides two additional qualitative examples for this ablation as well as quantitative results when using the same protocol as for the above state-of-the-art comparison.

\begin{figure}
	\centering
	\begin{tblr}{
			vline{3} = {2,4,6}{dashed},
			hline{3} = {1-5}{},
			hline{5} = {1-5}{},
		}
		{Ref.} & \raisebox{-.5\height}{\includegraphics[width=0.07\textwidth]{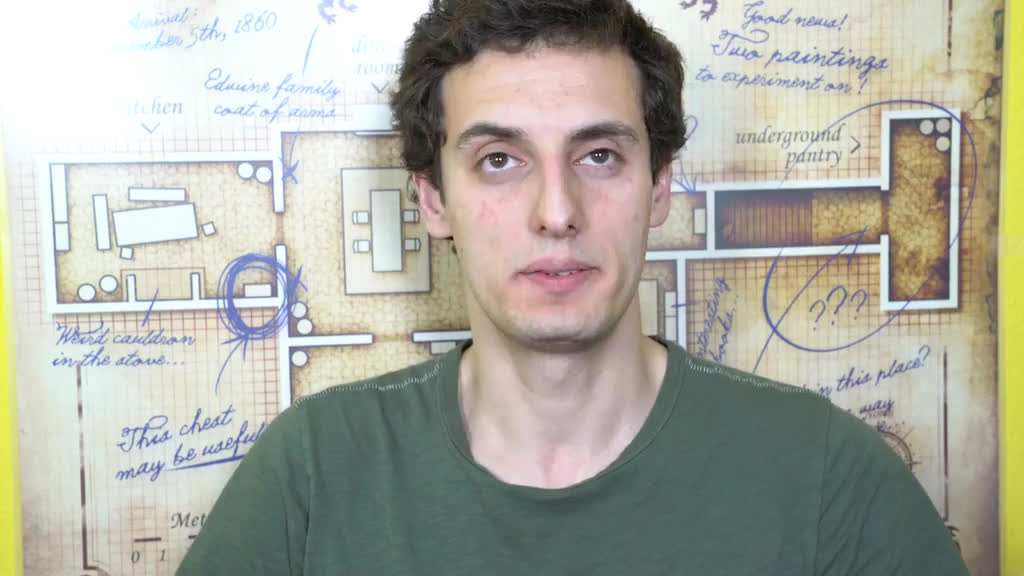}} & \raisebox{-.5\height}{\includegraphics[width=0.07\textwidth]{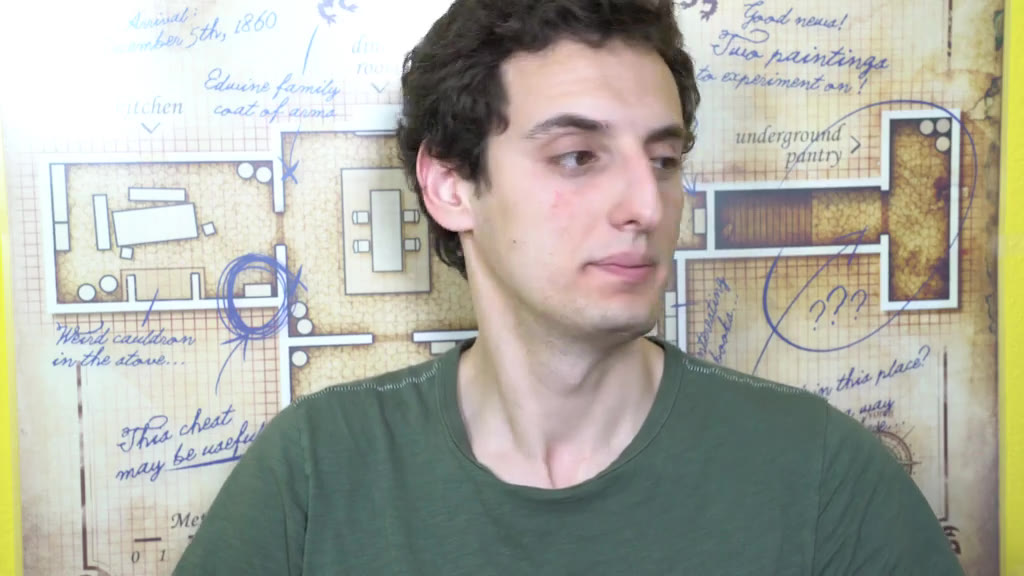}} \raisebox{-.5\height}{\includegraphics[width=0.07\textwidth]{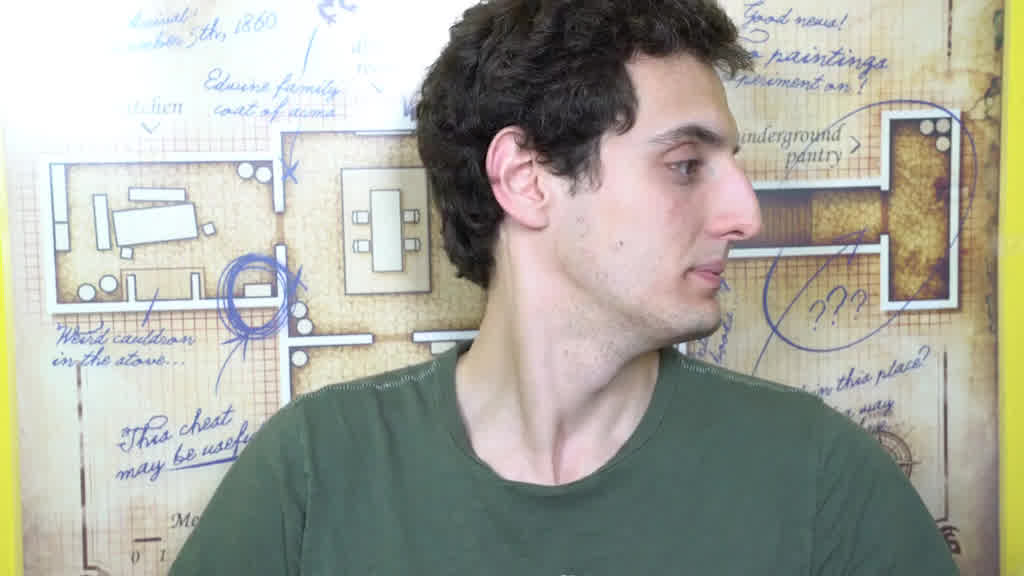}} \raisebox{-.5\height}{\includegraphics[width=0.07\textwidth]{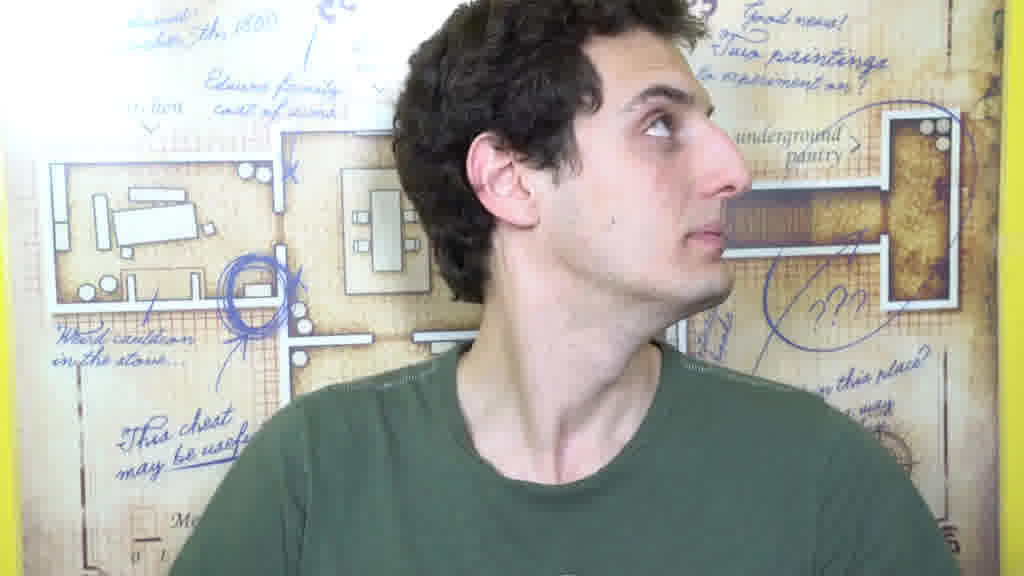}} \\
		{Gen.}  & \raisebox{-.5\height}{\includegraphics[width=0.07\textwidth]{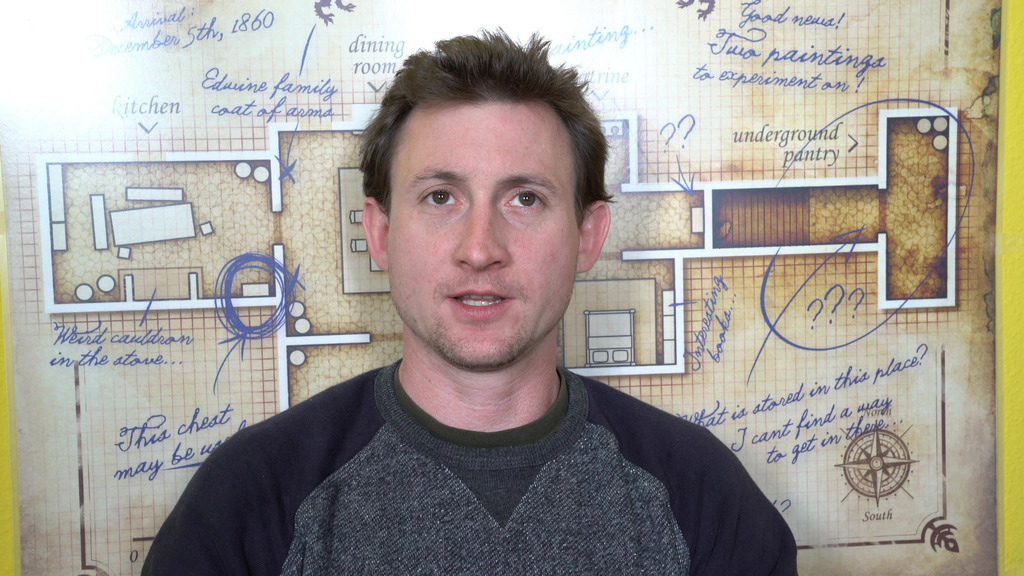}} & \raisebox{-.5\height}{\includegraphics[width=0.07\textwidth]{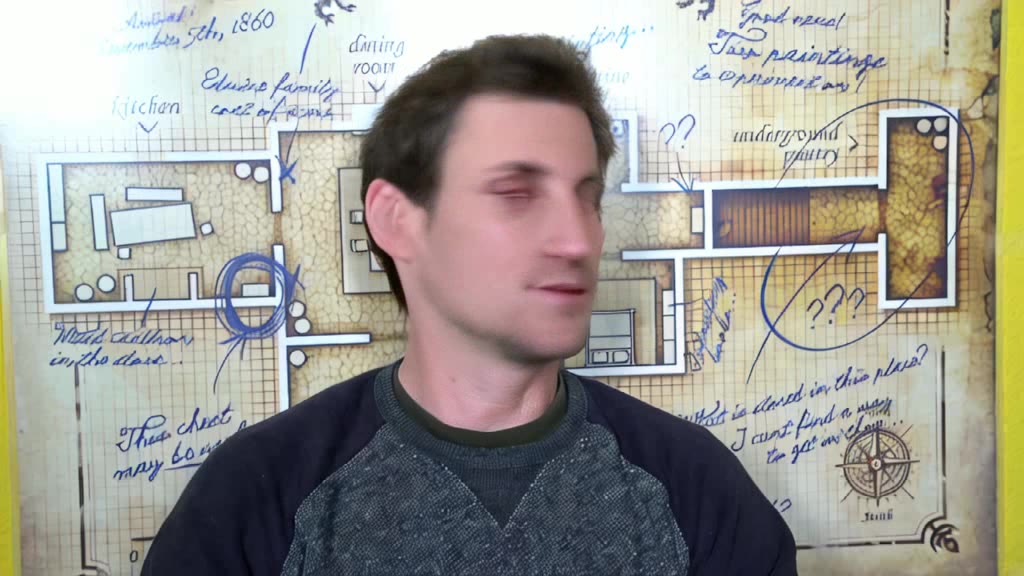}} \raisebox{-.5\height}{\includegraphics[width=0.07\textwidth]{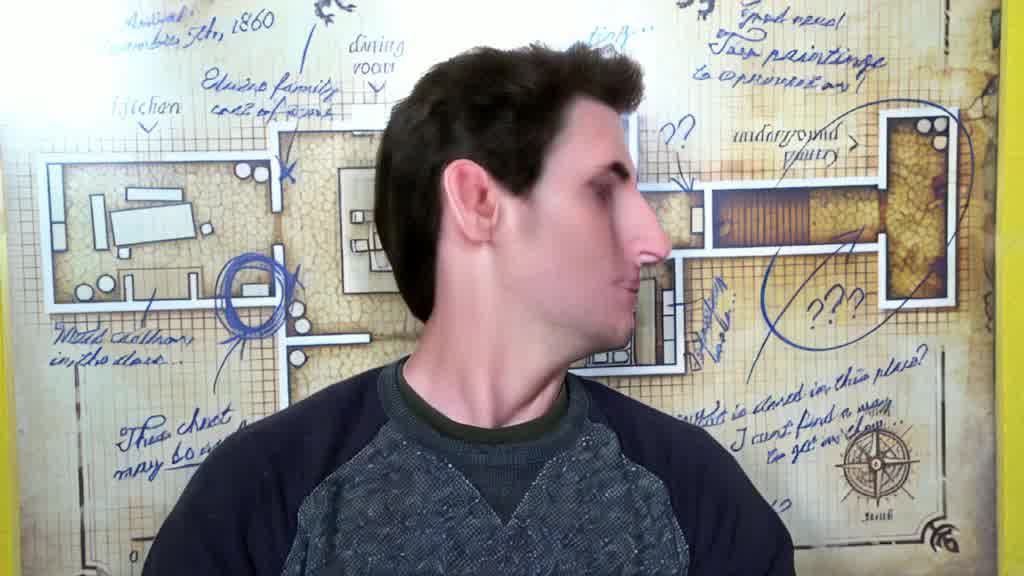}} \raisebox{-.5\height}{\includegraphics[width=0.07\textwidth]{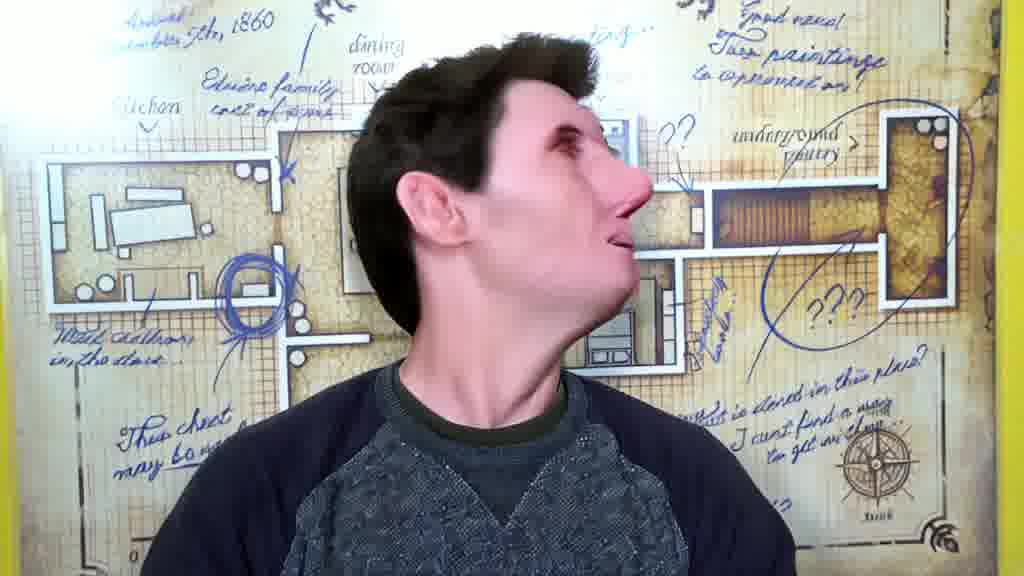}} \\
		{Ref.} & \raisebox{-.5\height}{\includegraphics[width=0.07\textwidth]{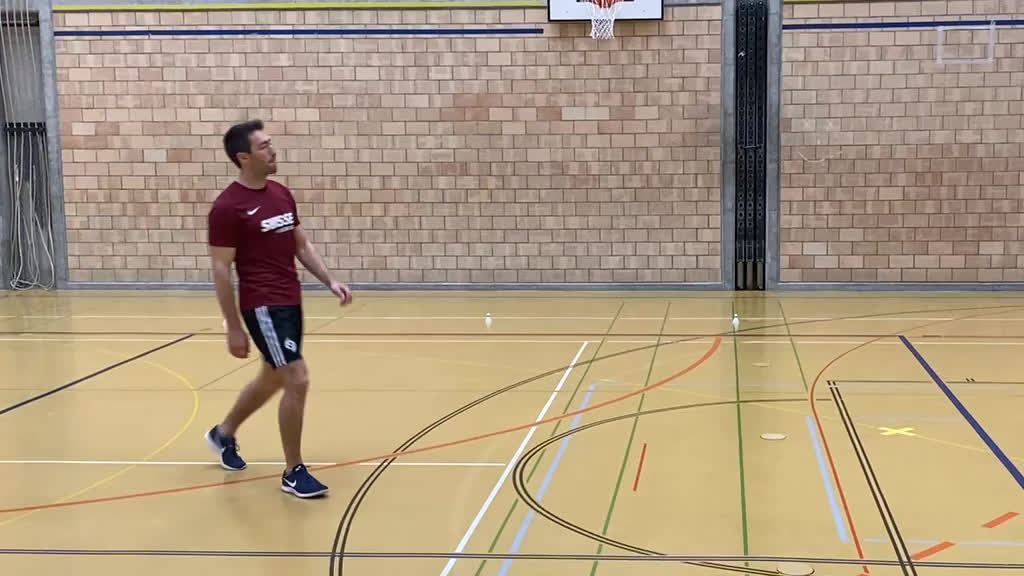}} & \raisebox{-.5\height}{\includegraphics[width=0.07\textwidth]{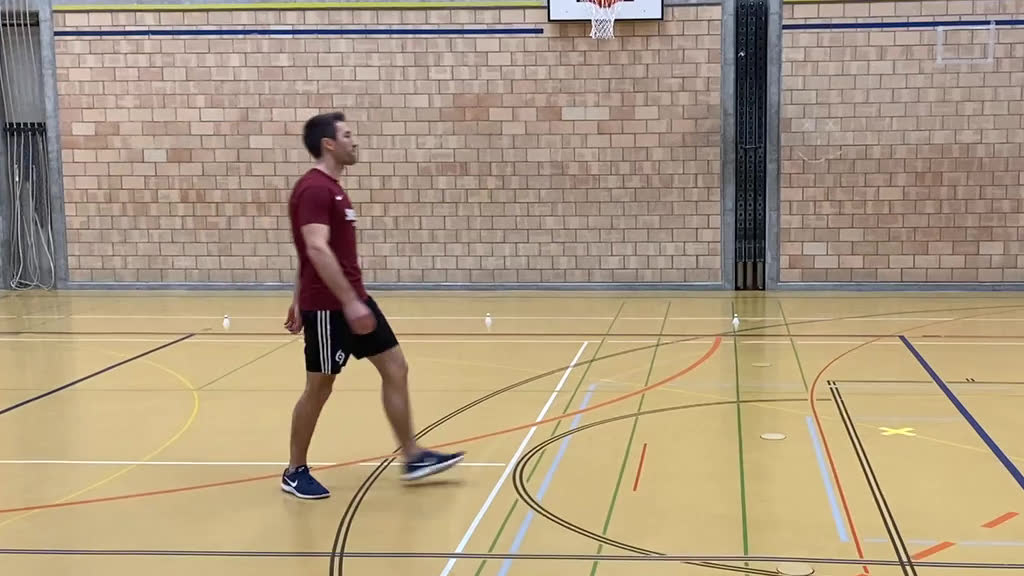}} \raisebox{-.5\height}{\includegraphics[width=0.07\textwidth]{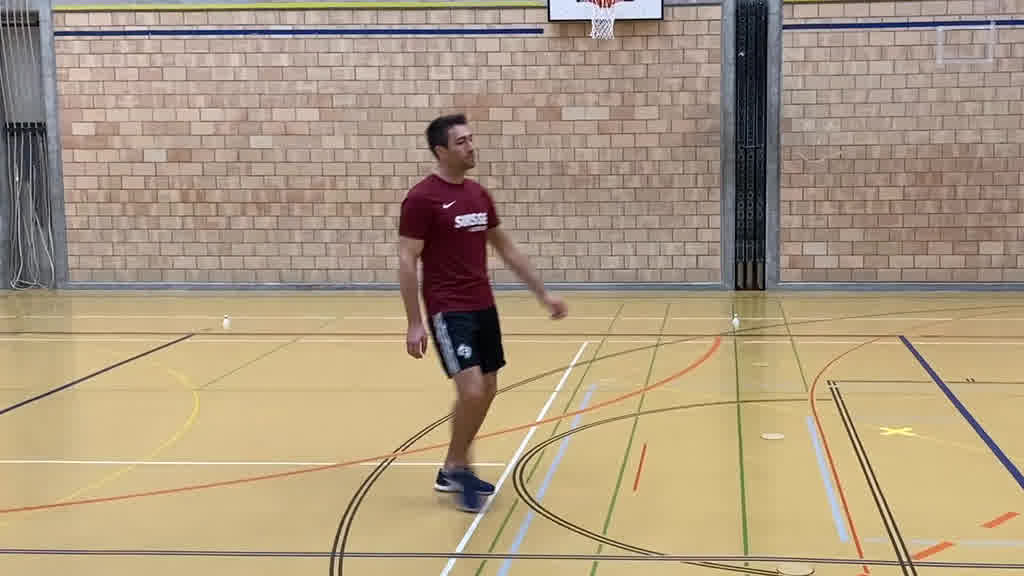}} \raisebox{-.5\height}{\includegraphics[width=0.07\textwidth]{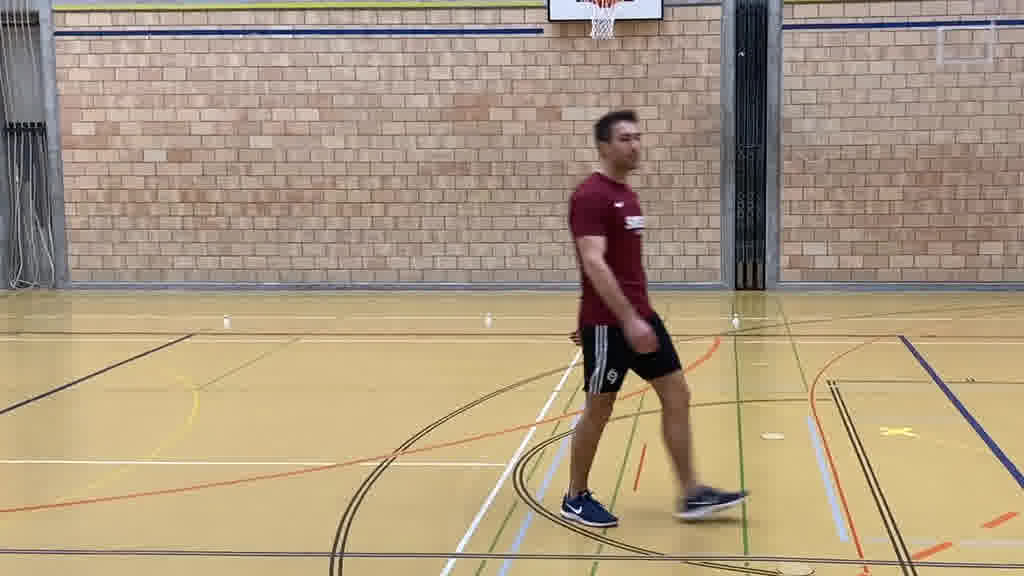}} \\
		{Gen.}  & \raisebox{-.5\height}{\includegraphics[width=0.07\textwidth]{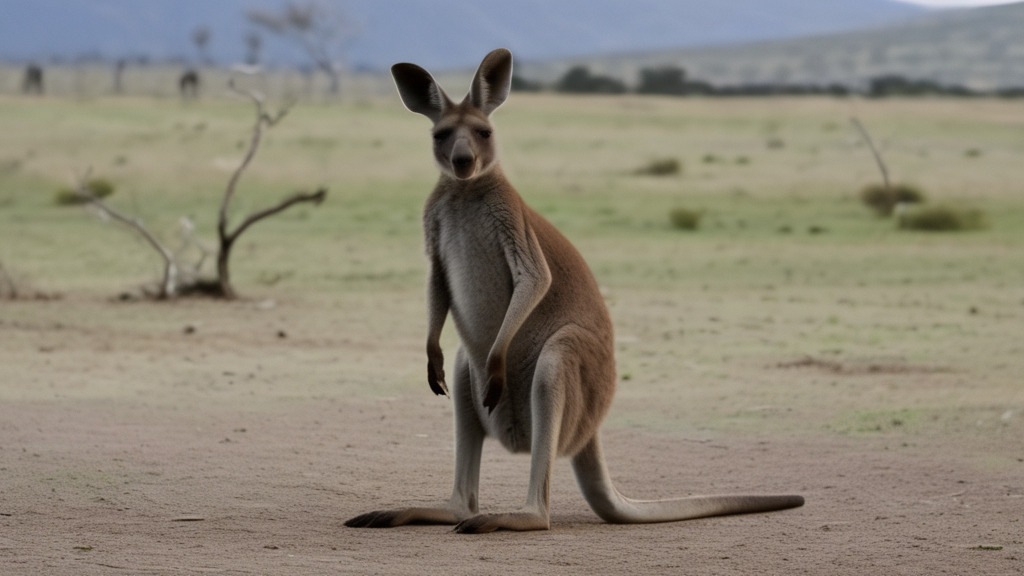}} & \raisebox{-.5\height}{\includegraphics[width=0.07\textwidth]{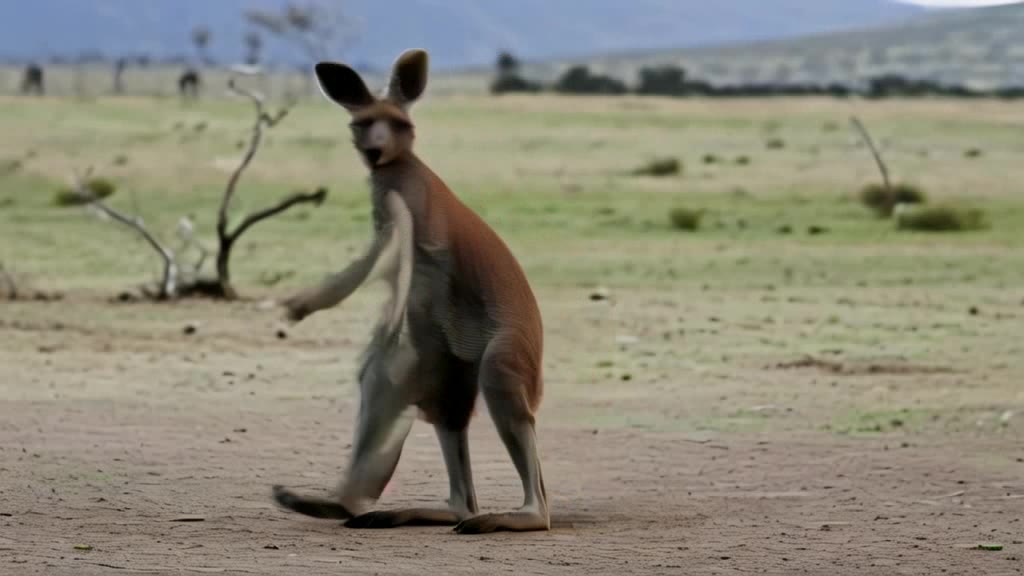}} \raisebox{-.5\height}{\includegraphics[width=0.07\textwidth]{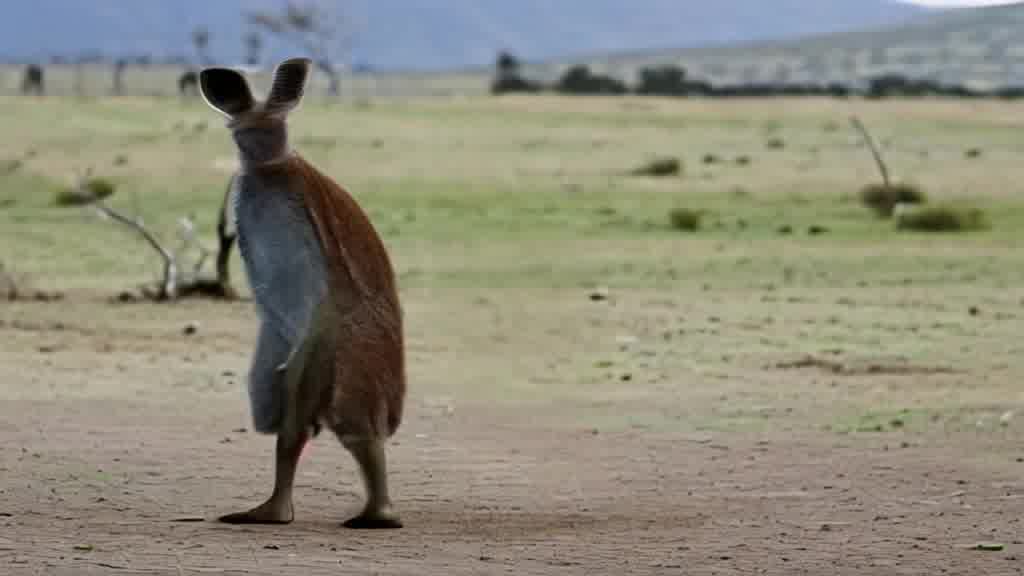}} \raisebox{-.5\height}{\includegraphics[width=0.07\textwidth]{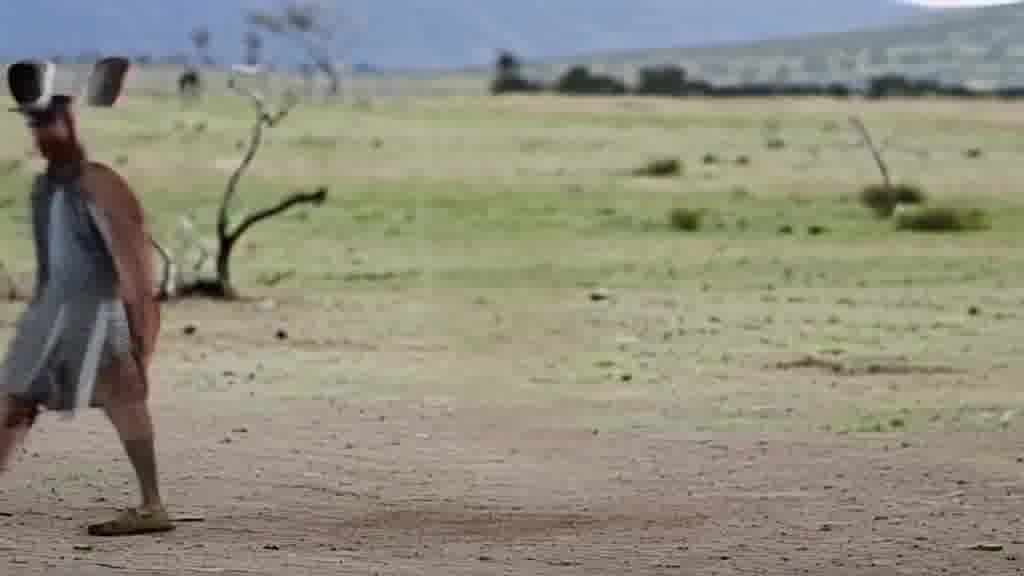}} \\
		{Ref.} & \raisebox{-.5\height}{\includegraphics[width=0.07\textwidth]{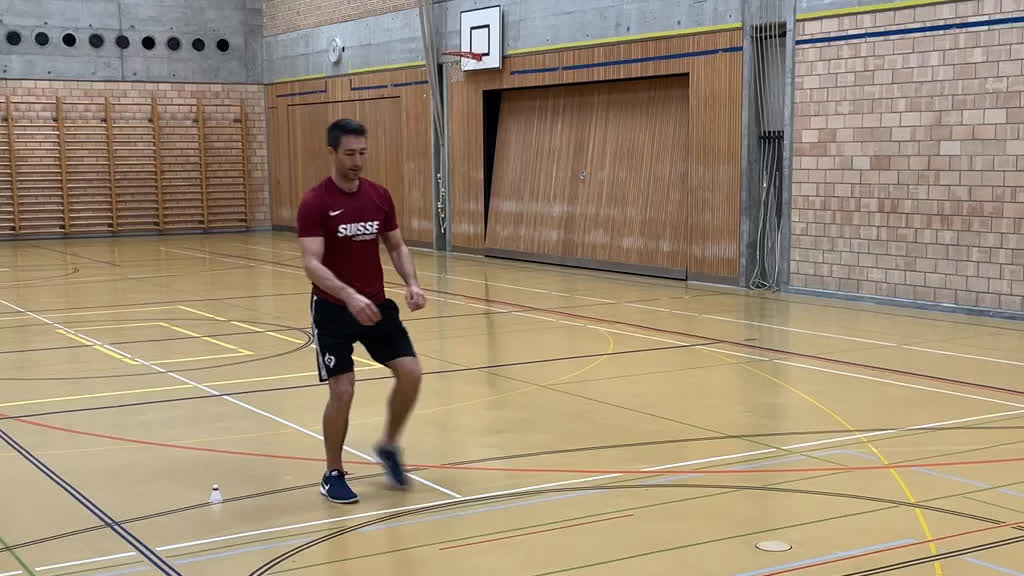}} & \raisebox{-.5\height}{\includegraphics[width=0.07\textwidth]{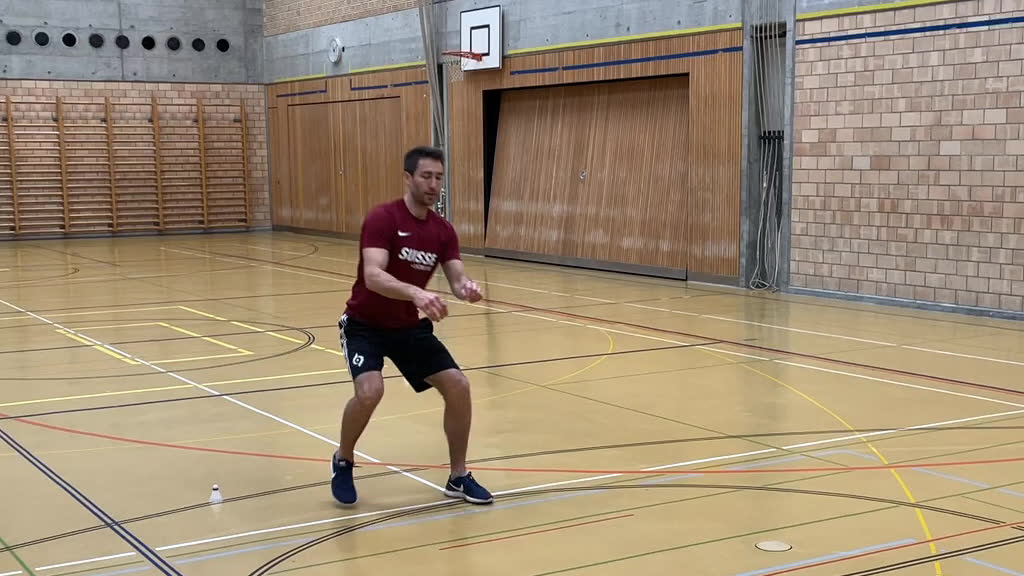}} \raisebox{-.5\height}{\includegraphics[width=0.07\textwidth]{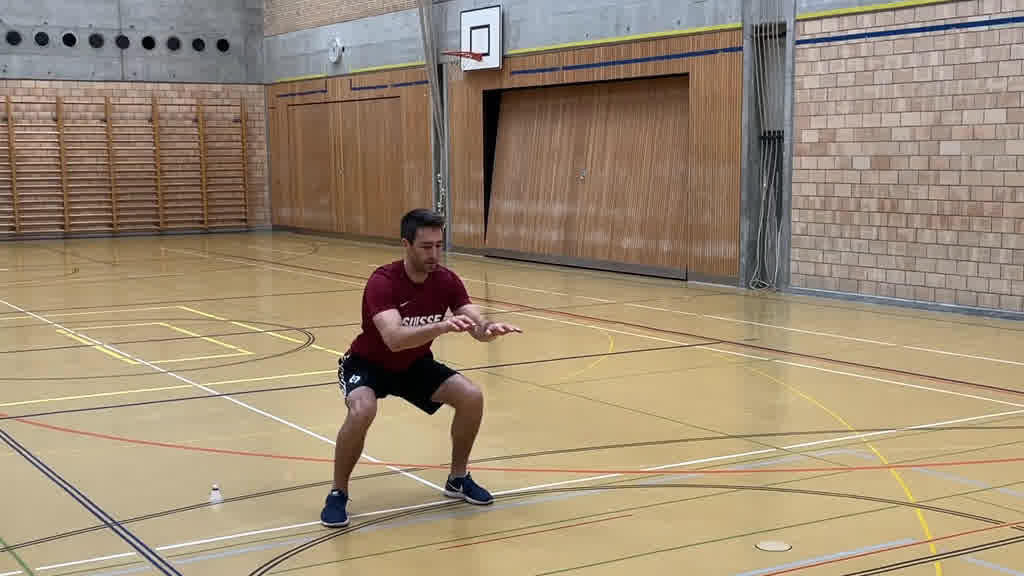}} \raisebox{-.5\height}{\includegraphics[width=0.07\textwidth]{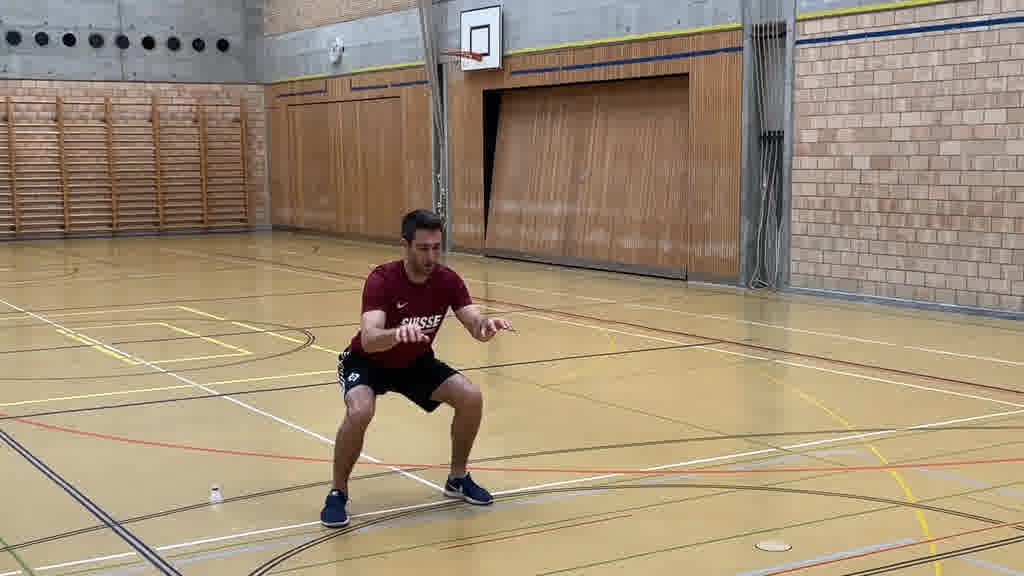}} \\
		{Gen.} & \raisebox{-.5\height}{\includegraphics[width=0.07\textwidth]{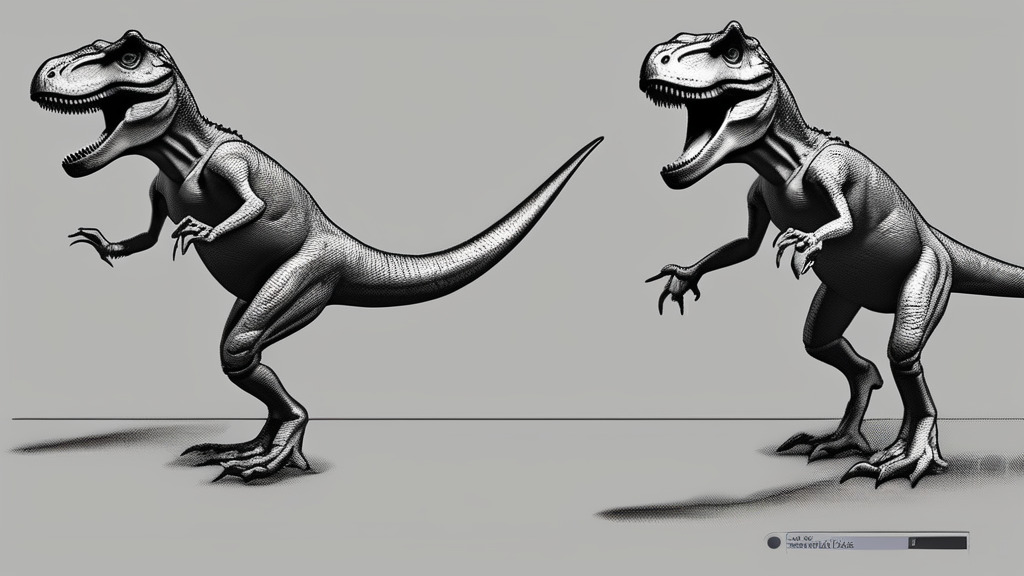}} & \raisebox{-.5\height}{\includegraphics[width=0.07\textwidth]{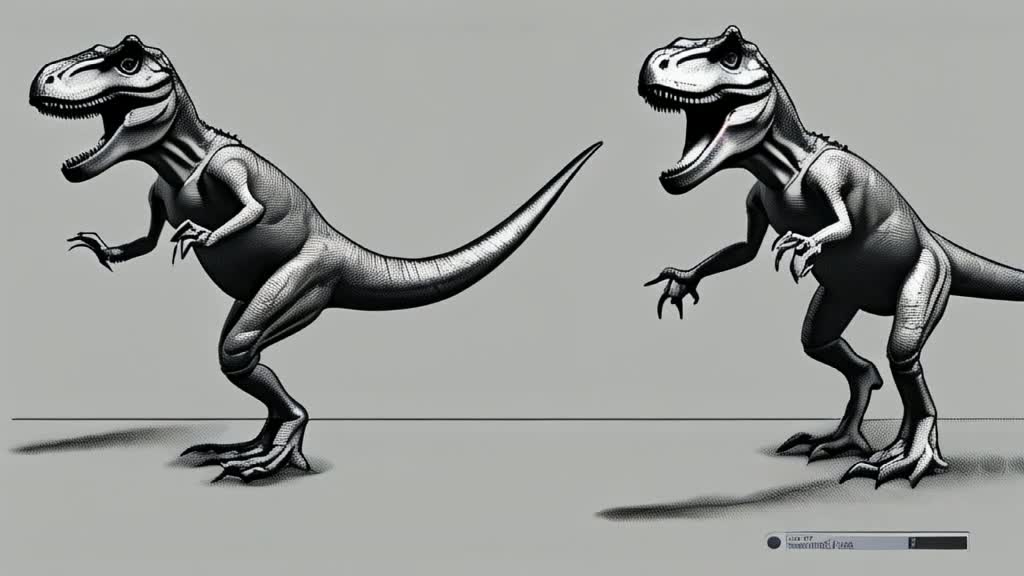}} \raisebox{-.5\height}{\includegraphics[width=0.07\textwidth]{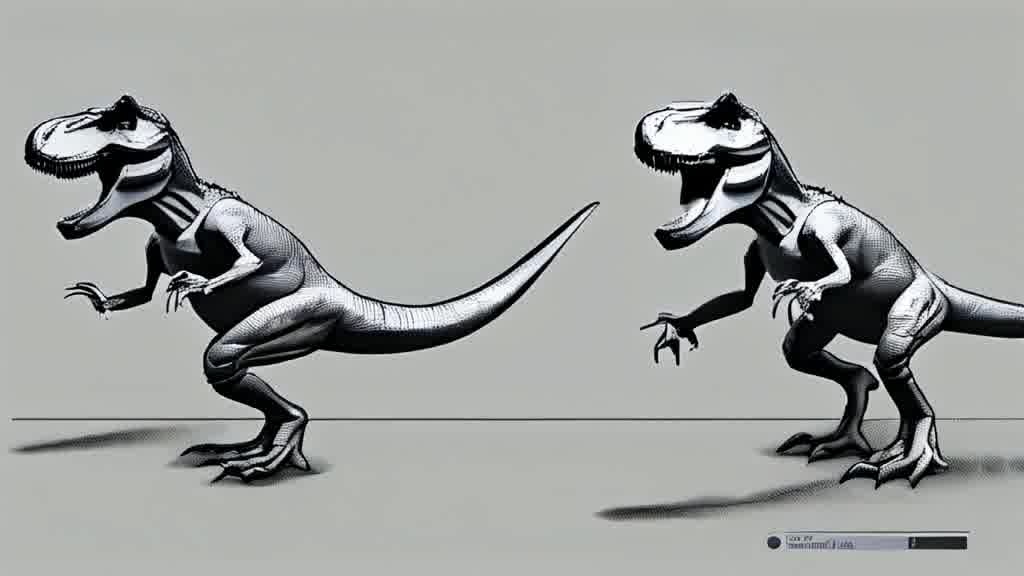}} \raisebox{-.5\height}{\includegraphics[width=0.07\textwidth]{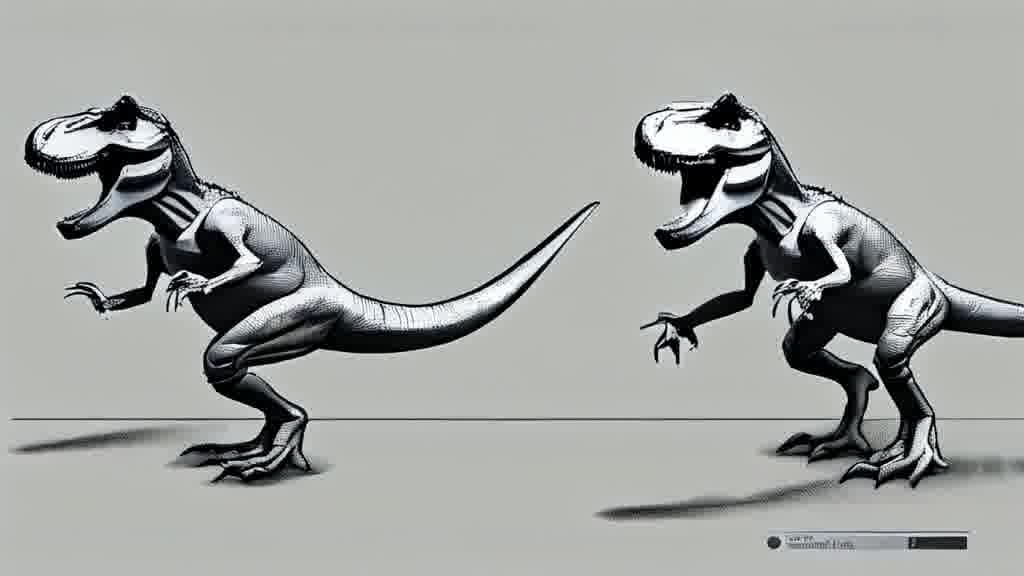}}
	\end{tblr}
	\caption{Failure cases. Our method is limited by the priors and quality of the pre-trained image-to-video model, which may lead to artifacts (e.g., identity changes as head moves in first example). Furthermore, there may be some structure leakage in some cases, leading to certain characteristics from the motion reference video being visible (e.g., human-like legs on a kangaroo in second example). Lastly, our method struggles to transfer spatially fine-grained motion at times (e.g., typing motion not transferred to dinosaurs in third example).}
	\Description{The grid shows three examples of common failure cases of our model. In the first set of videos, there are some artifacts when the subject turns its head, but the motion is correct. In the second set, a kangaroo's legs start becoming human-like when it starts walking like the human in the motion reference video. In the third set, a person sits down and does a typing motion in the air. While the dinosaurs sit down, their fingers do not move.}
	\label{fig:failure}
\end{figure}

\subsection{Results}

\begin{figure*}
	\centering
	\begin{tblr}{
			vline{3} = {2,4,6,8}{dashed},
			vline{4} = {1-8}{},
			vline{5} = {2,4,6,8}{dashed},
			hline{3} = {1-5}{},
			hline{5} = {1-5}{},
			hline{7} = {1-5}{},
		}
		{Ref.} & \raisebox{-.5\height}{\includegraphics[width=0.09\textwidth]{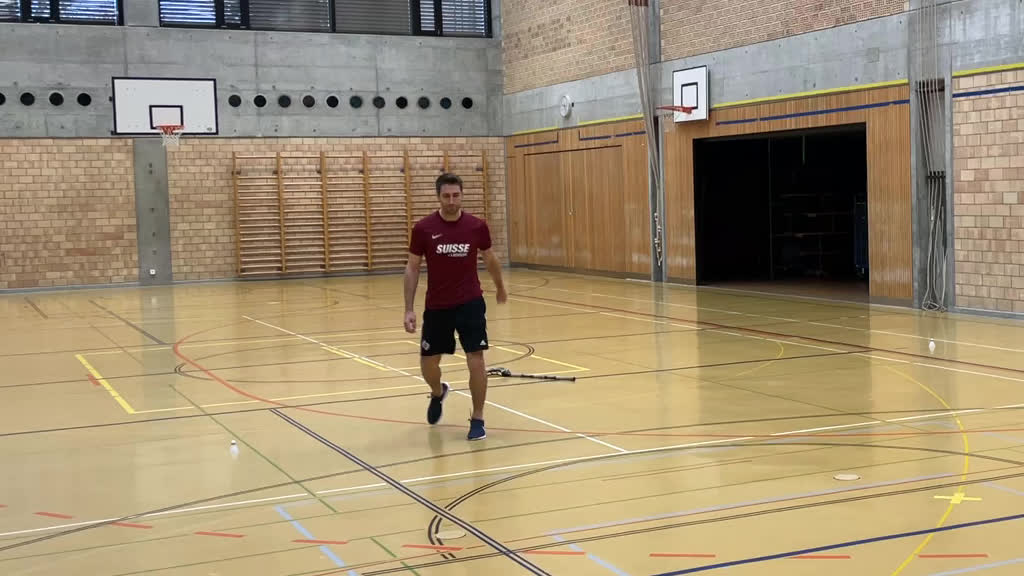}} & \raisebox{-.5\height}{\includegraphics[width=0.09\textwidth]{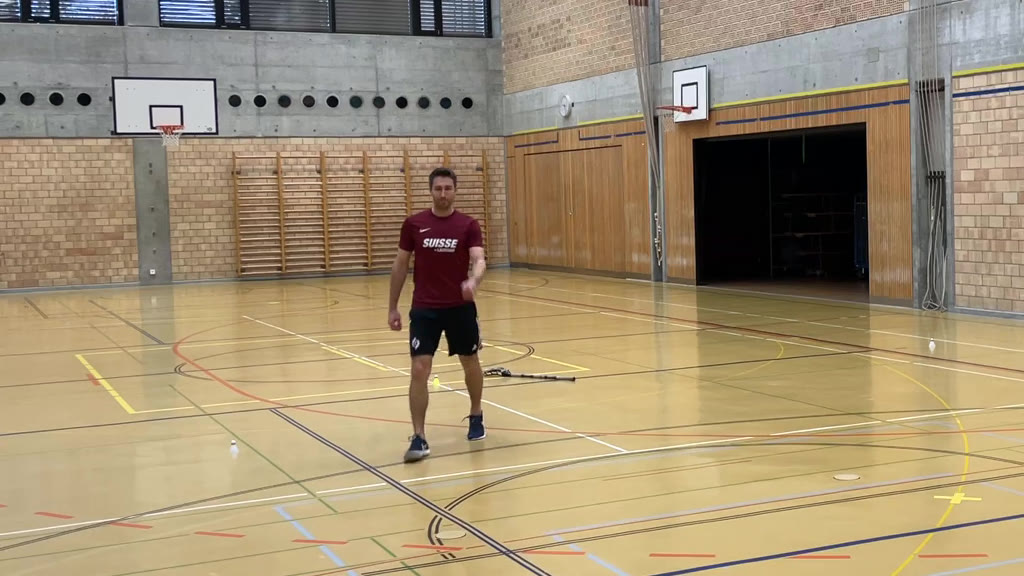}} \raisebox{-.5\height}{\includegraphics[width=0.09\textwidth]{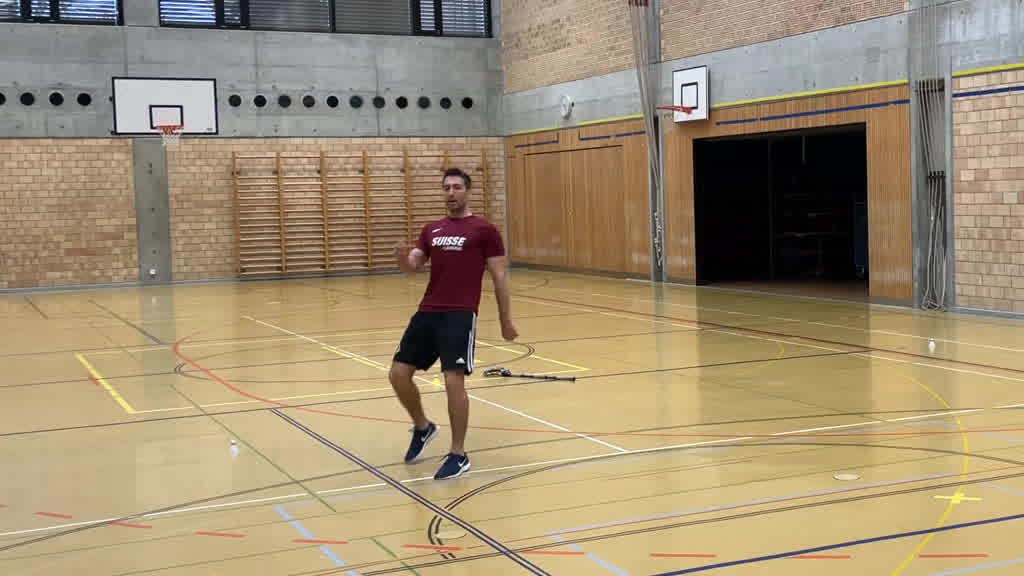}} \raisebox{-.5\height}{\includegraphics[width=0.09\textwidth]{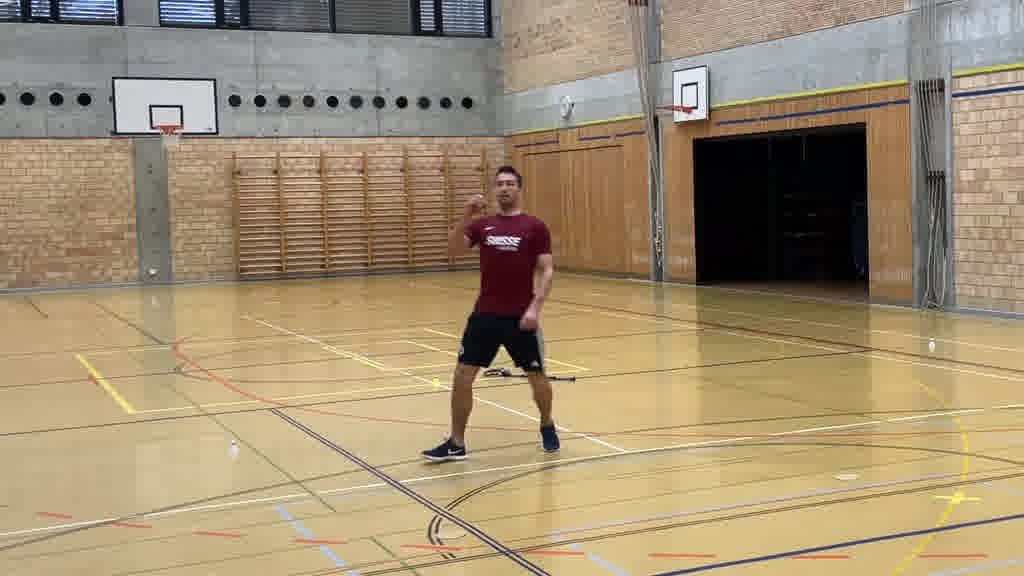}} & \raisebox{-.5\height}{\includegraphics[width=0.09\textwidth]{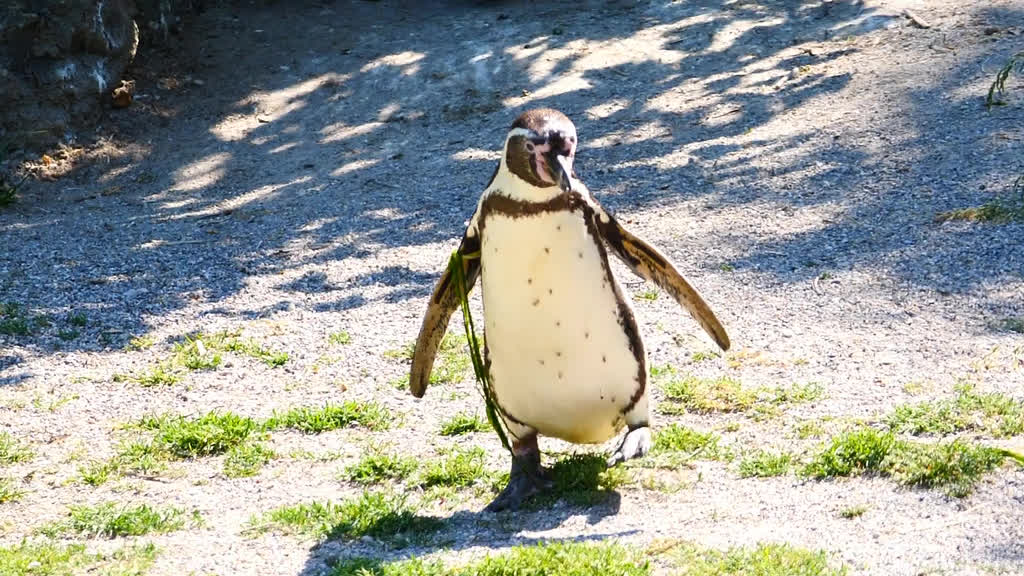}} & \raisebox{-.5\height}{\includegraphics[width=0.09\textwidth]{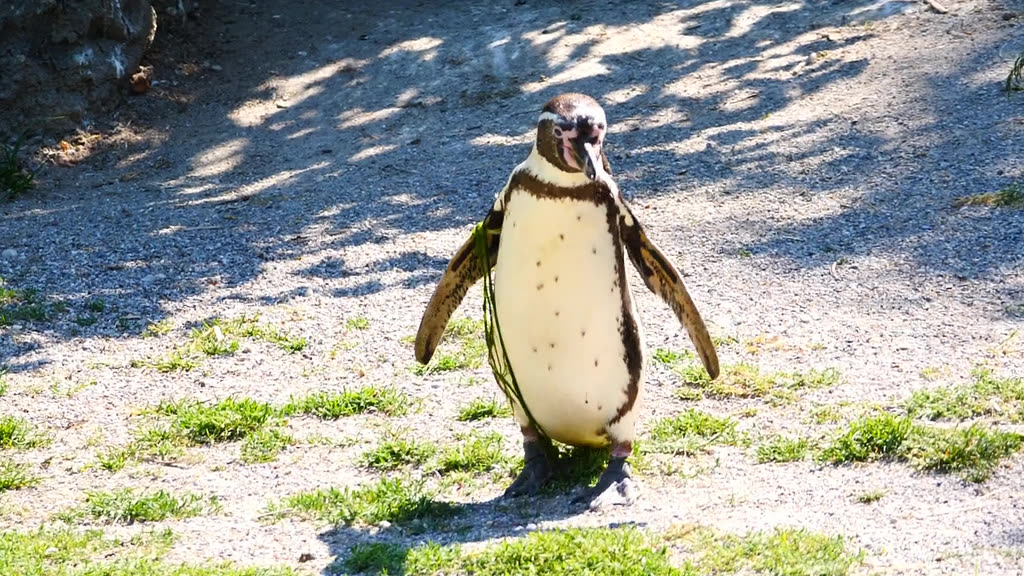}} \raisebox{-.5\height}{\includegraphics[width=0.09\textwidth]{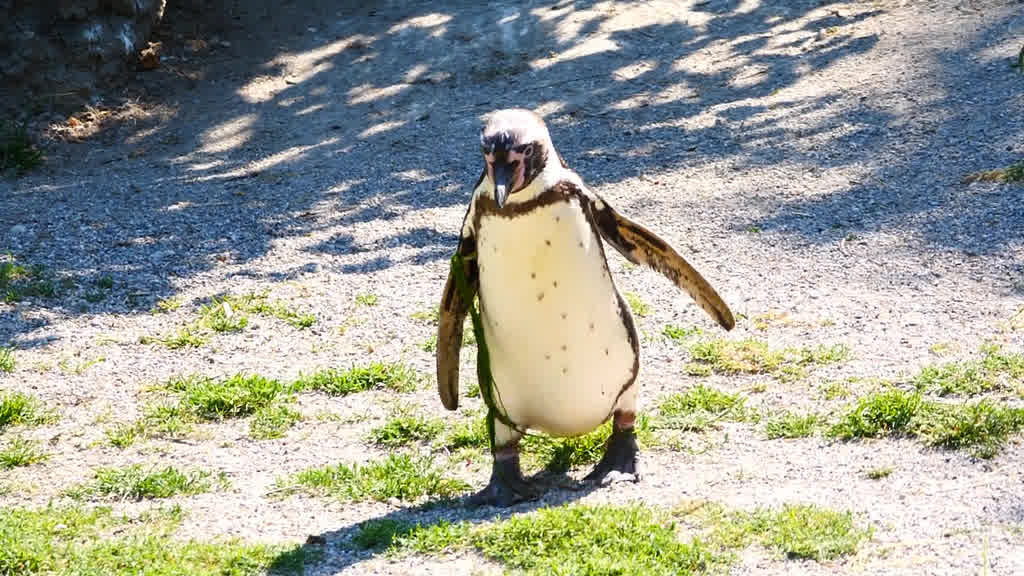}} \raisebox{-.5\height}{\includegraphics[width=0.09\textwidth]{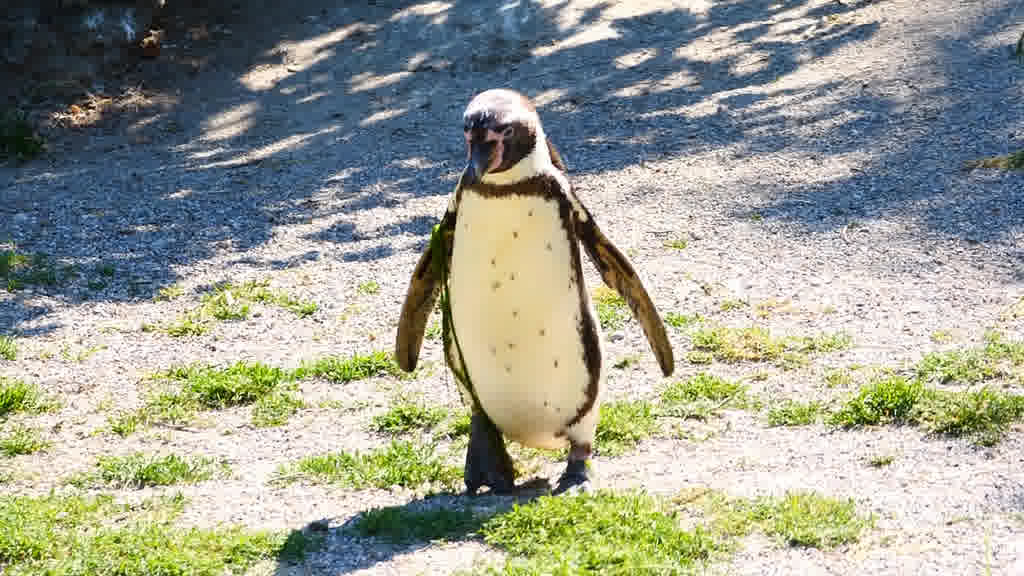}} \\
		{Gen.}  & \raisebox{-.5\height}{\includegraphics[width=0.09\textwidth]{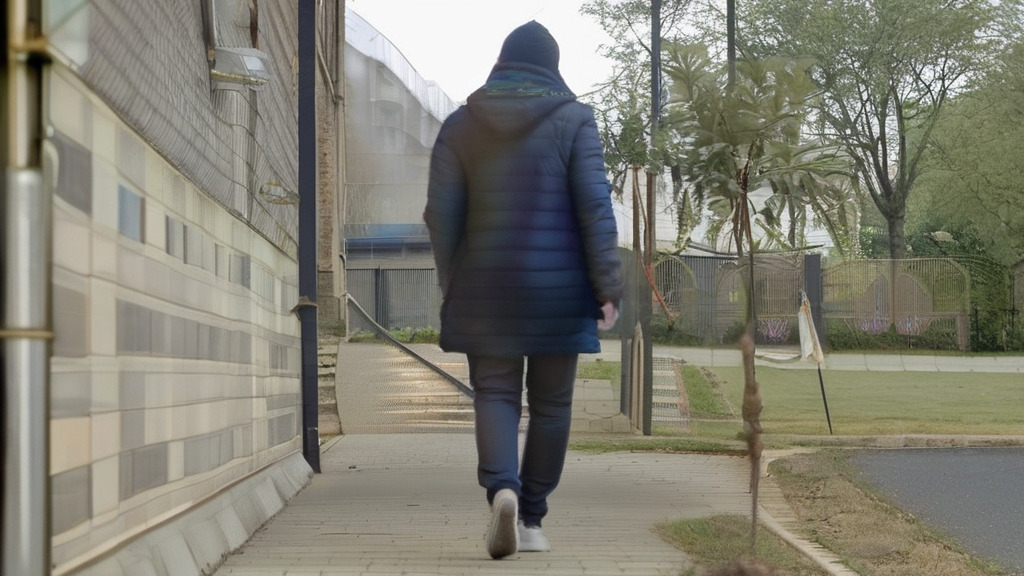}} & \raisebox{-.5\height}{\includegraphics[width=0.09\textwidth]{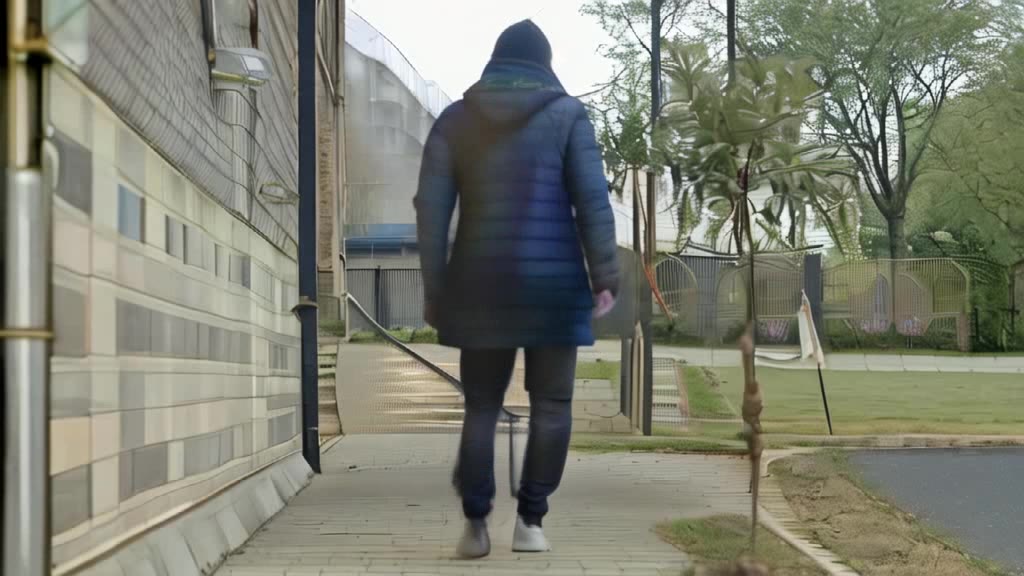}} \raisebox{-.5\height}{\includegraphics[width=0.09\textwidth]{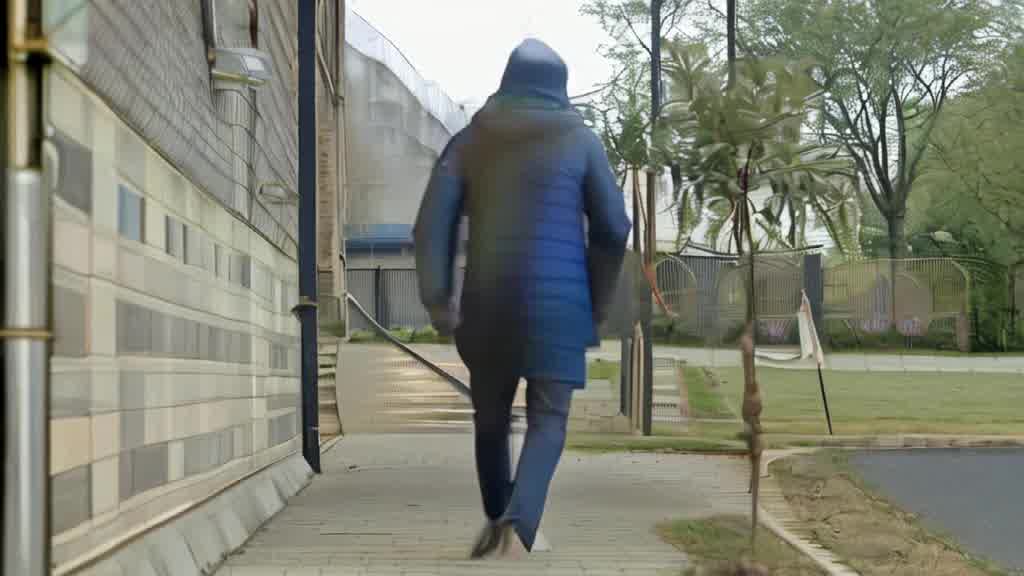}} \raisebox{-.5\height}{\includegraphics[width=0.09\textwidth]{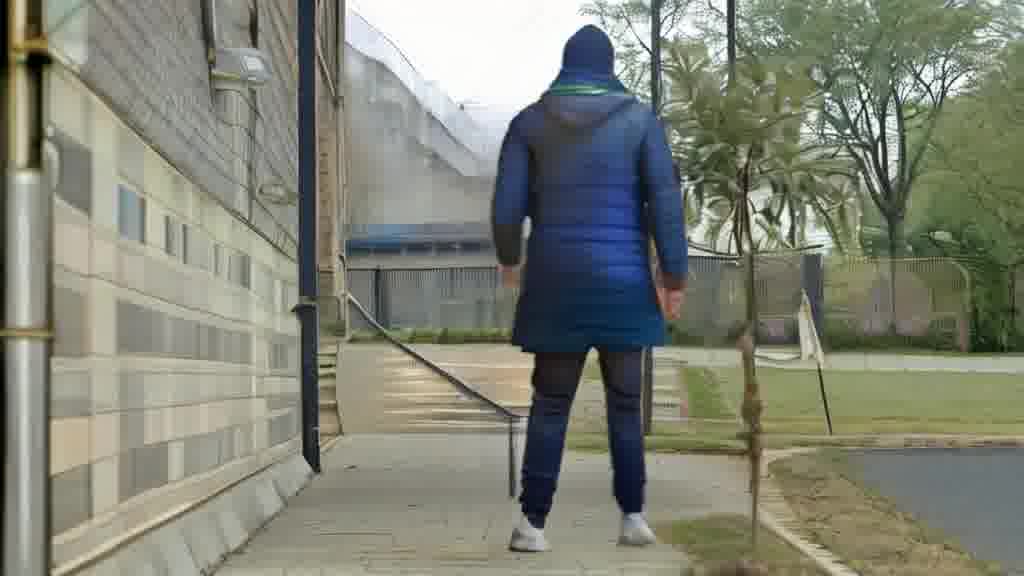}} & \raisebox{-.5\height}{\includegraphics[width=0.09\textwidth]{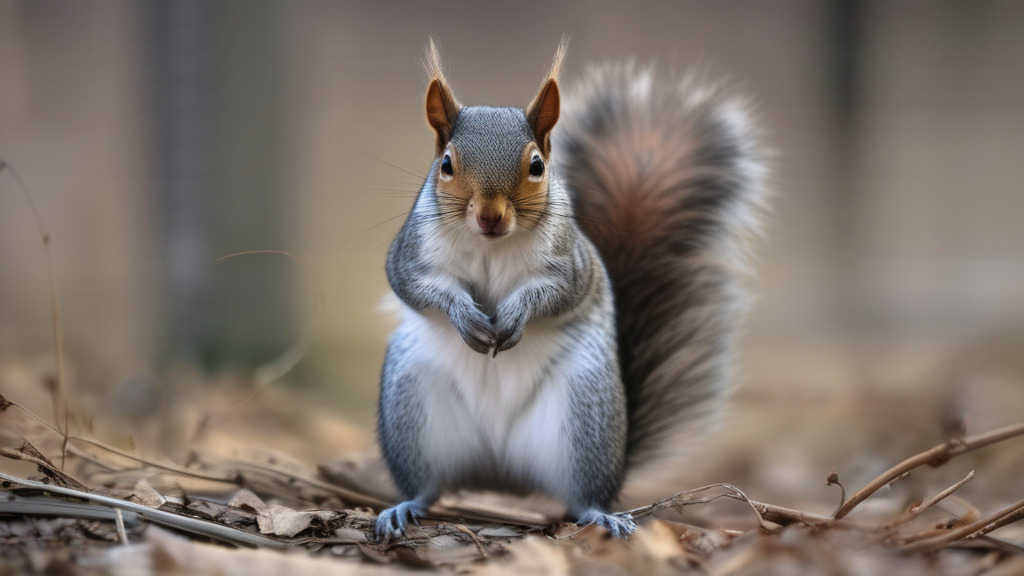}} & \raisebox{-.5\height}{\includegraphics[width=0.09\textwidth]{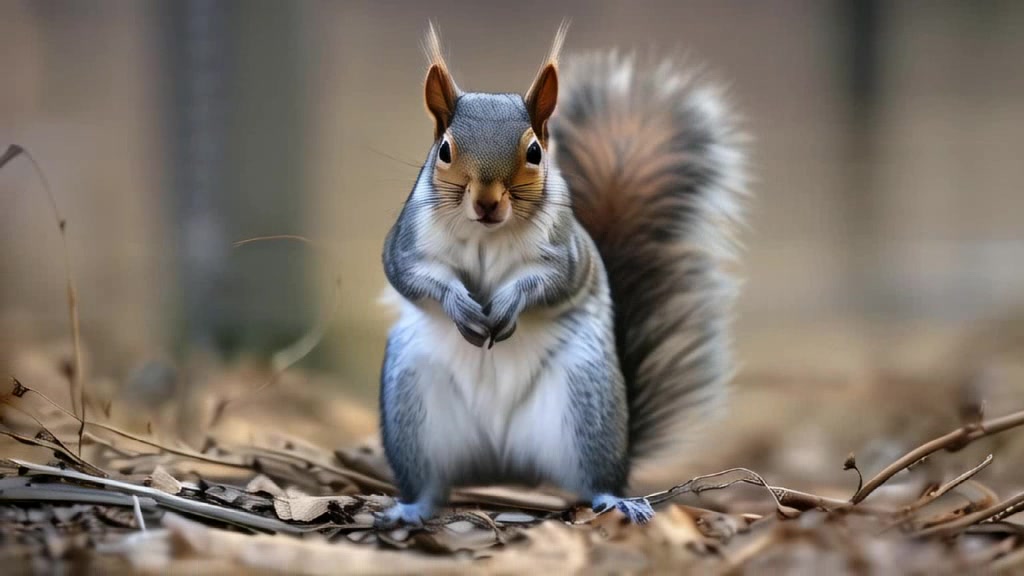}} \raisebox{-.5\height}{\includegraphics[width=0.09\textwidth]{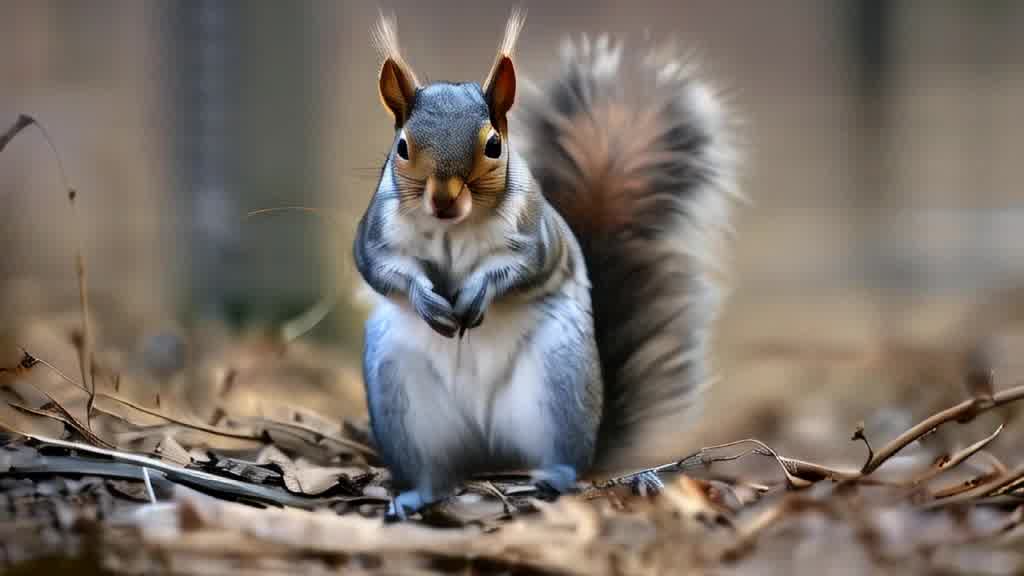}} \raisebox{-.5\height}{\includegraphics[width=0.09\textwidth]{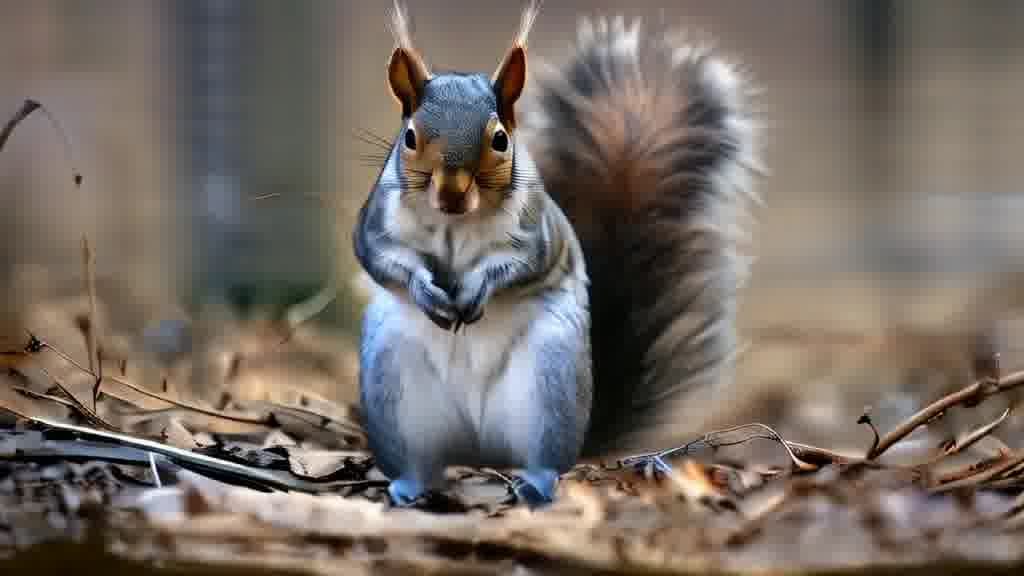}} \\
		{Ref.} & \raisebox{-.5\height}{\includegraphics[width=0.09\textwidth]{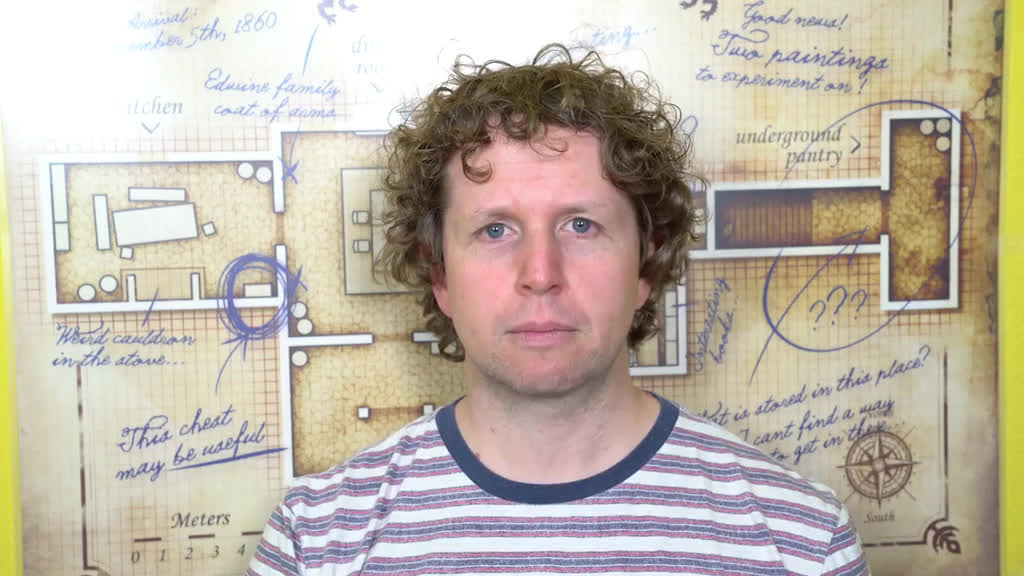}} & \raisebox{-.5\height}{\includegraphics[width=0.09\textwidth]{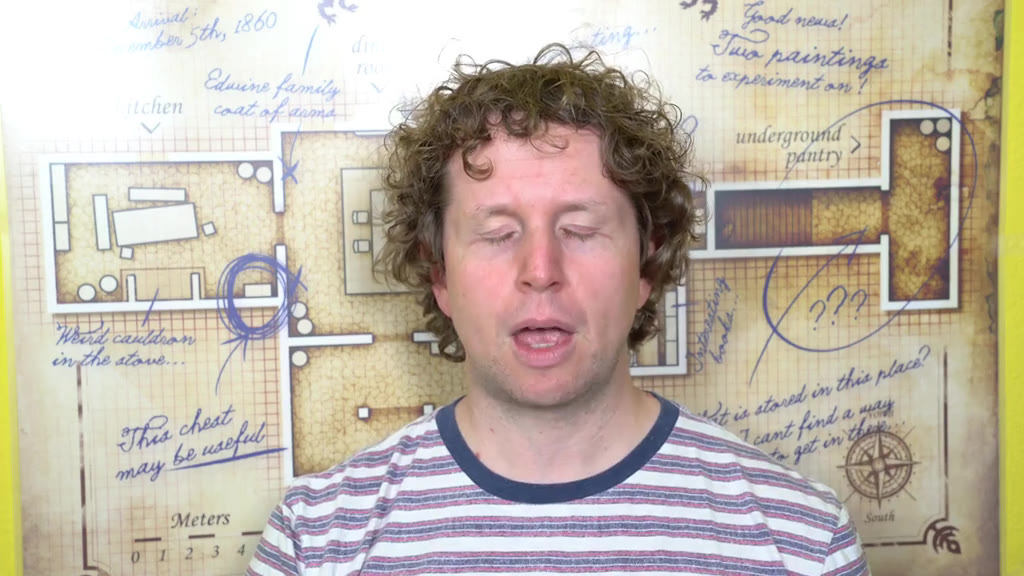}} \raisebox{-.5\height}{\includegraphics[width=0.09\textwidth]{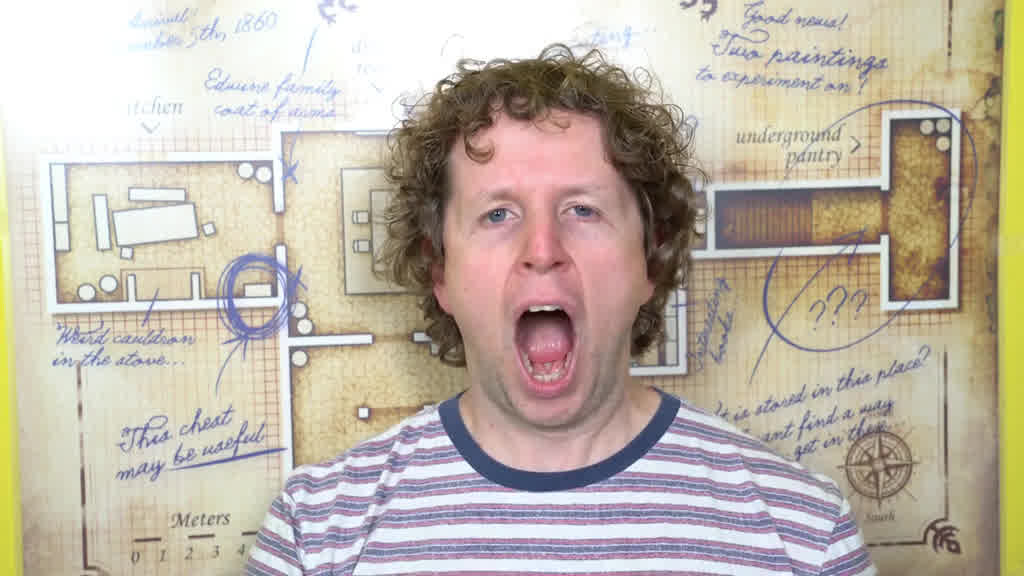}} \raisebox{-.5\height}{\includegraphics[width=0.09\textwidth]{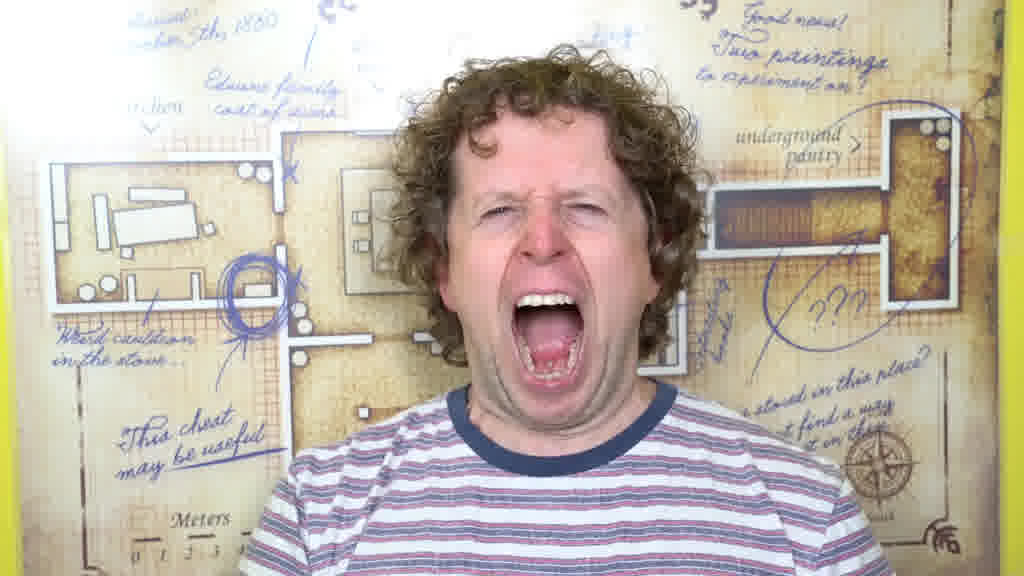}} & \raisebox{-.5\height}{\includegraphics[width=0.09\textwidth]{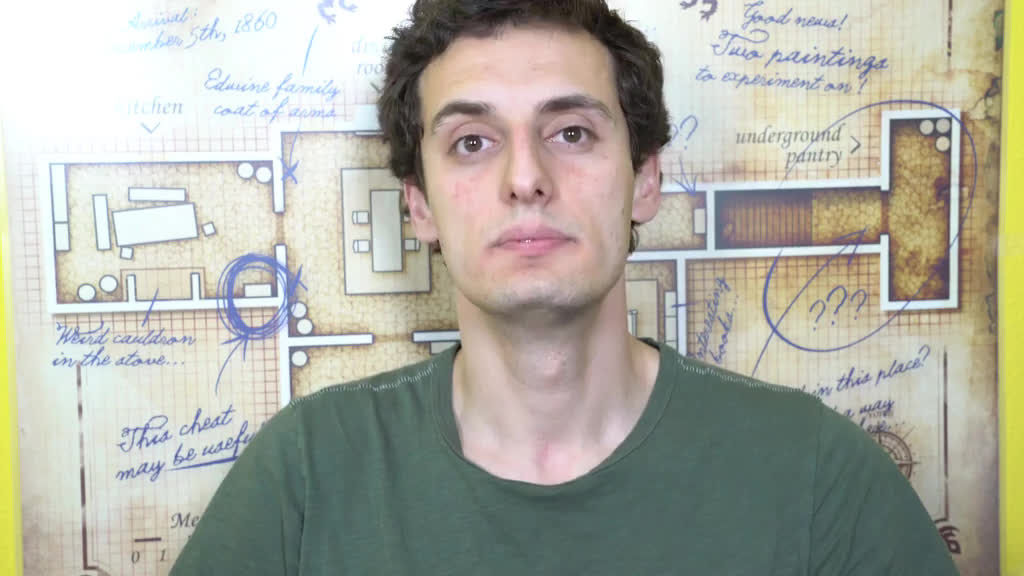}} & \raisebox{-.5\height}{\includegraphics[width=0.09\textwidth]{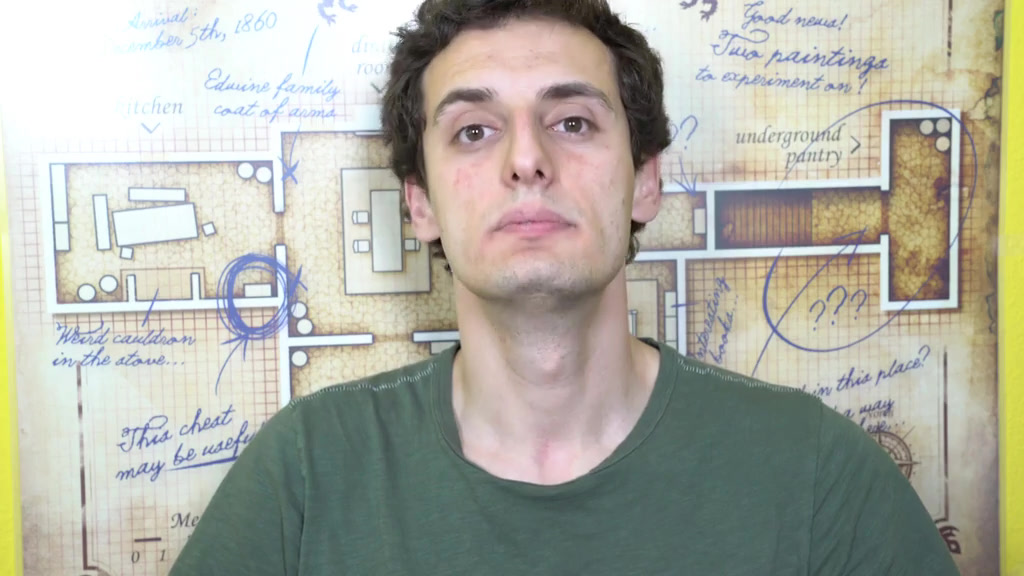}} \raisebox{-.5\height}{\includegraphics[width=0.09\textwidth]{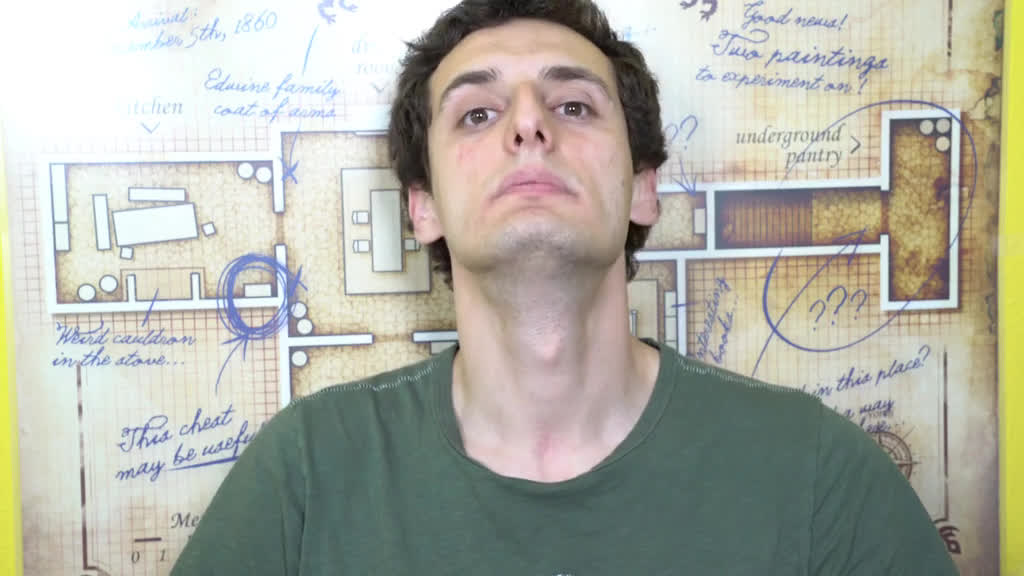}} \raisebox{-.5\height}{\includegraphics[width=0.09\textwidth]{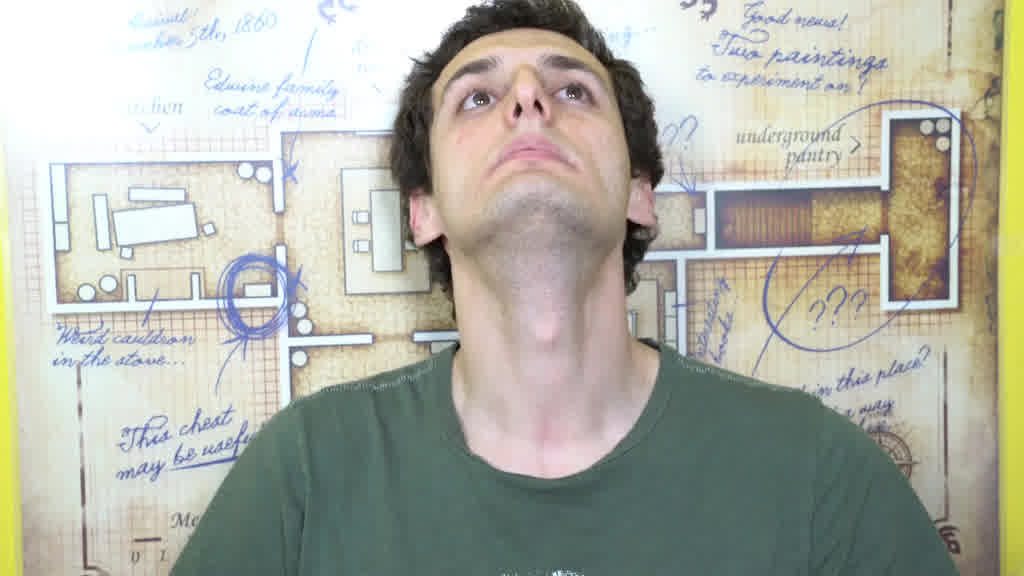}} \\
		{Gen.}  & \raisebox{-.5\height}{\includegraphics[width=0.09\textwidth]{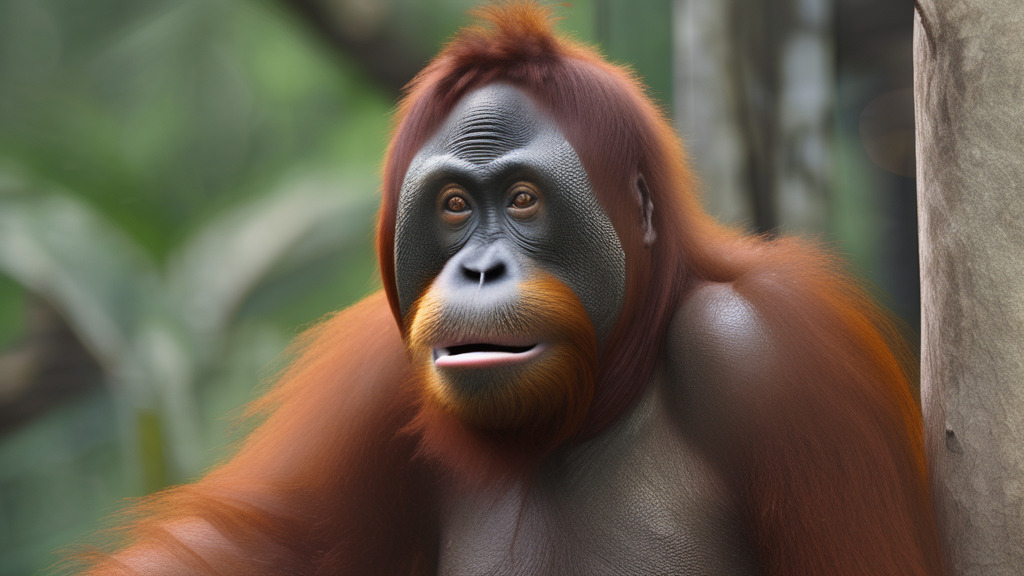}} & \raisebox{-.5\height}{\includegraphics[width=0.09\textwidth]{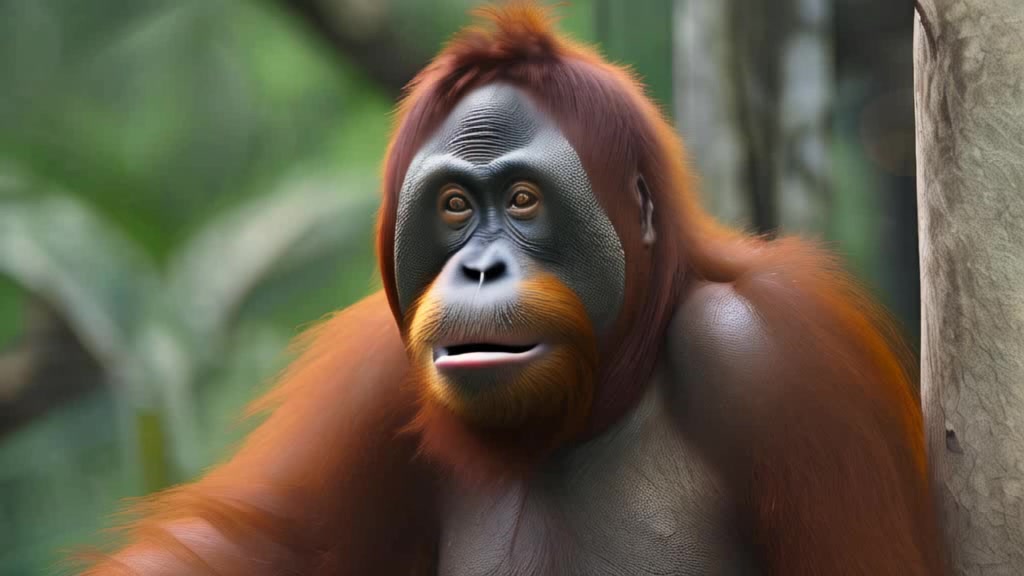}} \raisebox{-.5\height}{\includegraphics[width=0.09\textwidth]{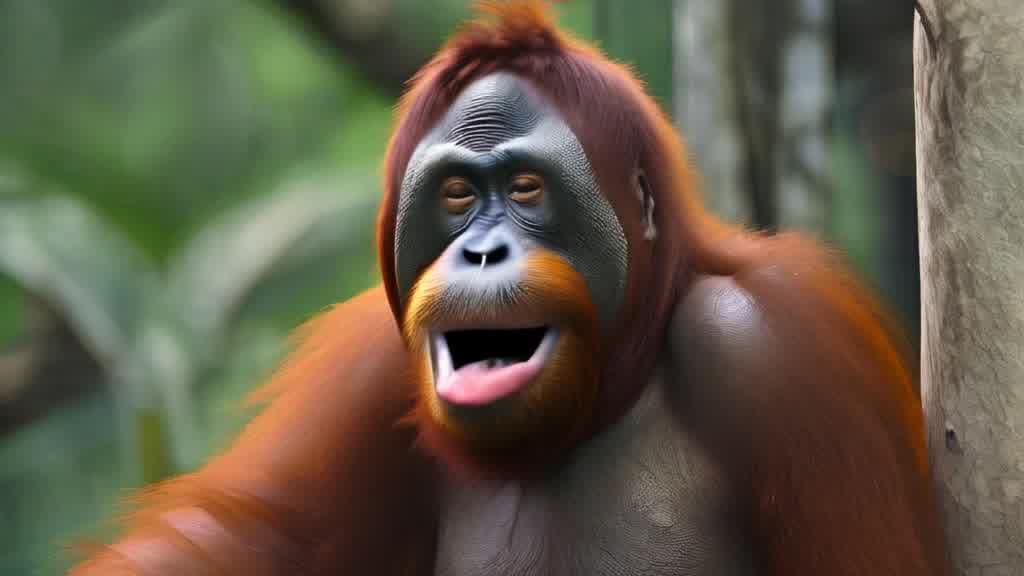}} \raisebox{-.5\height}{\includegraphics[width=0.09\textwidth]{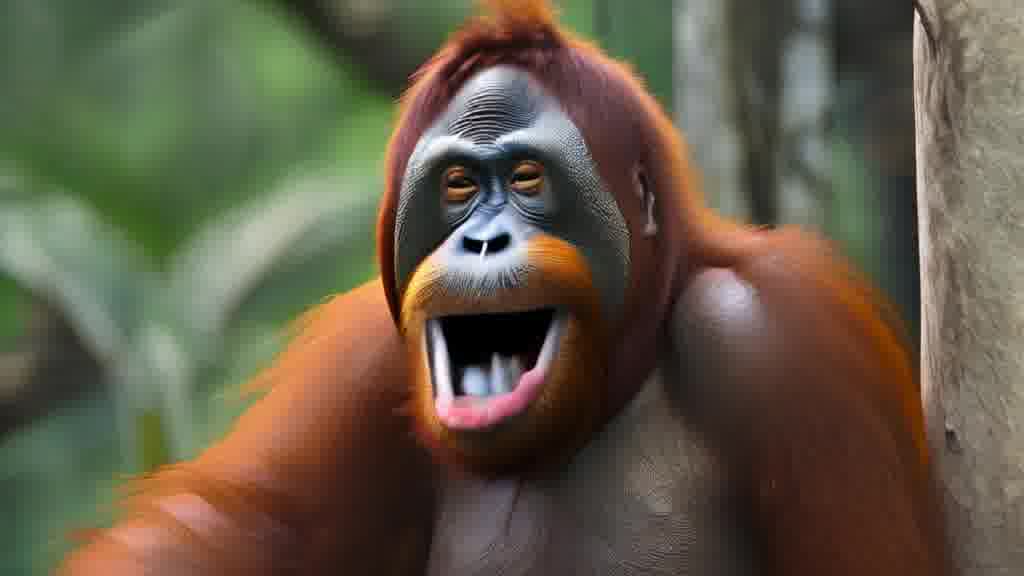}} & \raisebox{-.5\height}{\includegraphics[width=0.09\textwidth]{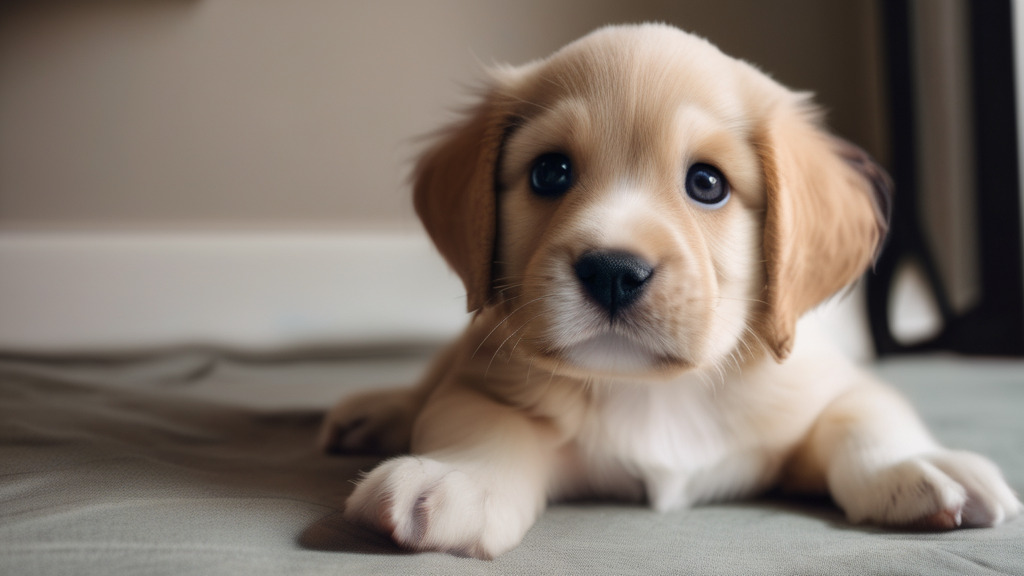}} & \raisebox{-.5\height}{\includegraphics[width=0.09\textwidth]{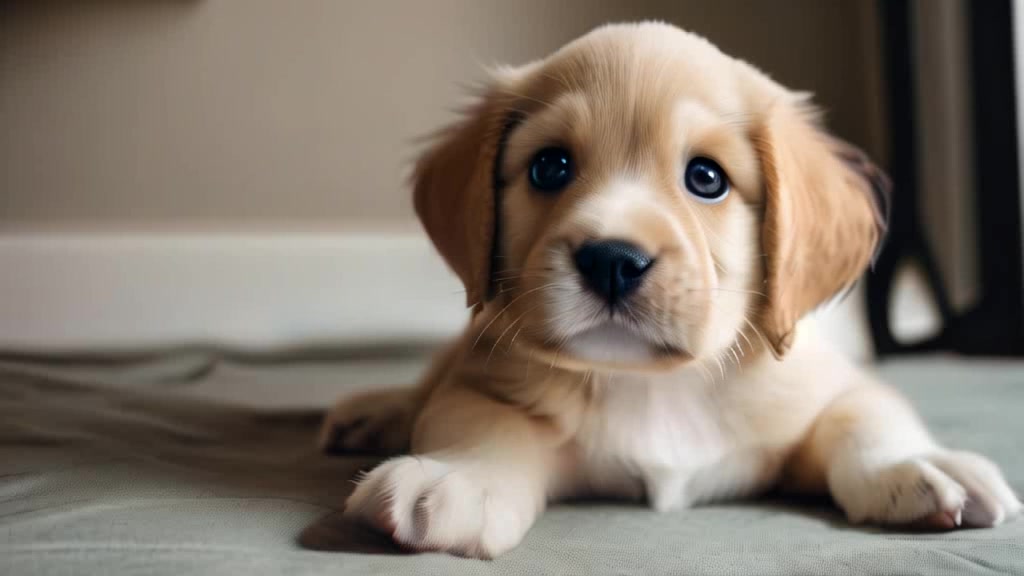}} \raisebox{-.5\height}{\includegraphics[width=0.09\textwidth]{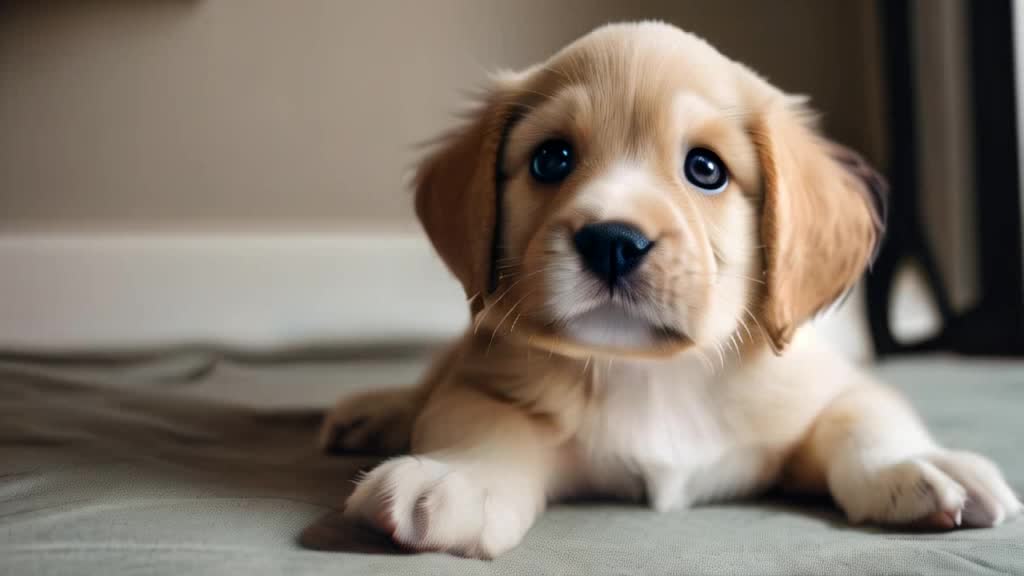}} \raisebox{-.5\height}{\includegraphics[width=0.09\textwidth]{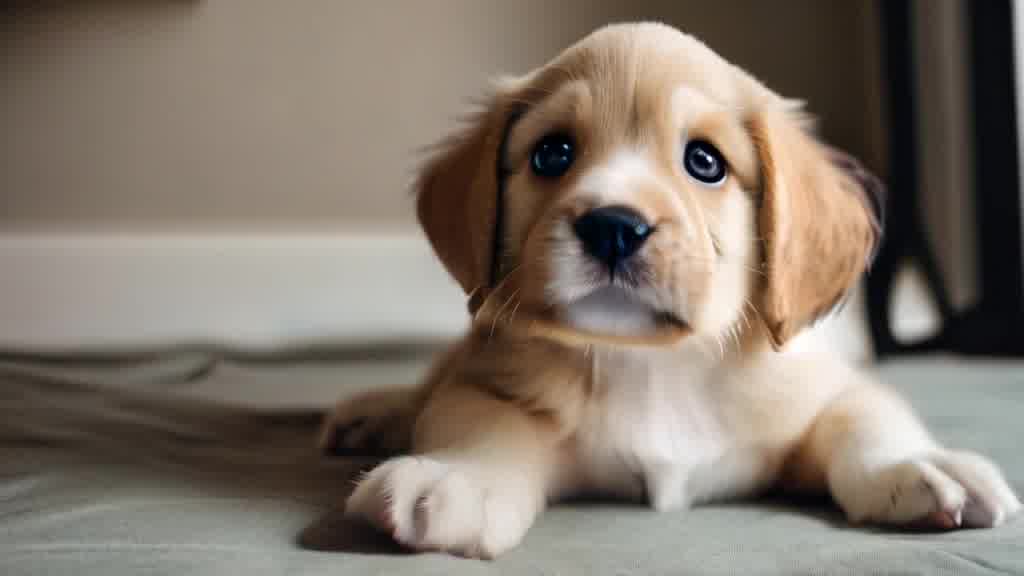}} \\
		{Ref.} & \raisebox{-.5\height}{\includegraphics[width=0.09\textwidth]{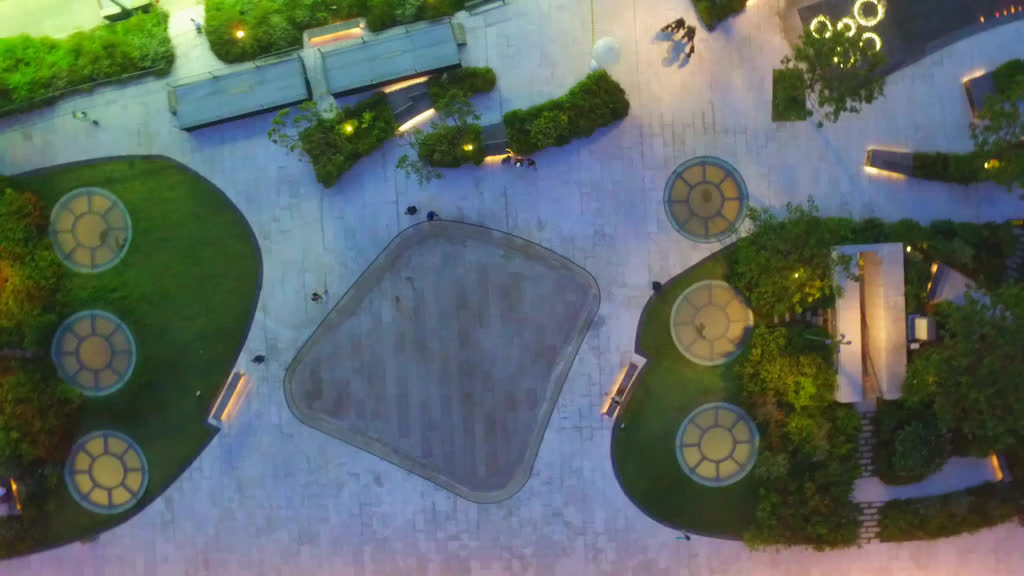}} & \raisebox{-.5\height}{\includegraphics[width=0.09\textwidth]{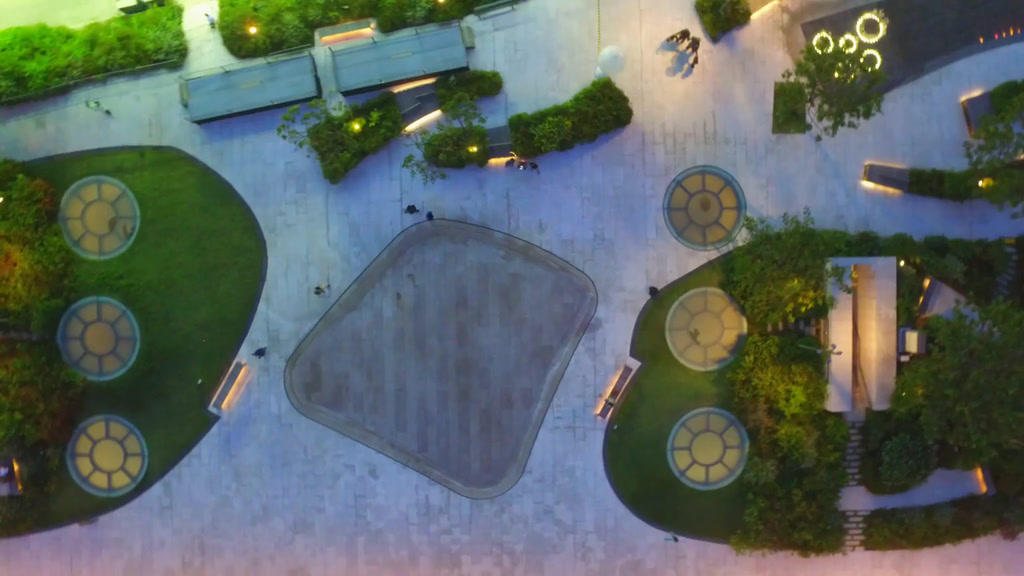}} \raisebox{-.5\height}{\includegraphics[width=0.09\textwidth]{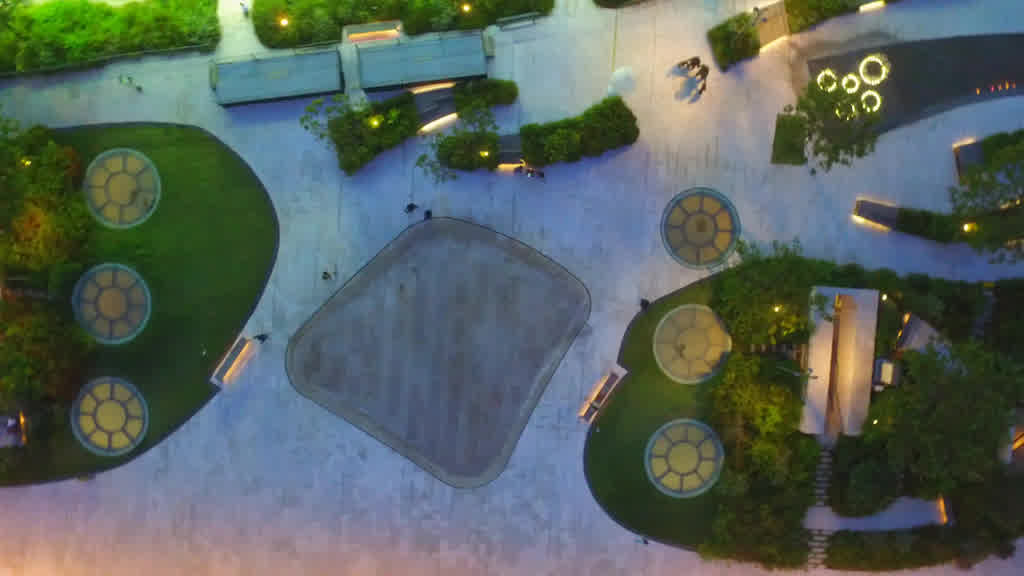}} \raisebox{-.5\height}{\includegraphics[width=0.09\textwidth]{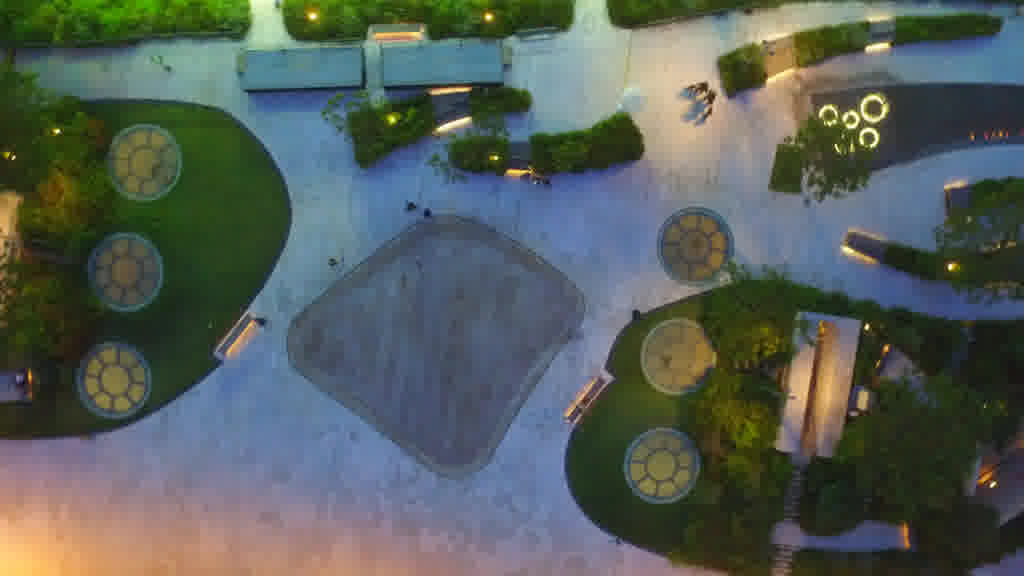}} & \raisebox{-.5\height}{\includegraphics[width=0.09\textwidth]{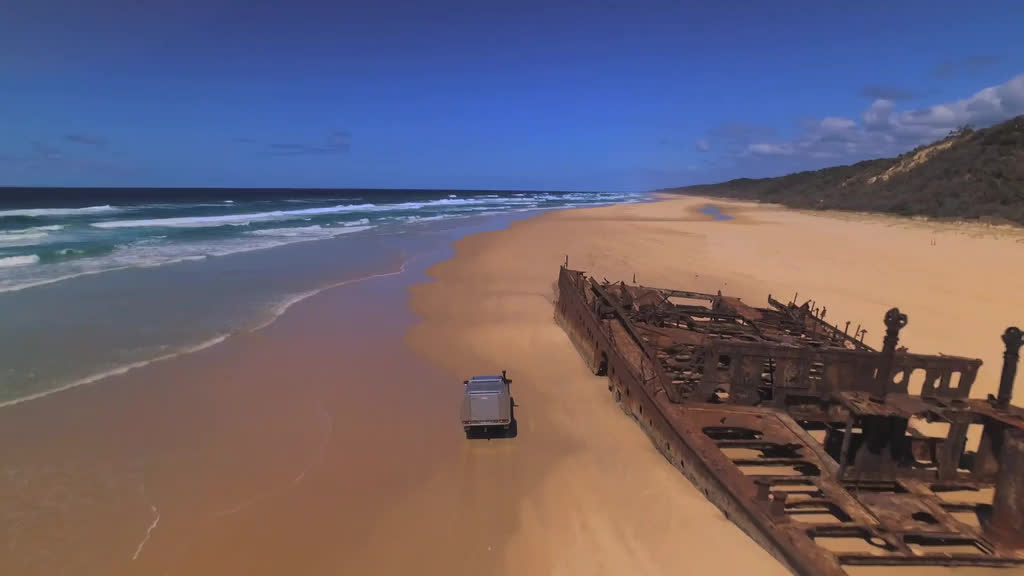}} & \raisebox{-.5\height}{\includegraphics[width=0.09\textwidth]{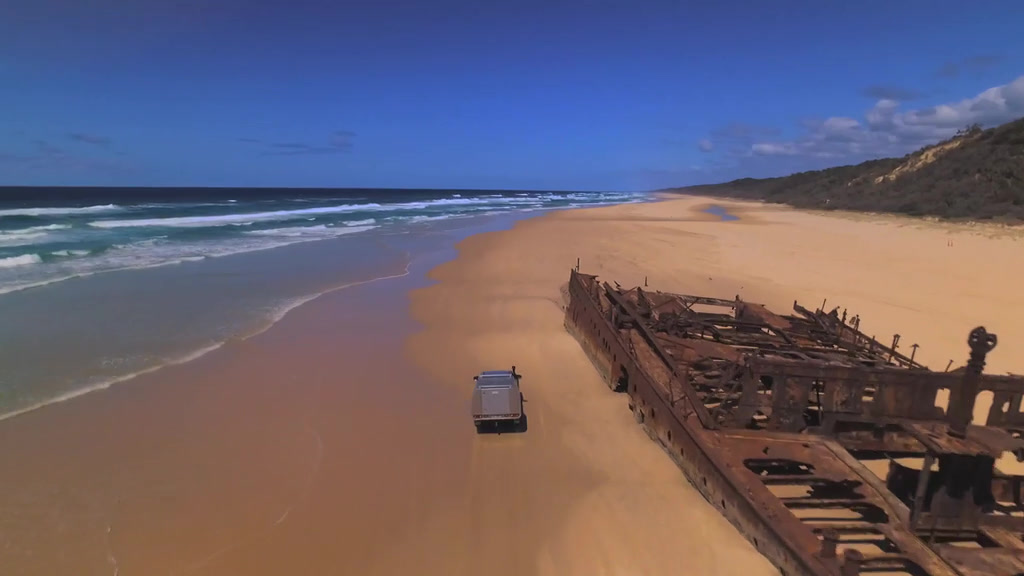}} \raisebox{-.5\height}{\includegraphics[width=0.09\textwidth]{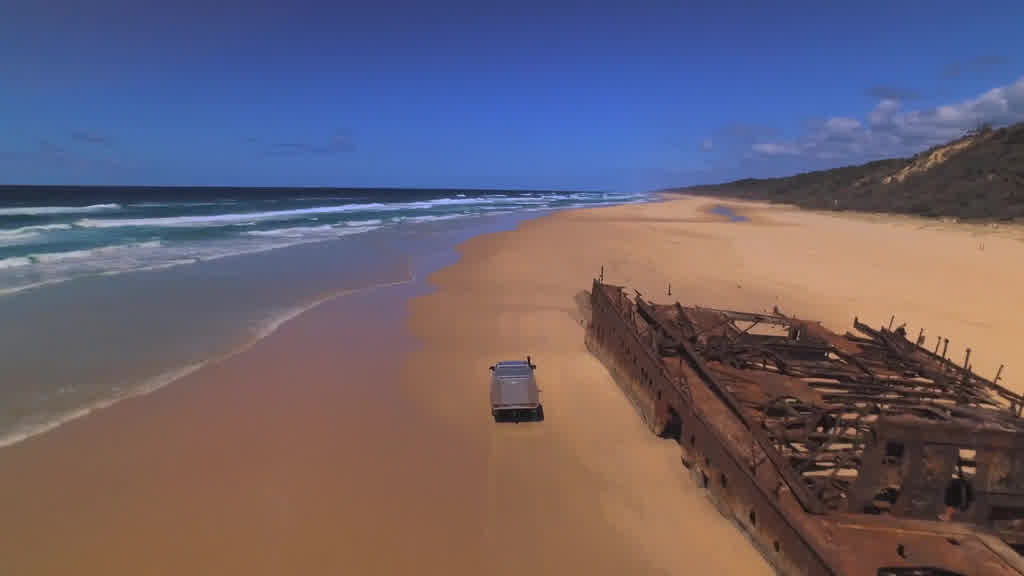}} \raisebox{-.5\height}{\includegraphics[width=0.09\textwidth]{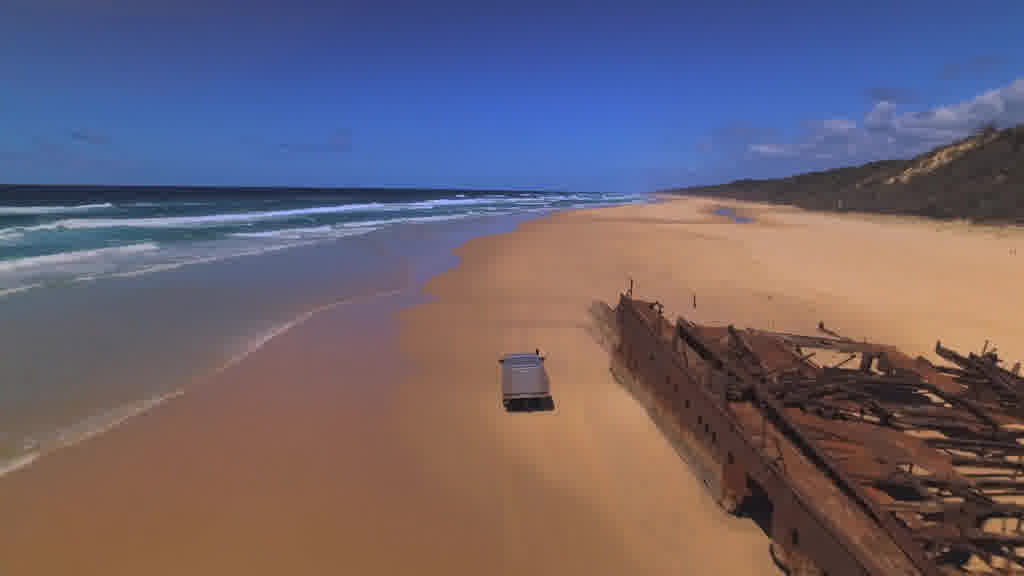}} \\
		{Gen.}  & \raisebox{-.5\height}{\includegraphics[width=0.09\textwidth]{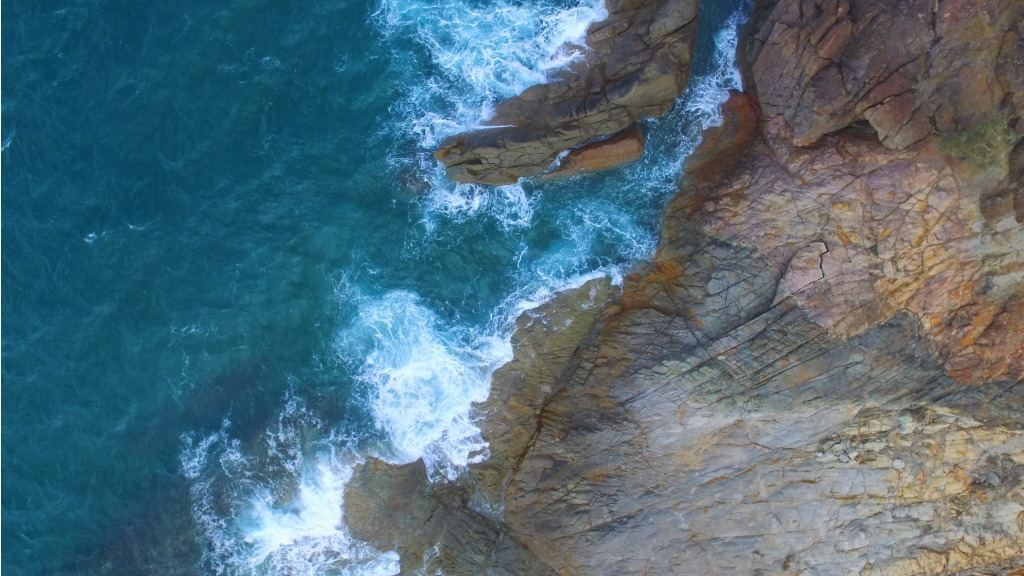}} & \raisebox{-.5\height}{\includegraphics[width=0.09\textwidth]{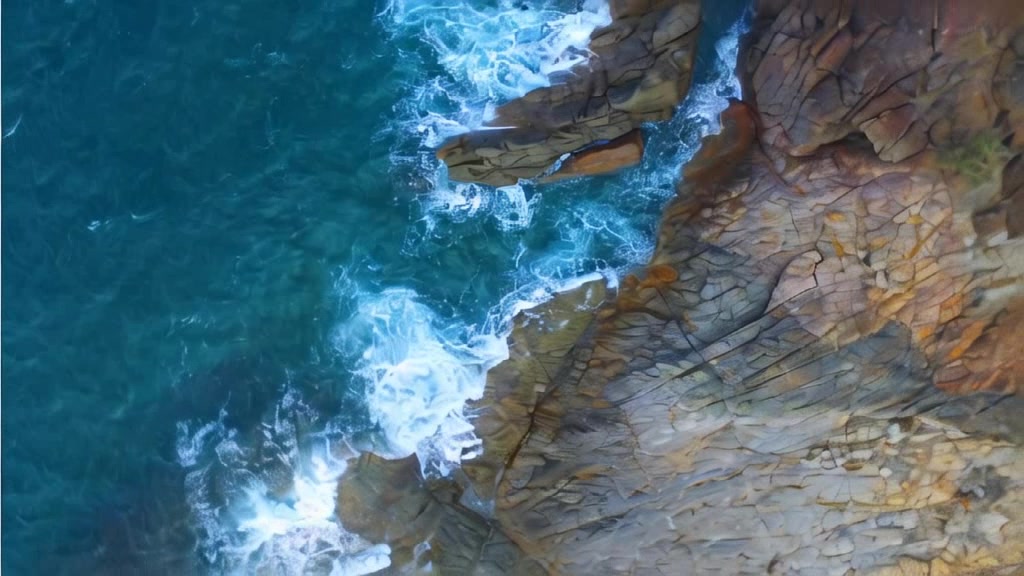}} \raisebox{-.5\height}{\includegraphics[width=0.09\textwidth]{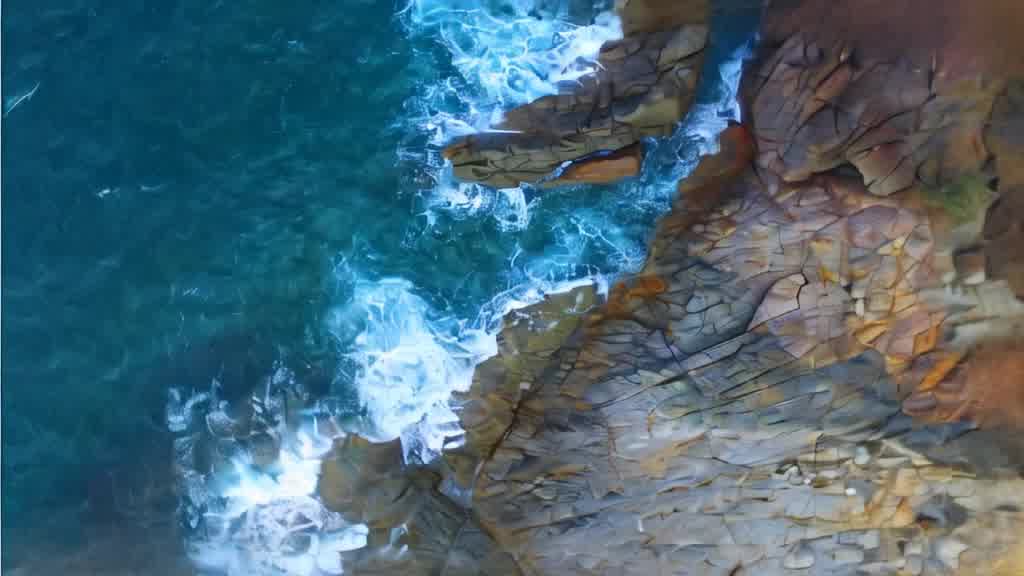}} \raisebox{-.5\height}{\includegraphics[width=0.09\textwidth]{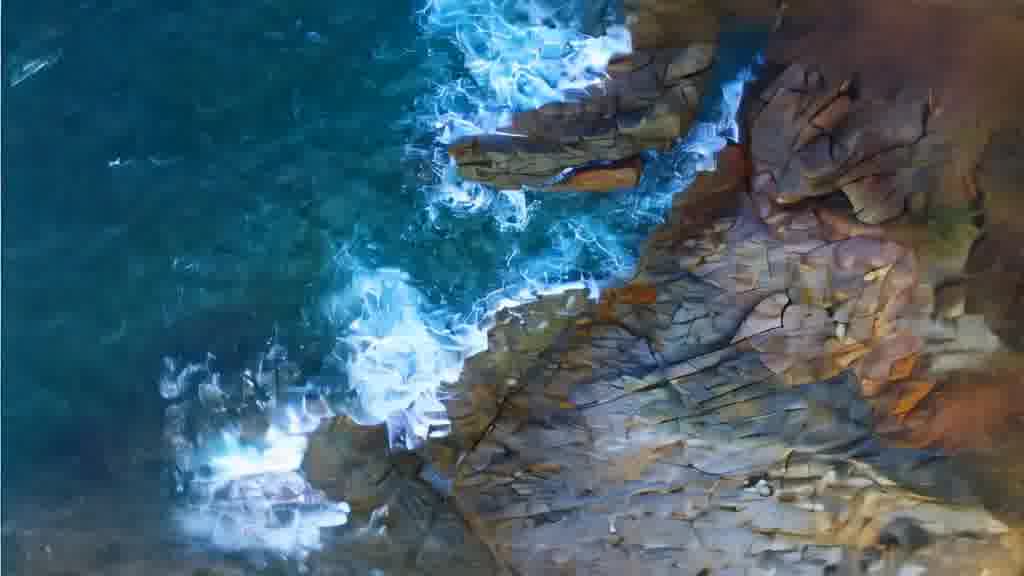}} & \raisebox{-.5\height}{\includegraphics[width=0.09\textwidth]{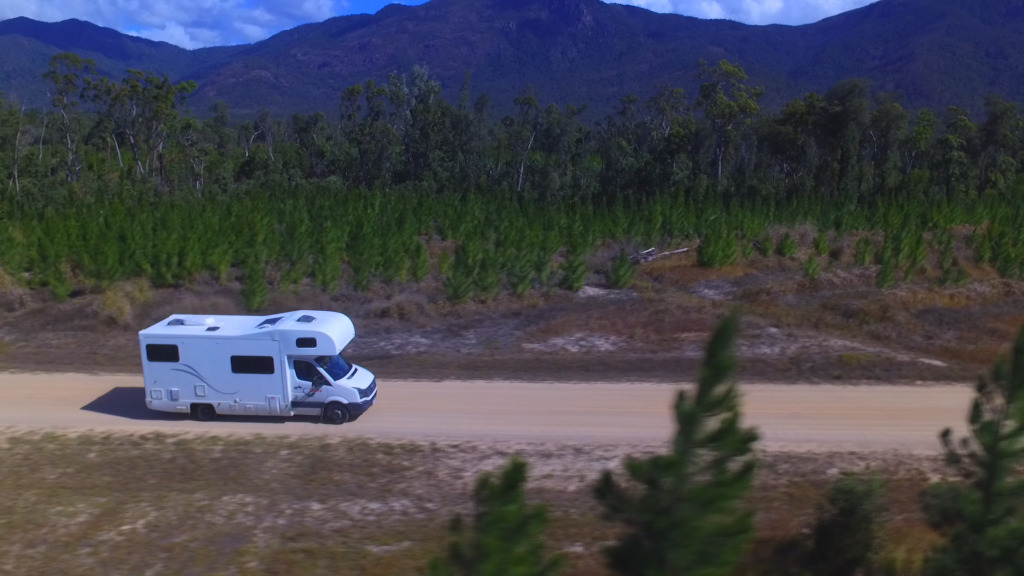}} & \raisebox{-.5\height}{\includegraphics[width=0.09\textwidth]{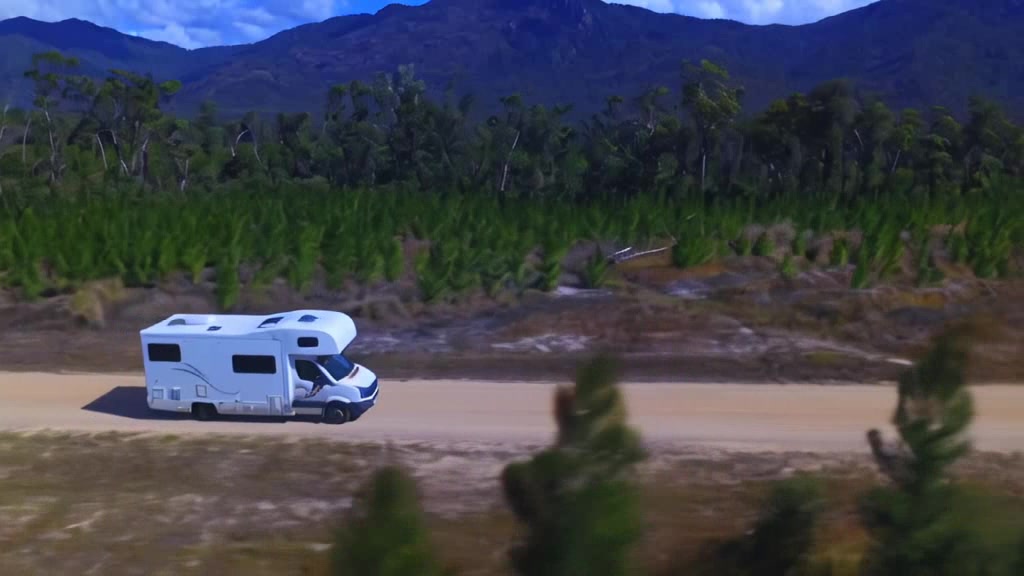}} \raisebox{-.5\height}{\includegraphics[width=0.09\textwidth]{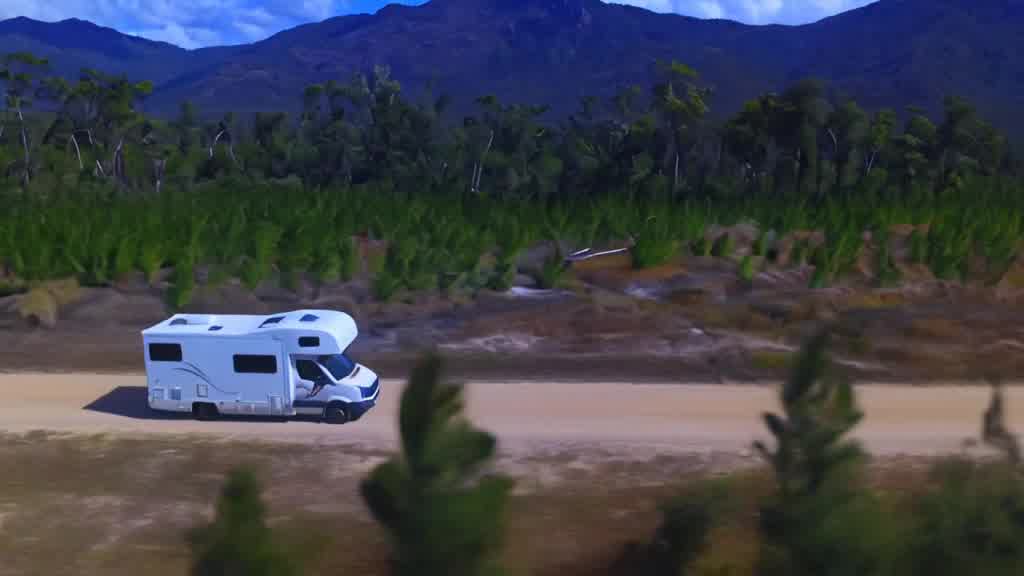}} \raisebox{-.5\height}{\includegraphics[width=0.09\textwidth]{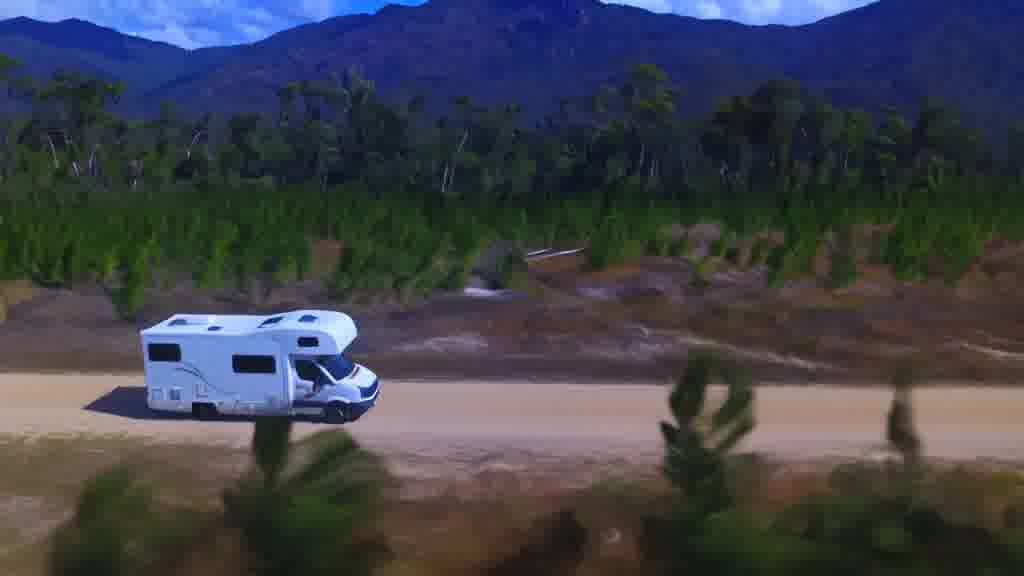}} \\
		{Ref.} & \raisebox{-.5\height}{\includegraphics[width=0.09\textwidth]{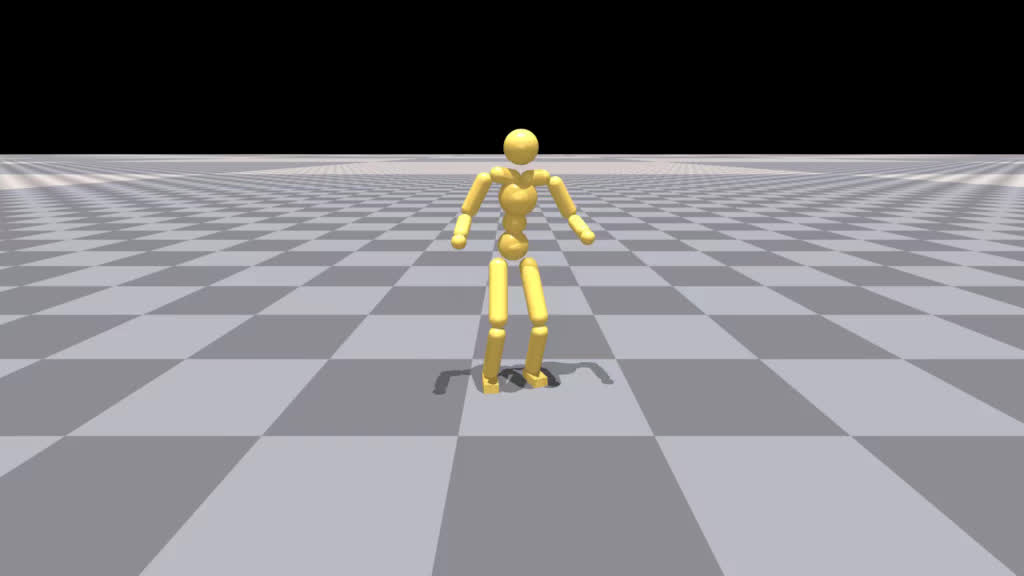}} & \raisebox{-.5\height}{\includegraphics[width=0.09\textwidth]{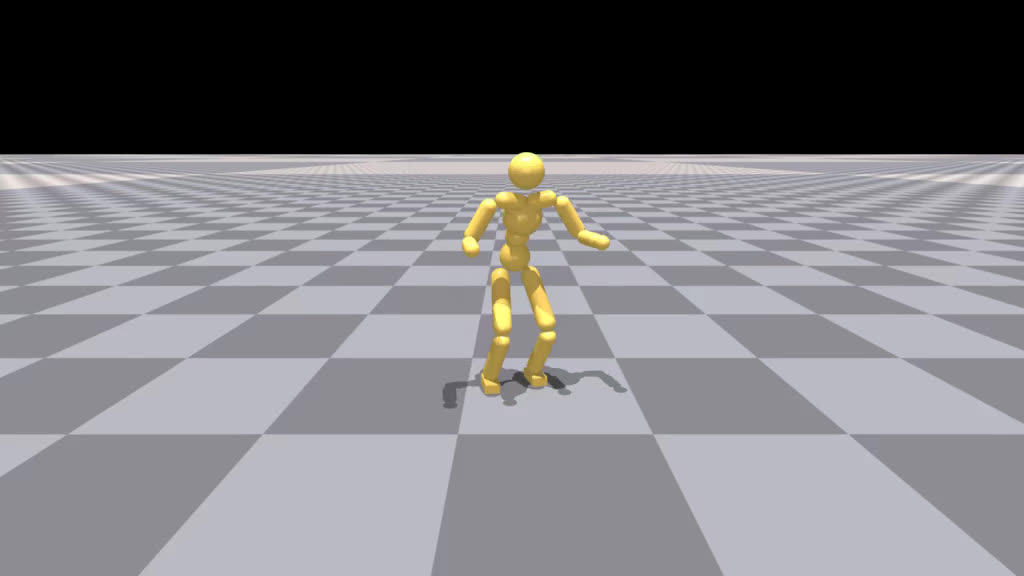}} \raisebox{-.5\height}{\includegraphics[width=0.09\textwidth]{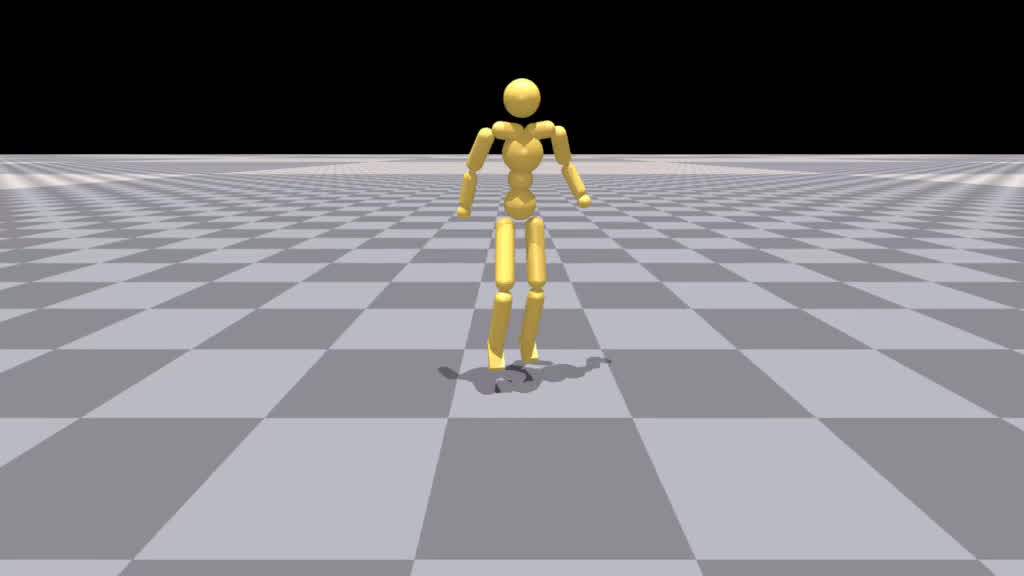}} \raisebox{-.5\height}{\includegraphics[width=0.09\textwidth]{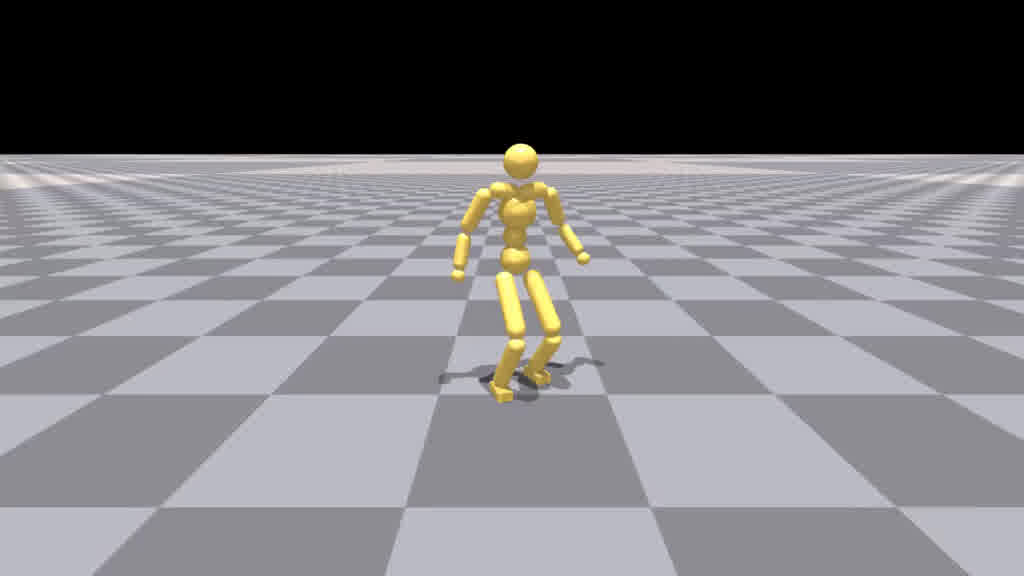}} & \raisebox{-.5\height}{\includegraphics[width=0.09\textwidth]{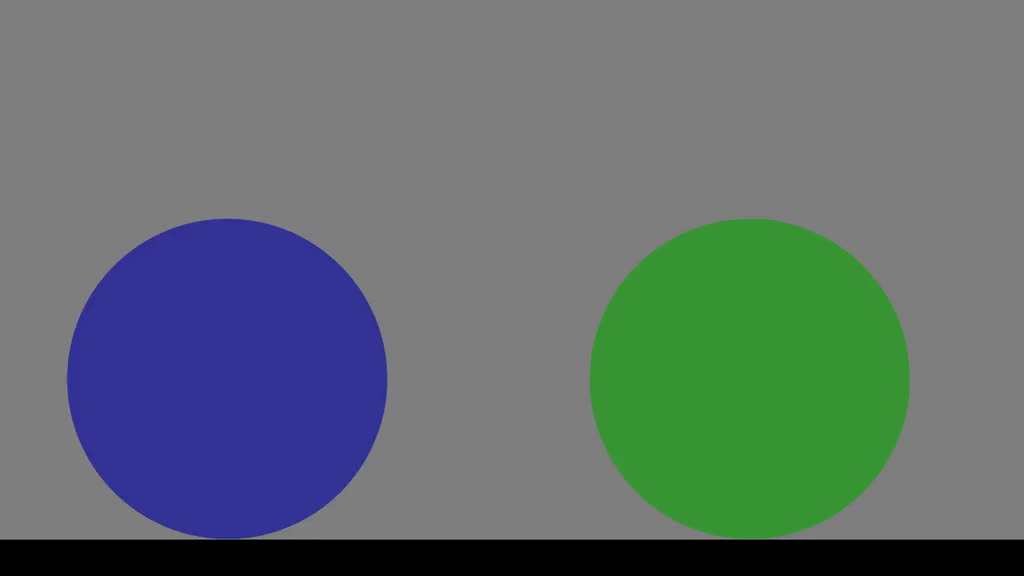}} & \raisebox{-.5\height}{\includegraphics[width=0.09\textwidth]{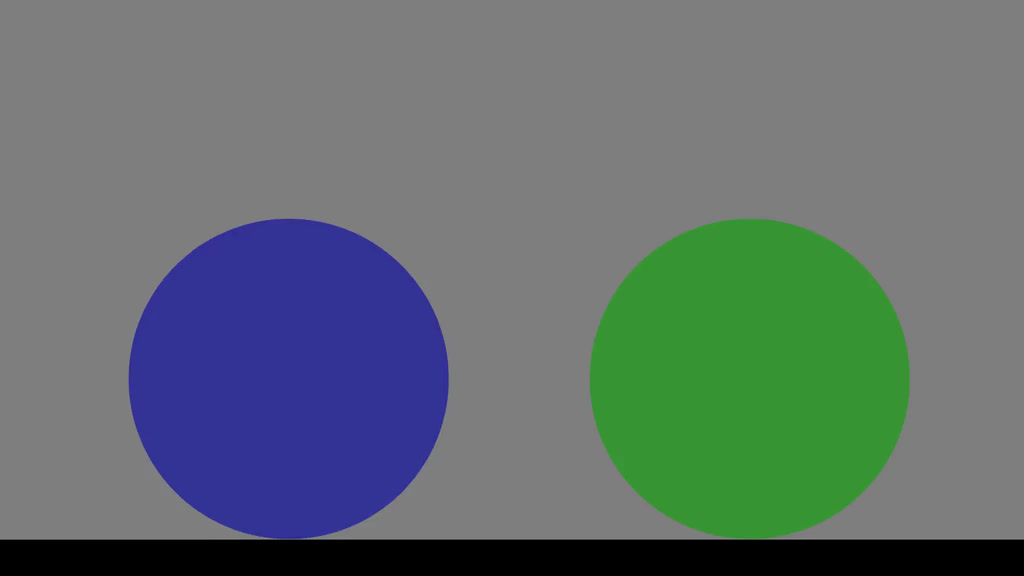}} \raisebox{-.5\height}{\includegraphics[width=0.09\textwidth]{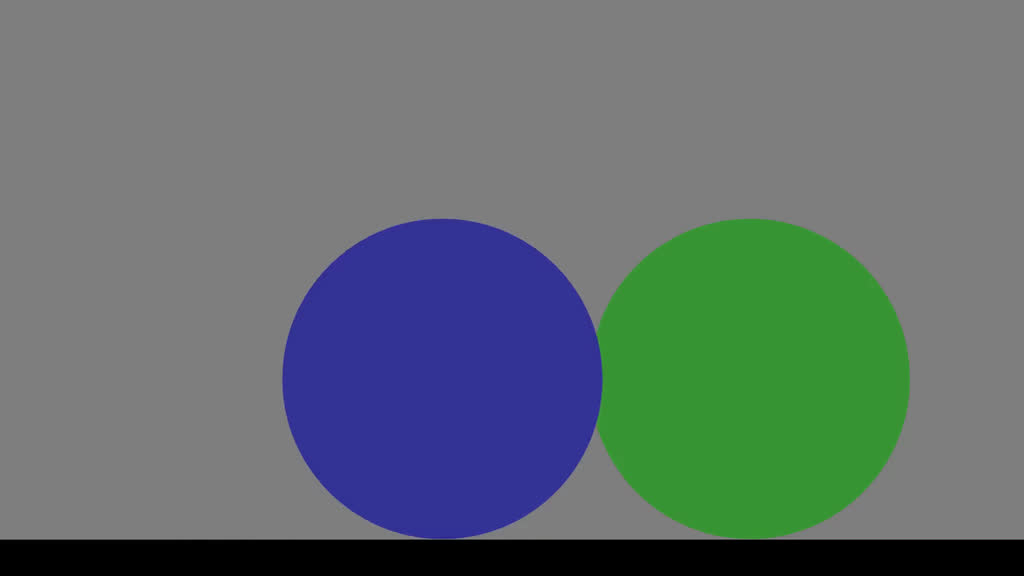}} \raisebox{-.5\height}{\includegraphics[width=0.09\textwidth]{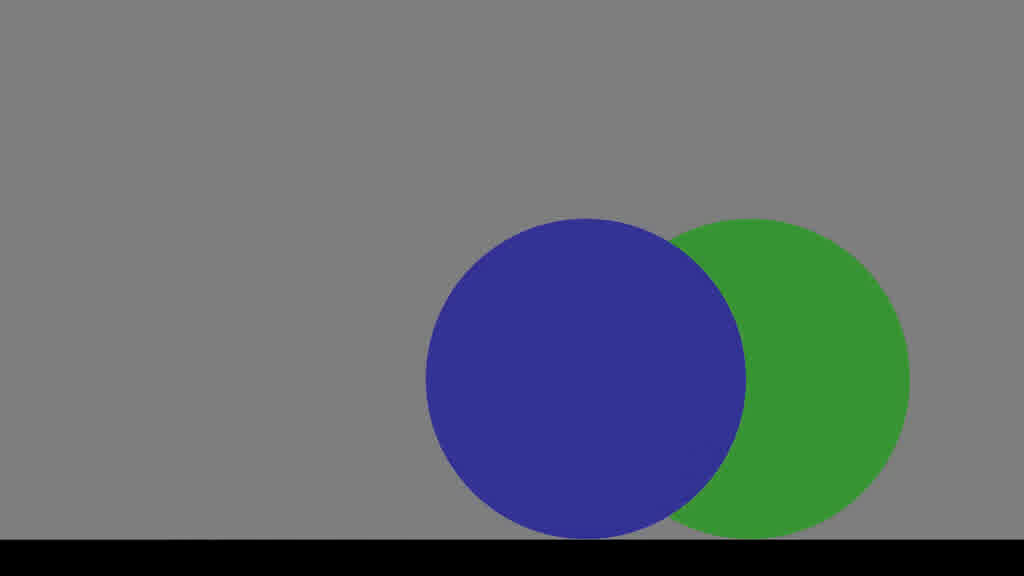}} \\
		{Gen.}  & \raisebox{-.5\height}{\includegraphics[width=0.09\textwidth]{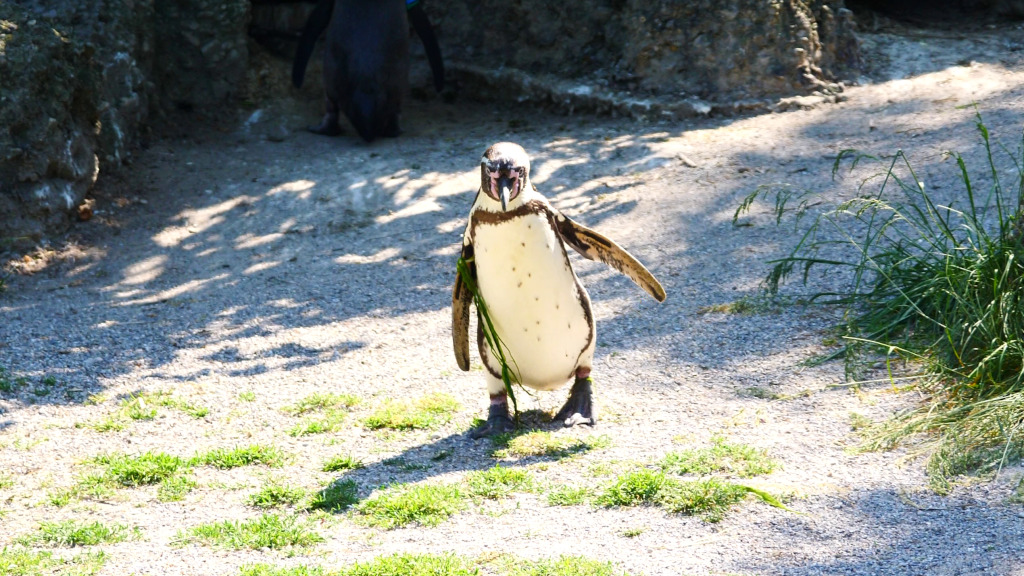}} & \raisebox{-.5\height}{\includegraphics[width=0.09\textwidth]{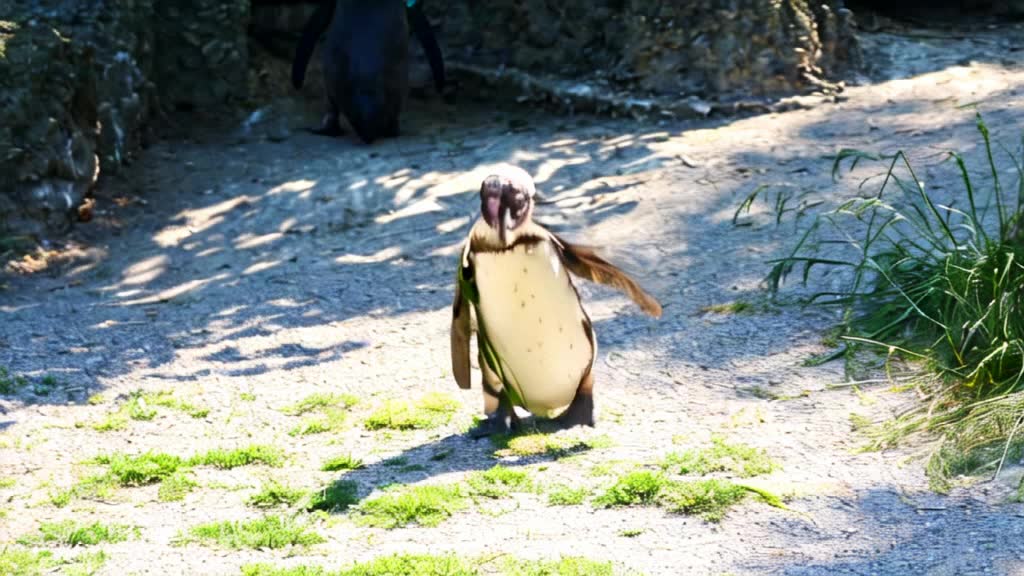}} \raisebox{-.5\height}{\includegraphics[width=0.09\textwidth]{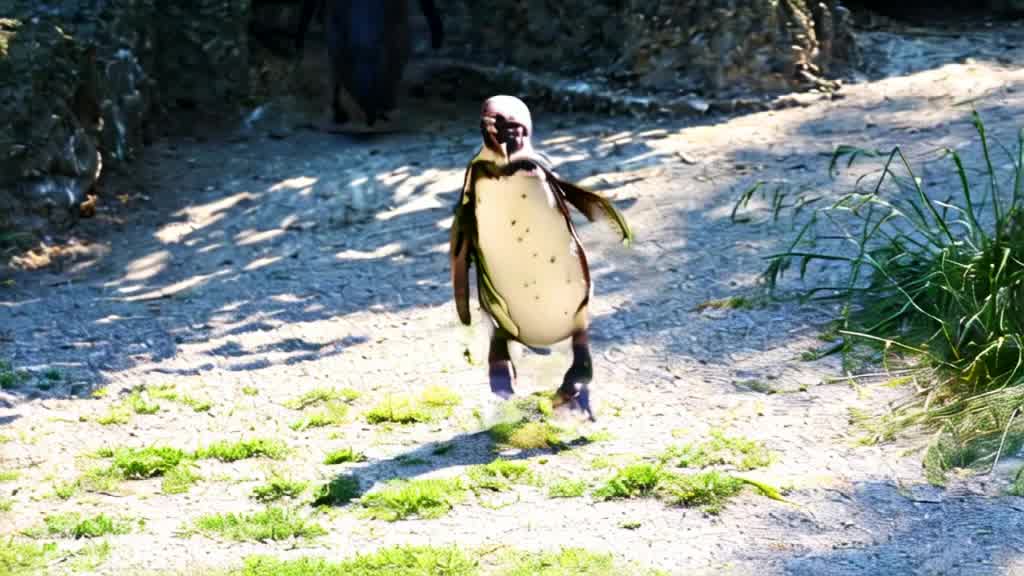}} \raisebox{-.5\height}{\includegraphics[width=0.09\textwidth]{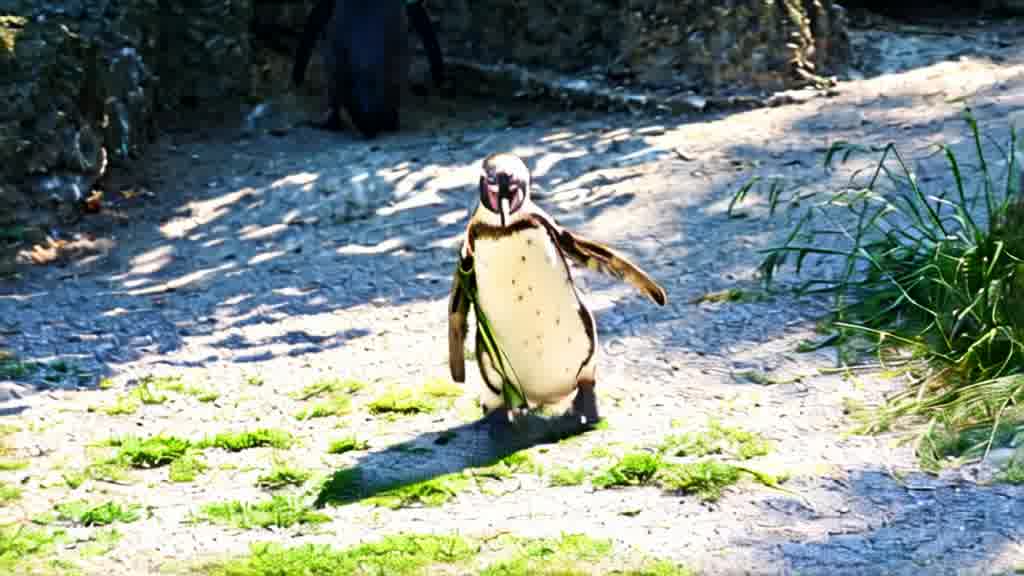}} & \raisebox{-.5\height}{\includegraphics[width=0.09\textwidth]{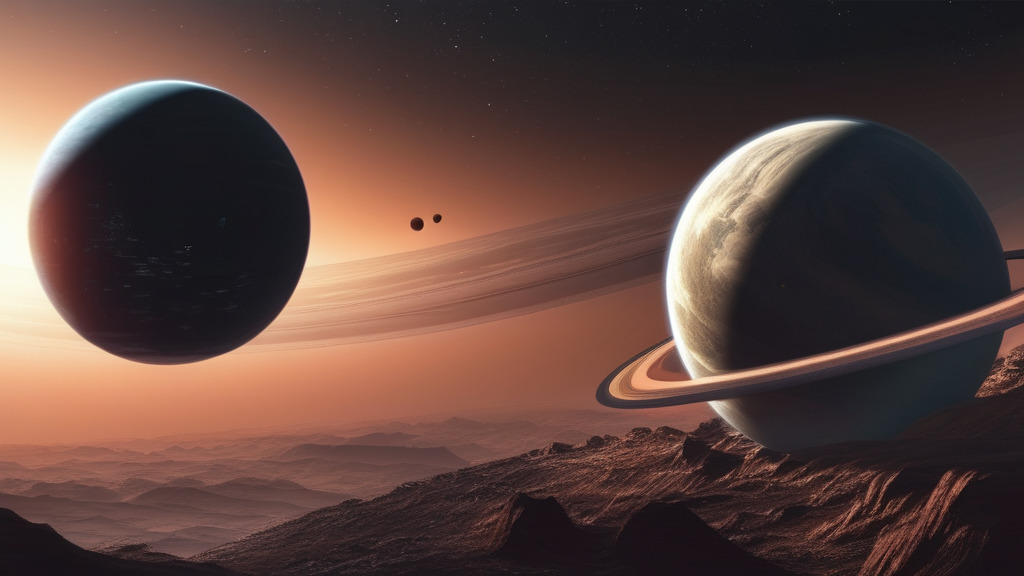}} & \raisebox{-.5\height}{\includegraphics[width=0.09\textwidth]{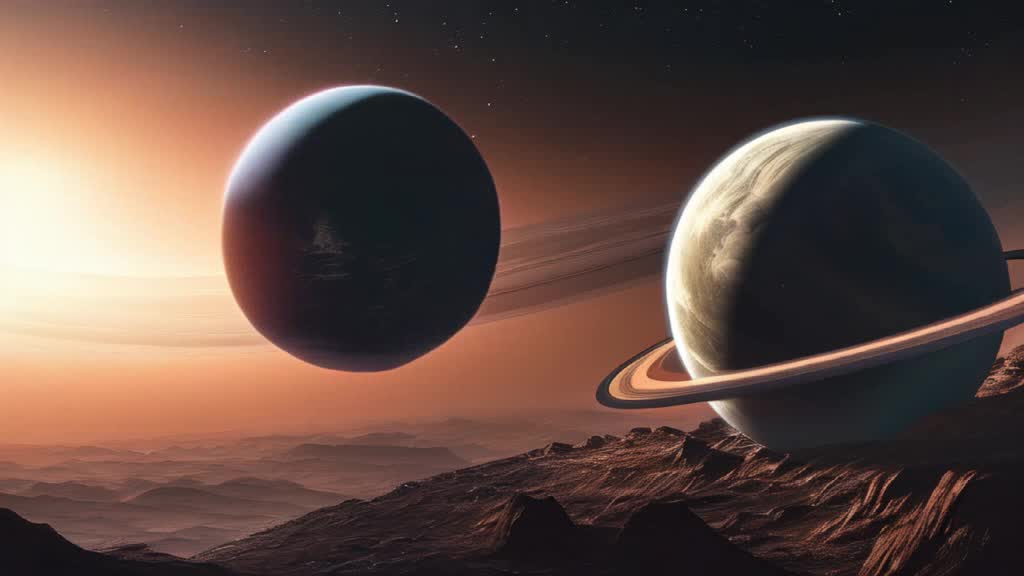}} \raisebox{-.5\height}{\includegraphics[width=0.09\textwidth]{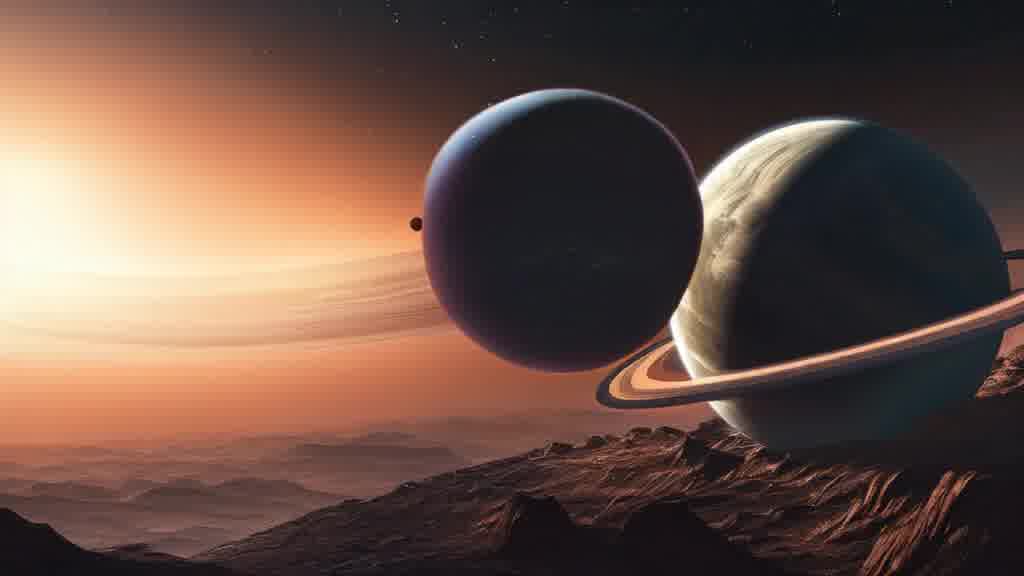}} \raisebox{-.5\height}{\includegraphics[width=0.09\textwidth]{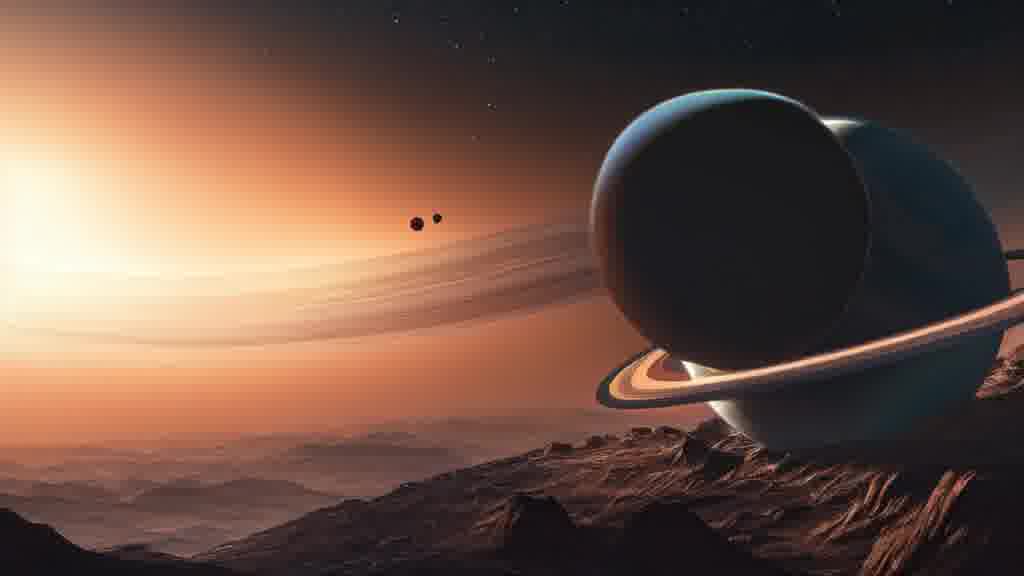}}
	\end{tblr}
	\caption{Results. Our method can successfully transfer semantic video motion across a wide number of domains and motions.}
	\Description{Grid showing eight different motion transfer examples arranged in a grid with four rows and two columns. From top to bottom and left to right: unsteady gait from human to human, waddling from penguin to squirrel, opening mouth from human to orangutan, tilting head back from human to dog, rotating bird's-eye view camera from landscape to landscape, following camera from car to car, simple animation of stick figure jumping to penguin, 2D ball passing in front of another 2D ball to a planet passing in front of another planet.}
	\label{fig:results}
\end{figure*}

Our motion representation is highly versatile, enabling motion transfer across diverse objects and motions, as demonstrated in Fig.~\ref{fig:teaser} and Fig.~\ref{fig:results}. Notably, we do not require a spatial alignment, as seen in row 6 (right) of Fig.~\ref{fig:results}, where the camera follows the moving camper van similar to how it follows the car in the fifth row, despite their misalignment. Our method also applies the motion to all semantically reasonable objects simultaneously ``for free.'' It even supports simple hand-crafted motions, enabling artists to sketch motions (e.g., stick figures) and apply them to complex scenes. For more results, including joint subject and camera motion, extreme cross-domain transfers, and applying the same motion to multiple target images, please refer to Section~\ref{sec:additional-results}.

\subsection{Limitations and Future Work}

Fig.~\ref{fig:failure} shows typical failure cases of our method. Since we do not fine-tune the model, our method inherits the priors and quality of our pre-trained image-to-video model. We observed that the SVD baseline often struggles with object motions, as can be seen in the head example in Fig.~\ref{fig:qual_eval}, where the appearance changes throughout the video. Our method's results have similar issues: in the first example of Fig.~\ref{fig:failure}, the identity of the target person changes when he moves his head to the side. We believe our motion-text embedding does not exacerbate these issues or temporal inconsistencies, as it primarily instructs the model on the desired motion without altering the rest of the model. Often, it seems that the model attempts to produce the desired motion, but its priors are insufficient to generate a satisfactory result. SVD also does not seem to be able to handle some combinations of motions and given input images, likely because they fall outside of the range of the training data set. When the domain gap between motion reference video and target image is too large, our method may leak the structure of the motion reference video into the generated video. In the second example of Fig.~\ref{fig:failure}, when applying a laid-back walking style to a kangaroo, the kangaroo starts walking, but its feet and overall structure become more human-like. Lastly, we found that some motions are not transferred or to a smaller extent. This is especially visible if a video has multiple motions, where the more fine-grained motion is sometimes not transferred. In the third example of Fig.~\ref{fig:failure}, the person pretends to squat down and type on a keyboard. The dinosaurs in the generated video do squat down, but their hands do not move. We hypothesize that fine-grained motions are also a general limitation of SVD. Overall, we expect better results of our method as image-to-video models improve.
In Section~\ref{sec:failure-rate-analysis}, we analyze our method's failure rate in more detail.

An important practical consideration is that the target image must be temporally aligned with the first frame of the motion reference video, as it serves as the starting frame. This is not a limitation of our method specifically, but rather a consequence of the task formulation. Alternatively, one could treat the image as an appearance reference (as in Animate Anyone~\cite{animateanyone}) and adapt or fine-tune the model accordingly.

While more accessible than methods requiring extensive training or fine-tuning, our approach requires an optimization procedure that takes about one hour per motion on an A100 (80 GB) GPU. We have also run it on 48 GB GPUs, albeit with slightly longer runtimes. We encourage future work on reducing the per-motion optimization time, or eliminating it entirely by learning to predict motion-text embeddings directly from motion reference videos, scaling our method to longer videos, as well as adapting it to newer architectures based on diffusion transformers~\cite{DiT}.

\section{Conclusion}

We introduce the general task of transferring the semantic motion of a reference video to any target image. We observe and exploit inherent advantages of image-to-video over text-to-video models for this task and find that text/image embedding tokens are well-suited as a motion representation. Specifically, our method, \emph{motion-textual inversion}, optimizes an inflated version of the text/image embedding for a given motion reference video. Due to its general nature, this motion can then be applied to a wide number of objects and domains. Our method thus enables completely novel applications and takes a significant step towards being able to reenact anything.

\begin{acks}

We would like to thank Michael Bernasconi, Dominik Borer, Jakob Buhmann, and Daniela Kansy for providing motion videos as well as all participants featured in our internal data sets. We also want to give special thanks to Bastian Amrhein, Michael Bernasconi, Vukasin Bozic, Karlis Briedis, Pascal Chang, Guilherme Haetinger, Christopher Otto, Lucas Relic, Seyedmorteza Sadat, and Agon Serifi for their valuable and insightful discussions throughout the project.

\end{acks}

\clearpage

\bibliographystyle{ACM-Reference-Format}
\bibliography{bibliography}
\clearpage

\onecolumn
\appendix
\section{Broader Impact and Ethics}

To the best of our knowledge, our method is the first that can reenact a wide array of objects and motions given a target image and motion reference video without training domain-specific models. We believe this represents a significant advancement in controllable video generation, as our approach can address multiple existing domain-specific scenarios within a single framework and even facilitate entirely new applications. That said, we acknowledge the potential for misuse of reenactment methods like ours, such as creating realistic deepfakes or videos depicting individuals or objects performing specified, potentially inappropriate actions. We strongly condemn such misuse and advocate for implementing safety mechanisms and procedures in real-world applications. Additionally, we support ongoing research into detecting fake videos to mitigate these risks.

For legal reasons, we cannot show images or videos from public data sets in the paper without individuals' written consents. For the qualitative evaluation, we therefore use motion reference videos and target images from internal data sets as well as target images generated with Stable Diffusion XL~\cite{sdxl}. 

\section{Extended Related Work} \label{sec:related_work_extra}

In this section, we provide an extended description of related work for interested readers.

\subsection{Domain-Specific Reenactment} \label{sec:related_work_extra-domain-specific}

Reenactment has been a significant research area, but much of the focus has been on domain-specific approaches like face reenactment~\cite{fsgan, wang2021one, hsu2022dual, megaportraits, li2023one, liveportrait} and human full-body motion transfer~\cite{transmomo, everybody_dance_now, follow_your_pose, dreampose, animateanyone, motioneditor, champ, edit-your-motion, vividpose, motionfollower}. While these methods perform well, their architectures and training data are tailored to specific domains, making it challenging to adapt them for use across multiple domains. 

\subsection{Keypoint-Based Motion Transfer}

Keypoint-based motion transfer has been a popular approach in reenactment, spanning both domain-specific and more general methods. Many techniques extract keypoints using pre-trained, domain-specific landmark detectors~\cite{transmomo, ni2023cross, everybody_dance_now, follow_your_pose, animateanyone, motioneditor, edit-your-motion, fsgan, hsu2022dual}, which limits their applicability to specific object categories like human bodies or faces. To move toward general motion transfer, other approaches learn keypoints in an unsupervised manner~\cite{fomm, mraa, tpsmm, anamodiff, wang2021one, megaportraits, liveportrait}. Although this strategy increases flexibility, it still typically requires a separate model per domain, making it impractical for applications involving diverse object types.

Several methods first find meaningful common keypoints and then warp features~\cite{ni2023cross, fomm, mraa, tpsmm} or latents~\cite{anamodiff} to transfer motion from the driving to the target object. However, such warping becomes nontrivial in the presence of 3D rotations, and methods like AnaMoDiff~\cite{anamodiff} are thus limited to flat 2D motions. JOKR~\cite{jokr}, while not relying on explicit warping, also focuses on relatively planar 2D motions and requires an affine alignment between the target and the driving video. Crucially, both JOKR and AnaMoDiff require a target video to learn target object motions, whereas our method works well even with a single target image by leveraging motion priors from a pre-trained image-to-video model.

Keypoint-based approaches also face challenges when applied to unseen domains or extreme cross-domain transfers (e.g., from animal to inanimate object). While recent advances in deep features from diffusion models~\cite{hedlin2023unsupervised, dift, diffusion-hyperfeatures, sd-complements-dino, telling-left-from-right} have made it easier to find correspondences between points across different images, a more fundamental problem remains: where to place keypoints in the first place to meaningfully capture motion. This becomes especially difficult for hand-crafted motions or for motion transfers with large structural differences between objects (e.g., Fig.~\ref{fig:results_style}), where there may be no obvious semantically meaningful anchors. To address these challenges, we propose using an implicit motion representation instead of relying on explicit keypoints. We show that priors from pre-trained diffusion models can be used more directly, rather than only as a tool to find keypoint correspondences.

\subsection{Video Generation}

Following the rise of text-to-image diffusion models~\cite{dalle2, imagen, ldm}, video generation models have also greatly improved in quality in recent years. Many text-to-video methods start with a pre-trained text-to-image model and inflate it by adding and training temporal convolution and attention blocks after each corresponding spatial block~\cite{animatediff, align_your_latents, modelscope, lumiere}. Similarly, many image-to-video diffusion models use a pre-trained text-to-image~\cite{i2vgen_xl} or text-to-video~\cite{svd} model as a starting point. They then adapt the model to the image-to-video task by conditioning the model on the image, e.g., by adding~\cite{i2vgen_xl} or concatenating~\cite{svd} it to the noisy input. The text embedding input from the pre-trained model is either kept~\cite{i2vgen_xl} or replaced with an image embedding input~\cite{svd}. Recently, video generation models~\cite{sora, cogvideox, hunyuanvideo} based on diffusion transformers~\cite{DiT} have gained significant popularity. While training a custom video generation model provides the most freedom in terms of design choices, it is very expensive in terms of computation and data. Even fine-tuning video models requires substantial resources, so we decided to use a pre-trained diffusion model, Stable Video Diffusion~\cite{svd}, and keep it frozen. Additionally, we aim for our method to be applicable to a wide range of motions and subjects. In contrast, approaches that involve training the model often focus on a single type of motion, such as human full-body motion~\cite{follow_your_pose, animateanyone}.

\subsection{Video Motion Editing with Explicit Motions}

\subsubsection{Based on Sparse Control Signals} \label{sec:related_work_extra-sparse}

In theory, the motion of all video generation models that have a text input can simply be controlled by text~\cite{dreamix, moca, animateanything, animateyourmotion}, but this approach struggles with complex motions in practice. For more precise spatial control, recent methods use bounding boxes, either with training~\cite{boximator, animateyourmotion} or without~\cite{trailblazer, motion_zero, peekaboo}, and trajectories~\cite{MCDiff, dragnuwa, draganything, ReVideo, MOFA-video, FreeTraj, image_conductor, puppet-master, trackgo, motionprompting}, but they rely on consistent spatial alignment for effective motion transfer. Similarly, keypoints are another option for describing motions~\cite{videoswap, anamodiff, MOFA-video}, but they suffer from the challenges outlined in Section~\ref{sec:related_work_extra-domain-specific}. Additionally, some methods focus specifically on camera motions~\cite{cameractrl, vd3d, CamTrol, CamCo, motionmaster, cami2v, cheong2024boosting} or combine camera and bounding box motions~\cite{motionctrl, direct_a_video, motionbooth}. However, all these approaches are either limited to simple motions or require significant effort to specify complex ones. For instance, a bounding box can specify an object's location (e.g., a person) but not the detailed motion within it (e.g., doing jumping jacks). Modeling complex motion with part-based boxes or trajectories~\cite{puppet-master} quickly becomes impractical, especially if a precise temporal alignment to a reference motion is desired. 

\subsubsection{Based on Dense Control Signals}

Dense control signals, such as motion vectors~\cite{videocomposer}, 3D tracking videos~\cite{diffusion_as_shader}, warped noise~\cite{go_with_the_flow}, and depth maps~\cite{videocomposer, controlvideo, control_a_video} allow for a more precise motion specification. However, using them for general motion transfer is challenging because they also encode information about image and object structure. This can result in unnatural motions when there is a mismatch between the structures of the target image and the reference video as shown in MotionCtrl~\cite{motionctrl}.

\subsection{Video Motion Editing with Implicit Motions}

This subsection covers methods for implicitly representing and transferring motion from a reference video. We thereby focus on the two main paradigms: fine-tuning approaches, which encode motion into model weights, and inversion-then-generation methods, which capture motion in model features and attention maps. Additionally, some techniques integrate elements of both paradigms.

When the layout of the subjects in the reference and generated videos match, a given transfer can be seen as either changing the appearance to match the target image or altering the motion to match the reference video. Our focus is on motion transfer where the layouts do not align, a less explored area in the literature, as discussed in Section~\ref{sec:related_work_extra-spatial-variations}.

\subsubsection{Fine-Tuning} \label{sec:related_work_extra-fine-tuning}

Many fine-tuning methods are inspired by image customization techniques like DreamBooth~\cite{dreambooth} and LoRA~\cite{lora}. Loosely speaking, the idea is to fine-tune the parts of the model responsible for motion but avoid training the parts responsible for appearance. Tune-A-Video~\cite{tune_a_video} inflates a text-to-image model by adding spatio-temporal attention and only trains some parts of the attention layers. Similarly, \citet{materzynska2024newmove} only fine-tune parts of the model and further focus the training more on earlier denoising steps to emphasize learning the general motion rather than fine appearance details. MotionDirector~\cite{motiondirector} proposes a dual-path LoRA architecture and an appearance-debiased temporal loss to disentangle appearance from motion. Similarly, DreamVideo~\cite{dreamvideo}, MotionCrafter~\cite{motioncrafter}, Customize-A-Video~\cite{ren2024customize}, and CustomTTT~\cite{CustomTTT} have separate branches for appearance and motion. CustomTTT~\cite{CustomTTT} further proposes a test-time training method to improve the results when combing the appearance and motion information. VMC~\cite{VMC} adapts temporal attention layers using a motion distillation strategy with residual vectors between consecutive noisy latent frames as the motion reference.

Fine-tuning a model carries the risk of appearance leakage, which is why many of the aforementioned methods focus on preventing it. We find that using an image-to-video model instead of a text-to-video model largely avoids these problems. LAMP~\cite{lamp} is the most similar method to ours in that sense, but they adapt a pre-trained text-to-image model to the image-to-video task and fine-tune it only briefly. In contrast, we employ a pre-trained, large-scale image-to-video model to leverage stronger priors for better generalization.

\subsubsection{Inversion-then-Generation} \label{sec:related_work_extra-inversion-then-generation}

The inversion-then-generation framework, initially developed for image editing~\cite{prompt_to_prompt, plug_and_play, pix2pix-zero}, involves first inverting a reference video into ``noise'' using methods like DDIM~\cite{ddim} to enable reconstruction through backward diffusion. Thereby, features such as self-attention maps are extracted from the reference video and then injected into the diffusion process of the video being generated. These features either directly replace existing features~\cite{plug_and_play} or are incorporated into a loss function~\cite{pix2pix-zero}, ensuring the generated video has a similar structure. Numerous methods have been proposed within this framework for video appearance editing~\cite{pix2video, zeroshot, rerender_a_video, video_p2p, tokenflow, uniedit, make-a-protagonist, genvideo, motionflow} and video motion editing~\cite{uniedit, space_time_diffusion}, mainly differing in their inversion techniques and feature choices.

The methods mentioned above face several inherent issues in motion transfer tasks. Most notably, they often assume or enforce that the features of the reference and target videos are identical, which leads to problems when generating videos with different geometries or spatial layouts. Some methods attempt to address this by collapsing the spatial dimension of features before using them in a loss~\cite{space_time_diffusion}, but they still typically produce motions with similar directions in pixel space. This limits control and diversity and can produce less natural results. Furthermore, these approaches require tuning numerous hyperparameters (choice of feature, layers, time steps) and necessitate inverting the video, which is challenging for high guidance scales~\cite{null_text_inversion} and when using few time steps~\cite{ReNoise}.

Another recent line of work~\cite{MOFT, DiTFlow, motionclone} extracts features from a reference video in line with the inversion-then-generation framework but without inversion. While these approaches bypass the costly inversion process, they still suffer from issues related to primarily replicating the spatial rather than semantic motion.

\subsubsection{With Different Spatial Layout} \label{sec:related_work_extra-spatial-variations}

To avoid being restricted to the layout of a single motion reference video, some methods use multiple motion videos~\cite{lamp, motiondirector, dreamvideo, materzynska2024newmove}. However, our goal is to transfer motion with precise temporal alignment to the reference video. This would require multiple temporally-aligned videos, which are often impractical to obtain. Additionally, many motion editing methods with spatial variations~\cite{materzynska2024newmove, wang2024motion, ren2024customize, motrans} use text to define the subject's appearance instead of an image, resulting in videos that only roughly match the input image. The concurrent work by \citet{wang2024motion} is most similar to ours as it keeps the model frozen and learns a motion embedding like we do, but it also suffers from the above limitation.

\section{Implementation Details} \label{sec:implementation-details}

\subsection{High-Level Overview of the Implementation}

To aid in reproducibility, we list the main steps of our method's implementation below:

\begin{enumerate}
    \item {[}Only once{]} Take pre-trained Stable Video Diffusion~(SVD) \cite{svd} and adapt code to inflate motion-text embedding and cross-attention. See high-level description in Section~\ref{sec:inflation} and details in Section~\ref{sec:inflation-details}.
    \item Initialize motion-text embedding of shape $(F + 1) \times N \times d$. See Section~\ref{sec:hyperparameters}.
    \item Repeat until convergence:
    \begin{itemize}
        \item Load same $F$ frames of reference video in data loader for each iteration.
        \item Augment data. See Section~\ref{sec:hyperparameters}.
        \item Input noisy version of frames, motion-text embedding, and other inputs into SVD.
        \item Apply loss to update motion-text embedding.
    \end{itemize}
    \item Save motion-text embedding.
    \item For all target images:
    \begin{itemize}
        \item Input learned motion-text embedding along with new target image to inflated SVD during inference to generate video with motion from reference video.
    \end{itemize}
\end{enumerate}

\subsection{Hyperparameters} \label{sec:hyperparameters}

Our implementation builds up on the diffusers implementation~\cite{diffusers} of Stable Video Diffusion~(SVD)~\cite{svd}. We use the default parameters of the $14$-frame version of SVD (e.g., micro-conditionings) unless specified otherwise. Like SVD, we generally employ a classifier-free guidance~\cite{cfg} scale that increases linearly from $1$ to $3$ across the frame axis. For the motion visualization (unconditional image input), however, we use a higher scale, i.e., increasing linearly from $1$ to $10$, to improve the visibility of the objects. We initialize the $F=14$ sets of $N=5$ tokens for the spatial cross-attention with the CLIP image embedding token of each corresponding frame and the $N=5$ tokens for the temporal cross-attention with the mean of the CLIP image embedding tokens across all frames. We additionally add Gaussian noise $\mathcal{N}(0, 0.1)$ to the combined motion-text embedding during initialization. In our experience, the initialization does not affect the results significantly, so other initializations are equally reasonable. During optimization, we always pick the same $F$ frames of a given video and apply the same spatial and color augmentations to all frames.\footnote{For horizontal camera motions, we turn of horizontal flipping} Since most of the video motion is determined in noisy diffusion steps, we shift the noise schedule towards higher noise values (from $P_\text{mean}=1.0, P_\text{std}=1.6$ to $P_\text{mean}=2.8, P_\text{std}=1.6$ where $\log \, \sigma \sim \mathcal{N}(P_\text{mean}, P_\text{std}^2)$ to speed up the optimization. We use Adam~\cite{adam} with a learning rate of $10^{-2}$ for $1000$ iterations with a batch size of $1$. 

\subsection{Hardware Requirements and Runtime}

The optimization for a motion reference video with a resolution of $1024 \times 576$ takes around $55$ GB of GPU memory and around one hour on an NVIDIA Tesla A100 (80 GB) GPU. The inference takes less than one minute per video. While the peak memory usage was measured at $55$ GB on the A100, we have also successfully run the method on a $48$ GB RTX A6000 GPU. Our current implementation has not been optimized extensively for memory efficiency or runtime, and further engineering could reduce the resource requirements.

\subsection{Motion-Text Embedding and Cross-Attention Inflation} \label{sec:inflation-details}

\begin{figure*}[htbp]
	\centering
	\subfloat[a][Default SVD~\cite{svd}: Since the image embedding $\mb{e}$ has only one token, the softmax operation causes all entries of the cross-attention maps to be $1$. Therefore, the section highlighted in yellow simplifies to a broadcasted version of the value vector of that token.\label{fig:inflation-technical-default}]{\includegraphics[trim = 5mm 62mm 158mm 58mm, width=0.65\textwidth, clip]{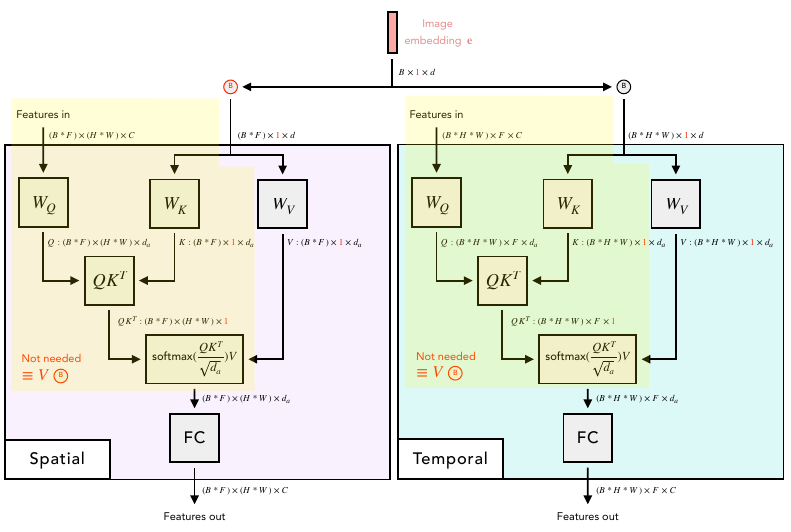}}
	\\
	\subfloat[b][Inflated SVD~\cite{svd} (Ours): We use $N$ tokens instead of $1$, so the model now dynamically attends to different tokens depending on the spatial and temporal location. Additionally, we use different sets of tokens per frame for the spatial cross-attention instead of broadcasting the same tokens to all frames.\label{fig:inflation-technical-ours}]{\includegraphics[trim = 157mm 62mm 6mm 56mm, width=0.65\textwidth, clip]{figures/inflation_technical.pdf}}
	\caption{Technical diagrams of the motion-text embedding and cross-attention inflation showing the dimensions of the features of the spatial and temporal cross-attention blocks. The changes between the default SVD~\cite{svd} and our inflated version are shown in red font. $B$ = batch size, $F$ = number of frames, $C$ = number of channels, $H$ = height, $W$ = width, $d$ = embedding dimension, $d_a$ = attention dimension, $N$ = token dimension, $W_Q$ = query weight matrix, $W_K$ = key weight matrix, $W_V$ = value weight matrix, $Q$ = queries, $K$ = keys, $V$ = values, FC = fully connected layer. For simplicity, the multiple attention heads and block level $i$ indices are not shown.}
	\Description{Two diagrams of the cross-attention layers of the default SVD~\cite{svd} in (a) and our proposed inflated version in (b). Each diagram shows the spatial cross-attention layer components and the feature dimensions on the left and the temporal cross-attention on the right. Diagram (a) shows the image embedding $\mb{e}$ with a single token on top, which is passed as input to the spatial and temporal cross-attention layers. A large section of both layers is highlighted in yellow because it can be simplified to a broadcast operation since there is only one token. Diagram (b) shows our motion-text embedding $\mb{m}$ with $F+1 \times N$ tokens which is split into two parts. The part with $F \times N$ tokens is passed as input to the spatial cross-attention layer whereas the other part with $N$ tokens is passed as input to the temporal cross-attention layer. The cross-attention layers look quite similar except that there is no highlighted section, the token dimension is $N$ instead of $1$, and the broadcast operation in the spatial cross-attention layer is replaced with a reshape operation.}
	\label{fig:inflation-technical}
\end{figure*}

This section provides more implementation details for the motion-text embedding and cross-attention inflation described in Section~\ref{sec:inflation}. Fig.~\ref{fig:inflation-technical} shows the spatial and temporal cross-attention layers of the default Stable Video Diffusion~(SVD)~\cite{svd} and our inflated version along with their tensor dimensions.

The image embedding of the default SVD consists of a single token and has dimensions $B \times 1 \times d$, where $B$ is the batch size (in our implementation typically $1$ when optimizing the motion-text embedding and $2$ during inference because of classifier-free guidance) and $d$ is the CLIP~\cite{clip} embedding dimension. For spatial cross-attention, the image embedding is broadcast to dimensions $(B*F) \times 1 \times d$, i.e., the same token is used for all $F$ frames. This results in an attention map $M$ of dimensions $(B*F) \times (H_i*W_i) \times 1$ where $H_i$ and $W_i$ are the spatial heights and widths respectively, and $C_i$ is the number of channels of level $i$ of the diffusion model. Notably, due to the softmax operation and the last dimension being $1$, every value of the attention map is $1$. This means that each spatial location attends $100\%$ to the single token. Similarly, for temporal cross-attention, the image embedding is broadcast from dimensions of $B \times 1 \times d$ to dimensions $(B*H_i*W_i) \times 1 \times d$, eventually leading to an attention map $M$ of dimensions $(B*H_i*W_i) \times F \times 1$  where every value is $1$. Having only one token thus leads to a degenerate case of the cross-attention where $\text{Attention}(Q, K, V) = V$ (broadcasted) and many of the components (e.g., queries and keys) have no effect on the result. 

\subsubsection{Multiple Tokens}

To avoid the above degenerate case and instead be able to dynamically attend to different tokens, we extend the token dimension from $1$ to $N$ where $N$ is a hyperparameter. For spatial cross-attention, this results in an attention map $M$ of dimensions $(B*F) \times (H_i*W_i) \times N$ where, in general, each spatial location has different values $\neq 1$ for the $N$ different tokens. Similarly, the temporal cross-attention map $M$ has dimensions $(B*H_i*W_i) \times F \times N$ with values $\neq 1$. Since SVD was pre-trained using multiple text embedding tokens as input, the code can already handle multiple tokens, so mainly the initialization of the motion-text embedding as well as some input dimensions have to be adapted slightly. 

\subsubsection{Different Tokens per Frame}

As explained in Section~\ref{sec:inflation-different-tokens-per-frame}, we propose to learn different sets of tokens per frame for the \textit{spatial} cross-attention to obtain a higher temporal granularity of the motion. The default SVD implementation broadcasts the embedding from dimensions $B \times N \times d$ across all frames to $(B*F) \times N \times d$ (where $N=1$ originally). We instead learn a larger spatial motion-text embedding of dimensions $B \times F \times N \times d$ and reshape it to $(B*F) \times N \times d$. We keep the dimensions of the temporal motion-text embedding at $B \times N \times d$ and learn it separately. Therefore, the dimensions of the combined spatial and temporal motion-text embedding is $B \times (F+1) \times N \times d$.

\subsubsection{Analogy}

To give an intuitive analogy for our motion-text embedding inflation, think of building a house. Instead of using a single tool for every part of the house, it is more efficient to have $N$ different tools depending on the spatial location on a given floor---like a hammer for the floor and a drill for the wall. Moreover, each of the $F$ floors of the house might need a different set of tools. For example, the roof requires different tools compared to the walls. Similarly, in our approach, we use multiple tokens to handle different aspects of the motion.

\section{Motion-Text Embedding Analysis} \label{sec:embedding-analysis}

SVD was pre-trained as a text-to-video model and dropped the image (latent) input for some percentage of training iterations for classifier-free guidance~\cite{cfg}. We find that SVD can produce somewhat reasonable videos with the image (latent) input zeroed out and only the CLIP~\cite{clip} image embedding as input, especially if we increase the classifier-free guidance scale (e.g., to $10$). We can use this to visualize our learned motion-text embedding with an unconditional appearance.

\begin{figure*}[htbp]
	\centering
	\begin{tblr}{
			cell{1}{2} = {c=2}{c},
			cell{1}{4} = {c=2}{c},
			cell{2}{4} = {c=1}{c},
			cell{2}{5} = {c=1}{c},
			vline{3} = {3-7}{dashed},
			vline{4} = {1-7}{},
			vline{5} = {2-7}{},
		}
		Iter. & Conditional & & Unconditional (Motion visualization) & \\
		& & & Seed 0 & Seed 1 \\
		0 & \raisebox{-.5\height}{\includegraphics[width=0.08\textwidth]{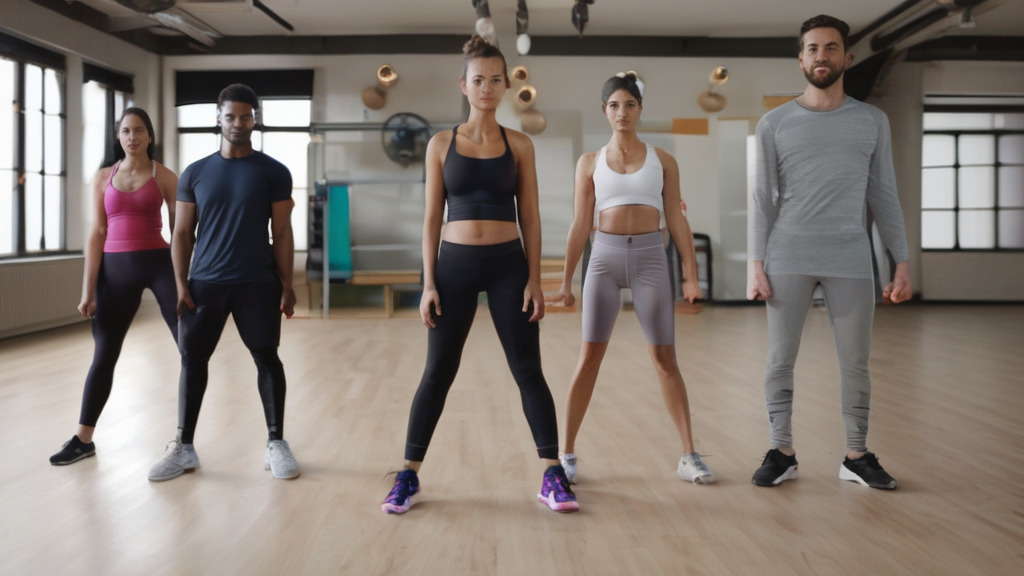}} & \raisebox{-.5\height}{\includegraphics[width=0.08\textwidth]{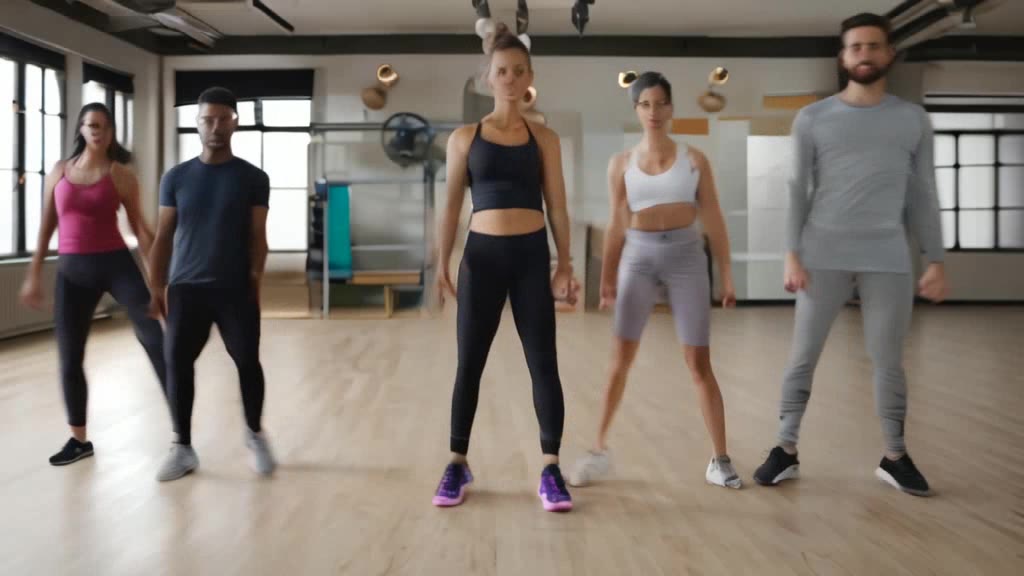}} \raisebox{-.5\height}{\includegraphics[width=0.08\textwidth]{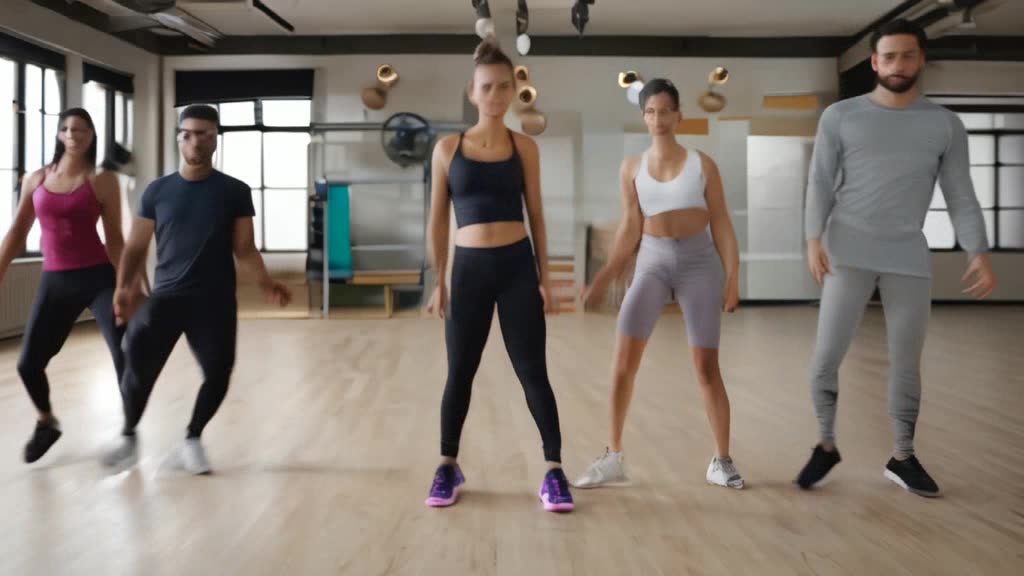}} \raisebox{-.5\height}{\includegraphics[width=0.08\textwidth]{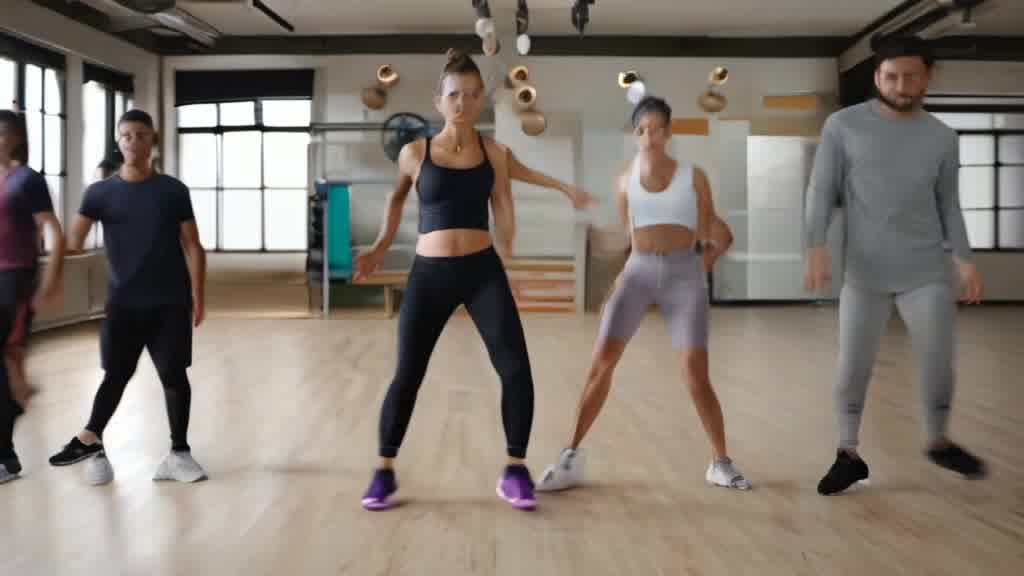}} & \raisebox{-.5\height}{\includegraphics[width=0.08\textwidth]{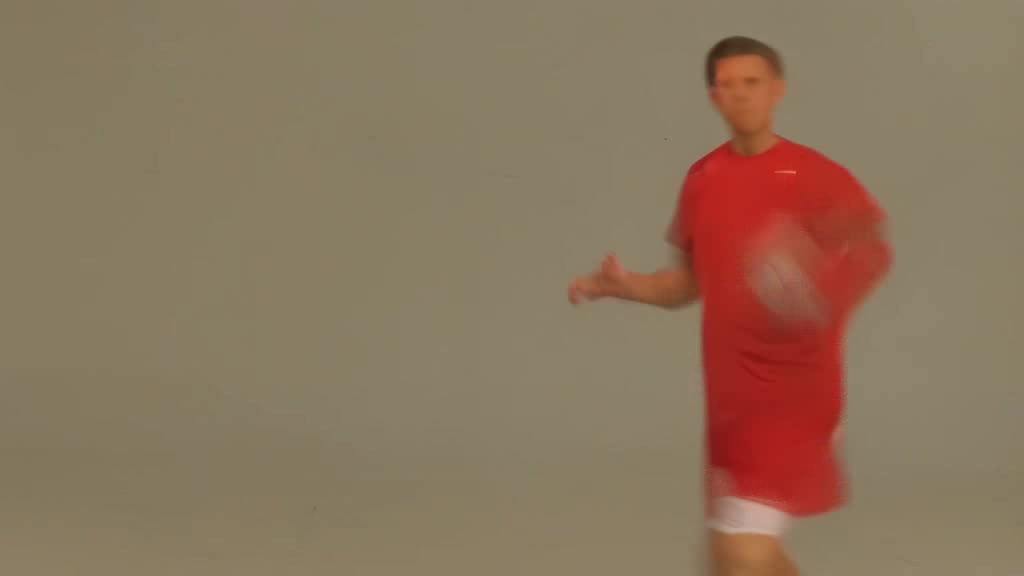}} \raisebox{-.5\height}{\includegraphics[width=0.08\textwidth]{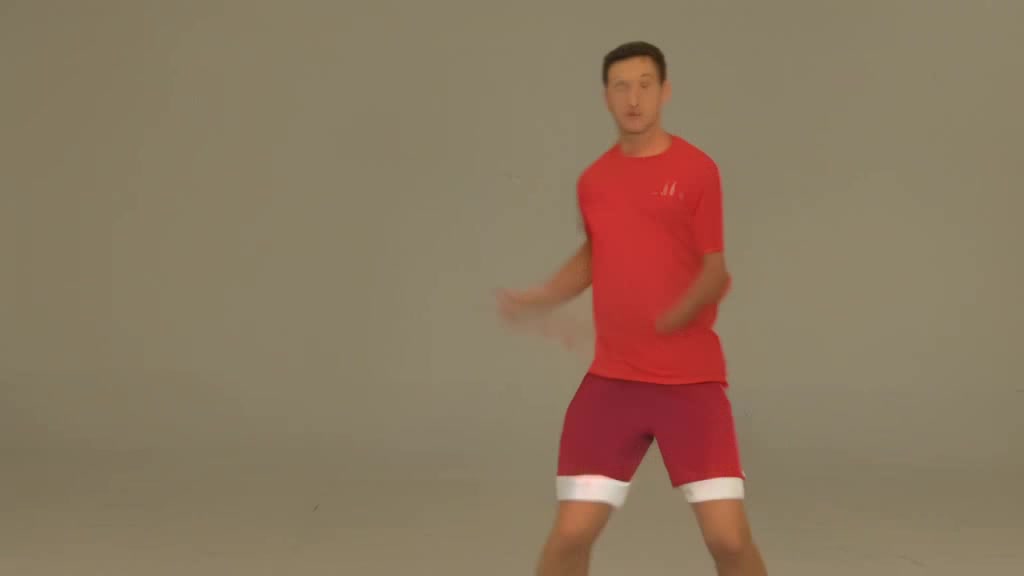}} \raisebox{-.5\height}{\includegraphics[width=0.08\textwidth]{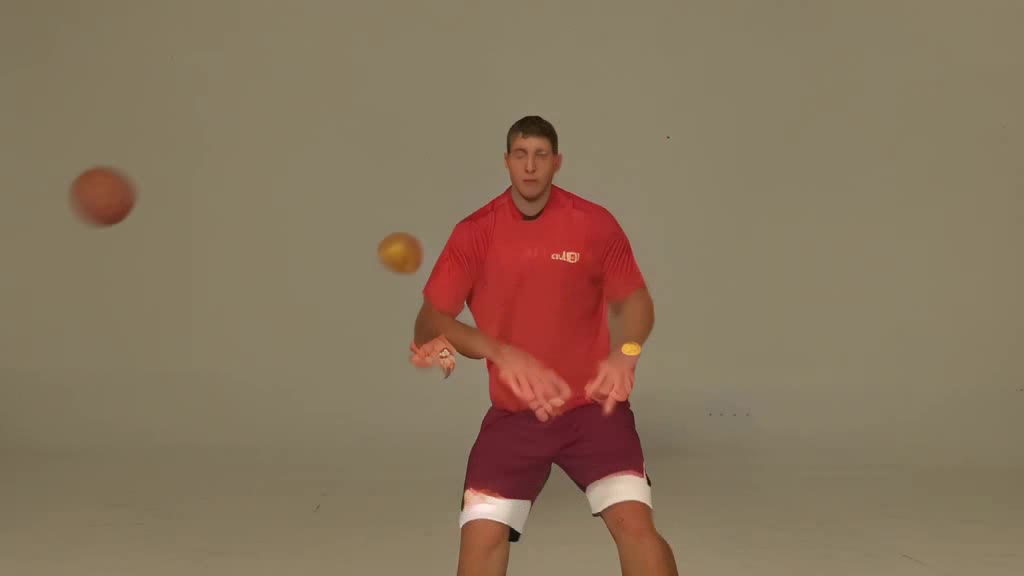}} & \raisebox{-.5\height}{\includegraphics[width=0.08\textwidth]{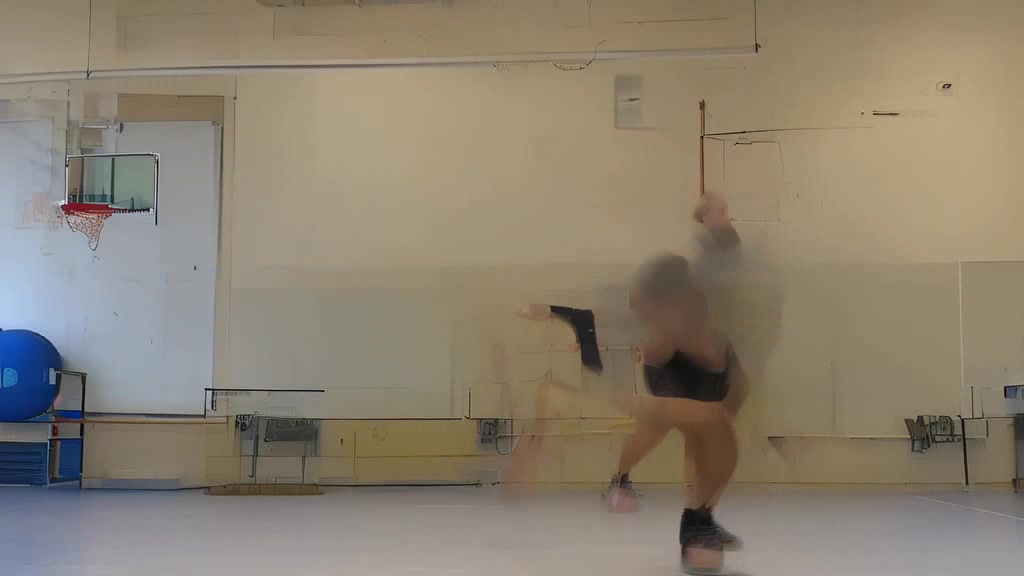}} \raisebox{-.5\height}{\includegraphics[width=0.08\textwidth]{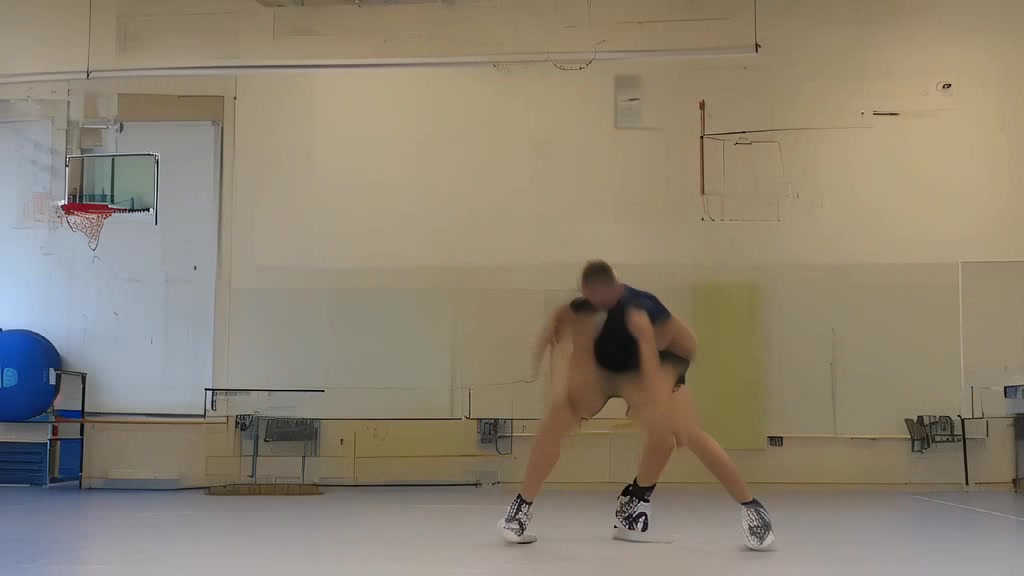}} \raisebox{-.5\height}{\includegraphics[width=0.08\textwidth]{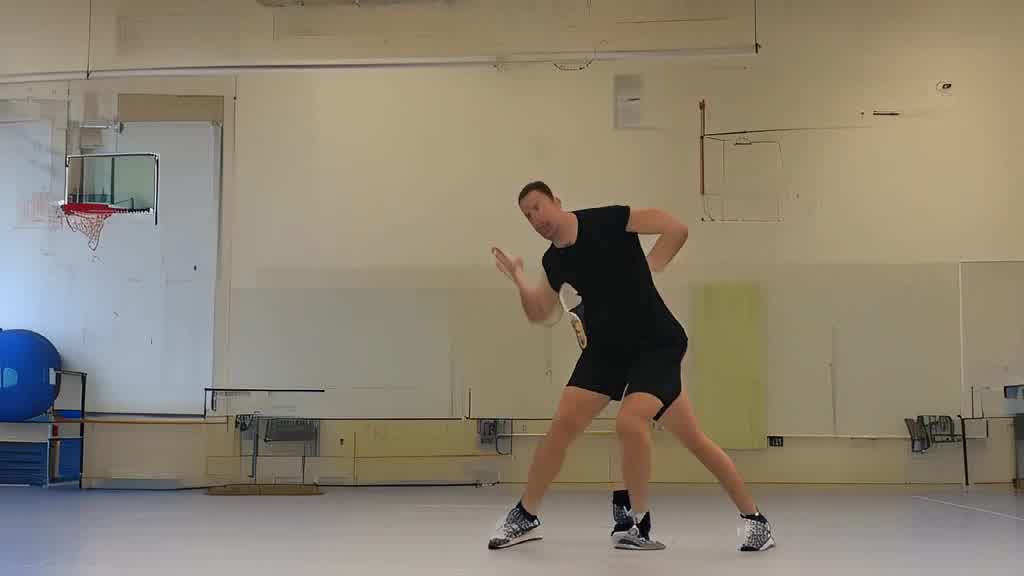}} \\
		200 & \raisebox{-.5\height}{\includegraphics[width=0.08\textwidth]{figures/motion_vis/first_frame.jpg}} & \raisebox{-.5\height}{\includegraphics[width=0.08\textwidth]{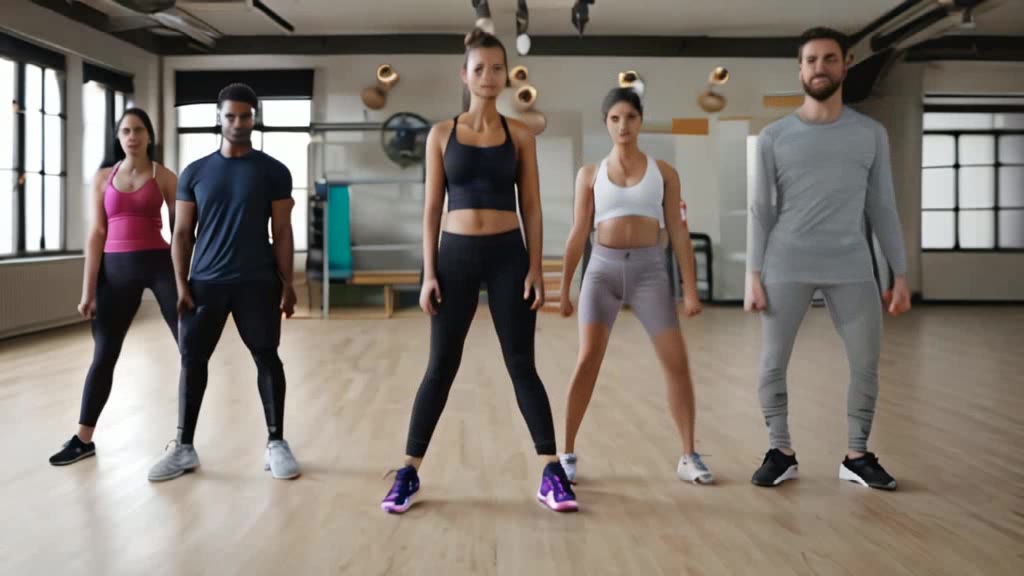}} \raisebox{-.5\height}{\includegraphics[width=0.08\textwidth]{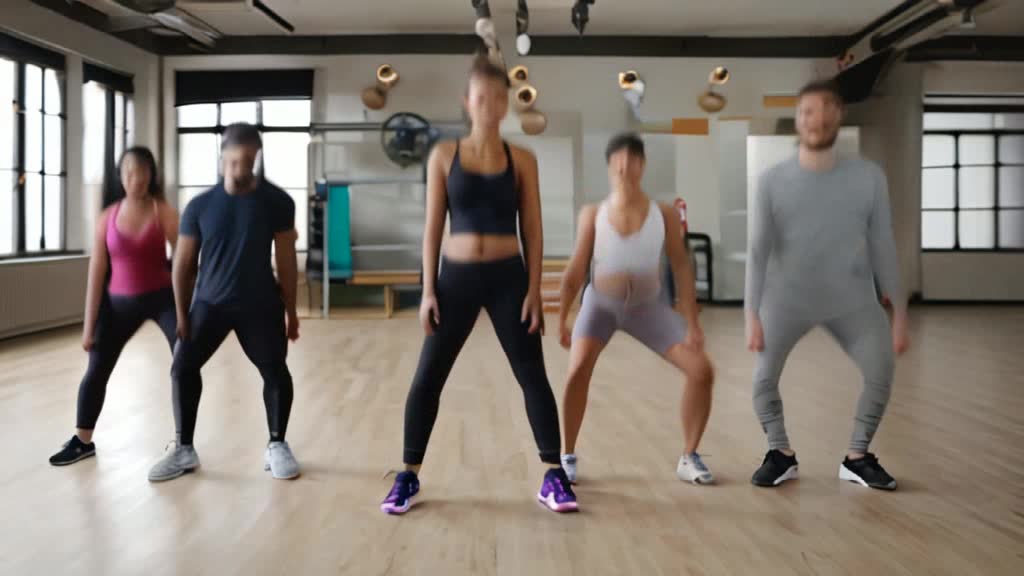}} \raisebox{-.5\height}{\includegraphics[width=0.08\textwidth]{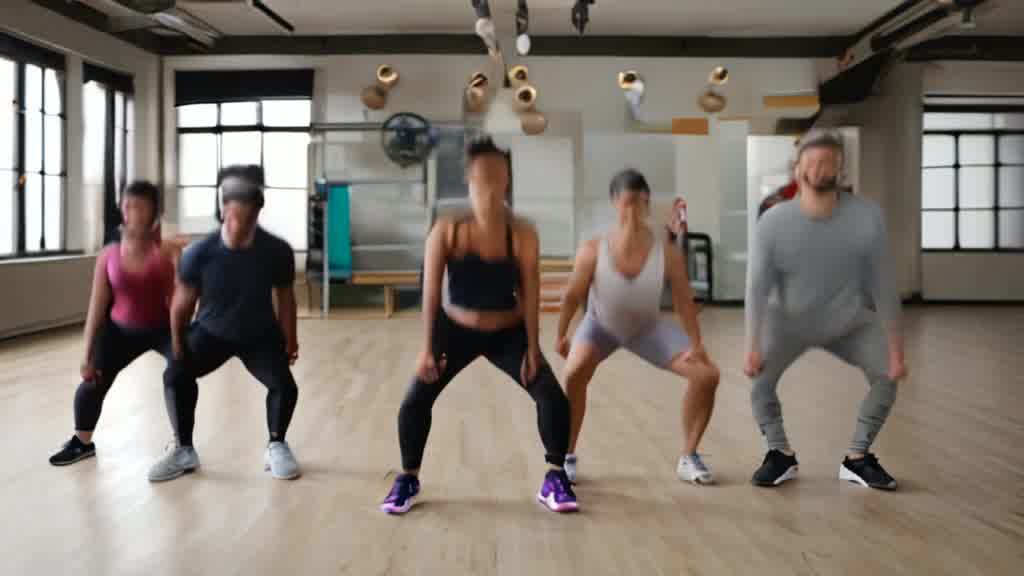}} & \raisebox{-.5\height}{\includegraphics[width=0.08\textwidth]{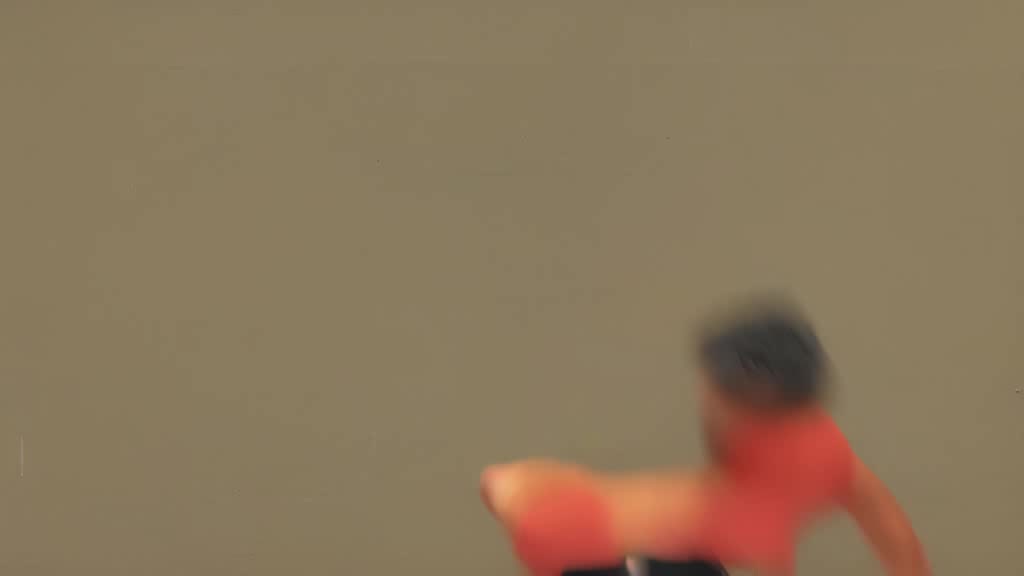}} \raisebox{-.5\height}{\includegraphics[width=0.08\textwidth]{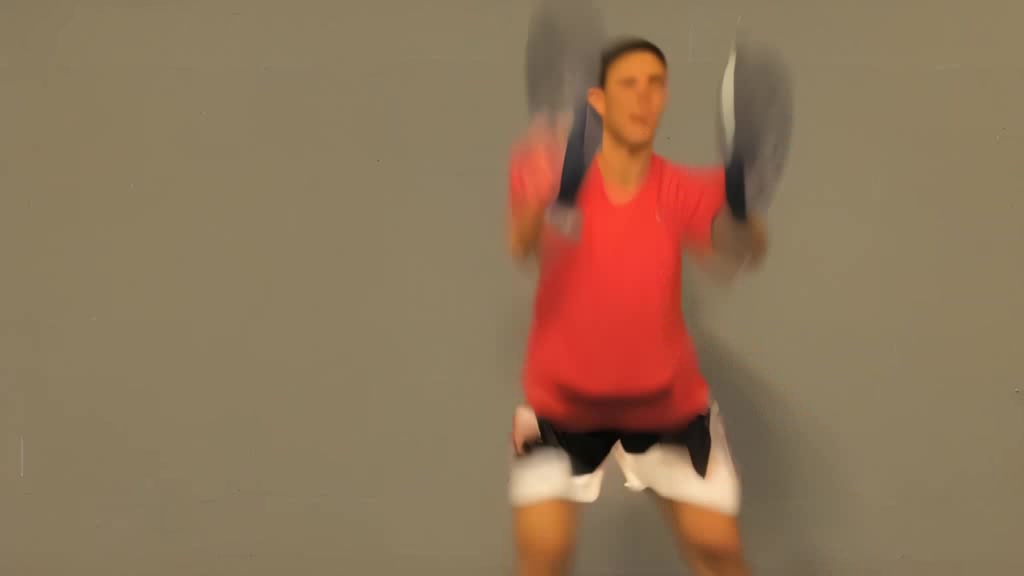}} \raisebox{-.5\height}{\includegraphics[width=0.08\textwidth]{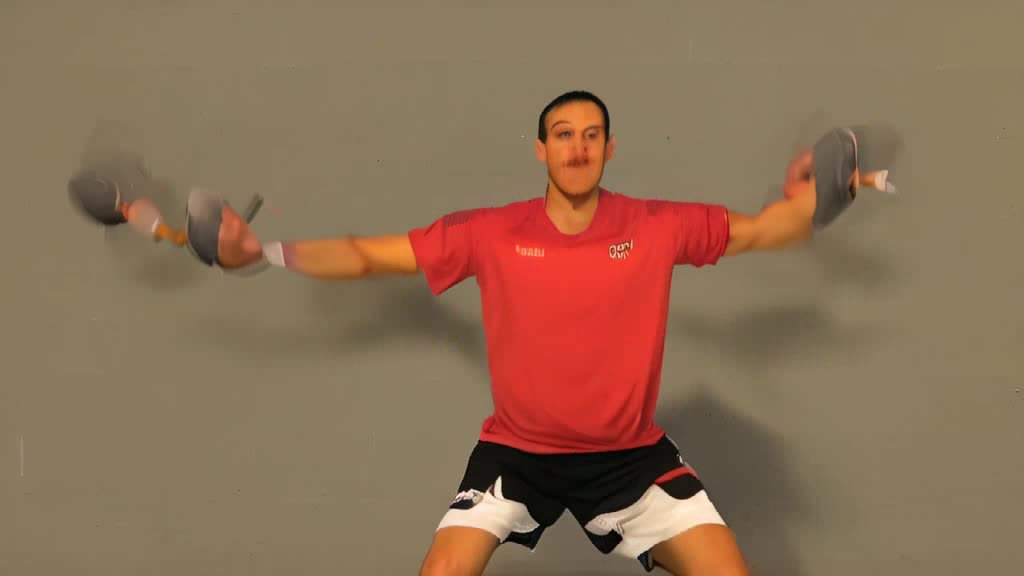}} & \raisebox{-.5\height}{\includegraphics[width=0.08\textwidth]{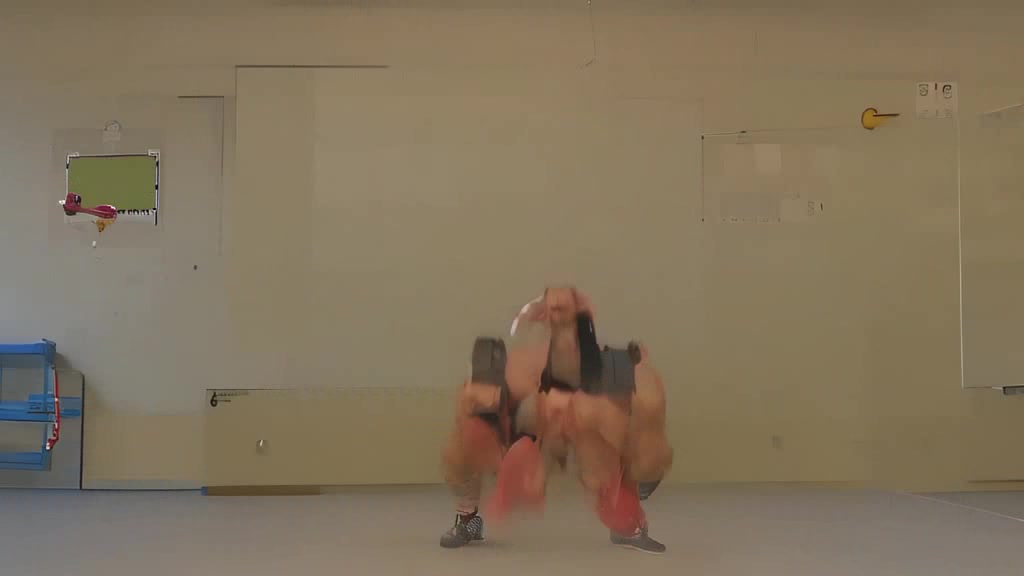}} \raisebox{-.5\height}{\includegraphics[width=0.08\textwidth]{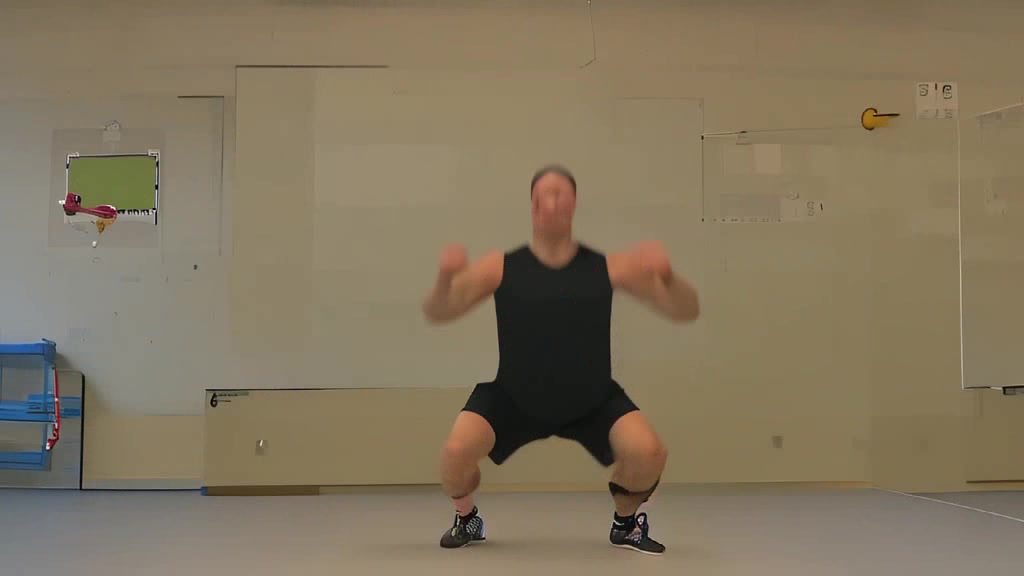}} \raisebox{-.5\height}{\includegraphics[width=0.08\textwidth]{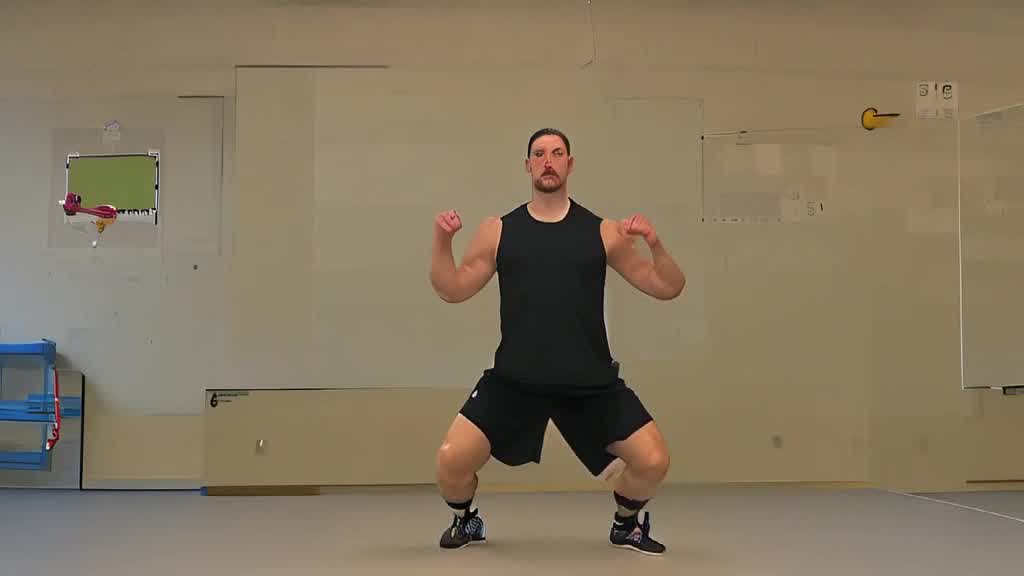}} \\
		500 & \raisebox{-.5\height}{\includegraphics[width=0.08\textwidth]{figures/motion_vis/first_frame.jpg}} & \raisebox{-.5\height}{\includegraphics[width=0.08\textwidth]{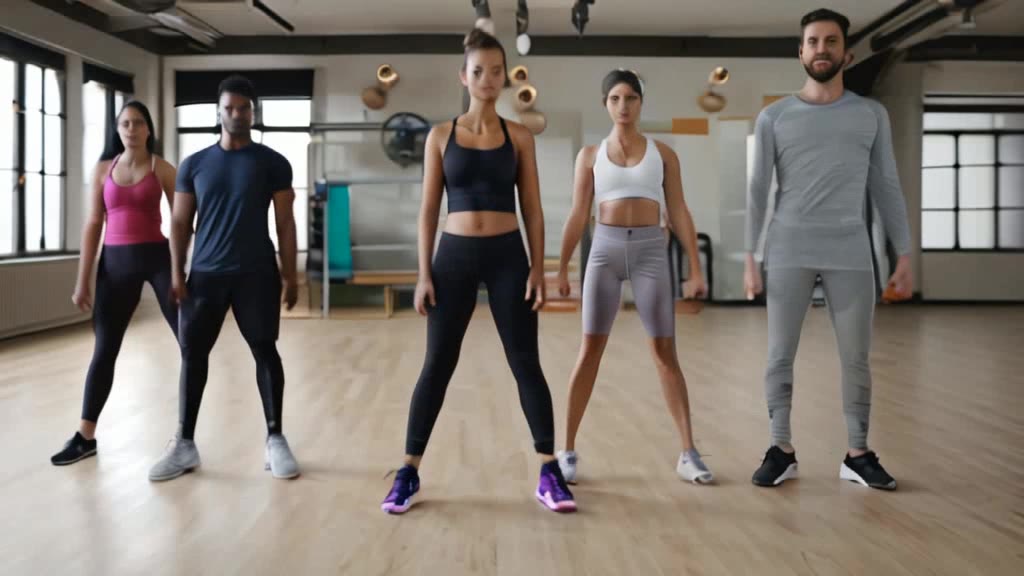}} \raisebox{-.5\height}{\includegraphics[width=0.08\textwidth]{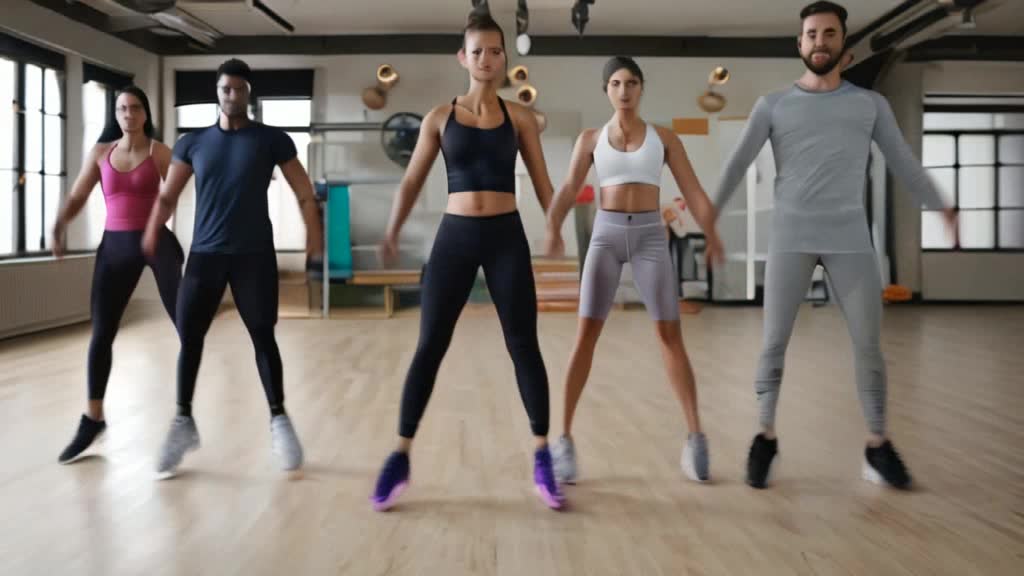}} \raisebox{-.5\height}{\includegraphics[width=0.08\textwidth]{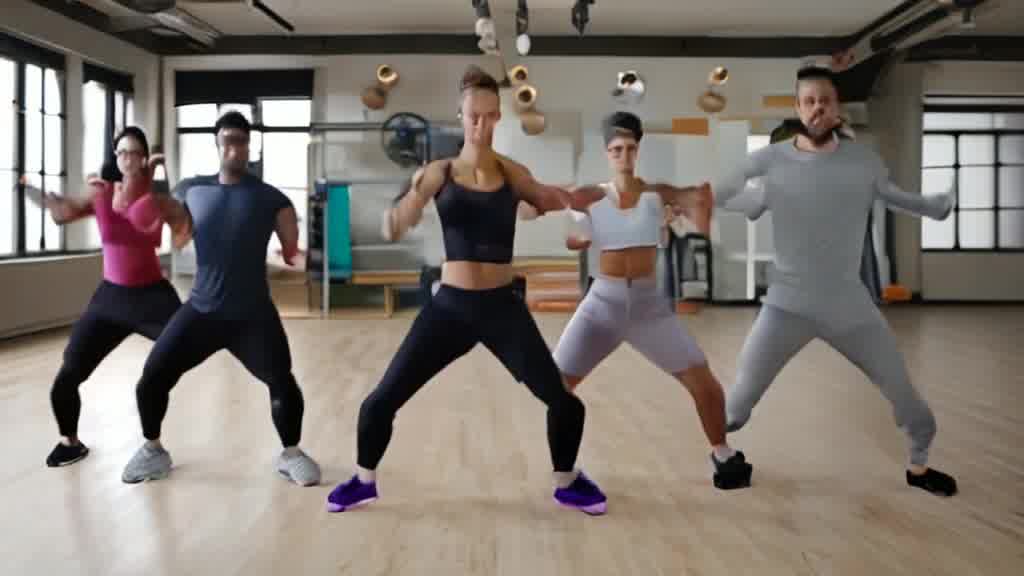}} & \raisebox{-.5\height}{\includegraphics[width=0.08\textwidth]{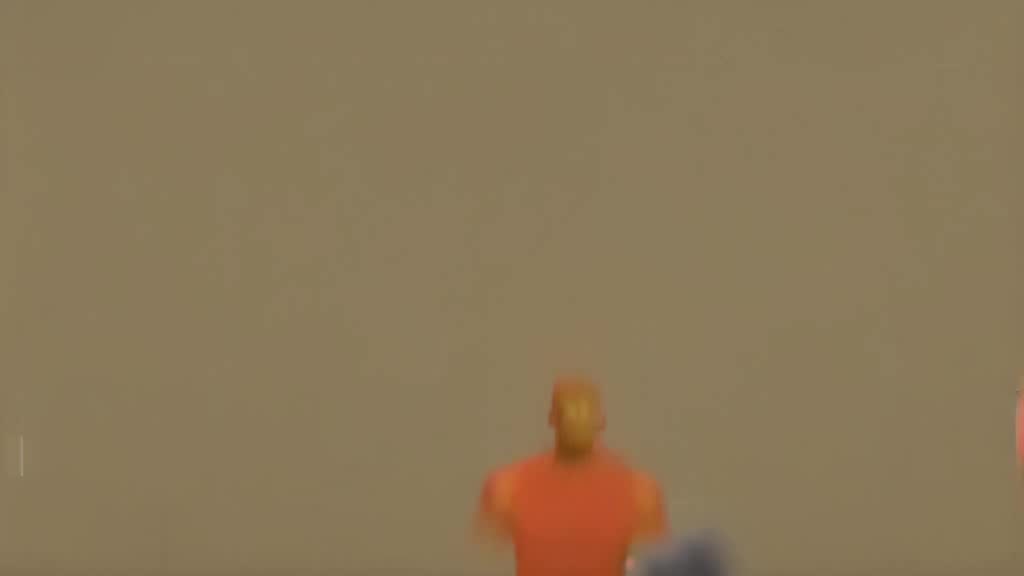}} \raisebox{-.5\height}{\includegraphics[width=0.08\textwidth]{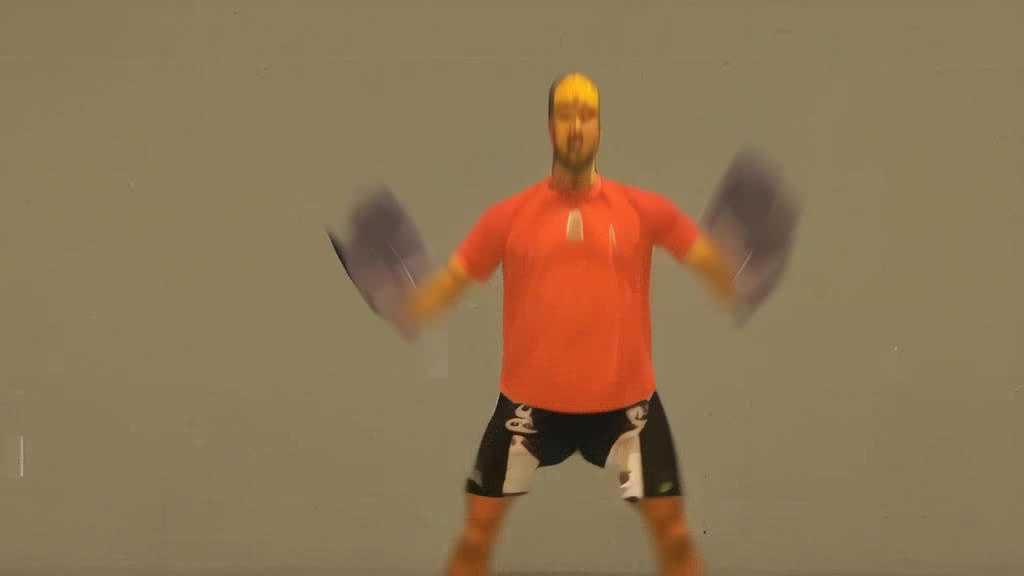}} \raisebox{-.5\height}{\includegraphics[width=0.08\textwidth]{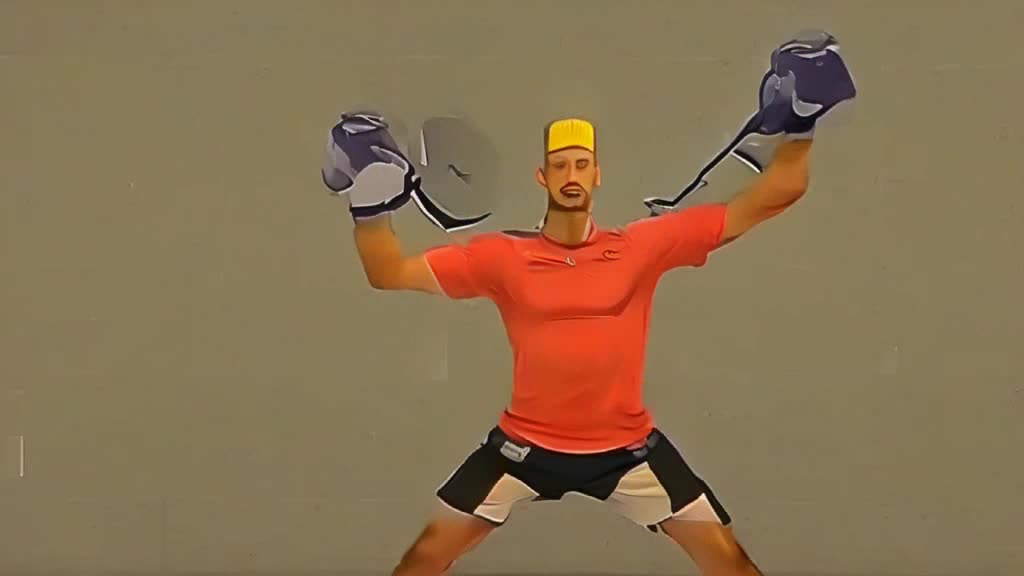}} & \raisebox{-.5\height}{\includegraphics[width=0.08\textwidth]{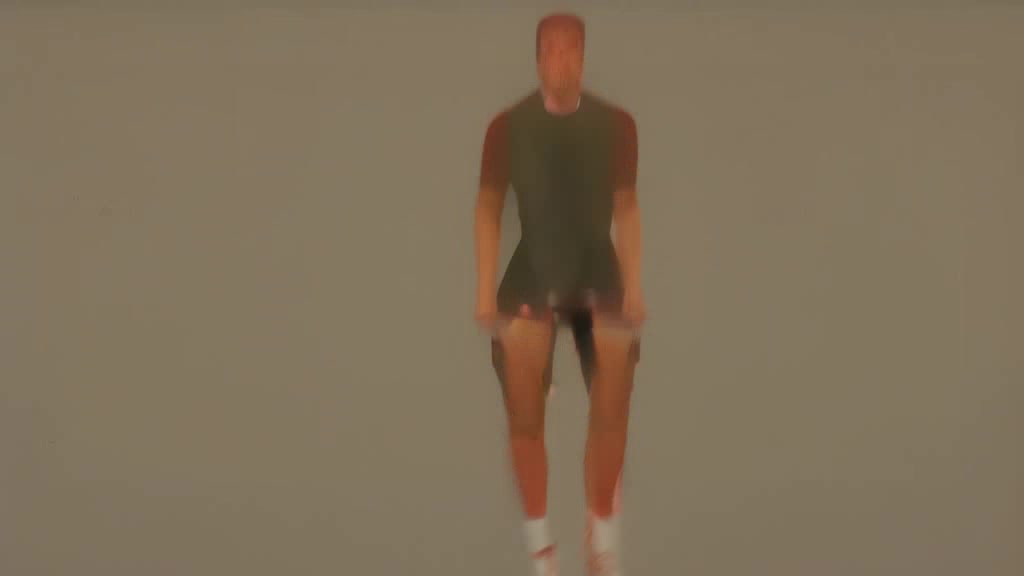}} \raisebox{-.5\height}{\includegraphics[width=0.08\textwidth]{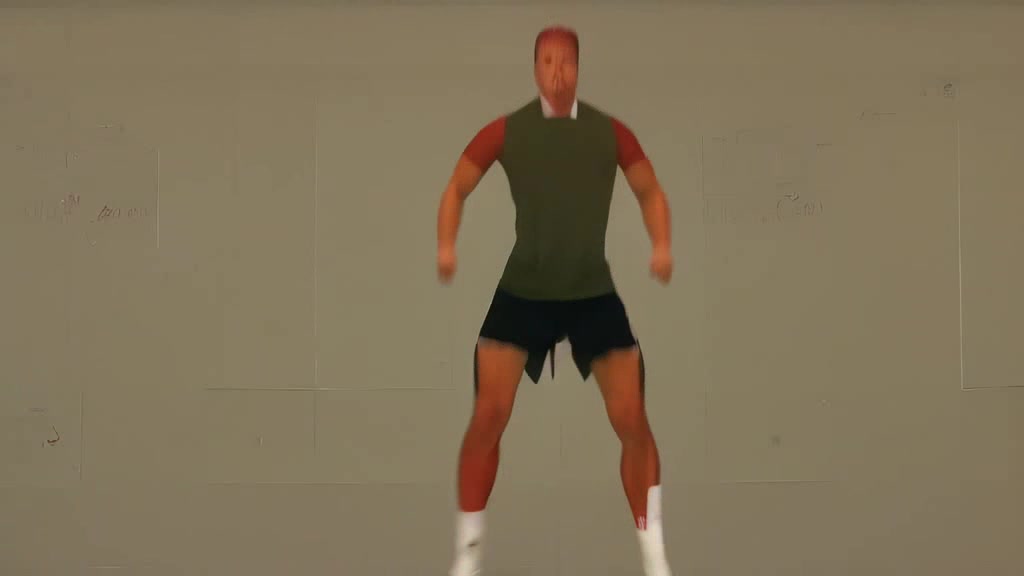}} \raisebox{-.5\height}{\includegraphics[width=0.08\textwidth]{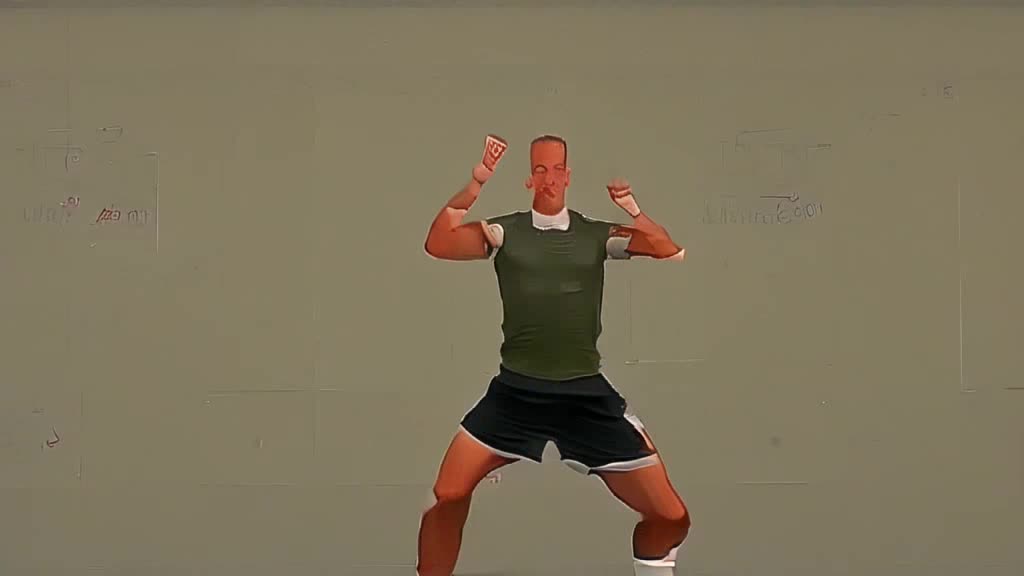}} \\
		1000 & \raisebox{-.5\height}{\includegraphics[width=0.08\textwidth]{figures/motion_vis/first_frame.jpg}} & \raisebox{-.5\height}{\includegraphics[width=0.08\textwidth]{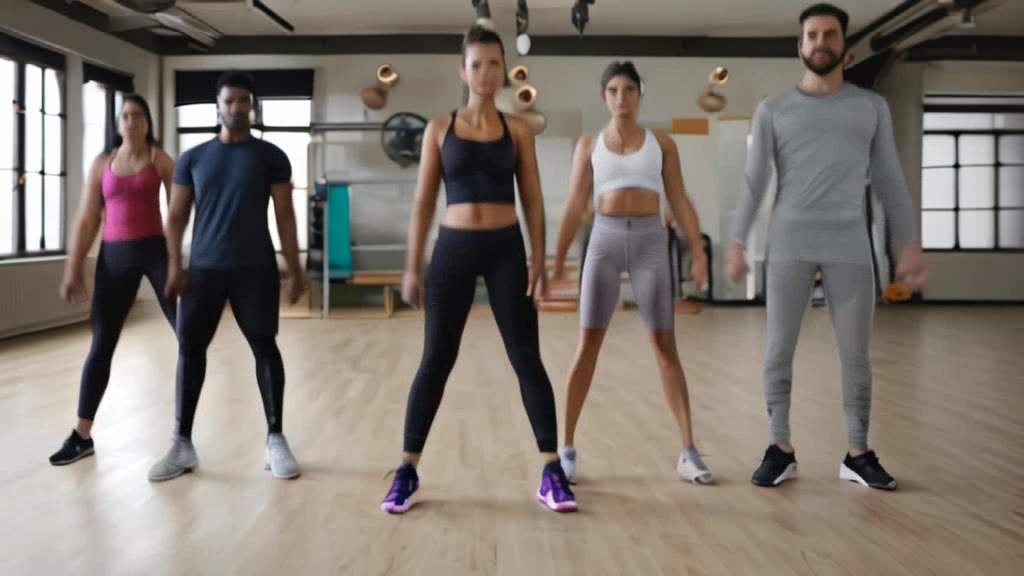}} \raisebox{-.5\height}{\includegraphics[width=0.08\textwidth]{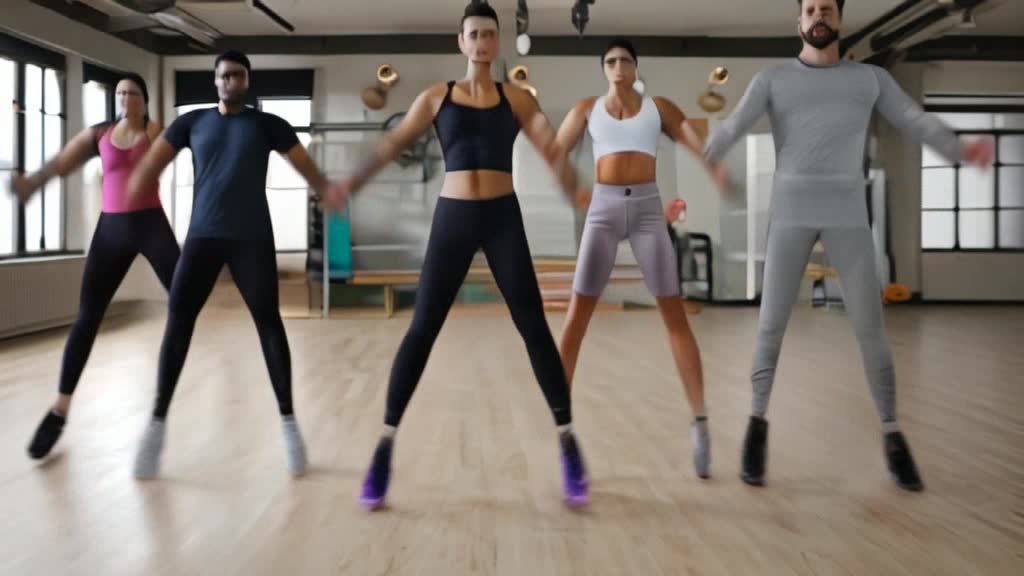}} \raisebox{-.5\height}{\includegraphics[width=0.08\textwidth]{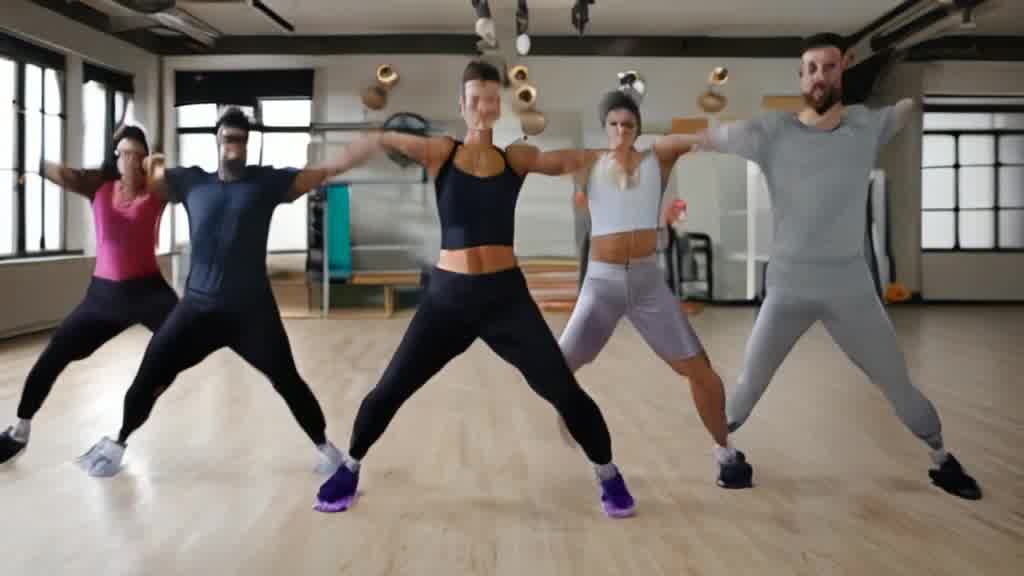}} & \raisebox{-.5\height}{\includegraphics[width=0.08\textwidth]{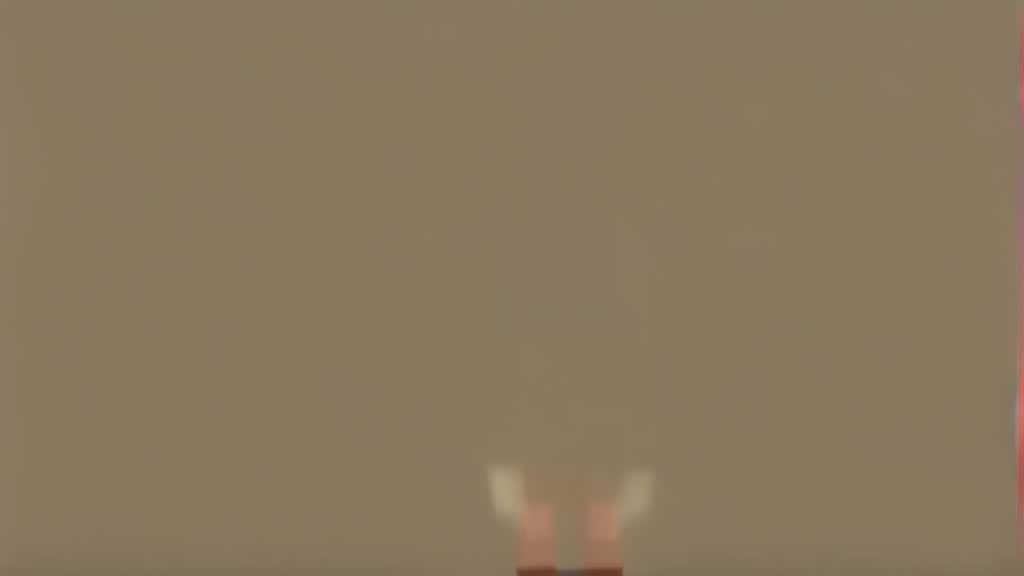}} \raisebox{-.5\height}{\includegraphics[width=0.08\textwidth]{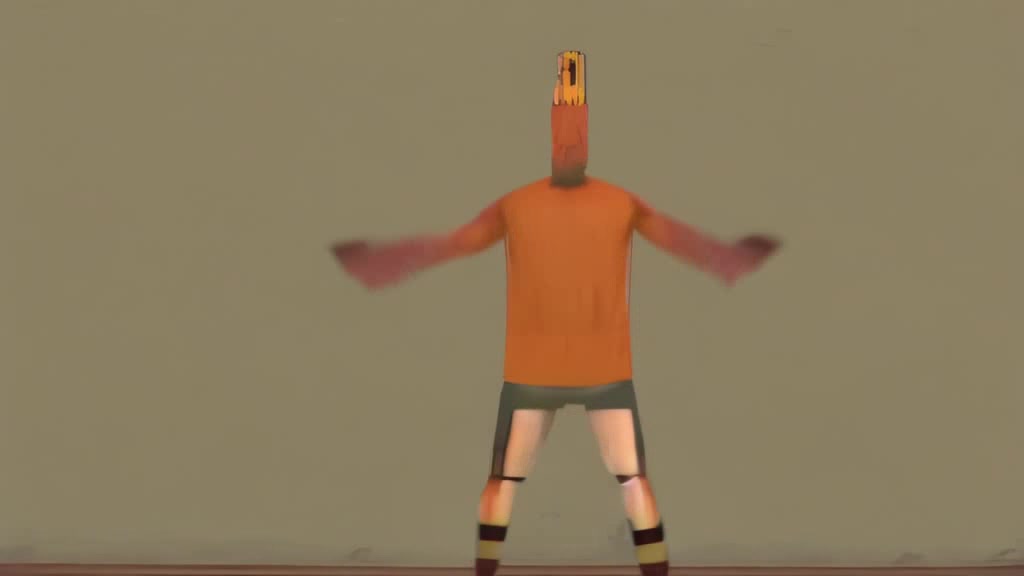}} \raisebox{-.5\height}{\includegraphics[width=0.08\textwidth]{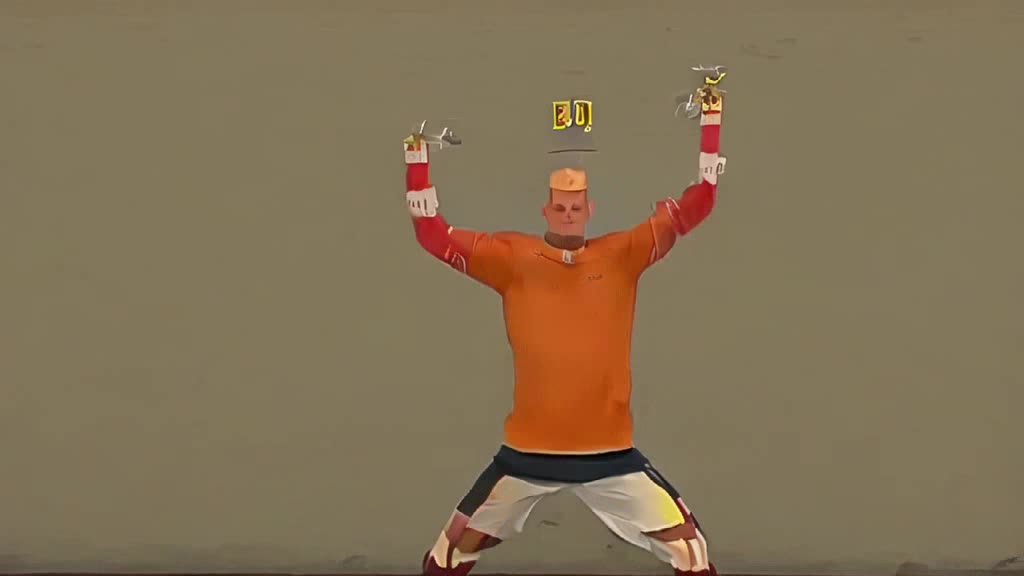}} & \raisebox{-.5\height}{\includegraphics[width=0.08\textwidth]{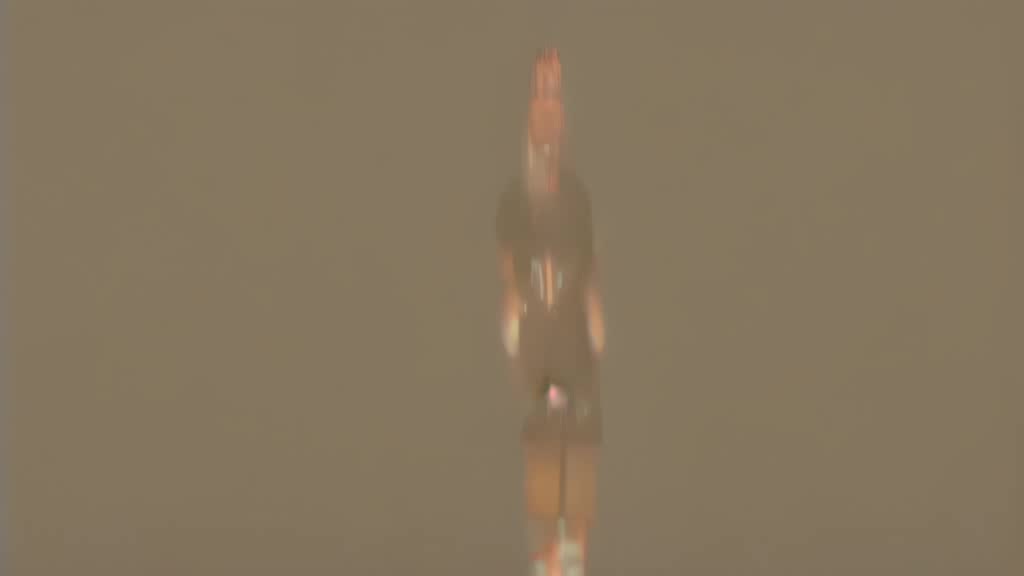}} \raisebox{-.5\height}{\includegraphics[width=0.08\textwidth]{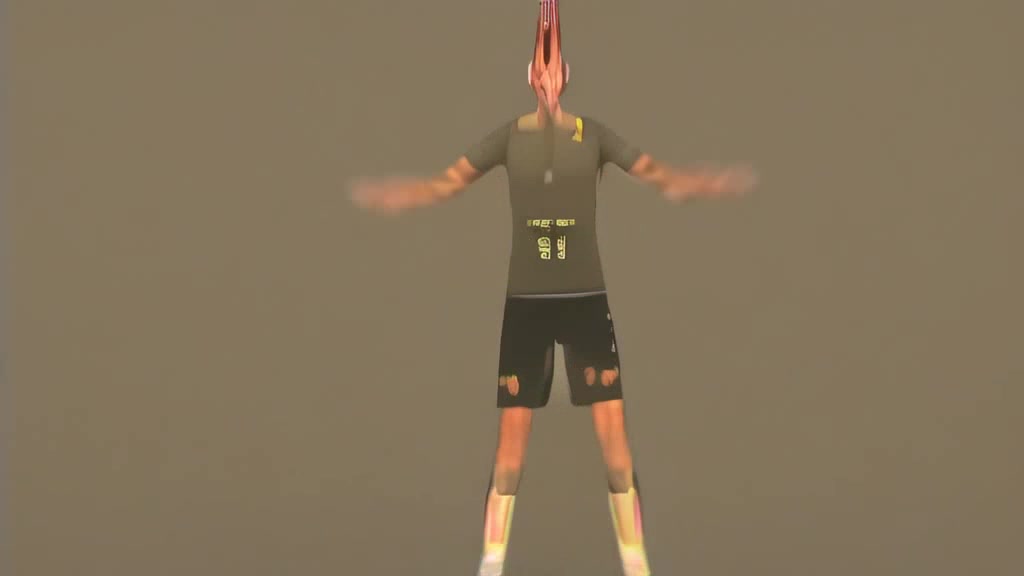}} \raisebox{-.5\height}{\includegraphics[width=0.08\textwidth]{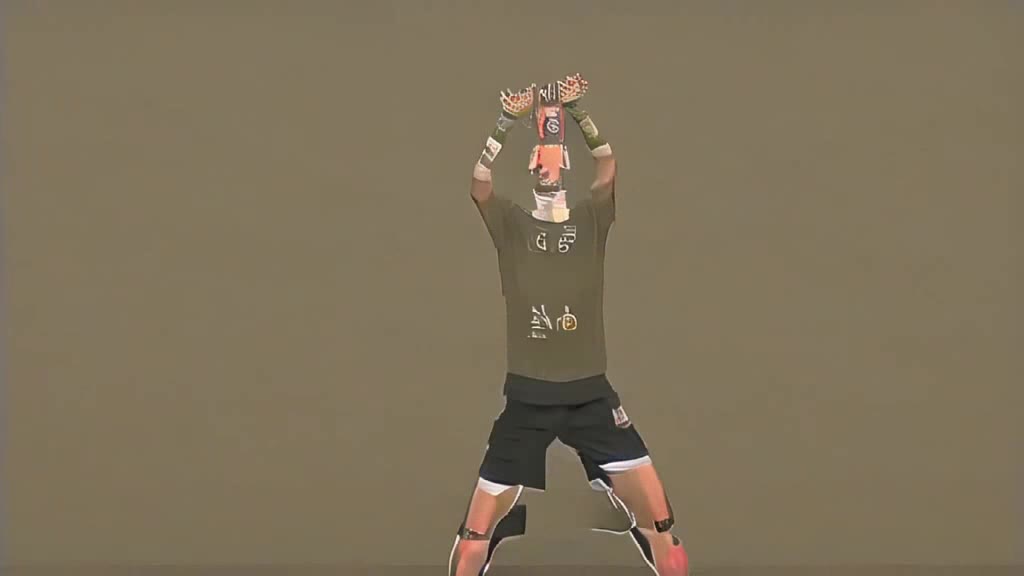}} \\
		2000 & \raisebox{-.5\height}{\includegraphics[width=0.08\textwidth]{figures/motion_vis/first_frame.jpg}} & \raisebox{-.5\height}{\includegraphics[width=0.08\textwidth]{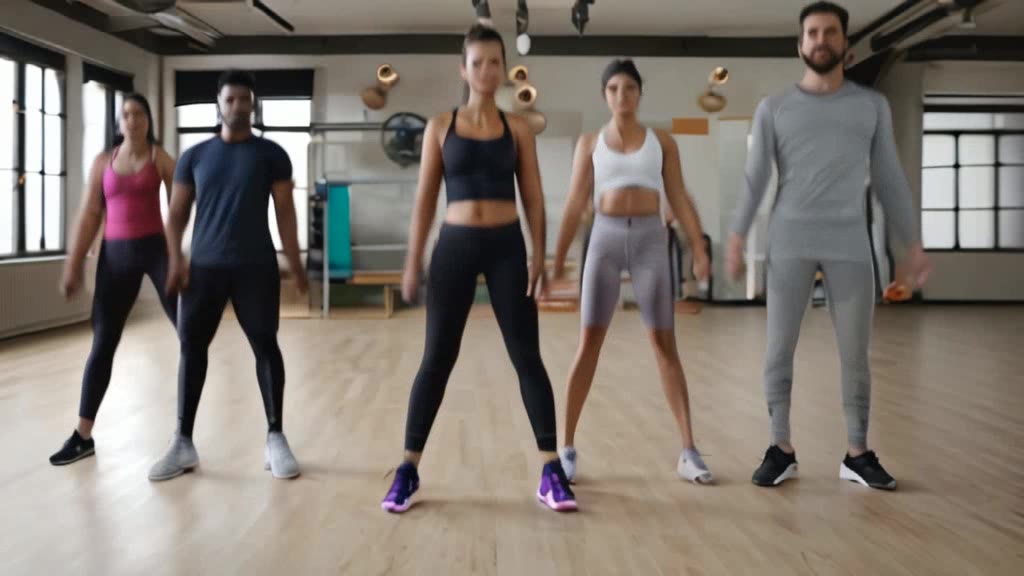}} \raisebox{-.5\height}{\includegraphics[width=0.08\textwidth]{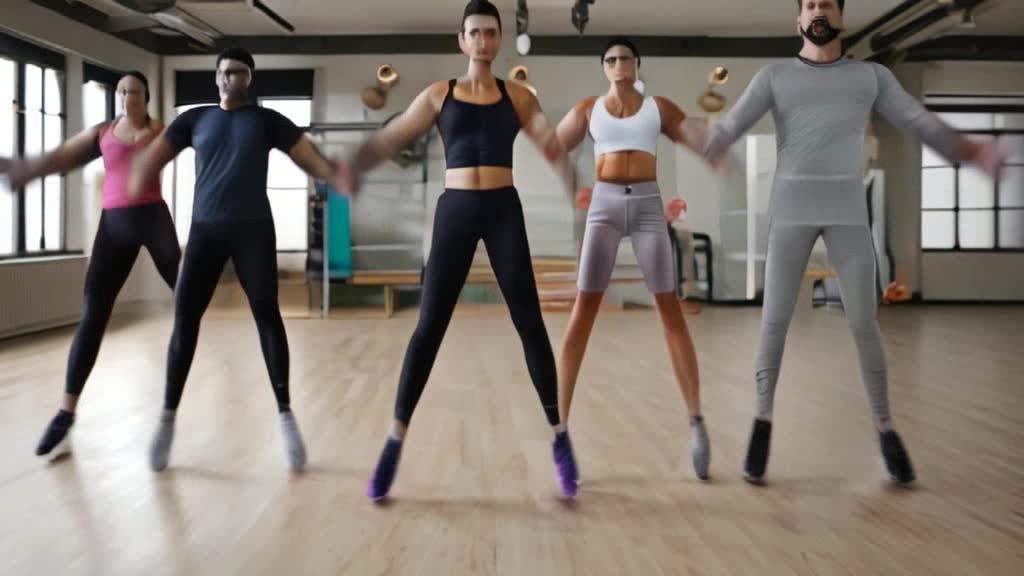}} \raisebox{-.5\height}{\includegraphics[width=0.08\textwidth]{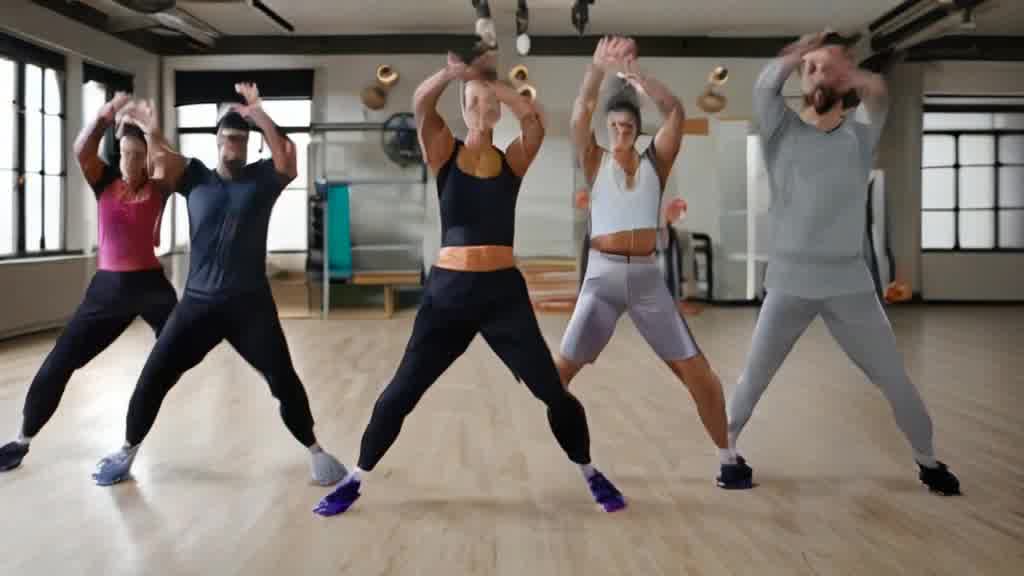}} & \raisebox{-.5\height}{\includegraphics[width=0.08\textwidth]{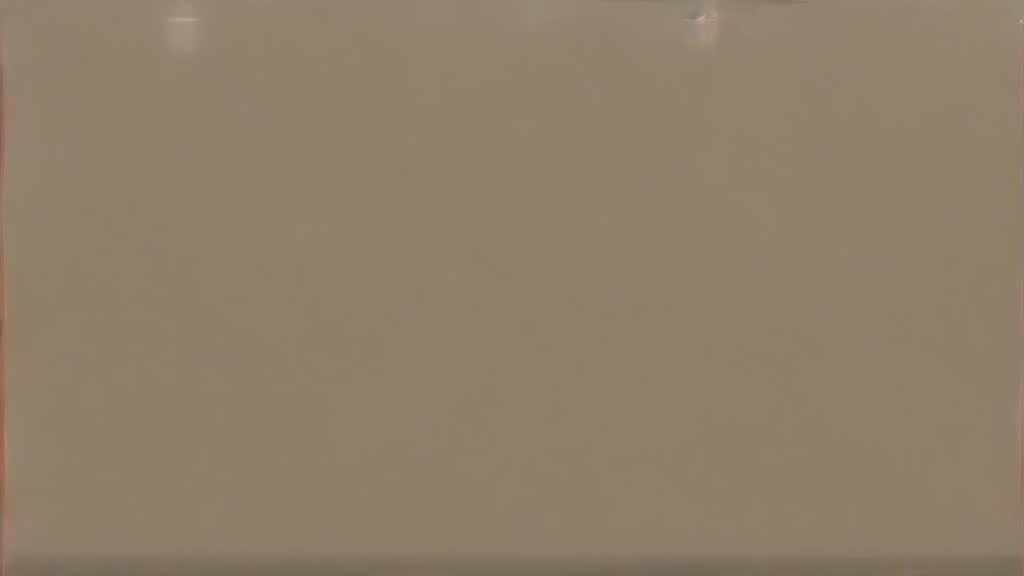}} \raisebox{-.5\height}{\includegraphics[width=0.08\textwidth]{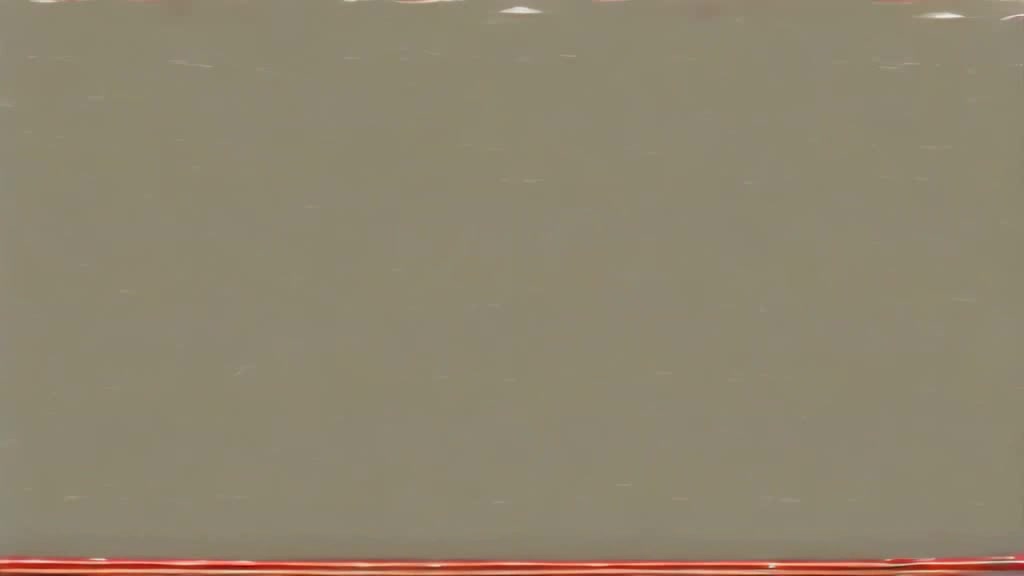}} \raisebox{-.5\height}{\includegraphics[width=0.08\textwidth]{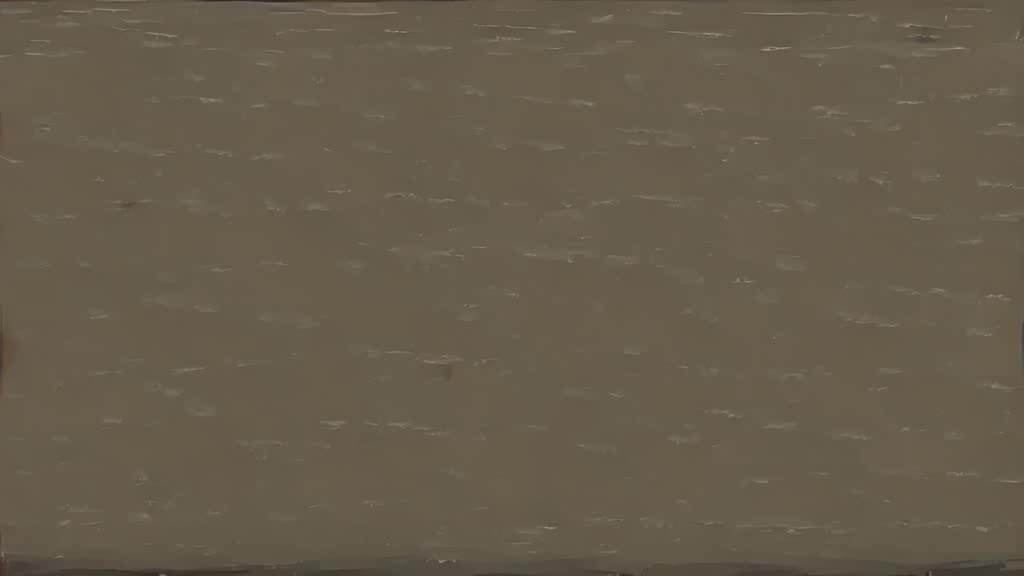}} & \raisebox{-.5\height}{\includegraphics[width=0.08\textwidth]{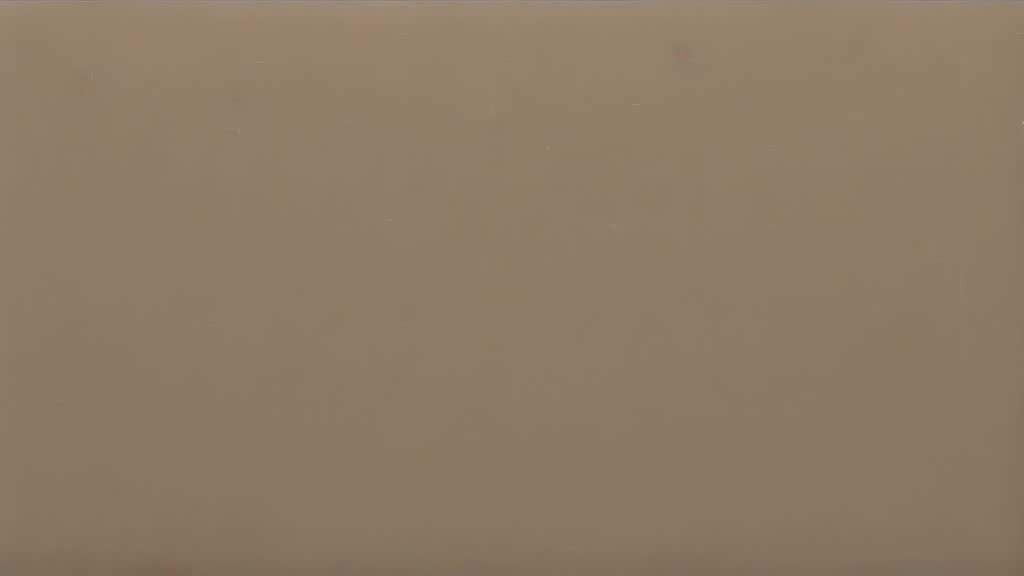}} \raisebox{-.5\height}{\includegraphics[width=0.08\textwidth]{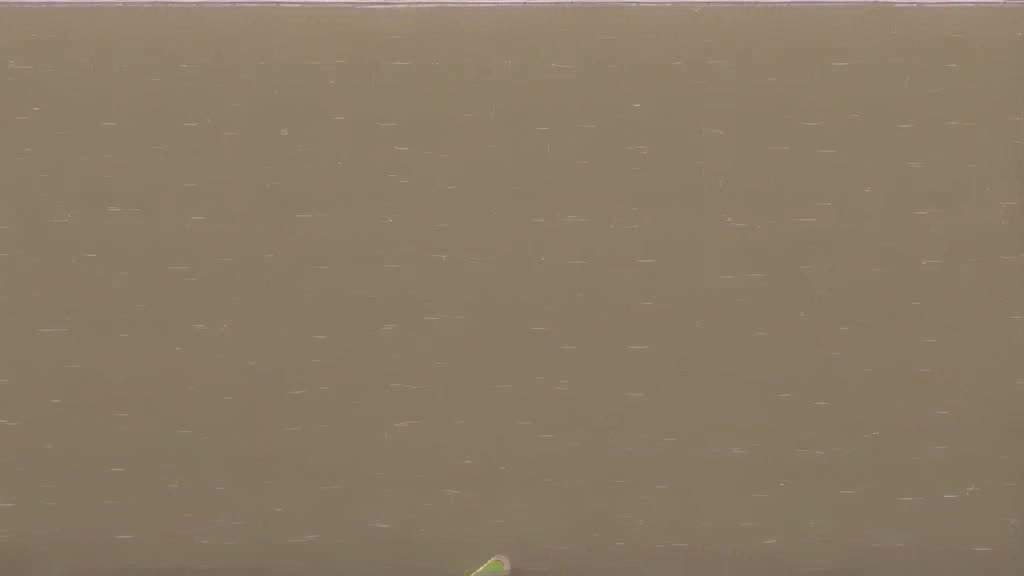}} \raisebox{-.5\height}{\includegraphics[width=0.08\textwidth]{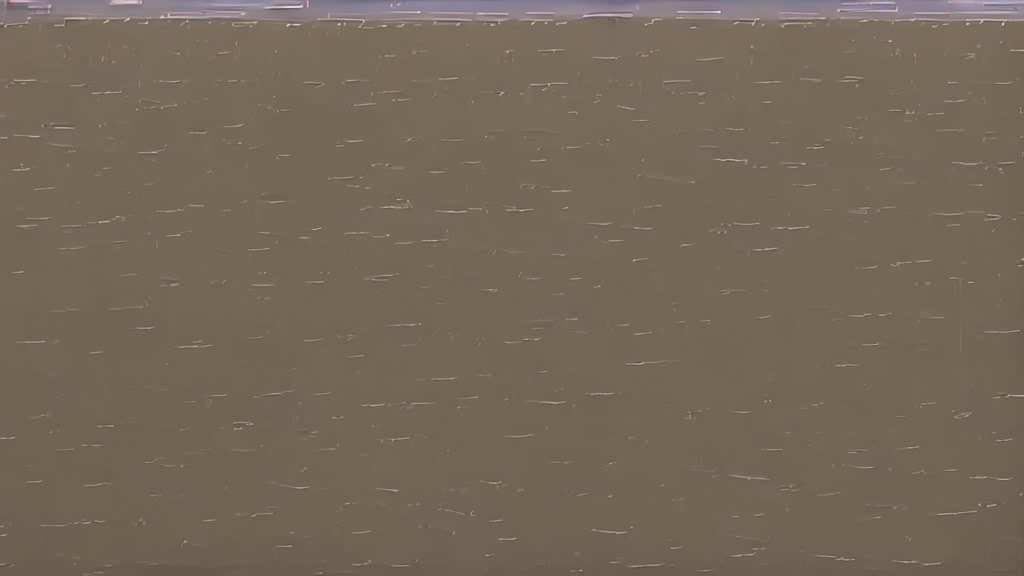}}
	\end{tblr}
	\caption{Motion visualization. We generate videos using our optimized motion-text embedding for a ``jumping jacks'' motion (reference from Fig.~\ref{fig:comp_svd}) both with the image input (conditional) and without (unconditional) after a different number of optimization iterations. Note how the appearance of the unconditional generations differs from the motion reference video and varies with different seeds. Further observe that our method effectively generates similar semantic motions without needing or enforcing spatial alignment.
		\vspace{-2mm}
	}
	\Description{Figure showing three sets of videos for a learned ``jumping jacks'' motion: one conditional video with multiple people on the left and two unconditional videos (different seeds) on the right. There are five videos for each of the above settings: for $0$, $200$, $500$, $1000$, and $2000$ optimization iterations. In general, as we optimize the motion-text embedding for more iterations, the resulting videos show the ``jumping jacks'' motion better. For the unconditional generations, i.e., the motion visualization, the videos become very abstract at $2000$ optimization iterations, whereas the conditional generation at the same number of iterations remains good.}
	\label{fig:motion_vis}
\end{figure*}

Fig.~\ref{fig:motion_vis} shows motion visualizations of our motion-text embedding for a ``jumping jacks'' motion after different numbers of optimization iterations and the generated videos for a given target image side-by-side. Starting around iteration $500$, a person doing a ``jumping jacks'' motion can be seen in the visualizations. Beyond $1000$ iterations, the motion visualizations become more abstract, but the generated motions in the conditional case remain of high quality. Notably, the appearance and position of the people do not match those of the motion reference video (from Fig.~\ref{fig:comp_svd}). Furthermore, the position of the people is different in the conditional and unconditional videos, but all videos have a similar semantic motion. This demonstrates that our motion-text embedding neither encodes the appearance nor the exact spatial positioning of the objects extensively, likely for reasons described in Section~\ref{sec:method_motivation}.

\section{Applicability to Other Video Diffusion Models}

We believe our approach should generalize to other architectures, including ones based on transformers, as long as the image-to-video model mainly extracts appearance from the image input and motion from text/image embeddings. This appears to hold for HunyuanVideo-I2V~\cite{hunyuanvideo}; when we repeated the experiment from Fig.~\ref{fig:i2v_ignores_text}, the horse remained white despite the text input specifying a ``pink'' horse. For video models with full spatio-temporal attention (e.g., HunyuanVideo-I2V), rather than SVD’s separate spatial and temporal attention, it remains to be investigated whether inflating the motion-text embedding to have different tokens per frame is strictly necessary for good performance, as it was for SVD.

\section{Additional Evaluation}

\subsection{Additional Information for the Compared Methods} \label{sec:additional-information-compared-methods}

\subsubsection{Choice of Compared Methods}

To the best of our knowledge, our method is the first to tackle the general motion transfer task in the image-to-video setting. As a result, there are no direct competitor methods. Instead, we evaluate the most closely related general methods, (which were originally designed for slightly different tasks) on our problem. We considered the three most similar classes of methodology and compared our method with a representative of each class:
\begin{enumerate}
	\item Image-to-video model with explicit, dense motion representation: VideoComposer~\cite{videocomposer}
	\item Image-to-video model with implicit motion representation: MotionClone~\cite{motionclone} (our method falls into this category)
	\item Text-to-video model with implicit motion representation: MotionDirector~\cite{motiondirector}
\end{enumerate}

Methods within each class tend to have certain inherent drawbacks in common. Specifically, methods based on explicit, dense motion representations (class (1)) transfer spatial but not semantic motion and may leak the reference video's structure; and methods based on text-to-video models (class (3)) do not directly take a target image input, compromising the preservation of the target's appearance and layout. We believe that comparing to one method from each class is sufficient to demonstrate the types of artifacts, as adding more methods would not address the inherent limitations shared within the class.

Additional practical considerations: The following related methods did not have corresponding code publicly available at the time of writing: Diffusion as Shader~\cite{diffusion_as_shader} (class (1)), Go-With-The-Flow~\cite{go_with_the_flow} (class (1)), GenVideo~\cite{genvideo}, and CustomTTT~\cite{CustomTTT} (class (3)). The following methods are computationally infeasible given the size of our evaluation data set and our computational resources available: LAMP~\cite{lamp} (class (2), $\approx 14$ GPU hours per reference video), and DreamVideo~\cite{dreamvideo} (class (3), $\approx 1$ GPU hours per motion reference video and $\approx 2$ GPU hours per target image).

Furthermore, we do not compare to methods using explicit, sparse motion representations (see Section~\ref{sec:related_work_extra-sparse}) because it is unclear how to automatically extract sparse motion inputs from motion reference video. We also do not compare to methods based on text-to-video models without learned appearance~\cite{motioncrafter, wang2024motion, space_time_diffusion, materzynska2024newmove} because defining appearance solely through text is insufficient to accurately preserve the target image appearance.

\subsubsection{Implementation Details}

We used the official implementations for all compared methods and followed their installation and usage instructions closely. For the methods requiring a text input, we manually captioned images and videos for the qualitative evaluation. We initially tried several image and video captioning methods, but their captions all led to worse results than manual captions that follow the captions used in the papers more closely. For the quantitative evaluation, we used the corresponding caption from the Something-Something V2 data set~\cite{something_something}.

\subsection{Additional Qualitative Comparisons to Baseline} \label{sec:additional-qual-comp-baseline}

\begin{figure*}[htbp]
	\centering
	\begin{tblr}{
			colspec = {l c c c c},
			cell{1}{2} = {c=2}{c},
			cell{1}{4} = {c=2}{c},
			vline{3} = {3-8,10-15,17-22}{dashed},
			vline{4} = {1-22}{},
			vline{5} = {3-8,10-15,17-22}{dashed},
			hline{9} = {1-5}{},
			hline{16} = {1-5}{},
		}
		& SVD & & Ours & \\
		{Ref.} & \raisebox{-.5\height}{\includegraphics[width=0.075\textwidth]{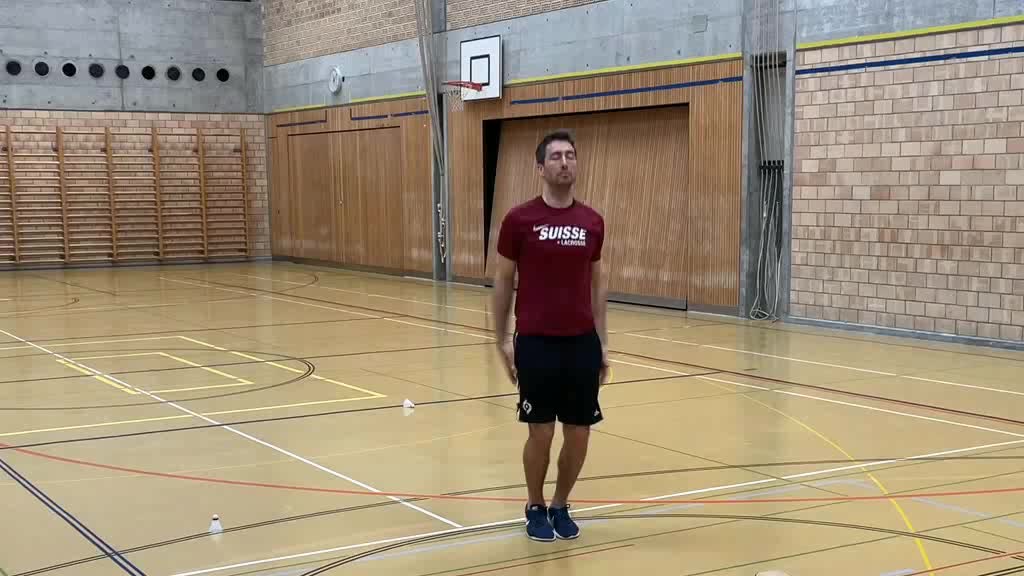}} & \raisebox{-.5\height}{\includegraphics[width=0.075\textwidth]{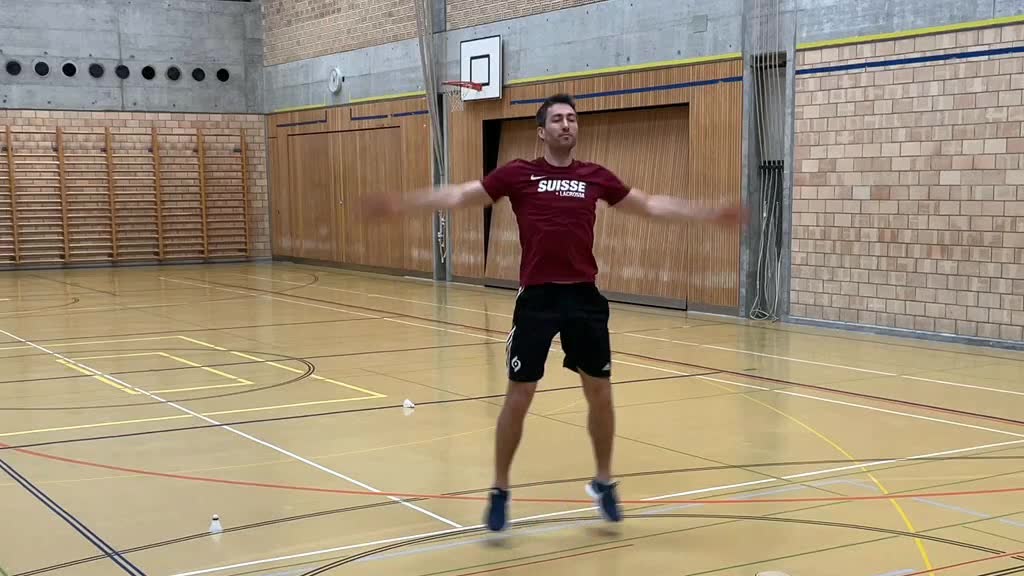}} \raisebox{-.5\height}{\includegraphics[width=0.075\textwidth]{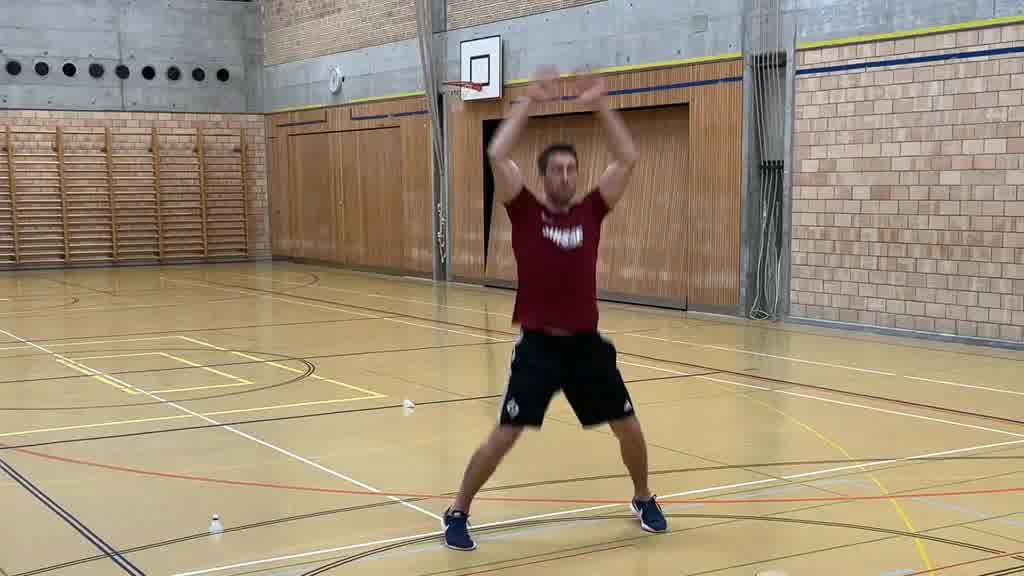}} \raisebox{-.5\height}{\includegraphics[width=0.075\textwidth]{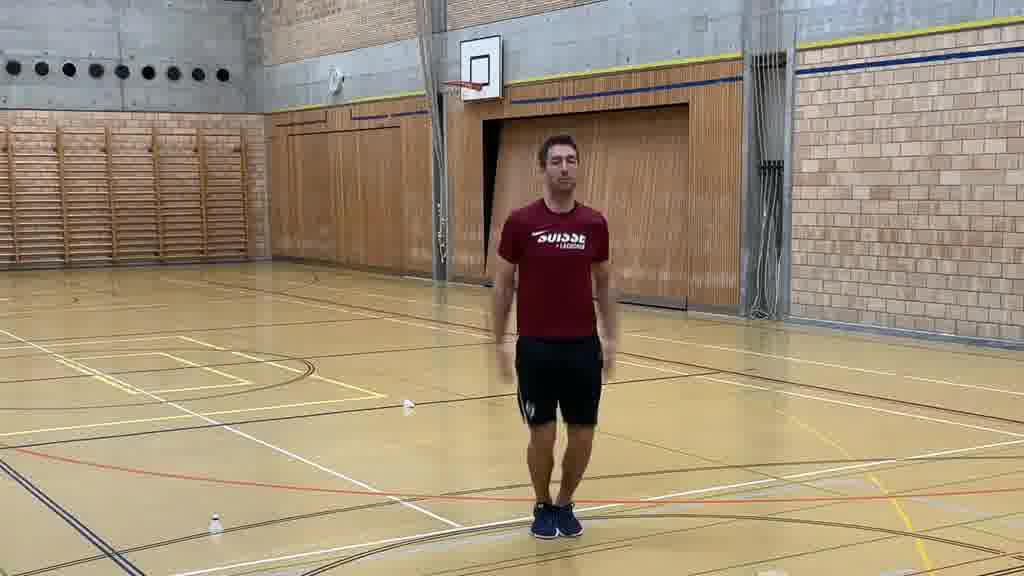}} & \raisebox{-.5\height}{\includegraphics[width=0.075\textwidth]{figures/svd_comparison/fullbody_standing/frames_input/001.jpg}} & \raisebox{-.5\height}{\includegraphics[width=0.075\textwidth]{figures/svd_comparison/fullbody_standing/frames_input/005.jpg}} \raisebox{-.5\height}{\includegraphics[width=0.075\textwidth]{figures/svd_comparison/fullbody_standing/frames_input/009.jpg}} \raisebox{-.5\height}{\includegraphics[width=0.075\textwidth]{figures/svd_comparison/fullbody_standing/frames_input/014.jpg}} \\
		{Seed 0} & \raisebox{-.5\height}{\includegraphics[width=0.075\textwidth]{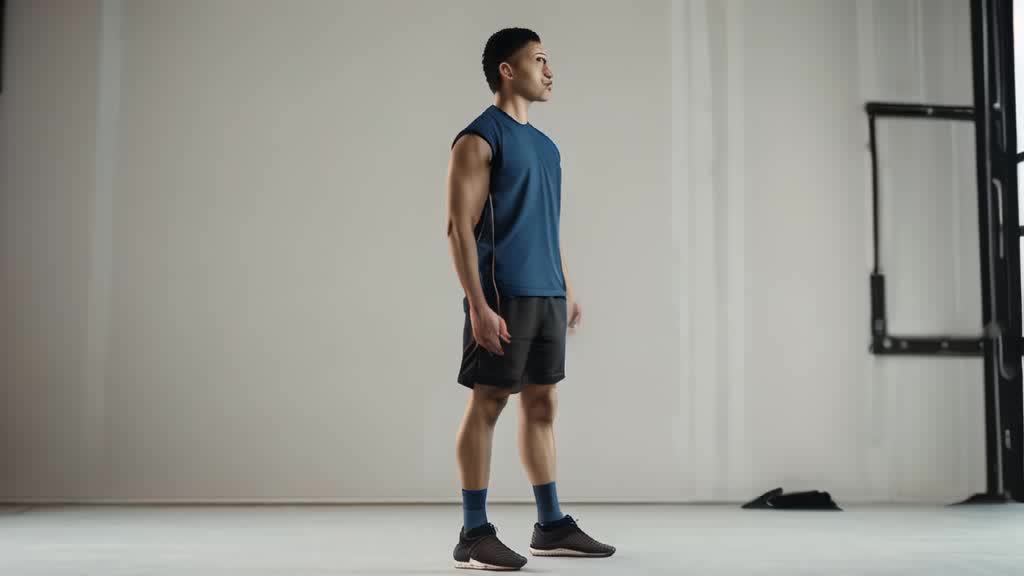}} & \raisebox{-.5\height}{\includegraphics[width=0.075\textwidth]{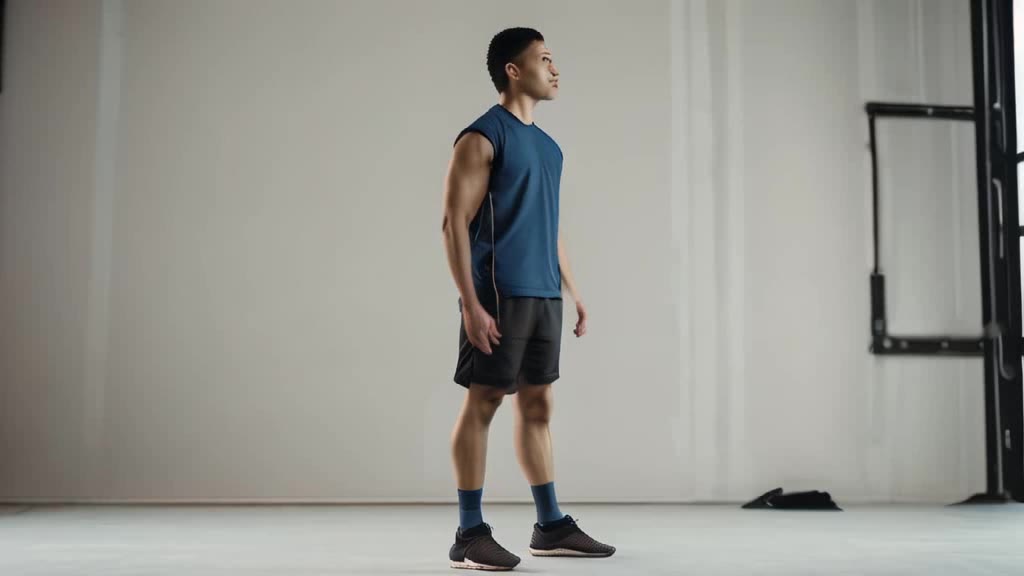}} \raisebox{-.5\height}{\includegraphics[width=0.075\textwidth]{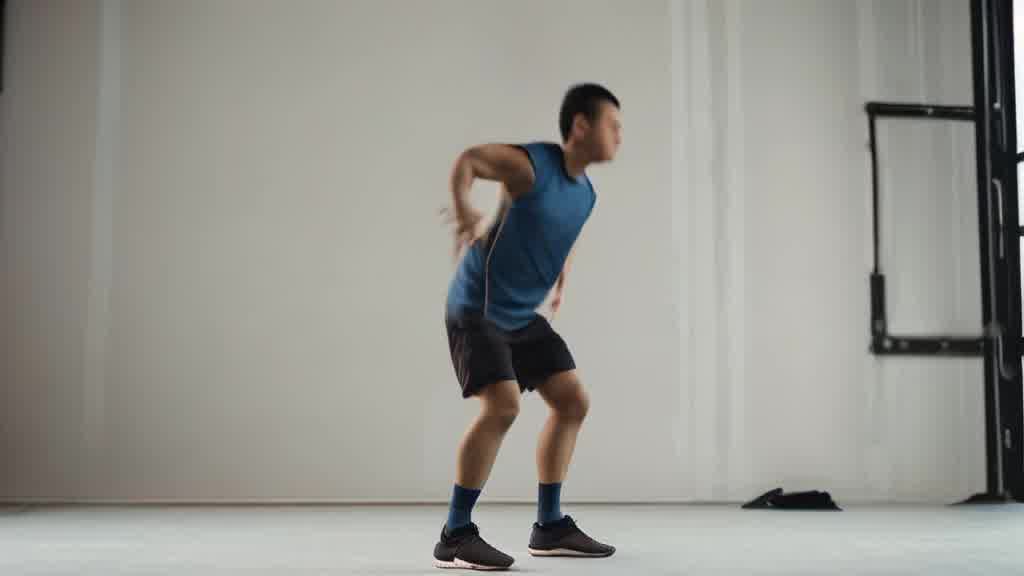}} \raisebox{-.5\height}{\includegraphics[width=0.075\textwidth]{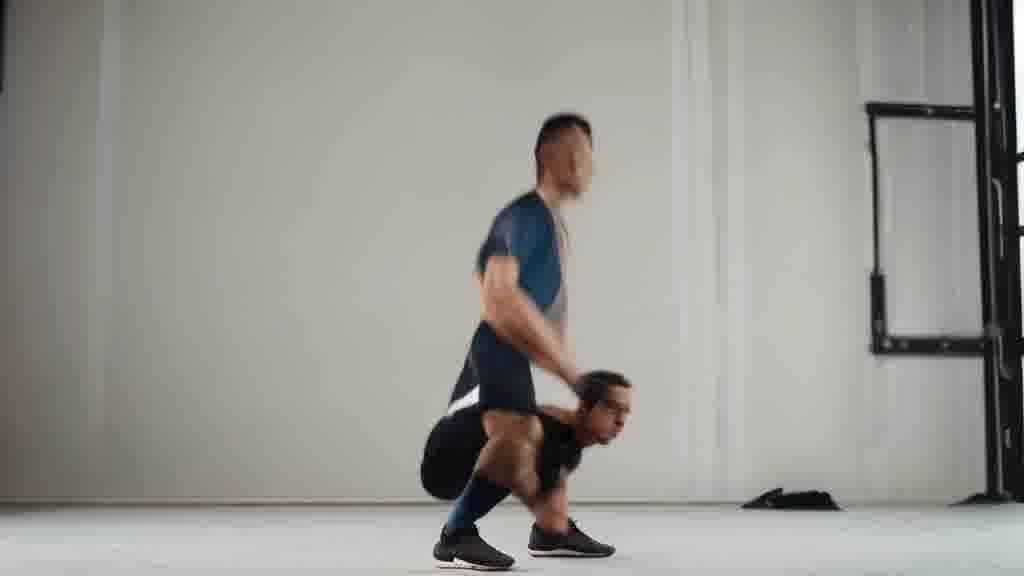}} & \raisebox{-.5\height}{\includegraphics[width=0.075\textwidth]{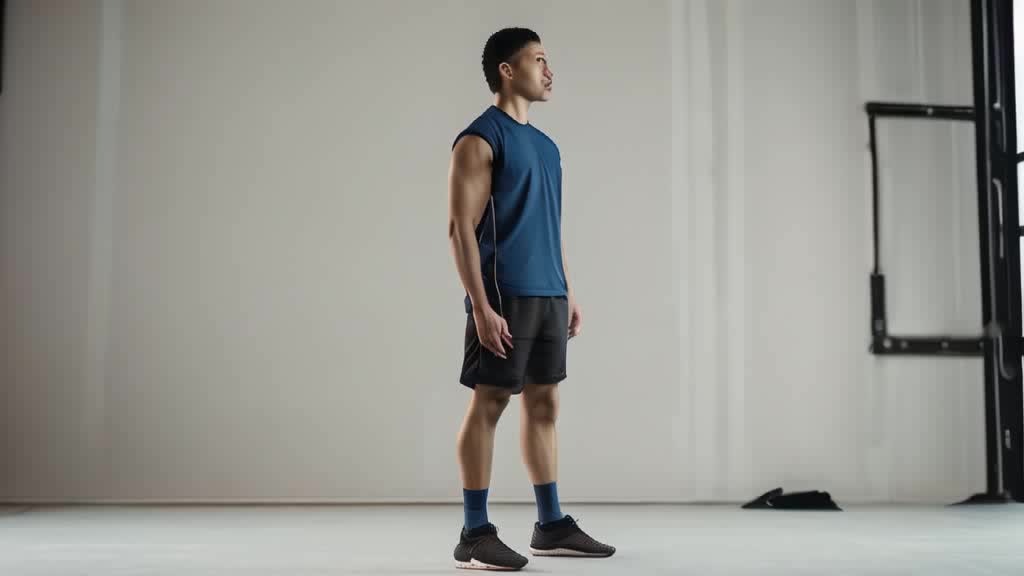}} & \raisebox{-.5\height}{\includegraphics[width=0.075\textwidth]{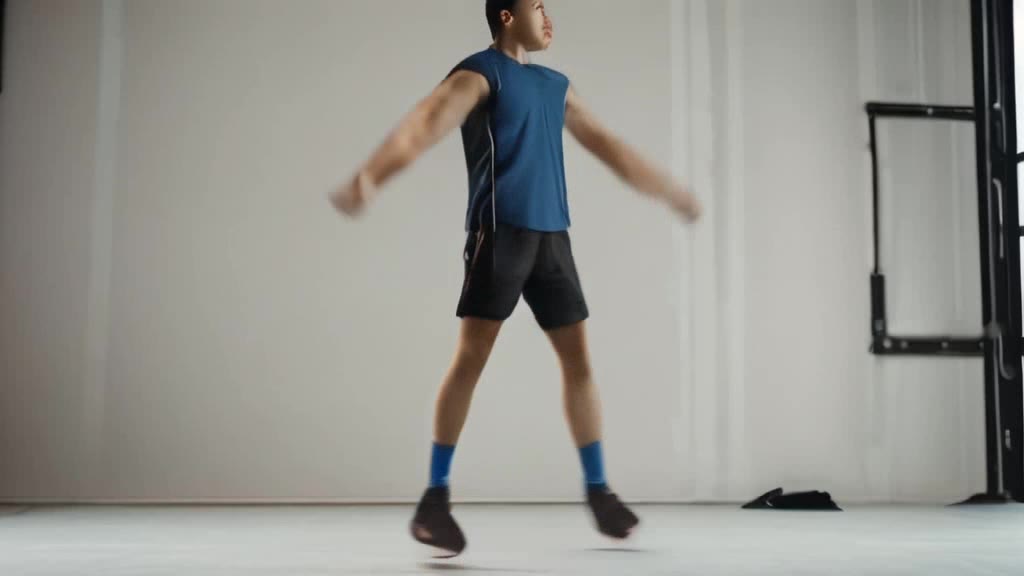}} \raisebox{-.5\height}{\includegraphics[width=0.075\textwidth]{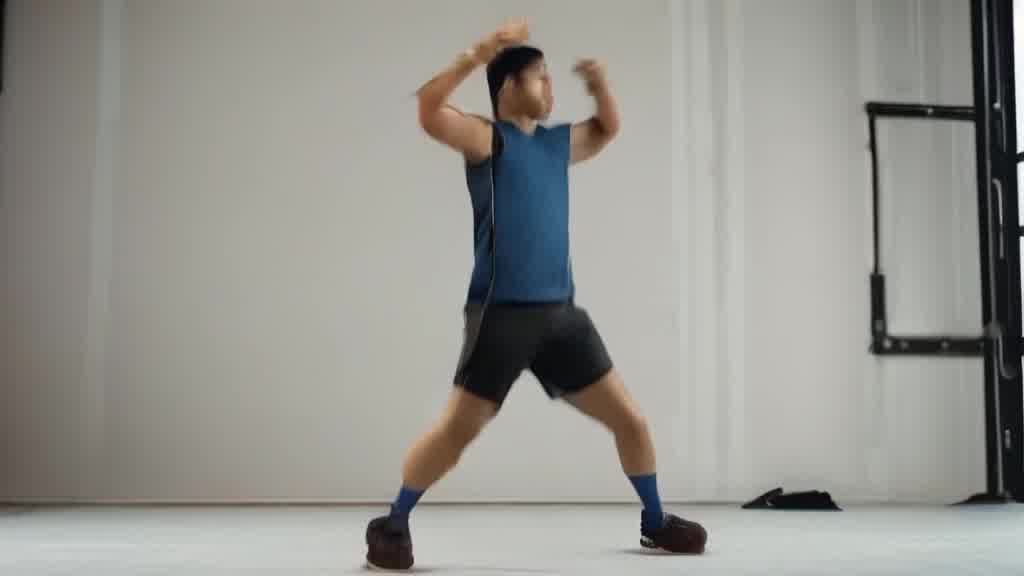}} \raisebox{-.5\height}{\includegraphics[width=0.075\textwidth]{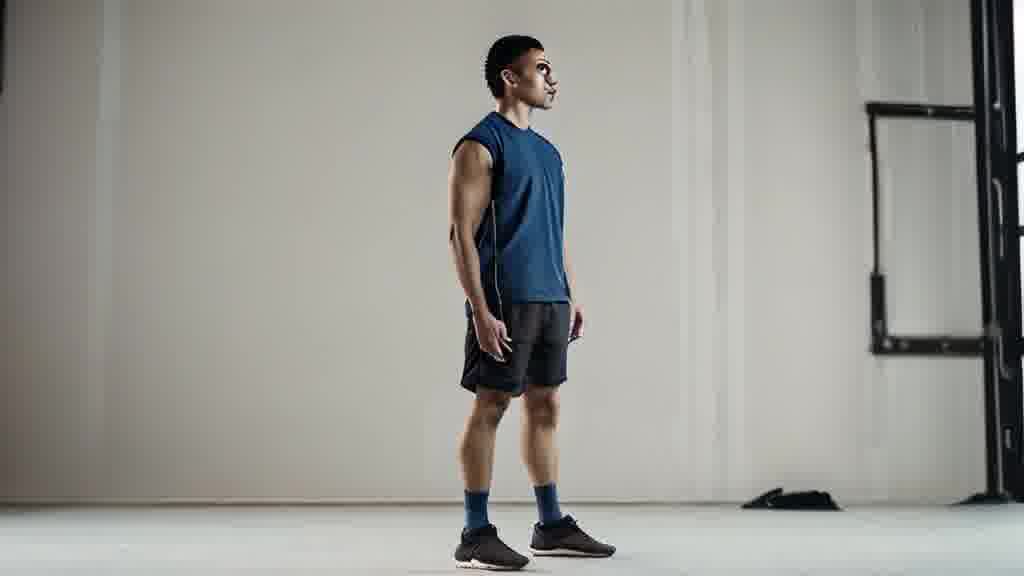}} \\
		{Seed 1} & \raisebox{-.5\height}{\includegraphics[width=0.075\textwidth]{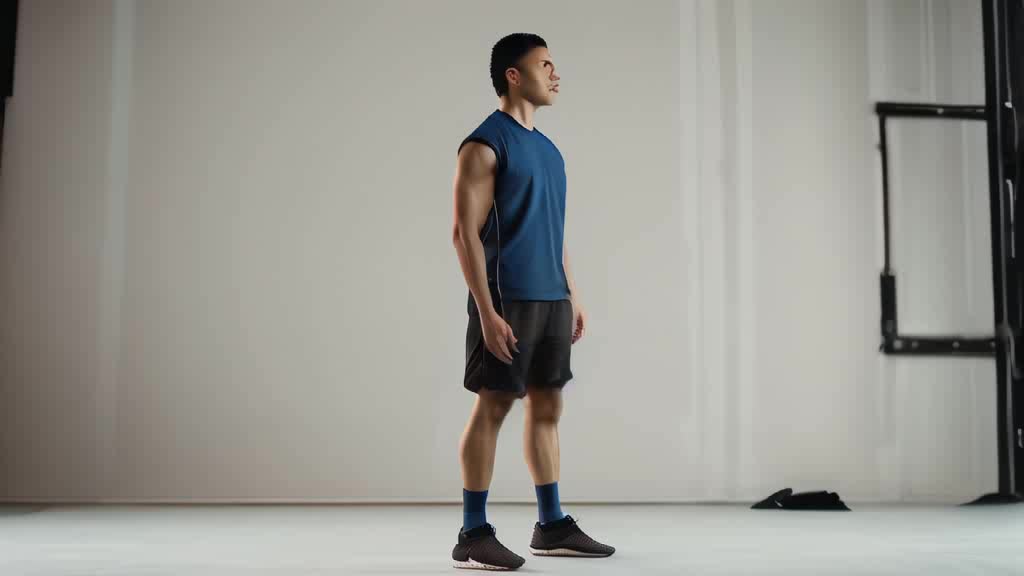}} & \raisebox{-.5\height}{\includegraphics[width=0.075\textwidth]{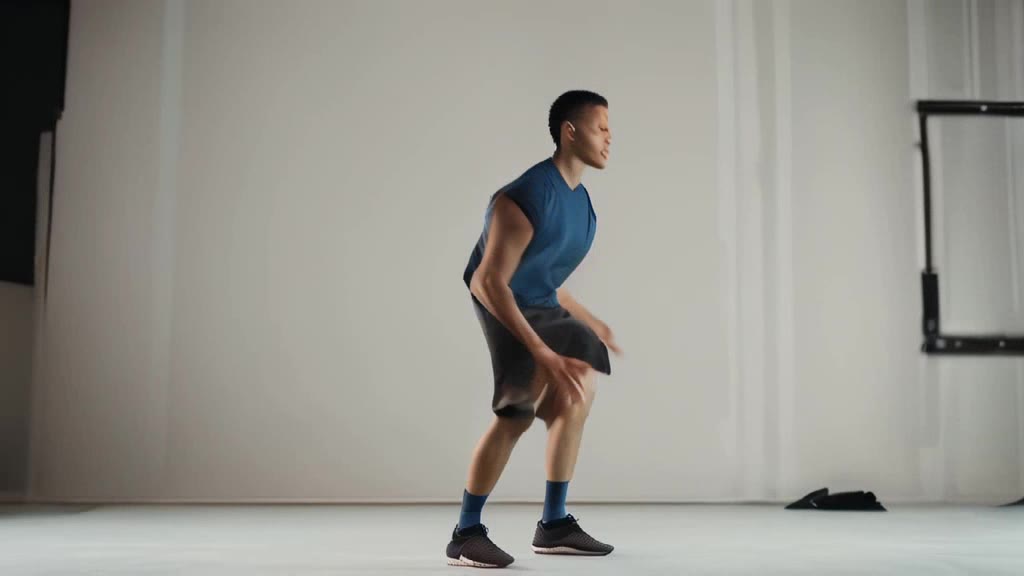}} \raisebox{-.5\height}{\includegraphics[width=0.075\textwidth]{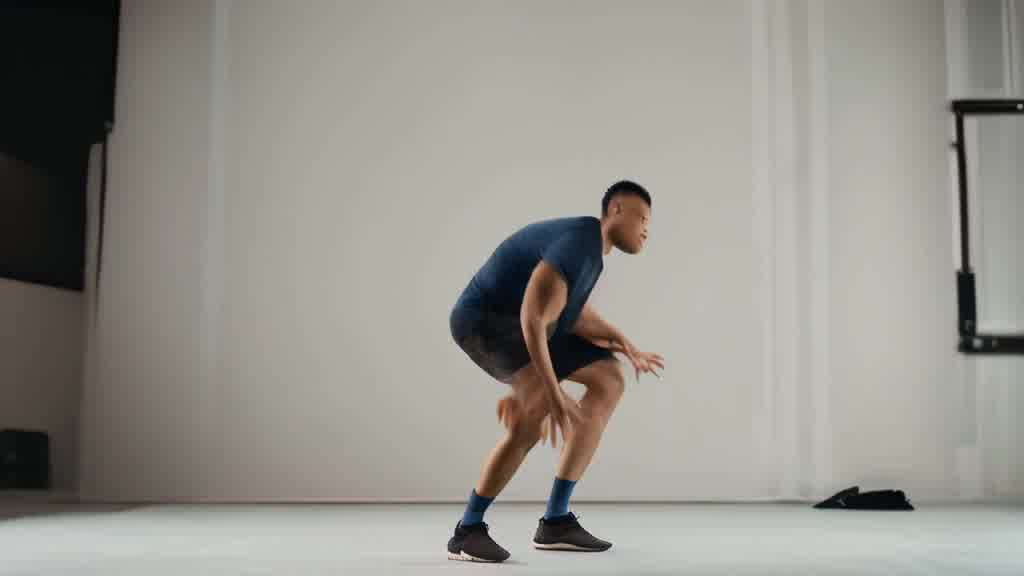}} \raisebox{-.5\height}{\includegraphics[width=0.075\textwidth]{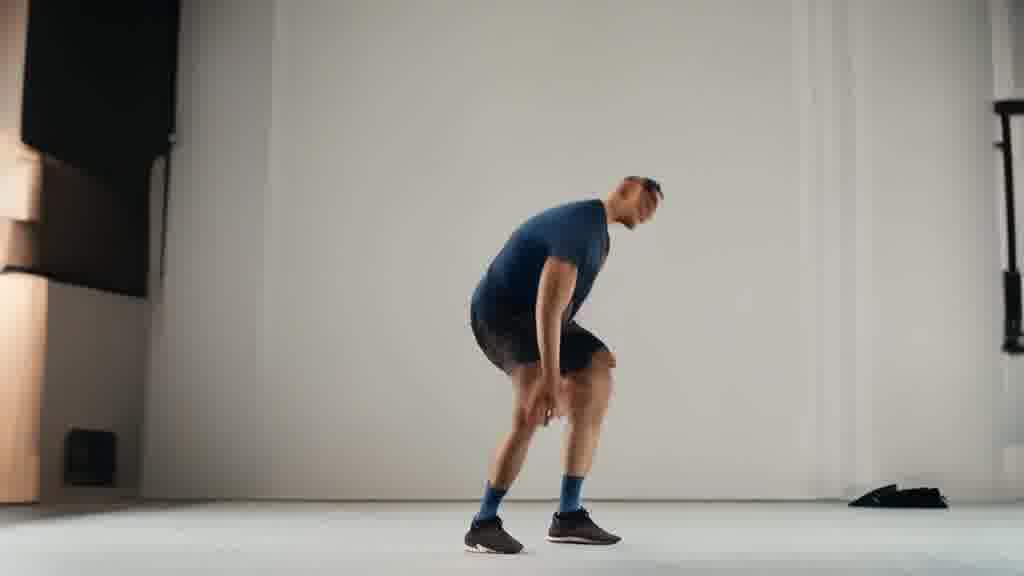}} & \raisebox{-.5\height}{\includegraphics[width=0.075\textwidth]{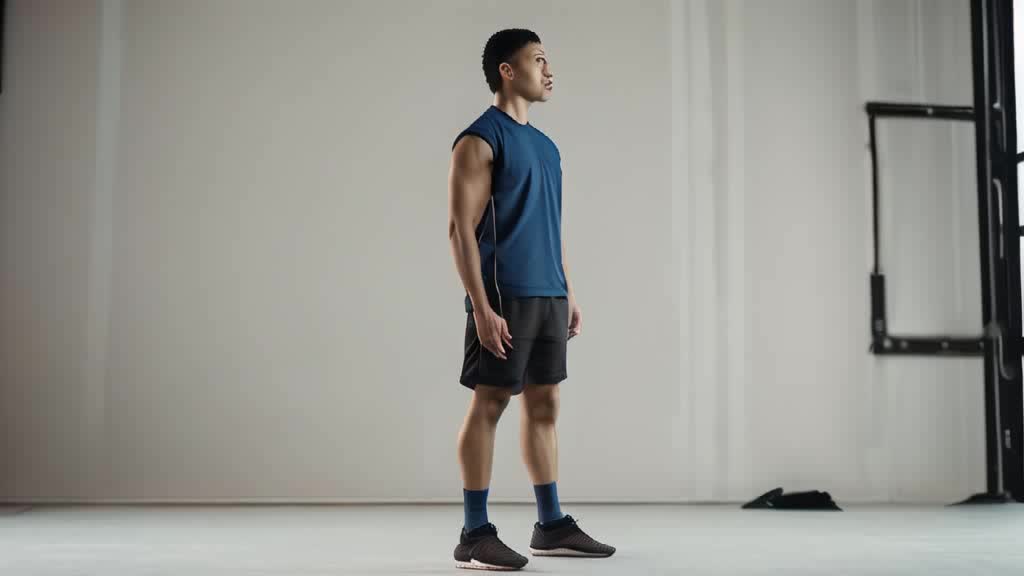}} & \raisebox{-.5\height}{\includegraphics[width=0.075\textwidth]{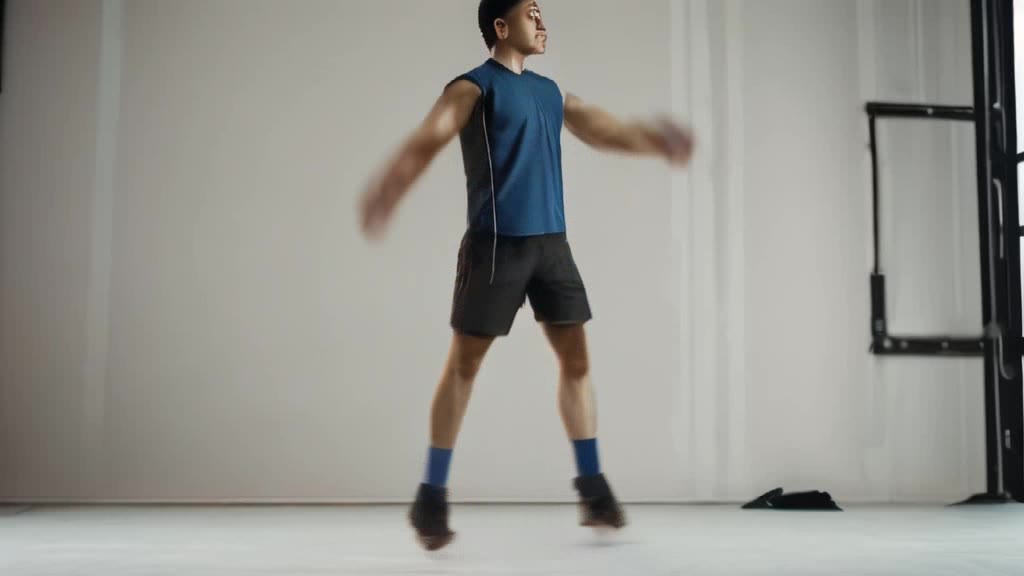}} \raisebox{-.5\height}{\includegraphics[width=0.075\textwidth]{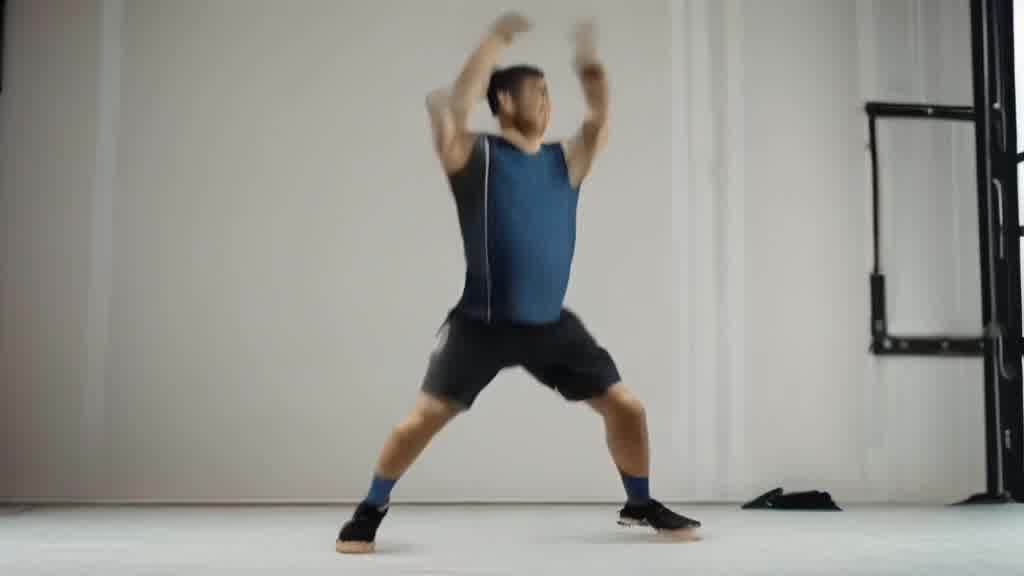}} \raisebox{-.5\height}{\includegraphics[width=0.075\textwidth]{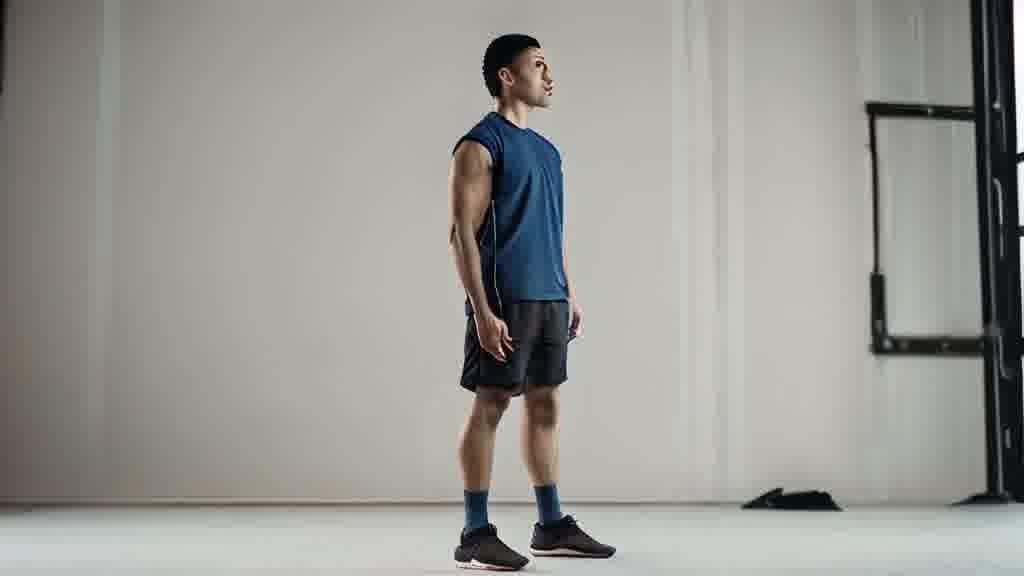}} \\
		{Seed 2} & \raisebox{-.5\height}{\includegraphics[width=0.075\textwidth]{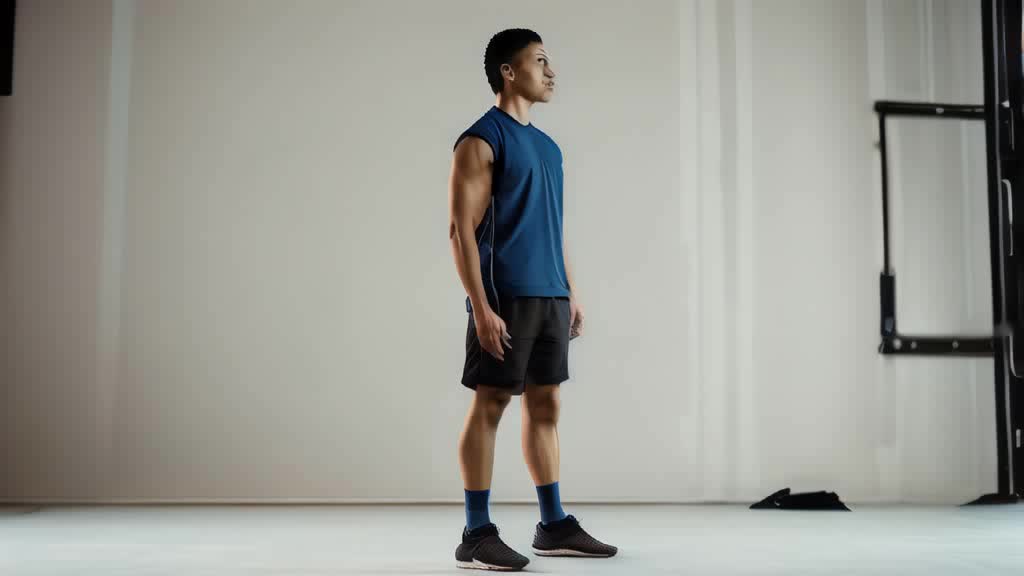}} & \raisebox{-.5\height}{\includegraphics[width=0.075\textwidth]{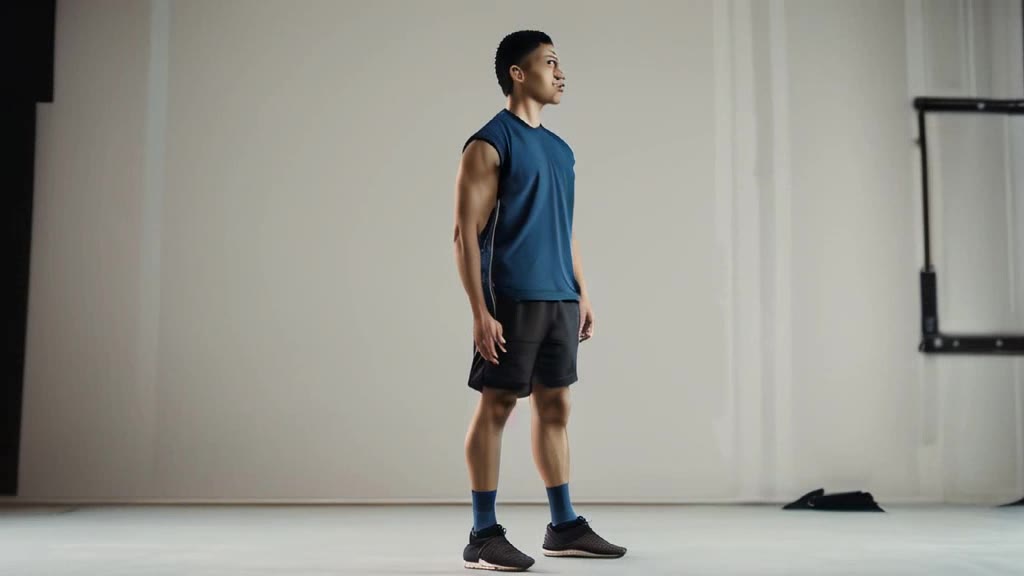}} \raisebox{-.5\height}{\includegraphics[width=0.075\textwidth]{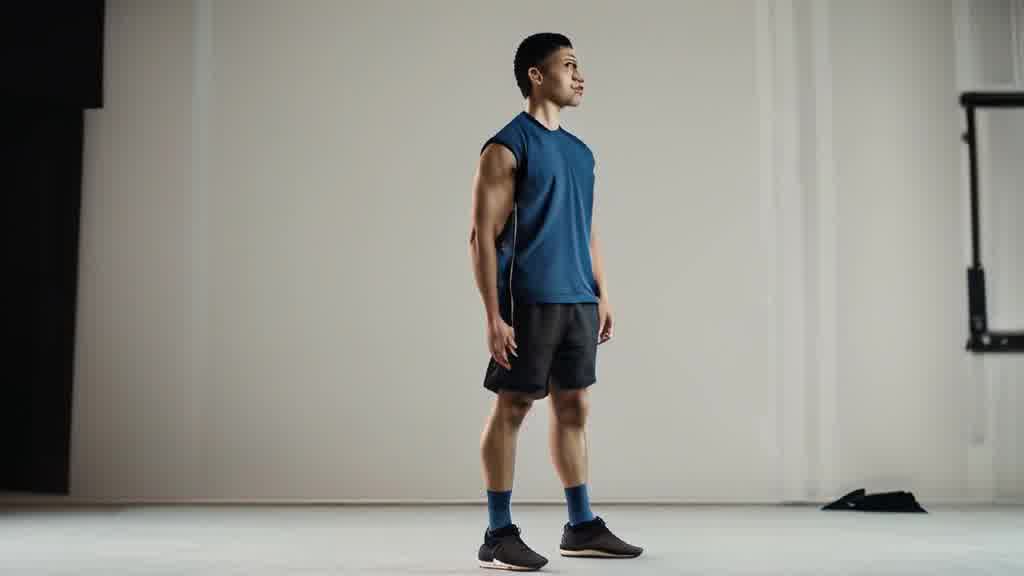}} \raisebox{-.5\height}{\includegraphics[width=0.075\textwidth]{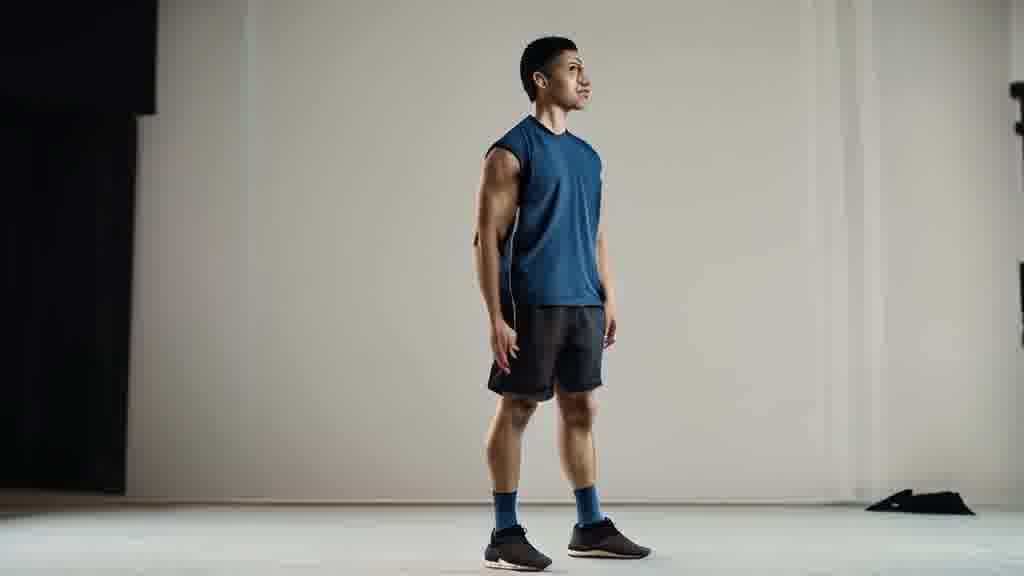}} & \raisebox{-.5\height}{\includegraphics[width=0.075\textwidth]{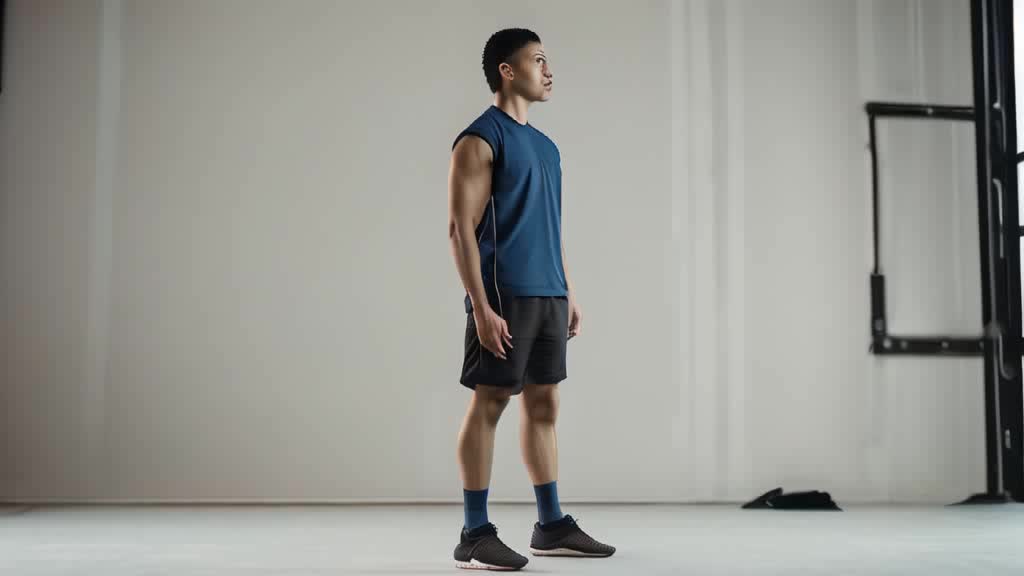}} & \raisebox{-.5\height}{\includegraphics[width=0.075\textwidth]{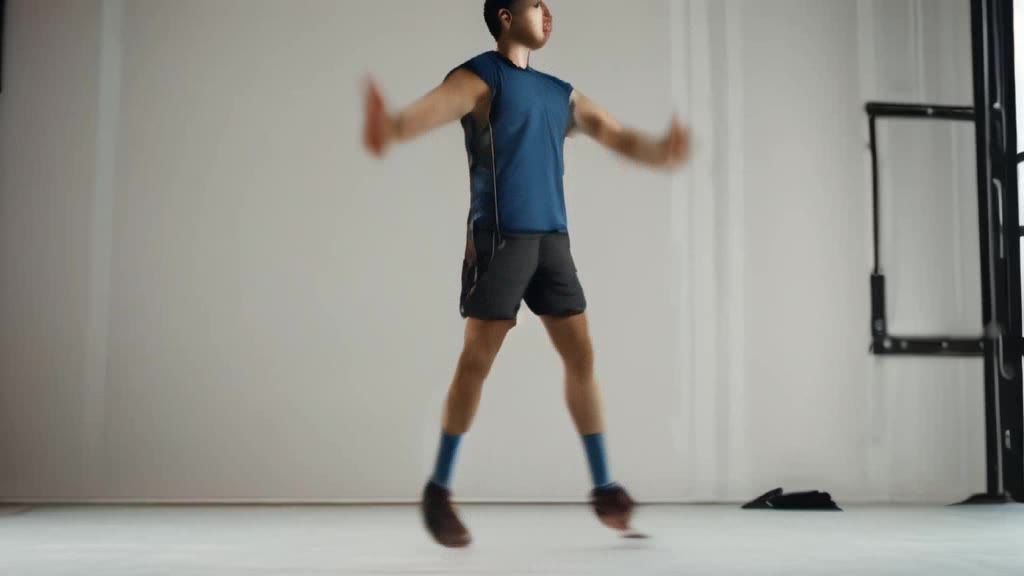}} \raisebox{-.5\height}{\includegraphics[width=0.075\textwidth]{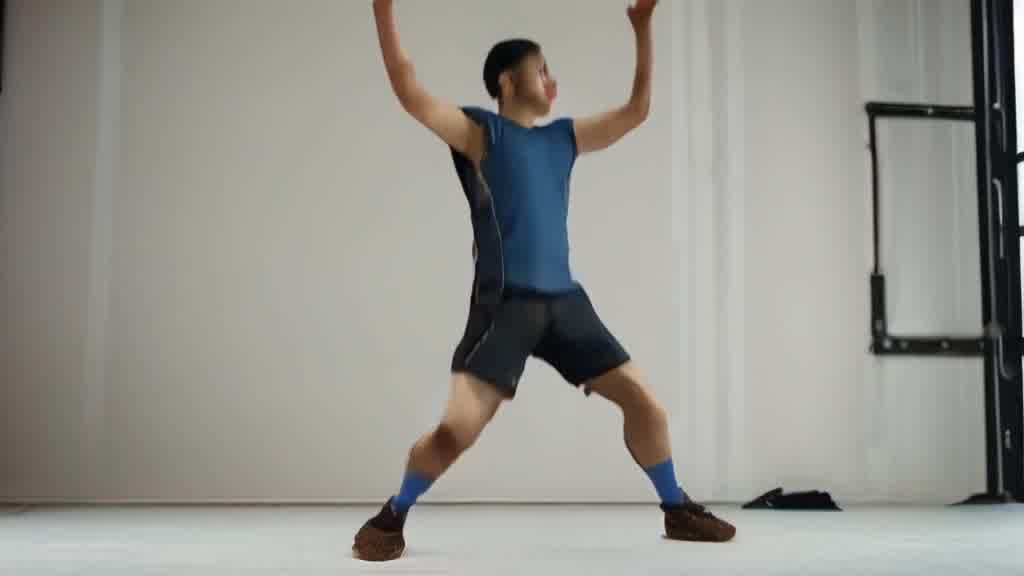}} \raisebox{-.5\height}{\includegraphics[width=0.075\textwidth]{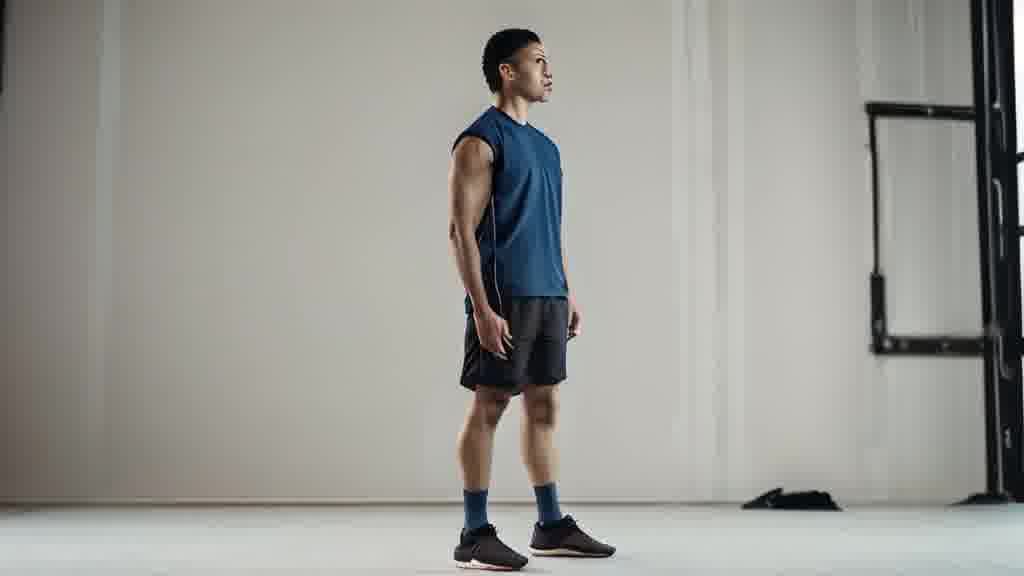}} \\
		{Uncond. Seed 0} & - & \raisebox{-.5\height}{\includegraphics[width=0.075\textwidth]{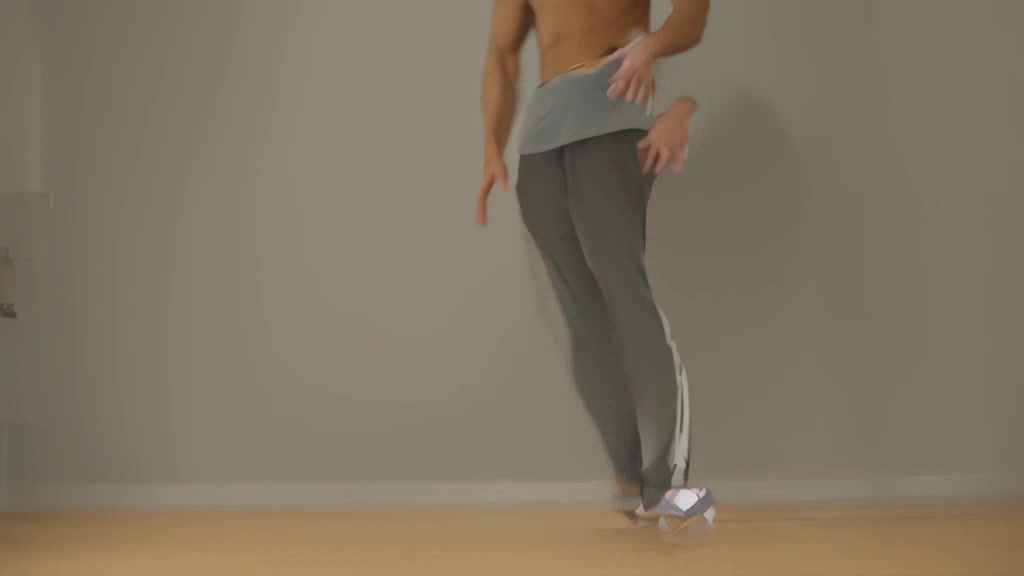}} \raisebox{-.5\height}{\includegraphics[width=0.075\textwidth]{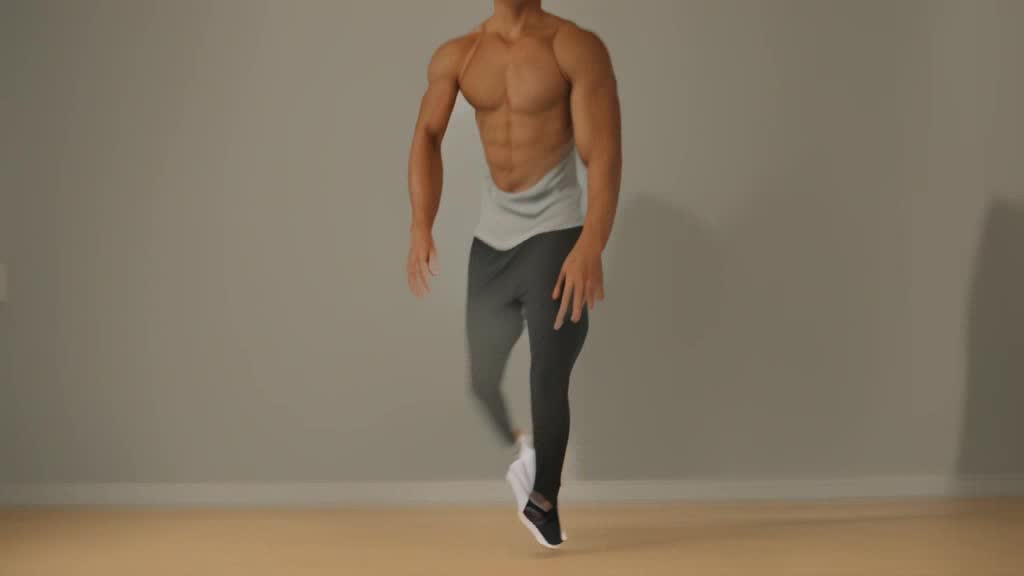}} \raisebox{-.5\height}{\includegraphics[width=0.075\textwidth]{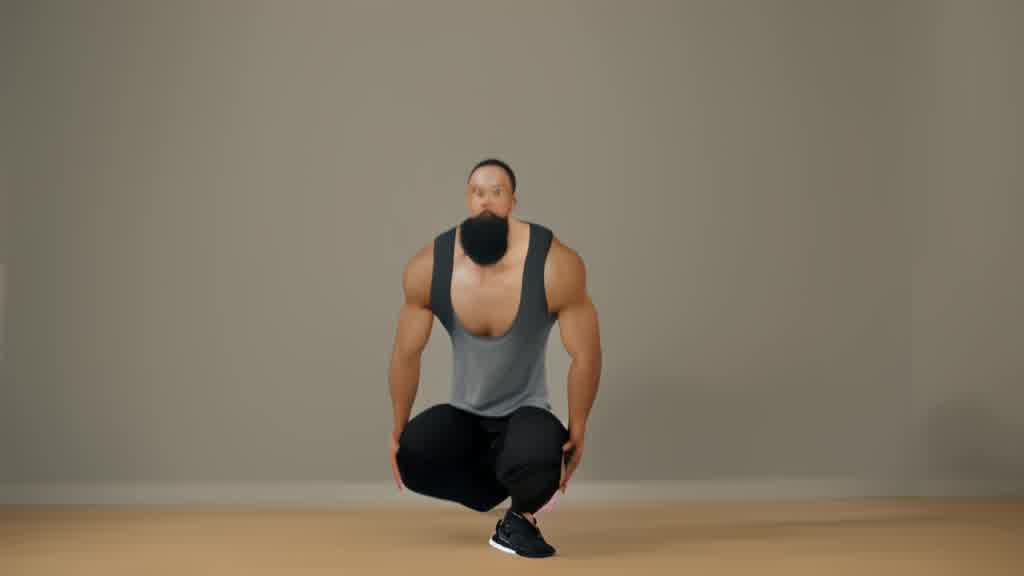}} & - & \raisebox{-.5\height}{\includegraphics[width=0.075\textwidth]{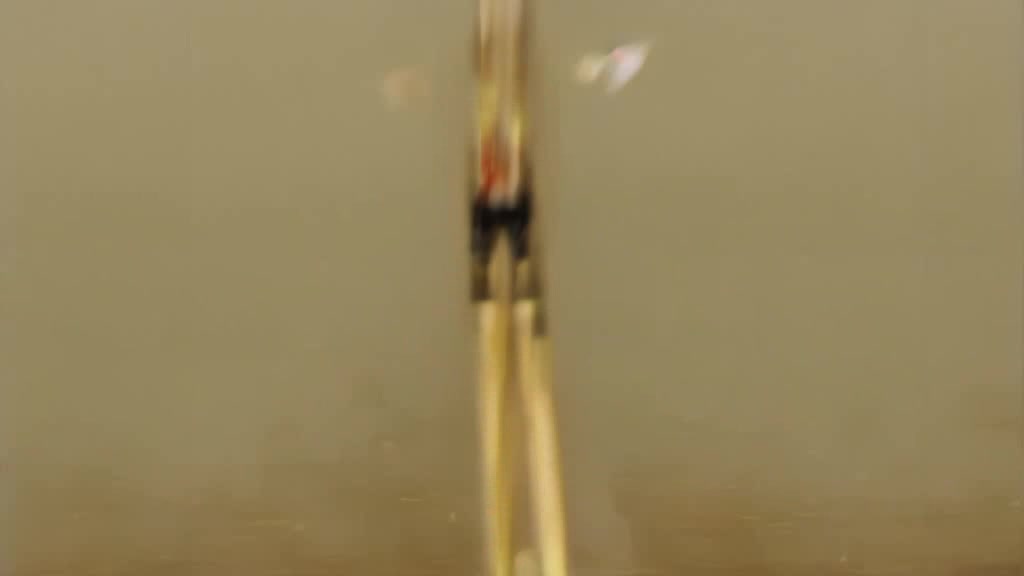}} \raisebox{-.5\height}{\includegraphics[width=0.075\textwidth]{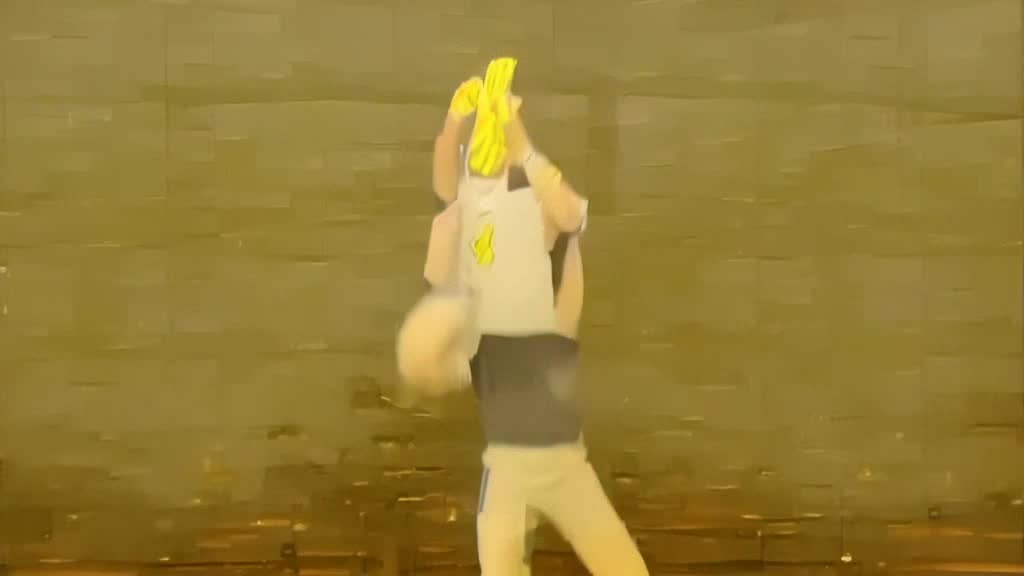}} \raisebox{-.5\height}{\includegraphics[width=0.075\textwidth]{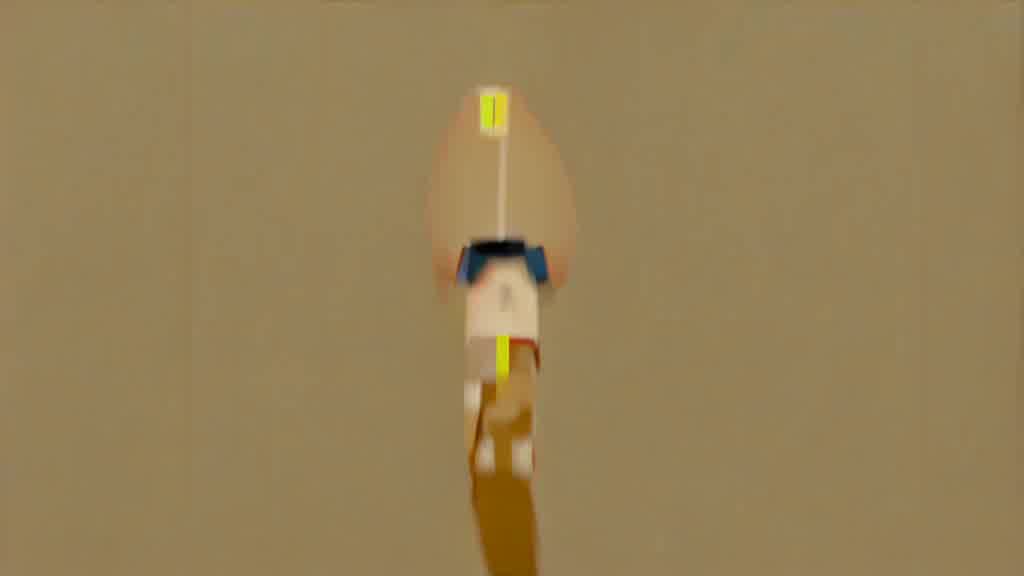}} \\
		{Uncond. Seed 1} & - & \raisebox{-.5\height}{\includegraphics[width=0.075\textwidth]{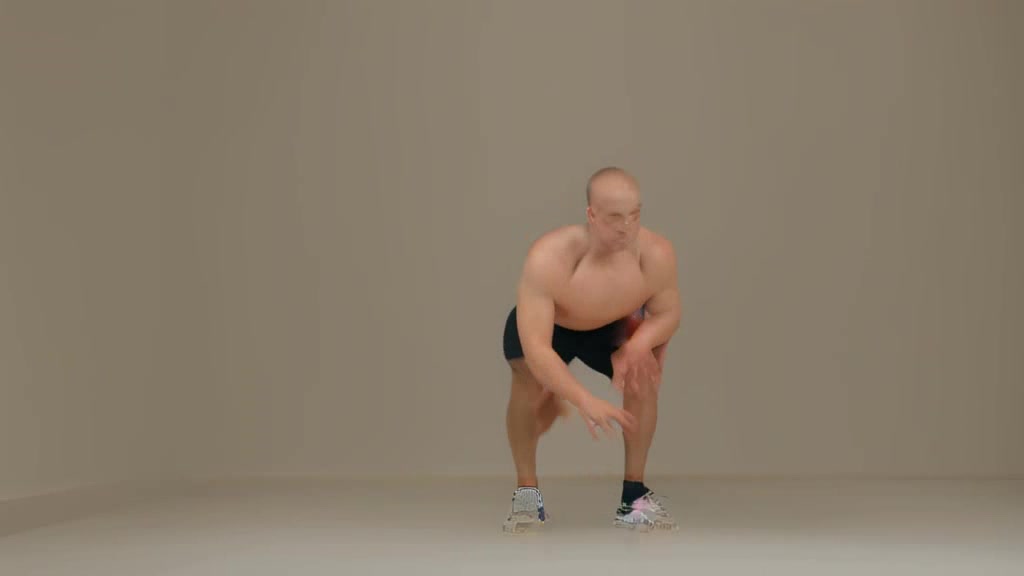}} \raisebox{-.5\height}{\includegraphics[width=0.075\textwidth]{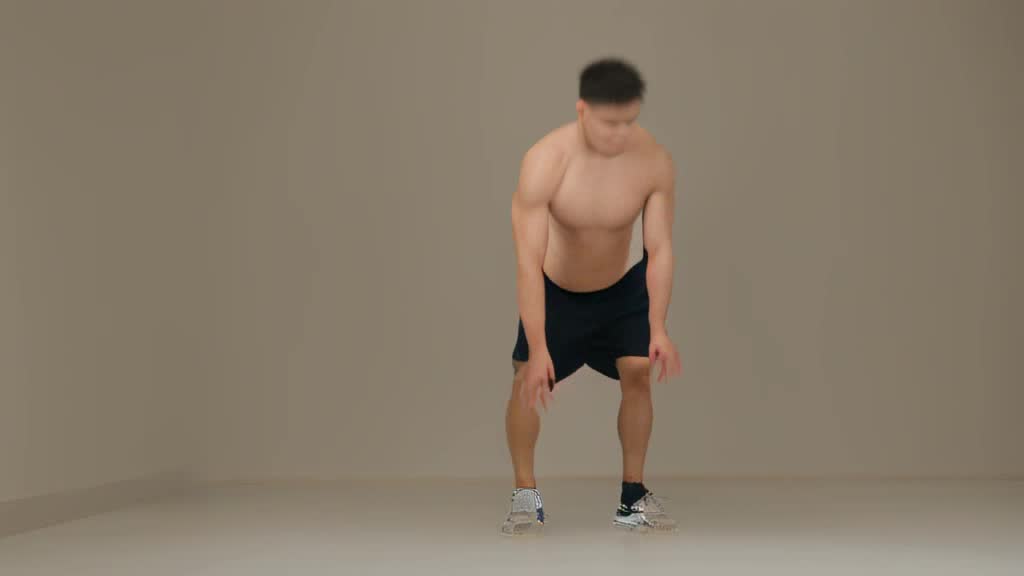}} \raisebox{-.5\height}{\includegraphics[width=0.075\textwidth]{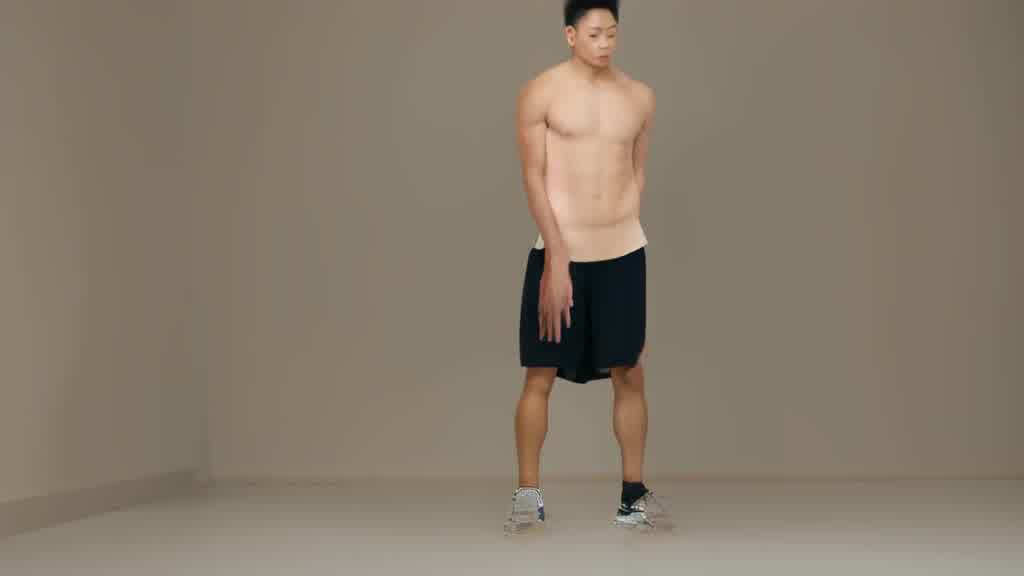}} & - & \raisebox{-.5\height}{\includegraphics[width=0.075\textwidth]{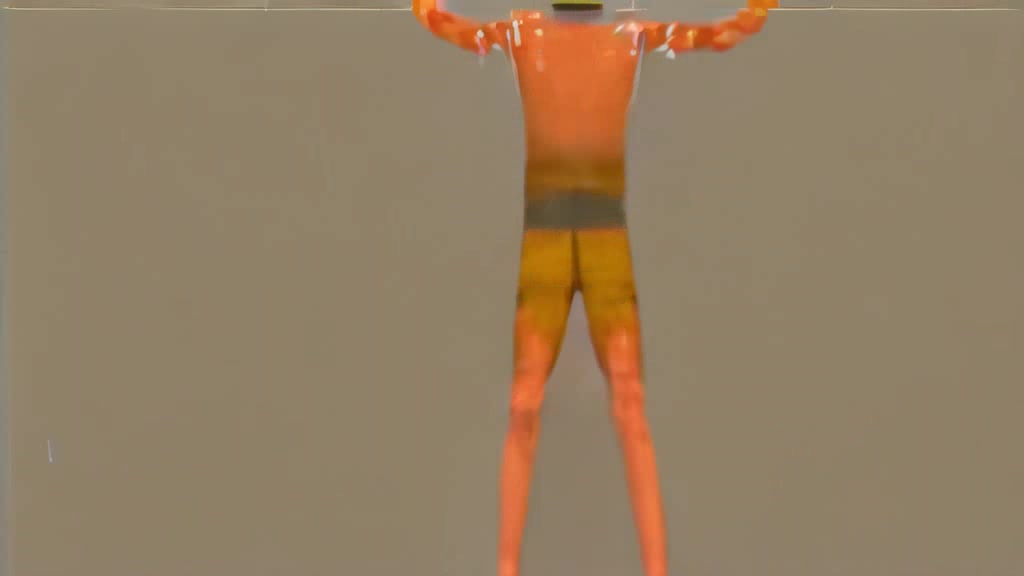}} \raisebox{-.5\height}{\includegraphics[width=0.075\textwidth]{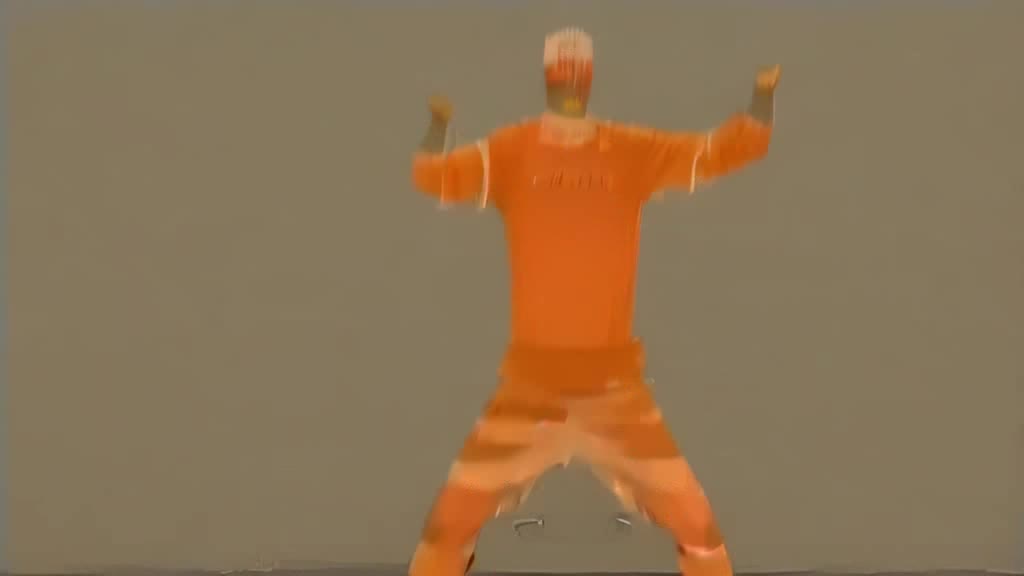}} \raisebox{-.5\height}{\includegraphics[width=0.075\textwidth]{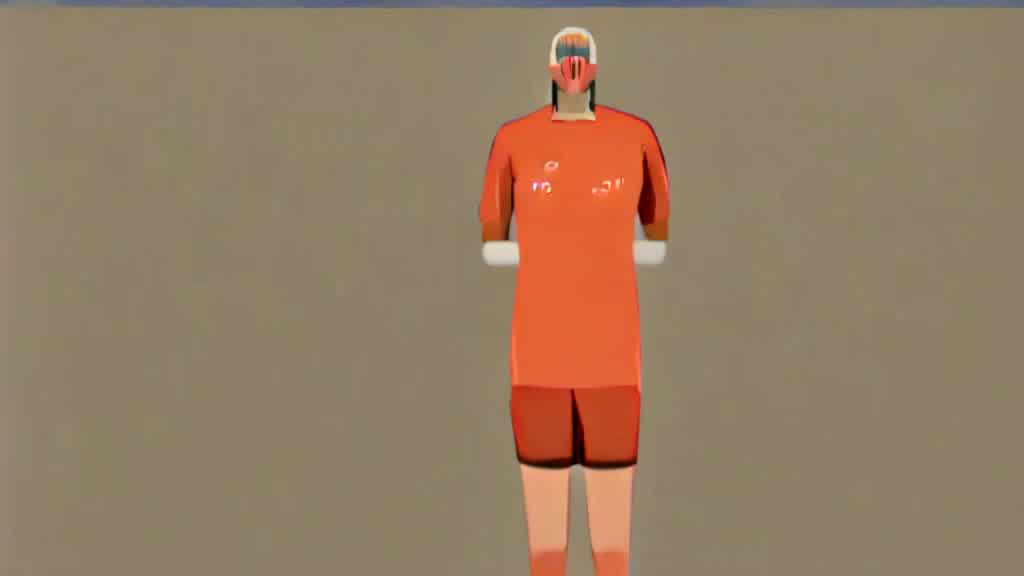}} \\
		{Uncond. Seed 2} & - & \raisebox{-.5\height}{\includegraphics[width=0.075\textwidth]{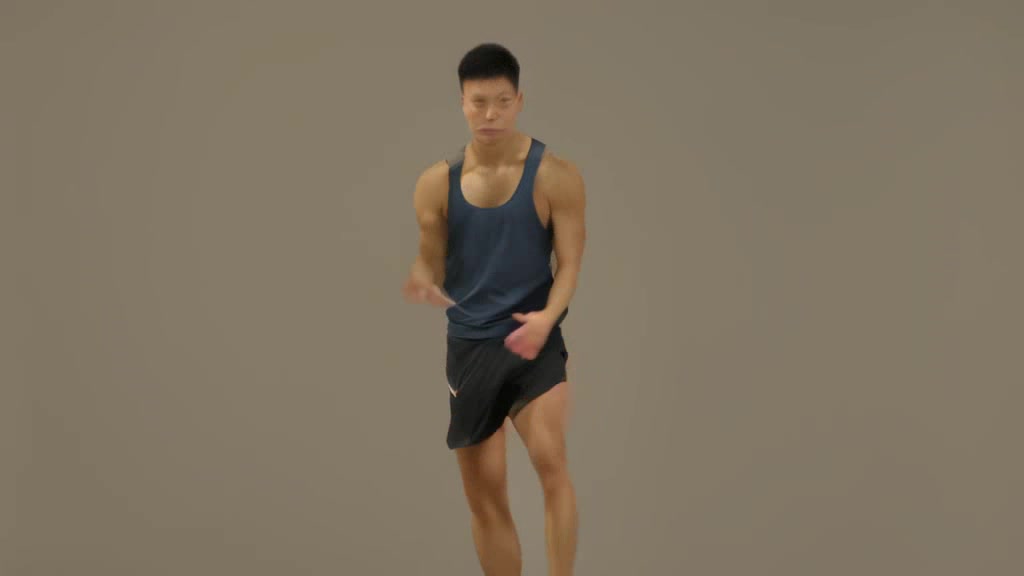}} \raisebox{-.5\height}{\includegraphics[width=0.075\textwidth]{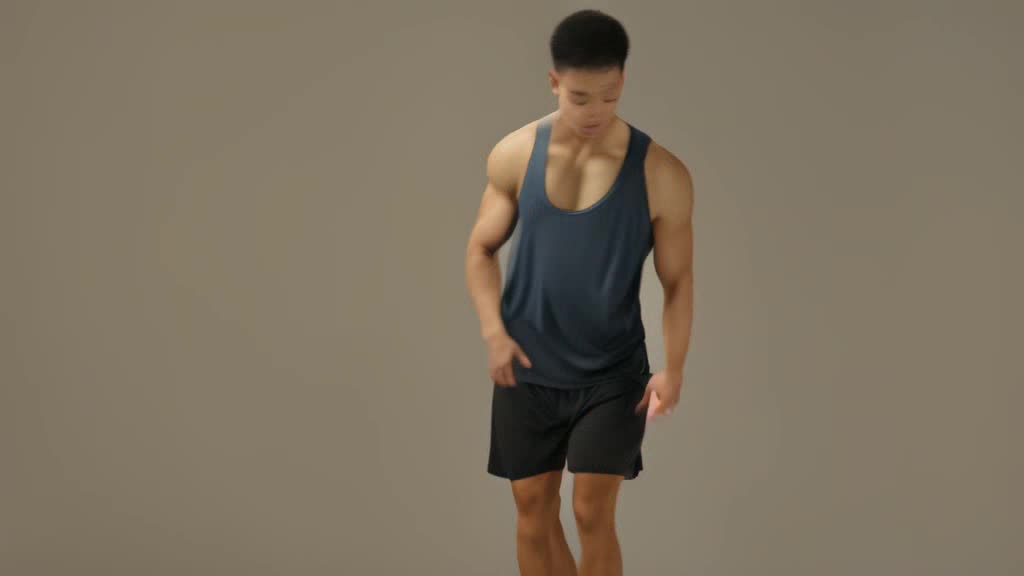}} \raisebox{-.5\height}{\includegraphics[width=0.075\textwidth]{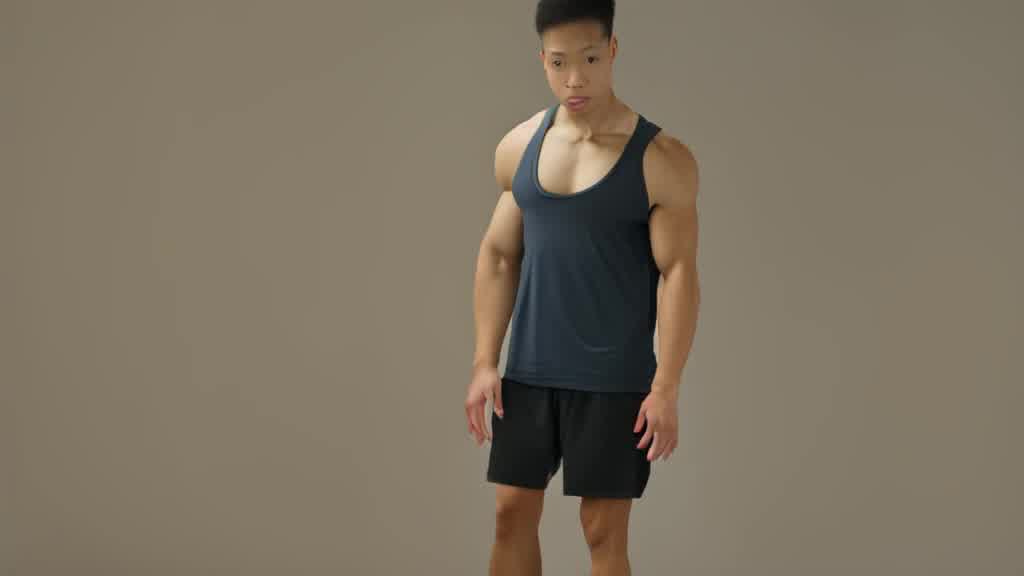}} & - & \raisebox{-.5\height}{\includegraphics[width=0.075\textwidth]{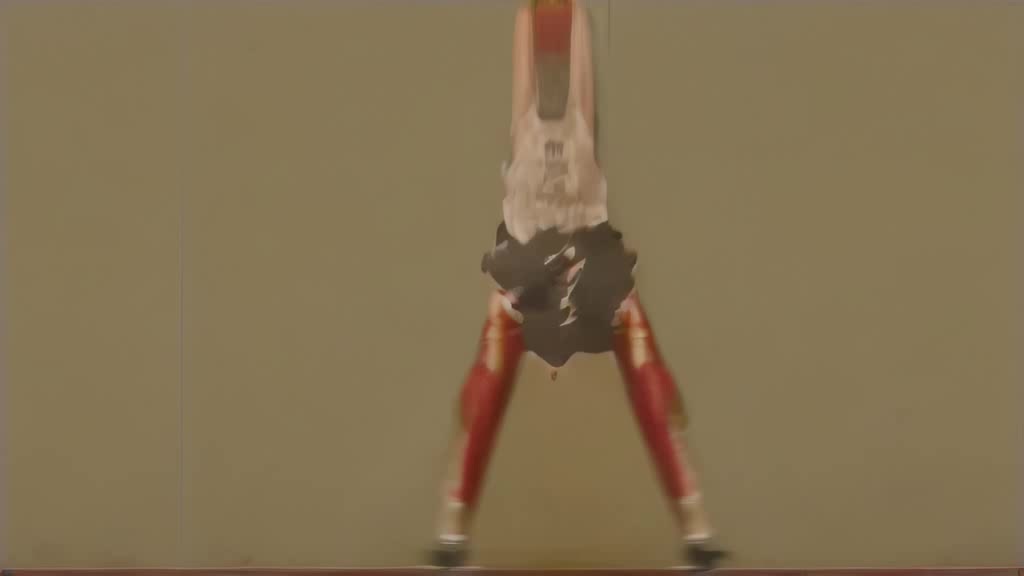}} \raisebox{-.5\height}{\includegraphics[width=0.075\textwidth]{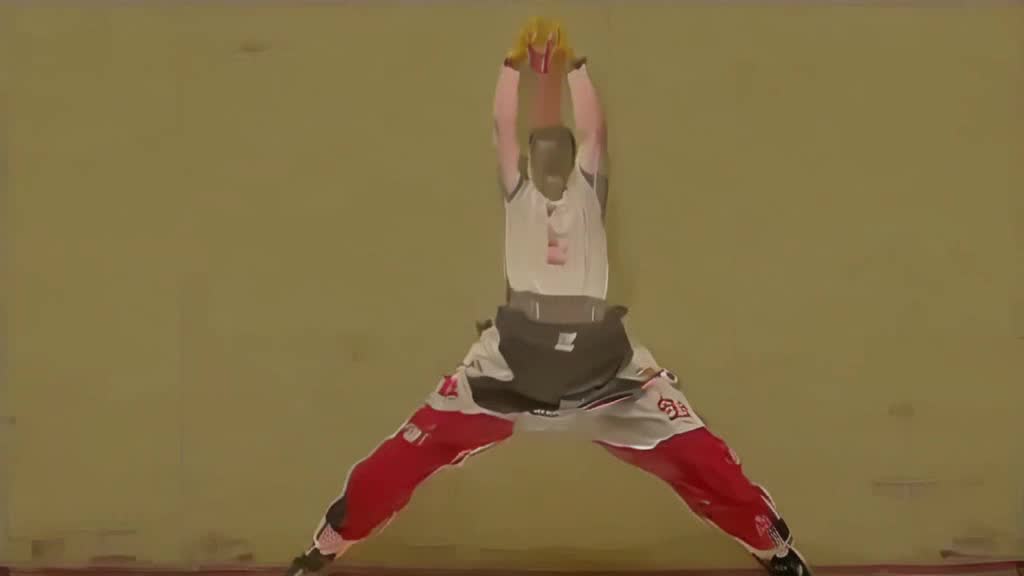}} \raisebox{-.5\height}{\includegraphics[width=0.075\textwidth]{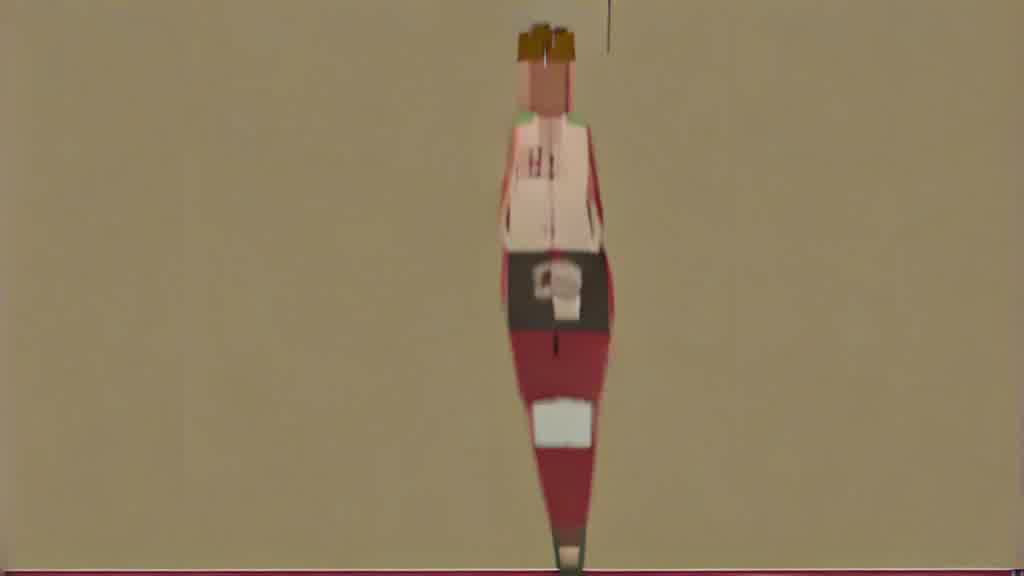}} \\
		{Ref.} & \raisebox{-.5\height}{\includegraphics[width=0.075\textwidth]{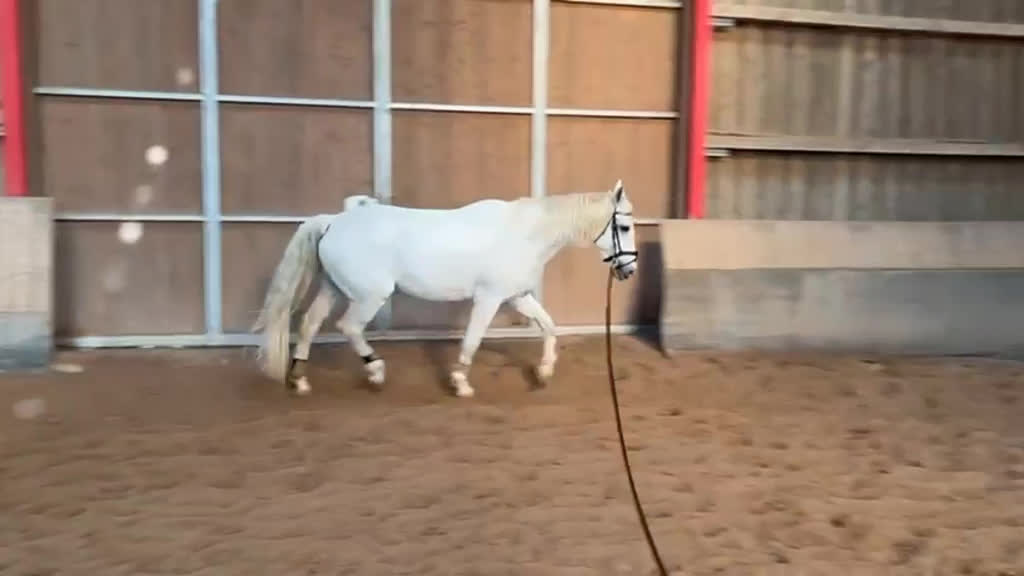}} & \raisebox{-.5\height}{\includegraphics[width=0.075\textwidth]{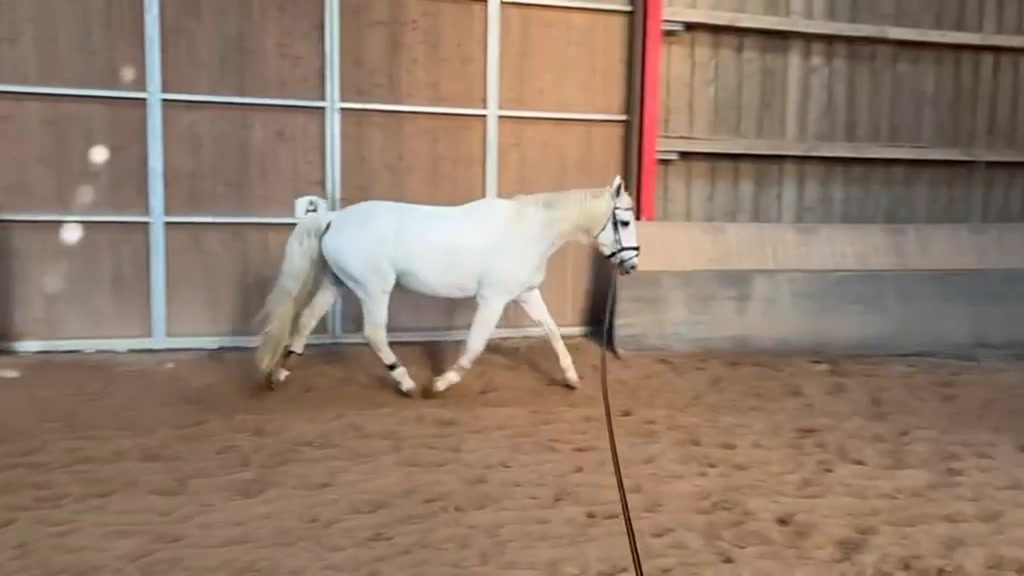}} \raisebox{-.5\height}{\includegraphics[width=0.075\textwidth]{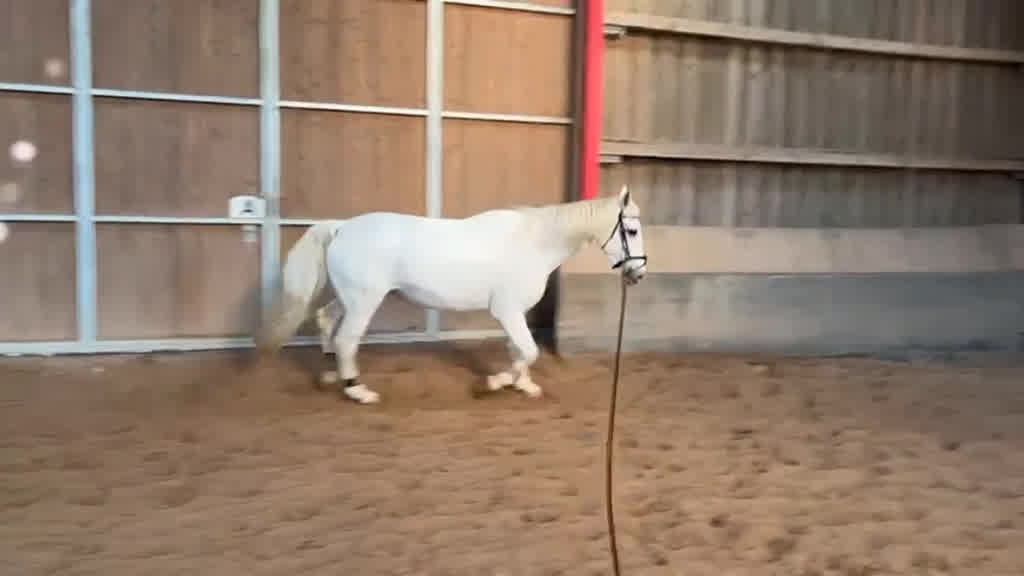}} \raisebox{-.5\height}{\includegraphics[width=0.075\textwidth]{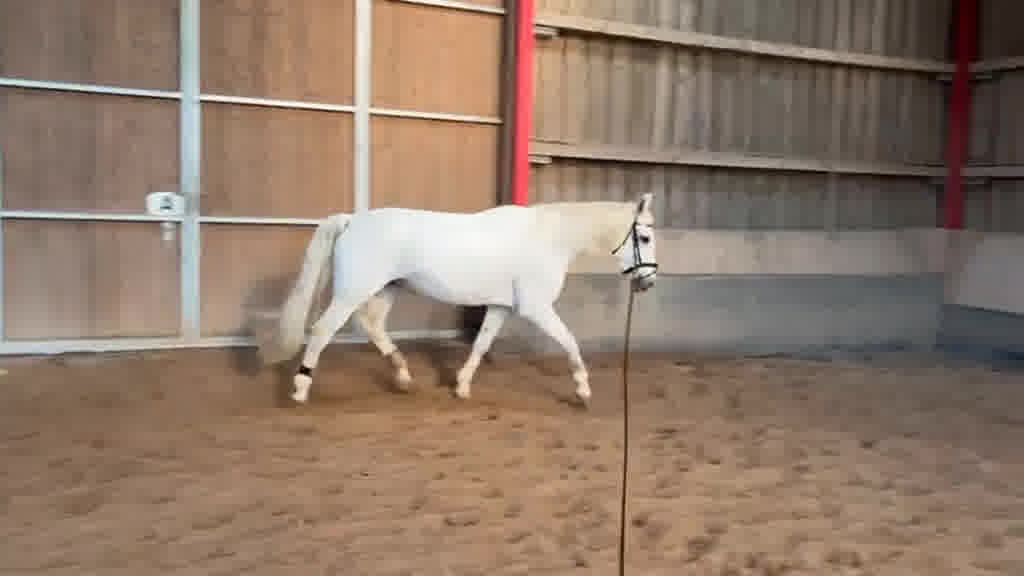}} & \raisebox{-.5\height}{\includegraphics[width=0.075\textwidth]{figures/svd_comparison/animal_fourlegged/frames_input/001.jpg}} & \raisebox{-.5\height}{\includegraphics[width=0.075\textwidth]{figures/svd_comparison/animal_fourlegged/frames_input/004.jpg}} \raisebox{-.5\height}{\includegraphics[width=0.075\textwidth]{figures/svd_comparison/animal_fourlegged/frames_input/008.jpg}} \raisebox{-.5\height}{\includegraphics[width=0.075\textwidth]{figures/svd_comparison/animal_fourlegged/frames_input/012.jpg}} \\
		{Seed 0} & \raisebox{-.5\height}{\includegraphics[width=0.075\textwidth]{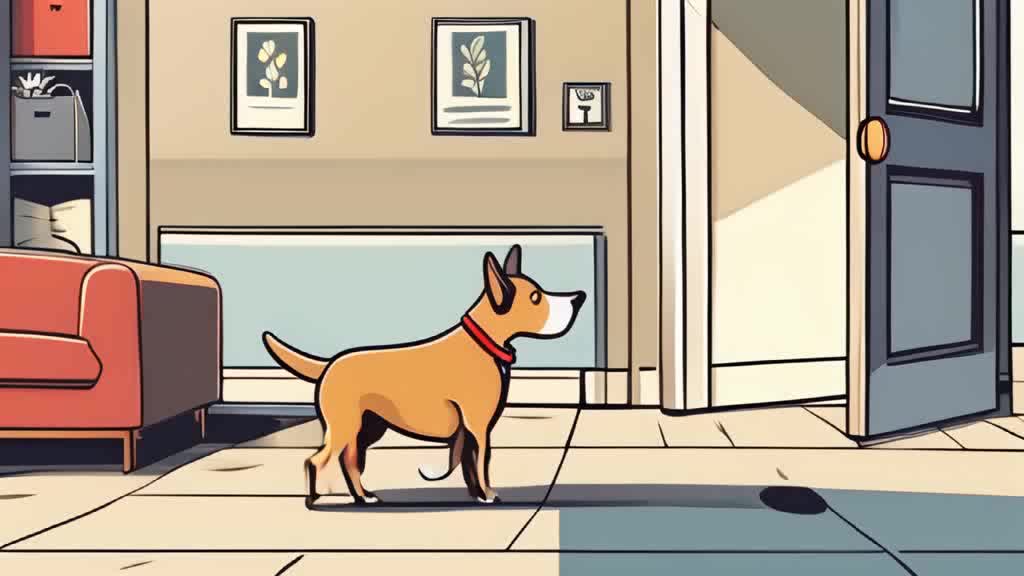}} & \raisebox{-.5\height}{\includegraphics[width=0.075\textwidth]{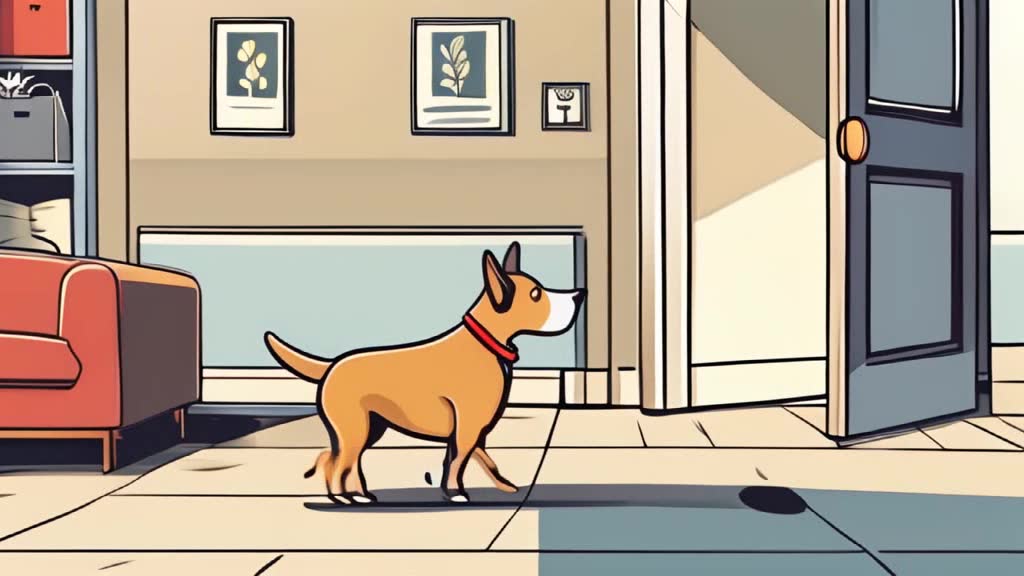}} \raisebox{-.5\height}{\includegraphics[width=0.075\textwidth]{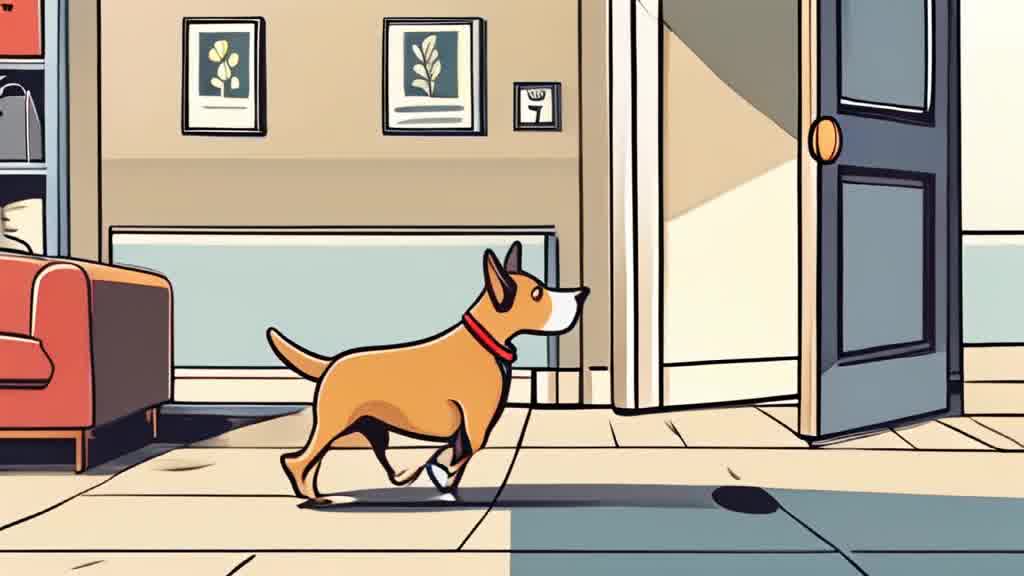}} \raisebox{-.5\height}{\includegraphics[width=0.075\textwidth]{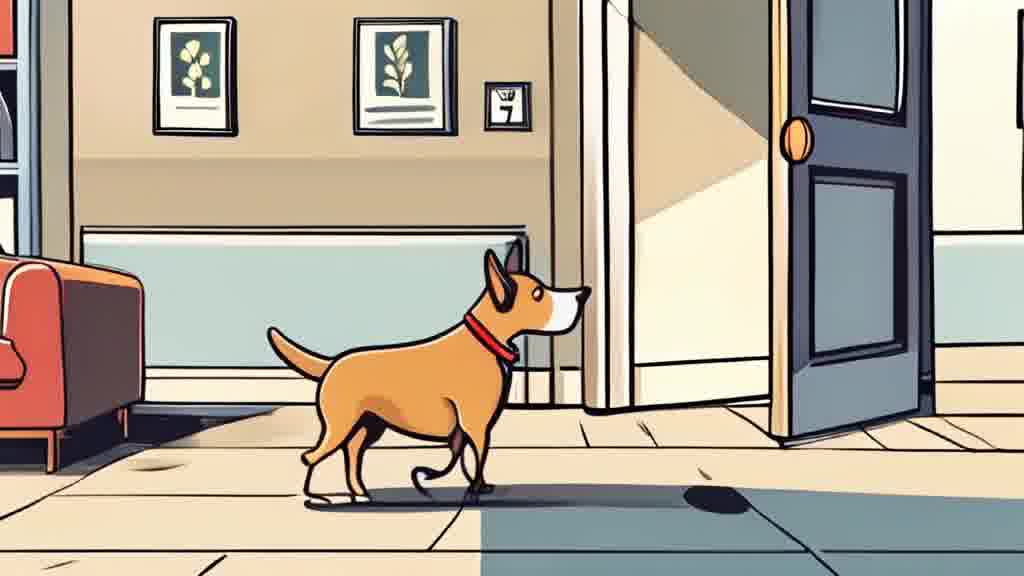}} & \raisebox{-.5\height}{\includegraphics[width=0.075\textwidth]{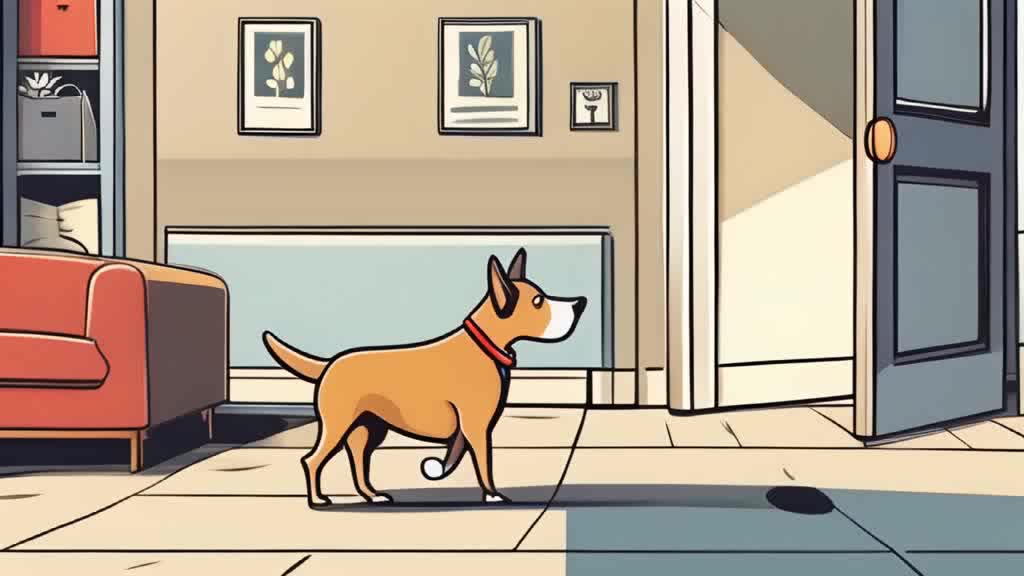}} & \raisebox{-.5\height}{\includegraphics[width=0.075\textwidth]{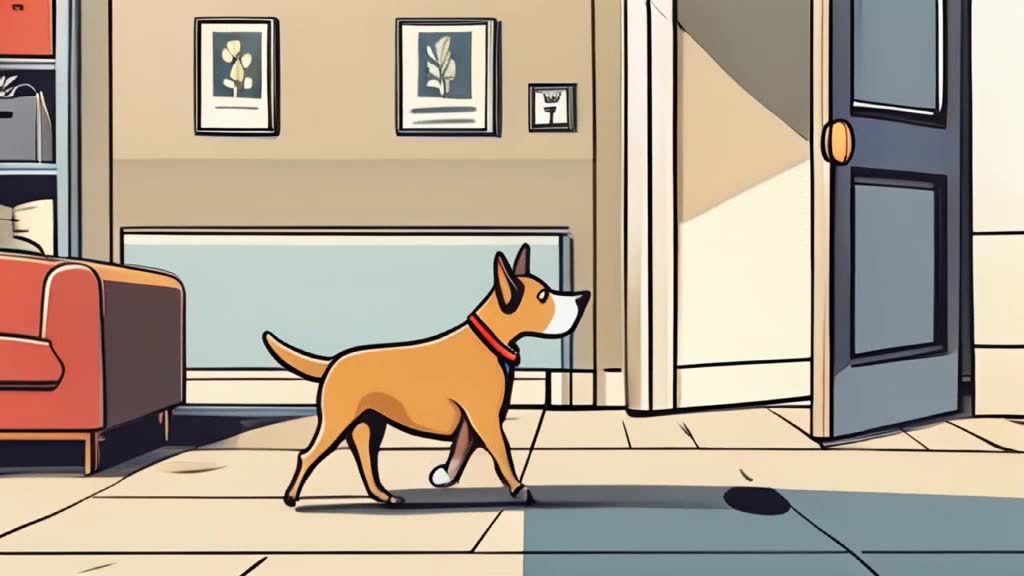}} \raisebox{-.5\height}{\includegraphics[width=0.075\textwidth]{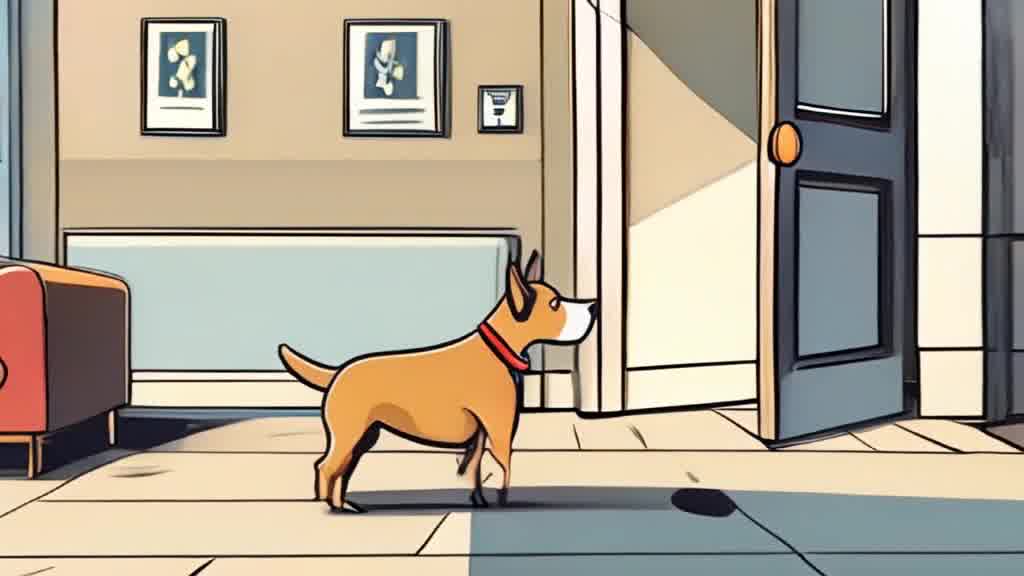}} \raisebox{-.5\height}{\includegraphics[width=0.075\textwidth]{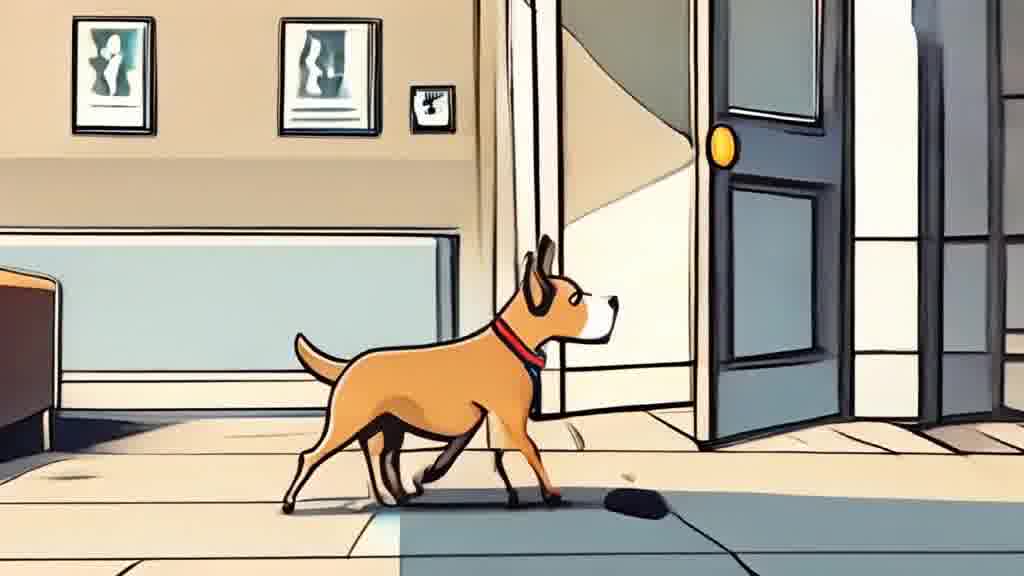}} \\
		{Seed 1} & \raisebox{-.5\height}{\includegraphics[width=0.075\textwidth]{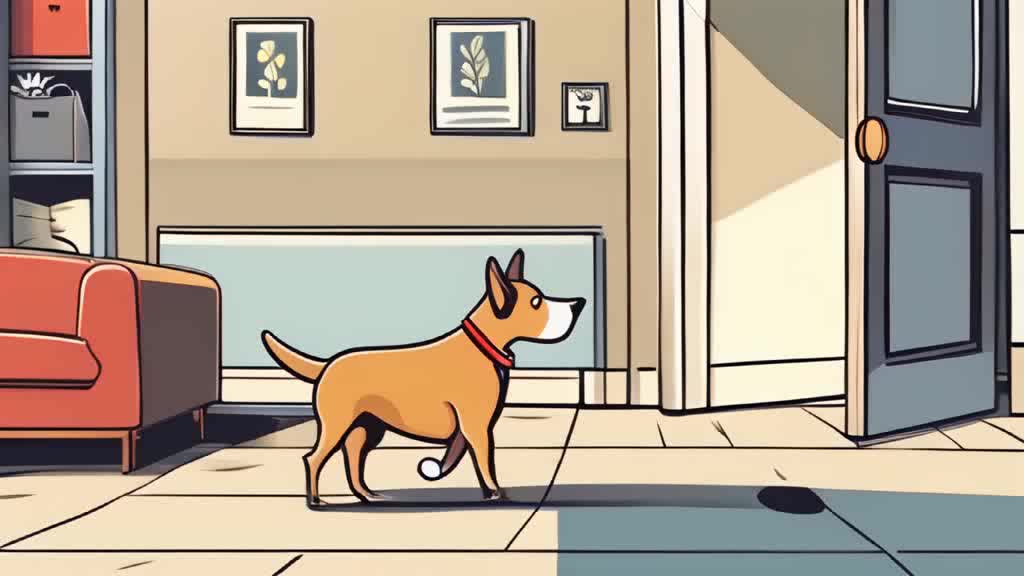}} & \raisebox{-.5\height}{\includegraphics[width=0.075\textwidth]{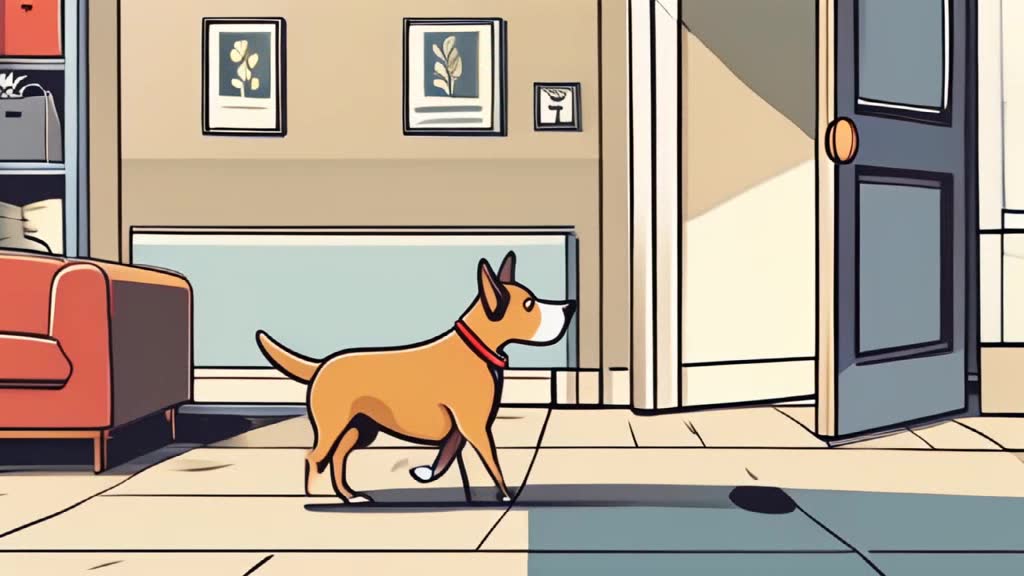}} \raisebox{-.5\height}{\includegraphics[width=0.075\textwidth]{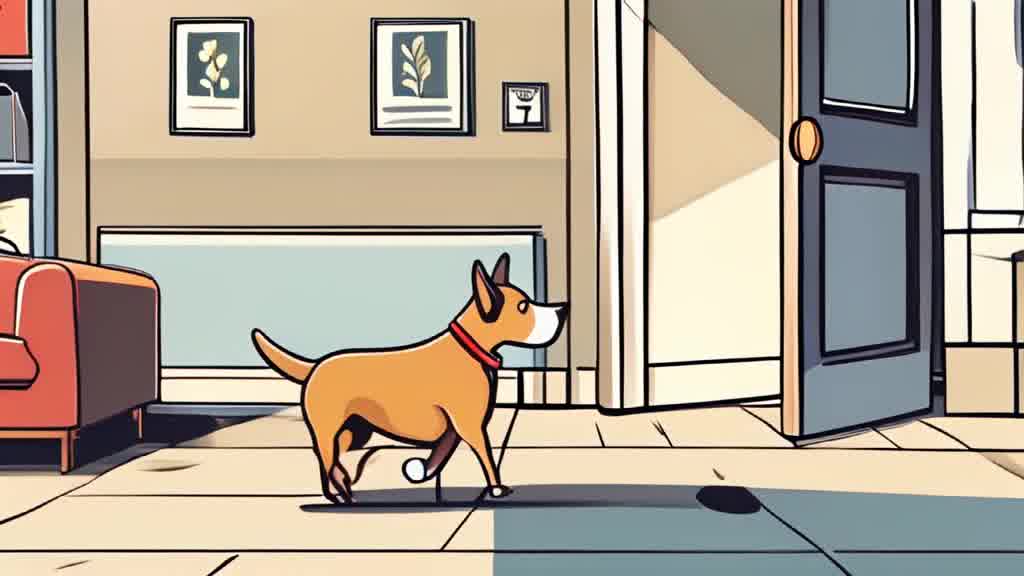}} \raisebox{-.5\height}{\includegraphics[width=0.075\textwidth]{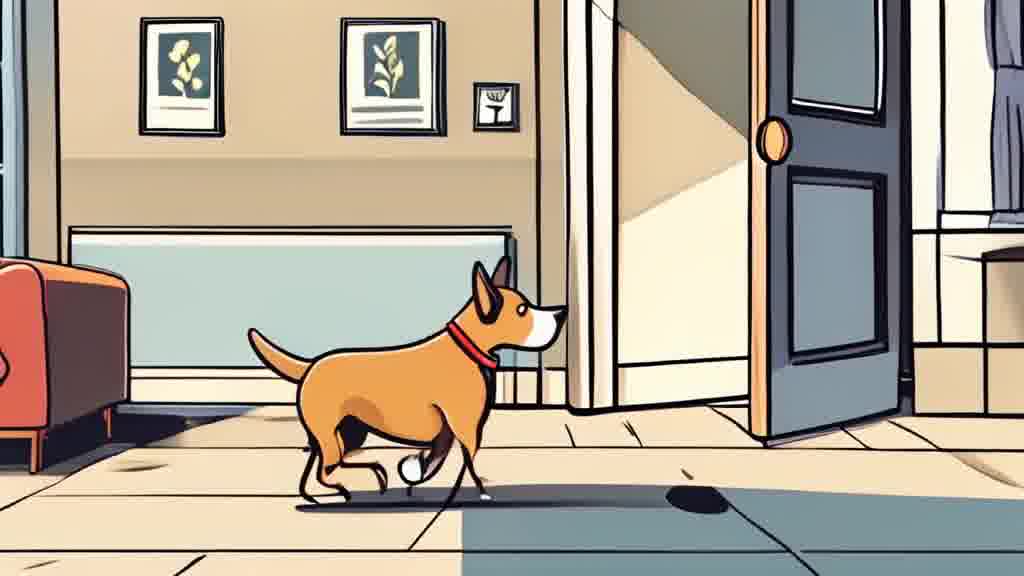}} & \raisebox{-.5\height}{\includegraphics[width=0.075\textwidth]{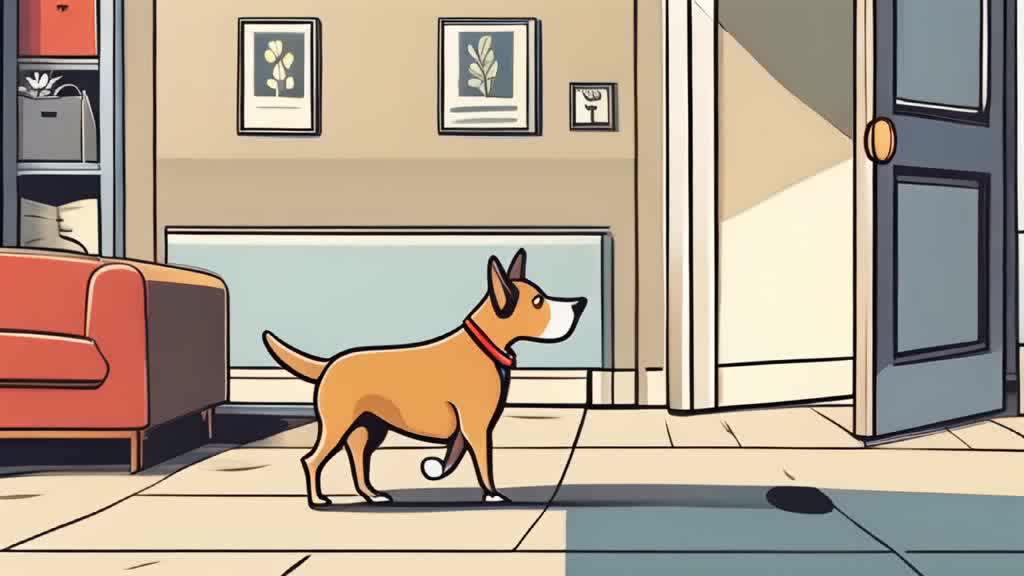}} & \raisebox{-.5\height}{\includegraphics[width=0.075\textwidth]{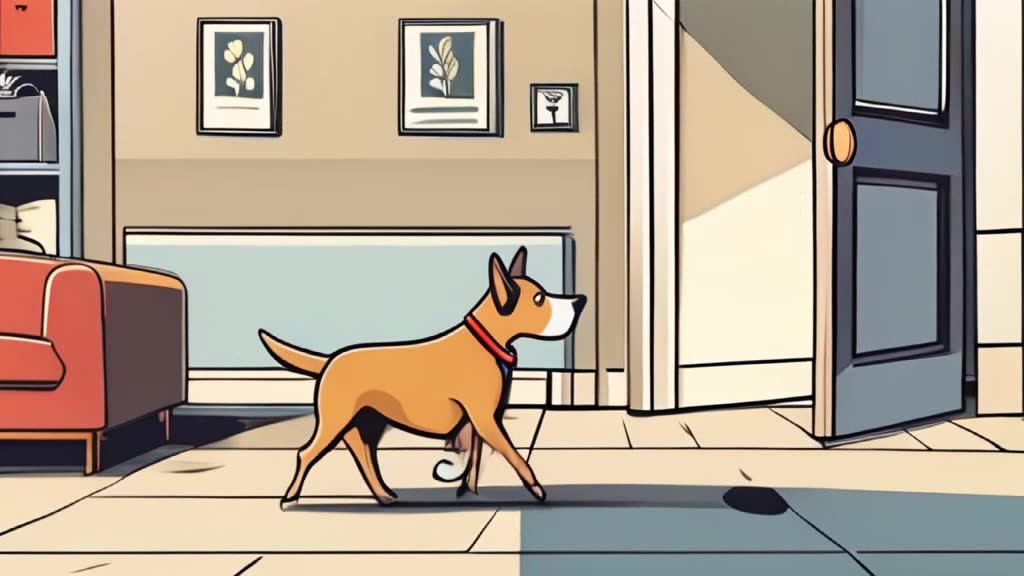}} \raisebox{-.5\height}{\includegraphics[width=0.075\textwidth]{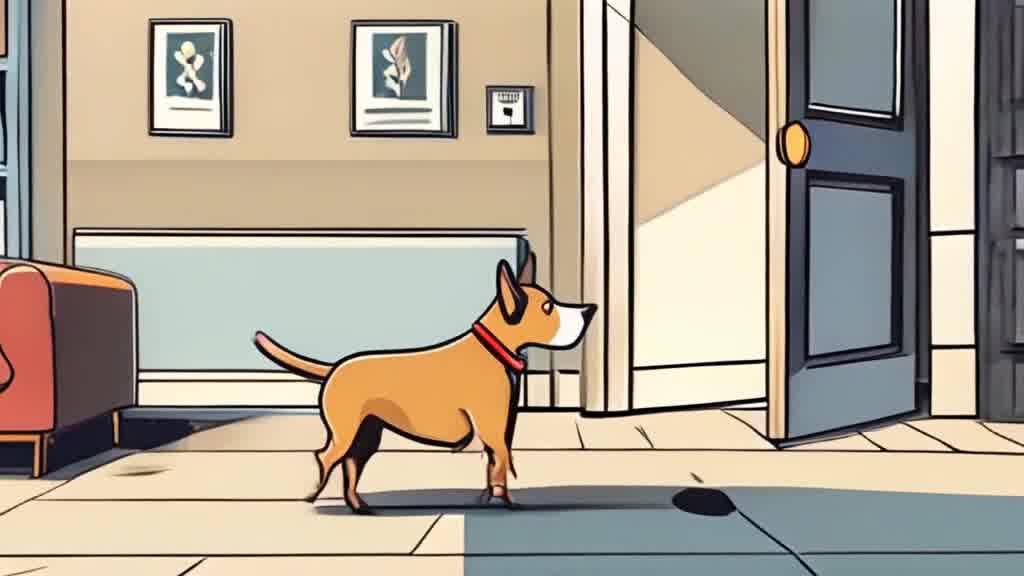}} \raisebox{-.5\height}{\includegraphics[width=0.075\textwidth]{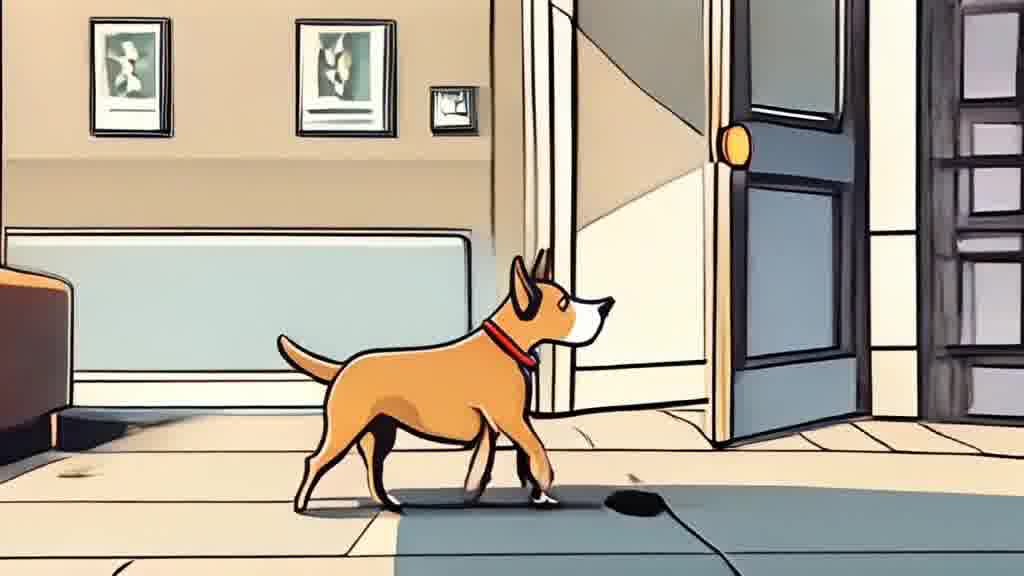}} \\
		{Seed 2} & \raisebox{-.5\height}{\includegraphics[width=0.075\textwidth]{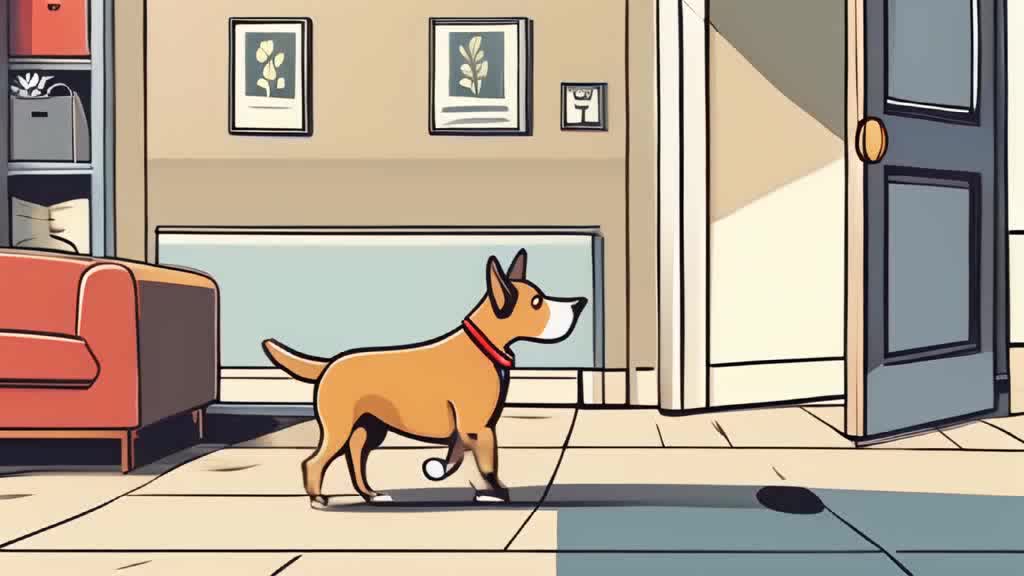}} & \raisebox{-.5\height}{\includegraphics[width=0.075\textwidth]{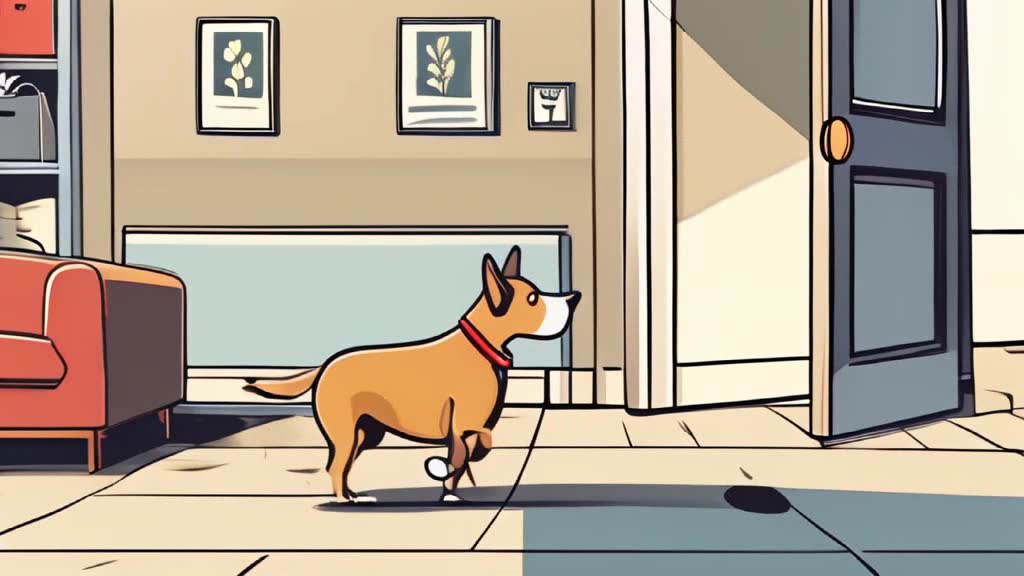}} \raisebox{-.5\height}{\includegraphics[width=0.075\textwidth]{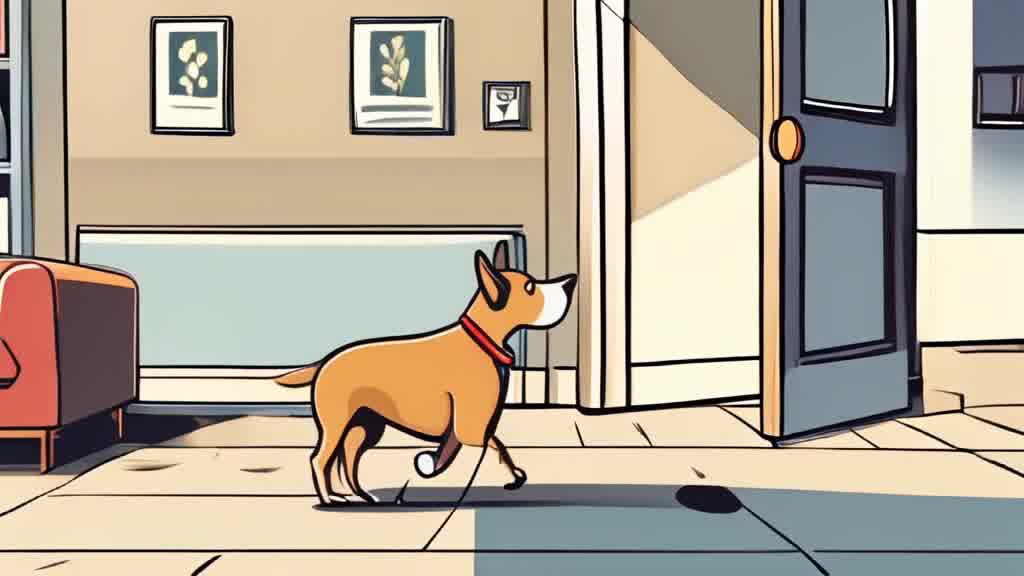}} \raisebox{-.5\height}{\includegraphics[width=0.075\textwidth]{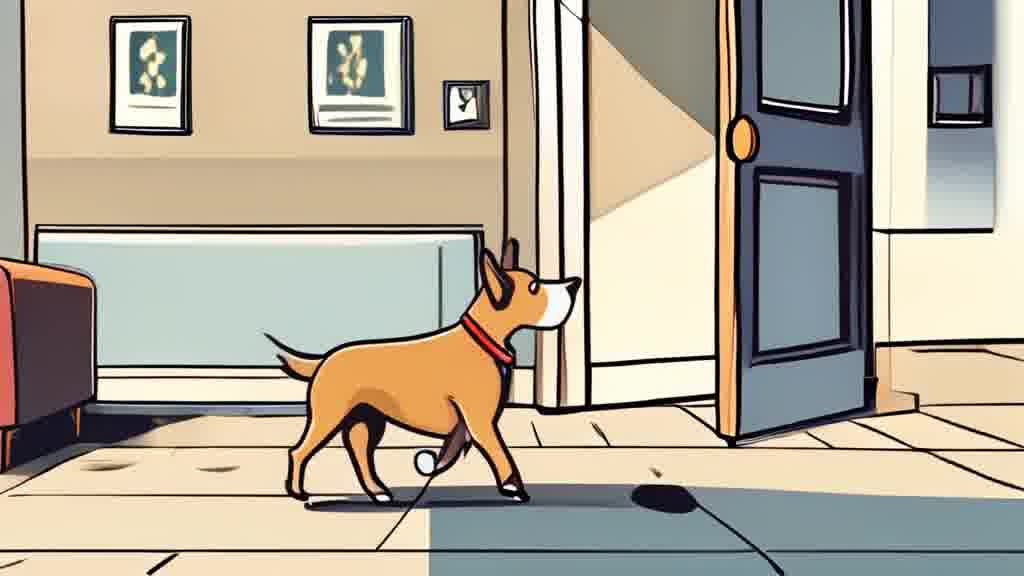}} & \raisebox{-.5\height}{\includegraphics[width=0.075\textwidth]{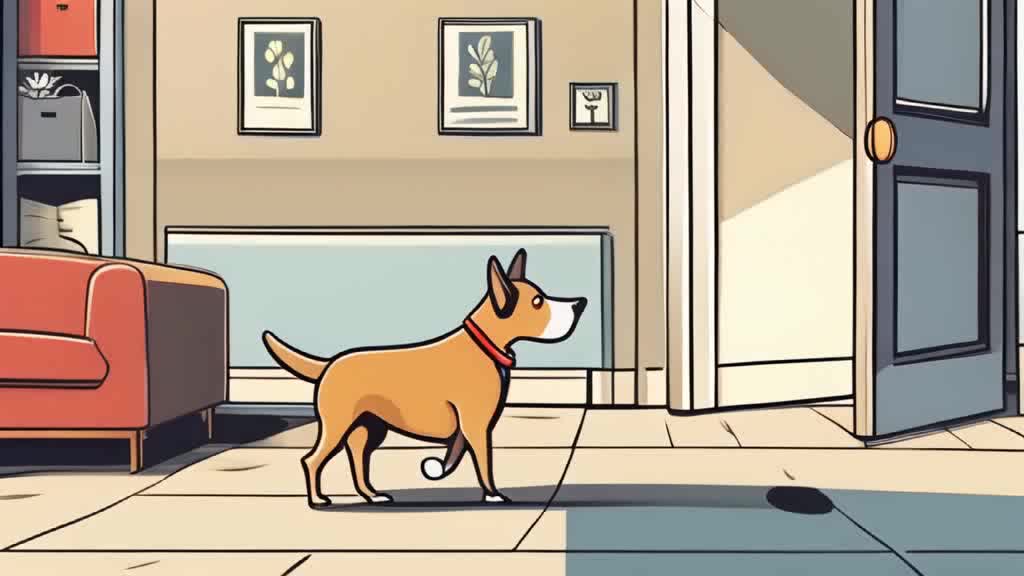}} & \raisebox{-.5\height}{\includegraphics[width=0.075\textwidth]{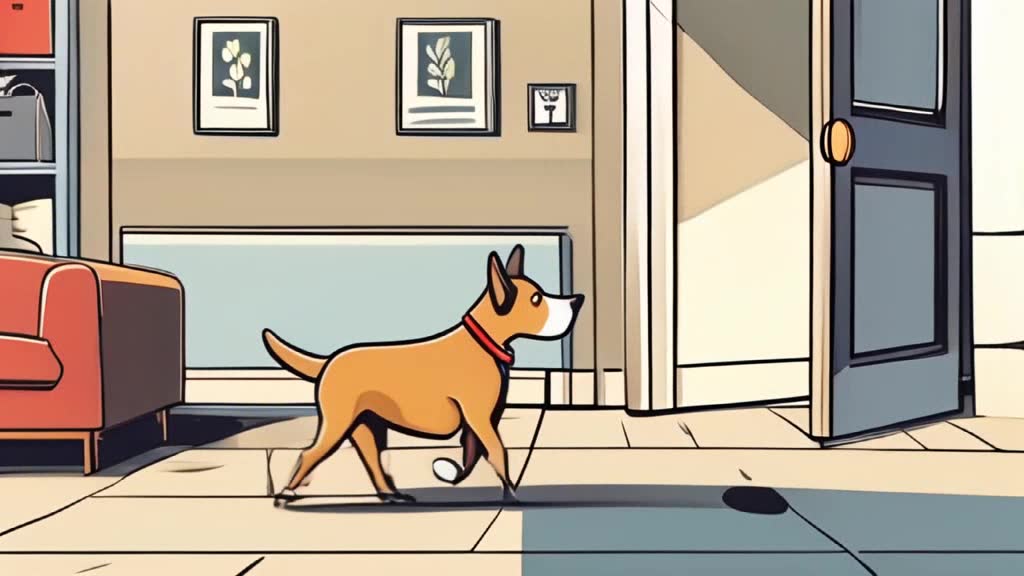}} \raisebox{-.5\height}{\includegraphics[width=0.075\textwidth]{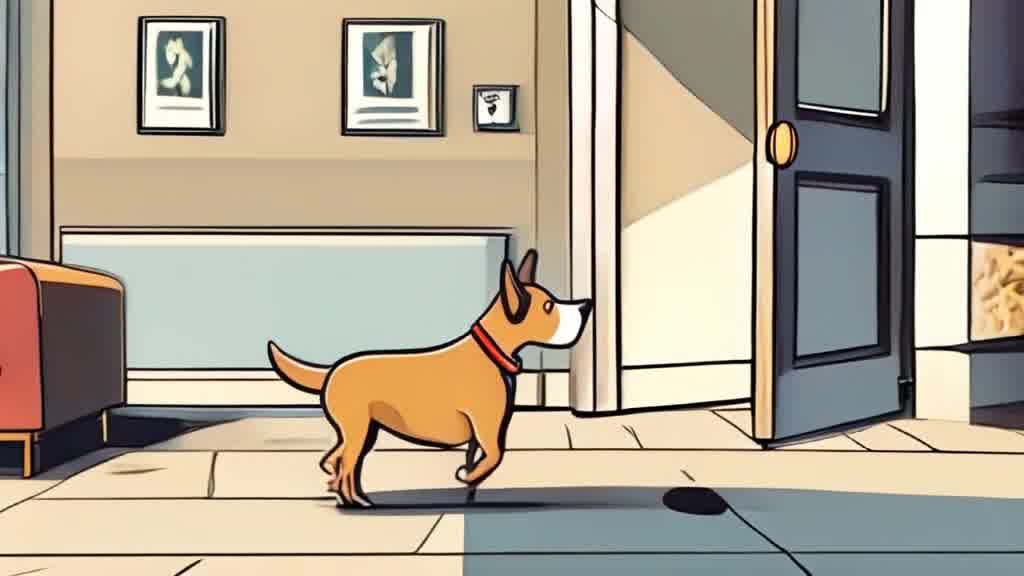}} \raisebox{-.5\height}{\includegraphics[width=0.075\textwidth]{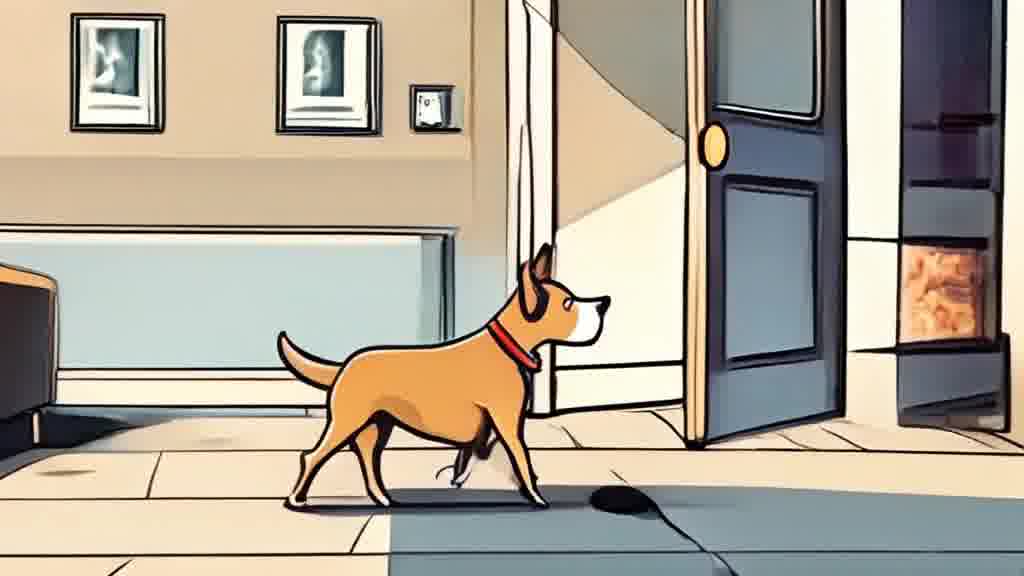}} \\
		{Uncond. Seed 0} & - & \raisebox{-.5\height}{\includegraphics[width=0.075\textwidth]{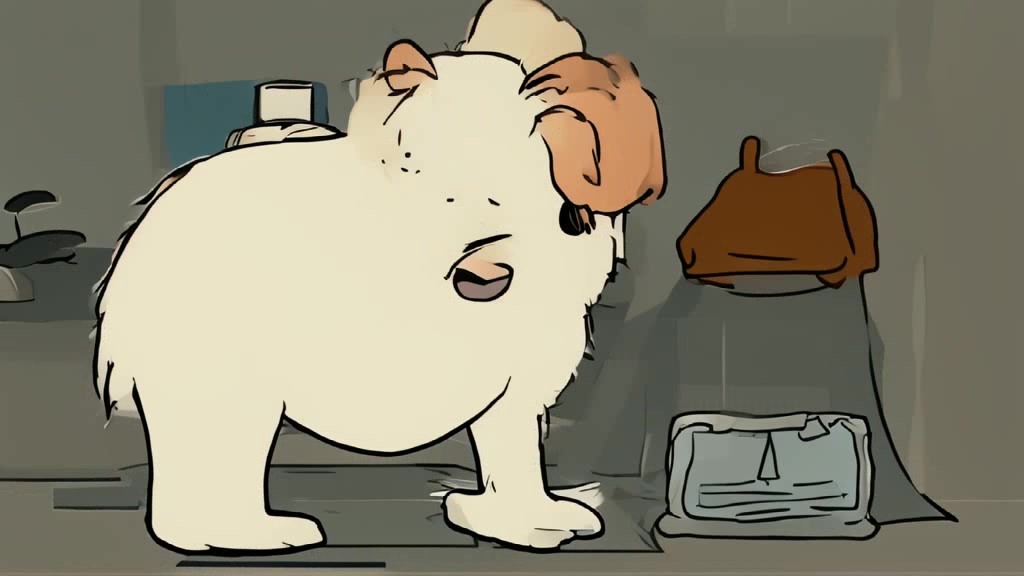}} \raisebox{-.5\height}{\includegraphics[width=0.075\textwidth]{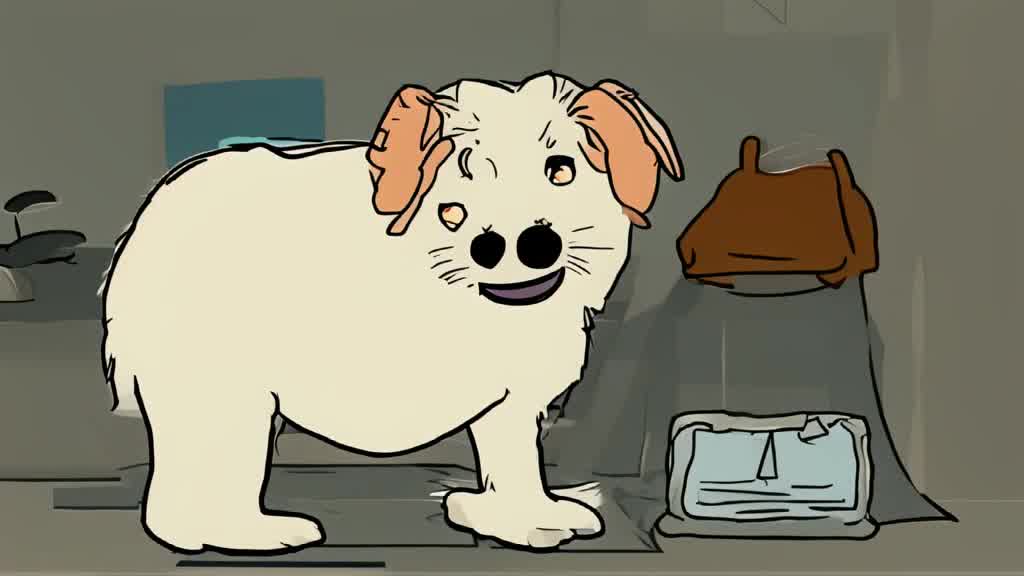}} \raisebox{-.5\height}{\includegraphics[width=0.075\textwidth]{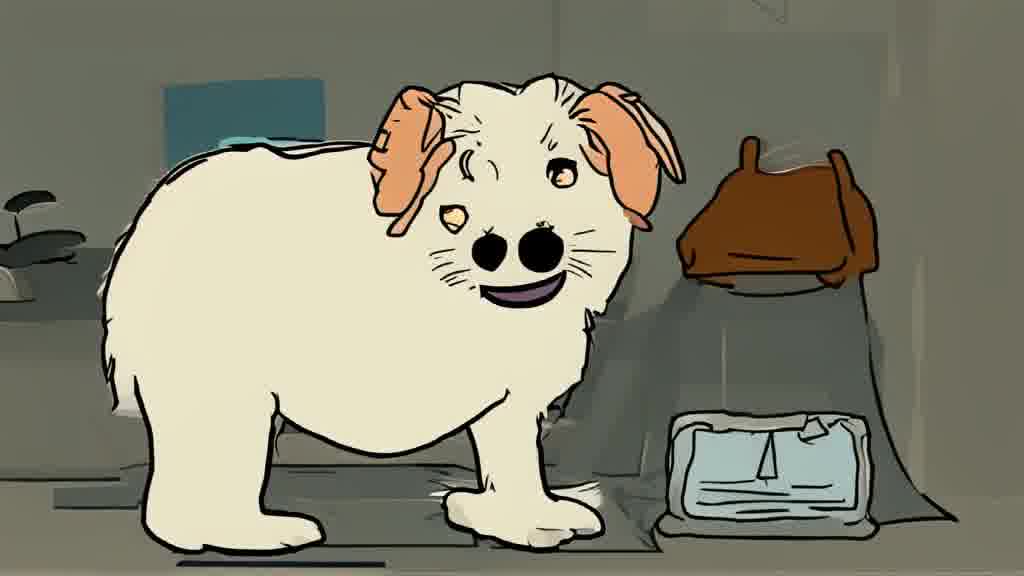}} & - & \raisebox{-.5\height}{\includegraphics[width=0.075\textwidth]{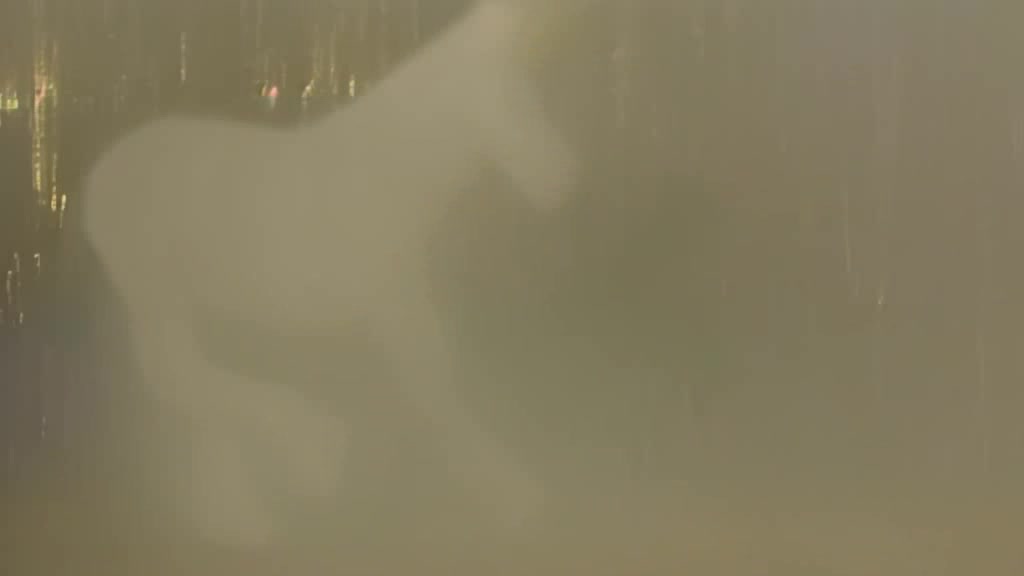}} \raisebox{-.5\height}{\includegraphics[width=0.075\textwidth]{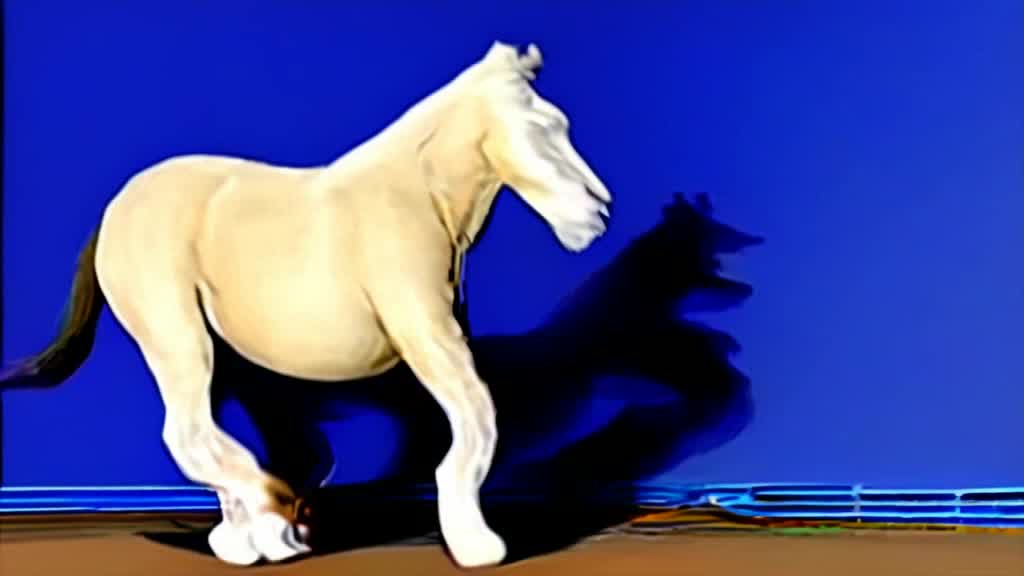}} \raisebox{-.5\height}{\includegraphics[width=0.075\textwidth]{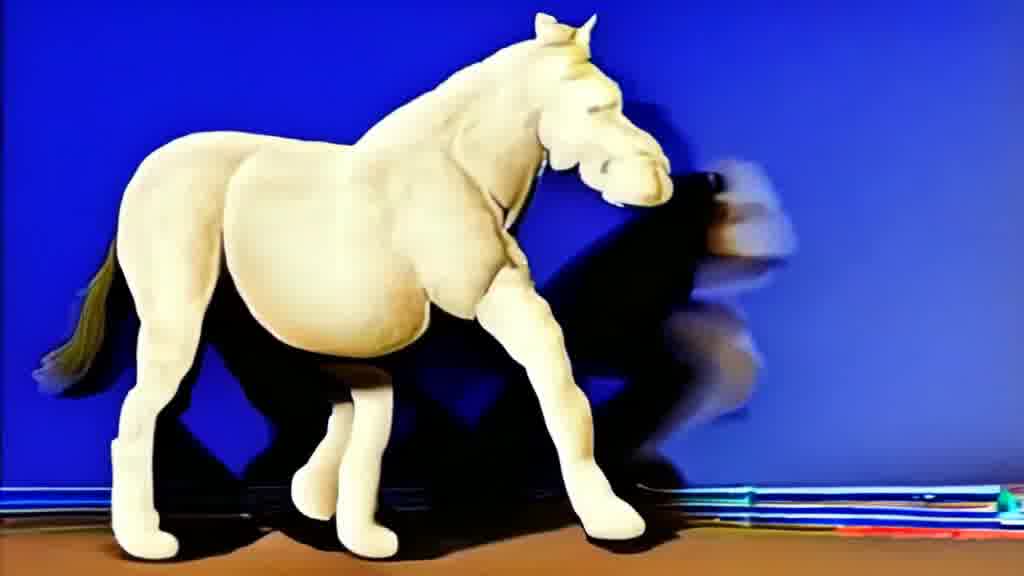}} \\
		{Uncond. Seed 1} & - & \raisebox{-.5\height}{\includegraphics[width=0.075\textwidth]{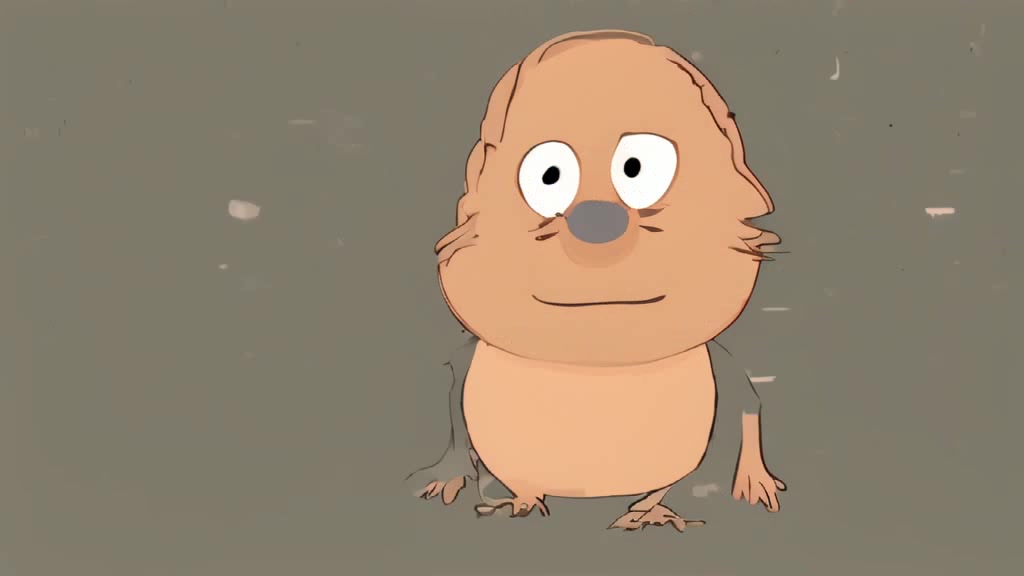}} \raisebox{-.5\height}{\includegraphics[width=0.075\textwidth]{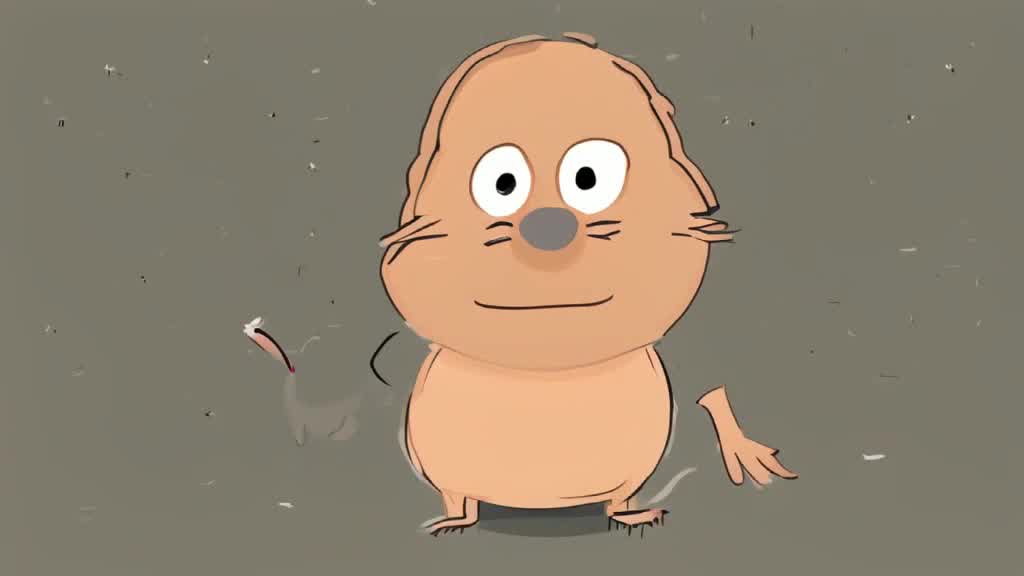}} \raisebox{-.5\height}{\includegraphics[width=0.075\textwidth]{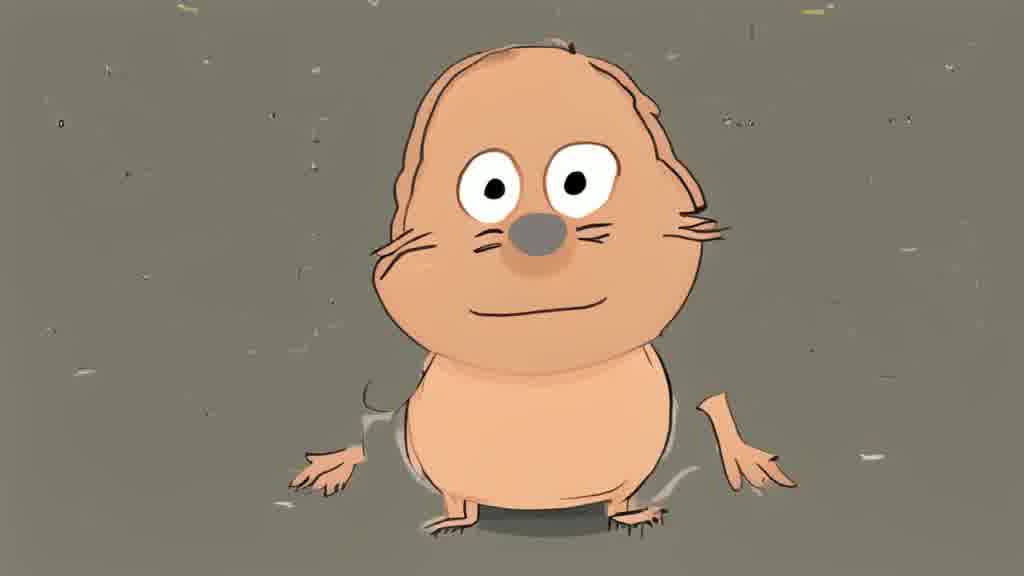}} & - & \raisebox{-.5\height}{\includegraphics[width=0.075\textwidth]{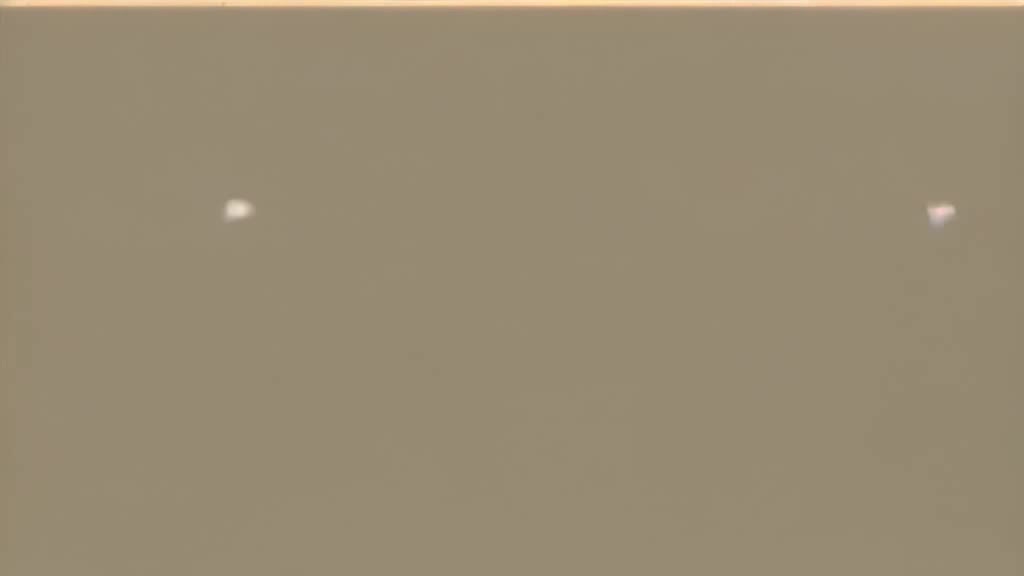}} \raisebox{-.5\height}{\includegraphics[width=0.075\textwidth]{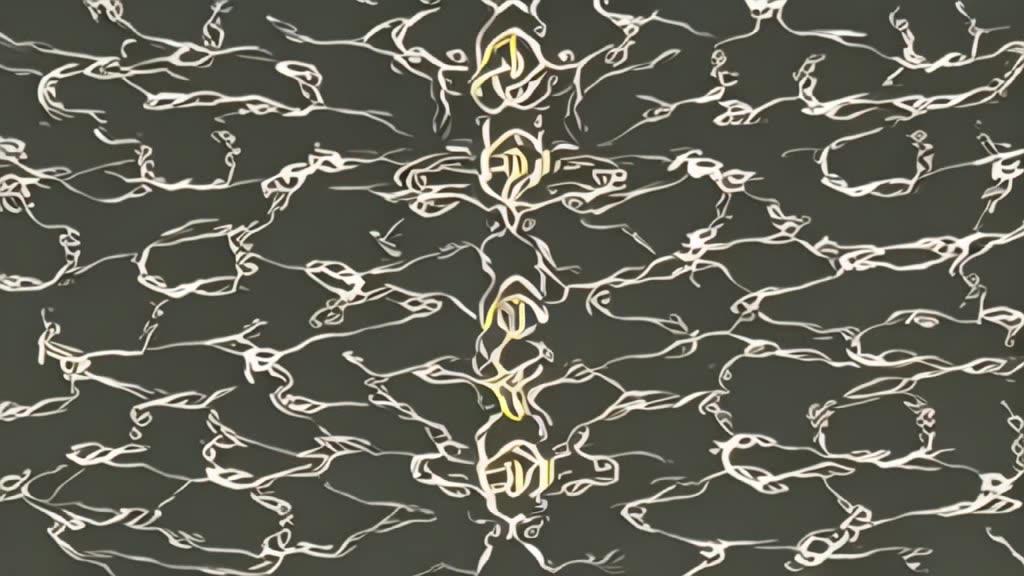}} \raisebox{-.5\height}{\includegraphics[width=0.075\textwidth]{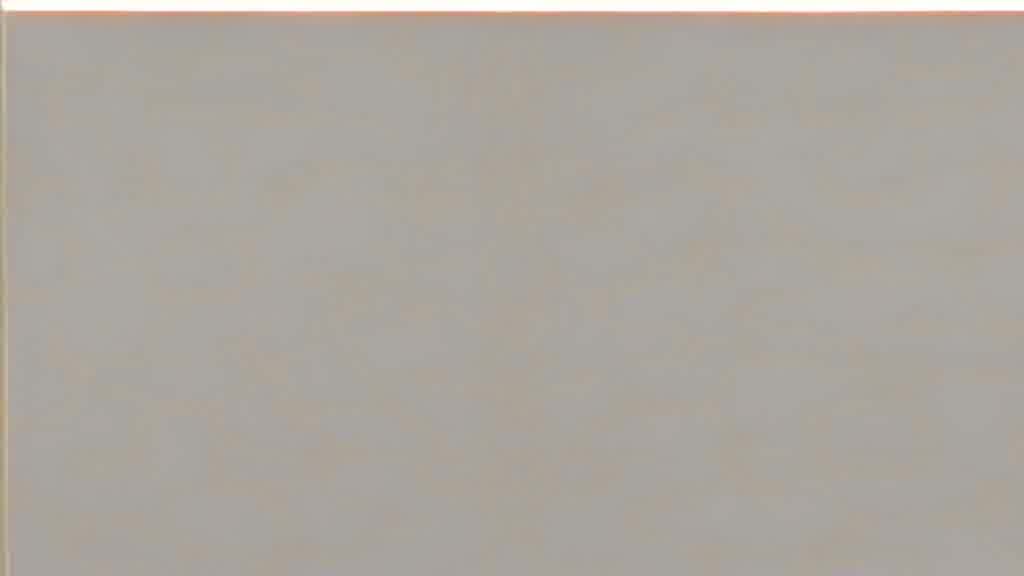}} \\
		{Uncond. Seed 2} & - & \raisebox{-.5\height}{\includegraphics[width=0.075\textwidth]{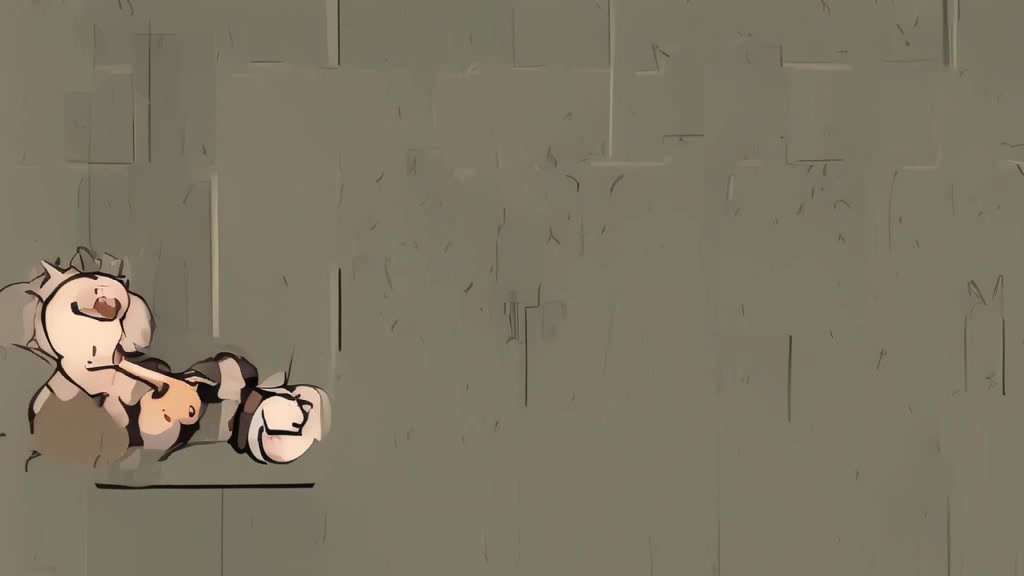}} \raisebox{-.5\height}{\includegraphics[width=0.075\textwidth]{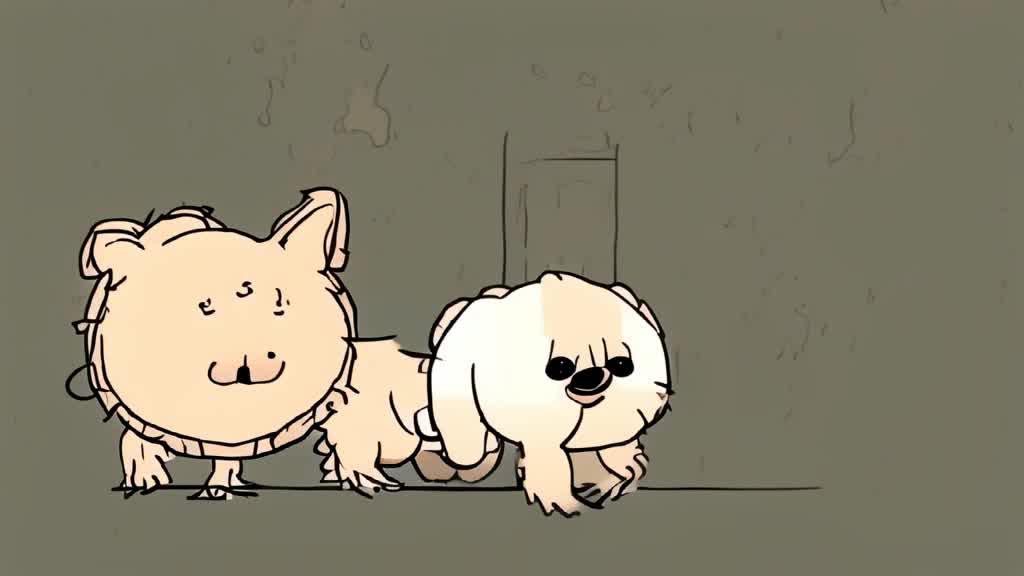}} \raisebox{-.5\height}{\includegraphics[width=0.075\textwidth]{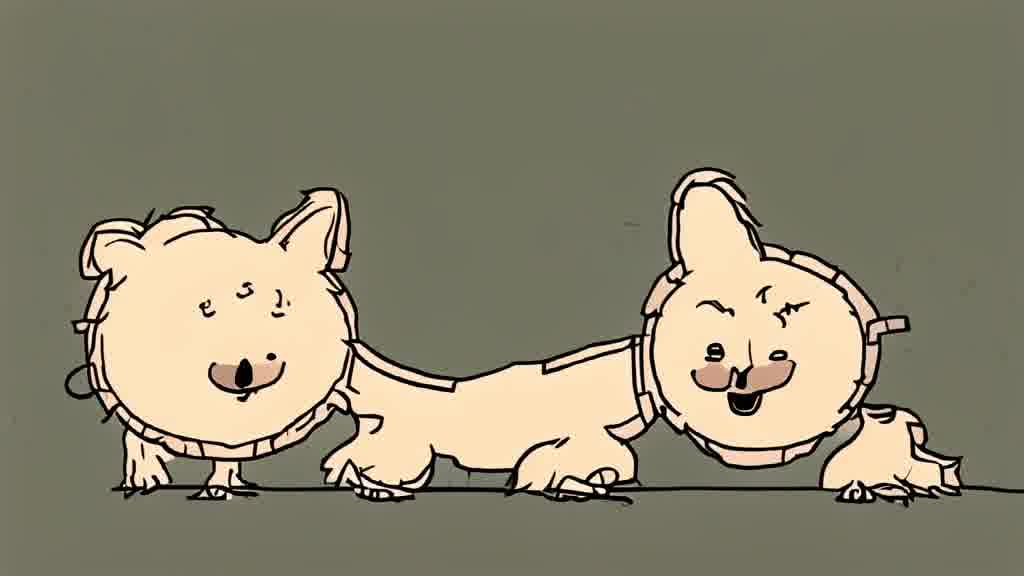}} & - & \raisebox{-.5\height}{\includegraphics[width=0.075\textwidth]{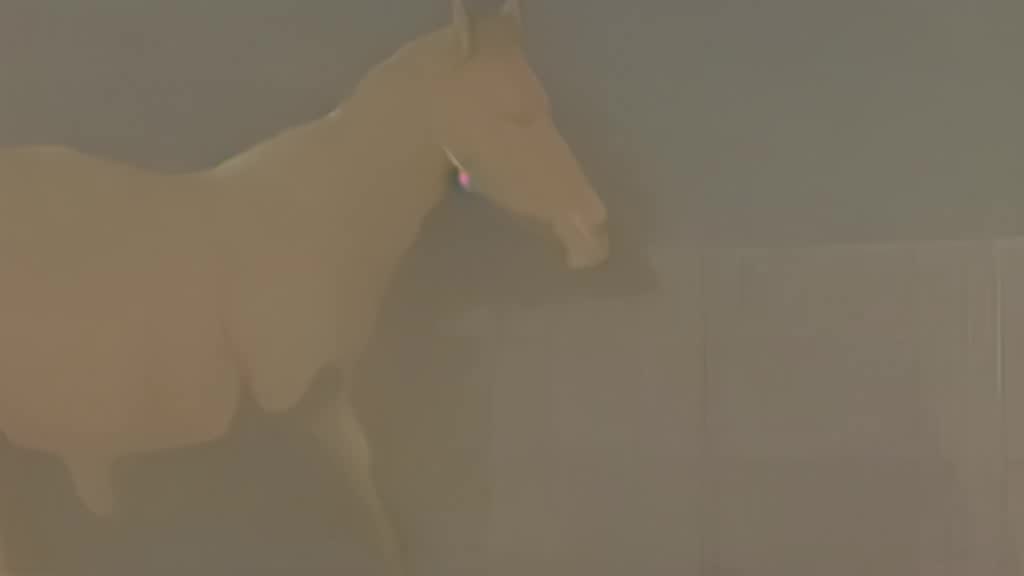}} \raisebox{-.5\height}{\includegraphics[width=0.075\textwidth]{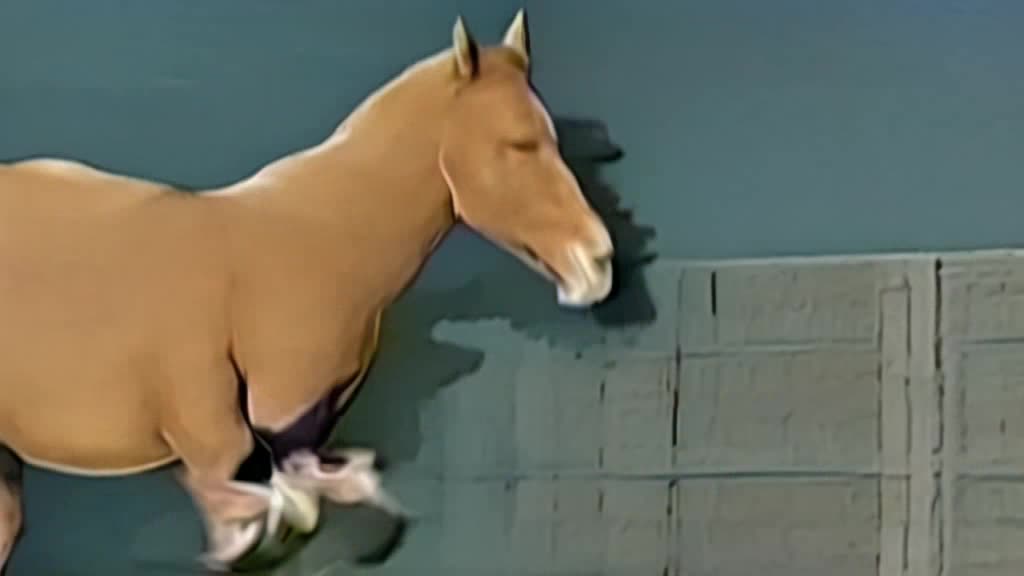}} \raisebox{-.5\height}{\includegraphics[width=0.075\textwidth]{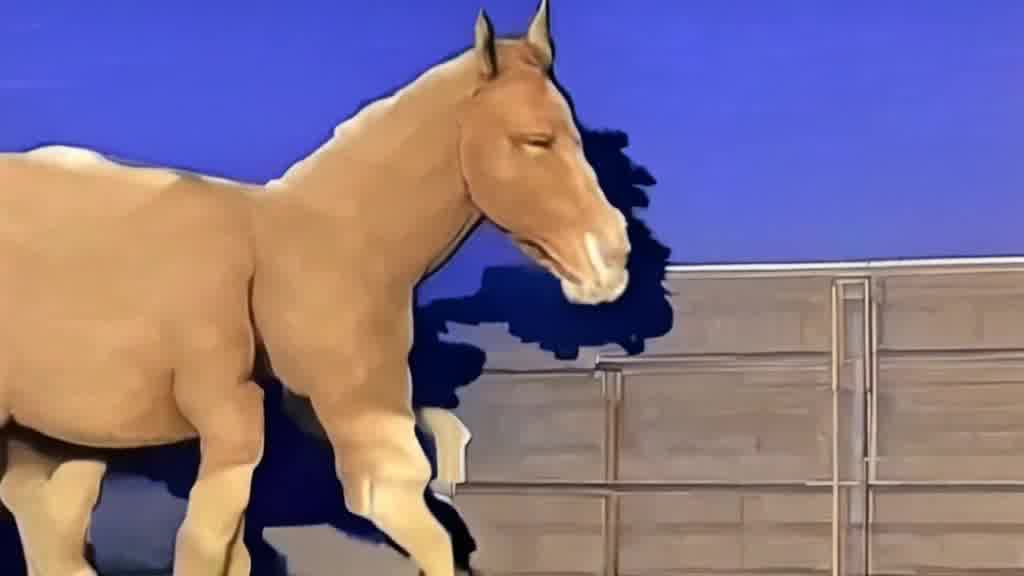}} \\
		{Ref.} & \raisebox{-.5\height}{\includegraphics[width=0.075\textwidth]{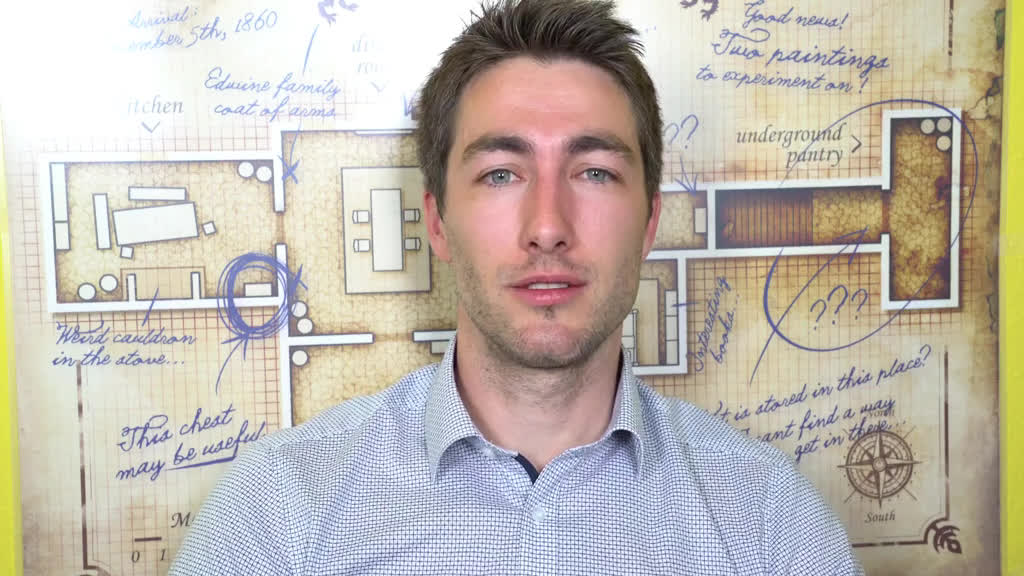}} & \raisebox{-.5\height}{\includegraphics[width=0.075\textwidth]{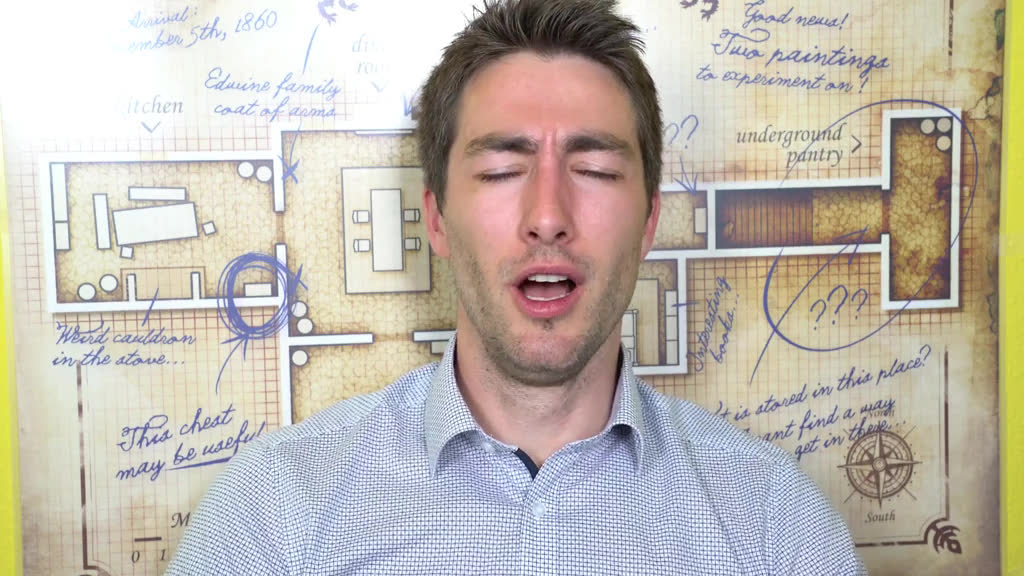}} \raisebox{-.5\height}{\includegraphics[width=0.075\textwidth]{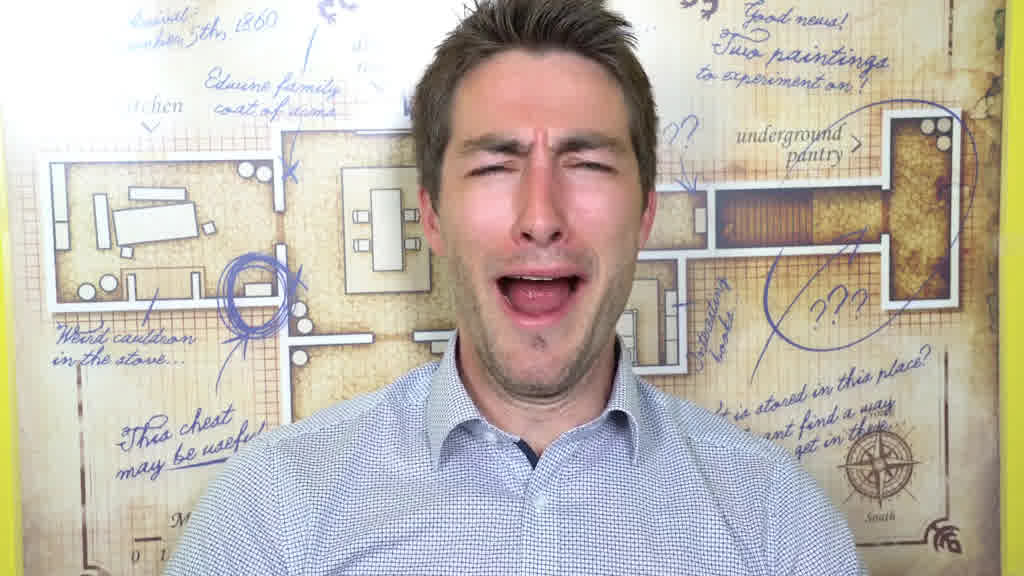}} \raisebox{-.5\height}{\includegraphics[width=0.075\textwidth]{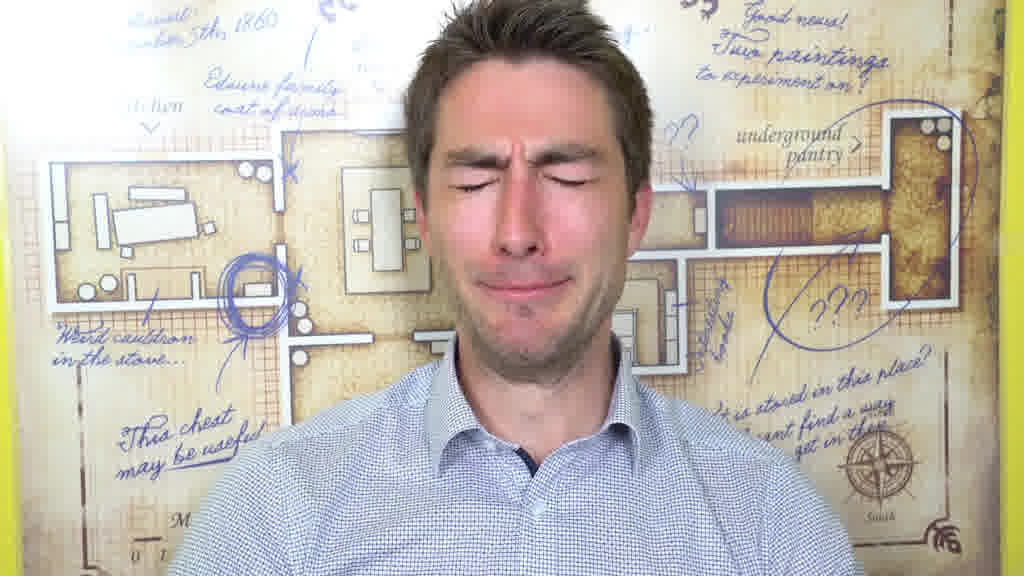}} & \raisebox{-.5\height}{\includegraphics[width=0.075\textwidth]{figures/svd_comparison/face_neutral/frames_input/001.jpg}} & \raisebox{-.5\height}{\includegraphics[width=0.075\textwidth]{figures/svd_comparison/face_neutral/frames_input/005.jpg}} \raisebox{-.5\height}{\includegraphics[width=0.075\textwidth]{figures/svd_comparison/face_neutral/frames_input/009.jpg}} \raisebox{-.5\height}{\includegraphics[width=0.075\textwidth]{figures/svd_comparison/face_neutral/frames_input/014.jpg}} \\
		{Seed 0} & \raisebox{-.5\height}{\includegraphics[width=0.075\textwidth]{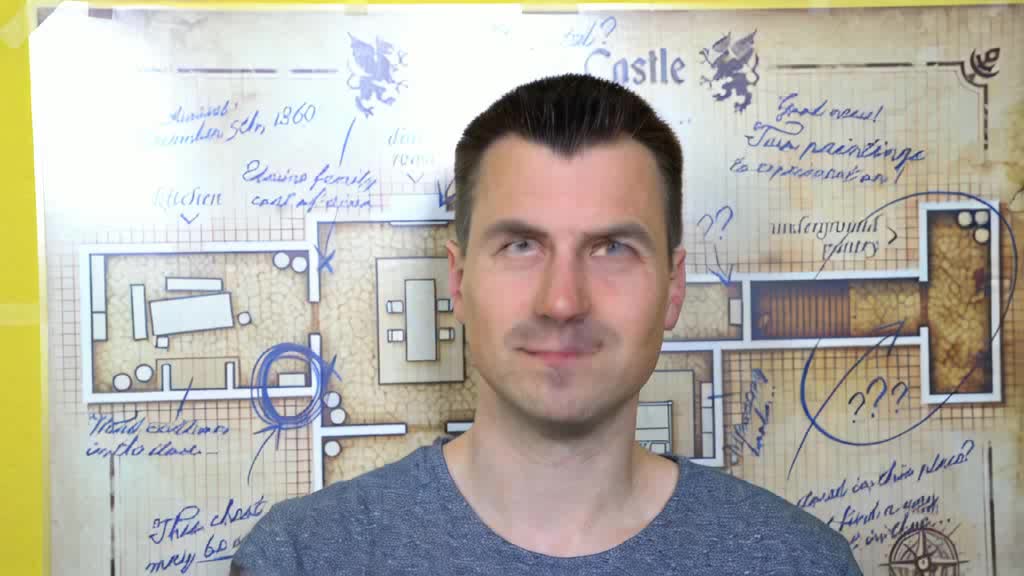}} & \raisebox{-.5\height}{\includegraphics[width=0.075\textwidth]{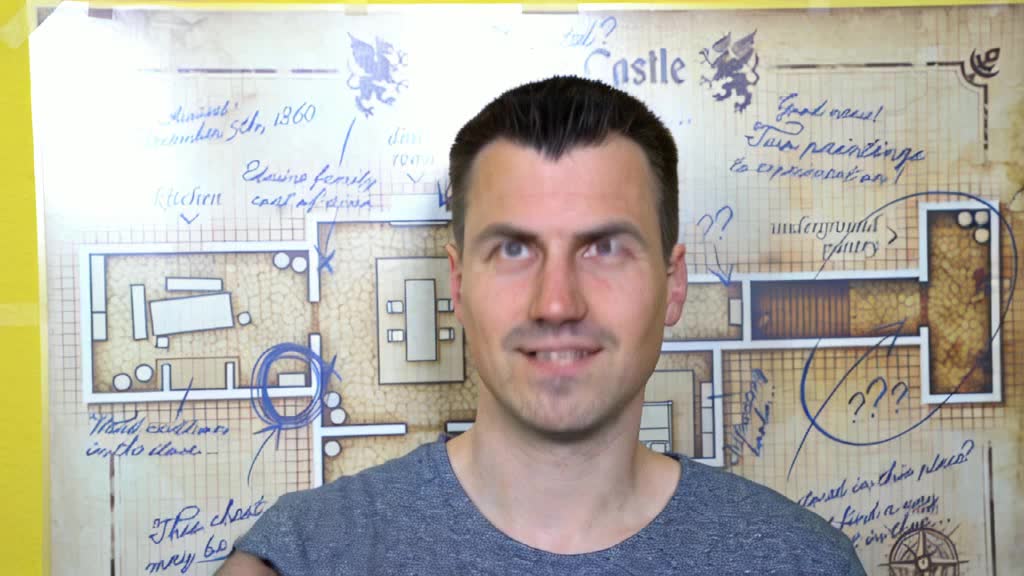}} \raisebox{-.5\height}{\includegraphics[width=0.075\textwidth]{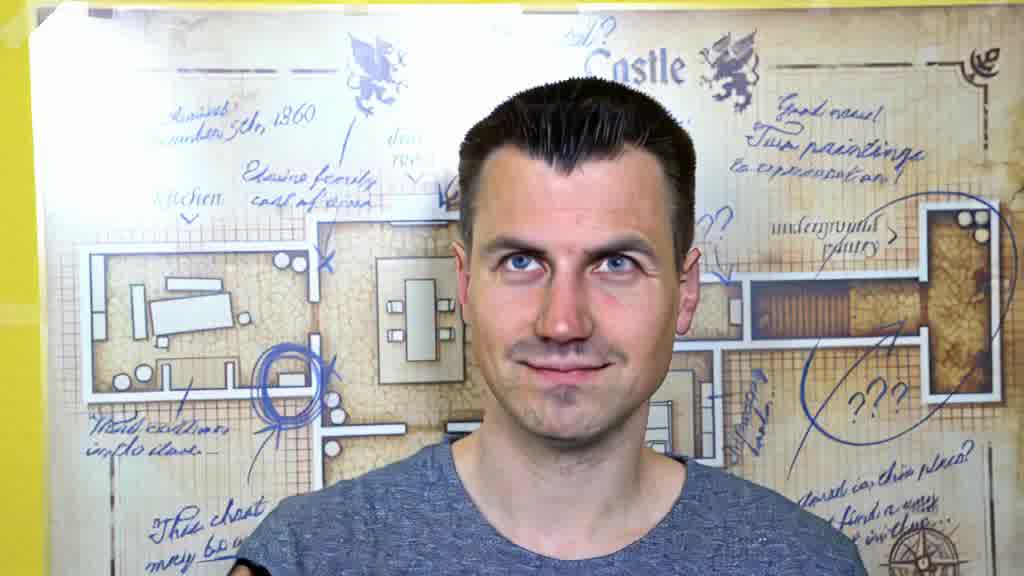}} \raisebox{-.5\height}{\includegraphics[width=0.075\textwidth]{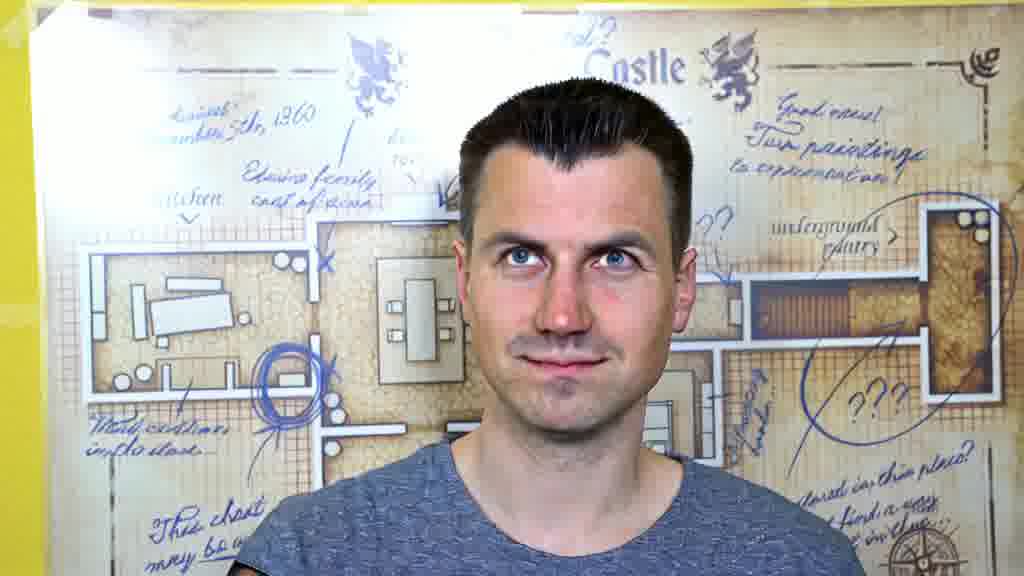}} & \raisebox{-.5\height}{\includegraphics[width=0.075\textwidth]{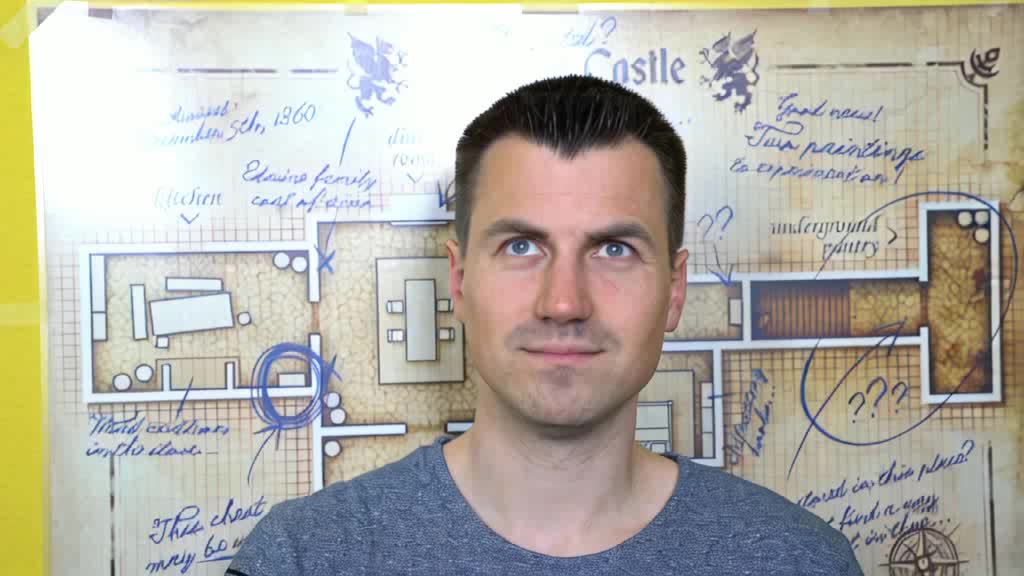}} & \raisebox{-.5\height}{\includegraphics[width=0.075\textwidth]{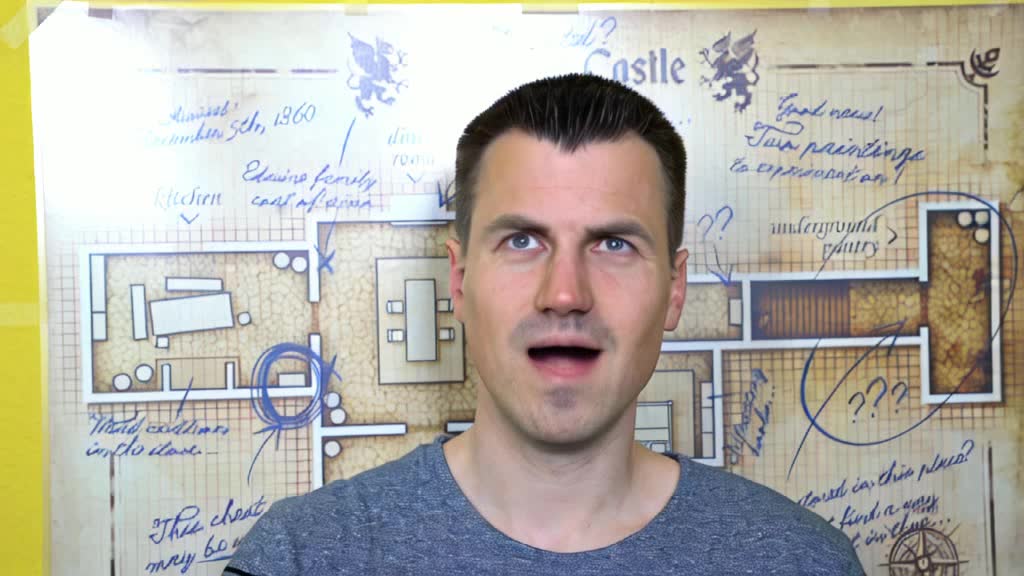}} \raisebox{-.5\height}{\includegraphics[width=0.075\textwidth]{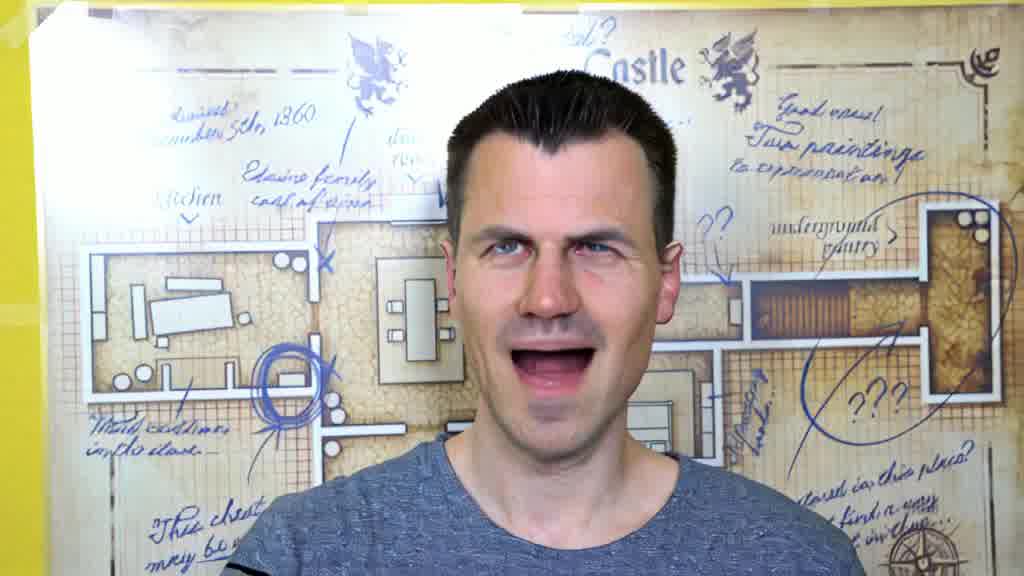}} \raisebox{-.5\height}{\includegraphics[width=0.075\textwidth]{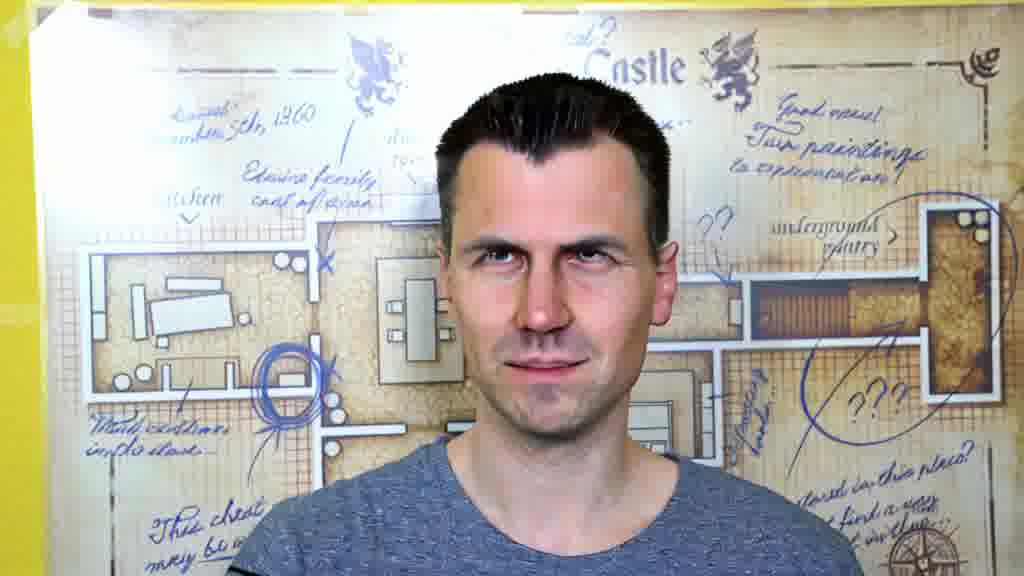}} \\
		{Seed 1} & \raisebox{-.5\height}{\includegraphics[width=0.075\textwidth]{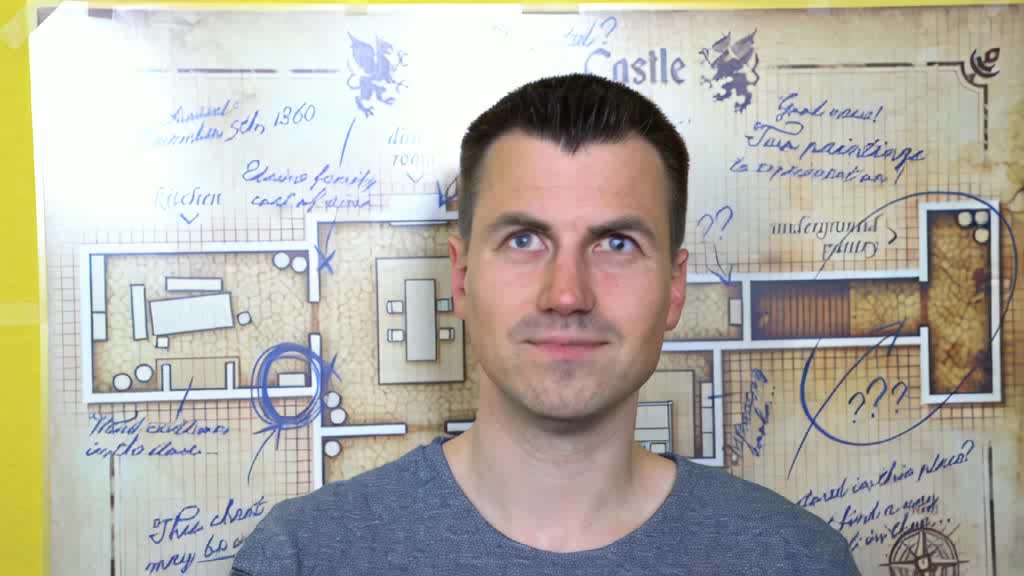}} & \raisebox{-.5\height}{\includegraphics[width=0.075\textwidth]{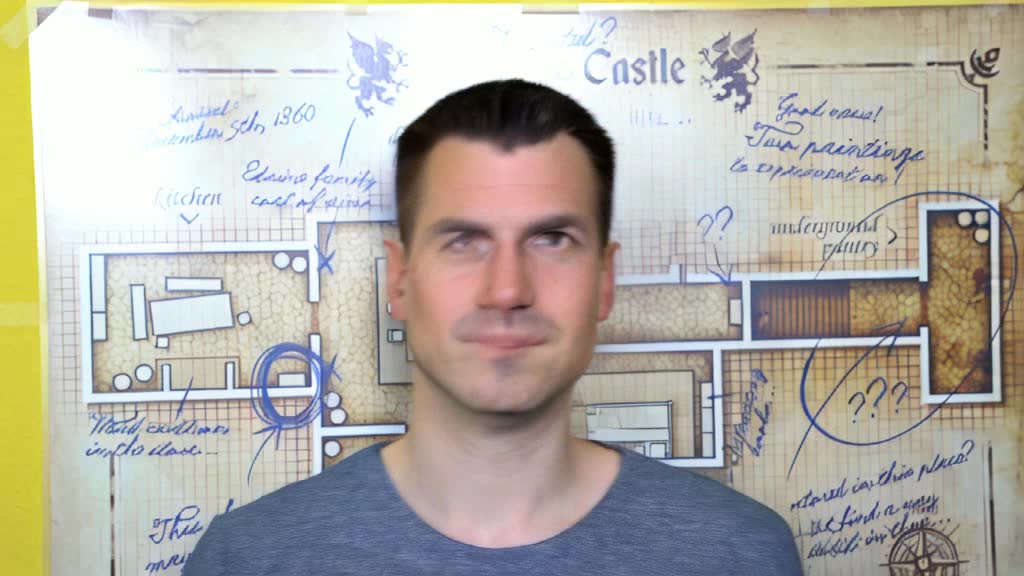}} \raisebox{-.5\height}{\includegraphics[width=0.075\textwidth]{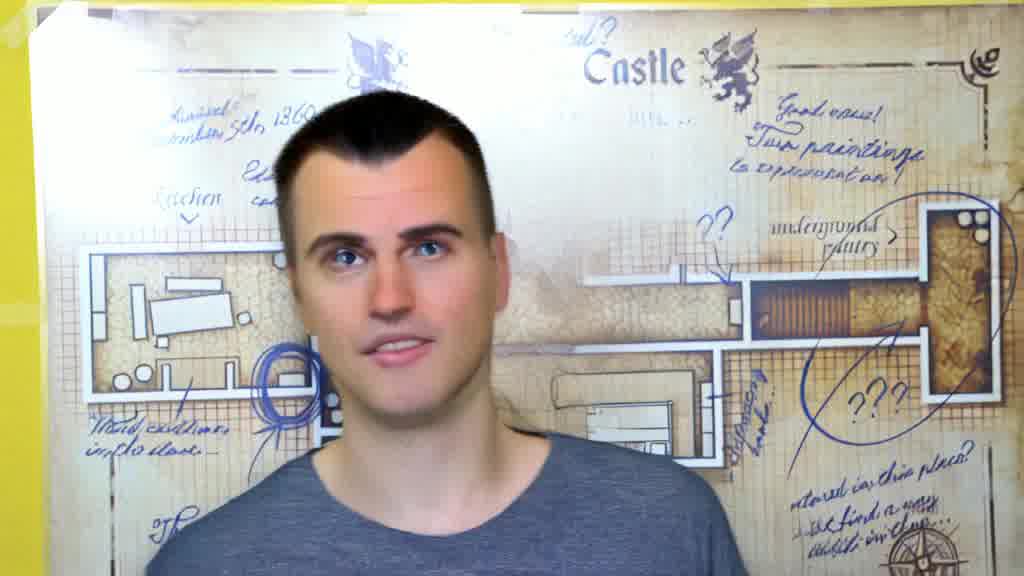}} \raisebox{-.5\height}{\includegraphics[width=0.075\textwidth]{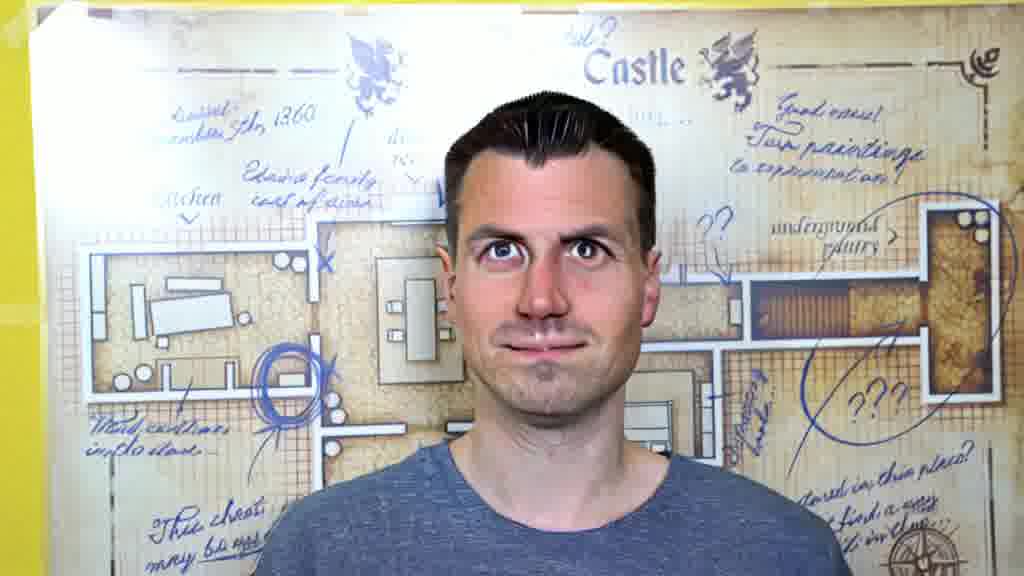}} & \raisebox{-.5\height}{\includegraphics[width=0.075\textwidth]{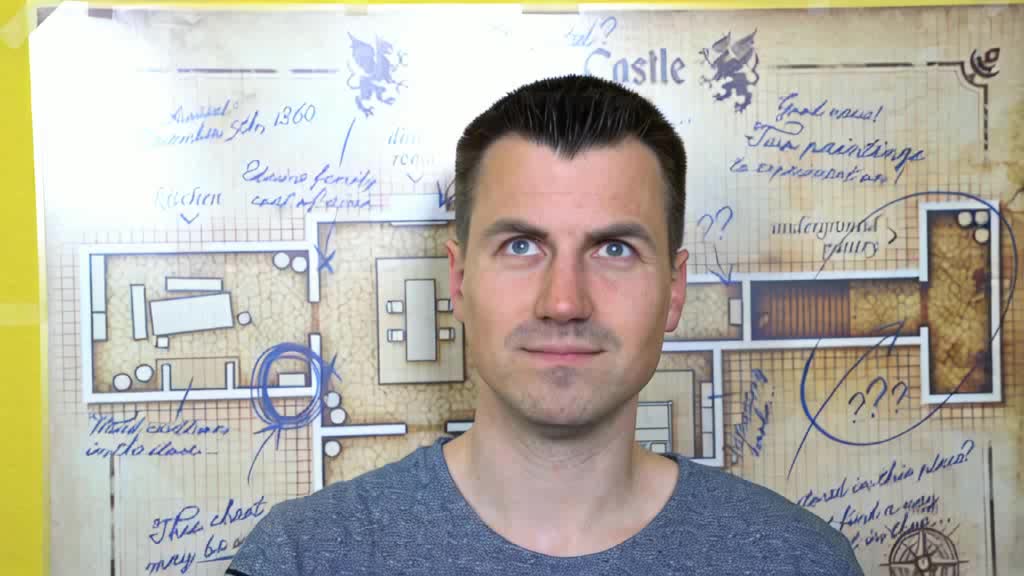}} & \raisebox{-.5\height}{\includegraphics[width=0.075\textwidth]{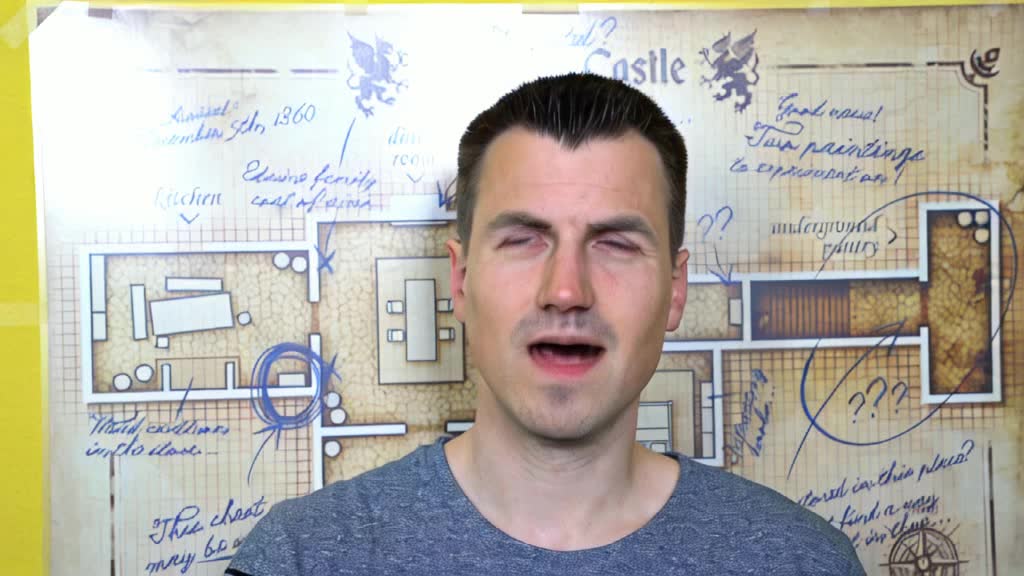}} \raisebox{-.5\height}{\includegraphics[width=0.075\textwidth]{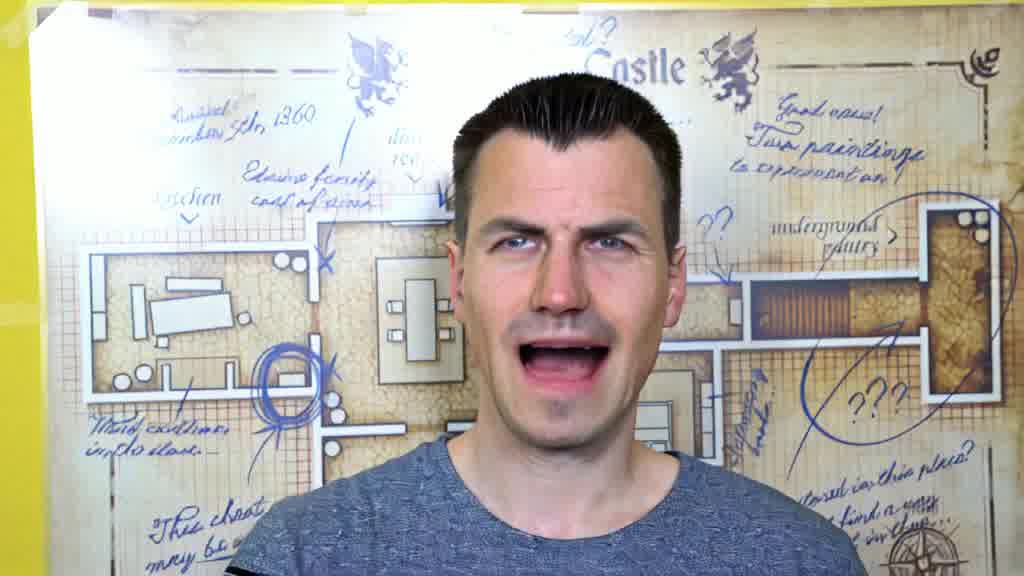}} \raisebox{-.5\height}{\includegraphics[width=0.075\textwidth]{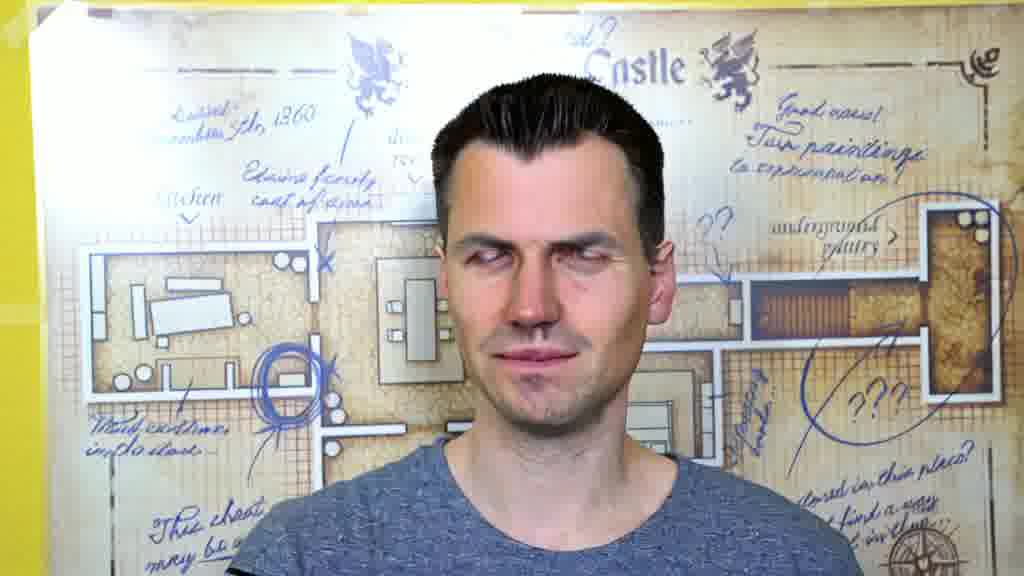}} \\
		{Seed 2} & \raisebox{-.5\height}{\includegraphics[width=0.075\textwidth]{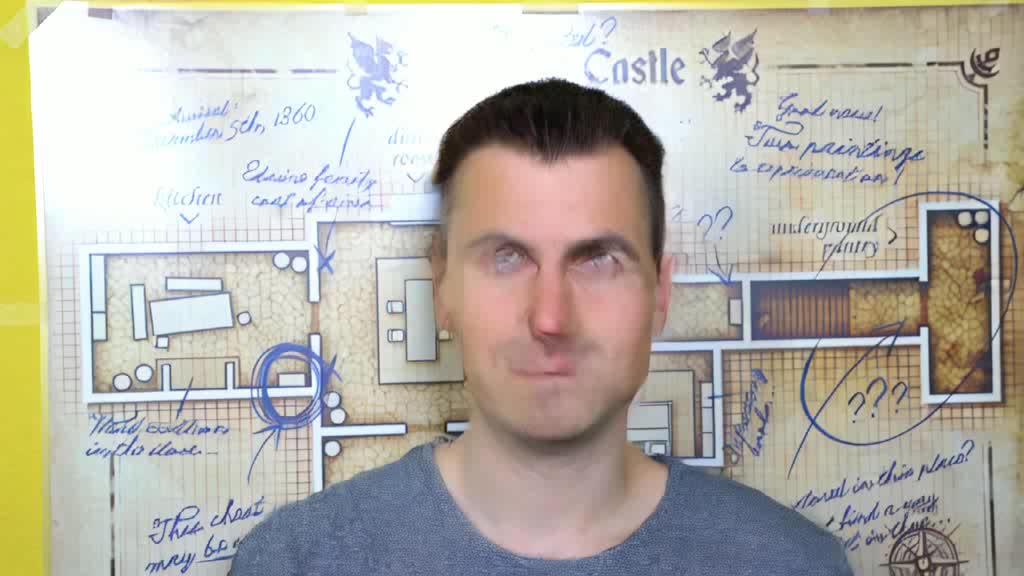}} & \raisebox{-.5\height}{\includegraphics[width=0.075\textwidth]{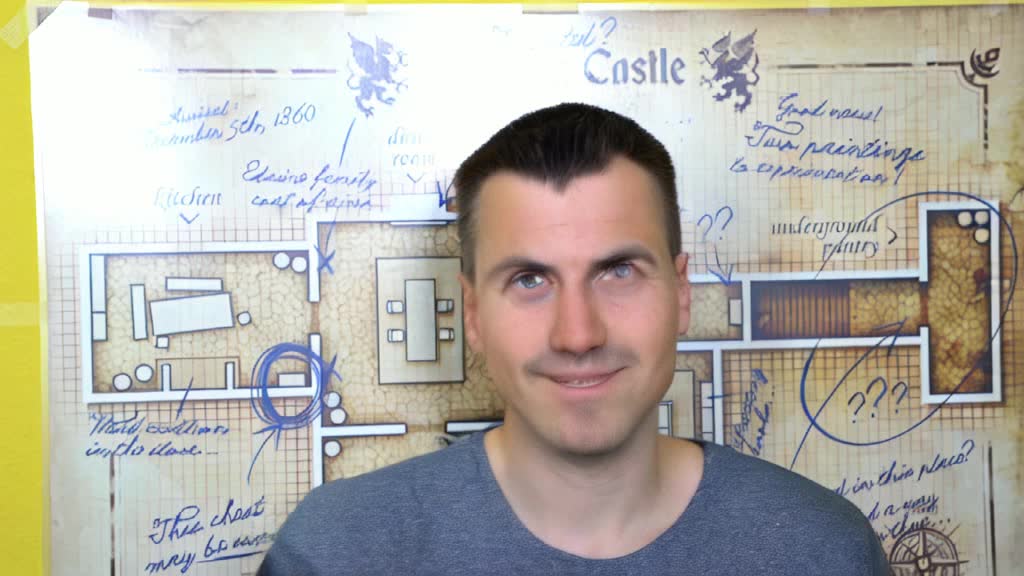}} \raisebox{-.5\height}{\includegraphics[width=0.075\textwidth]{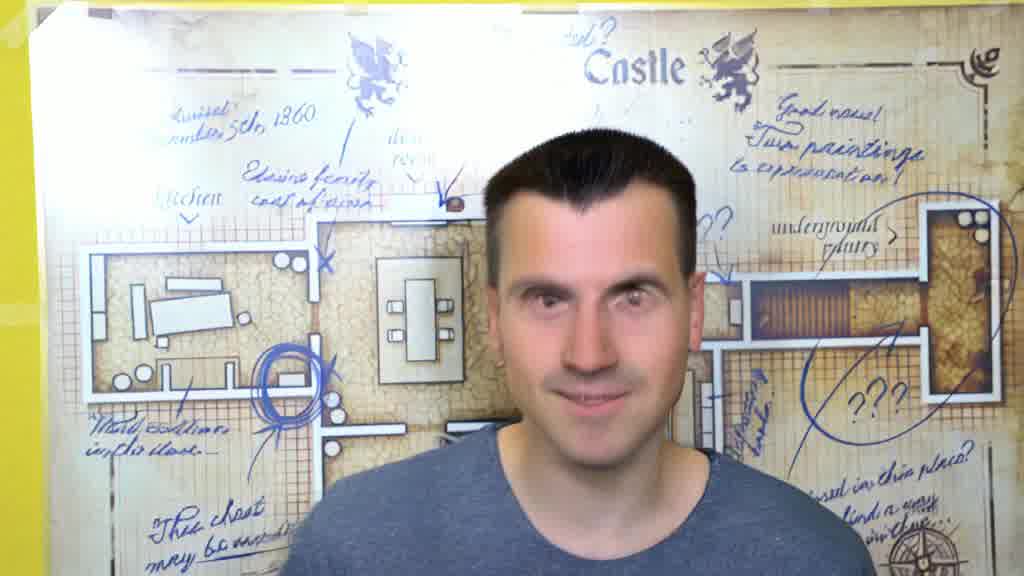}} \raisebox{-.5\height}{\includegraphics[width=0.075\textwidth]{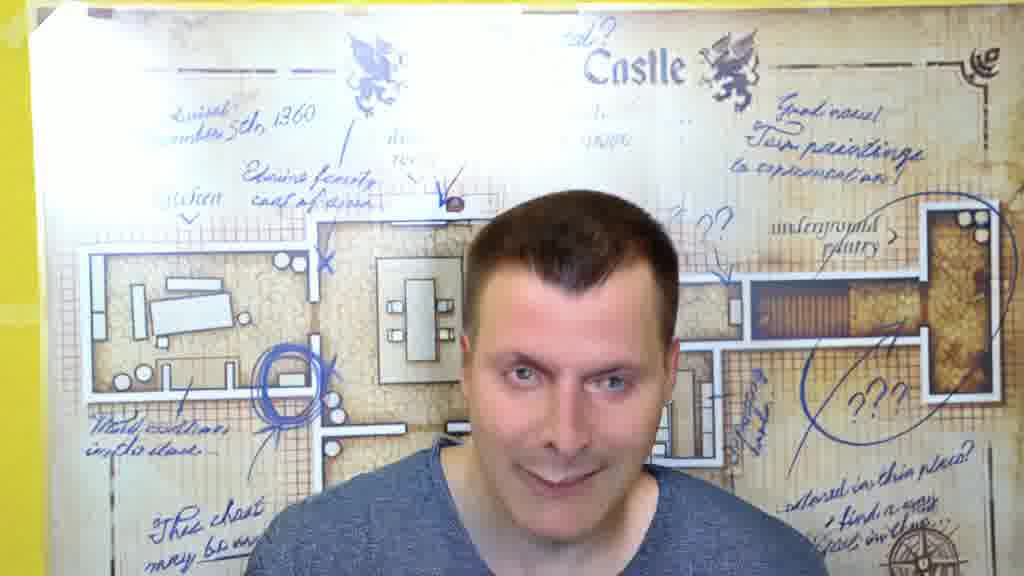}} & \raisebox{-.5\height}{\includegraphics[width=0.075\textwidth]{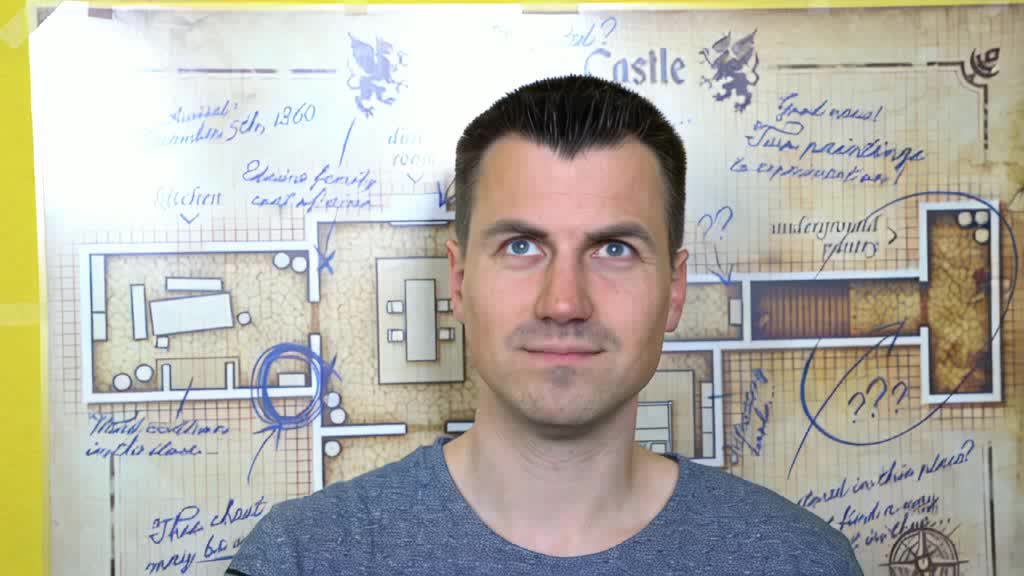}} & \raisebox{-.5\height}{\includegraphics[width=0.075\textwidth]{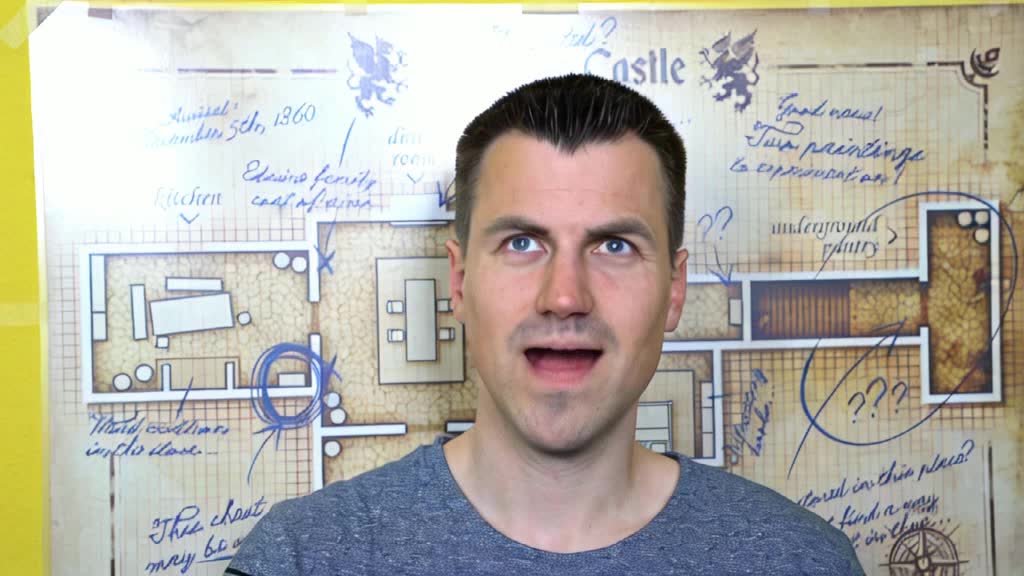}} \raisebox{-.5\height}{\includegraphics[width=0.075\textwidth]{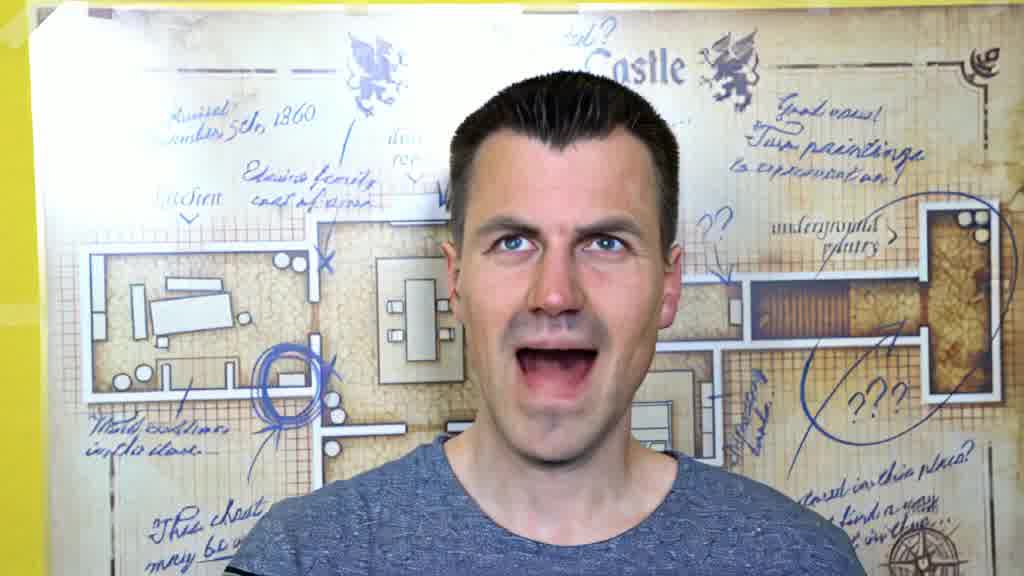}} \raisebox{-.5\height}{\includegraphics[width=0.075\textwidth]{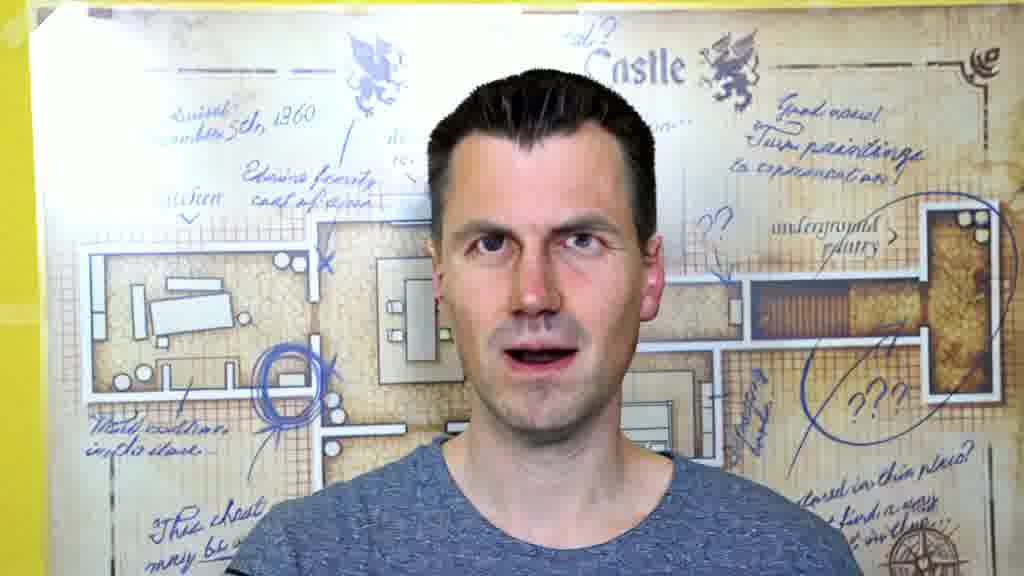}} \\
		{Uncond. Seed 0} & - & \raisebox{-.5\height}{\includegraphics[width=0.075\textwidth]{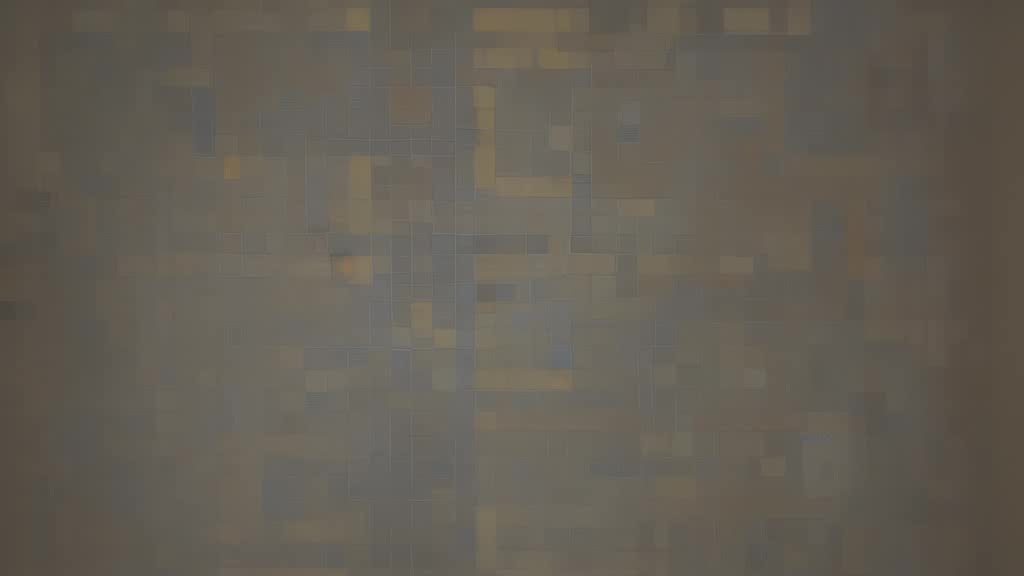}} \raisebox{-.5\height}{\includegraphics[width=0.075\textwidth]{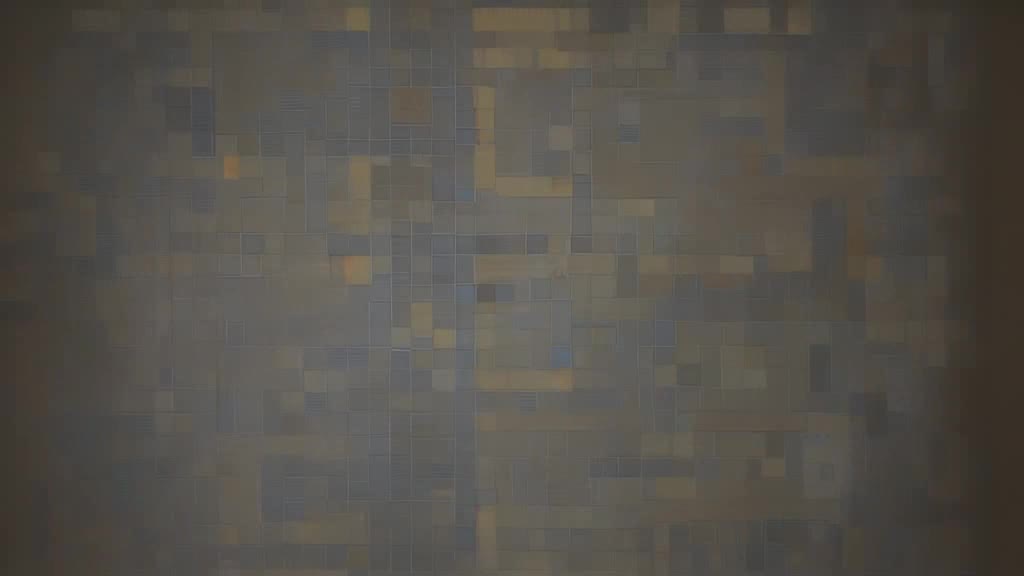}} \raisebox{-.5\height}{\includegraphics[width=0.075\textwidth]{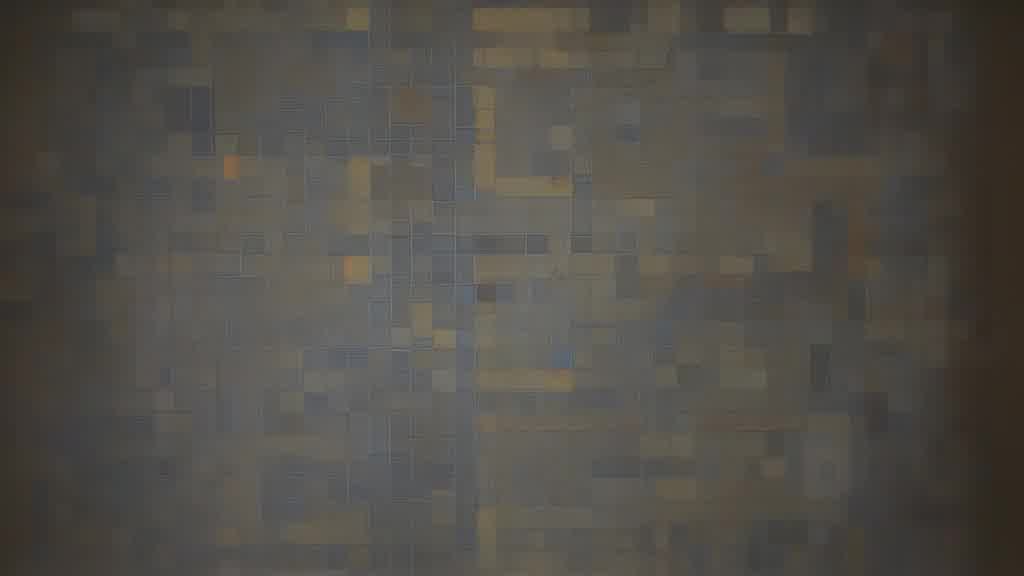}} & - & \raisebox{-.5\height}{\includegraphics[width=0.075\textwidth]{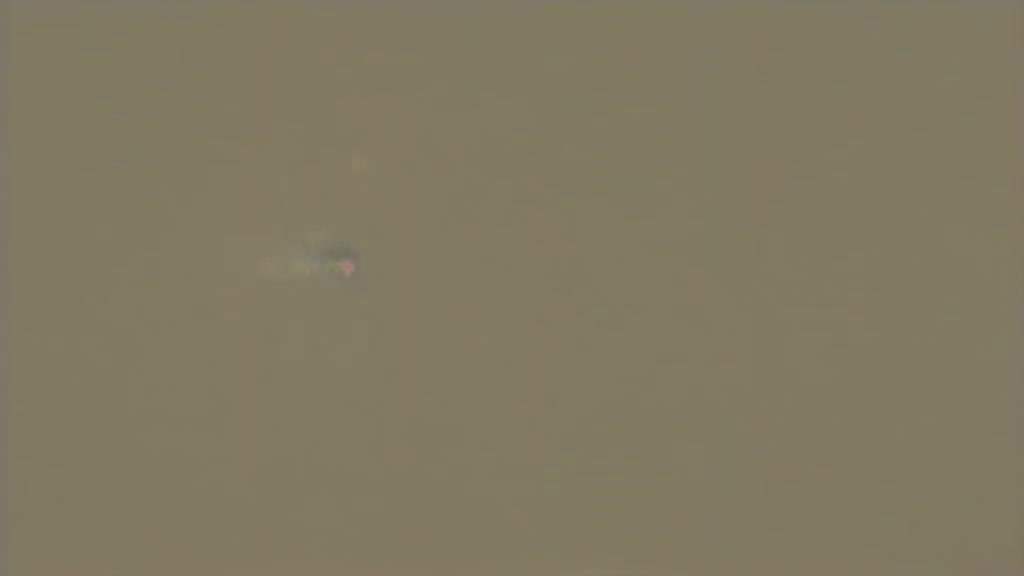}} \raisebox{-.5\height}{\includegraphics[width=0.075\textwidth]{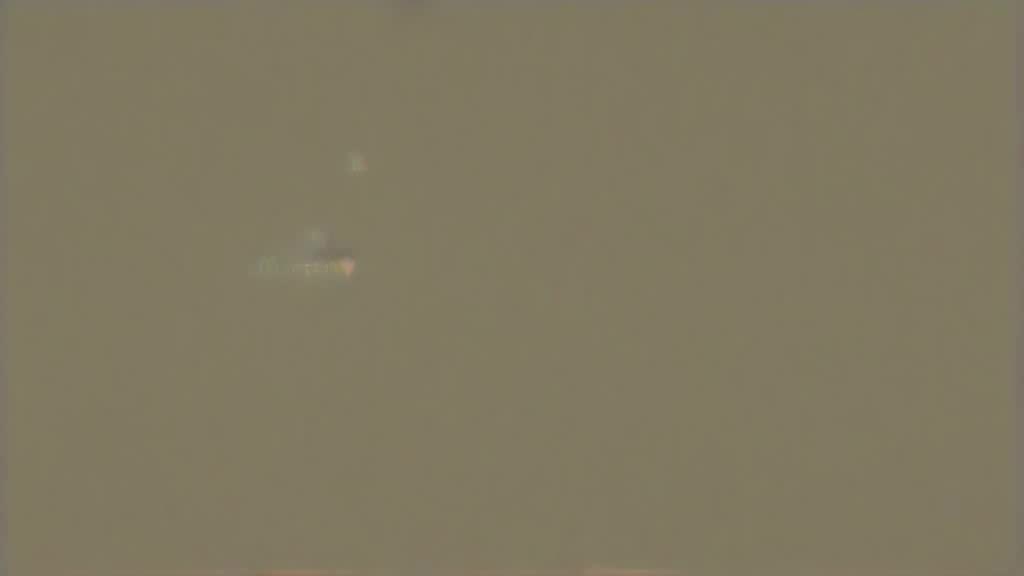}} \raisebox{-.5\height}{\includegraphics[width=0.075\textwidth]{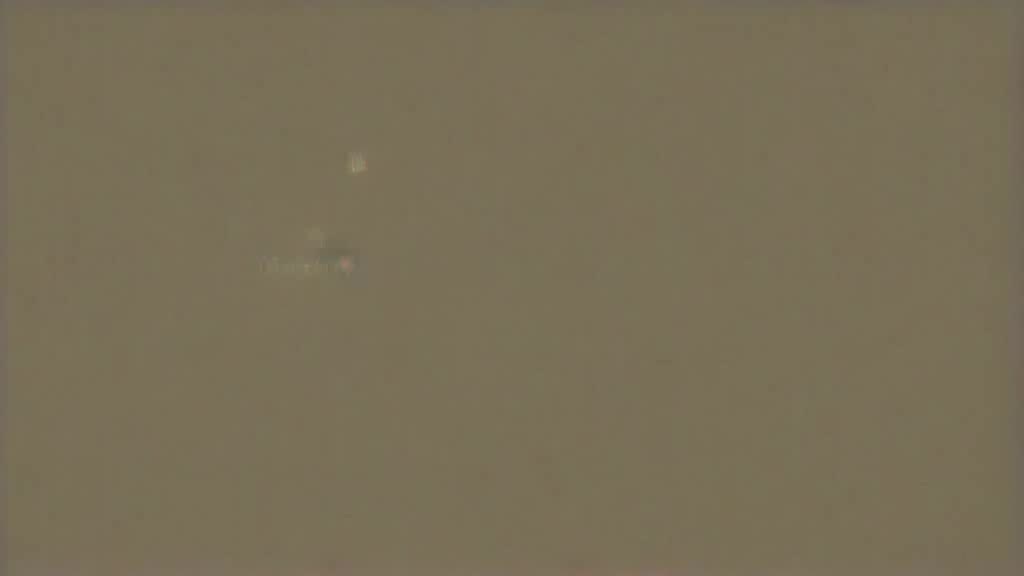}} \\
		{Uncond. Seed 1} & - & \raisebox{-.5\height}{\includegraphics[width=0.075\textwidth]{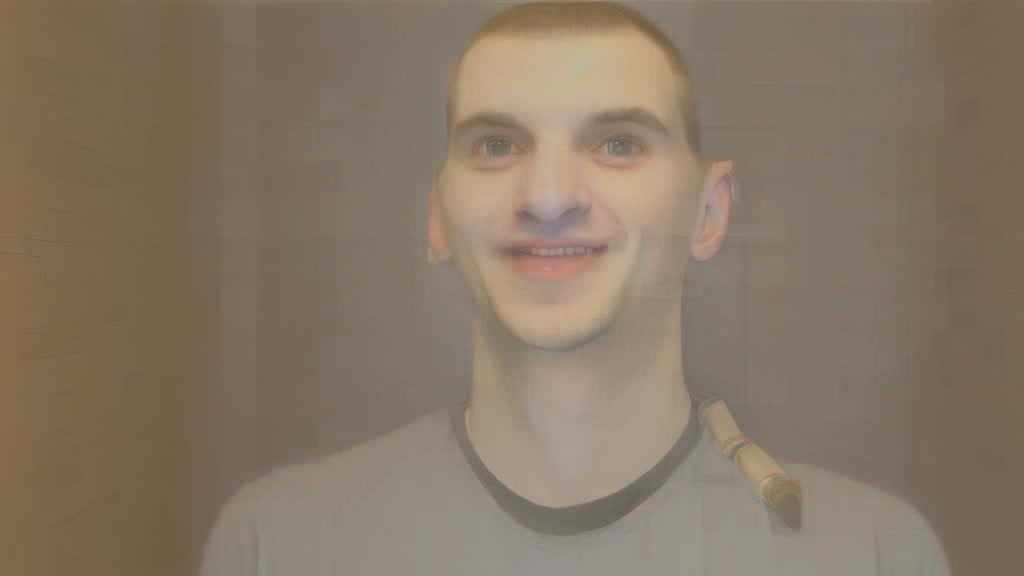}} \raisebox{-.5\height}{\includegraphics[width=0.075\textwidth]{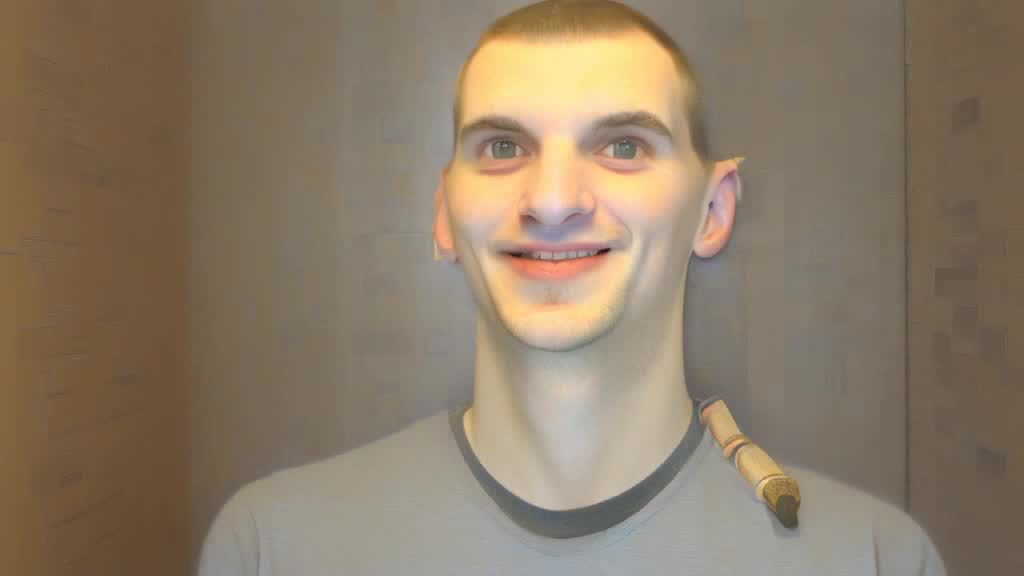}} \raisebox{-.5\height}{\includegraphics[width=0.075\textwidth]{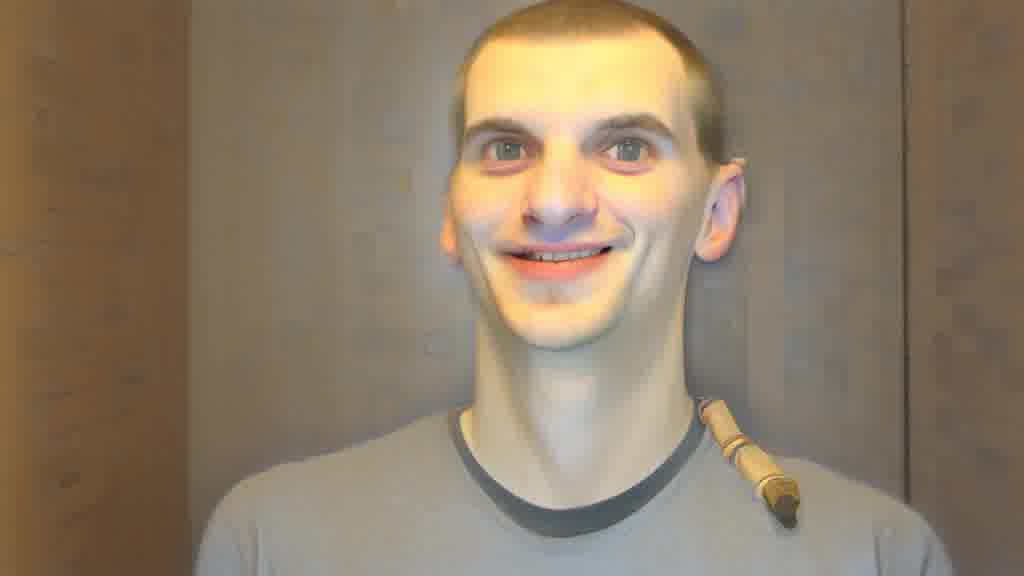}} & - & \raisebox{-.5\height}{\includegraphics[width=0.075\textwidth]{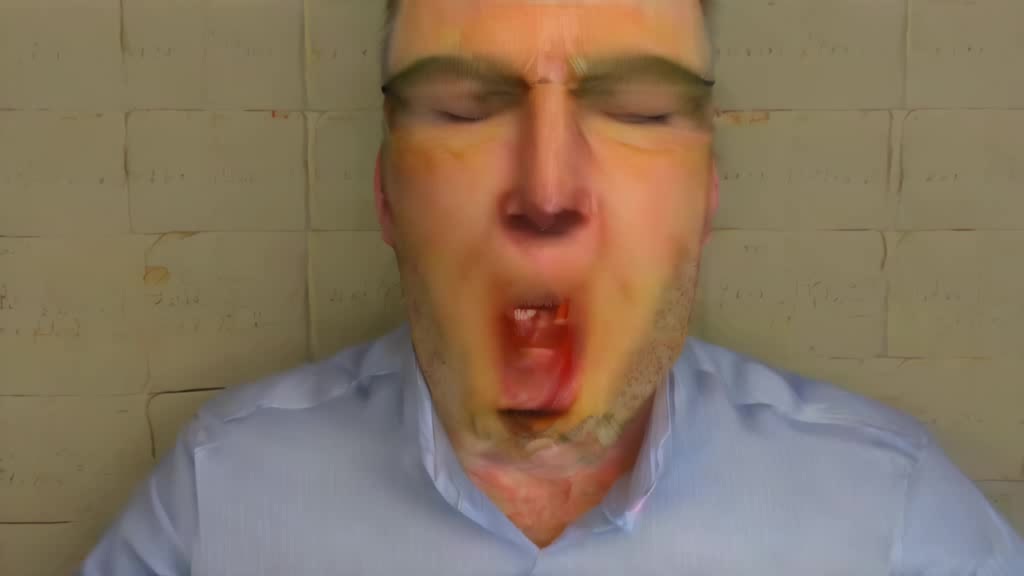}} \raisebox{-.5\height}{\includegraphics[width=0.075\textwidth]{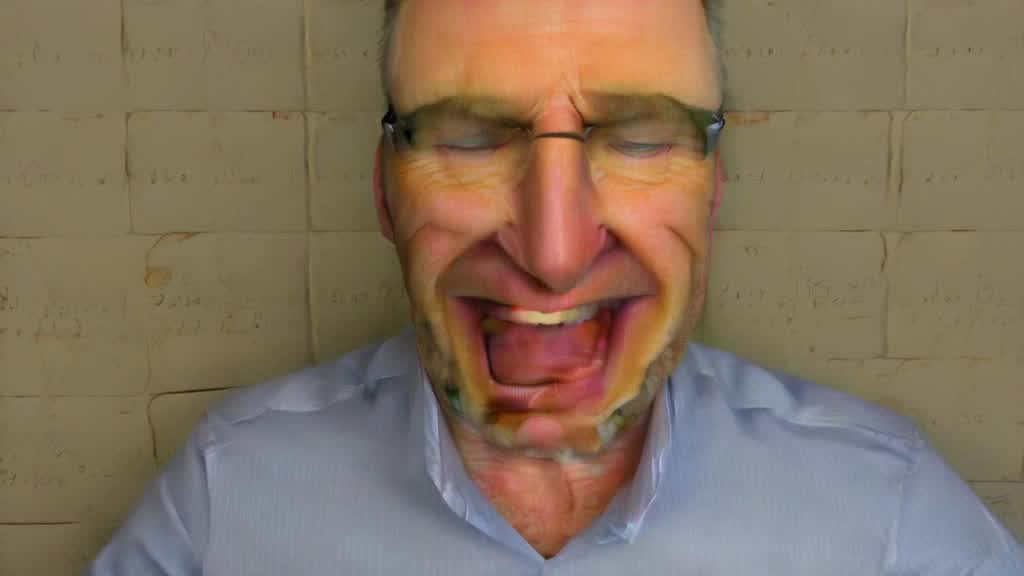}} \raisebox{-.5\height}{\includegraphics[width=0.075\textwidth]{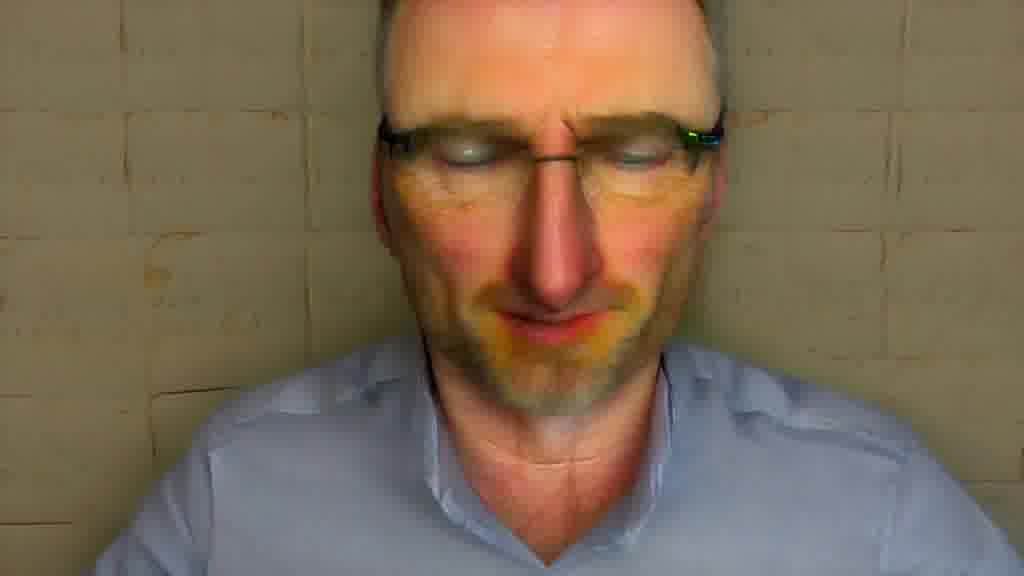}} \\
		{Uncond. Seed 2} & - & \raisebox{-.5\height}{\includegraphics[width=0.075\textwidth]{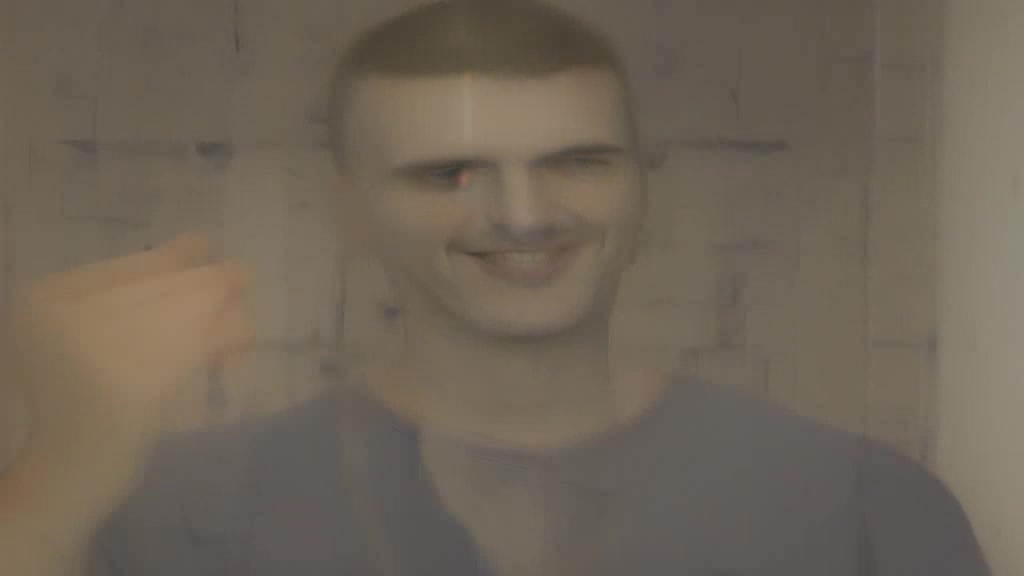}} \raisebox{-.5\height}{\includegraphics[width=0.075\textwidth]{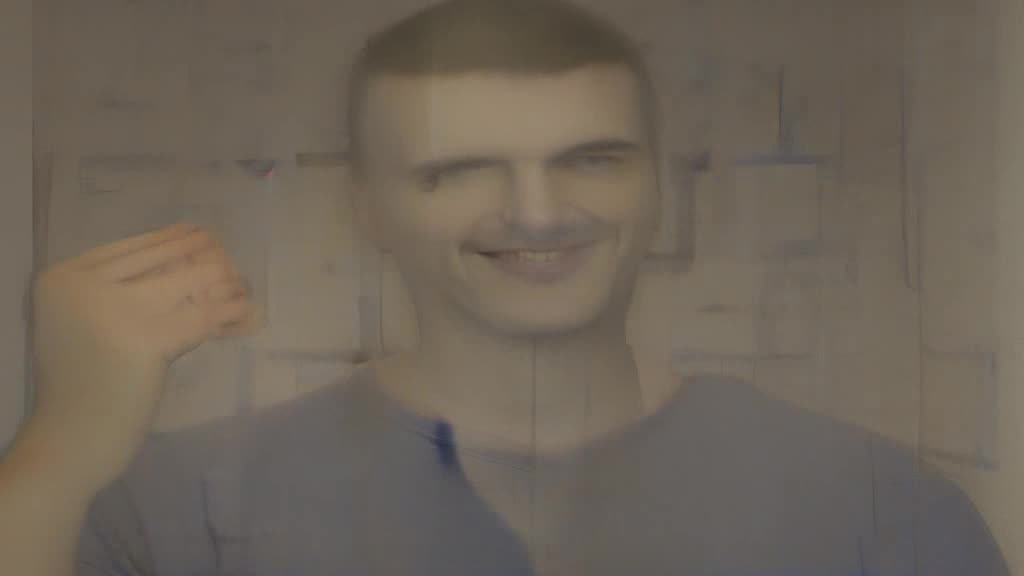}} \raisebox{-.5\height}{\includegraphics[width=0.075\textwidth]{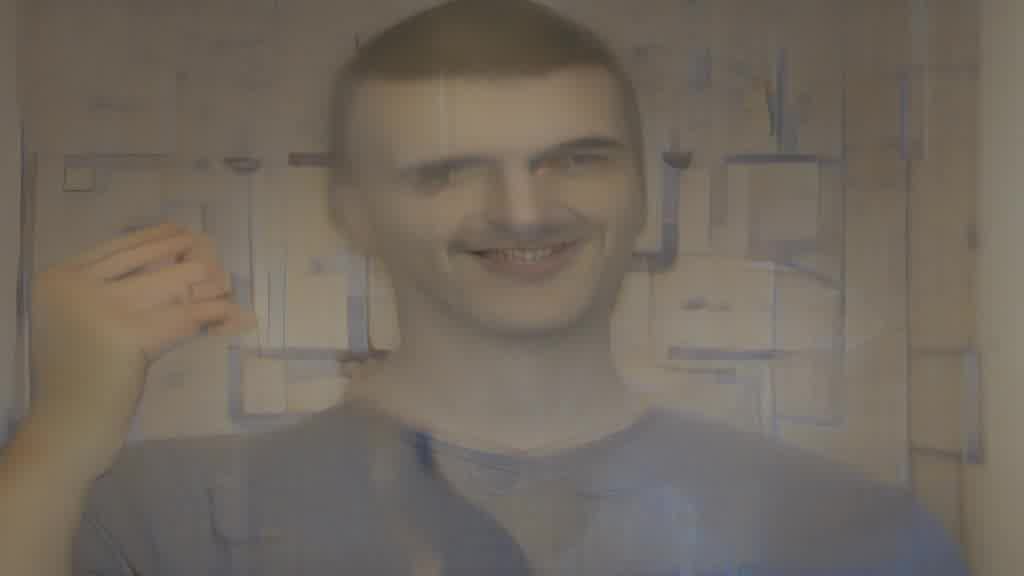}} & - & \raisebox{-.5\height}{\includegraphics[width=0.075\textwidth]{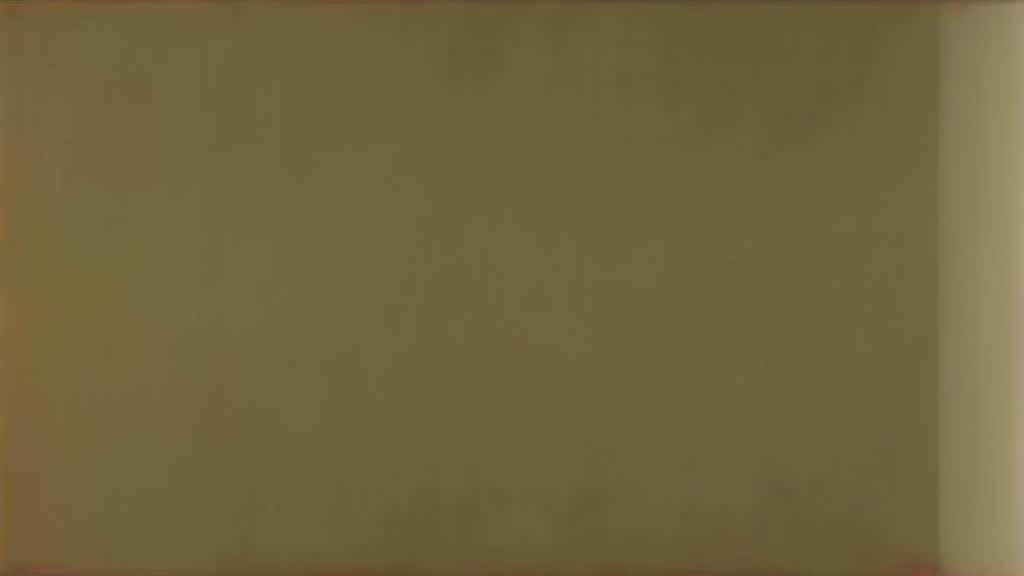}} \raisebox{-.5\height}{\includegraphics[width=0.075\textwidth]{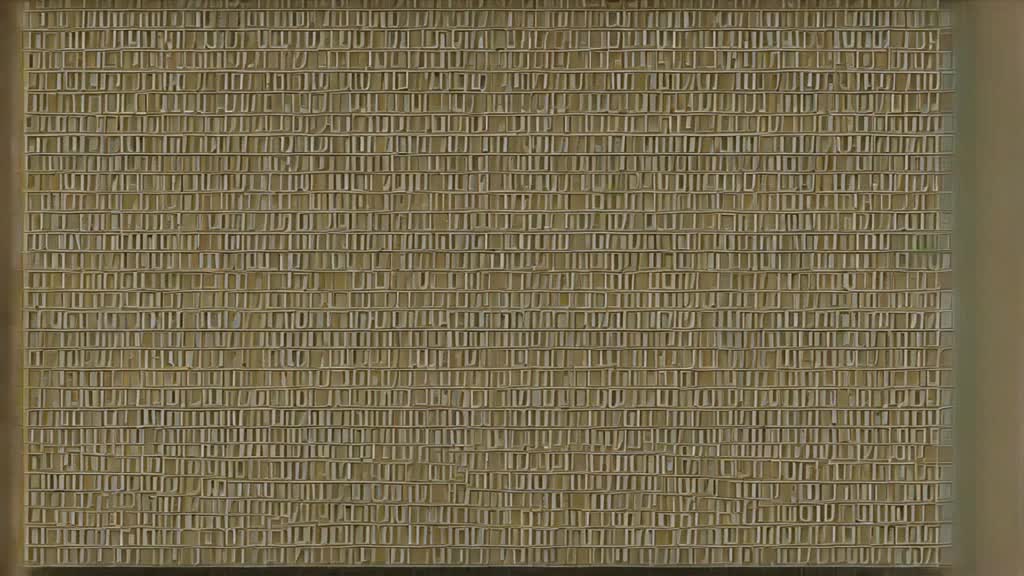}} \raisebox{-.5\height}{\includegraphics[width=0.075\textwidth]{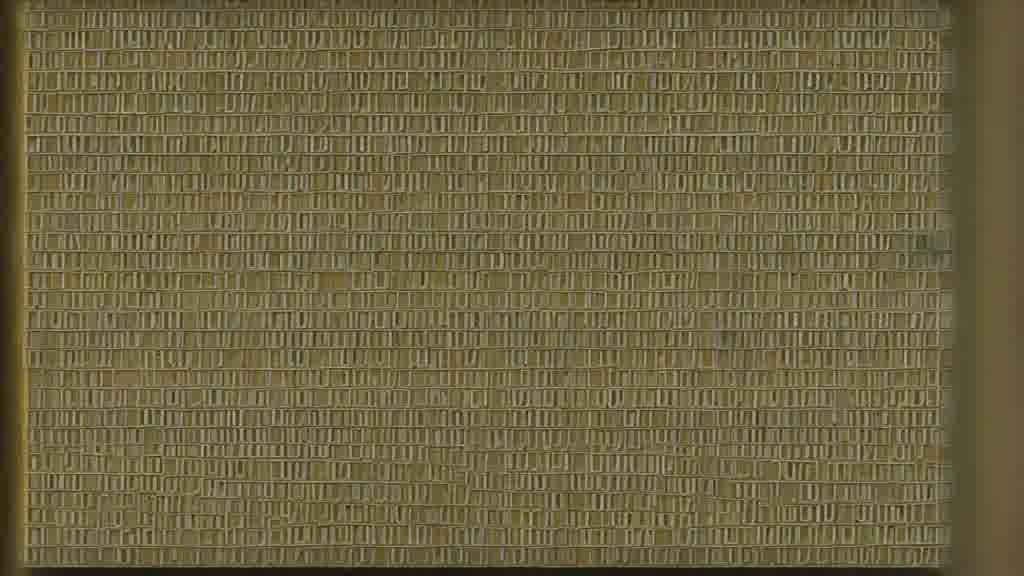}} \\
	\end{tblr}
	\caption{Comparison to Stable Video Diffusion~\cite{svd} baseline. We compare our method to Stable Video Diffusion (SVD) for multiple motions and seeds. While SVD often fails to align with the motion reference and is highly influenced by the seed, our motion-text embedding guides the model to generate videos with matching motion, minimizing variations caused by the seed.}
	\Description{Grid showing three different motions for SVD and our method: a human doing jumping jacks, a horse walking, and a person yawning. For each motion, the first row shows the motion reference video, the next three rows show the generated videos with the image (latent) input, and the last three rows show the generated videos without the image (latent) input, i.e., our visualizations.}
	\label{fig:comp_svd}
\end{figure*}

Fig.~\ref{fig:comp_svd} compares our method with the Stable Video Diffusion (SVD) \cite{svd} baseline for multiple motions and seeds. It further visualizes our motion-text embeddings and SVD's image embeddings with unconditional appearances. While this is not a fair comparison---since SVD does not incorporate the motion reference video---the goal is to analyze and better understand the capabilities of both methods.

As expected, SVD's generated results generally do not follow the reference motions. In rare cases, the motion does match somewhat, likely because the expected motion of the target image is similar to reference motion, as seen in the horse/dog example. However, close inspection reveals that the gaits of the generated videos do differ and that the dog's tail wiggles in the third example. Our method's motion-text embeddings seem to capture the motion of the reference videos well, i.e., replacing the image embedding of the target image with the motion-text embedding leads to successful motion transfers for all three seeds. In our method, different seeds produce varying artifacts (e.g., arms for the jumping jacks example) while maintaining largely consistent motions. For the horse/dog example, our method generates videos where the motion closely follows the horse's gait, as explored further in Fig.~\ref{fig:results_style}. 

Generating results with an unconditional appearance, i.e., where the image (latent) input is zeroed out, provides insight into the information encoded in the embeddings. However, note that the visualization is not always easily interpretable, depending on the motion, the optimization iteration, and the seed. SVD uses the CLIP~\cite{clip} image embedding of the target image, resulting in videos that depict characters semantically similar to those in the target image. The motions vary with the seed and do not consistently align with those in videos generated with the image (latent) condition. In contrast, our method uses the motion-text embeddings optimized on the motion reference video. While the exact appearance (e.g., colors) varies with the seed, the object types seem to resemble those of the motion reference video. This may stem from initializing the motion-text embedding with image embeddings extracted from the motion reference video. The encoding of object types in the motion-text embedding may also explain the occasional structure leakage noted in the limitations section.

Results generated with SVD frequently exhibit significant artifacts (e.g., first two seeds for the jumping jacks example) and appearance changes (e.g., last two seeds for the yawning example). As our method builds on SVD's frozen weights, we inherit some of SVD's issues, as described in the limitations section. However, by conditioning the model on a reference motion, our results tend to appear more realistic and contain fewer artifacts. We hypothesize that this improvement arises because the model leverages the provided (realistic) motion rather than needing to hallucinate it from scratch, simplifying the overall task. Additionally, SVD often generates static objects with moving cameras in our experience. We suggest that motion transfer methods, like ours, can help generate more natural and diverse motions.

\subsection{Additional Qualitative Comparisons to State-of-the-Art Methods} \label{sec:additional-qual-comp-sota}

\begin{figure*}[htbp]
	\centering
	\begin{tblr}{
			cell{1}{2} = {c=4}{c},
			cell{1}{6} = {c=4}{c},
			columns = {c},
			column{3} = {rightsep=1pt},
			column{4} = {leftsep=1pt,rightsep=1pt},
			column{5} = {leftsep=1pt},
			column{7} = {rightsep=1pt},
			column{8} = {leftsep=1pt,rightsep=1pt},
			column{9} = {leftsep=1pt},
			vline{3} = {3-7}{dashed},
			vline{6} = {1-7}{},
			vline{7} = {3-7}{dashed},
			hline{3} = {1-10}{},
			hline{4} = {1-10}{},
			hline{5} = {1-10}{},
			hline{6} = {1-10}{},
			hline{7} = {1-10}{},
		}
		& Walking unsteadily & & & & Waddling towards the camera & & & \\
		{Ref.} & \raisebox{-.5\height}{\includegraphics[width=0.09\textwidth]{figures/results/fullbody_drunk/frames_input/001.jpg}} & \raisebox{-.5\height}{\includegraphics[width=0.09\textwidth]{figures/results/fullbody_drunk/frames_input/004.jpg}} & \raisebox{-.5\height}{\includegraphics[width=0.09\textwidth]{figures/results/fullbody_drunk/frames_input/008.jpg}} & \raisebox{-.5\height}{\includegraphics[width=0.09\textwidth]{figures/results/fullbody_drunk/frames_input/012.jpg}} & \raisebox{-.5\height}{\includegraphics[width=0.09\textwidth]{figures/results/fullbody_animal/frames_input/001.jpg}} & \raisebox{-.5\height}{\includegraphics[width=0.09\textwidth]{figures/results/fullbody_animal/frames_input/004.jpg}} & \raisebox{-.5\height}{\includegraphics[width=0.09\textwidth]{figures/results/fullbody_animal/frames_input/008.jpg}} & \raisebox{-.5\height}{\includegraphics[width=0.09\textwidth]{figures/results/fullbody_animal/frames_input/012.jpg}} \\
		{SVD}  & \raisebox{-.5\height}{\includegraphics[width=0.09\textwidth]{figures/results/fullbody_drunk/first_frame.jpg}} & \raisebox{-.5\height}{\includegraphics[width=0.09\textwidth]{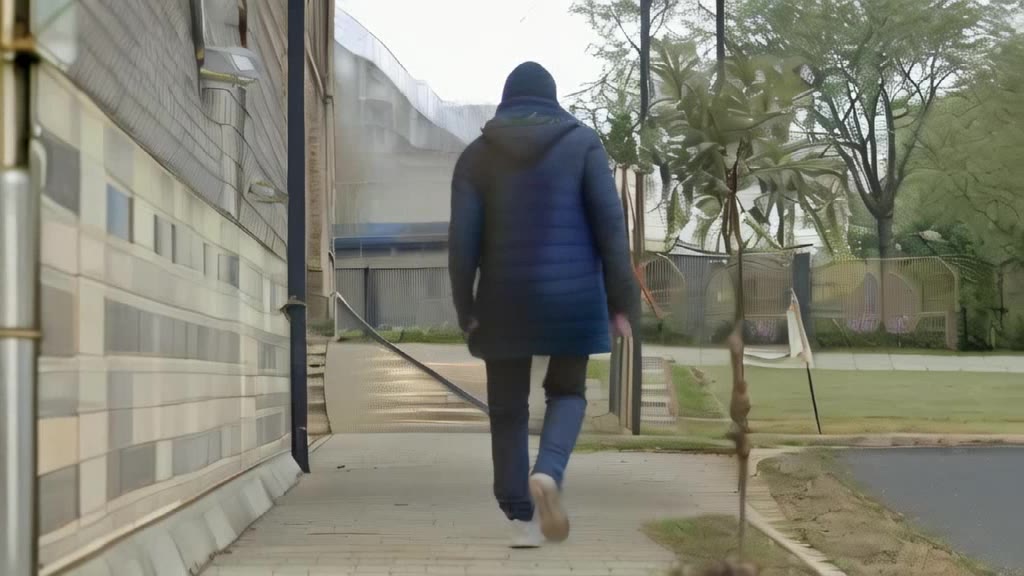}} & \raisebox{-.5\height}{\includegraphics[width=0.09\textwidth]{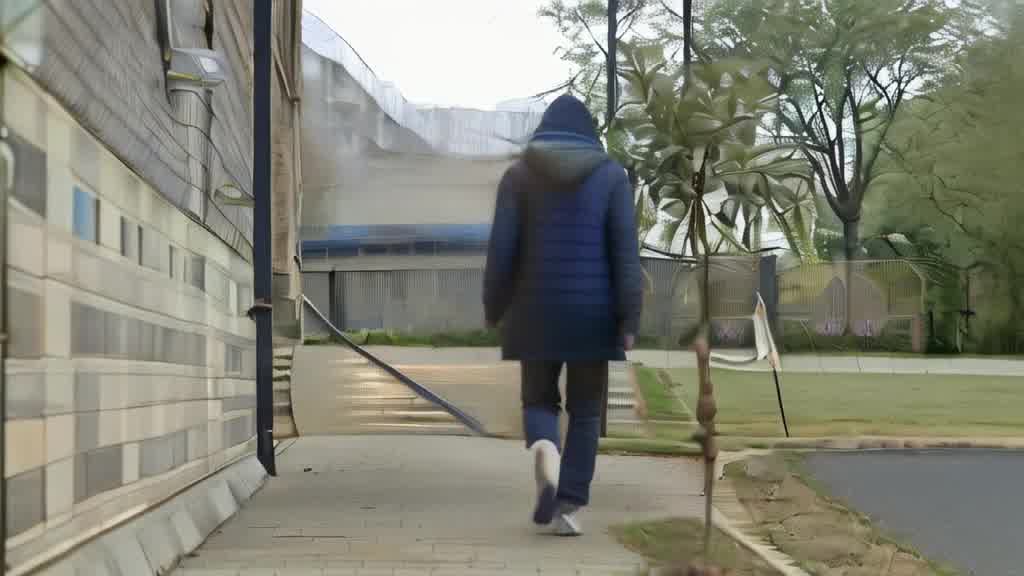}} & \raisebox{-.5\height}{\includegraphics[width=0.09\textwidth]{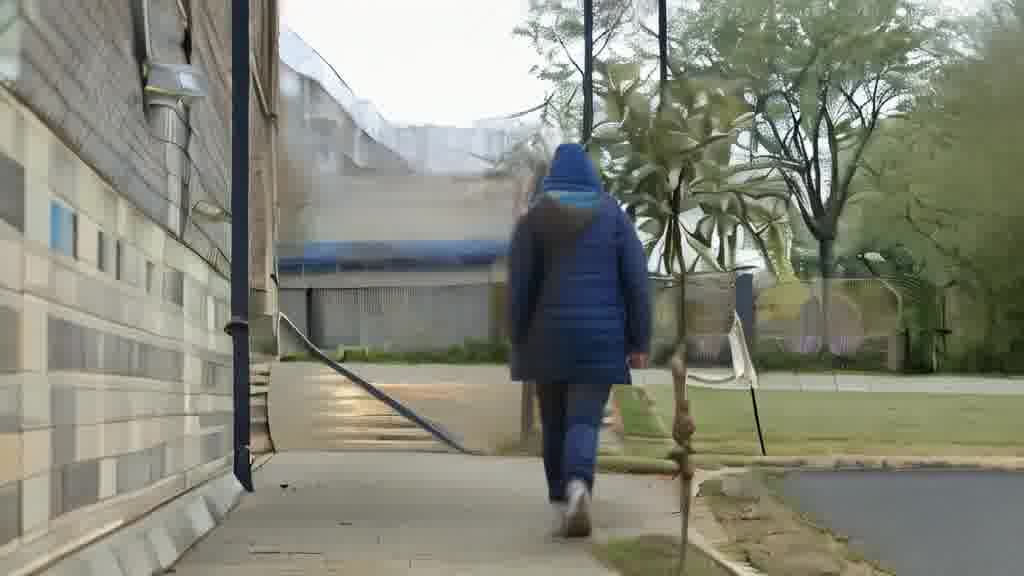}} & \raisebox{-.5\height}{\includegraphics[width=0.09\textwidth]{figures/results/fullbody_animal/first_frame.jpg}} & \raisebox{-.5\height}{\includegraphics[width=0.09\textwidth]{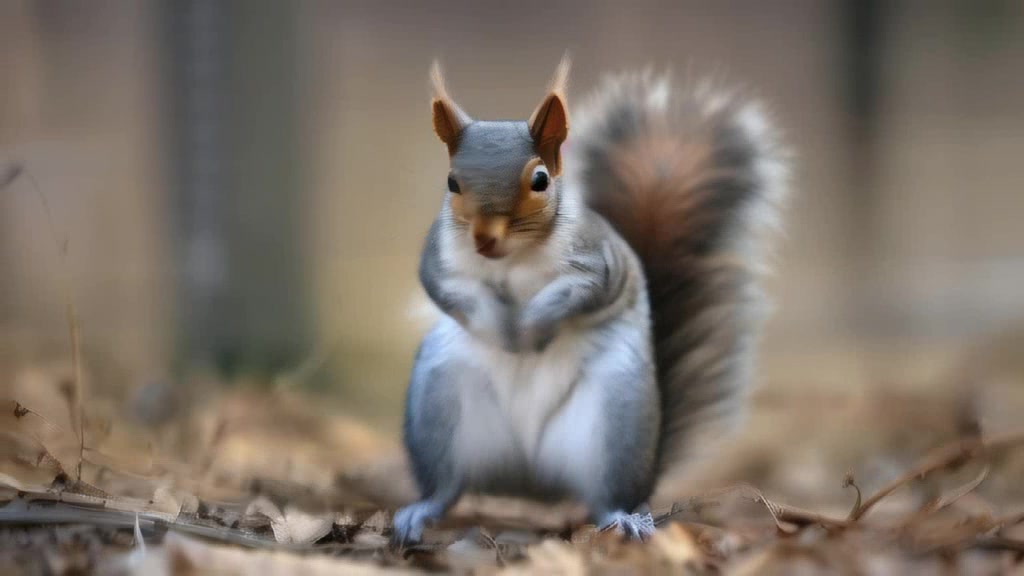}} & \raisebox{-.5\height}{\includegraphics[width=0.09\textwidth]{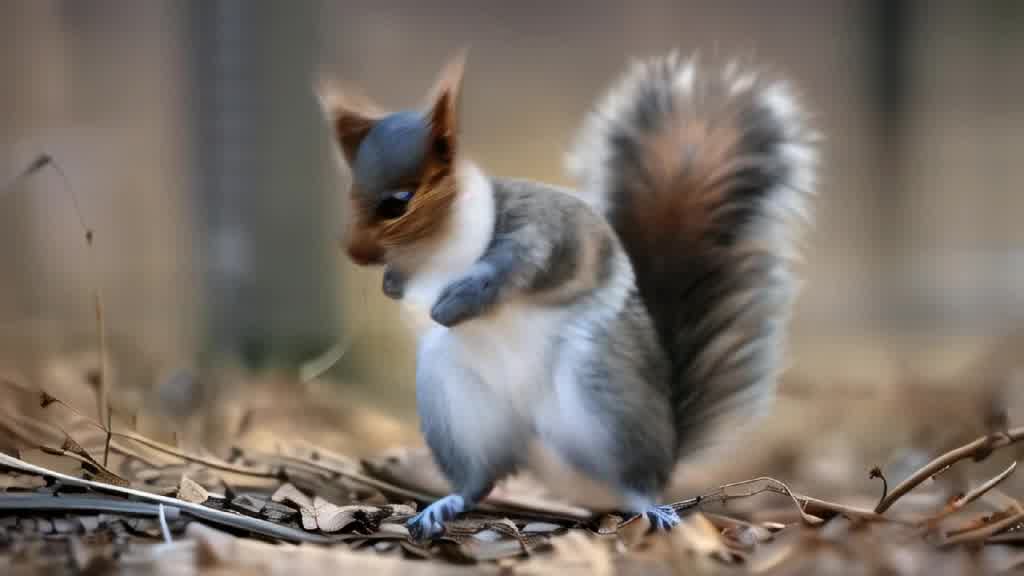}} & \raisebox{-.5\height}{\includegraphics[width=0.09\textwidth]{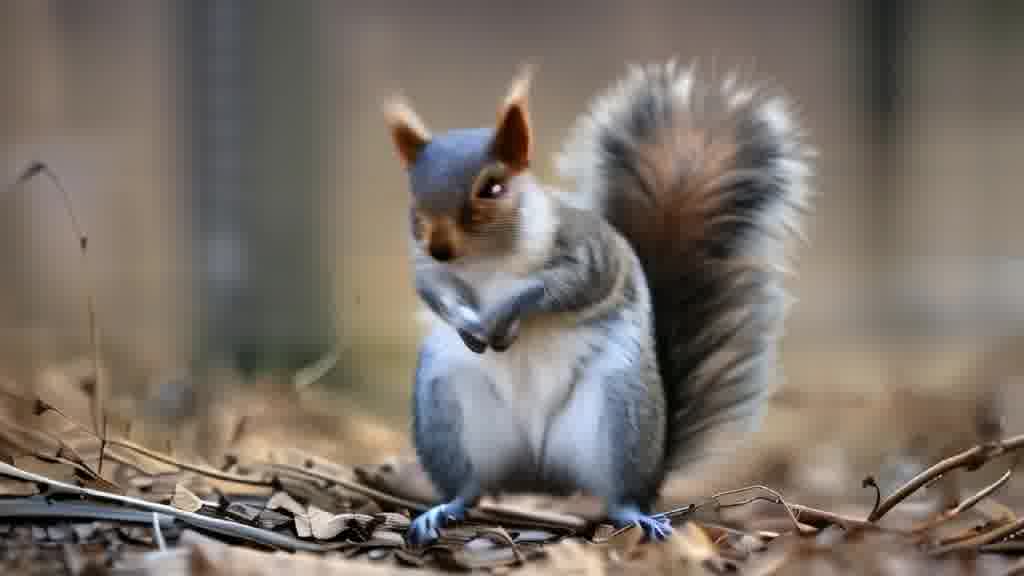}} \\
		{VC} & \raisebox{-.5\height}{\includegraphics[width=0.09\textwidth]{figures/results/fullbody_drunk/first_frame.jpg}} & \raisebox{-.5\height}{\includegraphics[width=0.050625\textwidth]{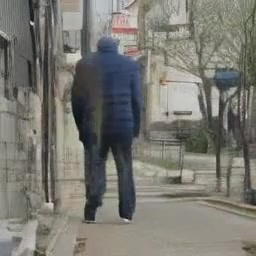}} & \raisebox{-.5\height}{\includegraphics[width=0.050625\textwidth]{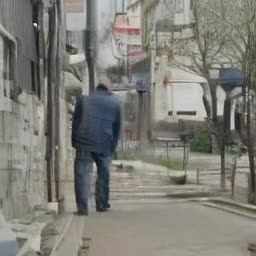}} & \raisebox{-.5\height}{\includegraphics[width=0.050625\textwidth]{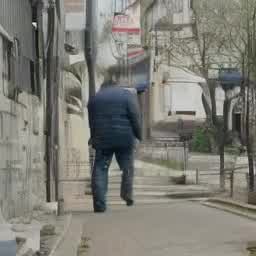}} & \raisebox{-.5\height}{\includegraphics[width=0.09\textwidth]{figures/results/fullbody_animal/first_frame.jpg}} & \raisebox{-.5\height}{\includegraphics[width=0.050625\textwidth]{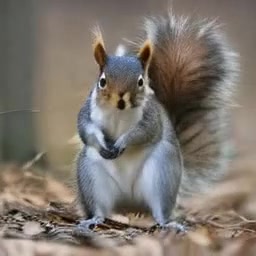}} & \raisebox{-.5\height}{\includegraphics[width=0.050625\textwidth]{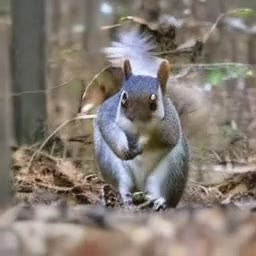}} & \raisebox{-.5\height}{\includegraphics[width=0.050625\textwidth]{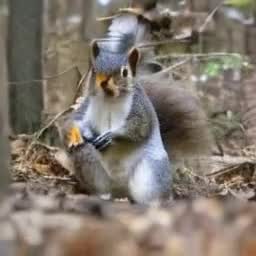}} \\
		{MC} & \raisebox{-.5\height}{\includegraphics[width=0.09\textwidth]{figures/results/fullbody_drunk/first_frame.jpg}} & \raisebox{-.5\height}{\includegraphics[width=0.050625\textwidth]{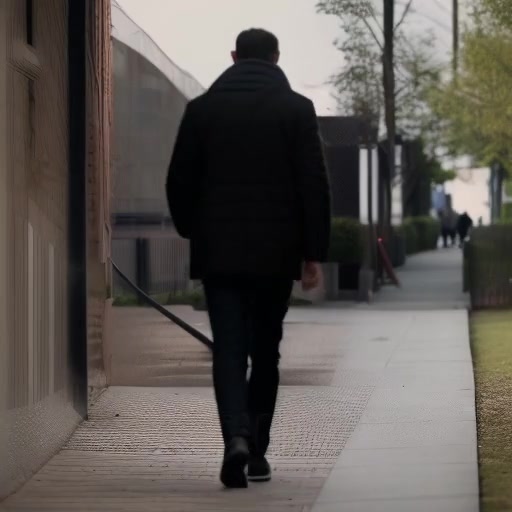}} & \raisebox{-.5\height}{\includegraphics[width=0.050625\textwidth]{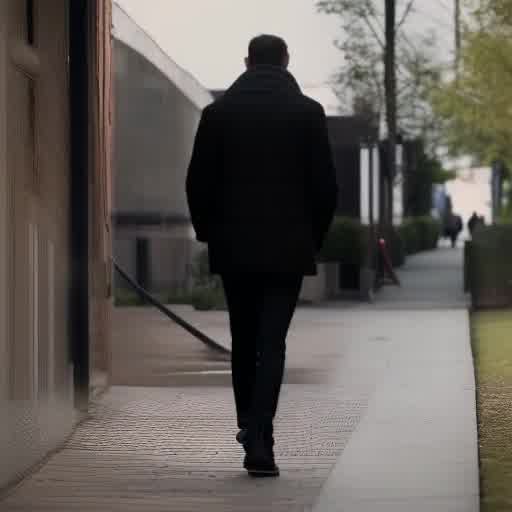}} & \raisebox{-.5\height}{\includegraphics[width=0.050625\textwidth]{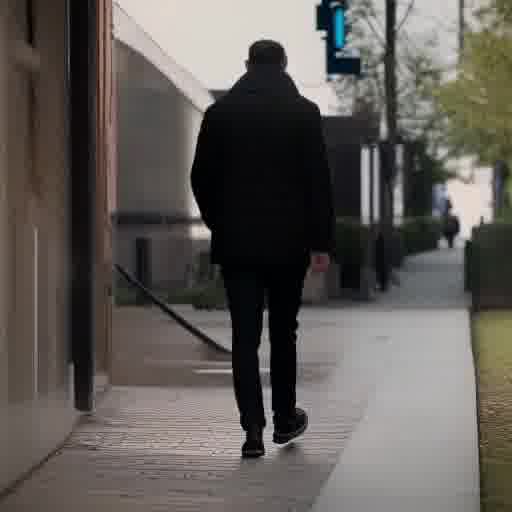}} & \raisebox{-.5\height}{\includegraphics[width=0.09\textwidth]{figures/results/fullbody_animal/first_frame.jpg}} & \raisebox{-.5\height}{\includegraphics[width=0.050625\textwidth]{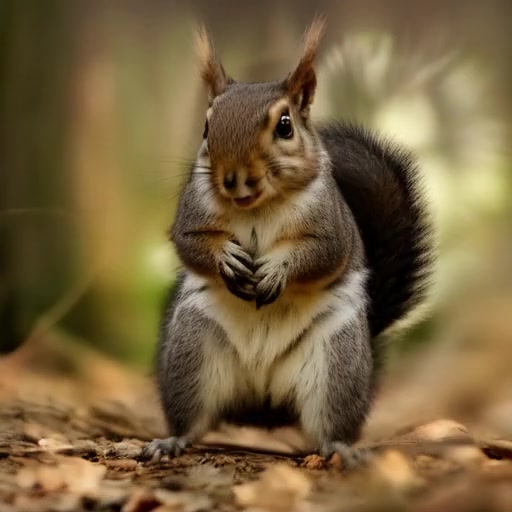}} & \raisebox{-.5\height}{\includegraphics[width=0.050625\textwidth]{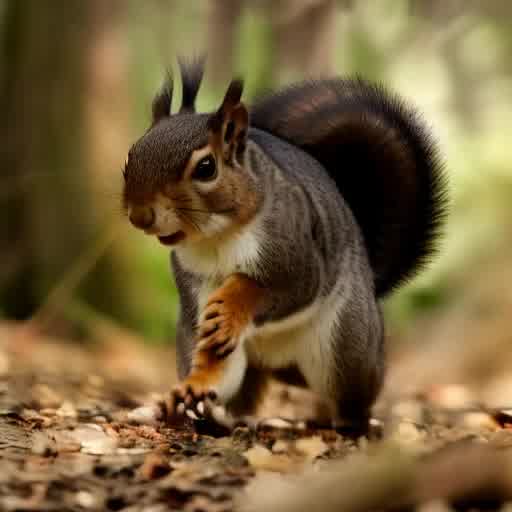}} & \raisebox{-.5\height}{\includegraphics[width=0.050625\textwidth]{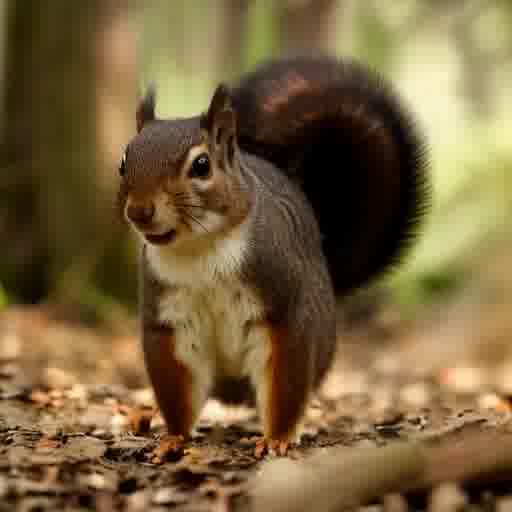}} \\
		{MD} & \raisebox{-.5\height}{\includegraphics[width=0.09\textwidth]{figures/results/fullbody_drunk/first_frame.jpg}} & \raisebox{-.5\height}{\includegraphics[width=0.050625\textwidth]{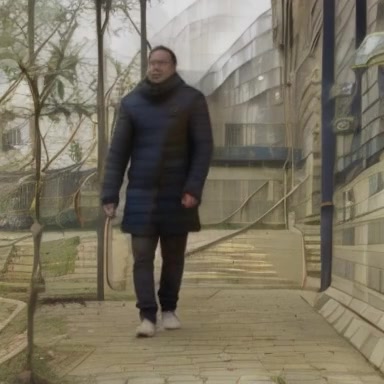}} & \raisebox{-.5\height}{\includegraphics[width=0.050625\textwidth]{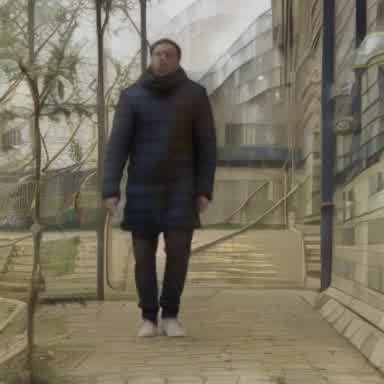}} & \raisebox{-.5\height}{\includegraphics[width=0.050625\textwidth]{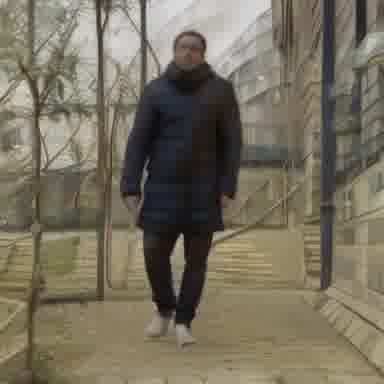}} & \raisebox{-.5\height}{\includegraphics[width=0.09\textwidth]{figures/results/fullbody_animal/first_frame.jpg}} & \raisebox{-.5\height}{\includegraphics[width=0.050625\textwidth]{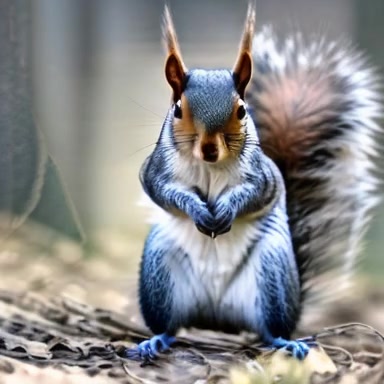}} & \raisebox{-.5\height}{\includegraphics[width=0.050625\textwidth]{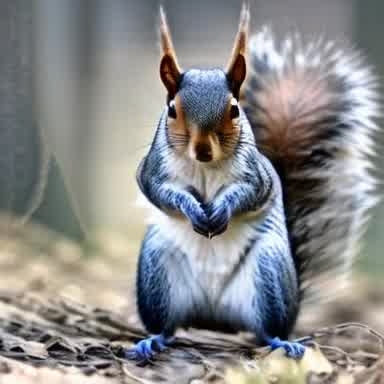}} & \raisebox{-.5\height}{\includegraphics[width=0.050625\textwidth]{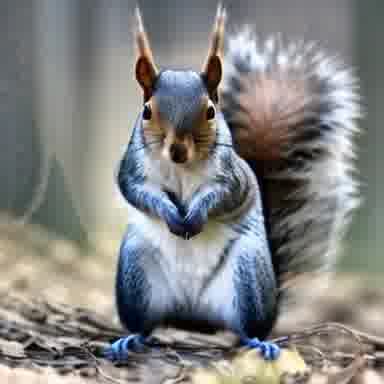}} \\
		{\textbf{Ours}} & \raisebox{-.5\height}{\includegraphics[width=0.09\textwidth]{figures/results/fullbody_drunk/first_frame.jpg}} & \raisebox{-.5\height}{\includegraphics[width=0.09\textwidth]{figures/results/fullbody_drunk/frames/004.jpg}} & \raisebox{-.5\height}{\includegraphics[width=0.09\textwidth]{figures/results/fullbody_drunk/frames/008.jpg}} & \raisebox{-.5\height}{\includegraphics[width=0.09\textwidth]{figures/results/fullbody_drunk/frames/012.jpg}} & \raisebox{-.5\height}{\includegraphics[width=0.09\textwidth]{figures/results/fullbody_animal/first_frame.jpg}} & \raisebox{-.5\height}{\includegraphics[width=0.09\textwidth]{figures/results/fullbody_animal/frames/004.jpg}} &  \raisebox{-.5\height}{\includegraphics[width=0.09\textwidth]{figures/results/fullbody_animal/frames/008.jpg}} &  \raisebox{-.5\height}{\includegraphics[width=0.09\textwidth]{figures/results/fullbody_animal/frames/012.jpg}}
	\end{tblr}
	\begin{tblr}{
			cell{1}{2} = {c=4}{c},
			cell{1}{6} = {c=4}{c},
			columns = {c},
			column{3} = {rightsep=1pt},
			column{4} = {leftsep=1pt,rightsep=1pt},
			column{5} = {leftsep=1pt},
			column{7} = {rightsep=1pt},
			column{8} = {leftsep=1pt,rightsep=1pt},
			column{9} = {leftsep=1pt},
			vline{3} = {3-7}{dashed},
			vline{6} = {1-7}{},
			vline{7} = {3-7}{dashed},
			hline{3} = {1-10}{},
			hline{4} = {1-10}{},
			hline{5} = {1-10}{},
			hline{6} = {1-10}{},
			hline{7} = {1-10}{},
		}
		& Opening mouth wide & & & & Tilting head back & & & \\
		{Ref.} & \raisebox{-.5\height}{\includegraphics[width=0.09\textwidth]{figures/results/face_opening_mouth/frames_input/001.jpg}} & \raisebox{-.5\height}{\includegraphics[width=0.09\textwidth]{figures/results/face_opening_mouth/frames_input/004.jpg}} & \raisebox{-.5\height}{\includegraphics[width=0.09\textwidth]{figures/results/face_opening_mouth/frames_input/008.jpg}} & \raisebox{-.5\height}{\includegraphics[width=0.09\textwidth]{figures/results/face_opening_mouth/frames_input/012.jpg}} & \raisebox{-.5\height}{\includegraphics[width=0.09\textwidth]{figures/results/face_tilting_up/frames_input/001.jpg}} & \raisebox{-.5\height}{\includegraphics[width=0.09\textwidth]{figures/results/face_tilting_up/frames_input/003.jpg}} & \raisebox{-.5\height}{\includegraphics[width=0.09\textwidth]{figures/results/face_tilting_up/frames_input/006.jpg}} & \raisebox{-.5\height}{\includegraphics[width=0.09\textwidth]{figures/results/face_tilting_up/frames_input/009.jpg}} \\
		{SVD}  & \raisebox{-.5\height}{\includegraphics[width=0.09\textwidth]{figures/results/face_opening_mouth/first_frame.jpg}} & \raisebox{-.5\height}{\includegraphics[width=0.09\textwidth]{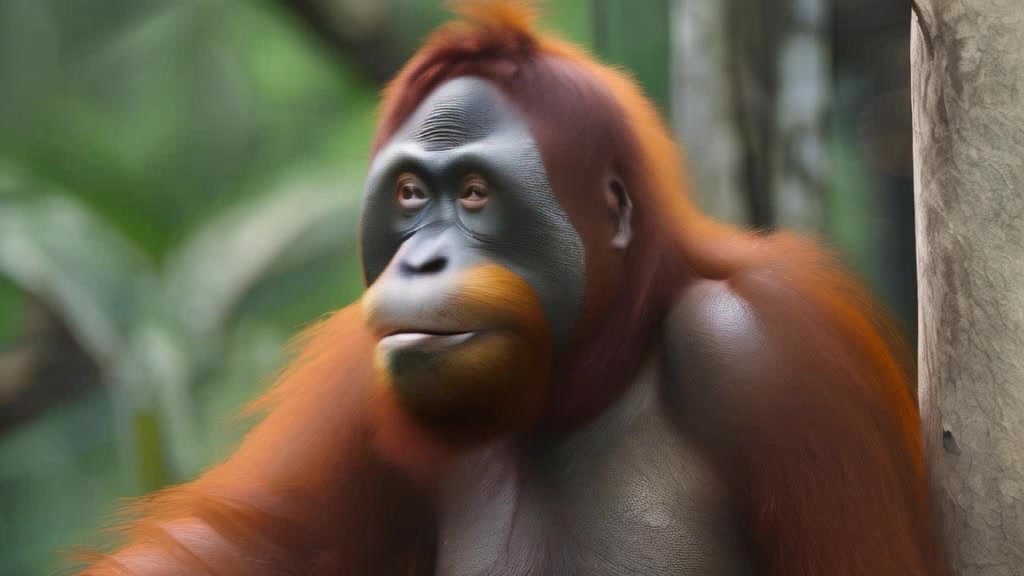}} & \raisebox{-.5\height}{\includegraphics[width=0.09\textwidth]{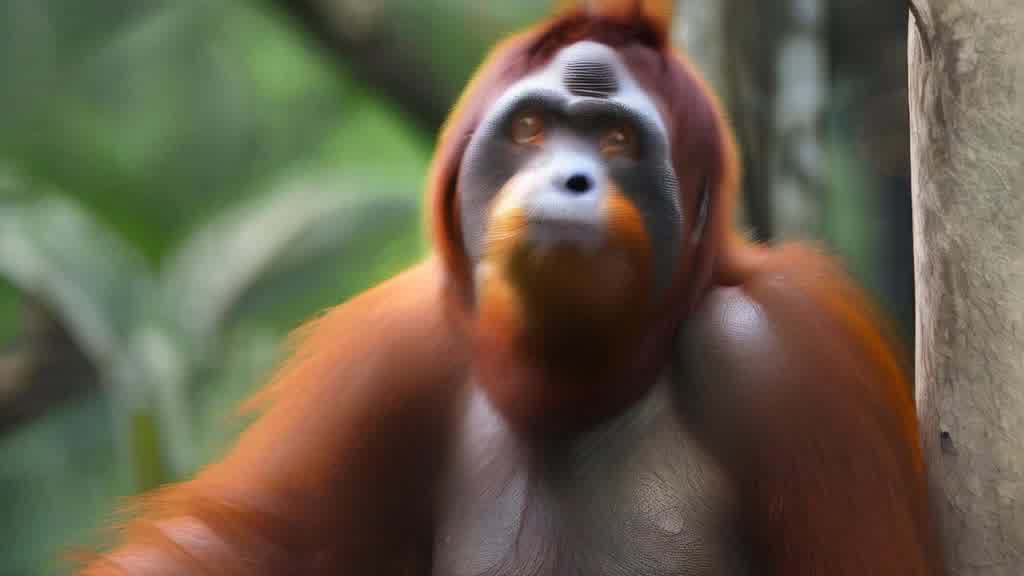}} & \raisebox{-.5\height}{\includegraphics[width=0.09\textwidth]{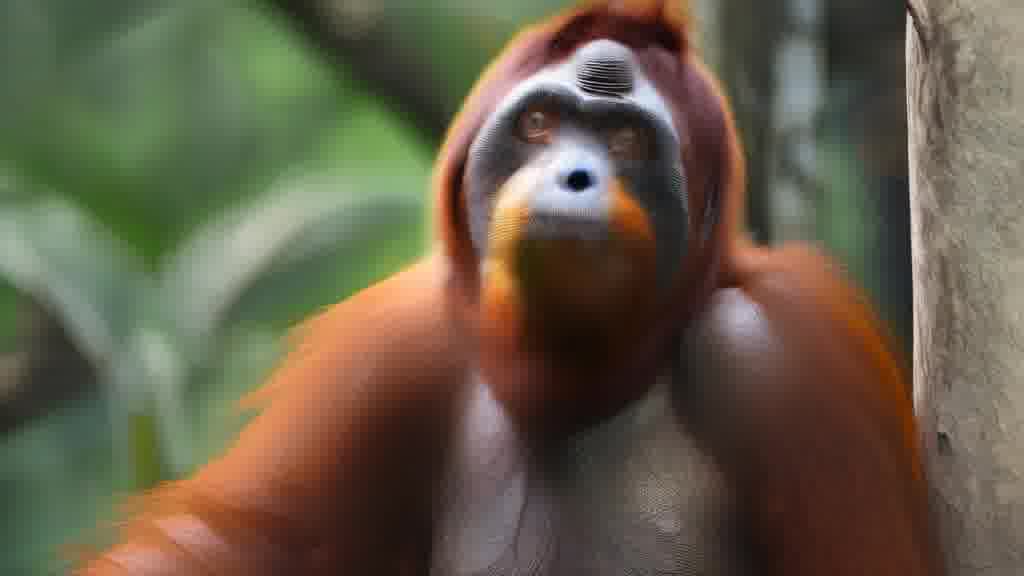}} & \raisebox{-.5\height}{\includegraphics[width=0.09\textwidth]{figures/results/face_tilting_up/first_frame.jpg}} & \raisebox{-.5\height}{\includegraphics[width=0.09\textwidth]{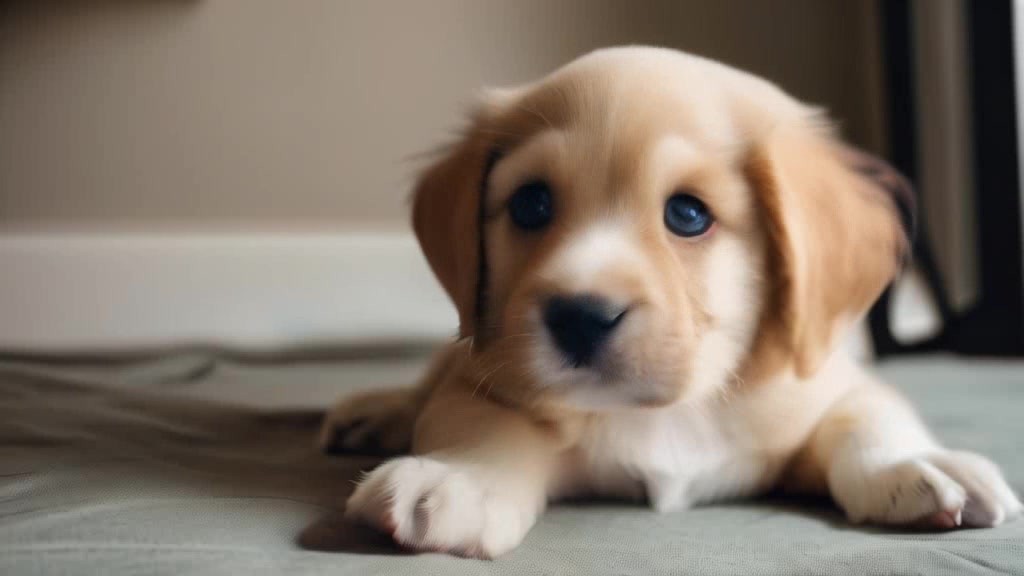}} & \raisebox{-.5\height}{\includegraphics[width=0.09\textwidth]{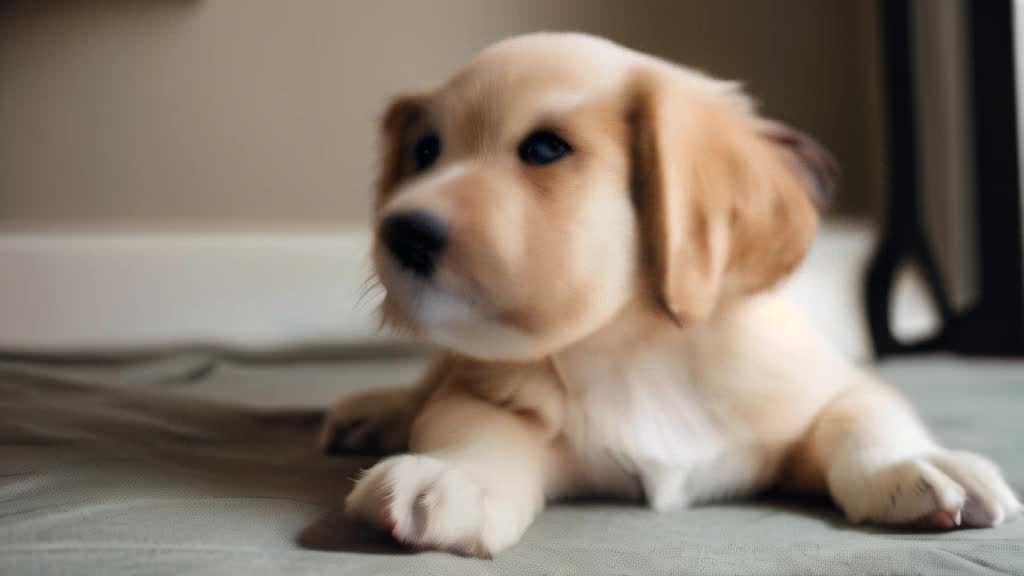}} & \raisebox{-.5\height}{\includegraphics[width=0.09\textwidth]{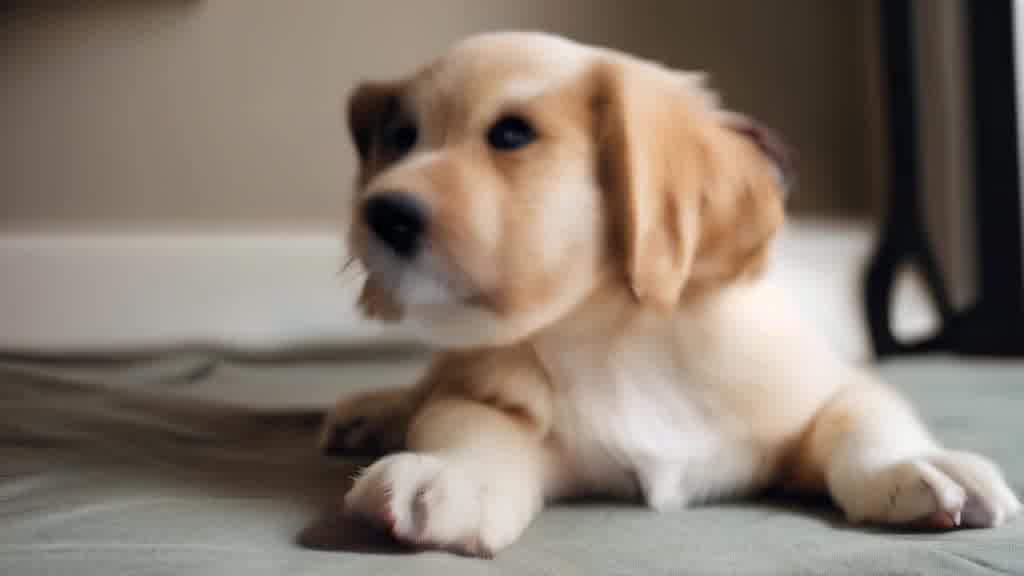}} \\
		{VC} & \raisebox{-.5\height}{\includegraphics[width=0.09\textwidth]{figures/results/face_opening_mouth/first_frame.jpg}} & \raisebox{-.5\height}{\includegraphics[width=0.050625\textwidth]{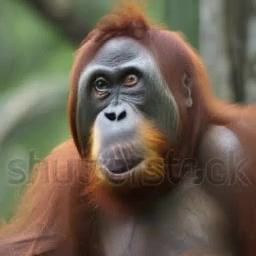}} & \raisebox{-.5\height}{\includegraphics[width=0.050625\textwidth]{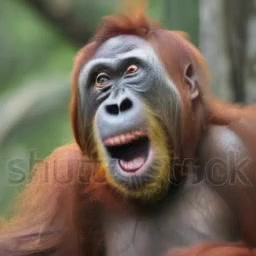}} & \raisebox{-.5\height}{\includegraphics[width=0.050625\textwidth]{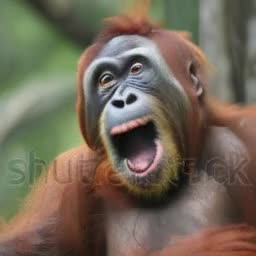}} & \raisebox{-.5\height}{\includegraphics[width=0.09\textwidth]{figures/results/face_tilting_up/first_frame.jpg}} & \raisebox{-.5\height}{\includegraphics[width=0.050625\textwidth]{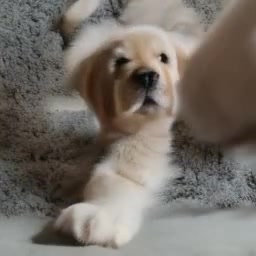}} & \raisebox{-.5\height}{\includegraphics[width=0.050625\textwidth]{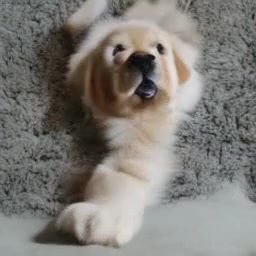}} & \raisebox{-.5\height}{\includegraphics[width=0.050625\textwidth]{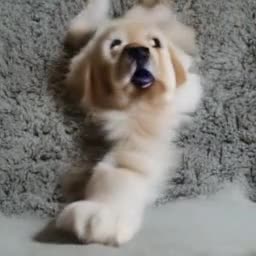}} \\
		{MC} & \raisebox{-.5\height}{\includegraphics[width=0.09\textwidth]{figures/results/face_opening_mouth/first_frame.jpg}} & \raisebox{-.5\height}{\includegraphics[width=0.050625\textwidth]{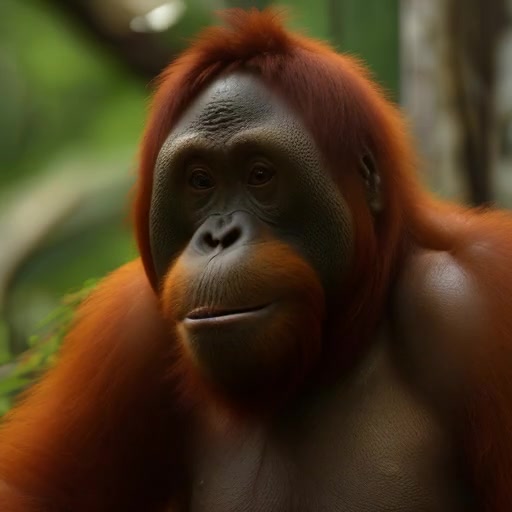}} & \raisebox{-.5\height}{\includegraphics[width=0.050625\textwidth]{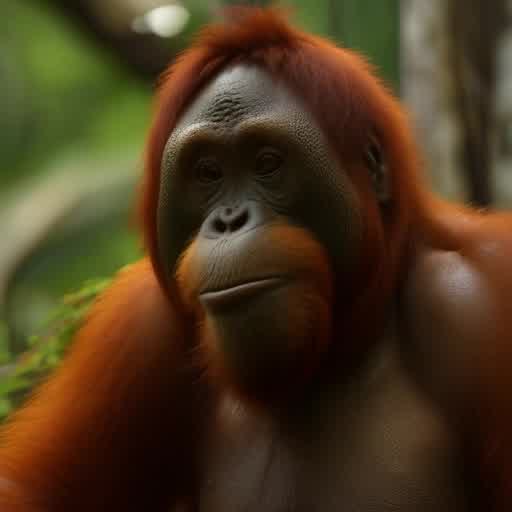}} & \raisebox{-.5\height}{\includegraphics[width=0.050625\textwidth]{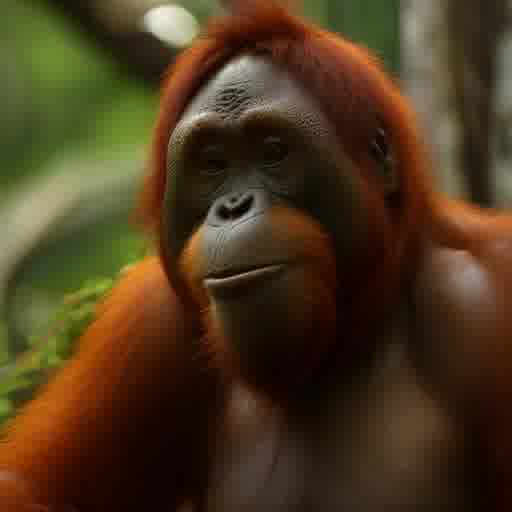}} & \raisebox{-.5\height}{\includegraphics[width=0.09\textwidth]{figures/results/face_tilting_up/first_frame.jpg}} & \raisebox{-.5\height}{\includegraphics[width=0.050625\textwidth]{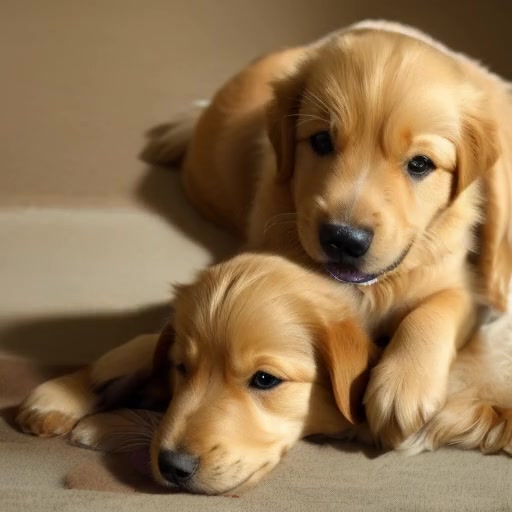}} & \raisebox{-.5\height}{\includegraphics[width=0.050625\textwidth]{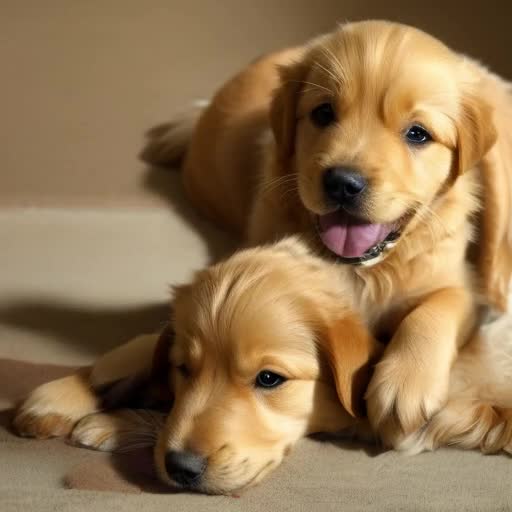}} & \raisebox{-.5\height}{\includegraphics[width=0.050625\textwidth]{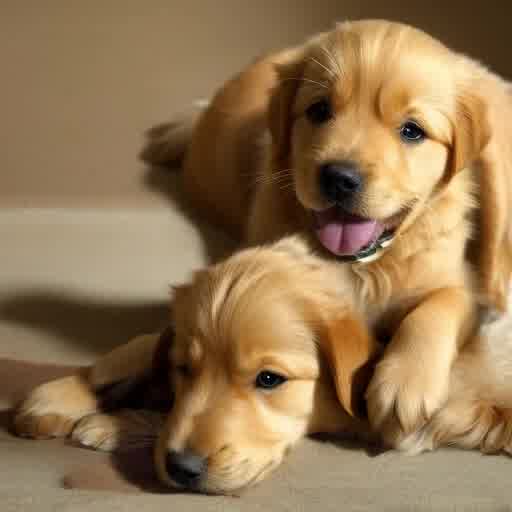}} \\
		{MD} & \raisebox{-.5\height}{\includegraphics[width=0.09\textwidth]{figures/results/face_opening_mouth/first_frame.jpg}} & \raisebox{-.5\height}{\includegraphics[width=0.050625\textwidth]{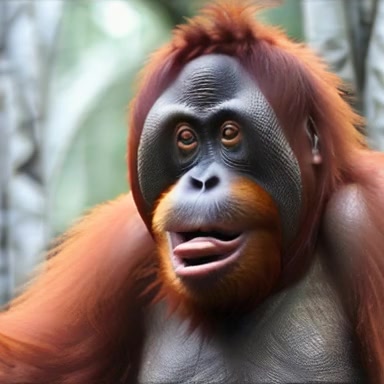}} & \raisebox{-.5\height}{\includegraphics[width=0.050625\textwidth]{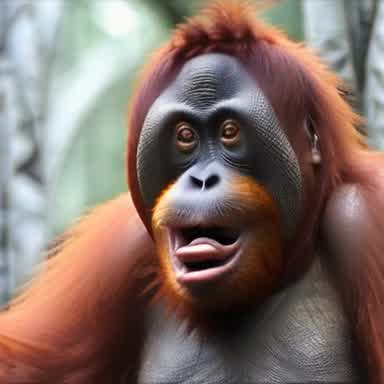}} & \raisebox{-.5\height}{\includegraphics[width=0.050625\textwidth]{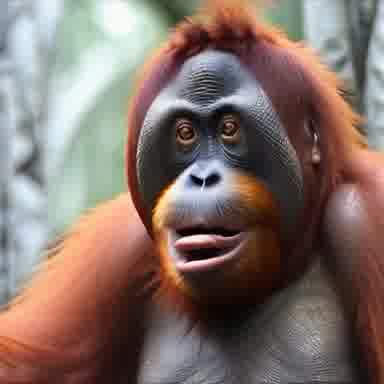}} & \raisebox{-.5\height}{\includegraphics[width=0.09\textwidth]{figures/results/face_tilting_up/first_frame.jpg}} & \raisebox{-.5\height}{\includegraphics[width=0.050625\textwidth]{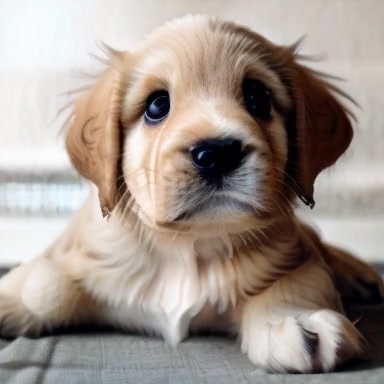}} & \raisebox{-.5\height}{\includegraphics[width=0.050625\textwidth]{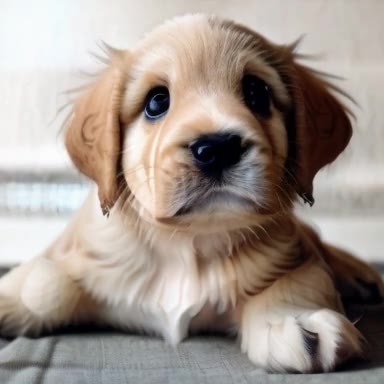}} & \raisebox{-.5\height}{\includegraphics[width=0.050625\textwidth]{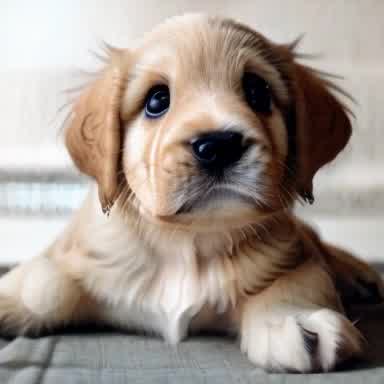}} \\
		{\textbf{Ours}} & \raisebox{-.5\height}{\includegraphics[width=0.09\textwidth]{figures/results/face_opening_mouth/first_frame.jpg}} & \raisebox{-.5\height}{\includegraphics[width=0.09\textwidth]{figures/results/face_opening_mouth/frames/004.jpg}} & \raisebox{-.5\height}{\includegraphics[width=0.09\textwidth]{figures/results/face_opening_mouth/frames/008.jpg}} & \raisebox{-.5\height}{\includegraphics[width=0.09\textwidth]{figures/results/face_opening_mouth/frames/012.jpg}} & \raisebox{-.5\height}{\includegraphics[width=0.09\textwidth]{figures/results/face_tilting_up/first_frame.jpg}} & \raisebox{-.5\height}{\includegraphics[width=0.09\textwidth]{figures/results/face_tilting_up/frames/003.jpg}} &  \raisebox{-.5\height}{\includegraphics[width=0.09\textwidth]{figures/results/face_tilting_up/frames/006.jpg}} &  \raisebox{-.5\height}{\includegraphics[width=0.09\textwidth]{figures/results/face_tilting_up/frames/009.jpg}}
	\end{tblr}
	\caption{Qualitative evaluation for additional examples (1/2). We compare our method to SVD = Stable Video Diffusion~\cite{svd} (baseline, no motion input), VC = VideoComposer~\cite{videocomposer}, MC = MotionClone~\cite{motionclone}, and MD = MotionDirector~\cite{motiondirector} for four different motions and target images. 
	}
	\Description{Grid with two blocks on top of each other. Each block has different methods in the rows and two different motion reference videos and starting frames in the columns. From top to bottom and left to right: unsteady gait from human to human, waddling from penguin to squirrel, opening mouth from human to orangutan, tilting head back from human to dog.}
	\label{fig:qual_eval_extra_1}
\end{figure*}

\begin{figure*}[htbp]
	\centering
	\begin{tblr}{
			cell{1}{2} = {c=4}{c},
			cell{1}{6} = {c=4}{c},
			columns = {c},
			column{3} = {rightsep=1pt},
			column{4} = {leftsep=1pt,rightsep=1pt},
			column{5} = {leftsep=1pt},
			column{7} = {rightsep=1pt},
			column{8} = {leftsep=1pt,rightsep=1pt},
			column{9} = {leftsep=1pt},
			vline{3} = {3-7}{dashed},
			vline{6} = {1-7}{},
			vline{7} = {3-7}{dashed},
			hline{3} = {1-10}{},
			hline{4} = {1-10}{},
			hline{5} = {1-10}{},
			hline{6} = {1-10}{},
			hline{7} = {1-10}{},
		}
		& Rotating bird's eye view camera & & & & Following vehicle with camera & & & \\
		{Ref.} & \raisebox{-.5\height}{\includegraphics[width=0.09\textwidth]{figures/results/camera_birdseye_rotate/frames_input/001.jpg}} & \raisebox{-.5\height}{\includegraphics[width=0.09\textwidth]{figures/results/camera_birdseye_rotate/frames_input/004.jpg}} & \raisebox{-.5\height}{\includegraphics[width=0.09\textwidth]{figures/results/camera_birdseye_rotate/frames_input/008.jpg}} & \raisebox{-.5\height}{\includegraphics[width=0.09\textwidth]{figures/results/camera_birdseye_rotate/frames_input/012.jpg}} & \raisebox{-.5\height}{\includegraphics[width=0.09\textwidth]{figures/results/camera_object_following/frames_input/001.jpg}} & \raisebox{-.5\height}{\includegraphics[width=0.09\textwidth]{figures/results/camera_object_following/frames_input/004.jpg}} & \raisebox{-.5\height}{\includegraphics[width=0.09\textwidth]{figures/results/camera_object_following/frames_input/008.jpg}} & \raisebox{-.5\height}{\includegraphics[width=0.09\textwidth]{figures/results/camera_object_following/frames_input/012.jpg}} \\
		{SVD}  & \raisebox{-.5\height}{\includegraphics[width=0.09\textwidth]{figures/results/camera_birdseye_rotate/first_frame.jpg}} & \raisebox{-.5\height}{\includegraphics[width=0.09\textwidth]{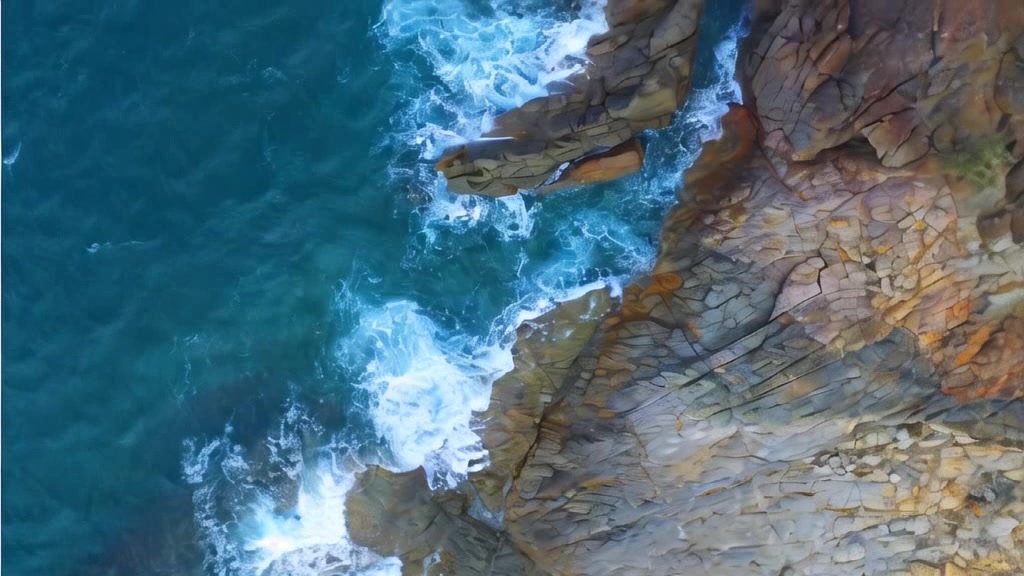}} & \raisebox{-.5\height}{\includegraphics[width=0.09\textwidth]{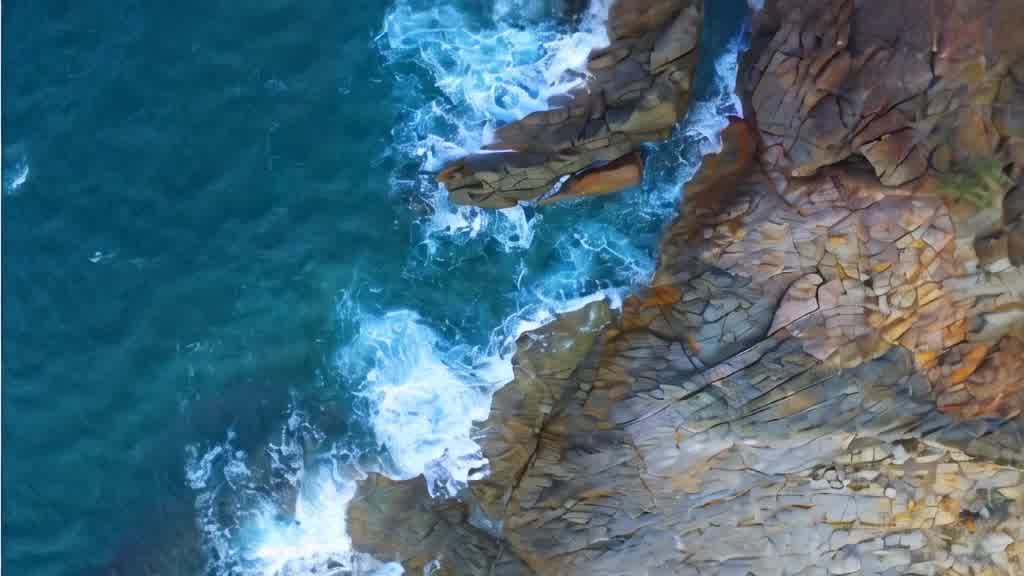}} & \raisebox{-.5\height}{\includegraphics[width=0.09\textwidth]{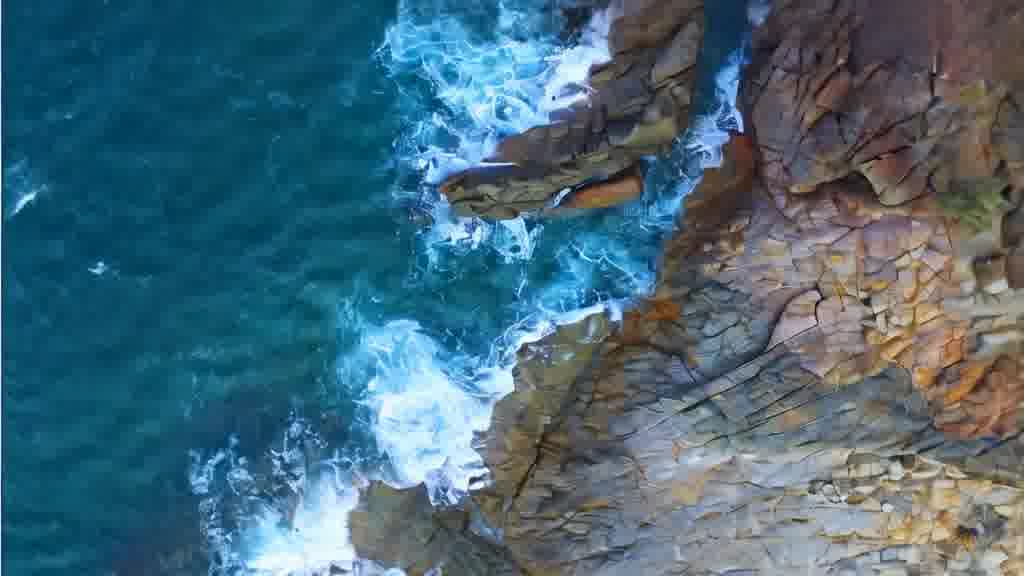}} & \raisebox{-.5\height}{\includegraphics[width=0.09\textwidth]{figures/results/camera_object_following/first_frame.jpg}} & \raisebox{-.5\height}{\includegraphics[width=0.09\textwidth]{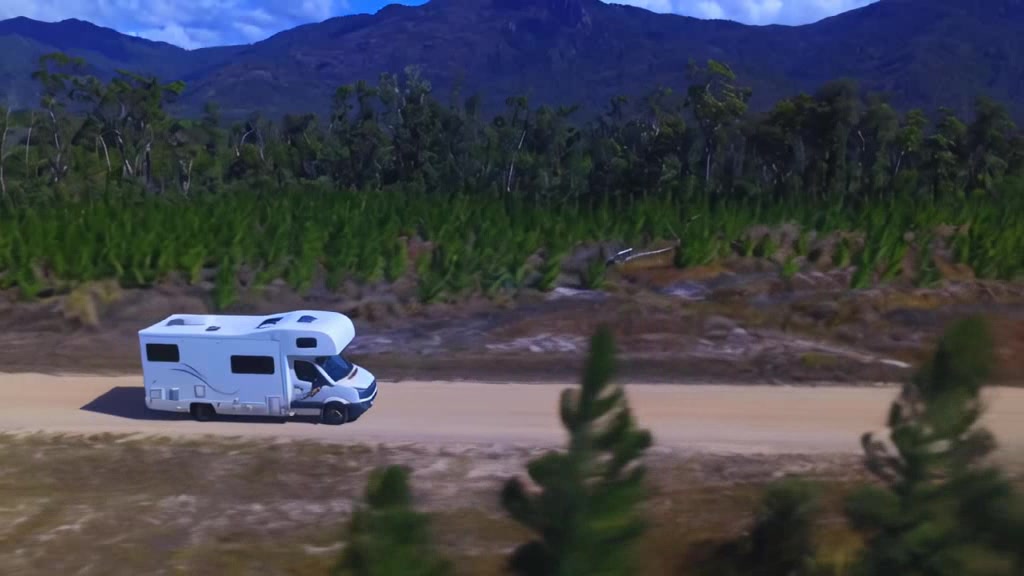}} & \raisebox{-.5\height}{\includegraphics[width=0.09\textwidth]{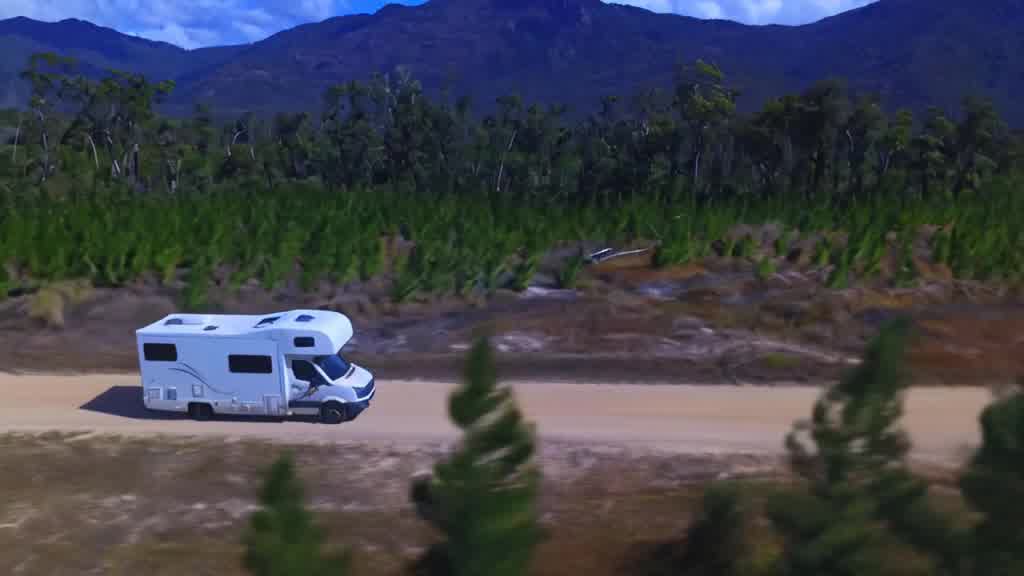}} & \raisebox{-.5\height}{\includegraphics[width=0.09\textwidth]{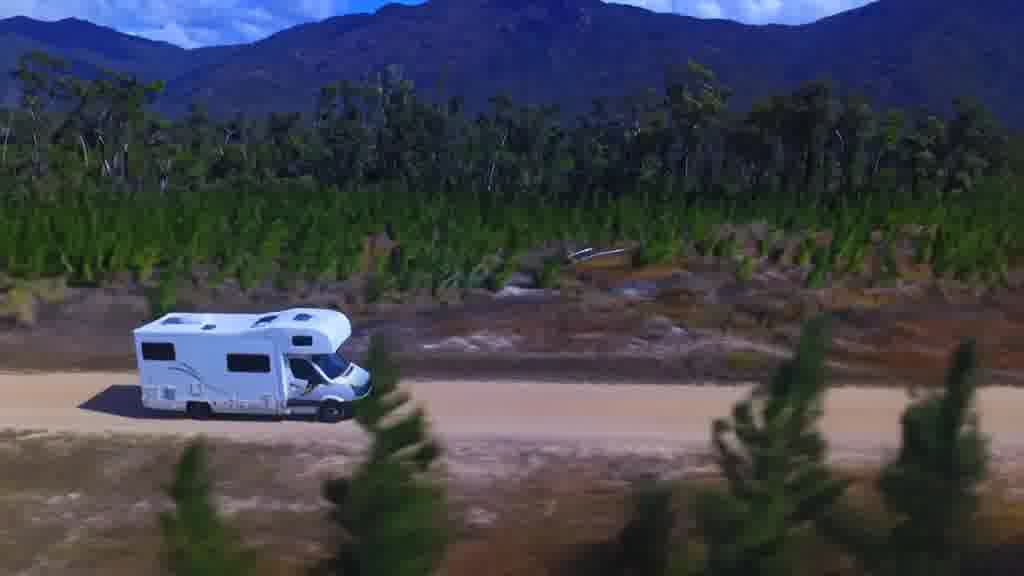}} \\
		{VC} & \raisebox{-.5\height}{\includegraphics[width=0.09\textwidth]{figures/results/camera_birdseye_rotate/first_frame.jpg}} & \raisebox{-.5\height}{\includegraphics[width=0.050625\textwidth]{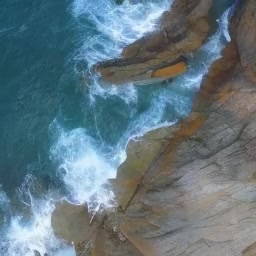}} & \raisebox{-.5\height}{\includegraphics[width=0.050625\textwidth]{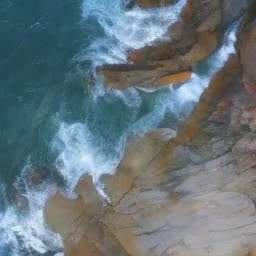}} & \raisebox{-.5\height}{\includegraphics[width=0.050625\textwidth]{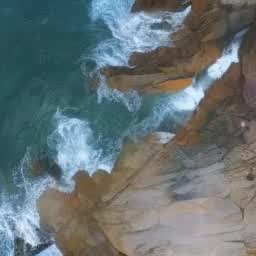}} & \raisebox{-.5\height}{\includegraphics[width=0.09\textwidth]{figures/results/camera_object_following/first_frame.jpg}} & \raisebox{-.5\height}{\includegraphics[width=0.050625\textwidth]{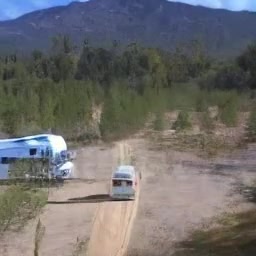}} & \raisebox{-.5\height}{\includegraphics[width=0.050625\textwidth]{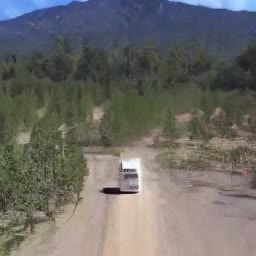}} & \raisebox{-.5\height}{\includegraphics[width=0.050625\textwidth]{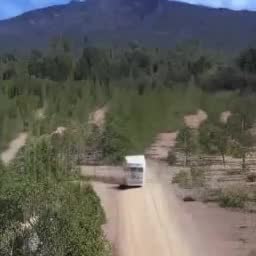}} \\
		{MC} & \raisebox{-.5\height}{\includegraphics[width=0.09\textwidth]{figures/results/camera_birdseye_rotate/first_frame.jpg}} & \raisebox{-.5\height}{\includegraphics[width=0.050625\textwidth]{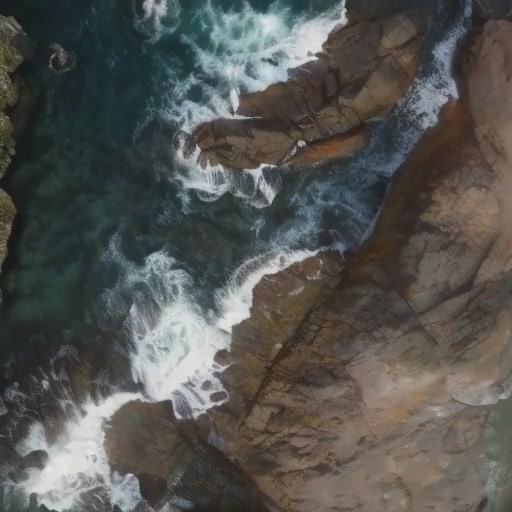}} & \raisebox{-.5\height}{\includegraphics[width=0.050625\textwidth]{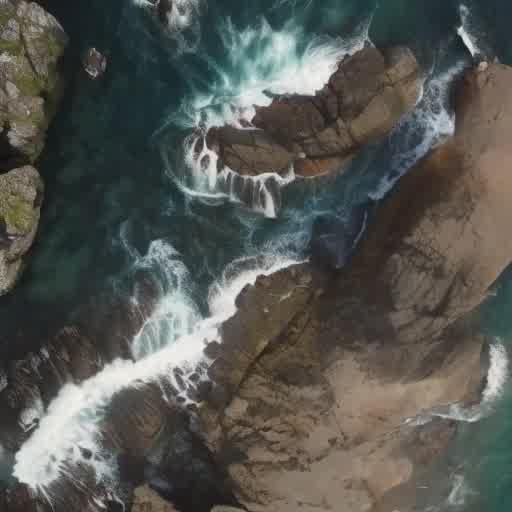}} & \raisebox{-.5\height}{\includegraphics[width=0.050625\textwidth]{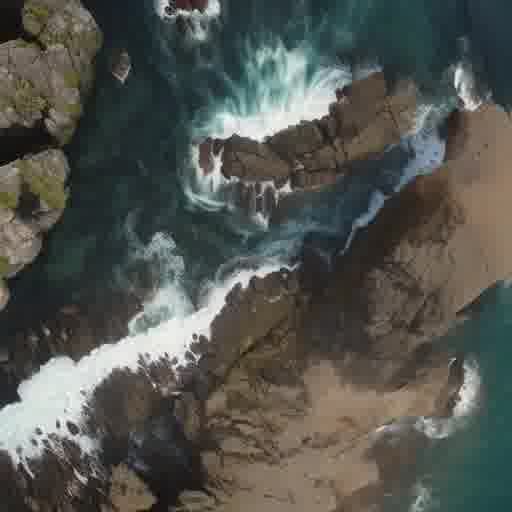}} & \raisebox{-.5\height}{\includegraphics[width=0.09\textwidth]{figures/results/camera_object_following/first_frame.jpg}} & \raisebox{-.5\height}{\includegraphics[width=0.050625\textwidth]{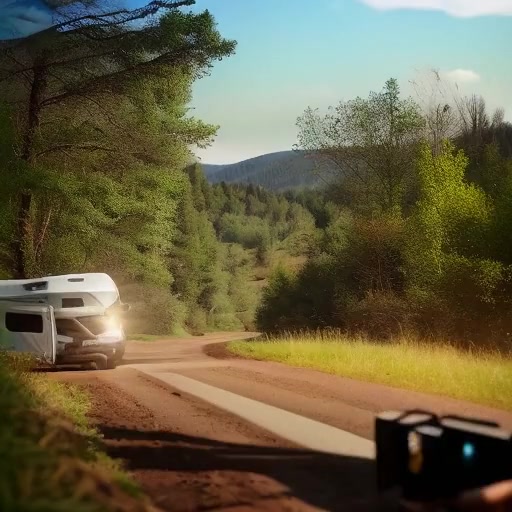}} & \raisebox{-.5\height}{\includegraphics[width=0.050625\textwidth]{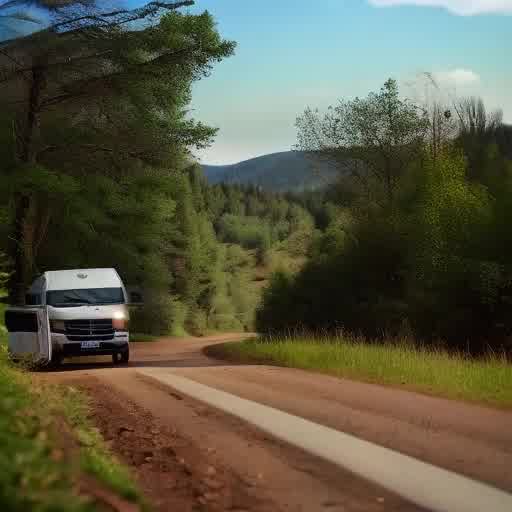}} & \raisebox{-.5\height}{\includegraphics[width=0.050625\textwidth]{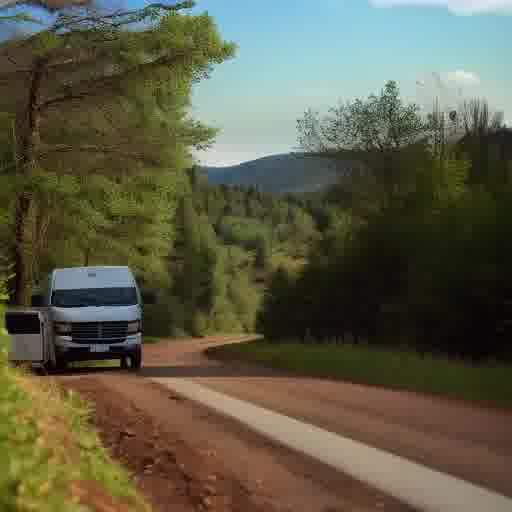}} \\
		{MD} & \raisebox{-.5\height}{\includegraphics[width=0.09\textwidth]{figures/results/camera_birdseye_rotate/first_frame.jpg}} & \raisebox{-.5\height}{\includegraphics[width=0.050625\textwidth]{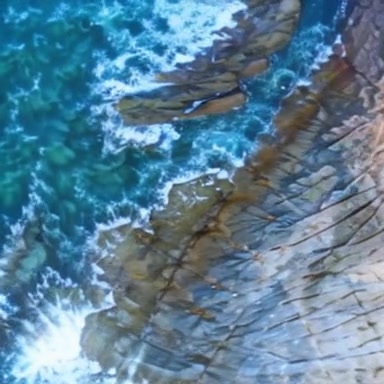}} & \raisebox{-.5\height}{\includegraphics[width=0.050625\textwidth]{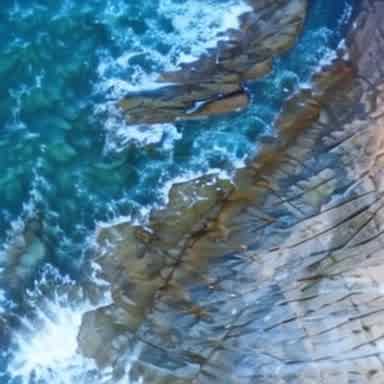}} & \raisebox{-.5\height}{\includegraphics[width=0.050625\textwidth]{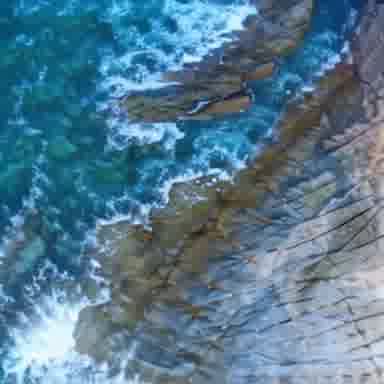}} & \raisebox{-.5\height}{\includegraphics[width=0.09\textwidth]{figures/results/camera_object_following/first_frame.jpg}} & \raisebox{-.5\height}{\includegraphics[width=0.050625\textwidth]{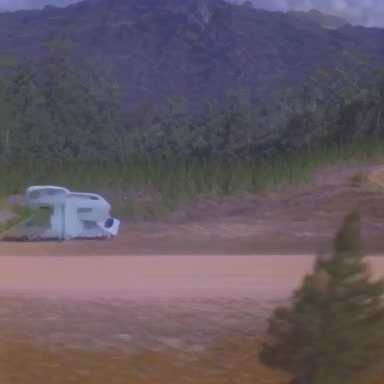}} & \raisebox{-.5\height}{\includegraphics[width=0.050625\textwidth]{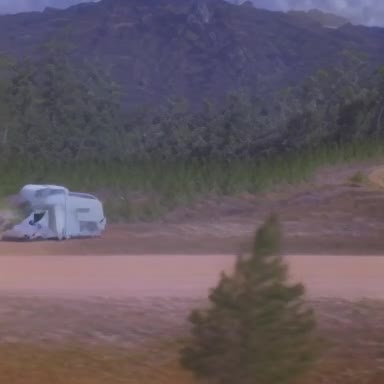}} & \raisebox{-.5\height}{\includegraphics[width=0.050625\textwidth]{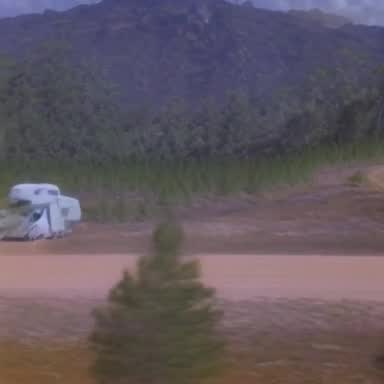}} \\
		{\textbf{Ours}} & \raisebox{-.5\height}{\includegraphics[width=0.09\textwidth]{figures/results/camera_birdseye_rotate/first_frame.jpg}} & \raisebox{-.5\height}{\includegraphics[width=0.09\textwidth]{figures/results/camera_birdseye_rotate/frames/004.jpg}} & \raisebox{-.5\height}{\includegraphics[width=0.09\textwidth]{figures/results/camera_birdseye_rotate/frames/008.jpg}} & \raisebox{-.5\height}{\includegraphics[width=0.09\textwidth]{figures/results/camera_birdseye_rotate/frames/012.jpg}} & \raisebox{-.5\height}{\includegraphics[width=0.09\textwidth]{figures/results/camera_object_following/first_frame.jpg}} & \raisebox{-.5\height}{\includegraphics[width=0.09\textwidth]{figures/results/camera_object_following/frames/004.jpg}} &  \raisebox{-.5\height}{\includegraphics[width=0.09\textwidth]{figures/results/camera_object_following/frames/008.jpg}} &  \raisebox{-.5\height}{\includegraphics[width=0.09\textwidth]{figures/results/camera_object_following/frames/012.jpg}}
	\end{tblr}
	\begin{tblr}{
			cell{1}{2} = {c=4}{c},
			cell{1}{6} = {c=4}{c},
			columns = {c},
			column{3} = {rightsep=1pt},
			column{4} = {leftsep=1pt,rightsep=1pt},
			column{5} = {leftsep=1pt},
			column{7} = {rightsep=1pt},
			column{8} = {leftsep=1pt,rightsep=1pt},
			column{9} = {leftsep=1pt},
			vline{3} = {3-7}{dashed},
			vline{6} = {1-7}{},
			vline{7} = {3-7}{dashed},
			hline{3} = {1-10}{},
			hline{4} = {1-10}{},
			hline{5} = {1-10}{},
			hline{6} = {1-10}{},
			hline{7} = {1-10}{},
		}
		& Jumping towards camera & & & & Object passing over other object & & & \\
		{Ref.} & \raisebox{-.5\height}{\includegraphics[width=0.09\textwidth]{figures/results/handcrafted_robot_jumping/frames_input/001.jpg}} & \raisebox{-.5\height}{\includegraphics[width=0.09\textwidth]{figures/results/handcrafted_robot_jumping/frames_input/005.jpg}} & \raisebox{-.5\height}{\includegraphics[width=0.09\textwidth]{figures/results/handcrafted_robot_jumping/frames_input/007.jpg}} & \raisebox{-.5\height}{\includegraphics[width=0.09\textwidth]{figures/results/handcrafted_robot_jumping/frames_input/010.jpg}} & \raisebox{-.5\height}{\includegraphics[width=0.09\textwidth]{figures/results/handcrafted_passing_front/frames_input/001.jpg}} & \raisebox{-.5\height}{\includegraphics[width=0.09\textwidth]{figures/results/handcrafted_passing_front/frames_input/004.jpg}} & \raisebox{-.5\height}{\includegraphics[width=0.09\textwidth]{figures/results/handcrafted_passing_front/frames_input/008.jpg}} & \raisebox{-.5\height}{\includegraphics[width=0.09\textwidth]{figures/results/handcrafted_passing_front/frames_input/012.jpg}} \\
		{SVD}  & \raisebox{-.5\height}{\includegraphics[width=0.09\textwidth]{figures/results/handcrafted_robot_jumping/first_frame.jpg}} & \raisebox{-.5\height}{\includegraphics[width=0.09\textwidth]{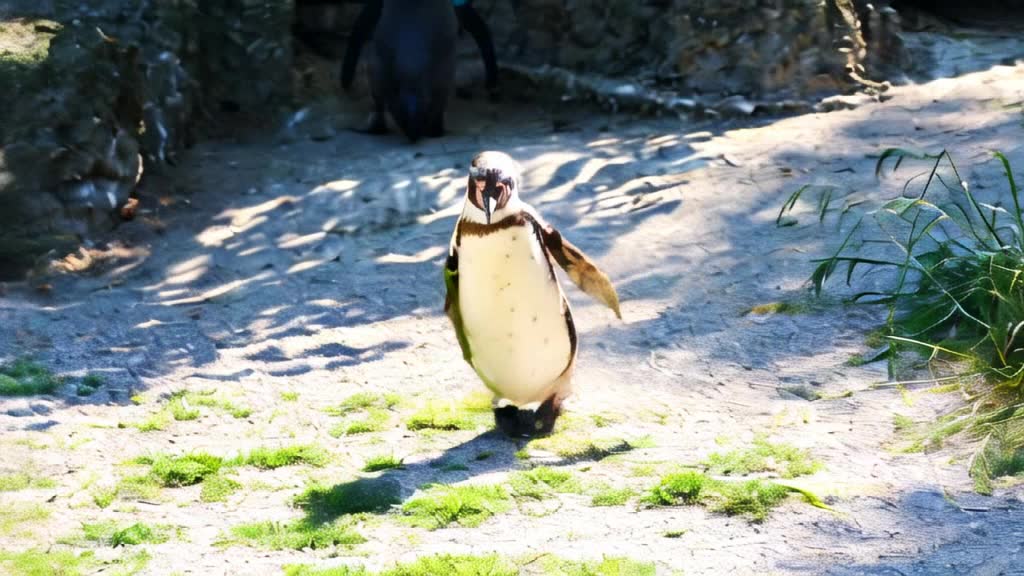}} & \raisebox{-.5\height}{\includegraphics[width=0.09\textwidth]{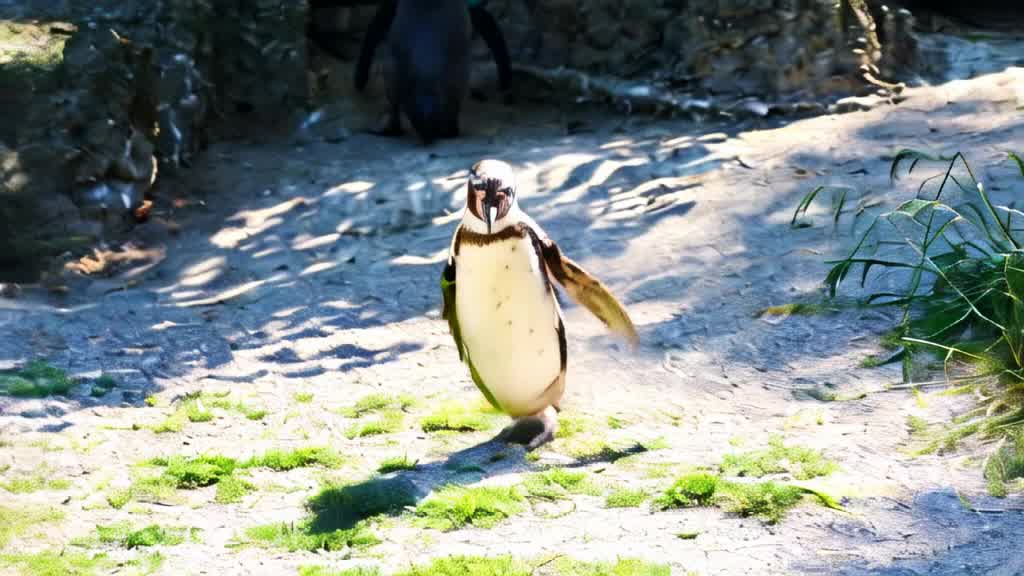}} & \raisebox{-.5\height}{\includegraphics[width=0.09\textwidth]{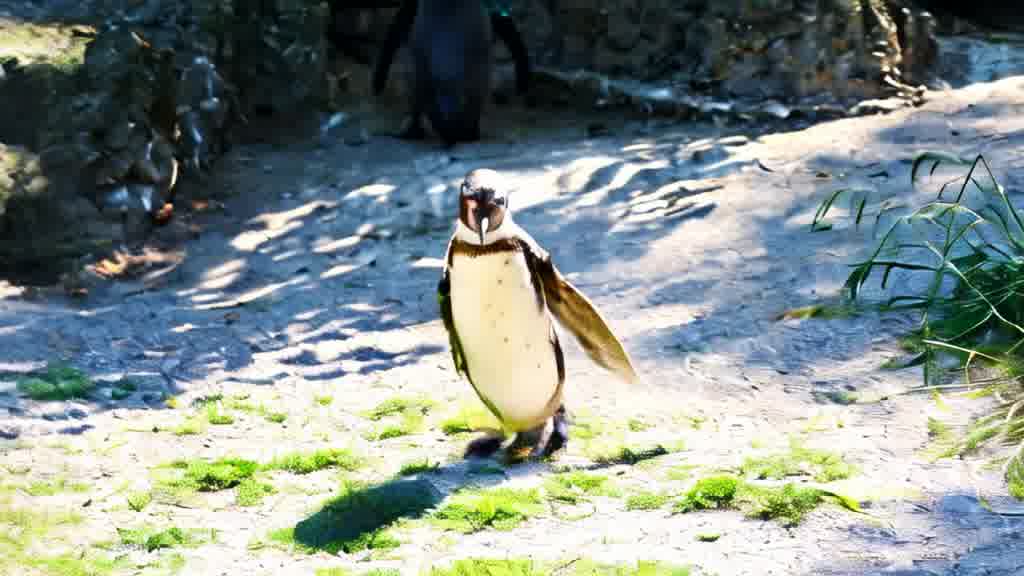}} & \raisebox{-.5\height}{\includegraphics[width=0.09\textwidth]{figures/results/handcrafted_passing_front/first_frame.jpg}} & \raisebox{-.5\height}{\includegraphics[width=0.09\textwidth]{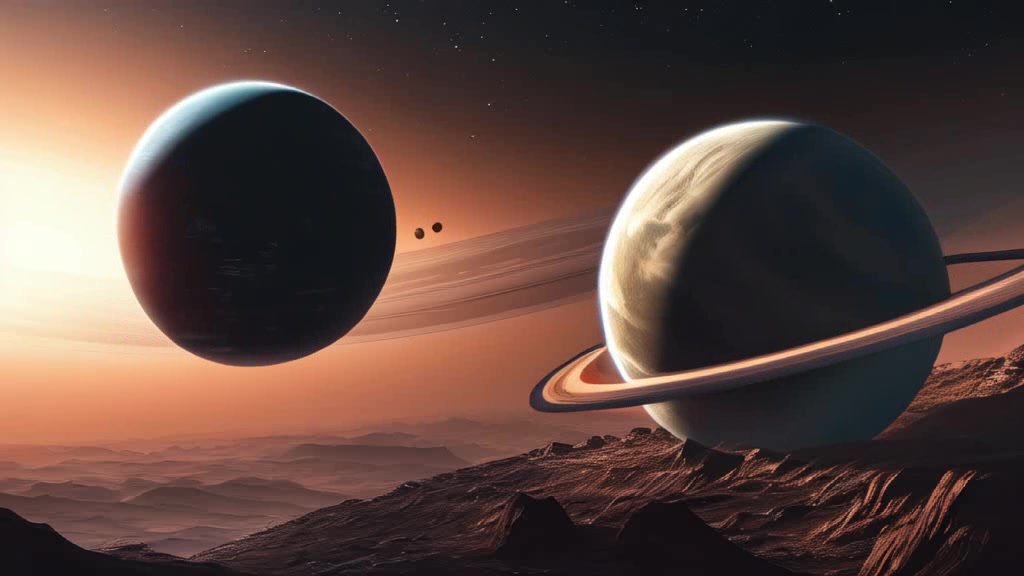}} & \raisebox{-.5\height}{\includegraphics[width=0.09\textwidth]{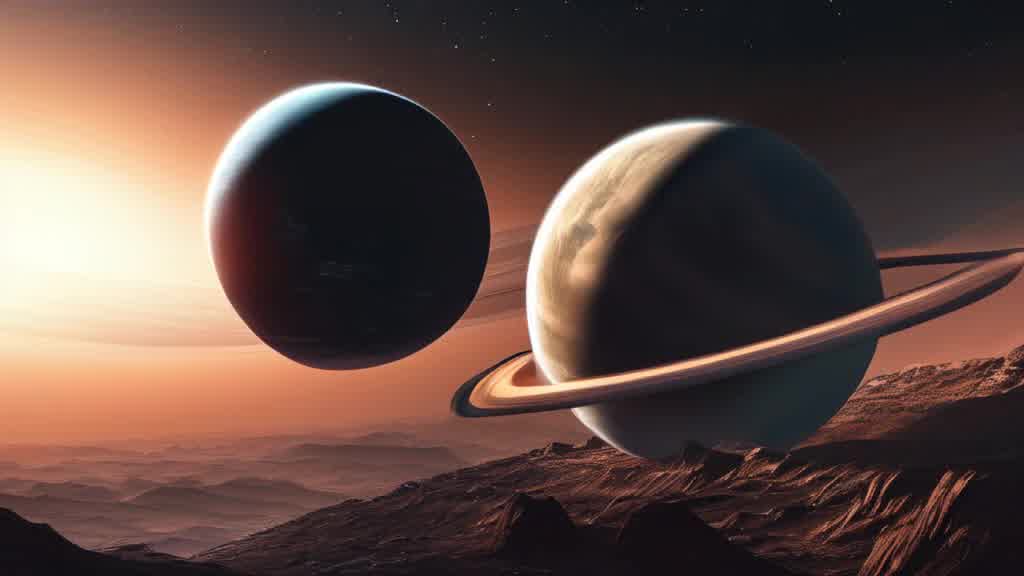}} & \raisebox{-.5\height}{\includegraphics[width=0.09\textwidth]{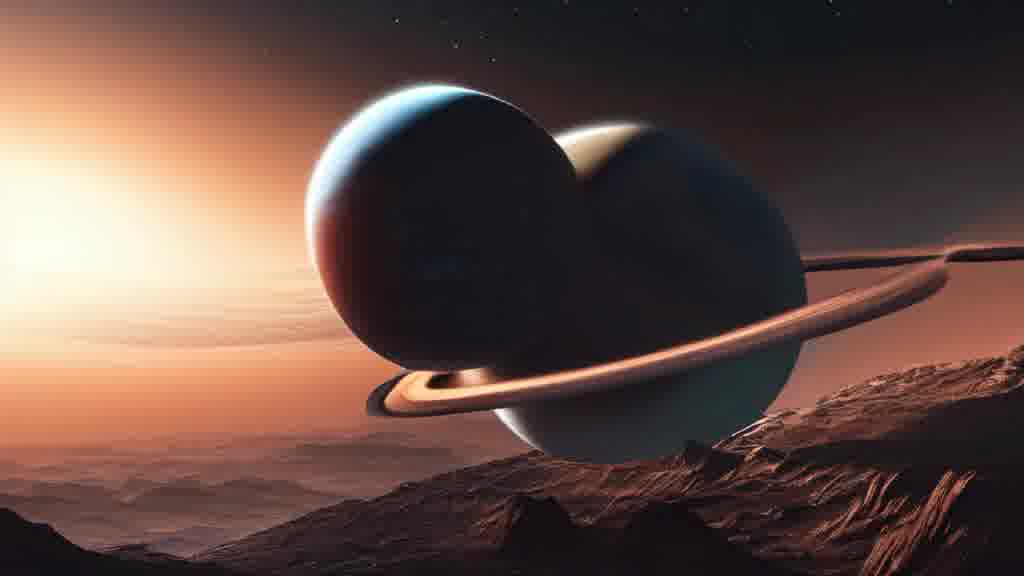}} \\
		{VC} & \raisebox{-.5\height}{\includegraphics[width=0.09\textwidth]{figures/results/handcrafted_robot_jumping/first_frame.jpg}} & \raisebox{-.5\height}{\includegraphics[width=0.050625\textwidth]{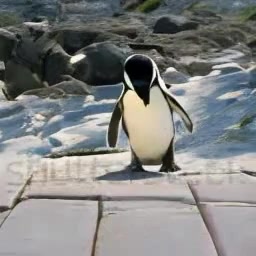}} & \raisebox{-.5\height}{\includegraphics[width=0.050625\textwidth]{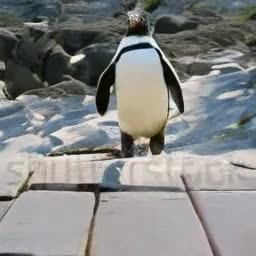}} & \raisebox{-.5\height}{\includegraphics[width=0.050625\textwidth]{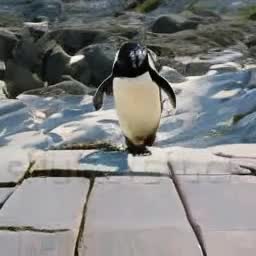}} & \raisebox{-.5\height}{\includegraphics[width=0.09\textwidth]{figures/results/handcrafted_passing_front/first_frame.jpg}} & \raisebox{-.5\height}{\includegraphics[width=0.050625\textwidth]{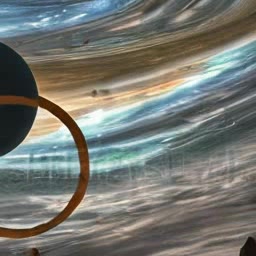}} & \raisebox{-.5\height}{\includegraphics[width=0.050625\textwidth]{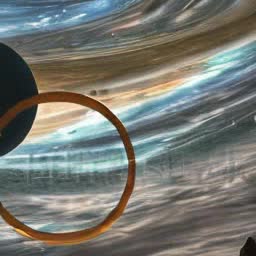}} & \raisebox{-.5\height}{\includegraphics[width=0.050625\textwidth]{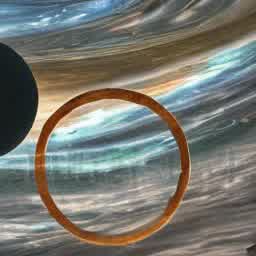}} \\
		{MC} & \raisebox{-.5\height}{\includegraphics[width=0.09\textwidth]{figures/results/handcrafted_robot_jumping/first_frame.jpg}} & \raisebox{-.5\height}{\includegraphics[width=0.050625\textwidth]{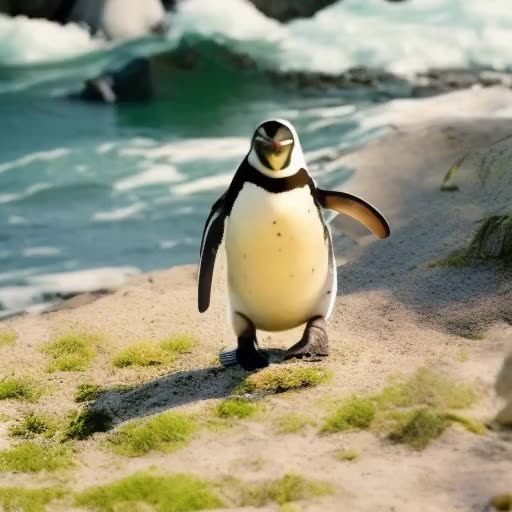}} & \raisebox{-.5\height}{\includegraphics[width=0.050625\textwidth]{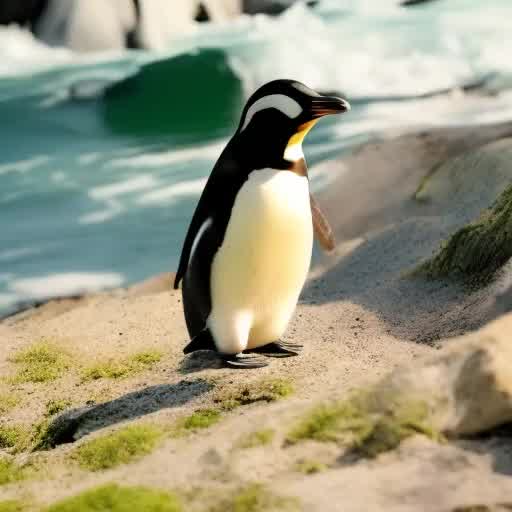}} & \raisebox{-.5\height}{\includegraphics[width=0.050625\textwidth]{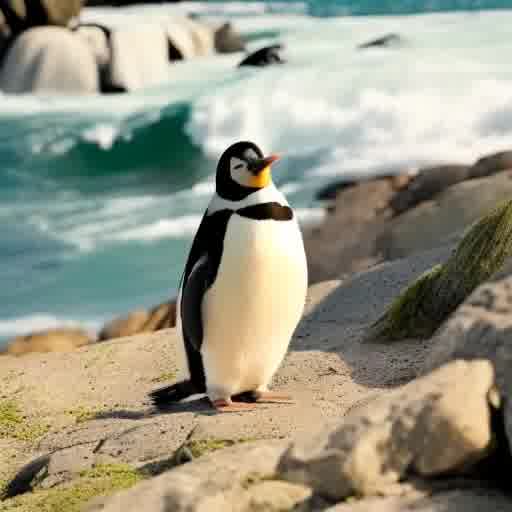}} & \raisebox{-.5\height}{\includegraphics[width=0.09\textwidth]{figures/results/handcrafted_passing_front/first_frame.jpg}} & \raisebox{-.5\height}{\includegraphics[width=0.050625\textwidth]{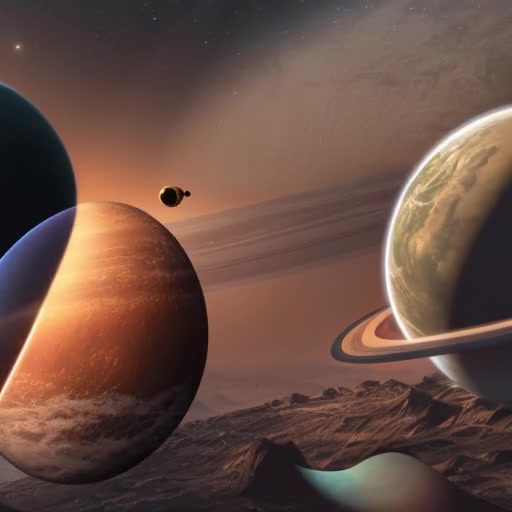}} & \raisebox{-.5\height}{\includegraphics[width=0.050625\textwidth]{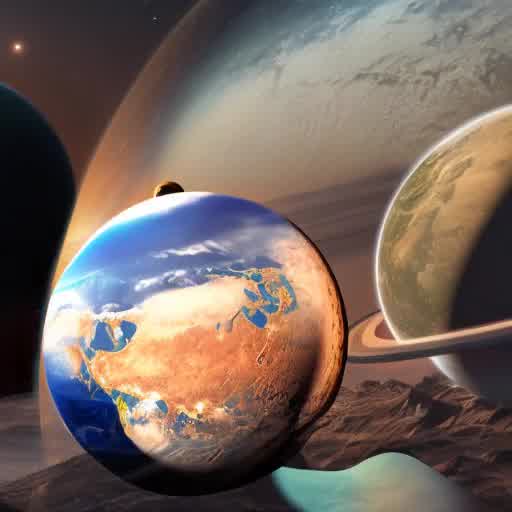}} & \raisebox{-.5\height}{\includegraphics[width=0.050625\textwidth]{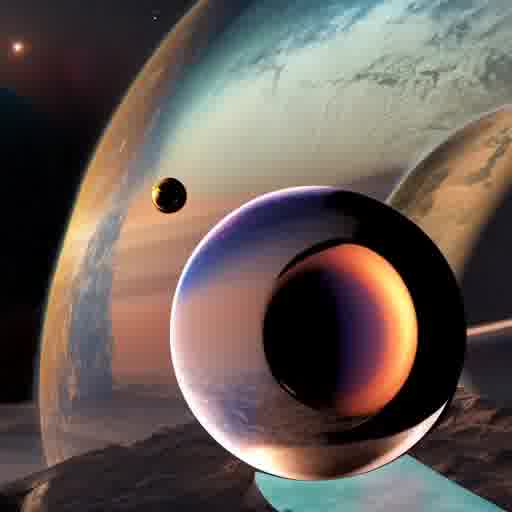}} \\
		{MD} & \raisebox{-.5\height}{\includegraphics[width=0.09\textwidth]{figures/results/handcrafted_robot_jumping/first_frame.jpg}} & \raisebox{-.5\height}{\includegraphics[width=0.050625\textwidth]{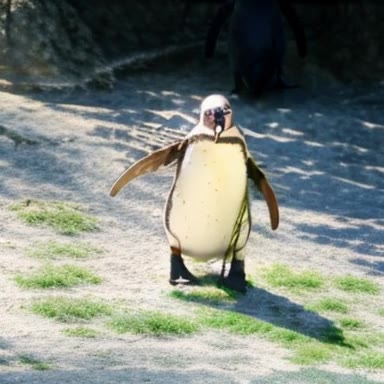}} & \raisebox{-.5\height}{\includegraphics[width=0.050625\textwidth]{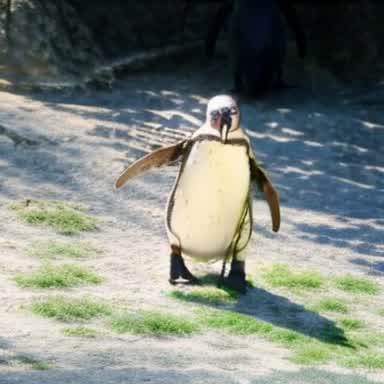}} & \raisebox{-.5\height}{\includegraphics[width=0.050625\textwidth]{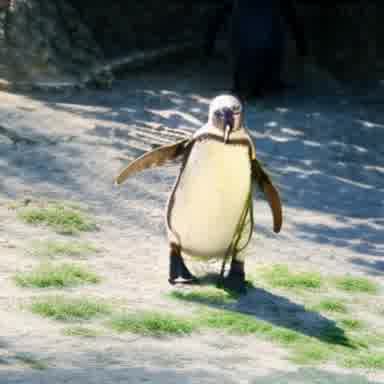}} & \raisebox{-.5\height}{\includegraphics[width=0.09\textwidth]{figures/results/handcrafted_passing_front/first_frame.jpg}} & \raisebox{-.5\height}{\includegraphics[width=0.050625\textwidth]{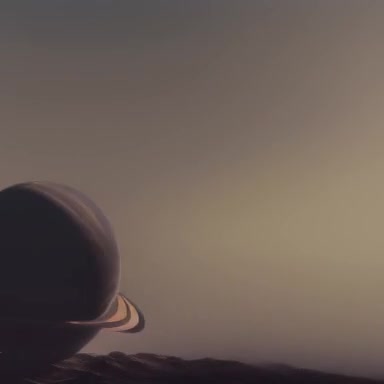}} & \raisebox{-.5\height}{\includegraphics[width=0.050625\textwidth]{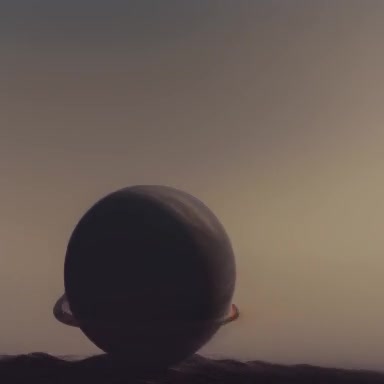}} & \raisebox{-.5\height}{\includegraphics[width=0.050625\textwidth]{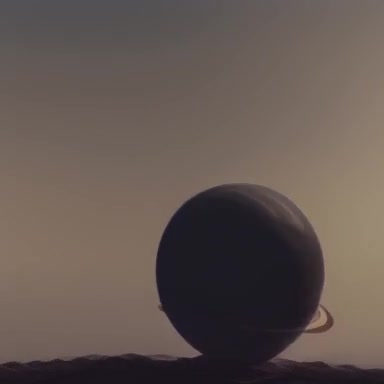}} \\
		{\textbf{Ours}} & \raisebox{-.5\height}{\includegraphics[width=0.09\textwidth]{figures/results/handcrafted_robot_jumping/first_frame.jpg}} & \raisebox{-.5\height}{\includegraphics[width=0.09\textwidth]{figures/results/handcrafted_robot_jumping/frames/005.jpg}} & \raisebox{-.5\height}{\includegraphics[width=0.09\textwidth]{figures/results/handcrafted_robot_jumping/frames/007.jpg}} & \raisebox{-.5\height}{\includegraphics[width=0.09\textwidth]{figures/results/handcrafted_robot_jumping/frames/010.jpg}} & \raisebox{-.5\height}{\includegraphics[width=0.09\textwidth]{figures/results/handcrafted_passing_front/first_frame.jpg}} & \raisebox{-.5\height}{\includegraphics[width=0.09\textwidth]{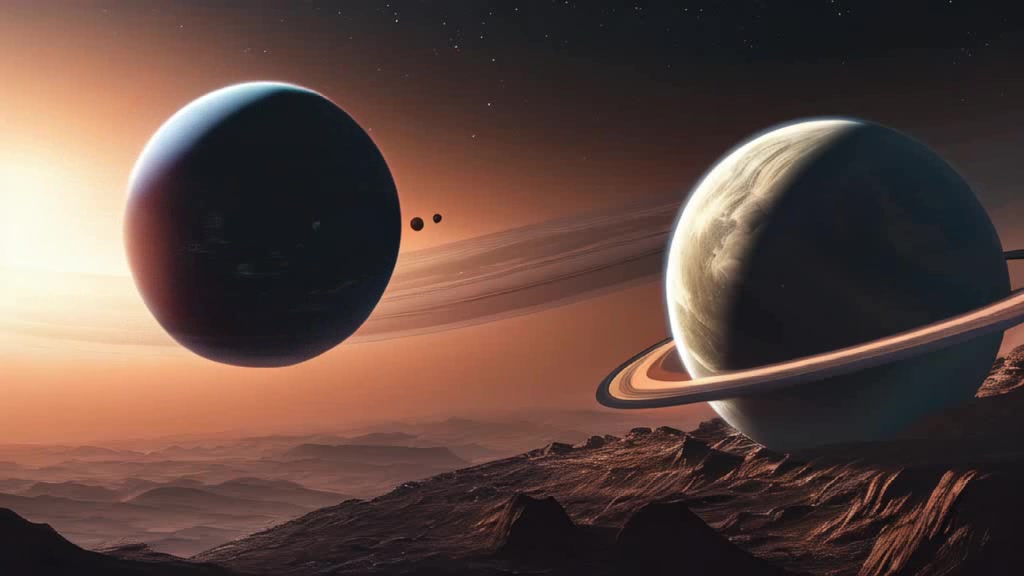}} &  \raisebox{-.5\height}{\includegraphics[width=0.09\textwidth]{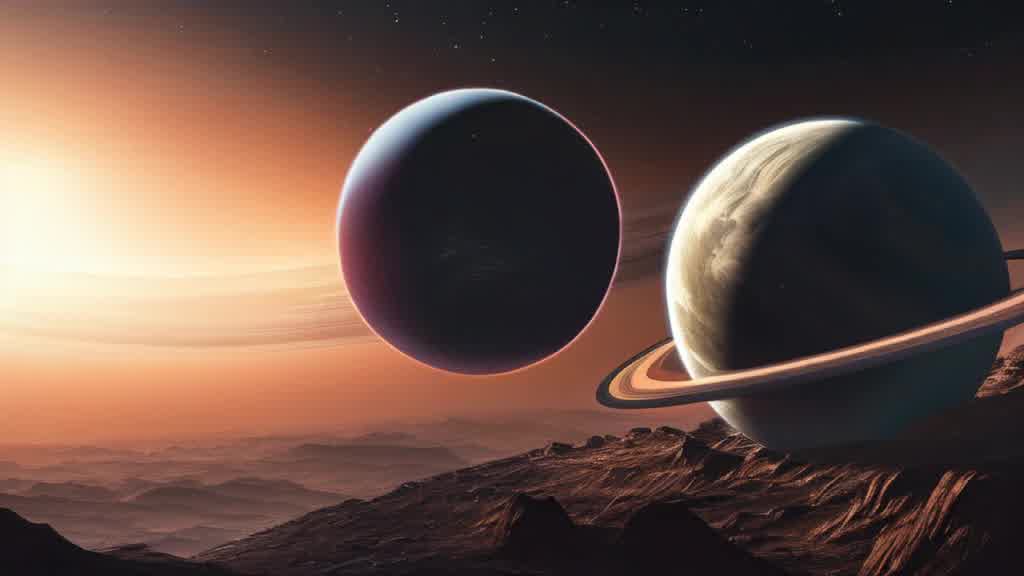}} &  \raisebox{-.5\height}{\includegraphics[width=0.09\textwidth]{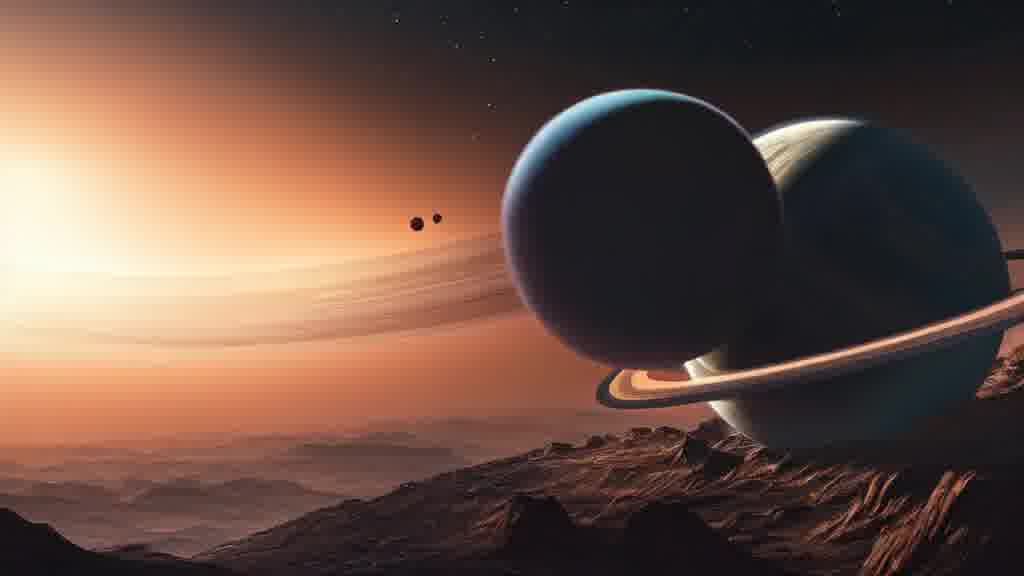}}
	\end{tblr}
	\caption{Qualitative evaluation for additional examples (2/2). We compare our method to SVD = Stable Video Diffusion~\cite{svd} (baseline, no motion input), VC = VideoComposer~\cite{videocomposer}, MC = MotionClone~\cite{motionclone}, and MD = MotionDirector~\cite{motiondirector} for four different motions and target images. 
	}
	\Description{Grid with two blocks on top of each other. Each block has different methods in the rows and two different motion reference videos and starting frames in the columns. From top to bottom and left to right: rotating bird's-eye view camera from landscape to landscape, following camera from car to car, simple animation of stick figure jumping to penguin, 2D ball passing in front of another 2D ball to a planet passing in front of another planet.}
	\label{fig:qual_eval_extra_2}
\end{figure*}

To further demonstrate the effectiveness of our method in transferring semantic motion from a reference video to target images, we generated videos using state-of-the-art competing methods for the same examples presented in Fig.~\ref{fig:results}. These results, covering a range of motion types and complexities, are provided in Fig.~\ref{fig:qual_eval_extra_1} and Fig.~\ref{fig:qual_eval_extra_2}. As before, competing methods suffer from problems inherent to their class of methods. Stable Video Diffusion~\cite{svd}, lacking a motion input, typically fails to follow the reference motion. VideoComposer~\cite{videocomposer}, an image-to-video method with dense motion inputs, struggles when the reference video's motions are not aligned with the input image. In such cases, the method applies the spatial but not semantic motion, leading to either unwanted background movement or the foreground object morphing into the spatial position where the motion occurs in the reference video. MotionDirector~\cite{motiondirector}, based on a text-to-video model, cannot directly use the target image as input and must instead learn its appearance. As a result, the generated videos often deviate in appearance and spatial layout from the target image.

\clearpage

\subsection{Additional Information for the Quantitative Evaluation} \label{sec:additional-info-quant-eval}

We selected the action classes from the Something-Something V2 data set~\cite{something_something} according to the following criteria:
\begin{itemize}
	\item All interacting objects typically appear in the start frame.
	\item The action is typically long enough, so that it appears in most of the frames.
	\item The class is sufficiently different from other classes.
\end{itemize}

We then extracted the first 11 examples of the given class (with some manual filtering in case the above criteria is not met) and took the first video as motion reference video and the first frames of the other 10 for the target images. Table~\ref{table:quant_eval_classes} lists the final class IDs and video IDs used.

\begin{table*}[htbp]
	\centering
	\caption{Quantitative evaluation data. List of video IDs from the Something-Something V2 data set~\cite{something_something} used in our quantitative evaluation.}
	\label{table:quant_eval_classes}
	\footnotesize{
		\begin{tabular}{ll}
			\toprule
			Class ID: Label & Video ID for Motion Reference Video: Video IDs for Target Images \\
			\midrule
			0: Approaching something with your camera & 31416: 174027, 49364, 179191, 58108, 219270, 124642, 18253, 112846, 75372, 201968 \\
			23: Letting something roll down a slanted surface & 97908 : 220450, 22070, 46282, 136926, 216643, 109913, 137160, 69704, 19903, 86892 \\
			27: Lifting something up completely without letting it drop down & 144105: 181548, 167709, 81608, 132100, 167837, 46057, 158390, 41755, 93247, 106014 \\
			32: Moving away from something with your camera & 121394: 3201, 100064, 35438, 44298, 123636, 4328, 178356, 76980, 71173, 33210 \\
			36: Moving something and something away from each other & 51295: 4443, 88084, 76718, 132951, 49285, 43627, 45186, 18456, 18788, 142654 \\
			37: Moving something and something closer to each other & 87711: 180193, 137350, 39979, 150128, 10055, 16205, 208340, 97632, 94171, 99258 \\
			41: Moving something away from the camera & 207150:  205156, 108506, 139808, 44794, 68922, 197965, 201362, 153856, 21809, 211202 \\
			44: Moving something towards the camera & 160529: 145447, 30260, 118270, 10405, 66666, 154312, 157137, 106357, 164212, 176798 \\
			92: Pulling two ends of something so that it separates into two pieces & 187909: 162071, 51196, 87892, 11780, 75398, 148274, 113149, 177507, 47061, 28237 \\
			165: Turning the camera downwards while filming something & 169117: 120585, 131318, 68372, 104829, 162135, 124382, 108641, 98914, 197549, 213899 \\
			\bottomrule
			\\
		\end{tabular}
	}
\end{table*}

The 10 action classes used in our evaluation can be grouped into two categories: five involving camera motion (IDs: 0, 32, 41, 44, 165) and five involving object motion (IDs: 23, 27, 36, 37, 92). Table~\ref{table:quant_eval_by_motion} provides the quantitative results from Table~\ref{table:quant_eval}, aggregated by motion category.
We observe that image appearance preservation is generally worse for camera motions. This is likely because strong camera movements cause significant changes in the visual content. In contrast, video motion fidelity is typically higher for camera motions, possibly because the movements are more uniform and linear, and spatial alignment between the motion reference video and target image is less critical. As a result, methods that mostly transfer spatial rather than semantic motion (e.g., VideoComposer~\cite{videocomposer}) can still perform well for camera motions.

\begin{table*}[htbp]
	\centering
	\caption{Quantitative evaluation aggregated by motion category (camera/object). As in Table~\ref{table:quant_eval}, we compare our method to Stable Video Diffusion~\cite{svd} (baseline, no motion input), VideoComposer~\cite{videocomposer}, MotionClone~\cite{motionclone}, and MotionDirector~\cite{motiondirector}. The first value in each cell corresponds to camera motions and the second to object motions. The best performing method per column is marked in bold.}
	\label{table:quant_eval_by_motion}
	\small{
		\begin{tabular}{llccclcccclc}
			\toprule
			\multirow{2}{*}{Method} & & \multicolumn{3}{c}{Image Appearance Preservation} & & \multicolumn{4}{c}{Video Motion Fidelity} & & \multicolumn{1}{c}{Overall} \\
			\cmidrule{3-5} \cmidrule{7-10} \cmidrule{12-12}
			& & {CLIP-Avg $\uparrow$} & {CLIP-1st $\uparrow$} & {User rank $\downarrow$} & & {Acc-Top-1 $\uparrow$} & {Acc-Top-5 $\uparrow$} & {Cos-Sim $\uparrow$} & {User rank $\downarrow$} & & {User rank $\downarrow$} \\
			\midrule
			Stable Video Diffusion & & \textbf{0.837}/\textbf{0.849} & 0.842/0.857 & \textbf{1.215}/\textbf{1.378} & & 4\%/2\% & 4\%/6\% & 0.398/0.342 & 4.689/3.733 & & 3.311/2.333 \\
			VideoComposer & & 0.713/0.726 & 0.853/0.860 & 3.867/3.704 & & 64\%/24\% & 82\%/42\% & 0.575/0.419 & 2.941/3.119 & & 3.407/3.696 \\
			MotionClone & & 0.610/0.664 & \textbf{0.881}/0.890 & 4.778/4.393 & & 48\%/26\% & 80\%/44\% & 0.555/0.491 & 3.215/3.059 & & 4.385/4.015 \\
			MotionDirector & & 0.738/0.762 & 0.752/0.774 & 3.185/3.859 & & 38\%/24\% & 58\%/58\% & 0.545/0.501 & 3.067/2.733 & & 2.785/3.333 \\
			\midrule
			Ours & & 0.745/0.813 & 0.873/\textbf{0.894} & 1.956/1.667 & & \textbf{72\%}/\textbf{36\%} & \textbf{86\%}/\textbf{66\%} & \textbf{0.785}/\textbf{0.606} & \textbf{1.089}/\textbf{2.356} & & \textbf{1.111}/\textbf{1.622} \\
			\bottomrule
			\\
		\end{tabular}
	}
\end{table*}

Our method consistently outperforms all compared methods across both motion categories in terms of video motion fidelity. Notably, for object motions, the advantage over MotionDirector~\cite{motiondirector} is even more pronounced than the mean user rank suggests: our method was selected as the best in 58\% of comparisons, compared to only 22\% for MotionDirector. The relatively high mean rank of our method can be attributed to occasional failure cases (further discussed in Section~\ref{sec:failure-rate-analysis}) which greatly affect the average.
In terms of appearance preservation, Stable Video Diffusion~(SVD)~\cite{svd} slightly outperforms our approach, though this may be because SVD often produces very limited motion, making it easier to maintain the appearance of the input image.
When considering the overall user preference, our method shows a substantial lead: it was voted best among the five compared methods in 90\% of the evaluations for camera motions and 65\% for object motions. Notably, for object motions, Stable Video Diffusion, despite lacking any motion input, was voted best in 33\% of cases, while all other methods combined accounted for just 2\%. We believe this can be explained as follows: when our method succeeds, it significantly outperforms all other methods; when it fails, e.g., due to challenging motion reference videos or target images, SVD's conservative, low-motion outputs tend to be the most visually coherent and thus the preferred choice.

\section{Additional Ablation Study Results} \label{sec:additional-ablation}

\begin{figure*}[htbp]
	\centering
	\begin{tblr}{
			vline{3} = {2-6}{dashed},
			vline{4} = {1-6}{},
			vline{5} = {2-6}{dashed},
		}
		Reference & \raisebox{-.5\height}{\includegraphics[width=0.09\textwidth]{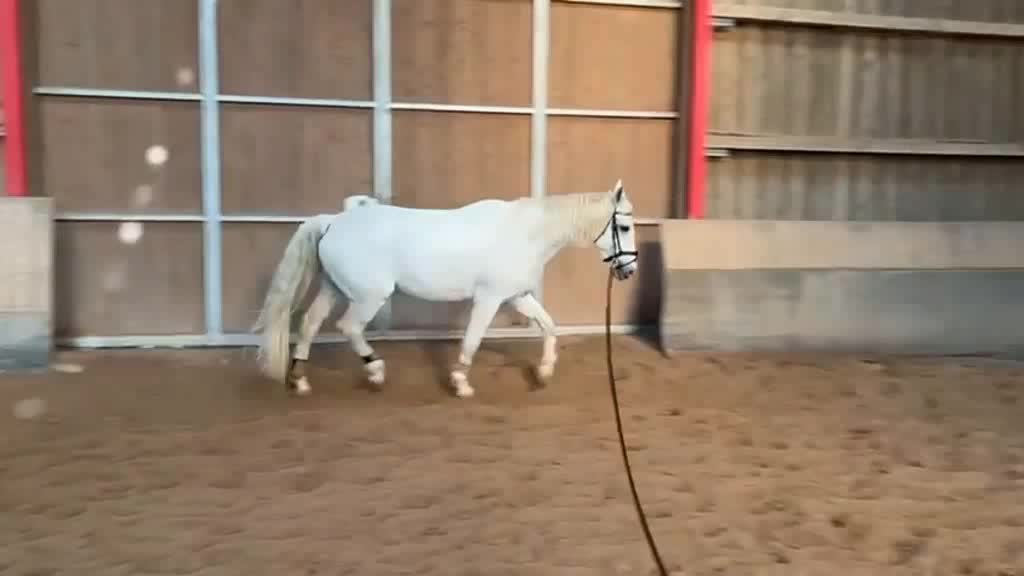}} & \raisebox{-.5\height}{\includegraphics[width=0.09\textwidth]{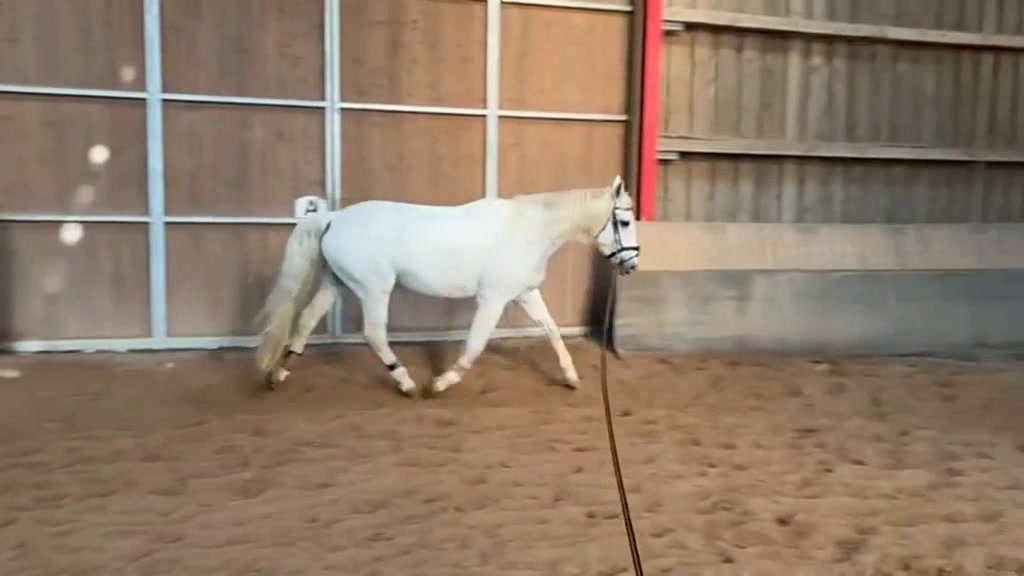}} \raisebox{-.5\height}{\includegraphics[width=0.09\textwidth]{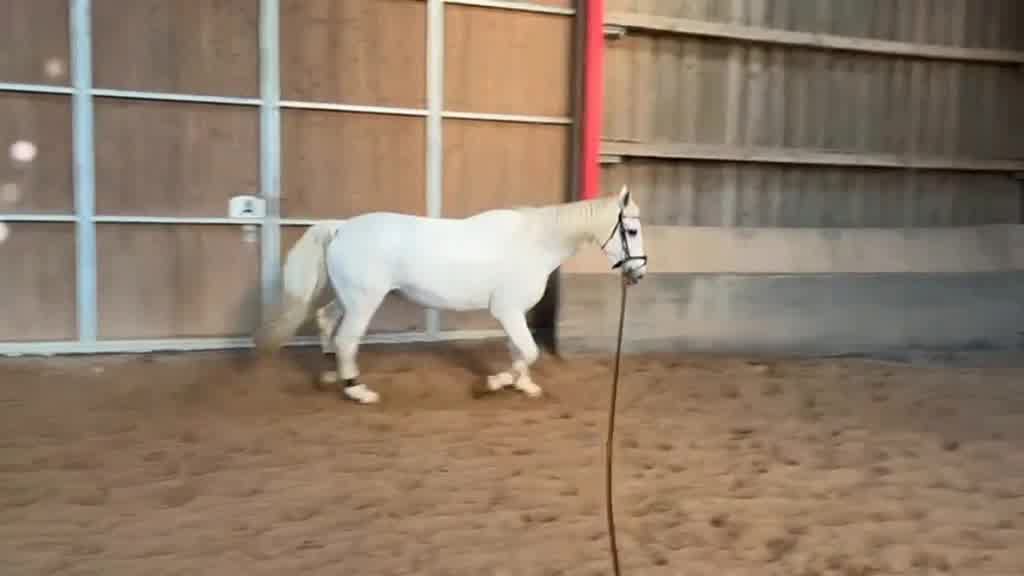}} \raisebox{-.5\height}{\includegraphics[width=0.09\textwidth]{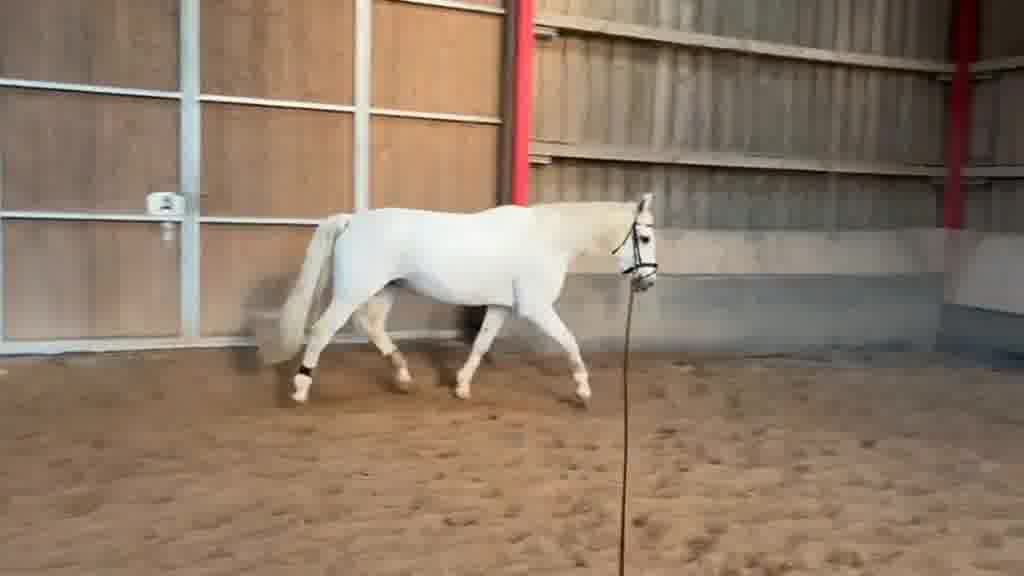}} & \raisebox{-.5\height}{\includegraphics[width=0.09\textwidth]{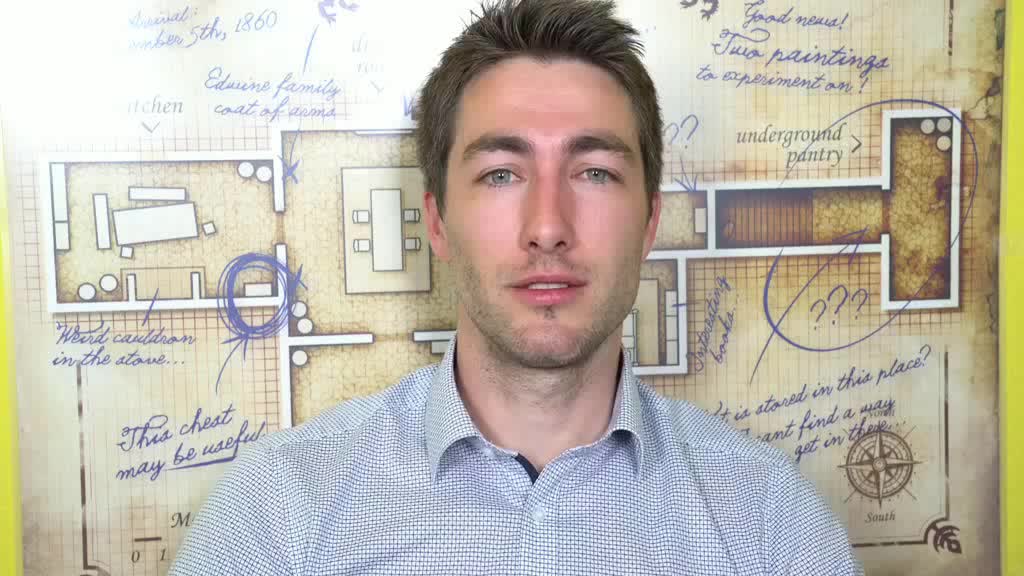}} & \raisebox{-.5\height}{\includegraphics[width=0.09\textwidth]{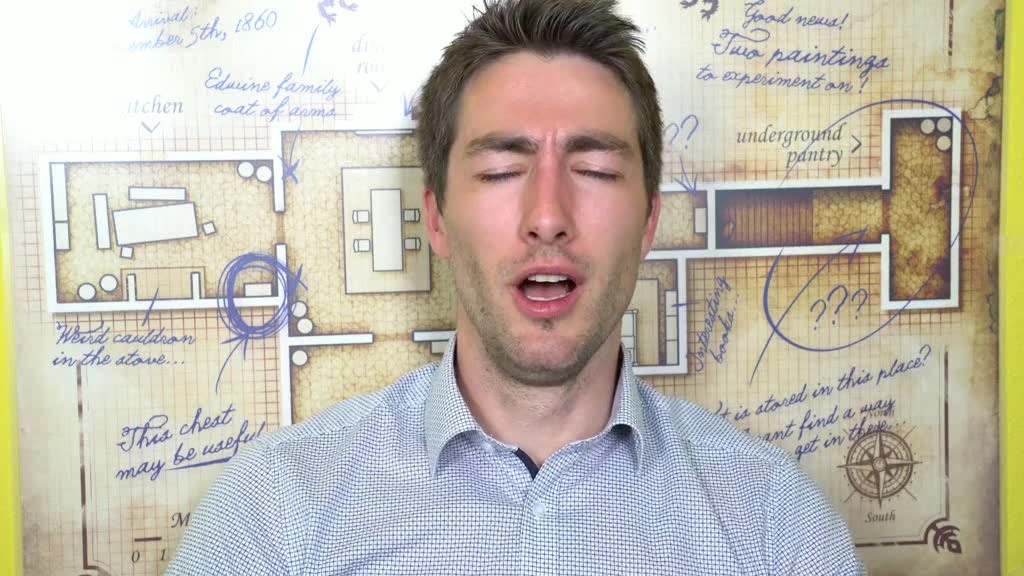}} \raisebox{-.5\height}{\includegraphics[width=0.09\textwidth]{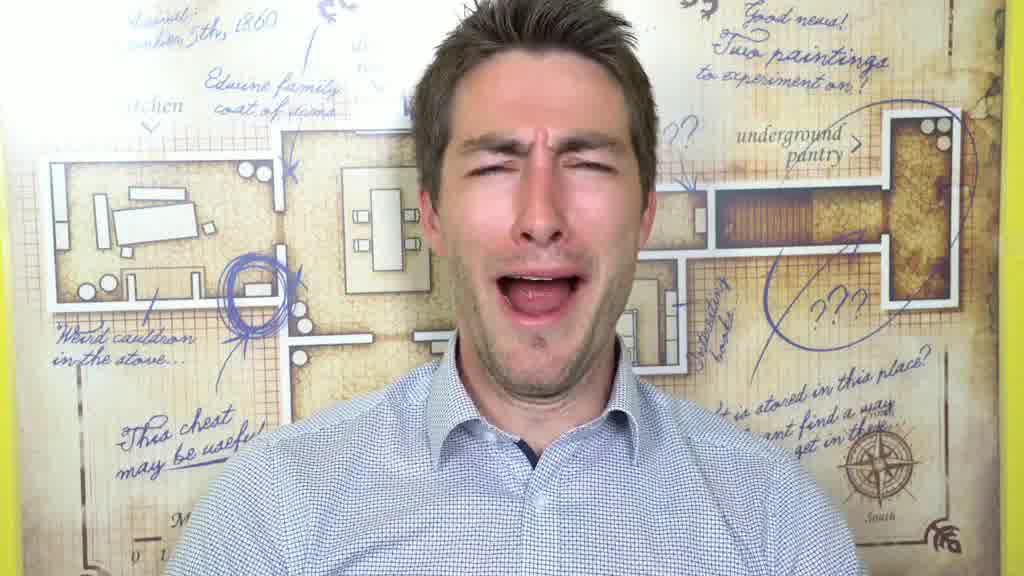}} \raisebox{-.5\height}{\includegraphics[width=0.09\textwidth]{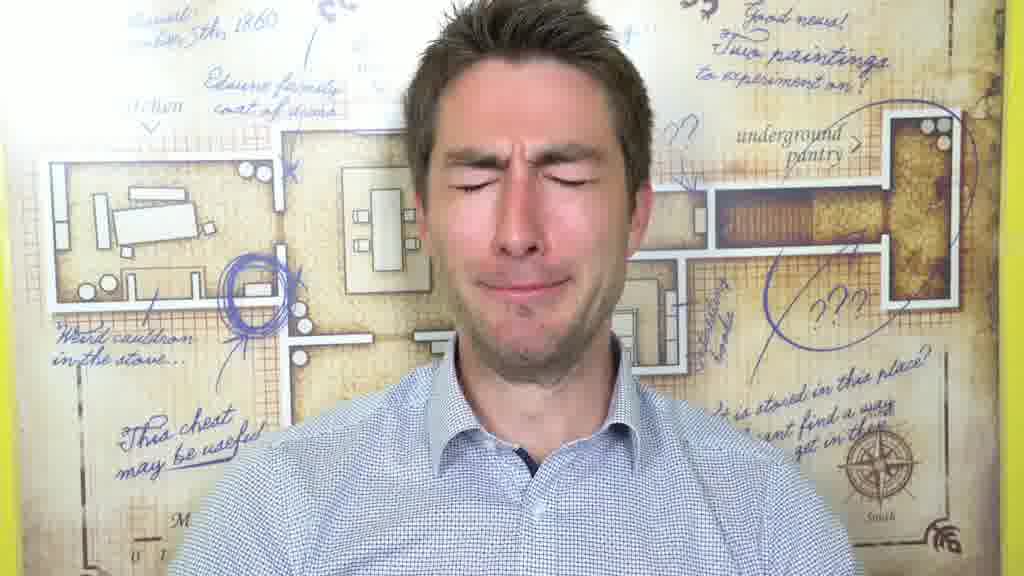}} \\
		$F'=1, N=1$ & \raisebox{-.5\height}{\includegraphics[width=0.09\textwidth]{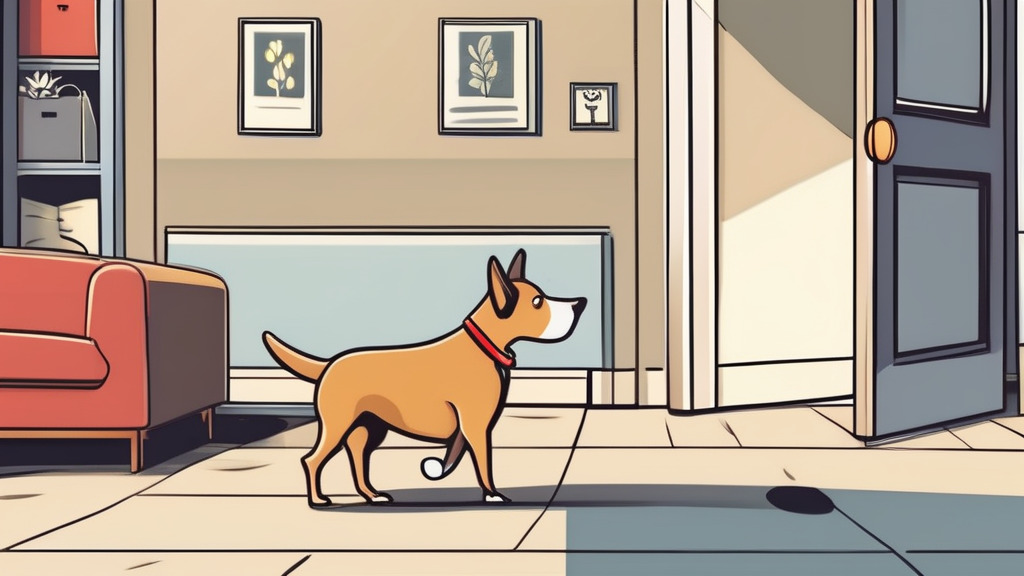}} & \raisebox{-.5\height}{\includegraphics[width=0.09\textwidth]{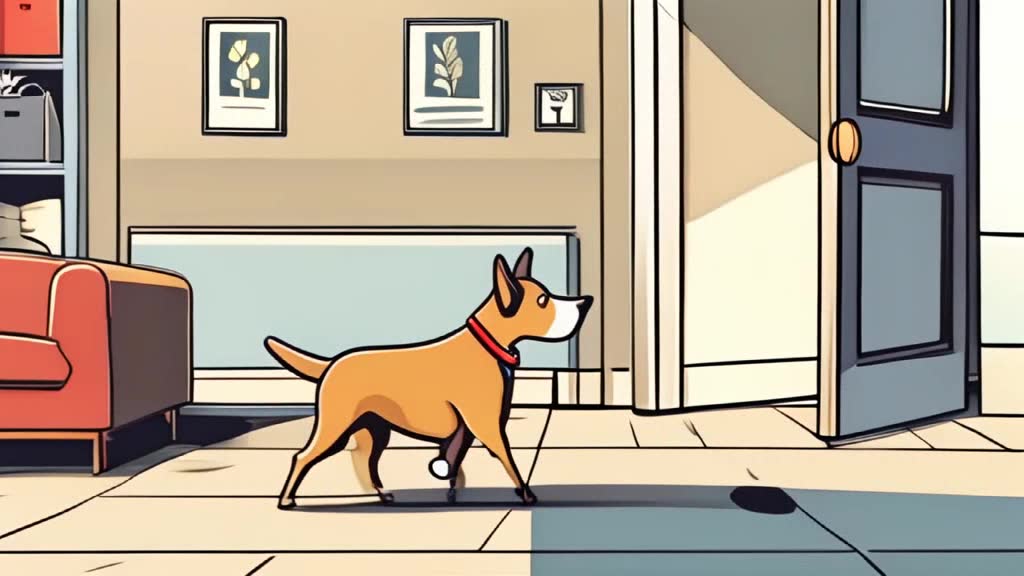}} \raisebox{-.5\height}{\includegraphics[width=0.09\textwidth]{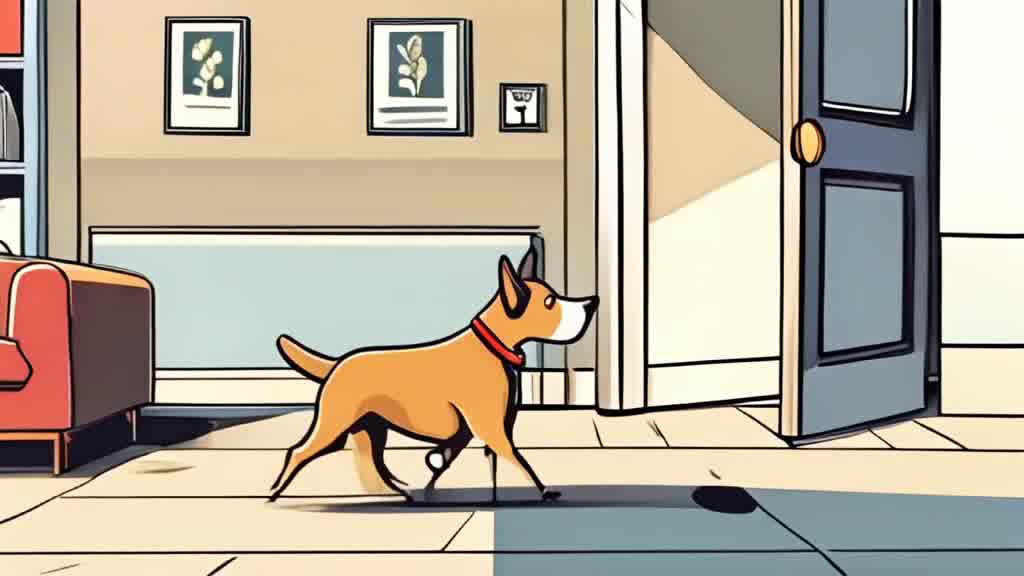}} \raisebox{-.5\height}{\includegraphics[width=0.09\textwidth]{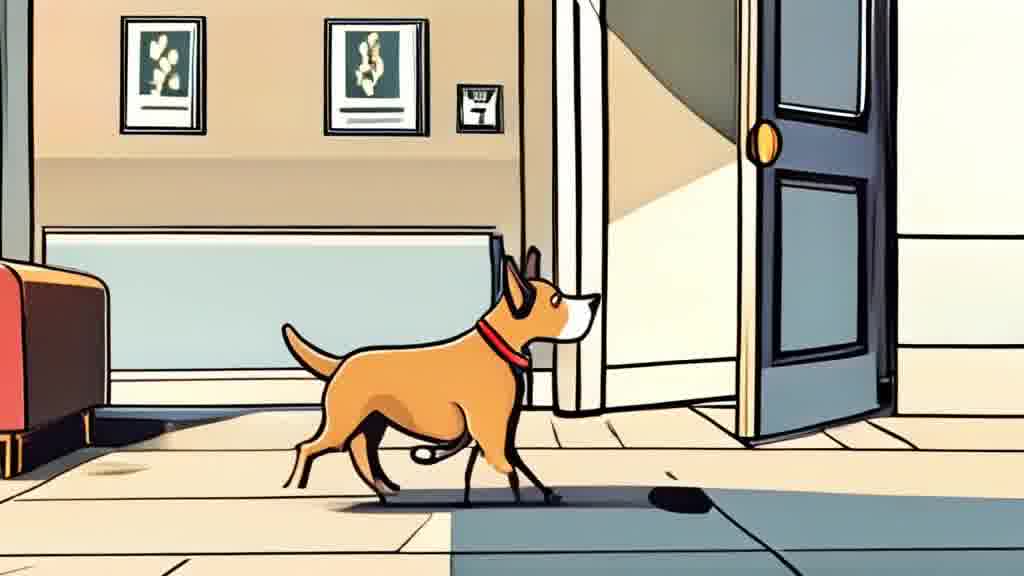}} & \raisebox{-.5\height}{\includegraphics[width=0.09\textwidth]{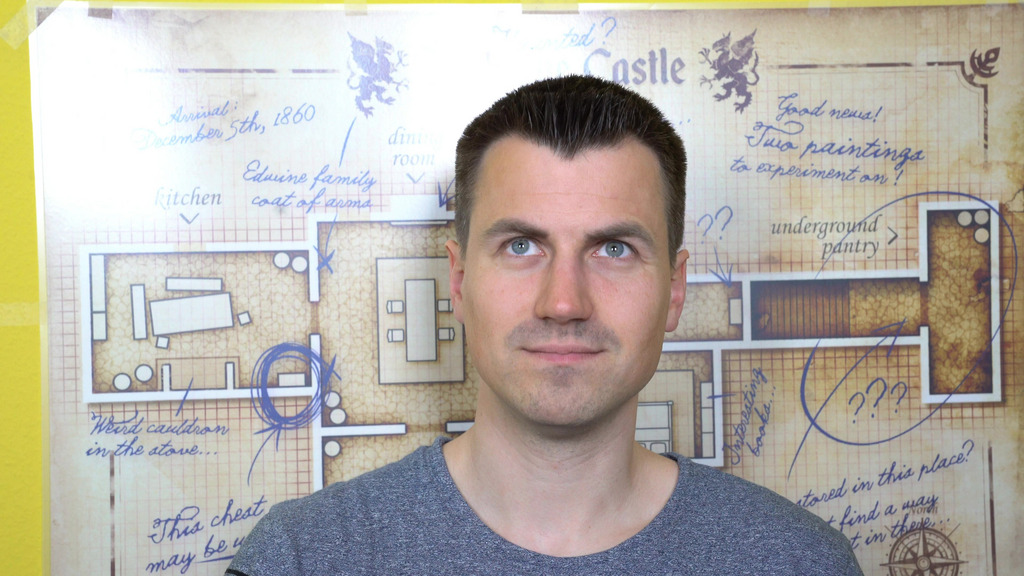}} & \raisebox{-.5\height}{\includegraphics[width=0.09\textwidth]{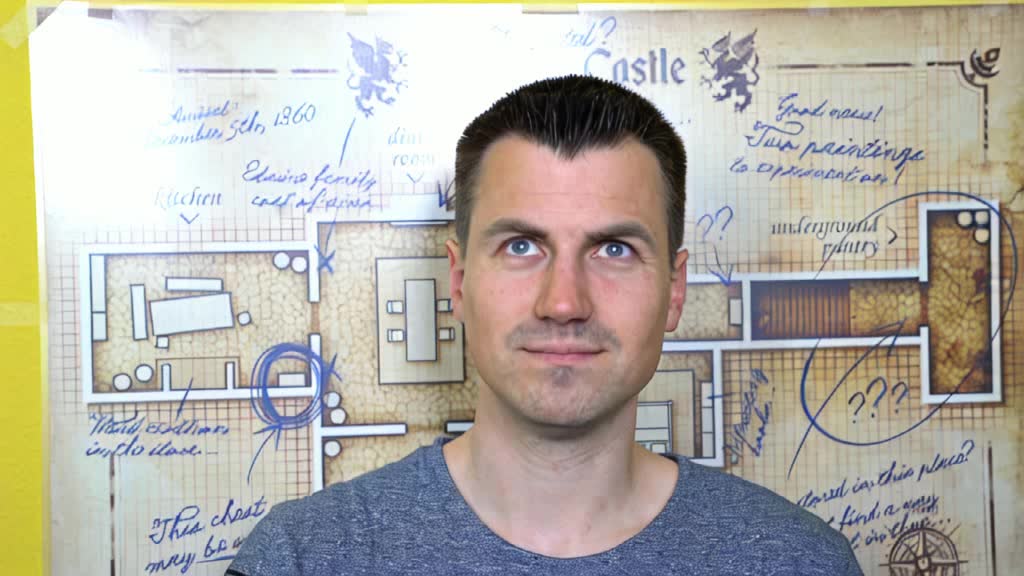}} \raisebox{-.5\height}{\includegraphics[width=0.09\textwidth]{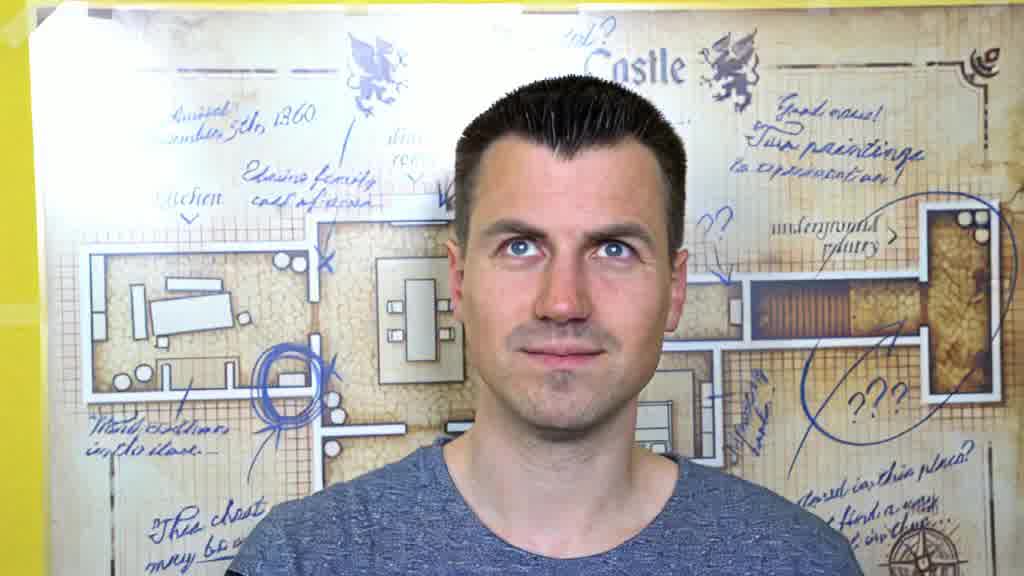}} \raisebox{-.5\height}{\includegraphics[width=0.09\textwidth]{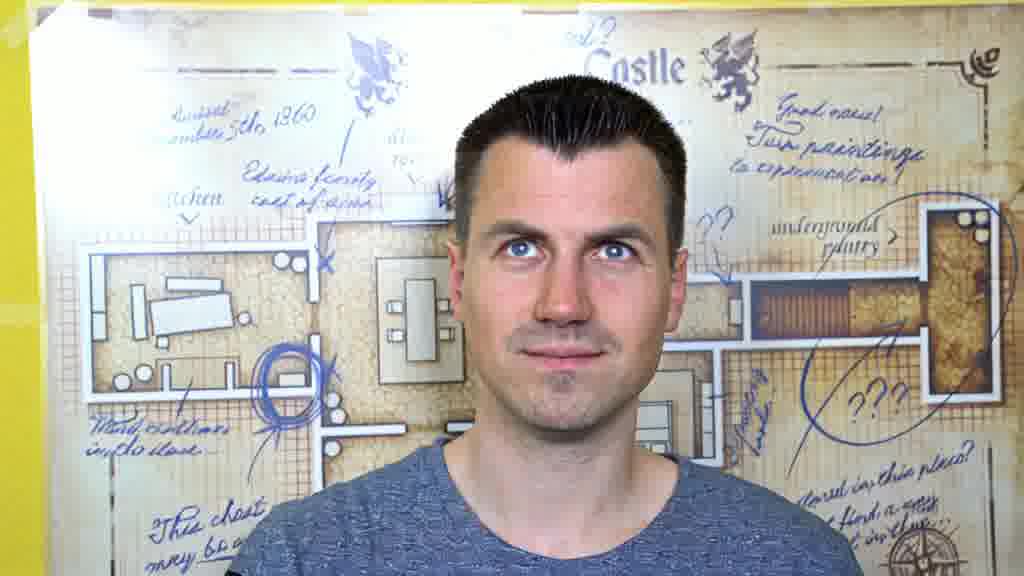}} \\
		$F'=1, N=15$ & \raisebox{-.5\height}{\includegraphics[width=0.09\textwidth]{figures/ablation_study_extra/animal_fourlegged/first_frame.jpg}} & \raisebox{-.5\height}{\includegraphics[width=0.09\textwidth]{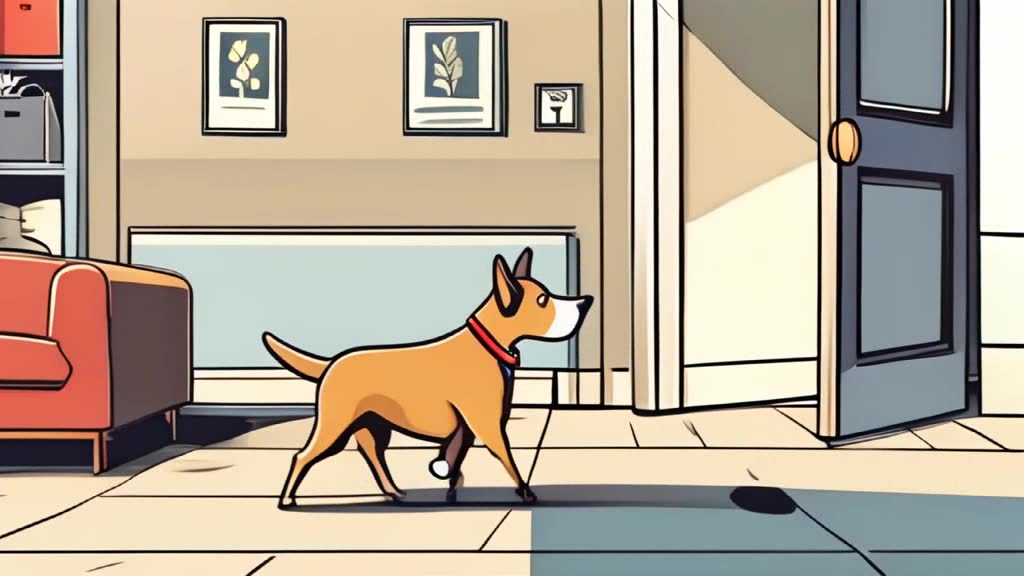}} \raisebox{-.5\height}{\includegraphics[width=0.09\textwidth]{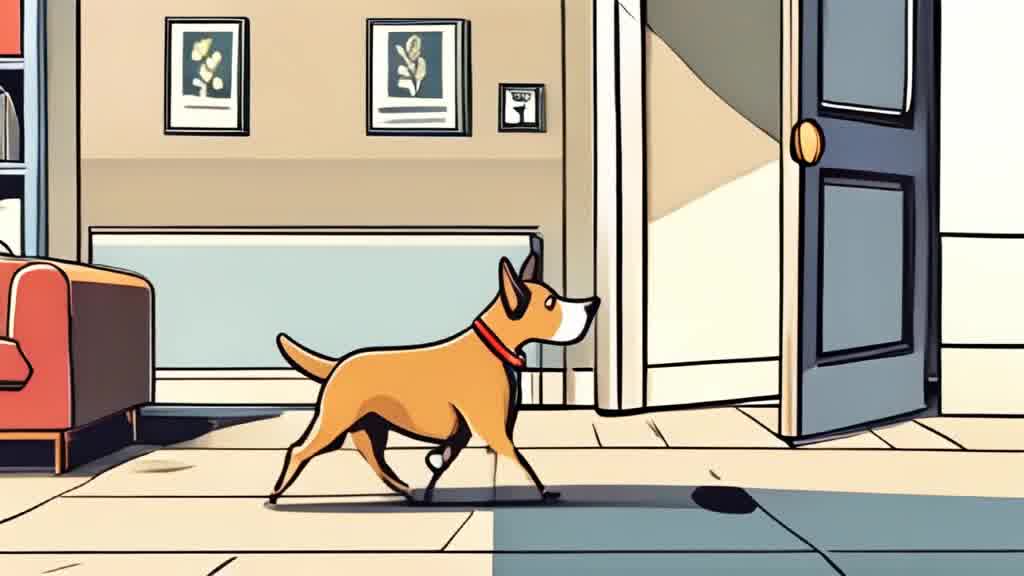}} \raisebox{-.5\height}{\includegraphics[width=0.09\textwidth]{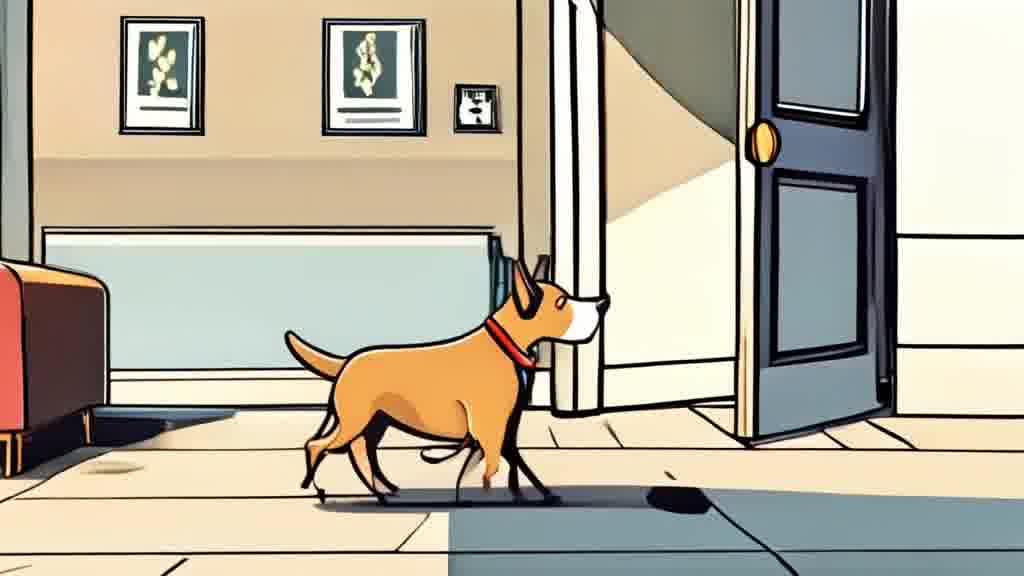}} & \raisebox{-.5\height}{\includegraphics[width=0.09\textwidth]{figures/ablation_study_extra/face_neutral/first_frame.jpg}} & \raisebox{-.5\height}{\includegraphics[width=0.09\textwidth]{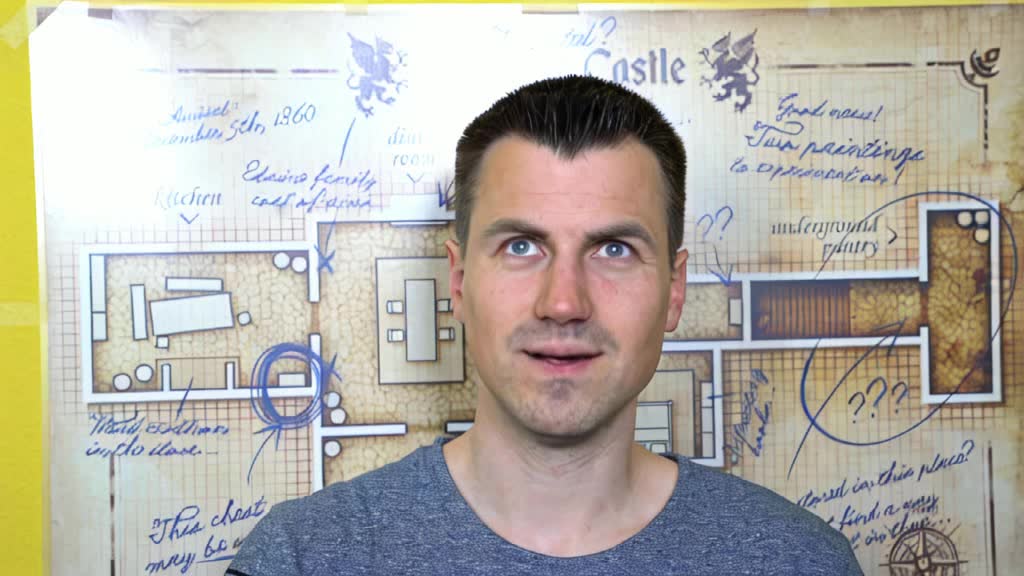}} \raisebox{-.5\height}{\includegraphics[width=0.09\textwidth]{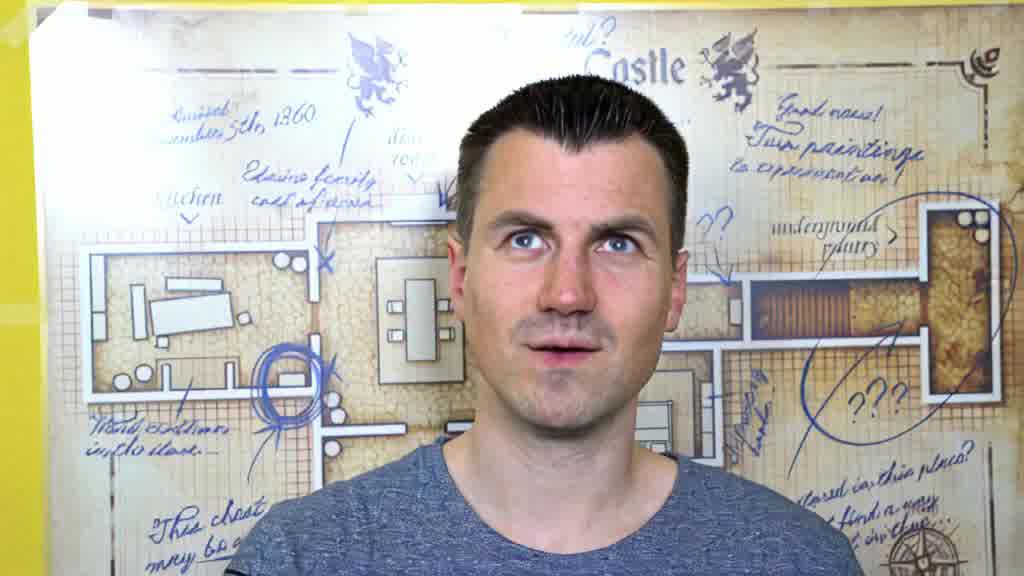}} \raisebox{-.5\height}{\includegraphics[width=0.09\textwidth]{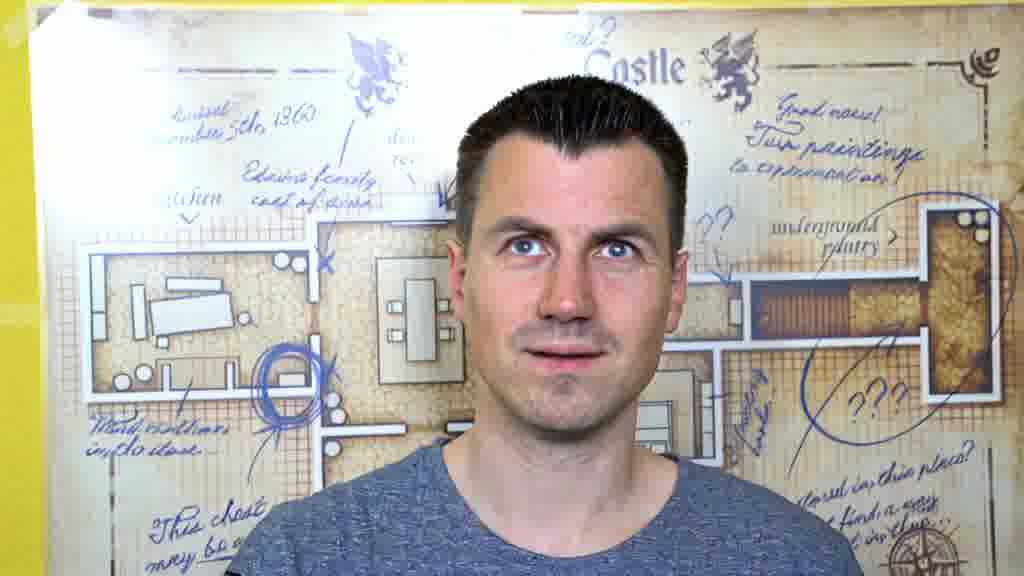}} \\
		$F'=15, N=1$ & \raisebox{-.5\height}{\includegraphics[width=0.09\textwidth]{figures/ablation_study_extra/animal_fourlegged/first_frame.jpg}} & \raisebox{-.5\height}{\includegraphics[width=0.09\textwidth]{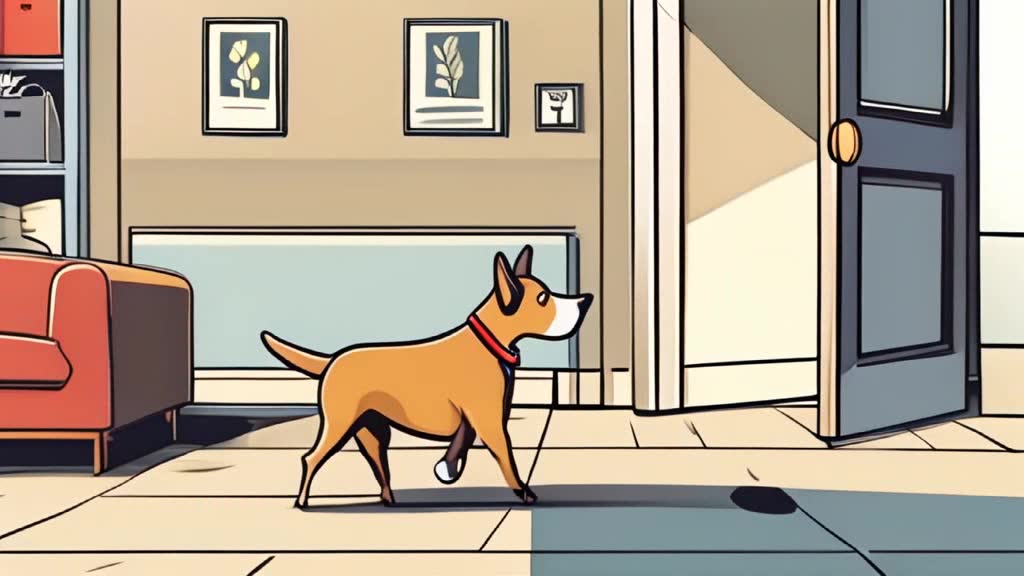}} \raisebox{-.5\height}{\includegraphics[width=0.09\textwidth]{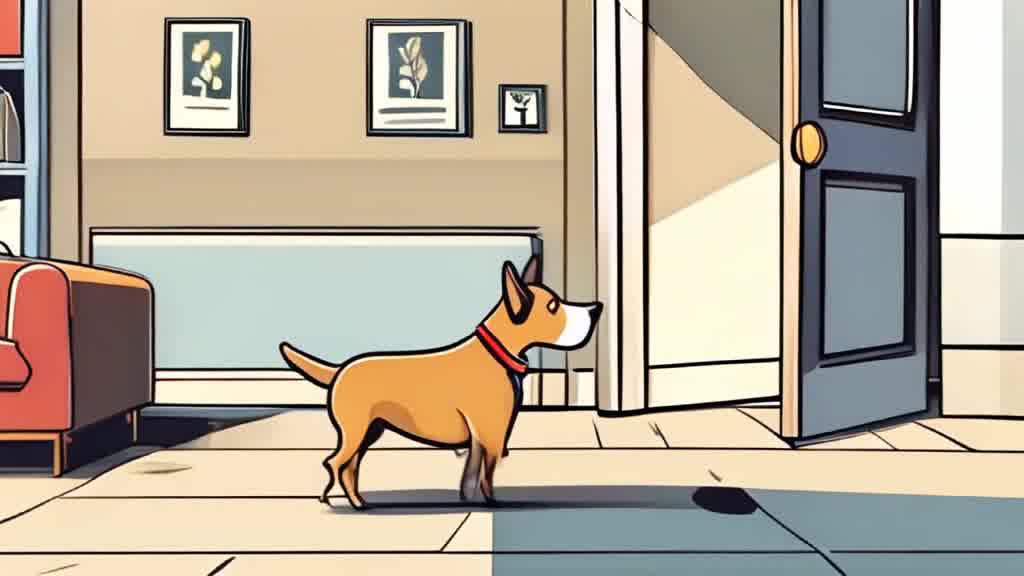}} \raisebox{-.5\height}{\includegraphics[width=0.09\textwidth]{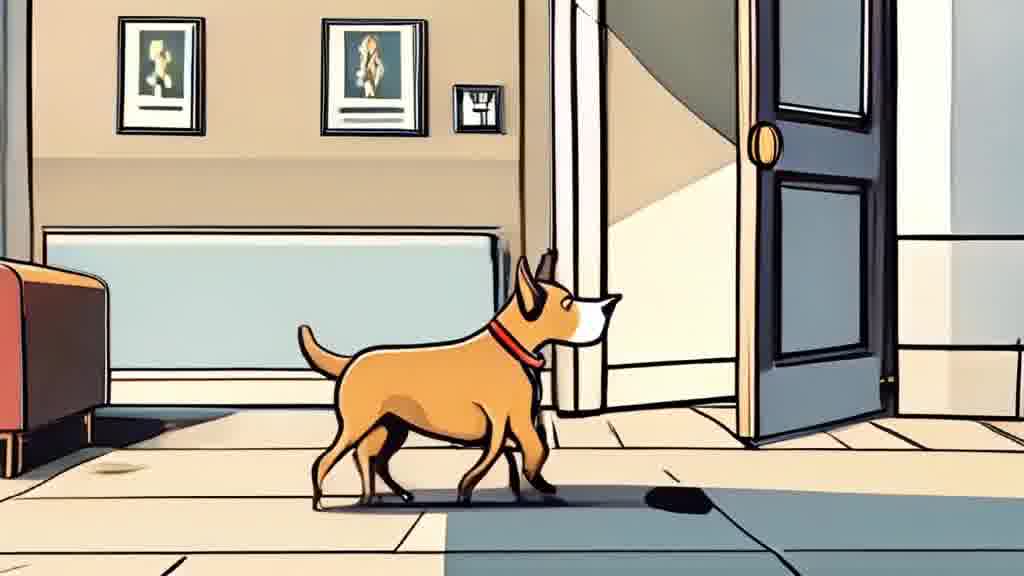}} & \raisebox{-.5\height}{\includegraphics[width=0.09\textwidth]{figures/ablation_study_extra/face_neutral/first_frame.jpg}} & \raisebox{-.5\height}{\includegraphics[width=0.09\textwidth]{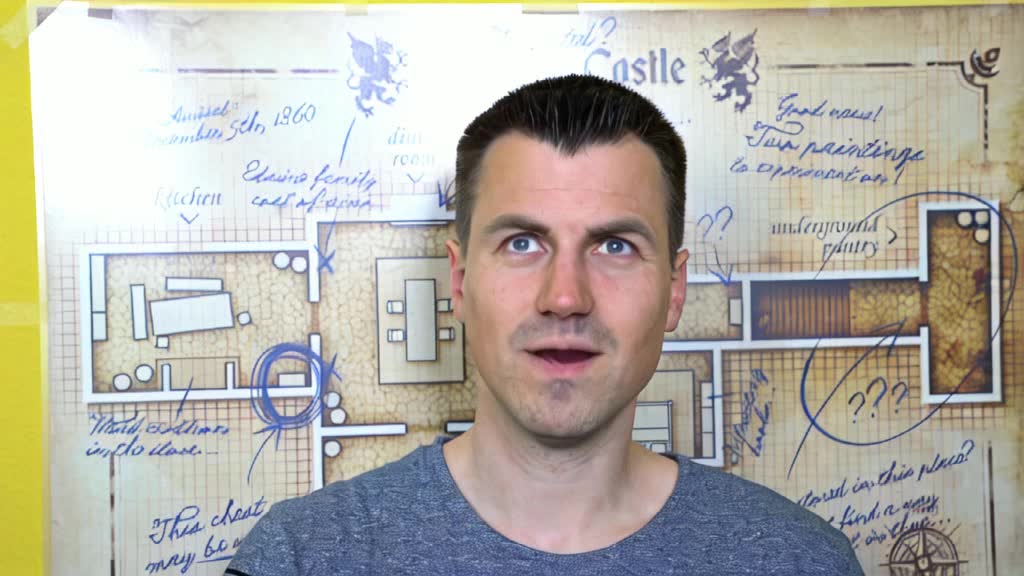}} \raisebox{-.5\height}{\includegraphics[width=0.09\textwidth]{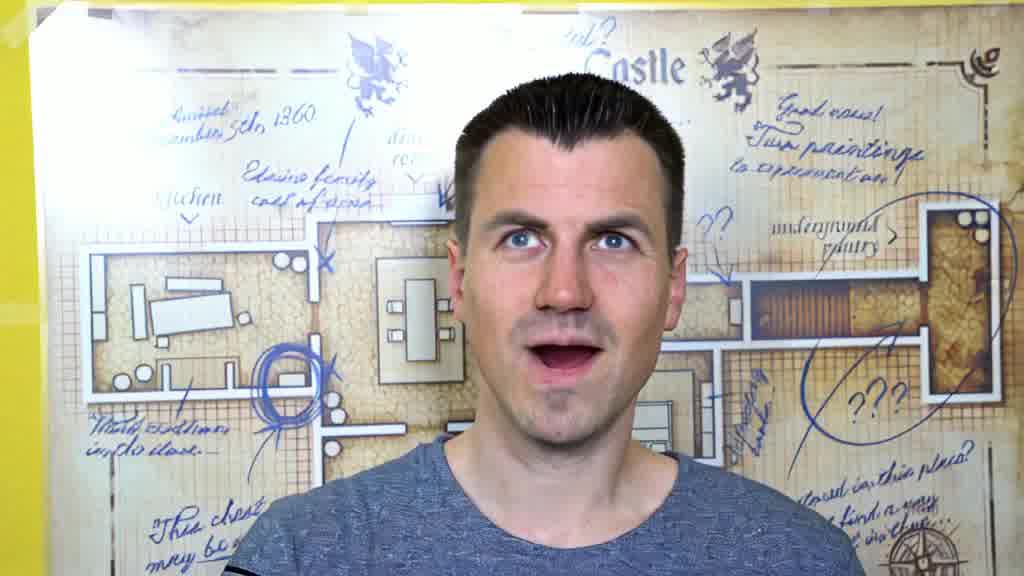}} \raisebox{-.5\height}{\includegraphics[width=0.09\textwidth]{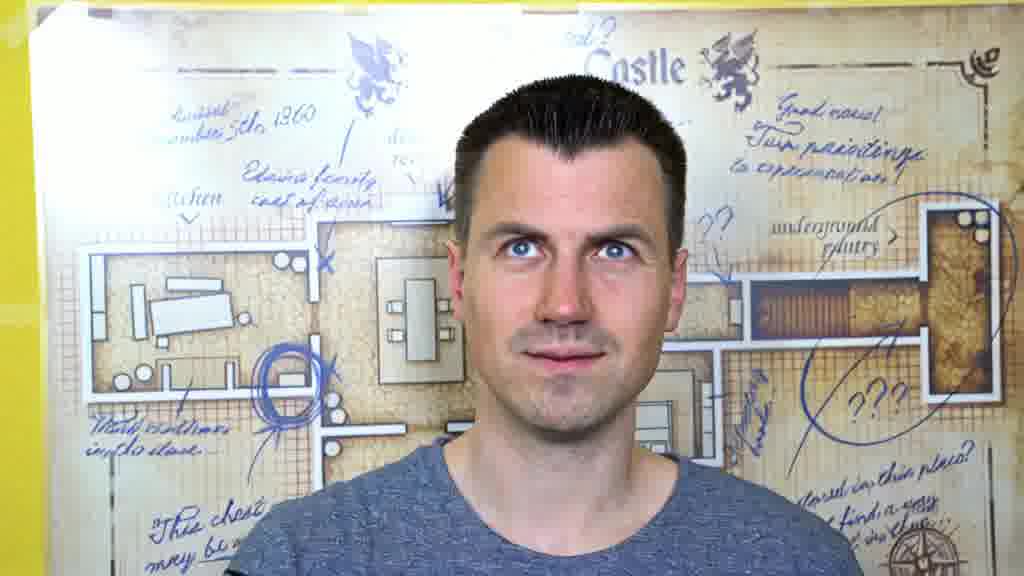}} \\
		{$F'=15, N=5$ \\ (Default)} & \raisebox{-.5\height}{\includegraphics[width=0.09\textwidth]{figures/ablation_study_extra/animal_fourlegged/first_frame.jpg}} & \raisebox{-.5\height}{\includegraphics[width=0.09\textwidth]{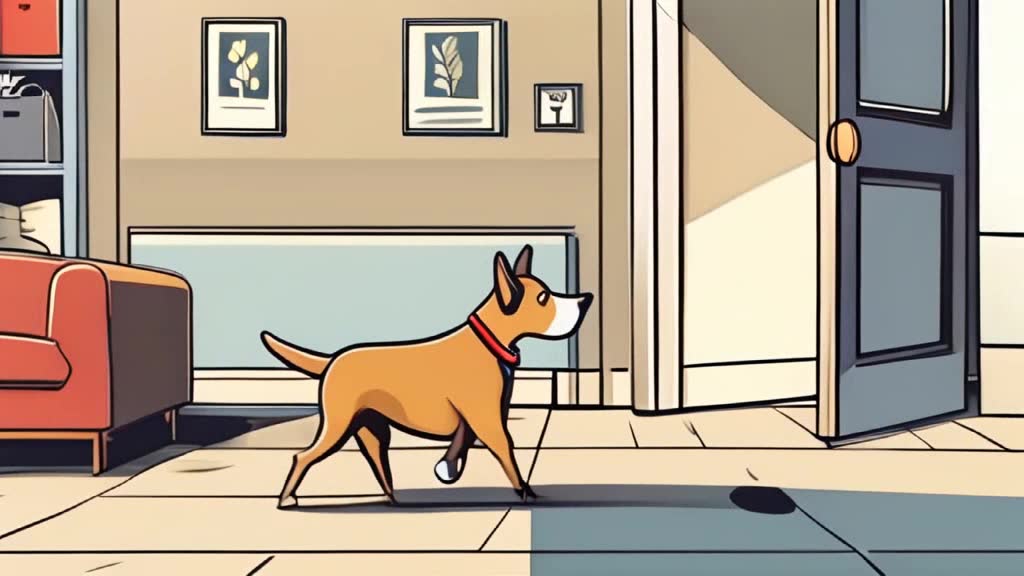}} \raisebox{-.5\height}{\includegraphics[width=0.09\textwidth]{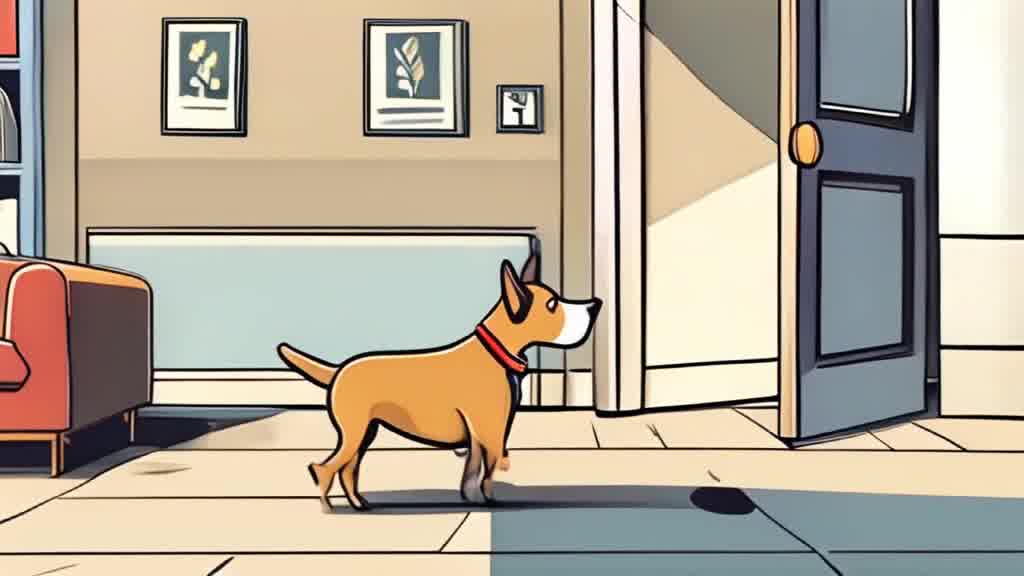}} \raisebox{-.5\height}{\includegraphics[width=0.09\textwidth]{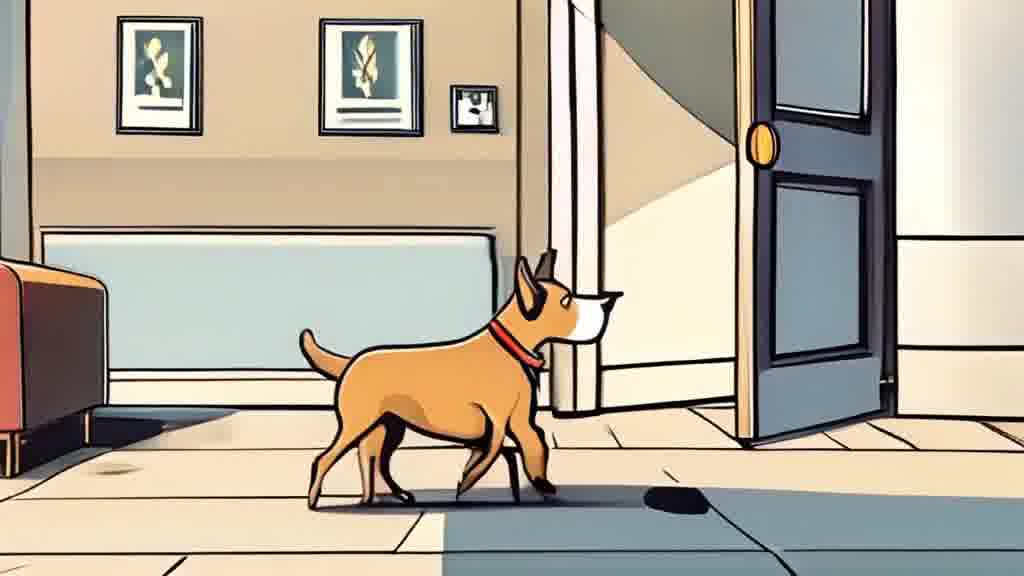}} & \raisebox{-.5\height}{\includegraphics[width=0.09\textwidth]{figures/ablation_study_extra/face_neutral/first_frame.jpg}} & \raisebox{-.5\height}{\includegraphics[width=0.09\textwidth]{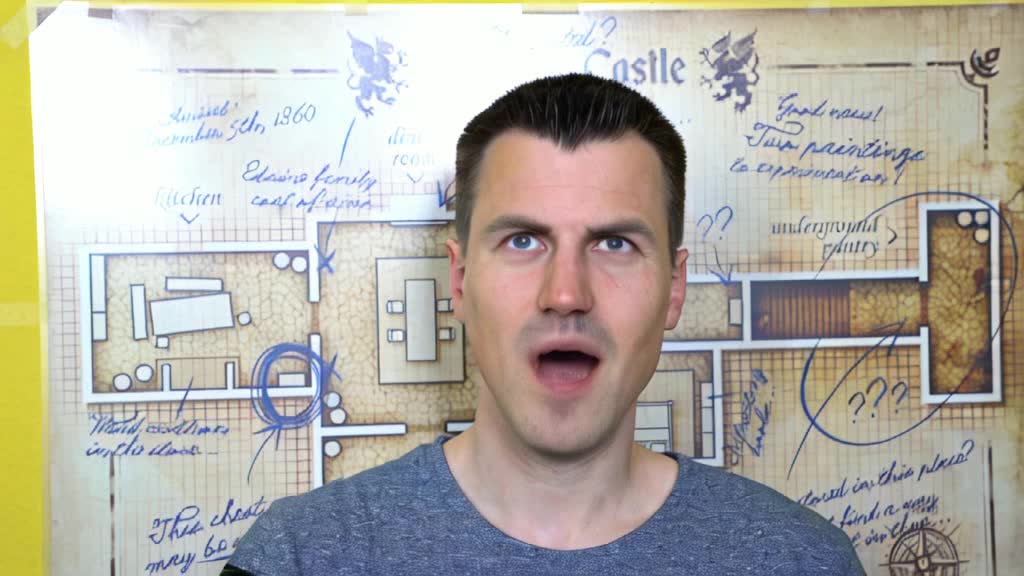}} \raisebox{-.5\height}{\includegraphics[width=0.09\textwidth]{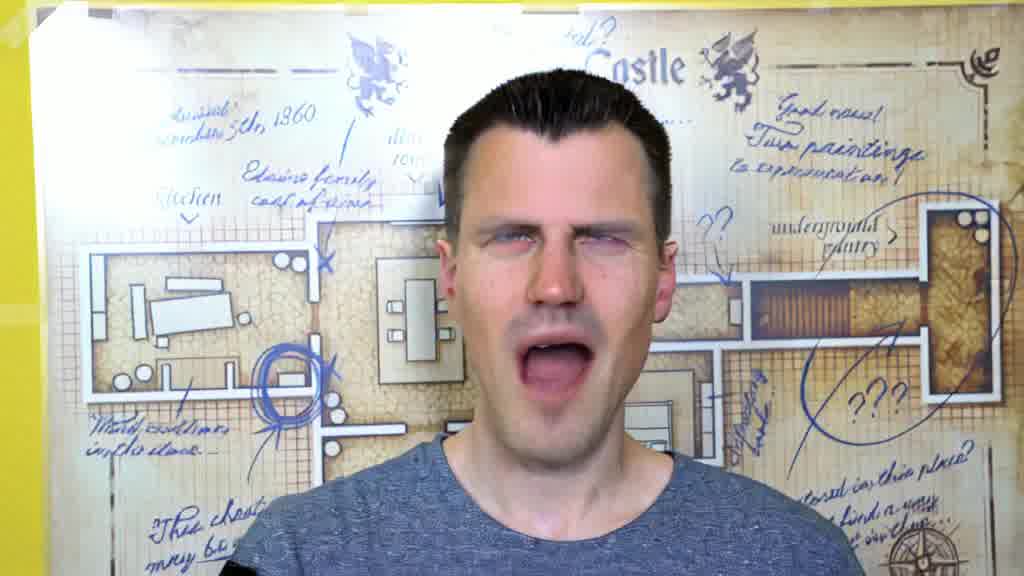}} \raisebox{-.5\height}{\includegraphics[width=0.09\textwidth]{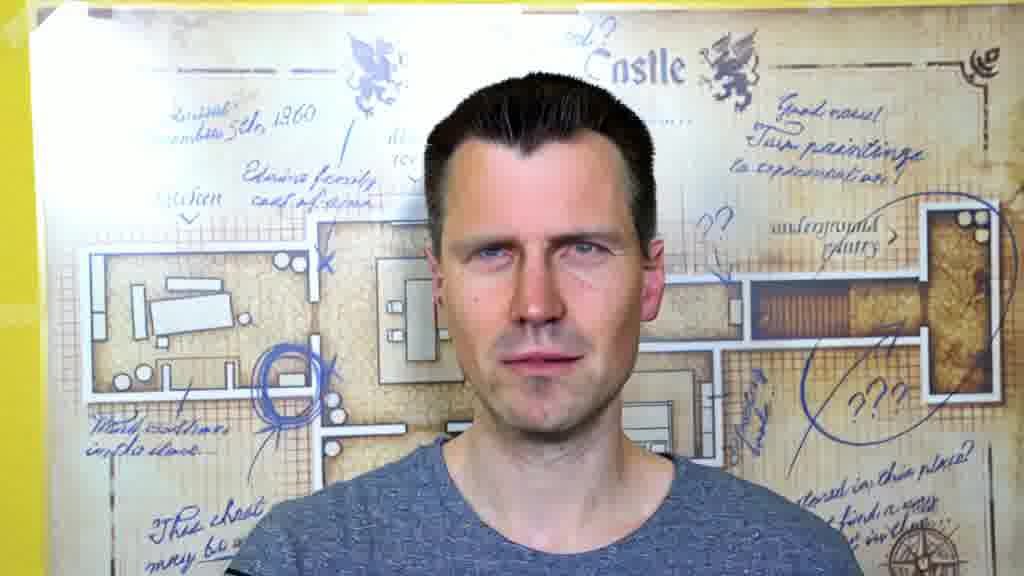}} \\
		$F'=15, N=15$ & \raisebox{-.5\height}{\includegraphics[width=0.09\textwidth]{figures/ablation_study_extra/animal_fourlegged/first_frame.jpg}} & \raisebox{-.5\height}{\includegraphics[width=0.09\textwidth]{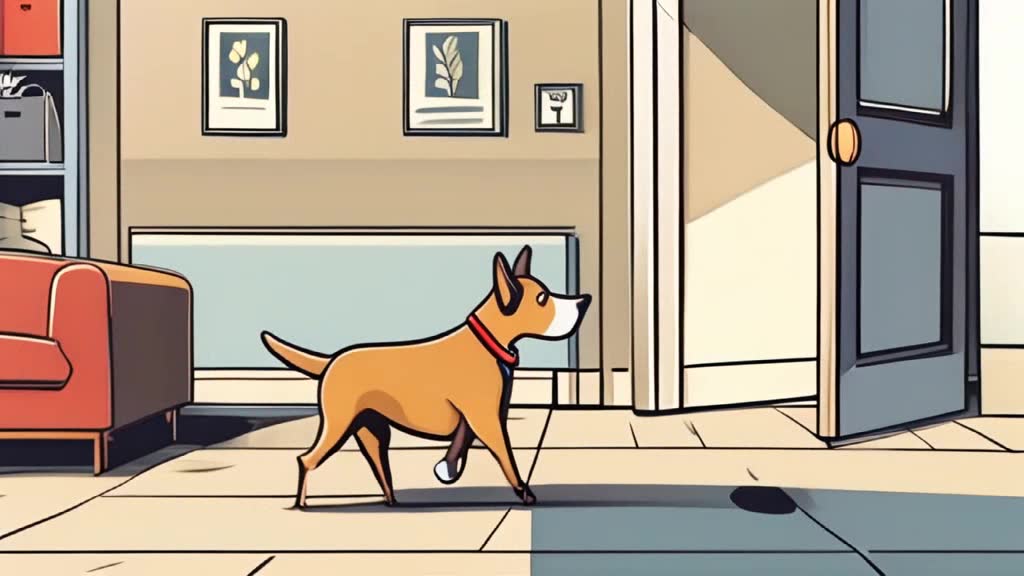}} \raisebox{-.5\height}{\includegraphics[width=0.09\textwidth]{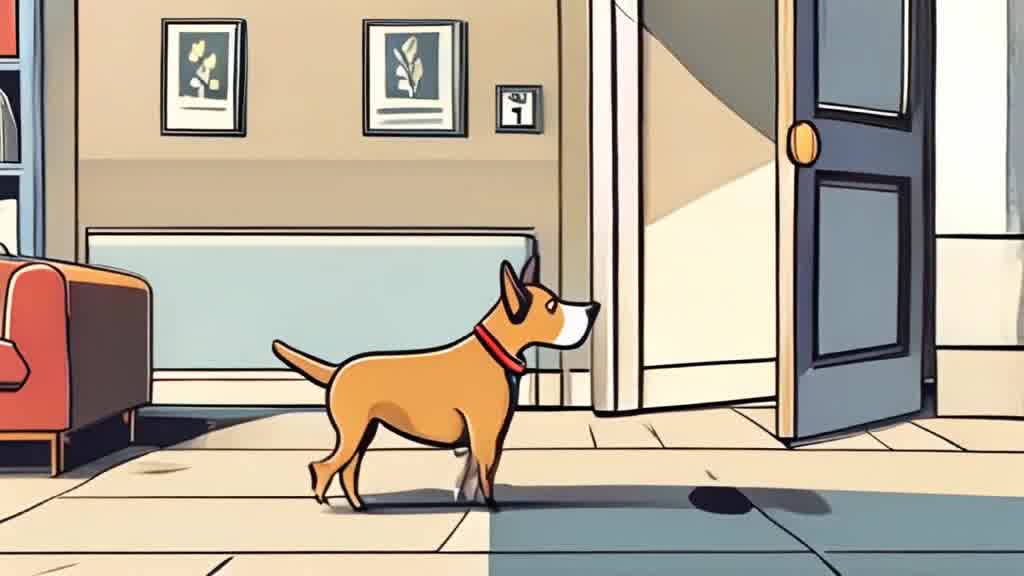}} \raisebox{-.5\height}{\includegraphics[width=0.09\textwidth]{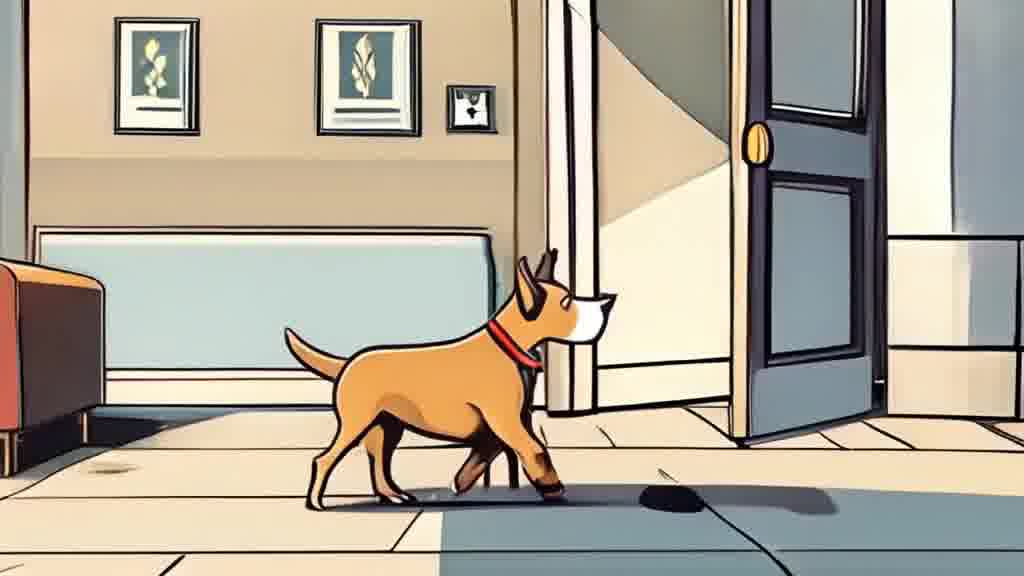}} & \raisebox{-.5\height}{\includegraphics[width=0.09\textwidth]{figures/ablation_study_extra/face_neutral/first_frame.jpg}} & \raisebox{-.5\height}{\includegraphics[width=0.09\textwidth]{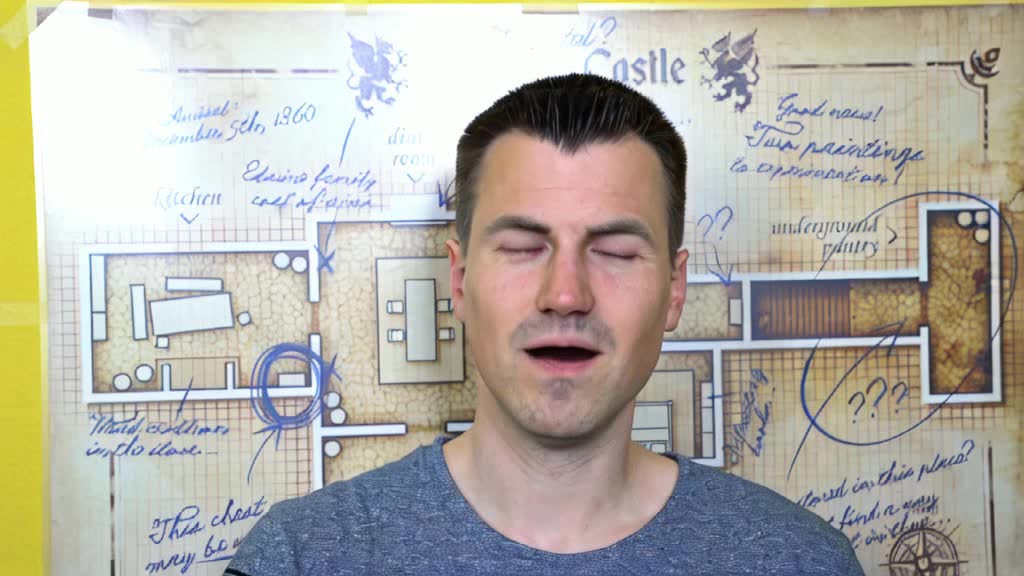}} \raisebox{-.5\height}{\includegraphics[width=0.09\textwidth]{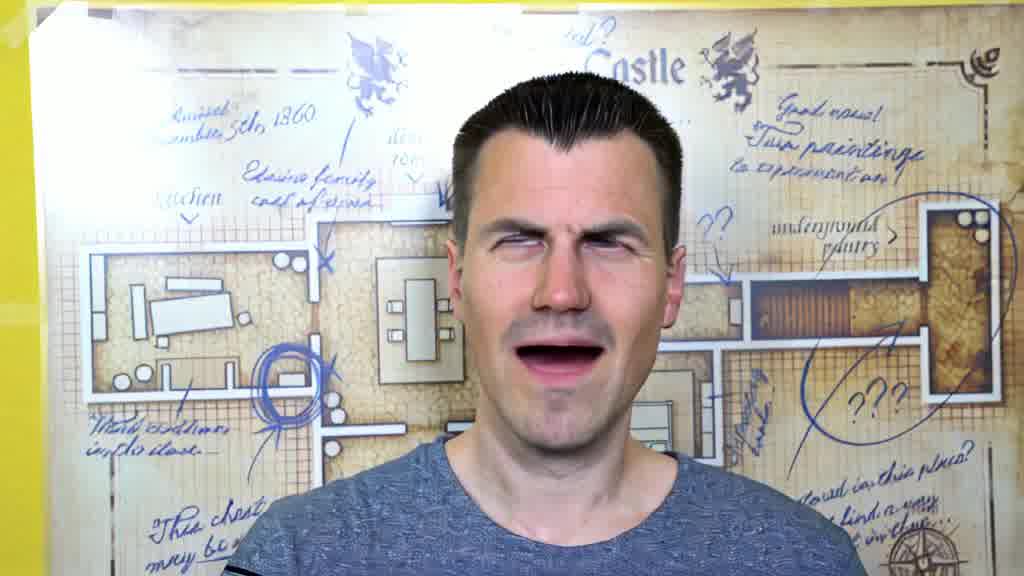}} \raisebox{-.5\height}{\includegraphics[width=0.09\textwidth]{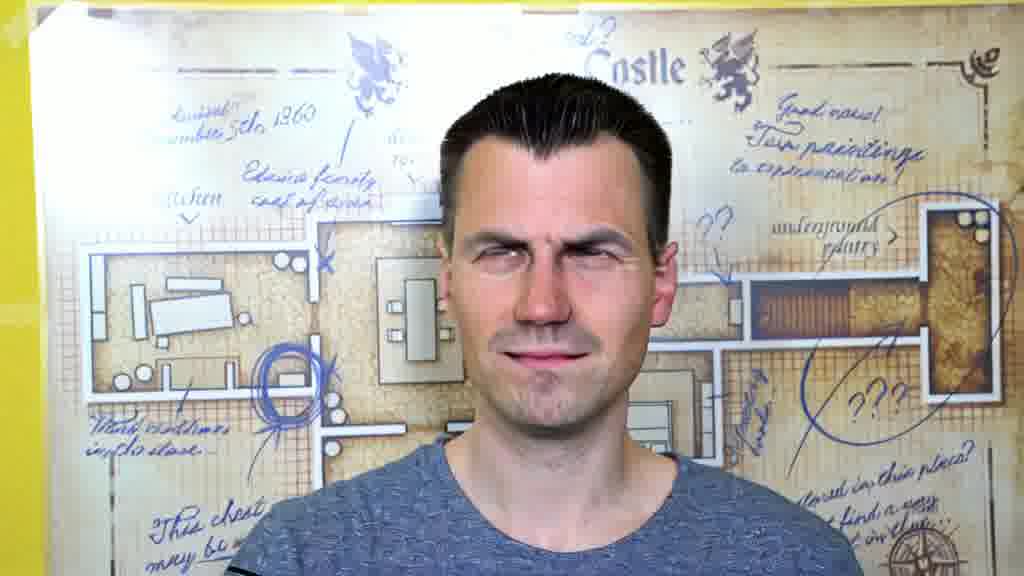}}
	\end{tblr}
	\caption{Ablation with additional examples. Inflating the motion-text embedding, by having more tokens $N$ or by having different tokens for each frame (where $F' = F+1 = 15$), greatly improves the motion transfer.}
	\Description{The first row shows frames of two motion reference videos: one of a horse walking and one of a person yawning. Below, each row shows one setting for the motion-text embedding. The lower the row, the more tokens are added, and the performance improves dramatically (especially once having different tokens per frame) but eventually saturates.}
	\label{fig:ablation_extra}
\end{figure*}

In Section~\ref{sec:ablation-study}, we show results for different settings of the motion-text embedding size for one motion. In Fig.~\ref{fig:ablation_extra}, we show two more examples for this ablation. As previously stated, the biggest performance improvement can be seen between rows 2 and 3 for each example, i.e., once there are \emph{different tokens per frame}. Note that the differences for the horse/dog example are best seen in the attached videos. While the dog is always moving to the right, the speed and style of the gait does not match the reference for the first two rows.

To quantitatively evaluate the settings of the motion-text embedding size, we followed the same protocol as for Table~\ref{table:quant_eval}. The results are listed in Table~\ref{table:quant_eval_ablation} and align well with our observations. Whereas the image appearance preservation is similar throughout, the motion fidelity improves slightly as we increase the token dimension $N$ (when $F'=1$) and significantly once we use \emph{different tokens per frame} ($F' = 15$). If $F' = 15$, the embedding dimension $N$ does not seem to affect the results much for the tested reference motion videos. In addition to the results aggregated over all evaluation videos in Table~\ref{table:quant_eval_ablation_overall}, we provide results aggregated by the motion category (camera/object) of the motion reference videos in Table~\ref{table:quant_eval_ablation_by_motion}. The results suggest that our proposed motion-text embedding inflation improves the performance for camera and object motions alike.

\begin{table*}[htbp]
	\centering
	\caption{Quantitative results for our ablation. Here, we compare various settings for the dimensions of the motion-text embedding. Table (a) shows the overall scores aggregated over all motion categories, whereas (b) shows the scores aggregated by the motion category of the motion reference videos, where the first value in each cell corresponds to camera motions and the second to object motions. The best performing method per column is marked in bold.}
	\label{table:quant_eval_ablation}
	\captionsetup{position=top} %
	\subfloat[a][Overall\label{table:quant_eval_ablation_overall}]{
		\small{
			\begin{tabular}{llcclccc}
				\toprule
				\multirow{2}{*}{Method} & & \multicolumn{2}{c}{Image Appearance Preservation} & & \multicolumn{3}{c}{Video Motion Fidelity} \\
				\cmidrule{3-4} \cmidrule{6-8}
				& & {CLIP-Avg $\uparrow$} & {CLIP-1st $\uparrow$} & & {Acc-Top-1 $\uparrow$} & {Acc-Top-5 $\uparrow$} & {Cos-Sim $\uparrow$} \\
				\midrule
				Ours ($F'=1, N=1$) & & \textbf{0.788} & 0.875 & & 44\% & 62\% & 0.619 \\
				Ours ($F'=1, N=15$) & & 0.785 & 0.878 & & 44\% & 65\% & 0.637 \\
				Ours ($F'=15, N=1$) & & 0.776 & 0.883 & & 52\% & \textbf{77\%} & 0.704 \\
				Ours ($F'=15, N=15$) & & 0.776 & \textbf{0.886} & & \textbf{56\%} & \textbf{77\%} & \textbf{0.705} \\
				\midrule
				Ours ($F'=15, N=5$, Default) & & 0.779 & 0.884 & & 54\% & 76\% & 0.696 \\
				\bottomrule
			\end{tabular}
		}
	}
	\\
	\subfloat[b][By motion category (camera/object)\label{table:quant_eval_ablation_by_motion}]{
		\small{
			\begin{tabular}{llcclccc}
				\toprule
				\multirow{2}{*}{Method} & & \multicolumn{2}{c}{Image Appearance Preservation} & & \multicolumn{3}{c}{Video Motion Fidelity} \\
				\cmidrule{3-4} \cmidrule{6-8}
				& & {CLIP-Avg $\uparrow$} & {CLIP-1st $\uparrow$} & & {Acc-Top-1 $\uparrow$} & {Acc-Top-5 $\uparrow$} & {Cos-Sim $\uparrow$} \\
				\midrule
				Ours ($F'=1, N=1$) & & \textbf{0.755}/\textbf{0.821} & 0.865/0.885 & & 64\%/24\% & 76\%/48\% & 0.722/0.516 \\
				Ours ($F'=1, N=15$) & & 0.754/0.817 & \textbf{0.874}/0.881 & & 70\%/18\% & 82\%/48\% & 0.758/0.517 \\
				Ours ($F'=15, N=1$) & & 0.743/0.810 & 0.872/0.894 & & 74\%/30\% & \textbf{86\%}/\textbf{68\%} & 0.807/0.600 \\
				Ours ($F'=15, N=15$) & & 0.740/0.813 & \textbf{0.874}/\textbf{0.899} & & \textbf{78\%}/34\% & \textbf{86\%}/\textbf{68\%} & \textbf{0.810}/0.601 \\
				\midrule
				Ours ($F'=15, N=5$, Default) & & 0.745/0.813 & 0.873/0.894 & & 72\%/\textbf{36\%} & \textbf{86\%}/66\% & 0.785/\textbf{0.606} \\
				\bottomrule
			\end{tabular}
		}
	}
\end{table*}

\clearpage

\section{Additional Results} \label{sec:additional-results}

\begin{figure*}[htbp]
	\centering
	\begin{tblr}{
			vline{3} = {2-5}{dashed},
			vline{4} = {1-5}{},
			vline{5} = {2-5}{dashed},
		}
		{Ref.} & \raisebox{-.5\height}{\includegraphics[width=0.09\textwidth]{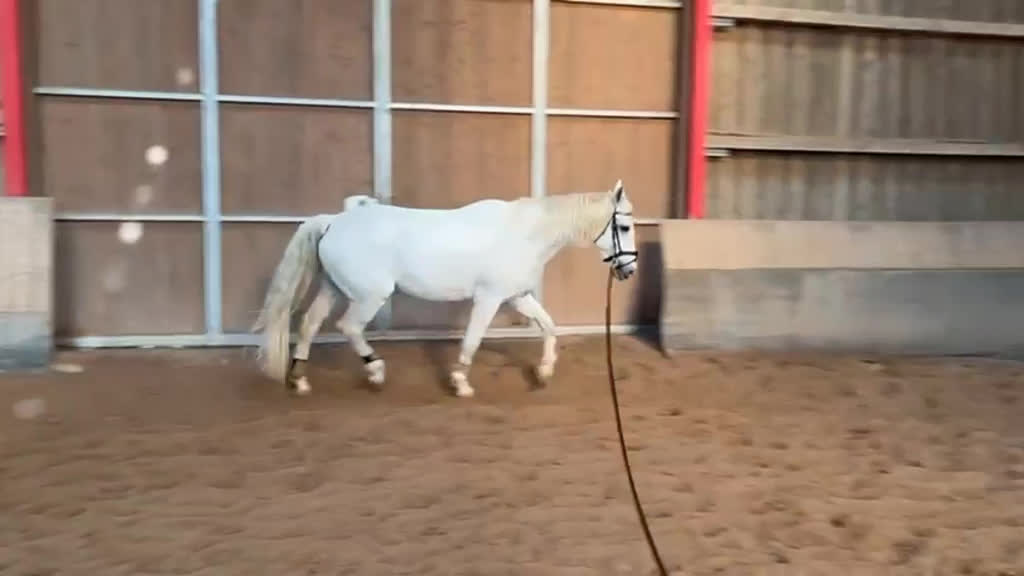}} & \raisebox{-.5\height}{\includegraphics[width=0.09\textwidth]{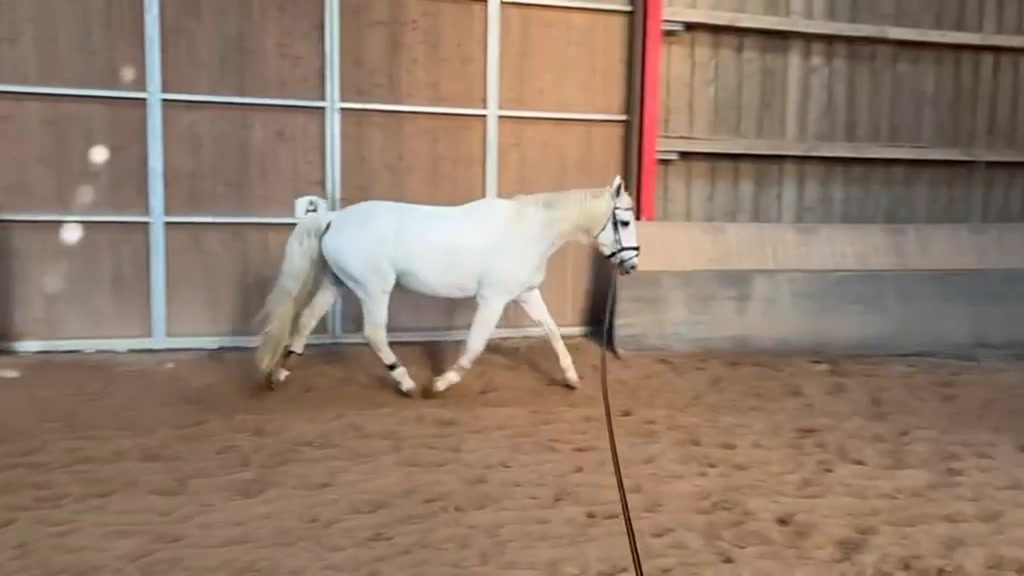}} \raisebox{-.5\height}{\includegraphics[width=0.09\textwidth]{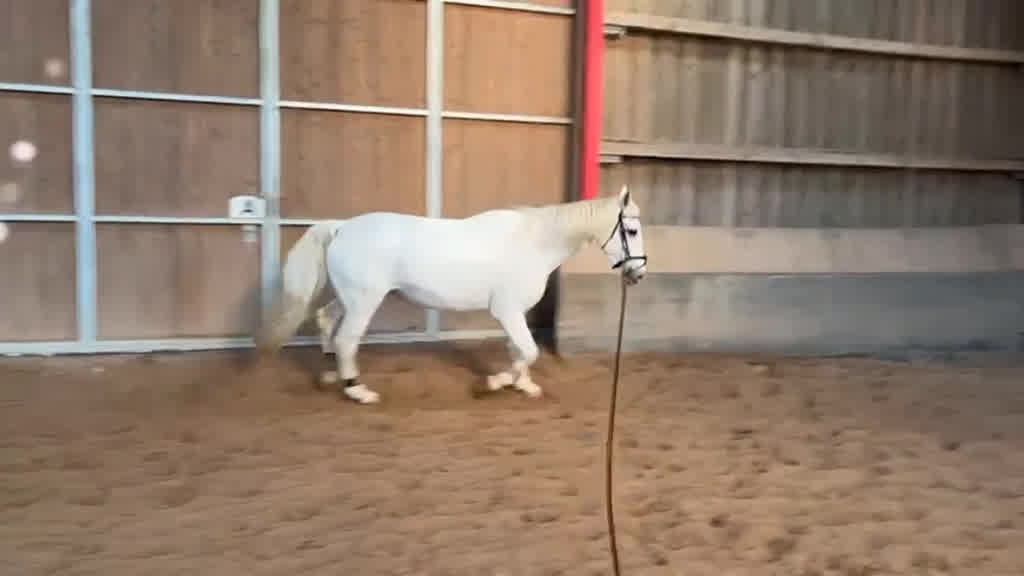}} \raisebox{-.5\height}{\includegraphics[width=0.09\textwidth]{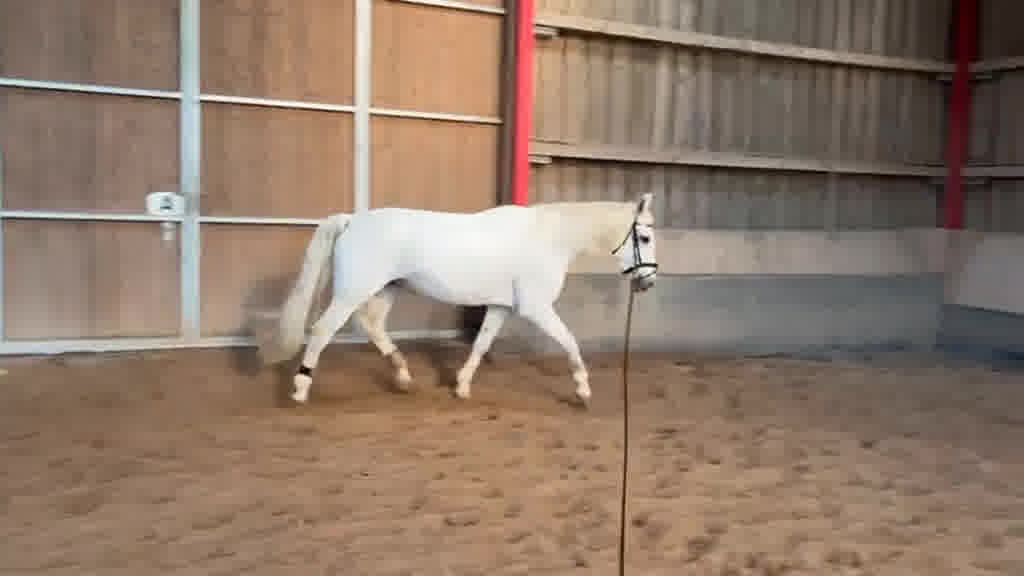}} & \raisebox{-.5\height}{\includegraphics[width=0.09\textwidth]{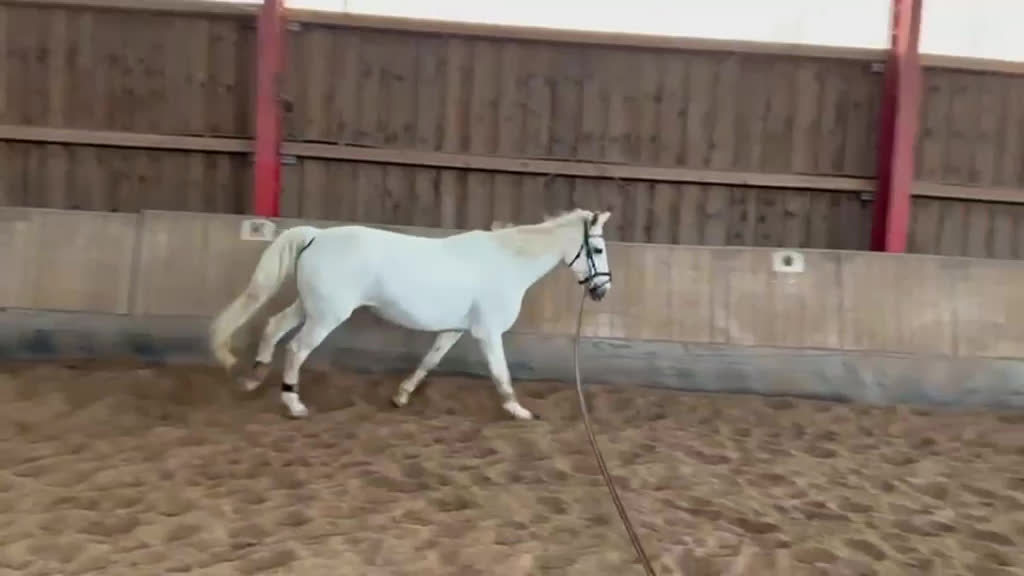}} & \raisebox{-.5\height}{\includegraphics[width=0.09\textwidth]{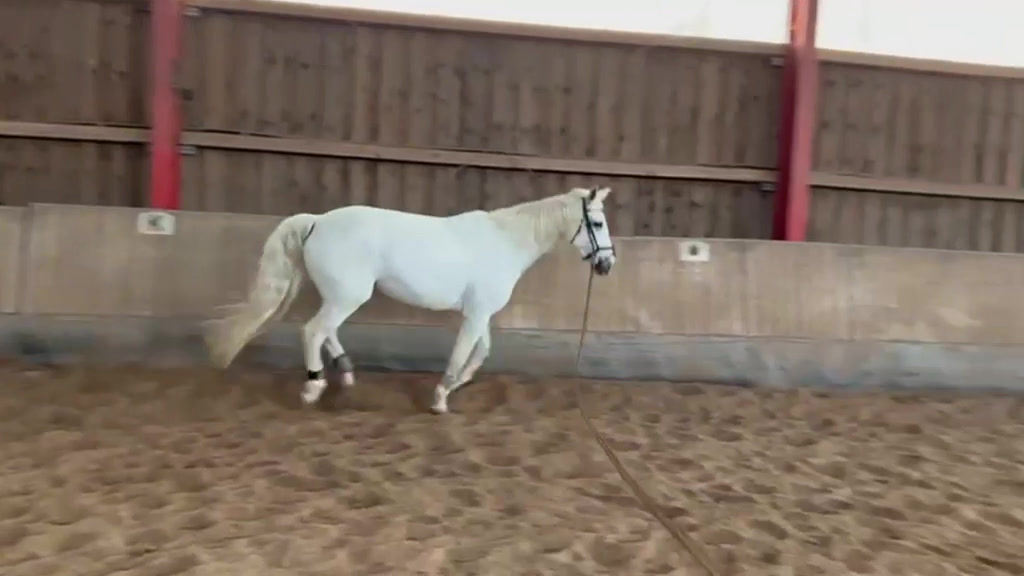}} \raisebox{-.5\height}{\includegraphics[width=0.09\textwidth]{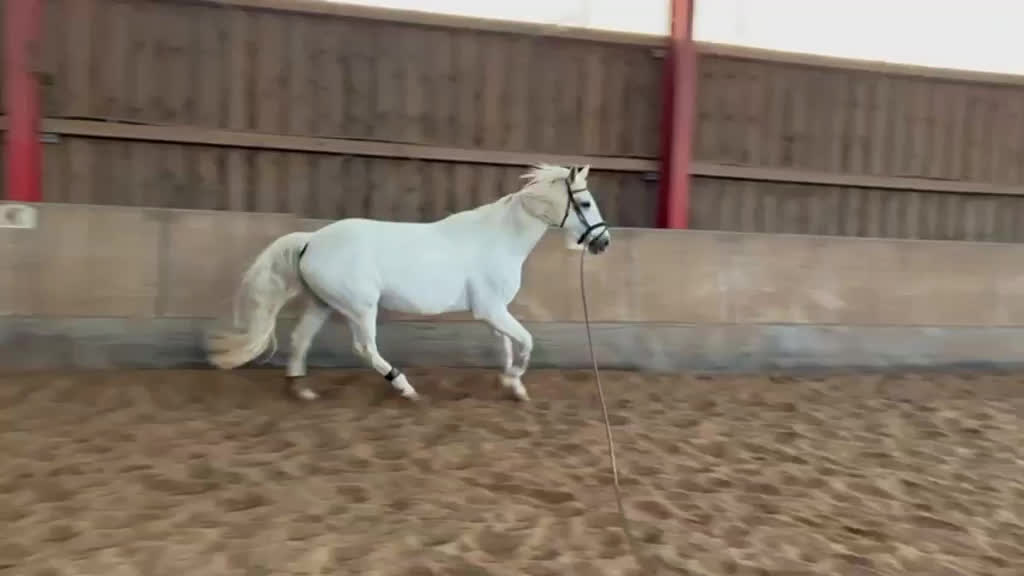}} \raisebox{-.5\height}{\includegraphics[width=0.09\textwidth]{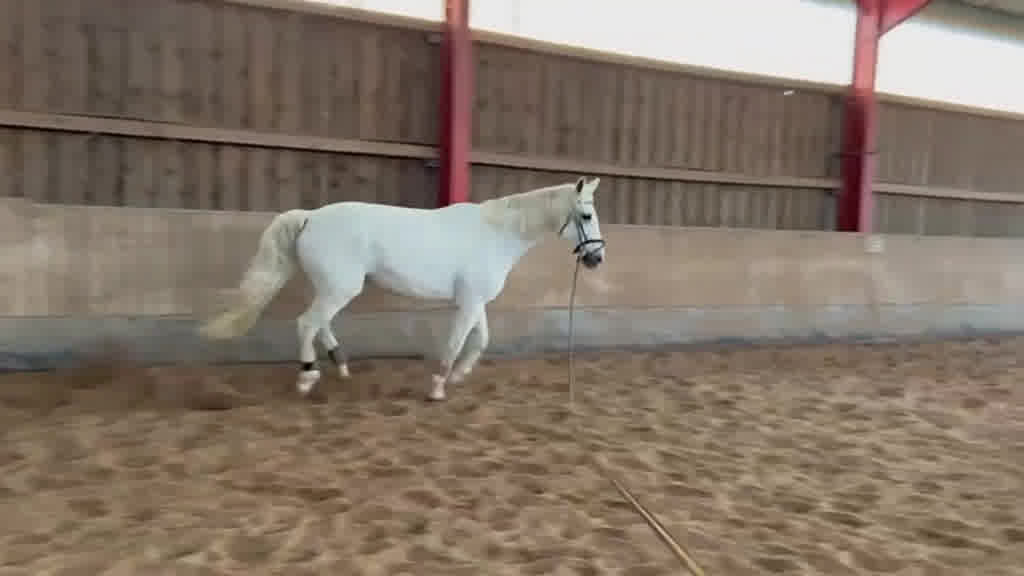}} \\
		{Gen. 1}  & \raisebox{-.5\height}{\includegraphics[width=0.09\textwidth]{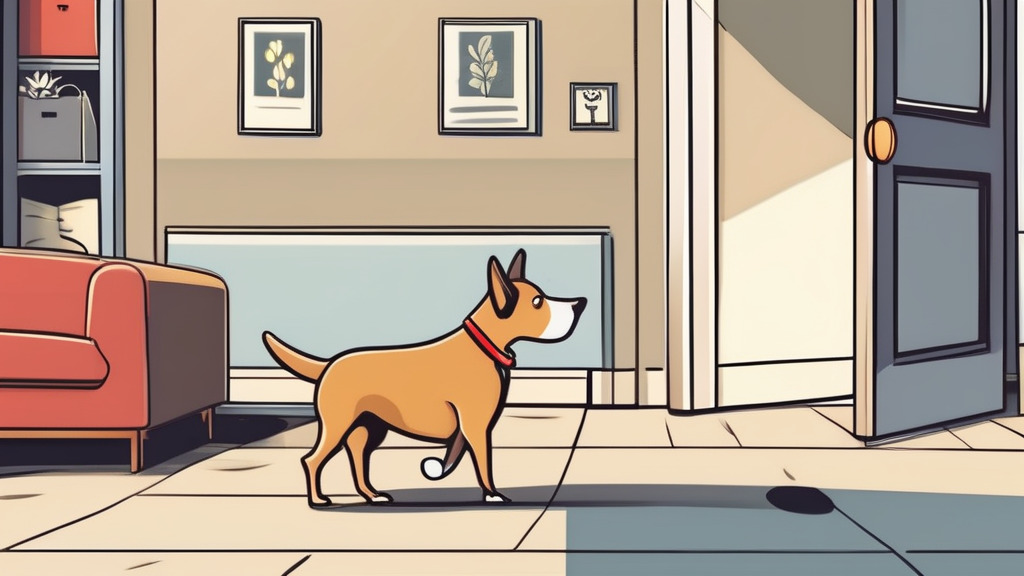}} & \raisebox{-.5\height}{\includegraphics[width=0.09\textwidth]{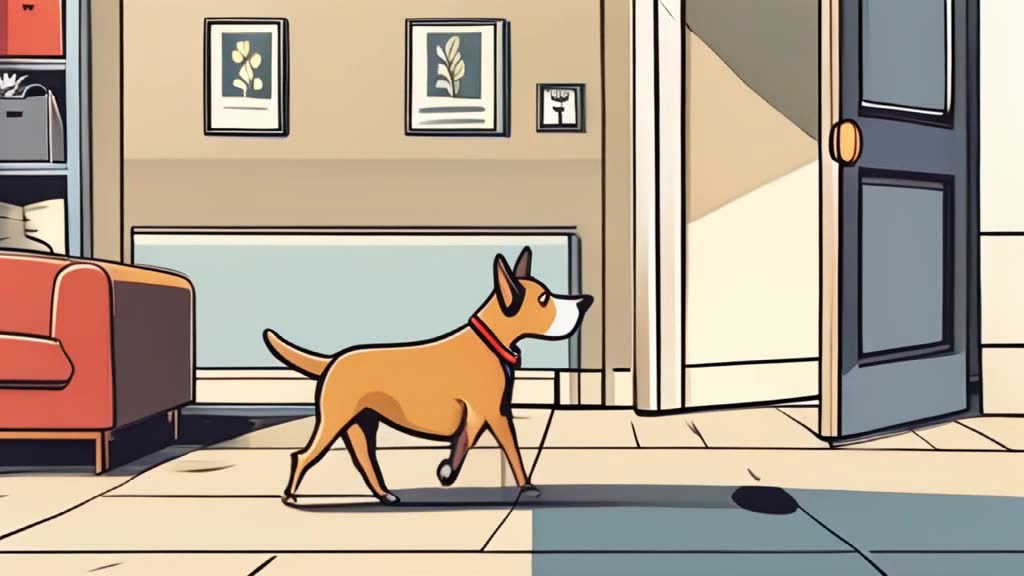}} \raisebox{-.5\height}{\includegraphics[width=0.09\textwidth]{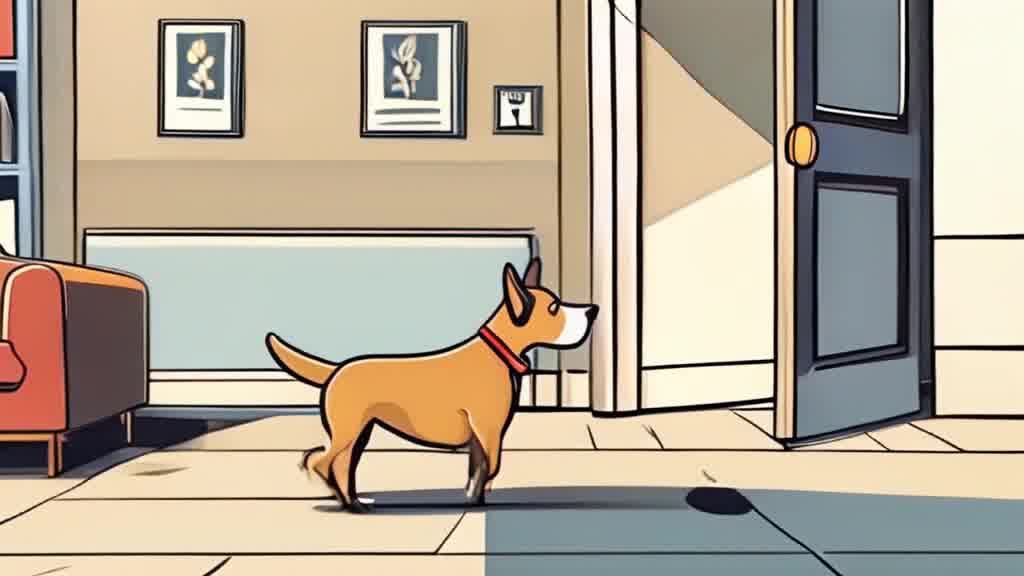}} \raisebox{-.5\height}{\includegraphics[width=0.09\textwidth]{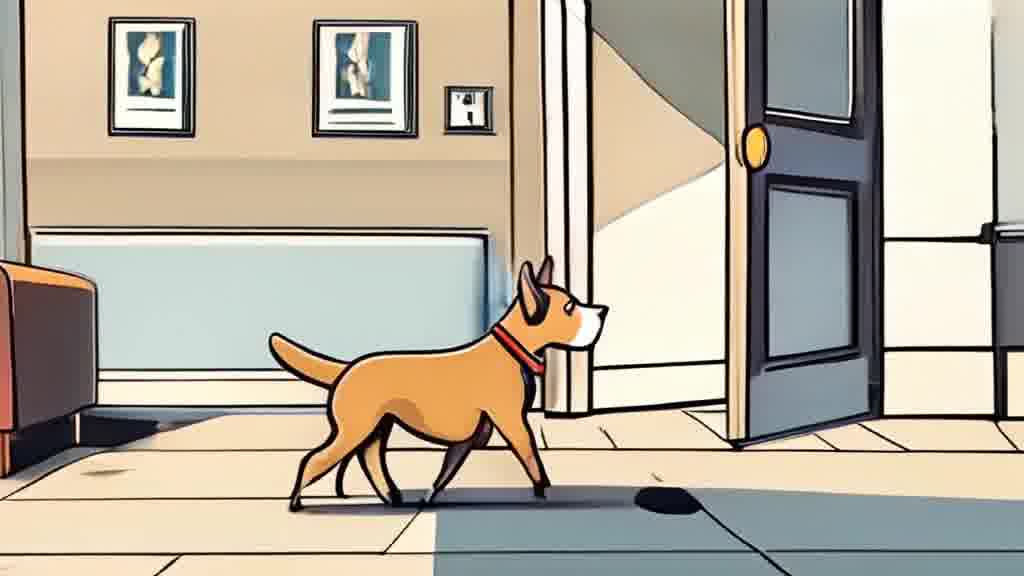}} & \raisebox{-.5\height}{\includegraphics[width=0.09\textwidth]{figures/style/trot/first_frame.jpg}} & \raisebox{-.5\height}{\includegraphics[width=0.09\textwidth]{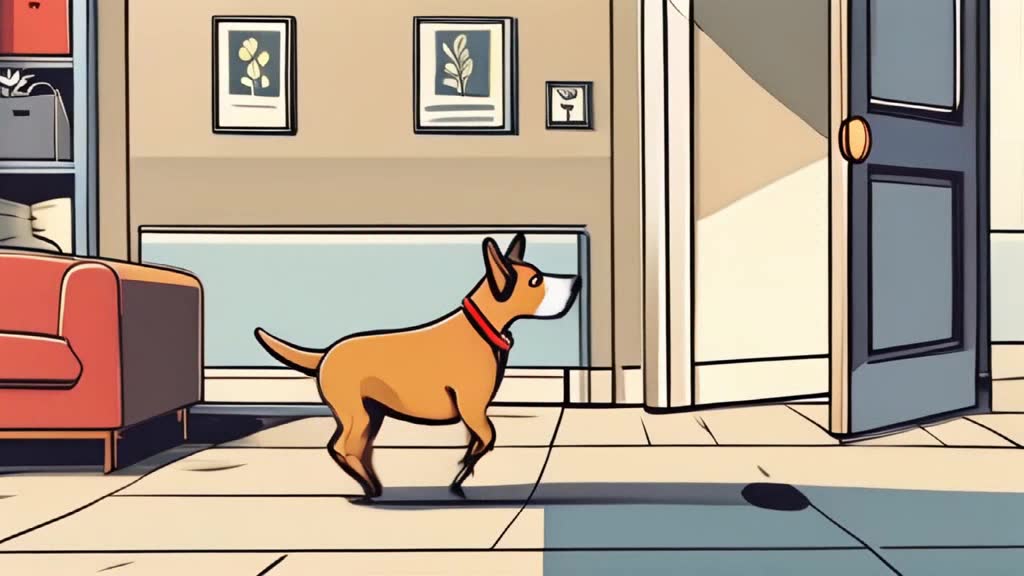}} \raisebox{-.5\height}{\includegraphics[width=0.09\textwidth]{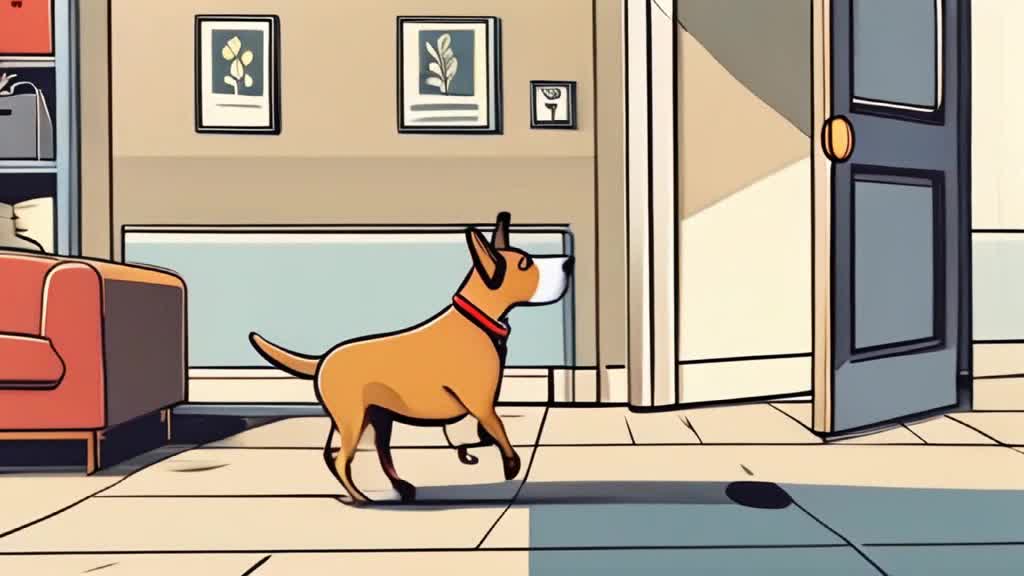}} \raisebox{-.5\height}{\includegraphics[width=0.09\textwidth]{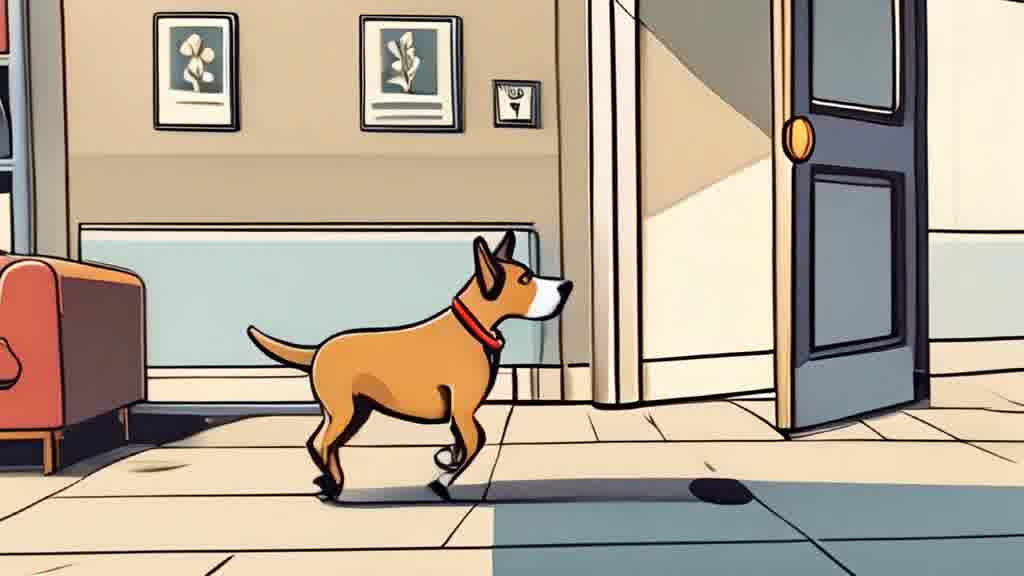}} \\
		{Gen. 2}  & \raisebox{-.5\height}{\includegraphics[width=0.09\textwidth]{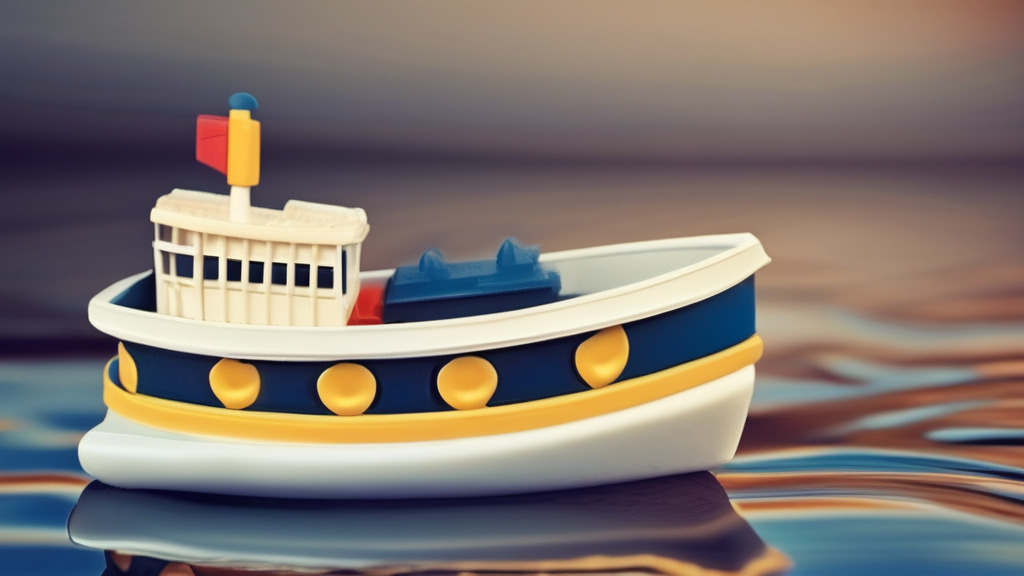}} & \raisebox{-.5\height}{\includegraphics[width=0.09\textwidth]{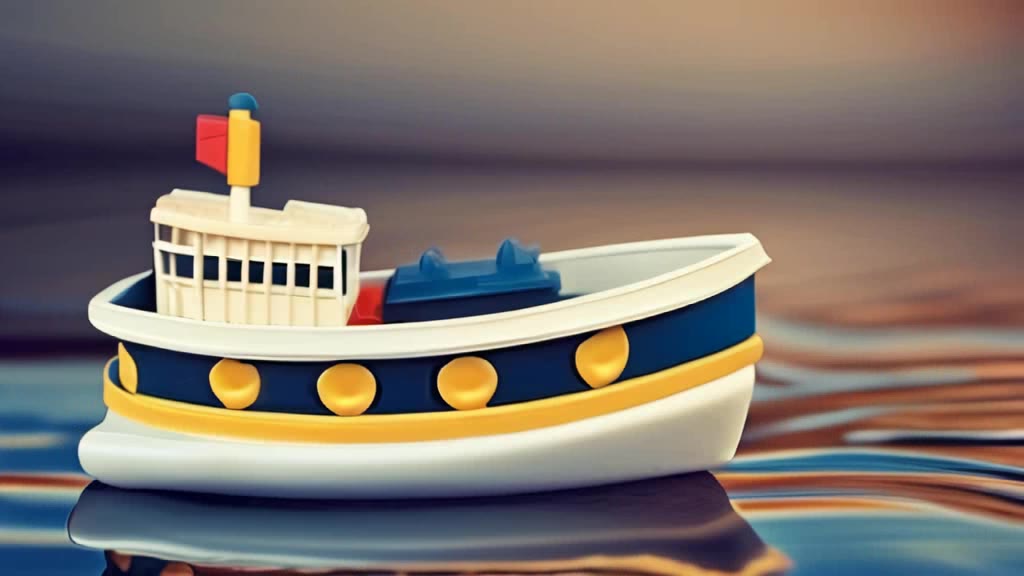}} \raisebox{-.5\height}{\includegraphics[width=0.09\textwidth]{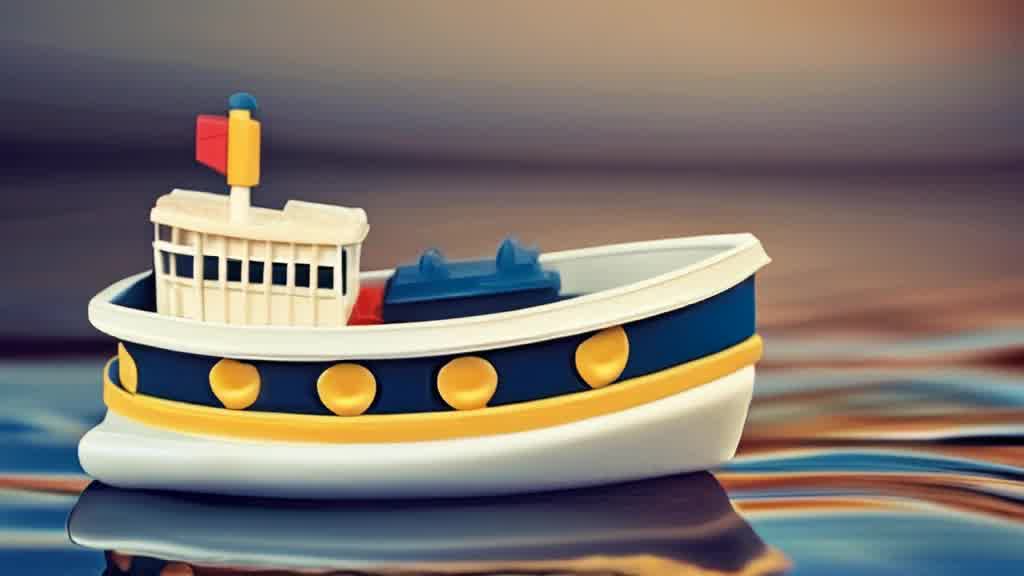}} \raisebox{-.5\height}{\includegraphics[width=0.09\textwidth]{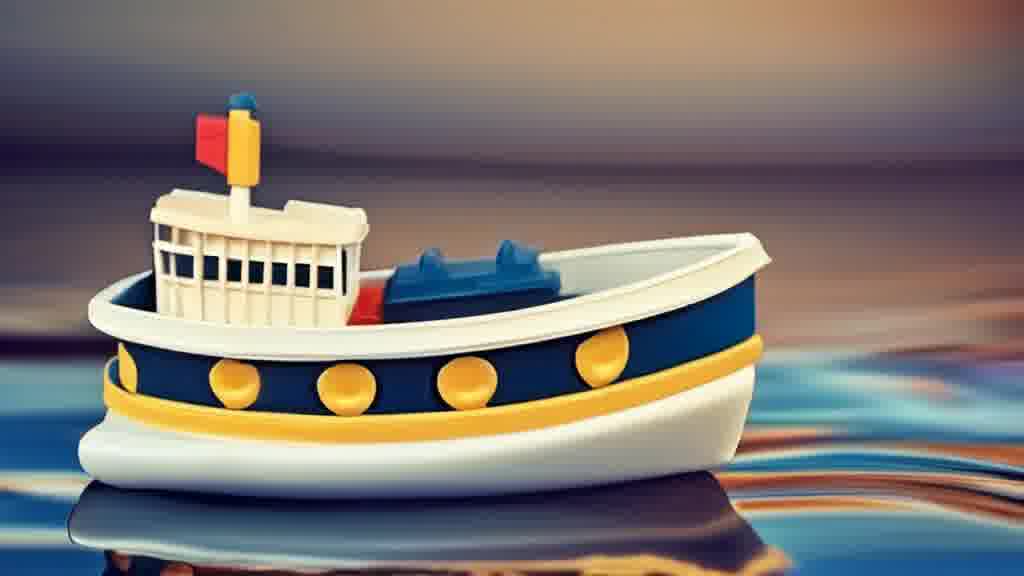}} & \raisebox{-.5\height}{\includegraphics[width=0.09\textwidth]{figures/style/trot/first_frame_2.jpg}} & \raisebox{-.5\height}{\includegraphics[width=0.09\textwidth]{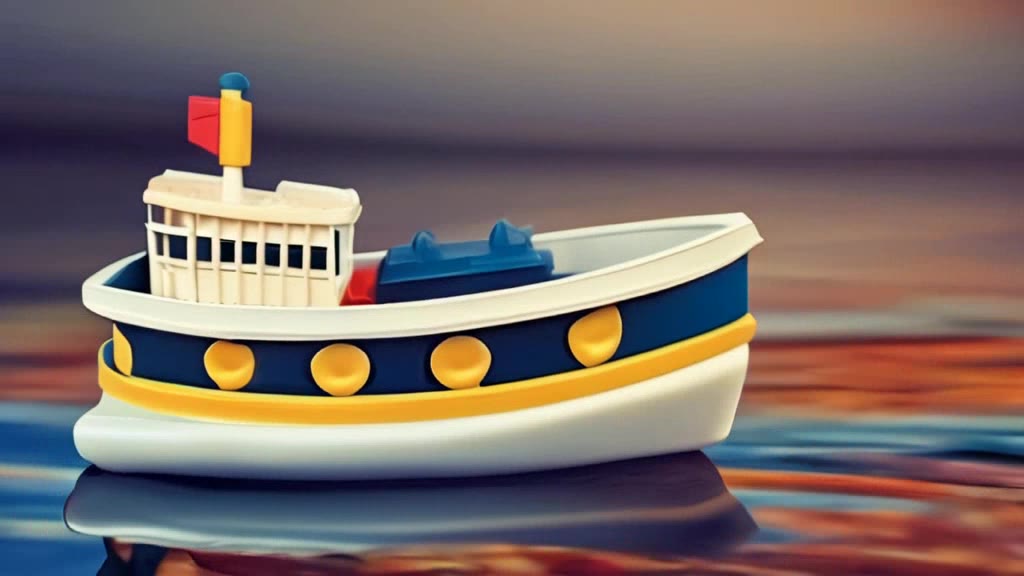}} \raisebox{-.5\height}{\includegraphics[width=0.09\textwidth]{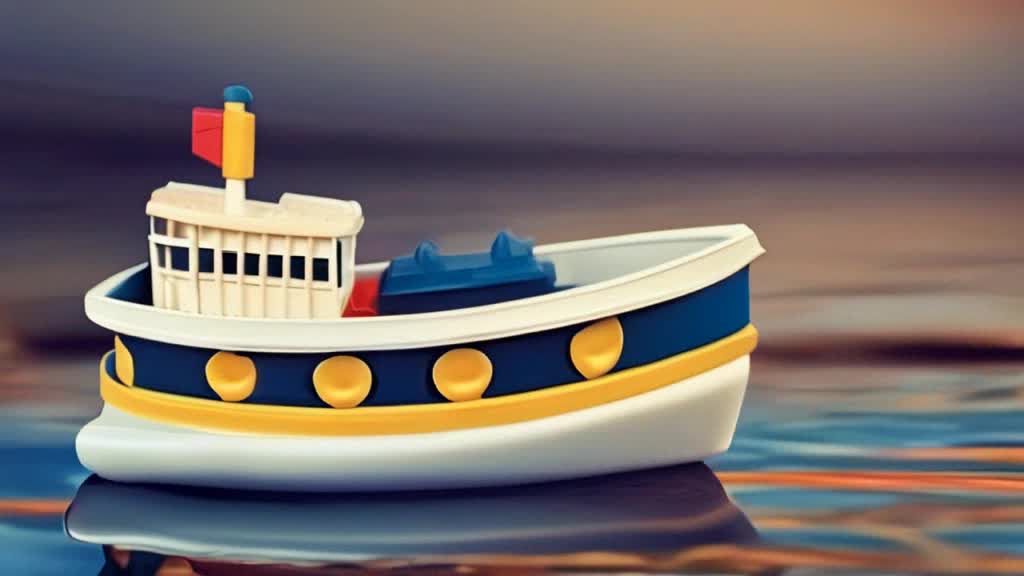}} \raisebox{-.5\height}{\includegraphics[width=0.09\textwidth]{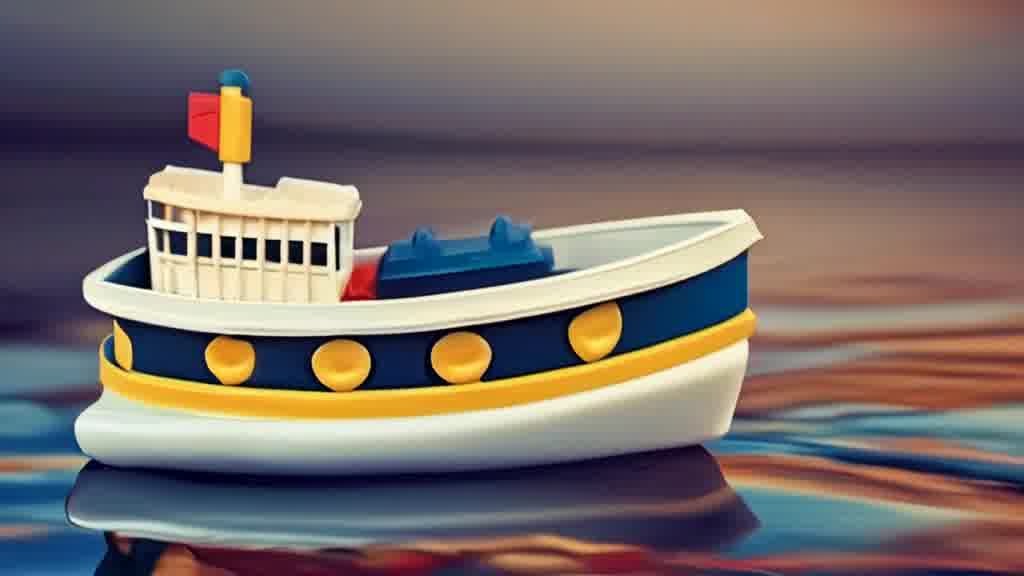}} \\
		{Gen. 3}  & \raisebox{-.5\height}{\includegraphics[width=0.09\textwidth]{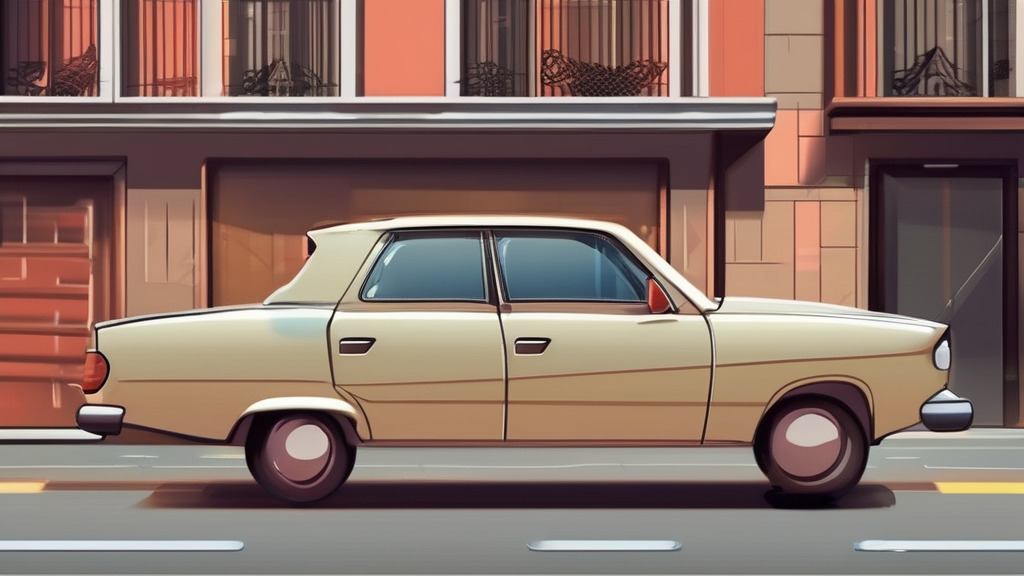}} & \raisebox{-.5\height}{\includegraphics[width=0.09\textwidth]{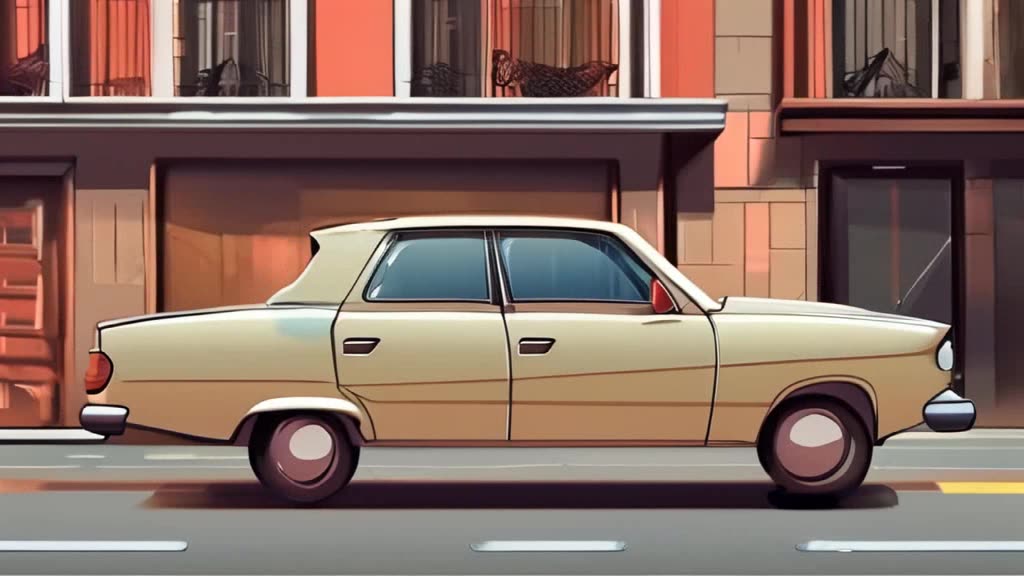}} \raisebox{-.5\height}{\includegraphics[width=0.09\textwidth]{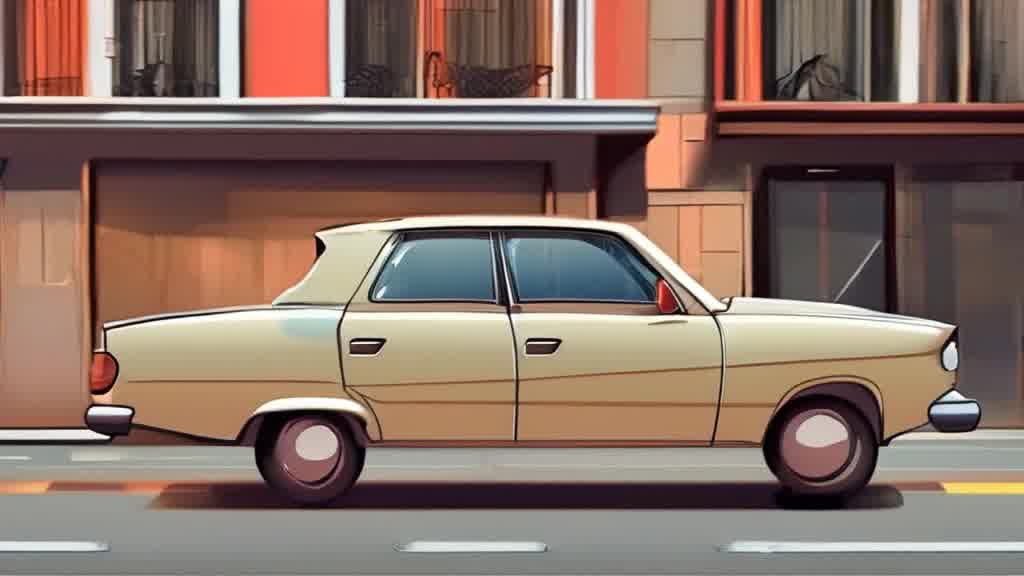}} \raisebox{-.5\height}{\includegraphics[width=0.09\textwidth]{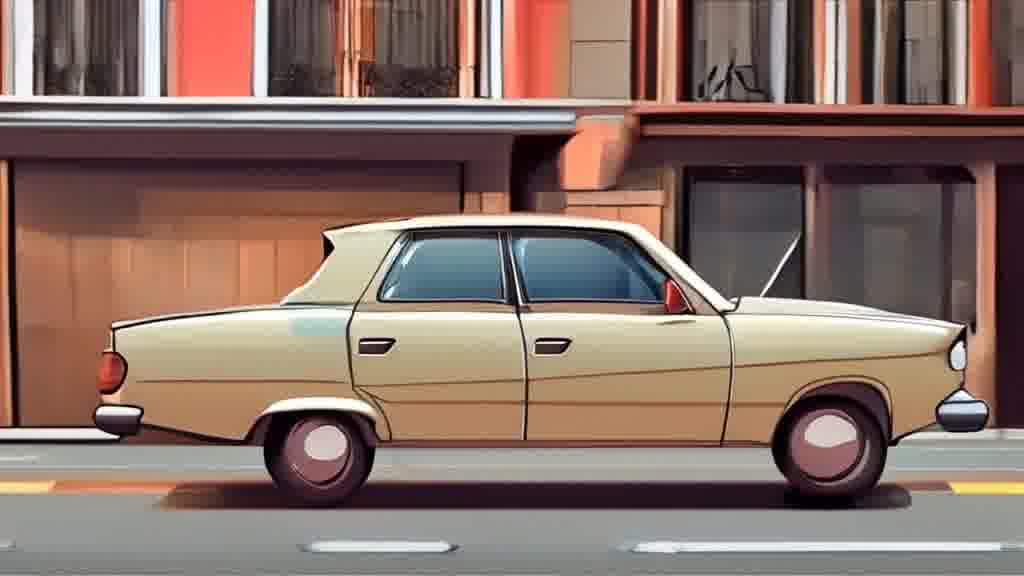}} & \raisebox{-.5\height}{\includegraphics[width=0.09\textwidth]{figures/style/trot/first_frame_3.jpg}} & \raisebox{-.5\height}{\includegraphics[width=0.09\textwidth]{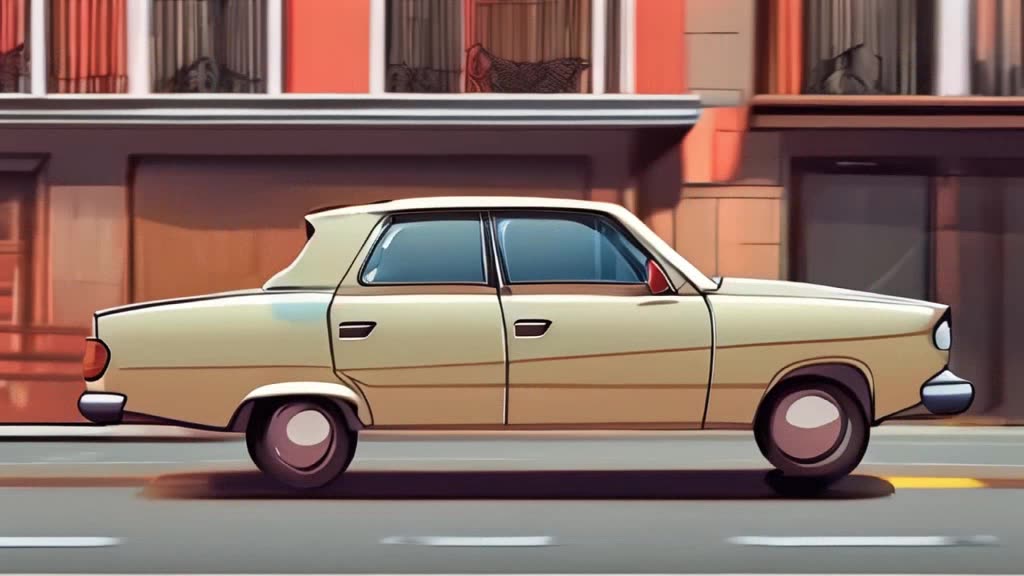}} \raisebox{-.5\height}{\includegraphics[width=0.09\textwidth]{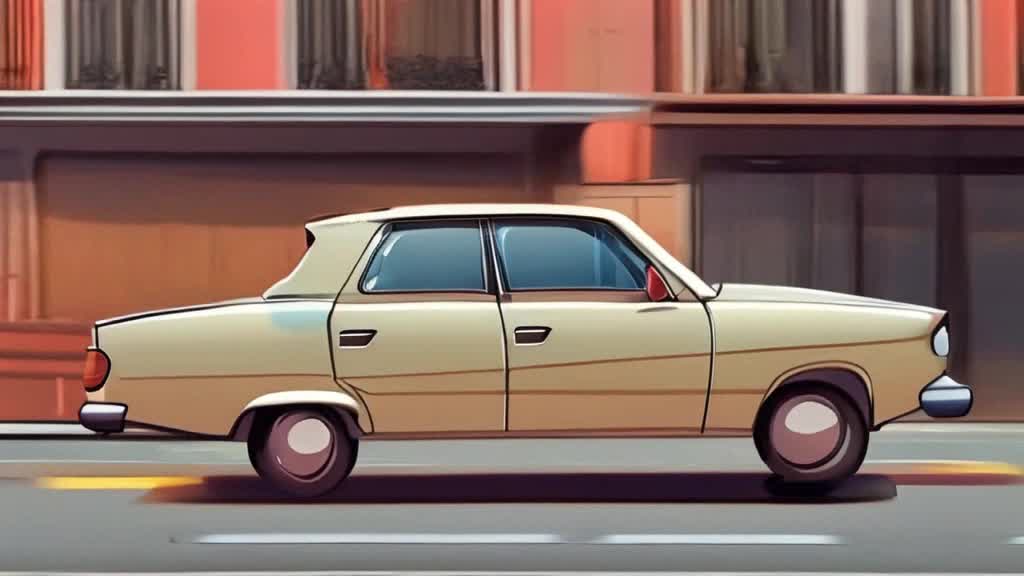}} \raisebox{-.5\height}{\includegraphics[width=0.09\textwidth]{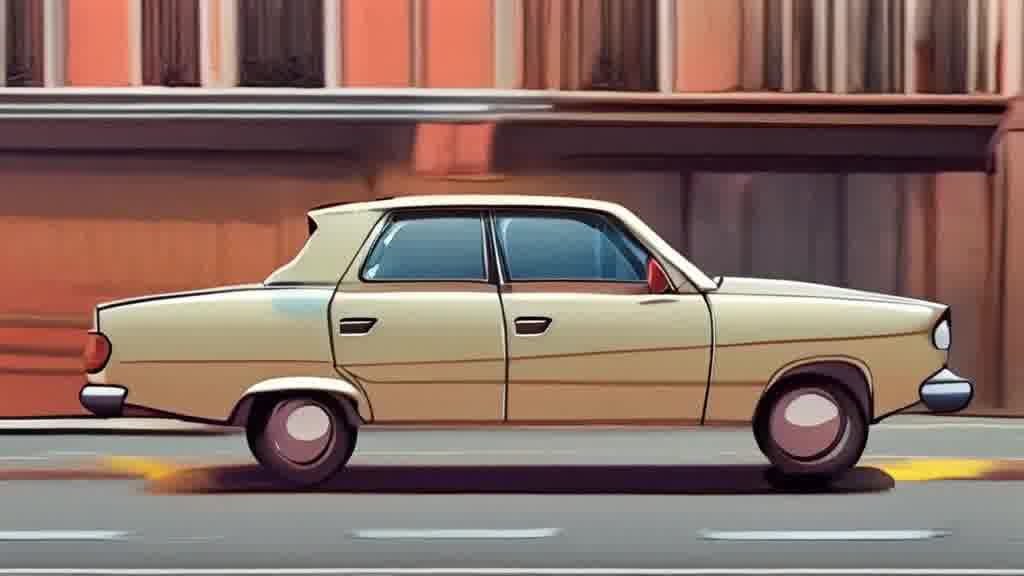}} \\
		{Gen. 4}  & \raisebox{-.5\height}{\includegraphics[width=0.09\textwidth]{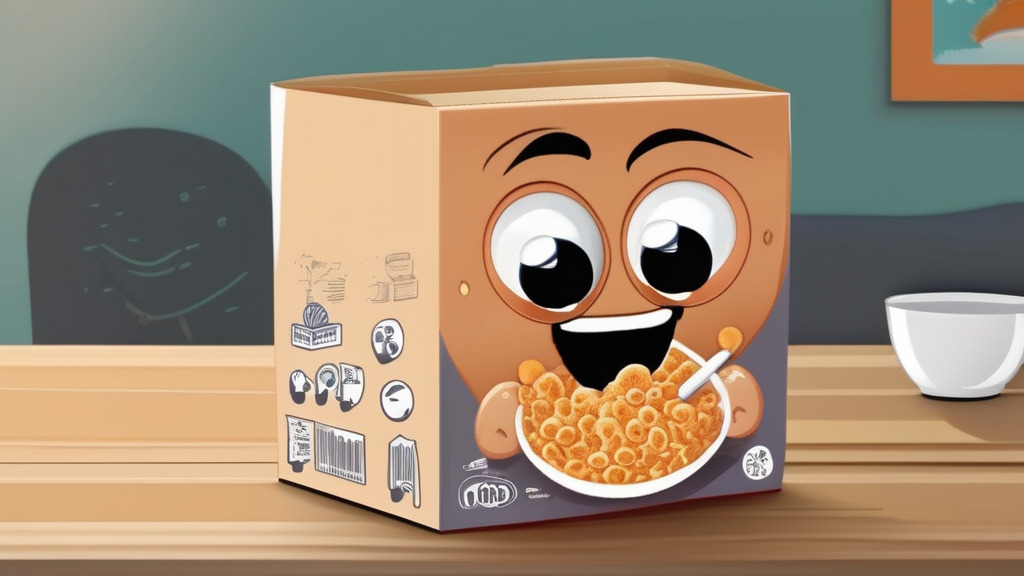}} & \raisebox{-.5\height}{\includegraphics[width=0.09\textwidth]{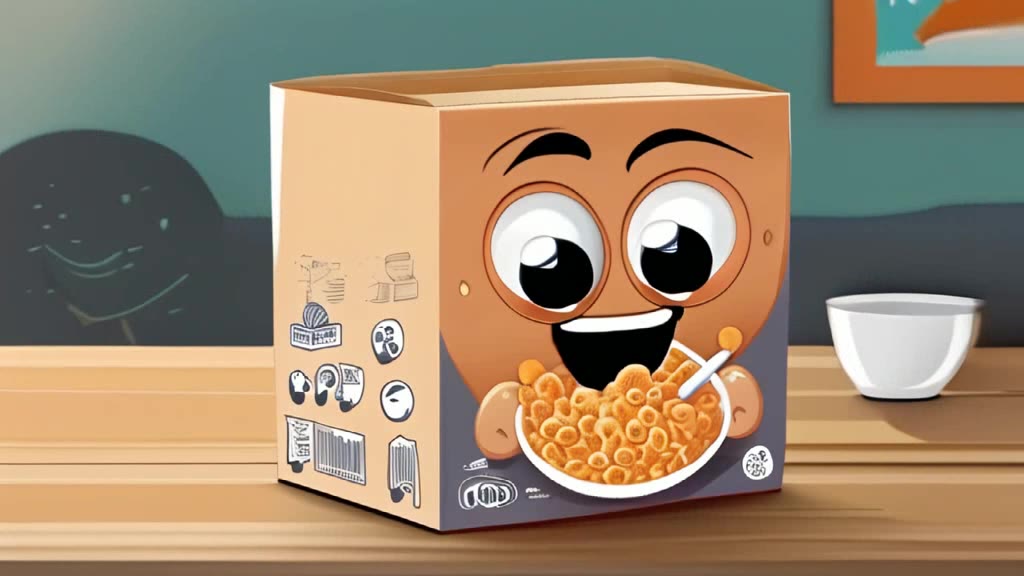}} \raisebox{-.5\height}{\includegraphics[width=0.09\textwidth]{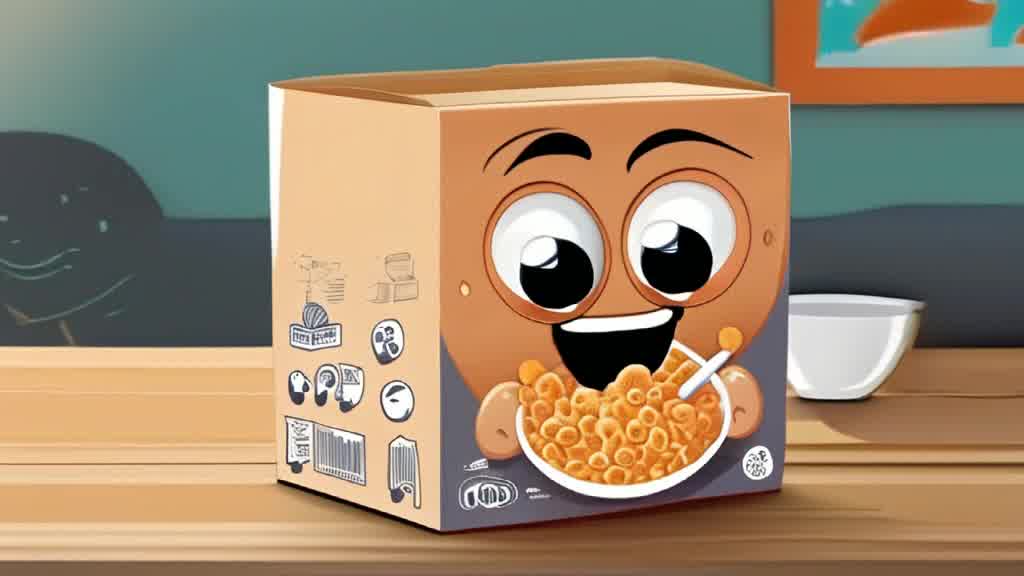}} \raisebox{-.5\height}{\includegraphics[width=0.09\textwidth]{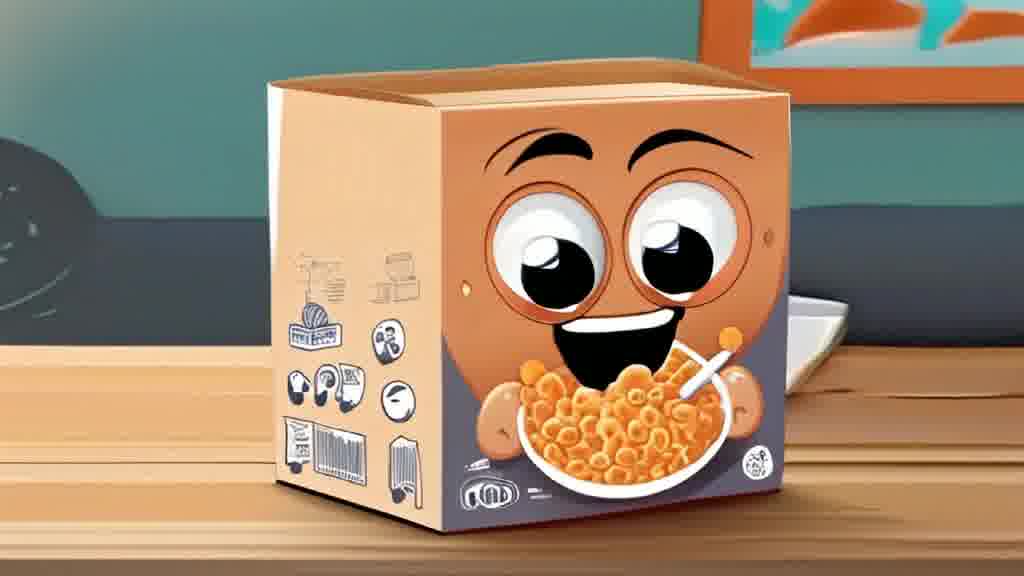}} & \raisebox{-.5\height}{\includegraphics[width=0.09\textwidth]{figures/style/trot/first_frame_4.jpg}} & \raisebox{-.5\height}{\includegraphics[width=0.09\textwidth]{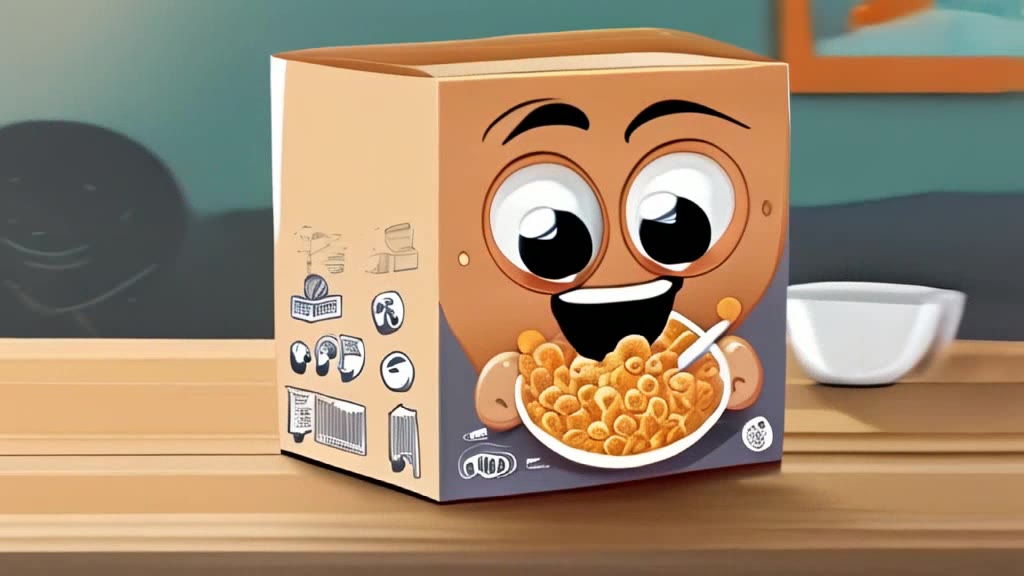}} \raisebox{-.5\height}{\includegraphics[width=0.09\textwidth]{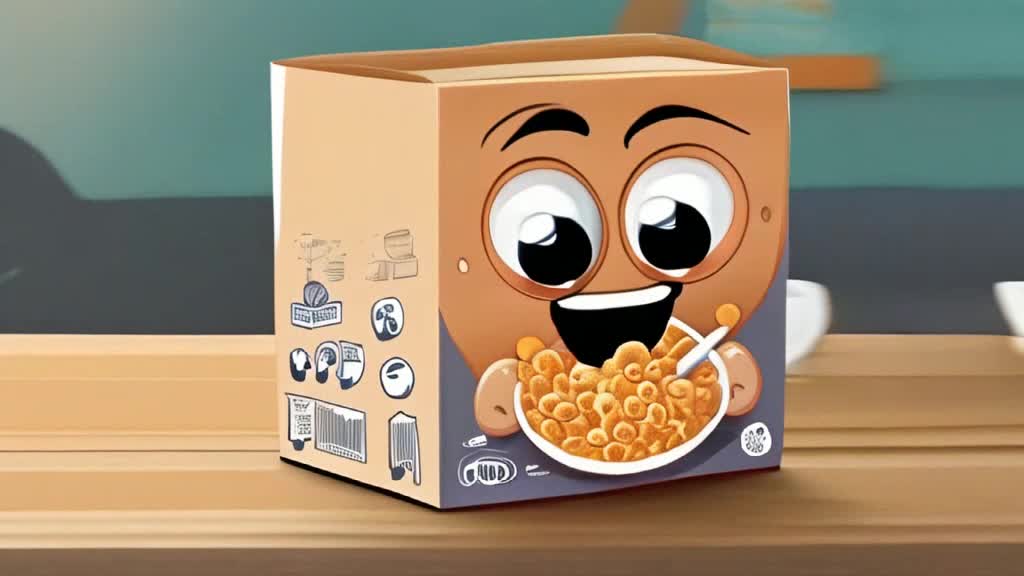}} \raisebox{-.5\height}{\includegraphics[width=0.09\textwidth]{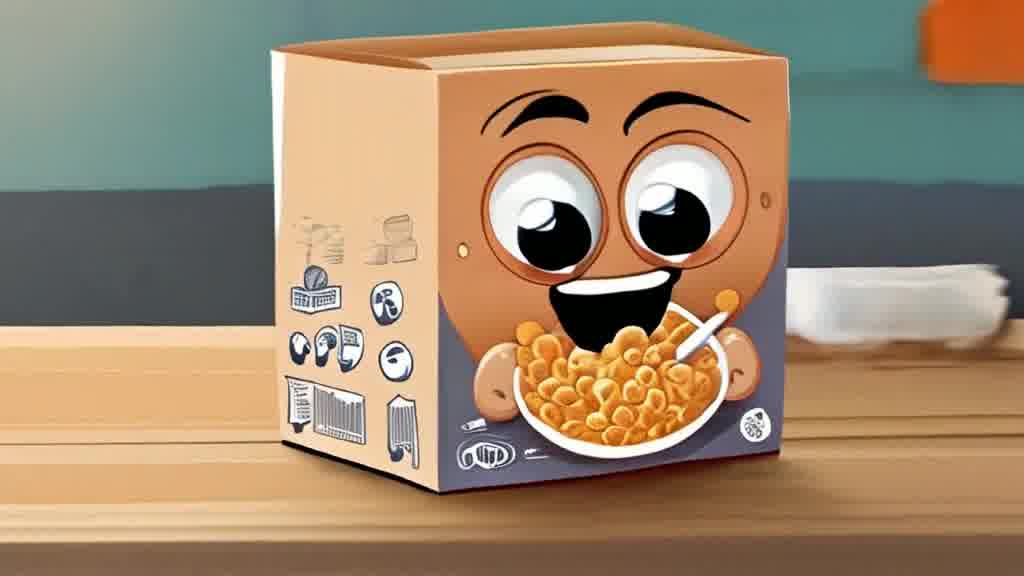}}
	\end{tblr}
	\caption{Motion style transfer. Our learned motion-text embeddings do not only store the rough motion category but also the style of the motion. Here, we apply two different gaits to the same target image: a horse trot (smooth) and a canter (rocking). The resulting videos for the cartoon dog are not only showing the dog moving, but their motions also closely match the motion reference video's gait style. Furthermore, the extreme cross-domain examples with the boat, car, and cereal box show that the essence of the motion style is preserved even across completely different objects.}
	\Description{Grid with two columns. The first row shows motion reference videos of two different horse gaits. The first column is a trot (smooth) whereas the second column is a canter (rocking). The second row shows the motions applied to a cartoon dog, the third to a boat, the fourth to a car, and the last to a cereal box.}
	\label{fig:results_style}
\end{figure*}

\begin{figure*}[htbp]
	\centering
	\begin{tblr}{
			vline{3} = {2-3}{dashed},
		}
		{Reference} & \raisebox{-.5\height}{\includegraphics[width=0.15\textwidth]{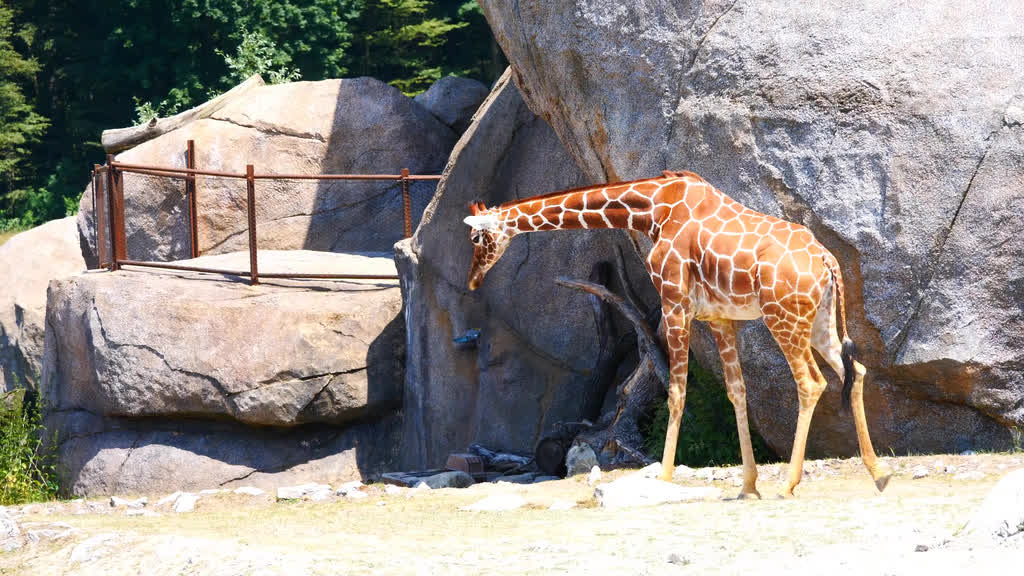}} & \raisebox{-.5\height}{\includegraphics[width=0.15\textwidth]{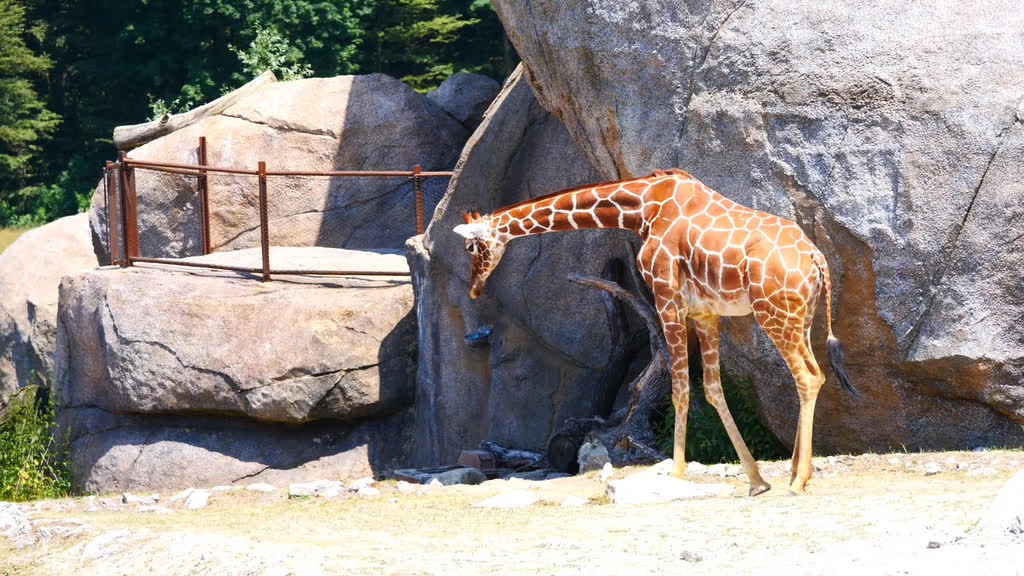}} \raisebox{-.5\height}{\includegraphics[width=0.15\textwidth]{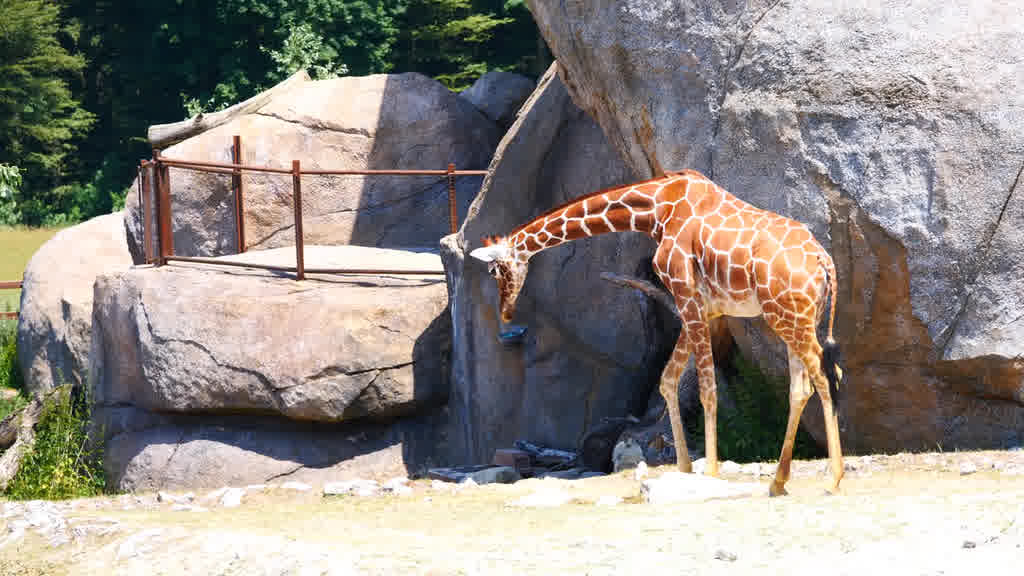}} \raisebox{-.5\height}{\includegraphics[width=0.15\textwidth]{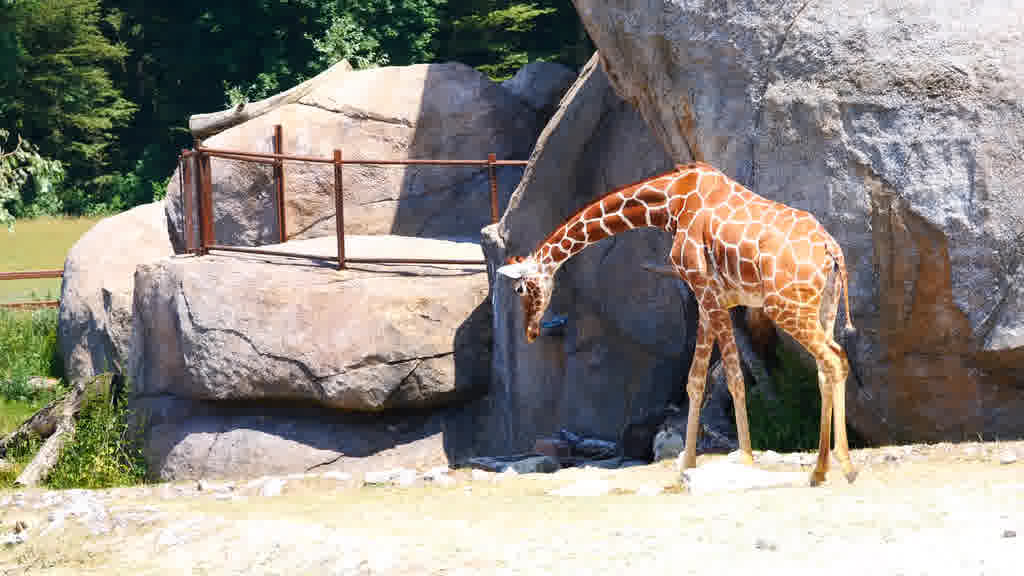}} \\
		{Regular input image}  & \raisebox{-.5\height}{\includegraphics[width=0.15\textwidth]{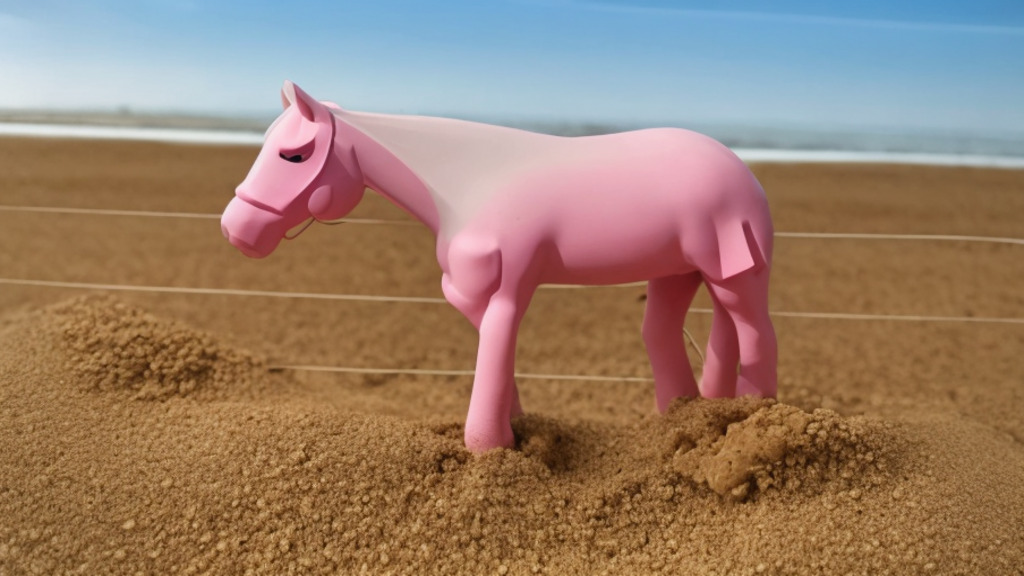}} & \raisebox{-.5\height}{\includegraphics[width=0.15\textwidth]{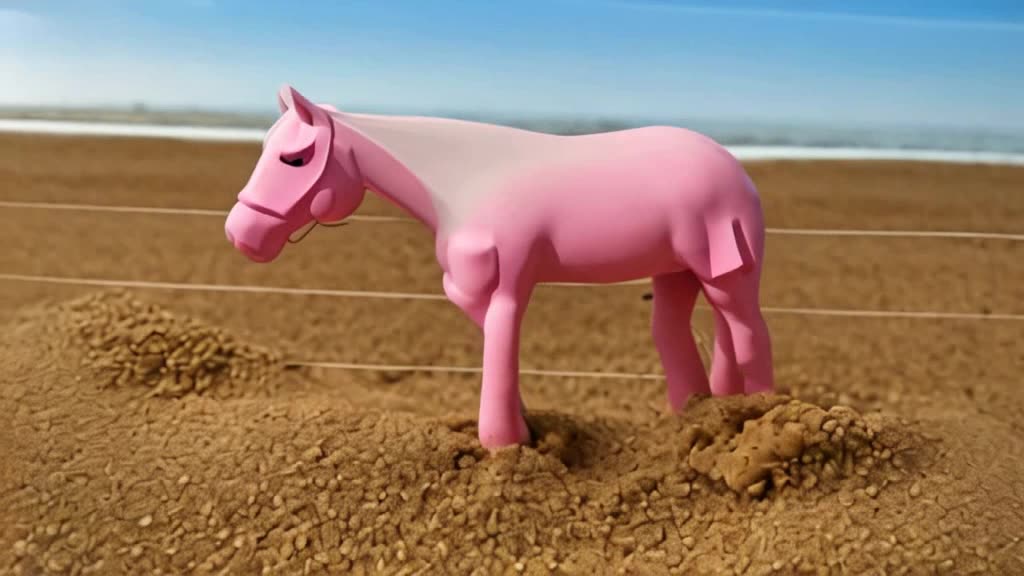}} \raisebox{-.5\height}{\includegraphics[width=0.15\textwidth]{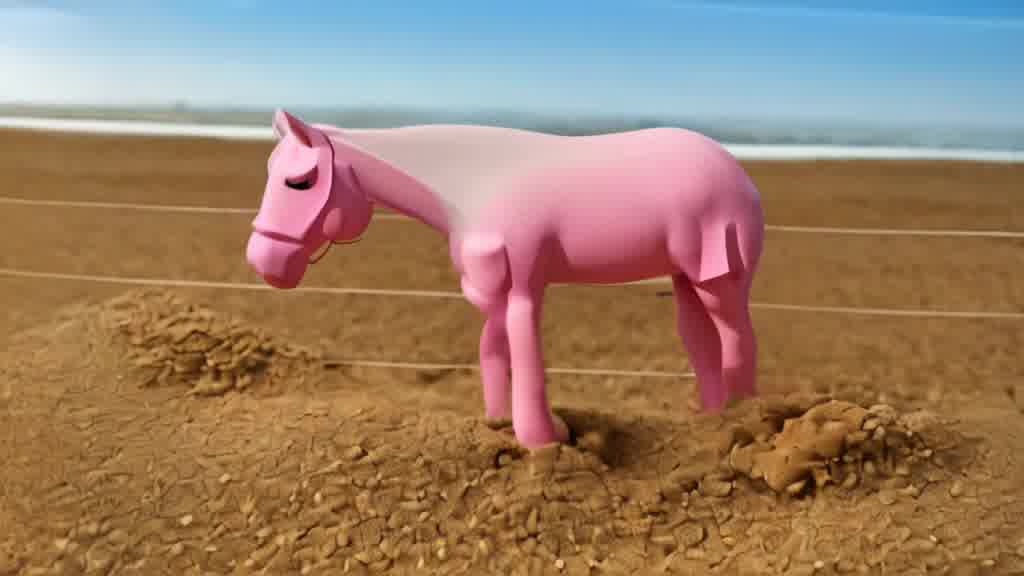}} \raisebox{-.5\height}{\includegraphics[width=0.15\textwidth]{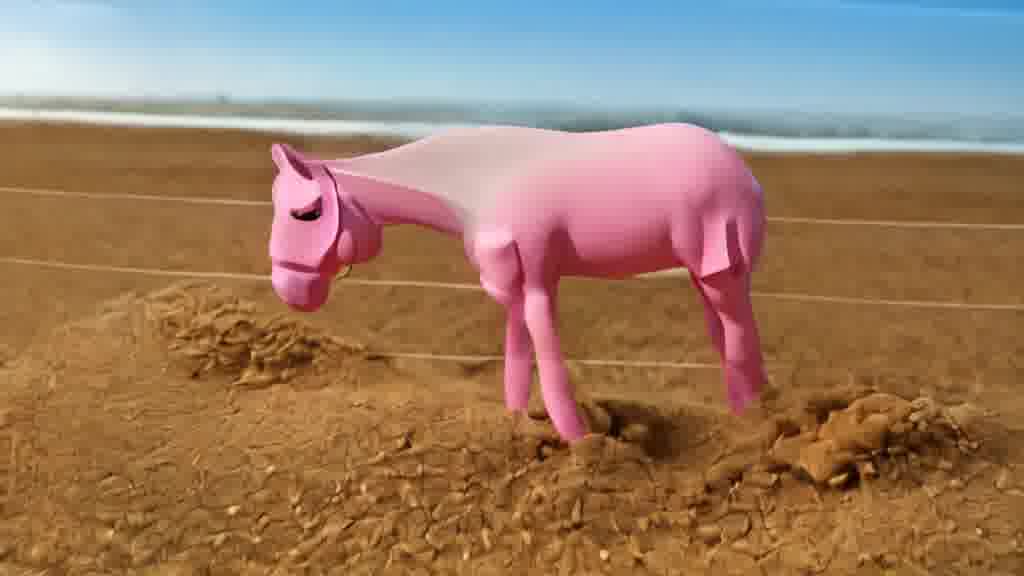}} \\
		{Flipped input image}  & \raisebox{-.5\height}{\includegraphics[width=0.15\textwidth]{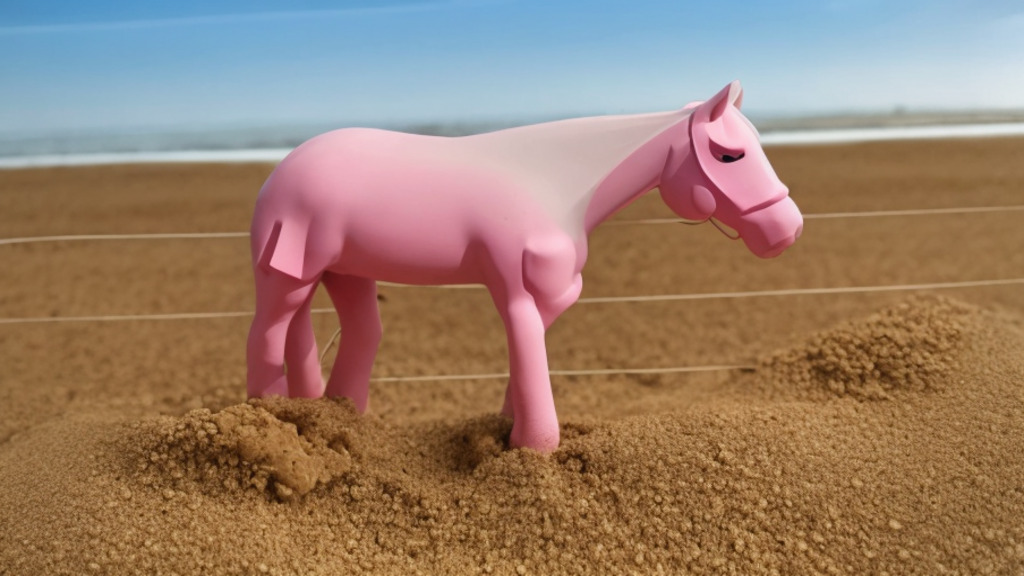}} & \raisebox{-.5\height}{\includegraphics[width=0.15\textwidth]{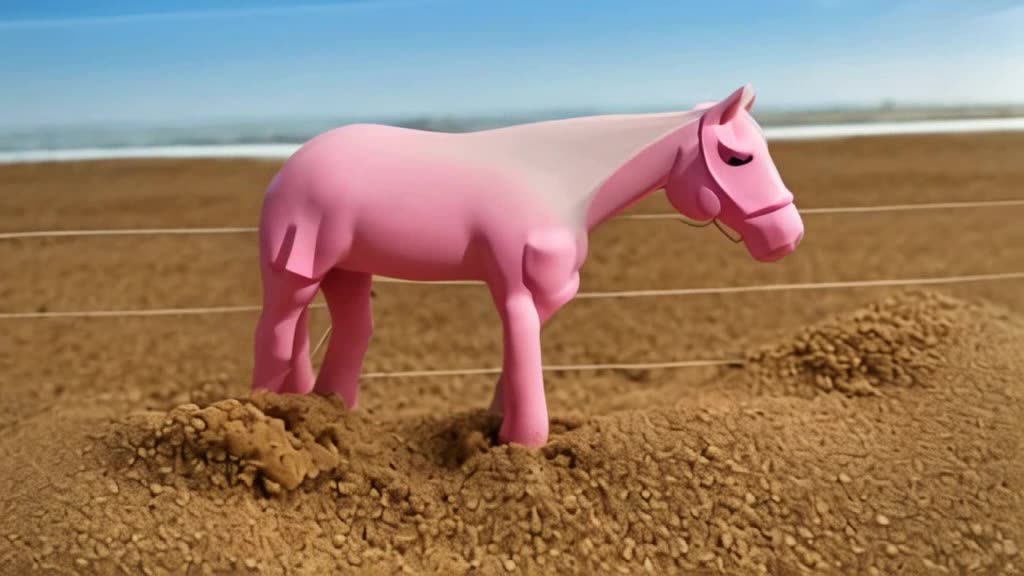}} \raisebox{-.5\height}{\includegraphics[width=0.15\textwidth]{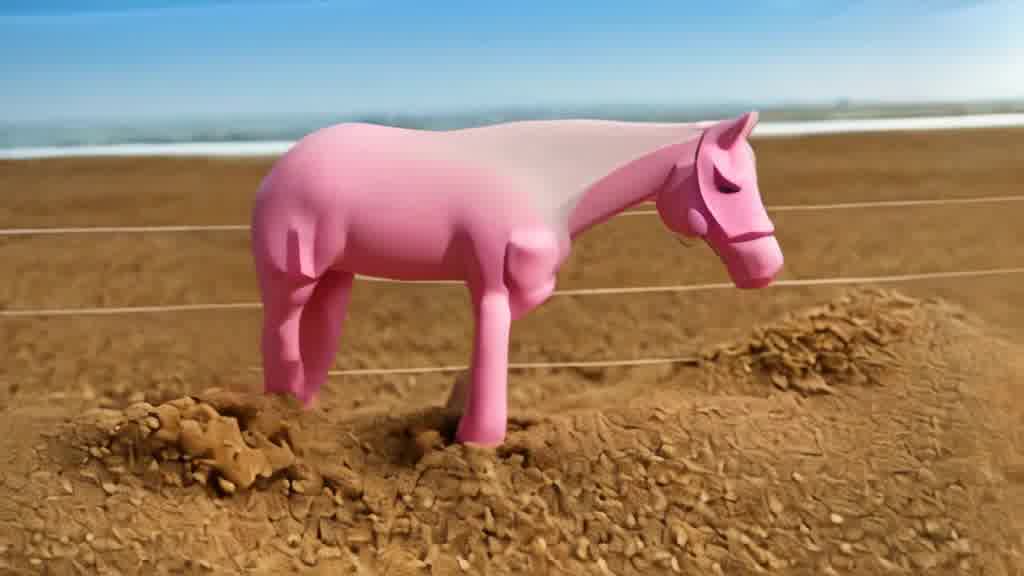}} \raisebox{-.5\height}{\includegraphics[width=0.15\textwidth]{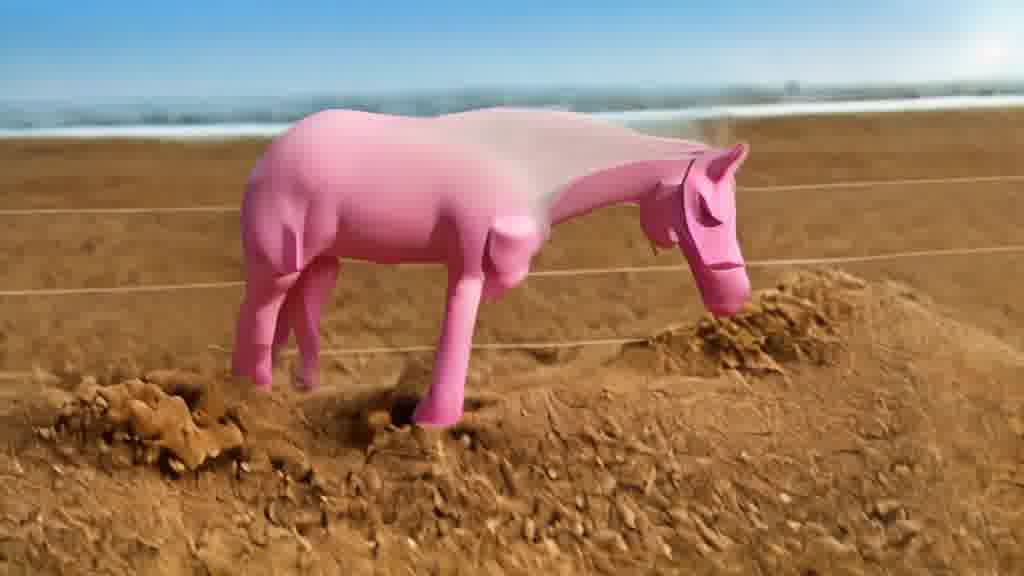}}
	\end{tblr}
	\caption{Semantic motion transfer. Our learned motion-text embeddings store the semantic motion (animal moving in the direction it is facing and moving its head down) rather than the spatial motion (animal moving from right to left and left part is going down). This can be seen in the above example where we apply the same learned motion-text embedding to a flipped input image, and our method produces semantically similar results.}
	\Description{In the top row, i.e., the reference video, a giraffe is moving from the right to the left in the direction it is facing and is moving its head down. In the second row, this motion is applied to a pink toy horse that is also facing the left side of the screen. In the third row, the motion is applied to a flipped version of the pink horse, where the horse is now facing the right side. Both resulting videos have similar semantic motions.}
	\label{fig:results_no_alignment}
\end{figure*}

\begin{figure*}[htbp]
	\centering
	\begin{tblr}{
			vline{3} = {2,3,5,6,8,9,11,12}{dashed},
			vline{4} = {1-12}{},
			vline{5} = {2,3,5,6,8,9,11,12}{dashed},
			hline{4} = {1-5}{},
			hline{7} = {1-5}{},
			hline{10} = {1-5}{},
		}
		{Ref.} & \raisebox{-.5\height}{\includegraphics[width=0.09\textwidth]{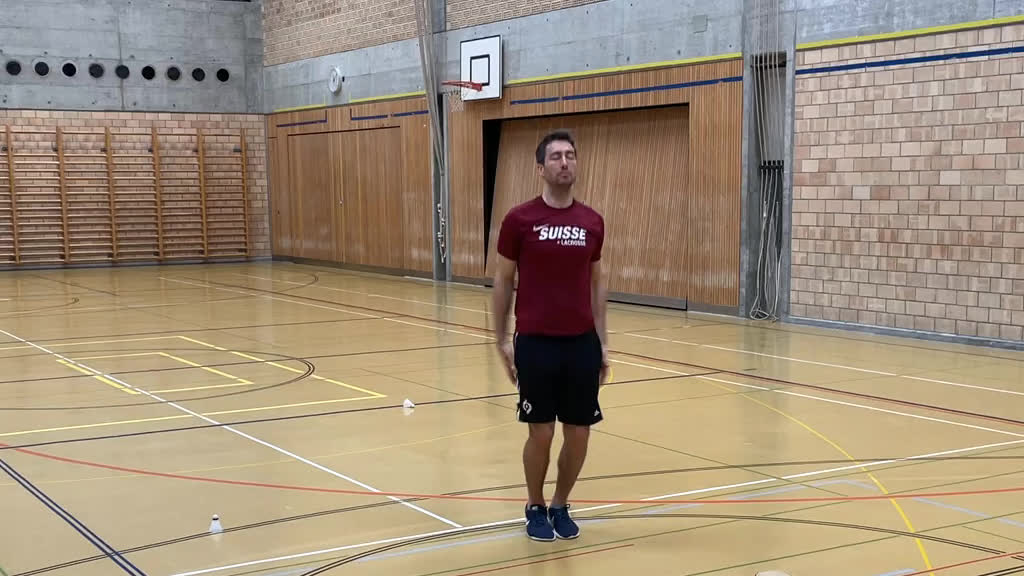}} & \raisebox{-.5\height}{\includegraphics[width=0.09\textwidth]{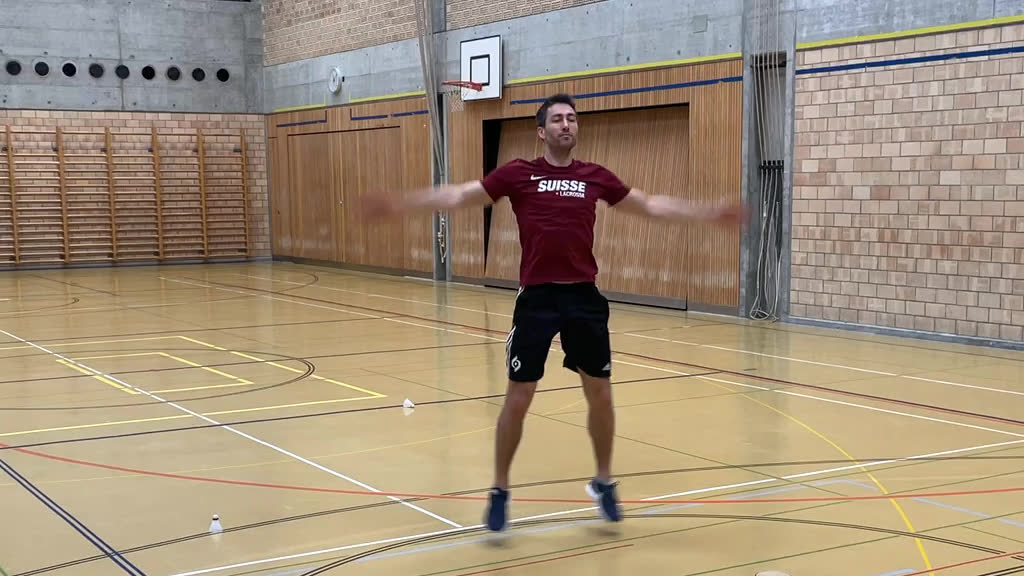}} \raisebox{-.5\height}{\includegraphics[width=0.09\textwidth]{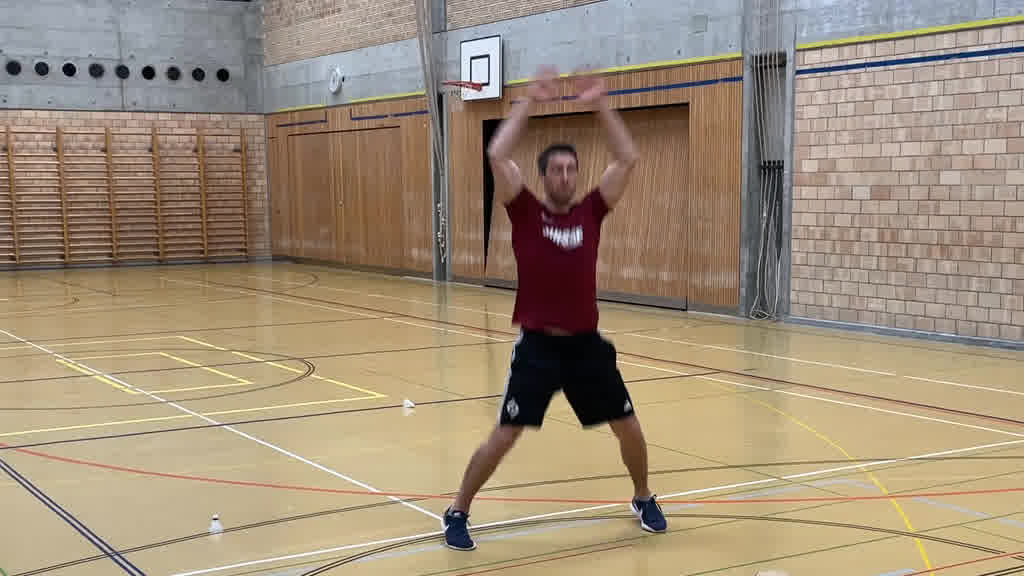}} \raisebox{-.5\height}{\includegraphics[width=0.09\textwidth]{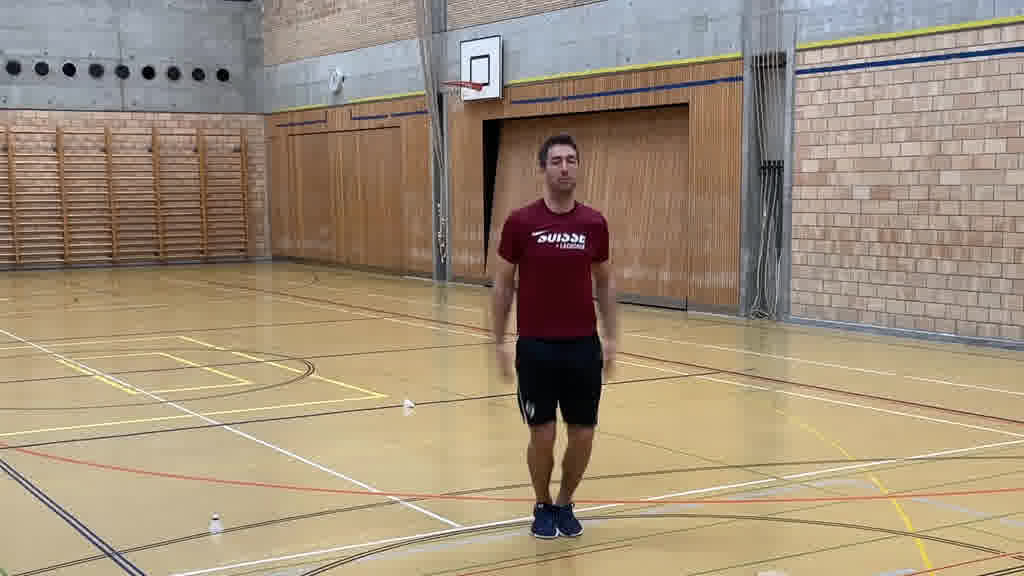}} & \raisebox{-.5\height}{\includegraphics[width=0.09\textwidth]{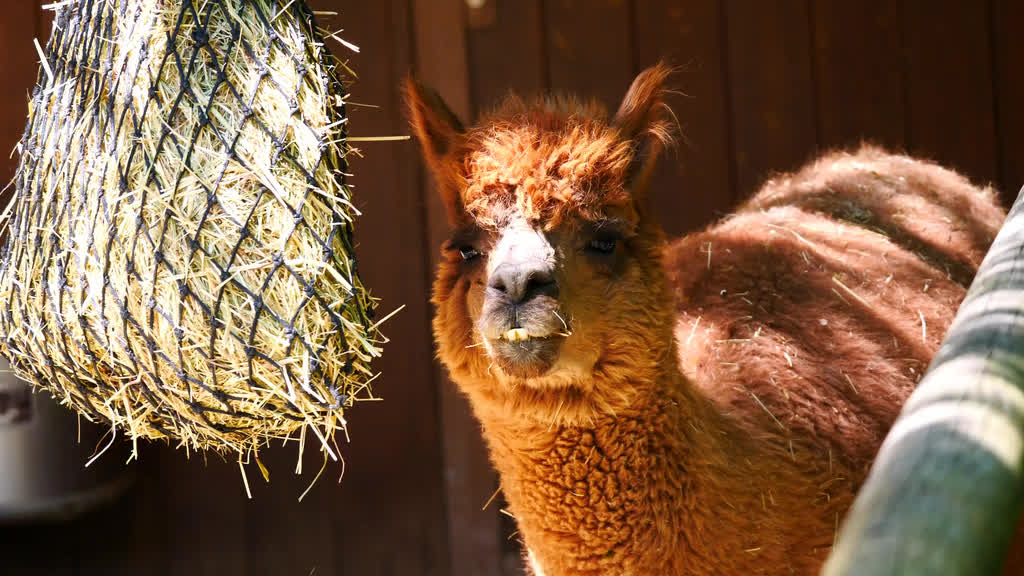}} & \raisebox{-.5\height}{\includegraphics[width=0.09\textwidth]{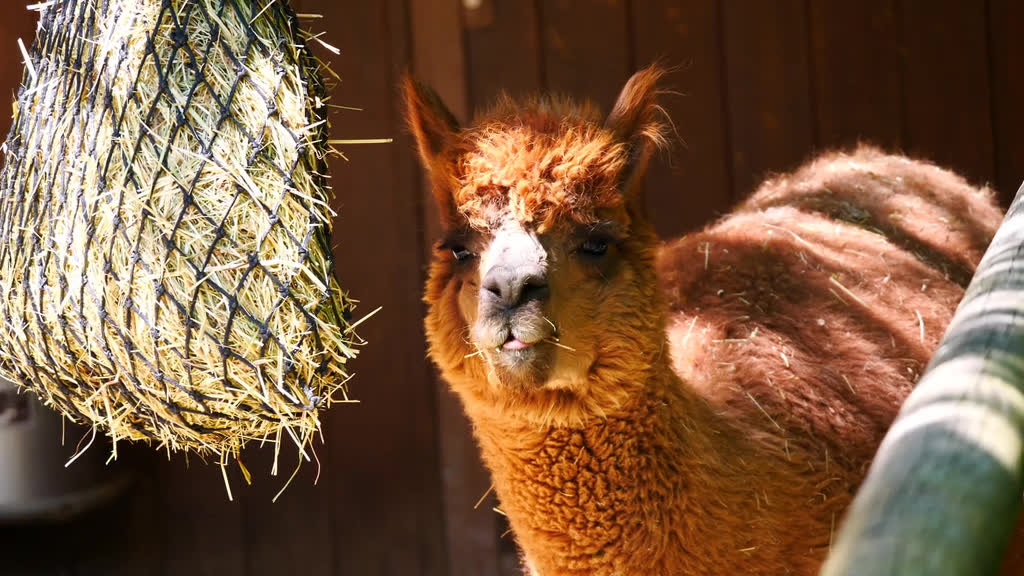}} \raisebox{-.5\height}{\includegraphics[width=0.09\textwidth]{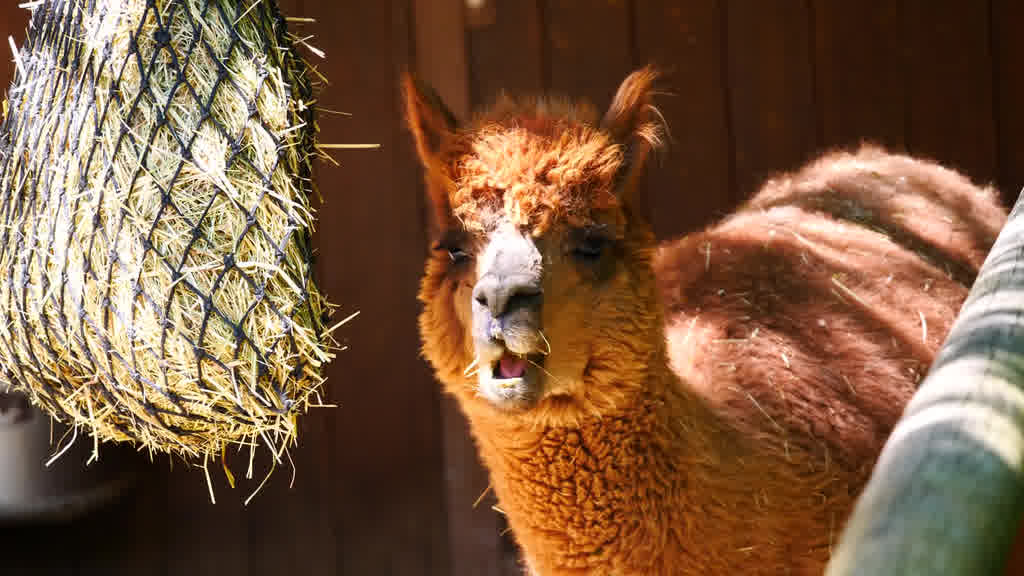}} \raisebox{-.5\height}{\includegraphics[width=0.09\textwidth]{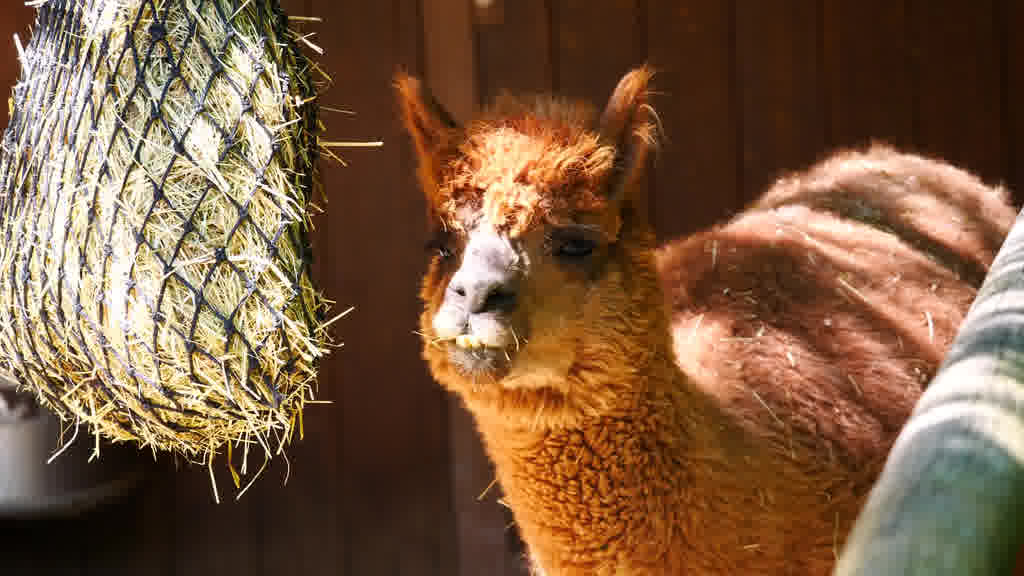}} \\
		{Gen. 1}  & \raisebox{-.5\height}{\includegraphics[width=0.09\textwidth]{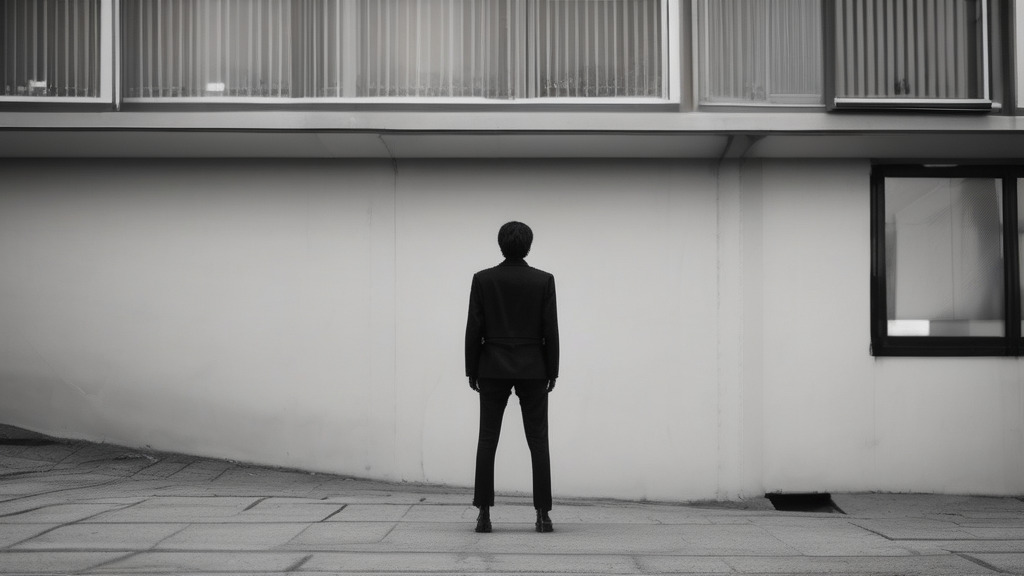}} & \raisebox{-.5\height}{\includegraphics[width=0.09\textwidth]{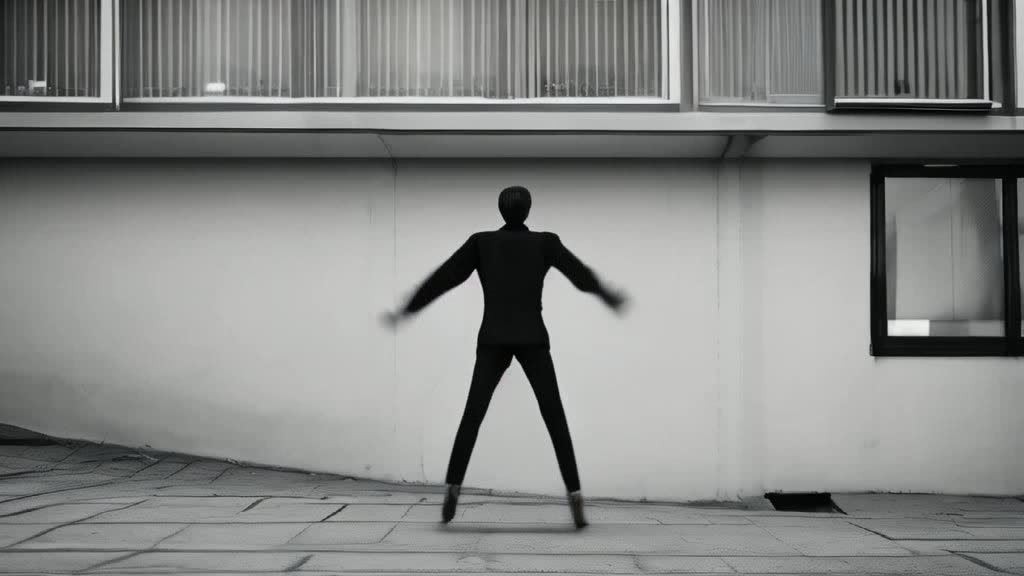}} \raisebox{-.5\height}{\includegraphics[width=0.09\textwidth]{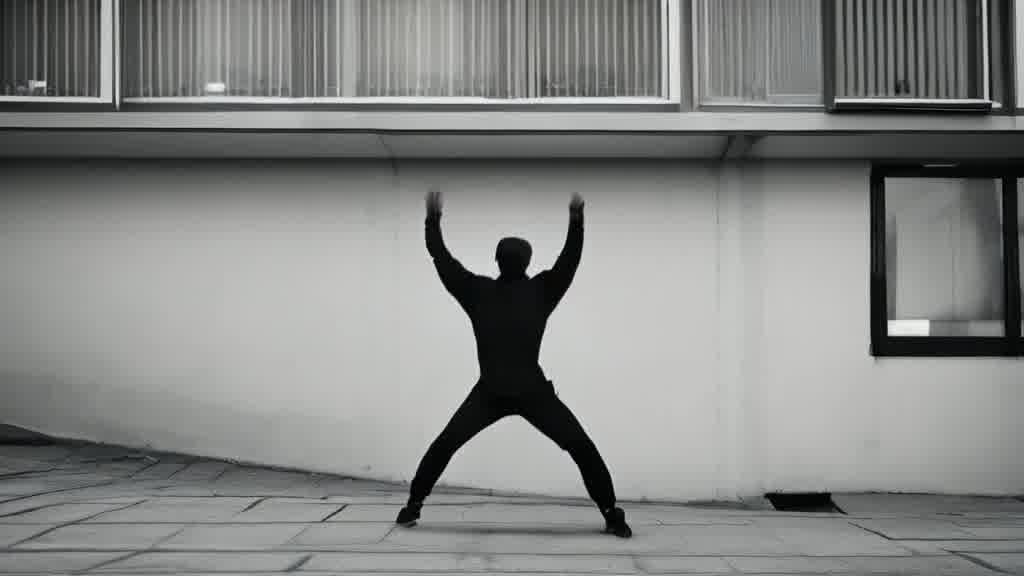}} \raisebox{-.5\height}{\includegraphics[width=0.09\textwidth]{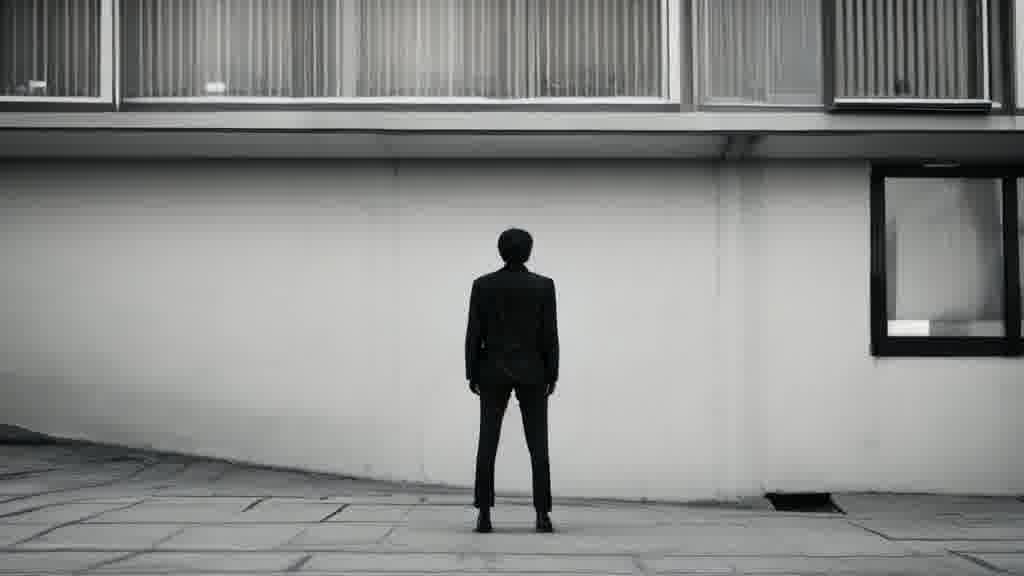}} & \raisebox{-.5\height}{\includegraphics[width=0.09\textwidth]{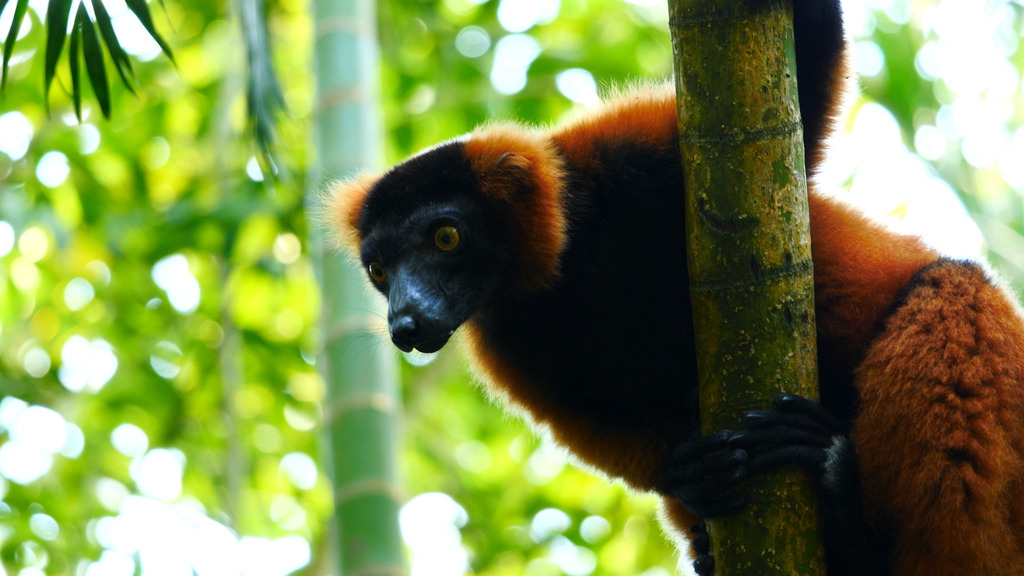}} & \raisebox{-.5\height}{\includegraphics[width=0.09\textwidth]{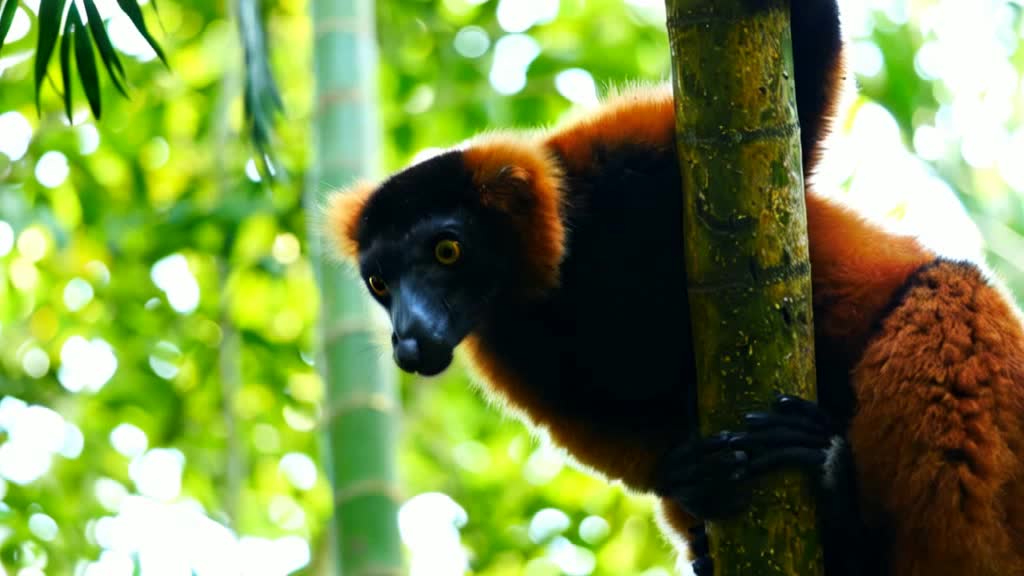}} \raisebox{-.5\height}{\includegraphics[width=0.09\textwidth]{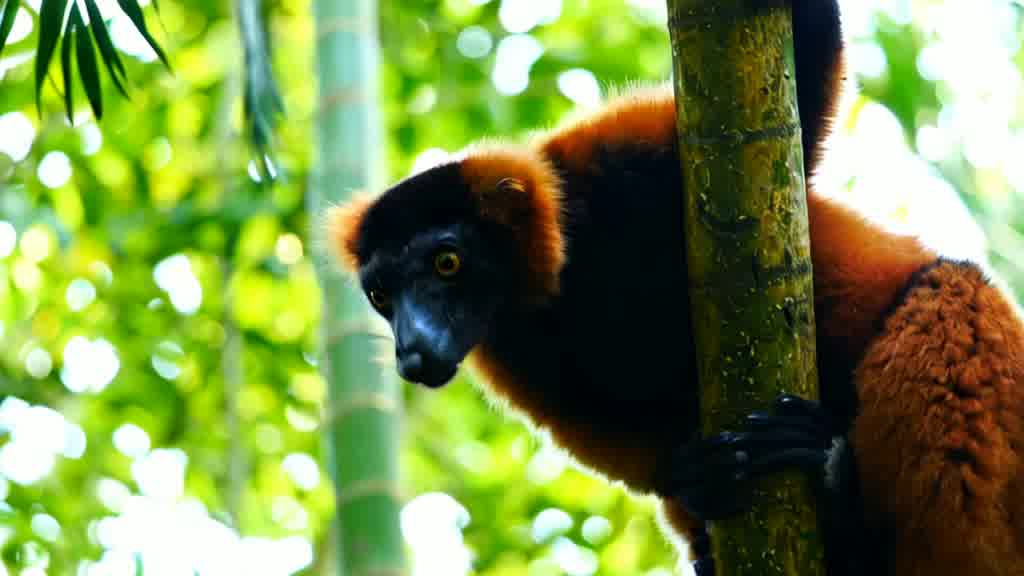}} \raisebox{-.5\height}{\includegraphics[width=0.09\textwidth]{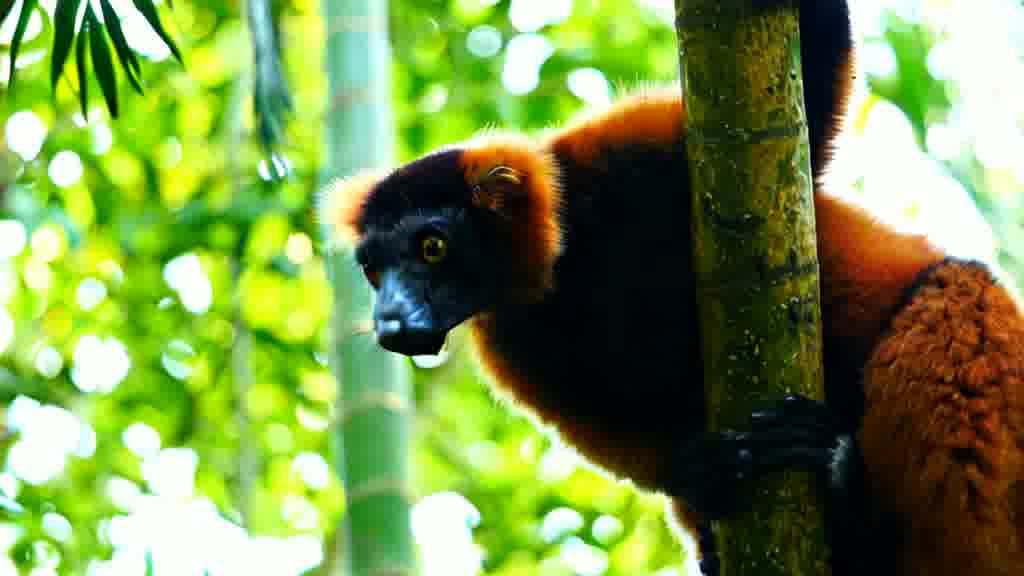}} \\
		{Gen. 2}  & \raisebox{-.5\height}{\includegraphics[width=0.09\textwidth]{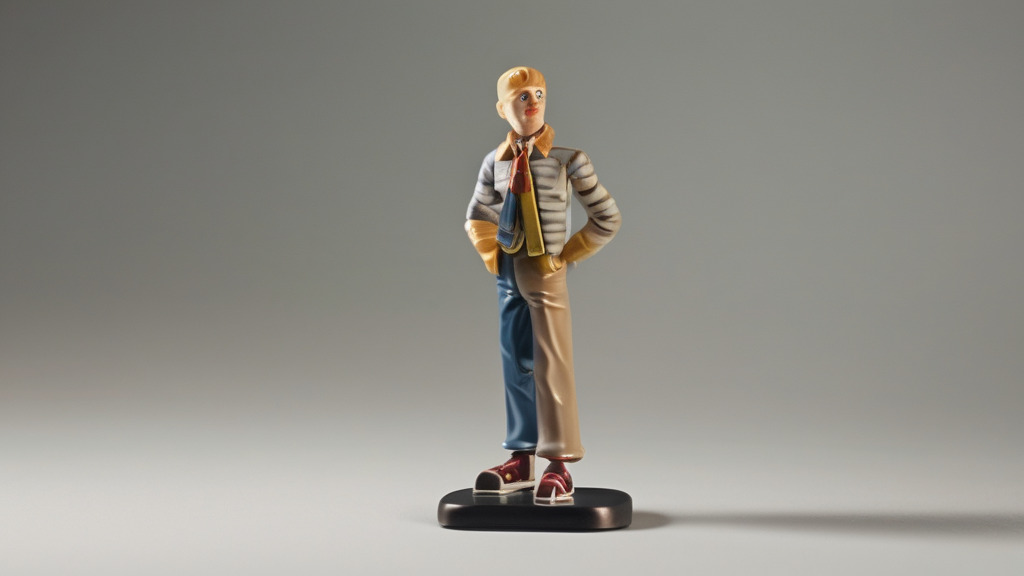}} & \raisebox{-.5\height}{\includegraphics[width=0.09\textwidth]{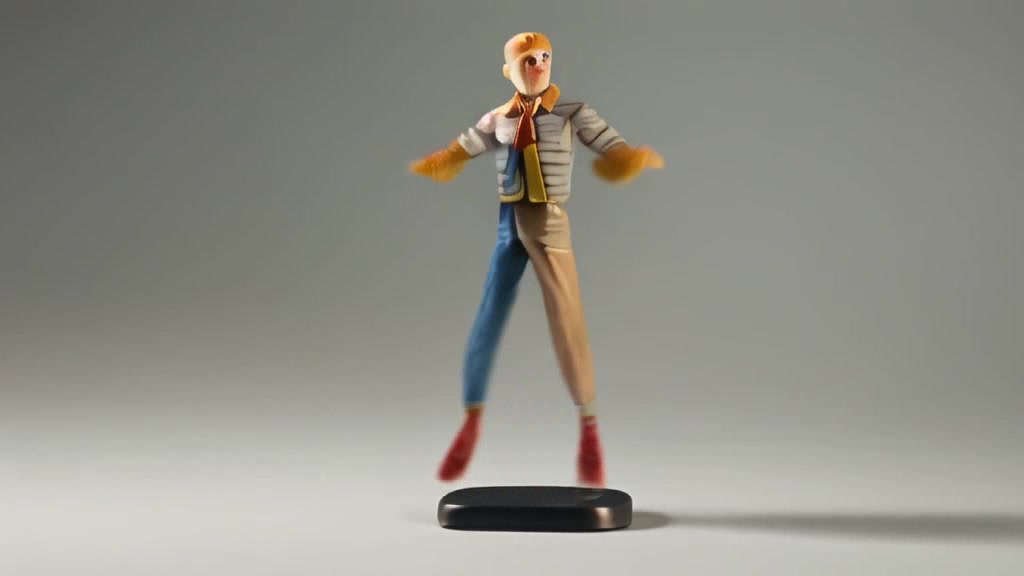}} \raisebox{-.5\height}{\includegraphics[width=0.09\textwidth]{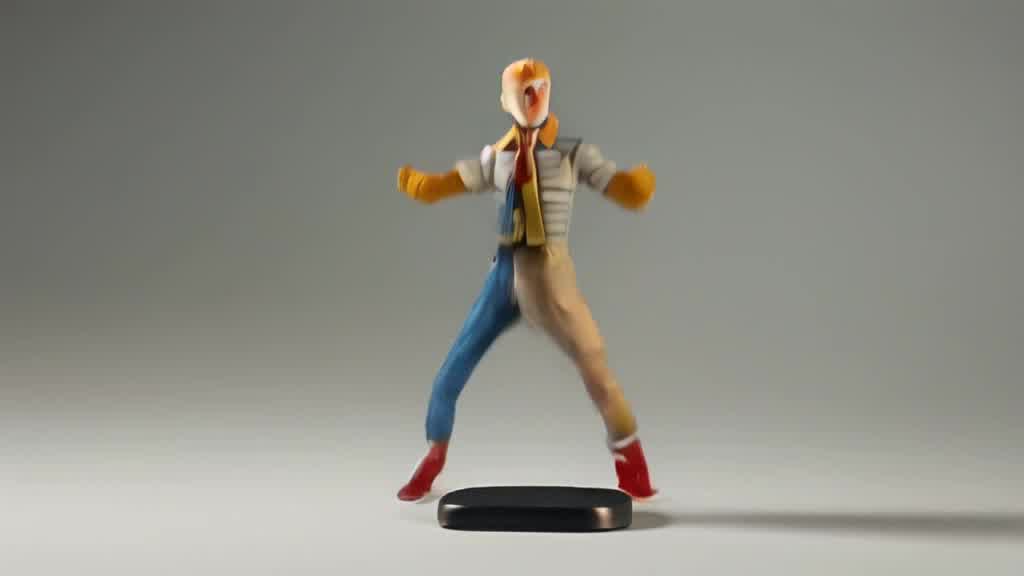}} \raisebox{-.5\height}{\includegraphics[width=0.09\textwidth]{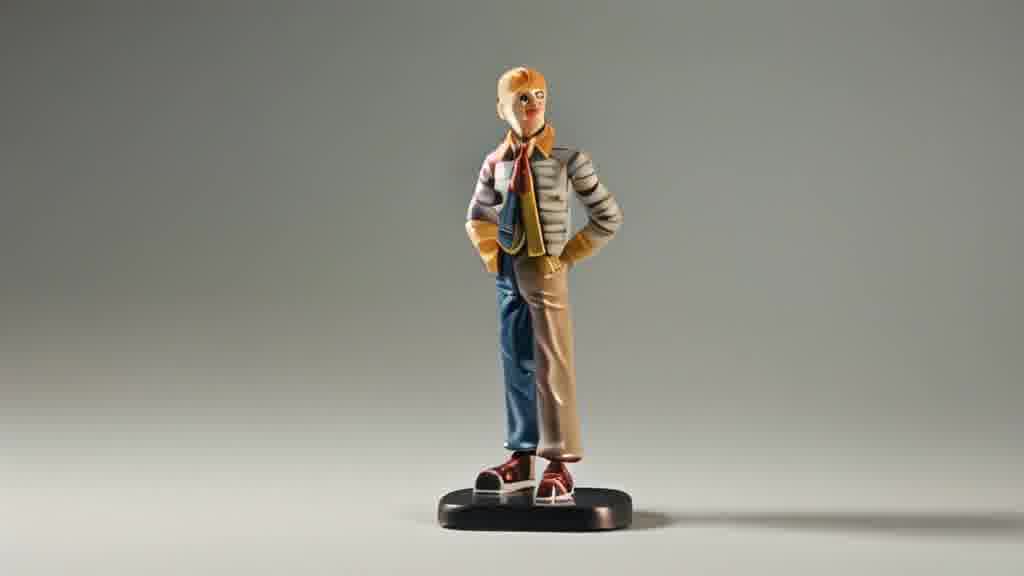}} & \raisebox{-.5\height}{\includegraphics[width=0.09\textwidth]{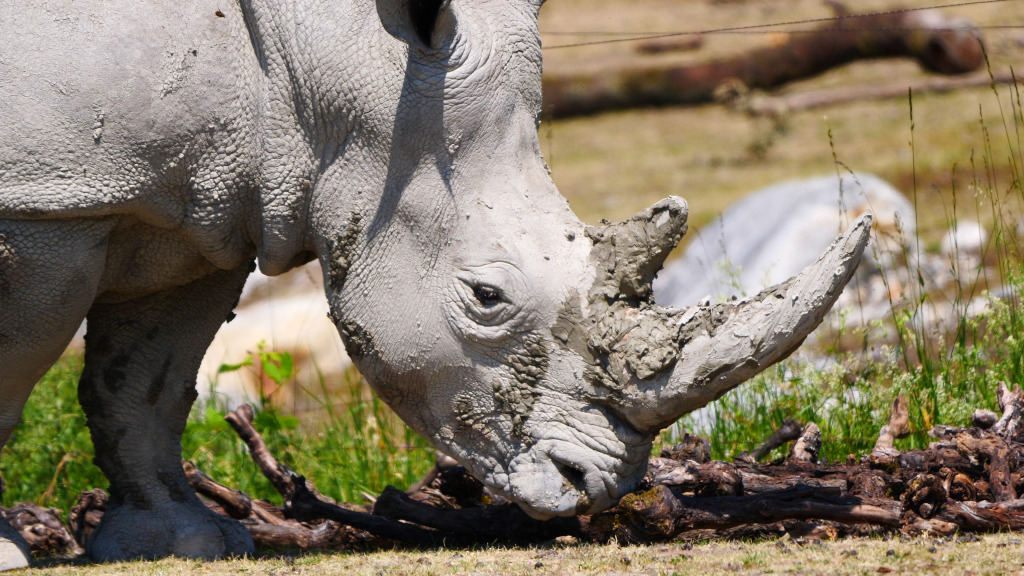}} & \raisebox{-.5\height}{\includegraphics[width=0.09\textwidth]{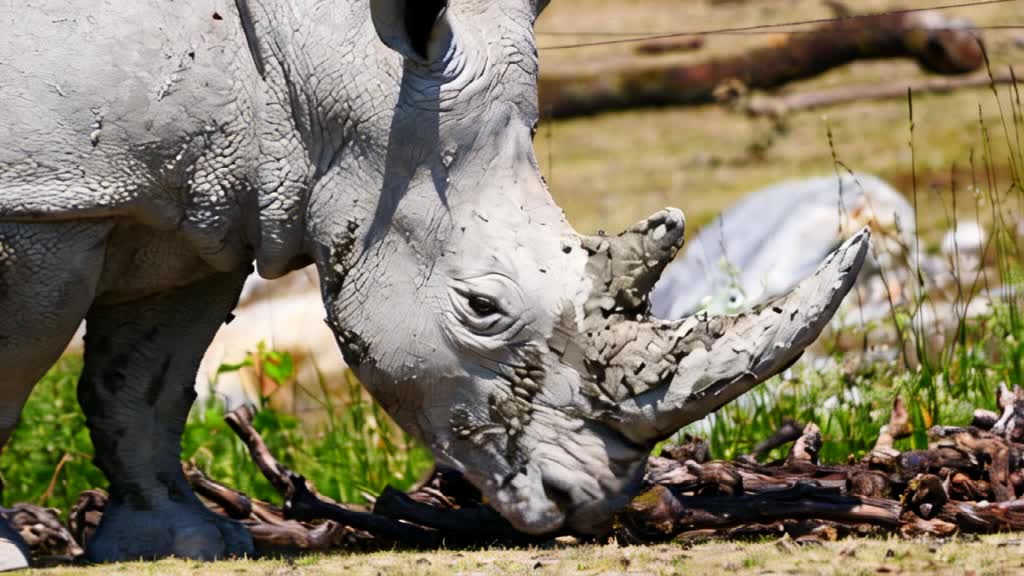}} \raisebox{-.5\height}{\includegraphics[width=0.09\textwidth]{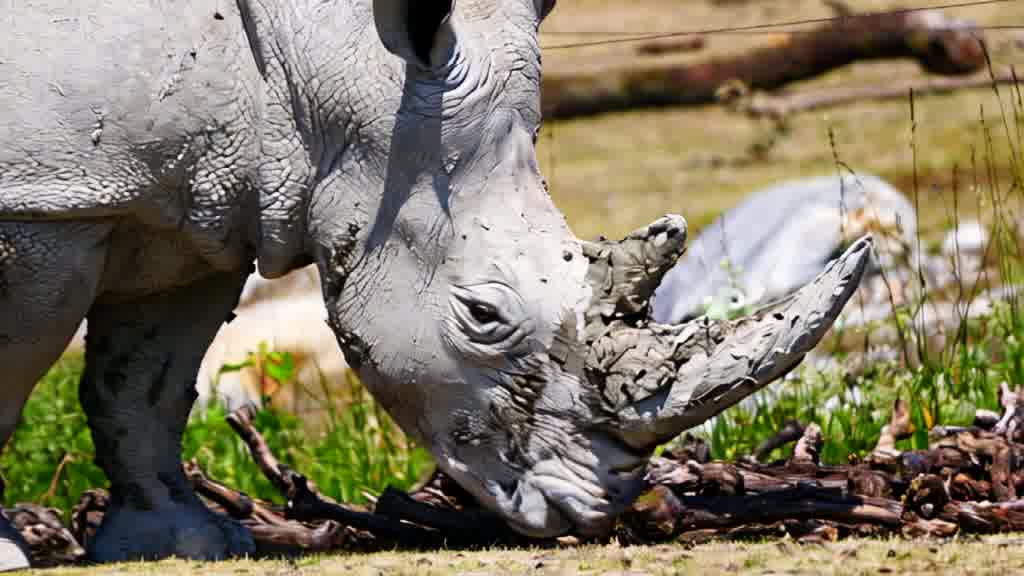}} \raisebox{-.5\height}{\includegraphics[width=0.09\textwidth]{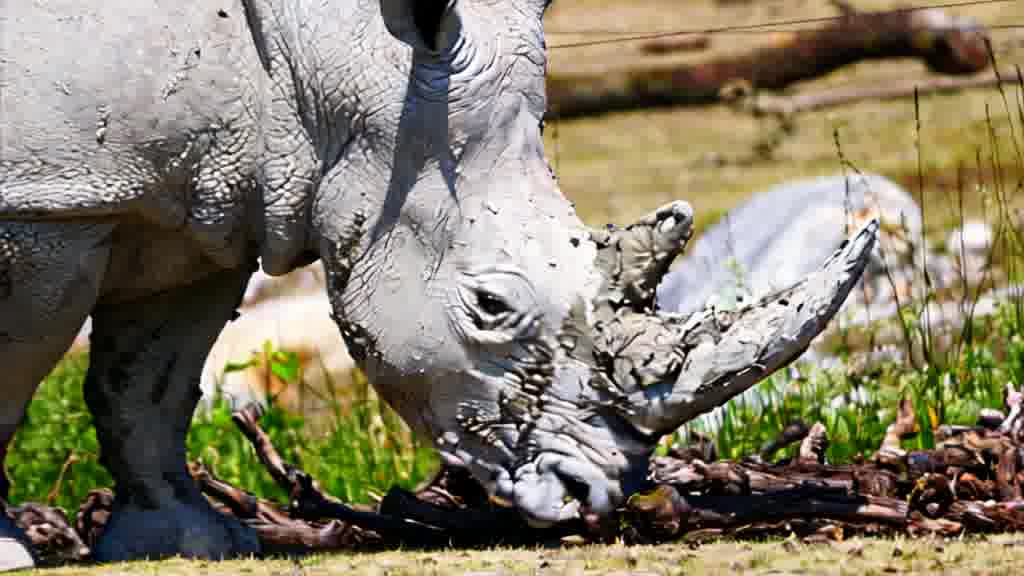}} \\
		{Ref.} & \raisebox{-.5\height}{\includegraphics[width=0.09\textwidth]{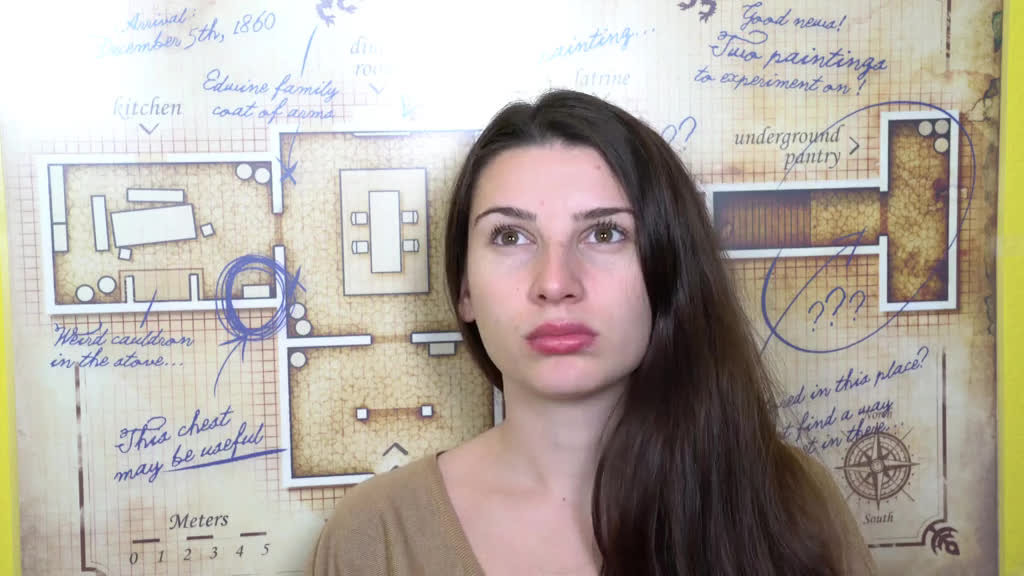}} & \raisebox{-.5\height}{\includegraphics[width=0.09\textwidth]{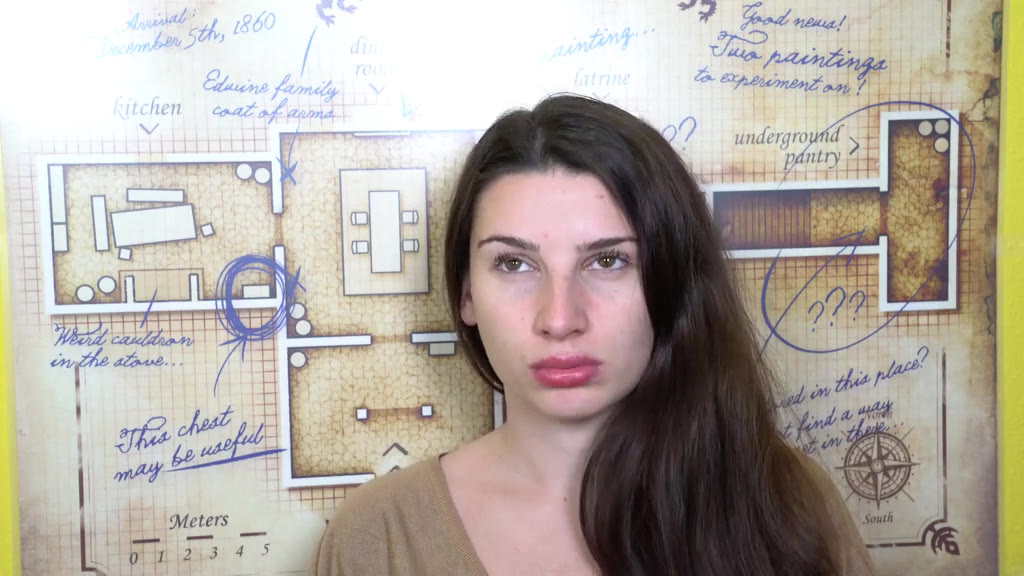}} \raisebox{-.5\height}{\includegraphics[width=0.09\textwidth]{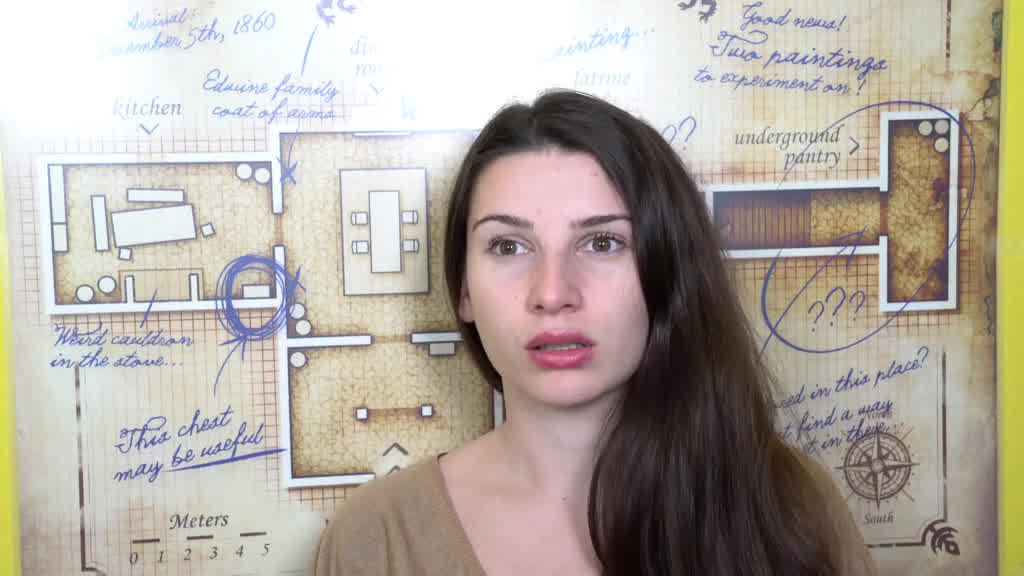}} \raisebox{-.5\height}{\includegraphics[width=0.09\textwidth]{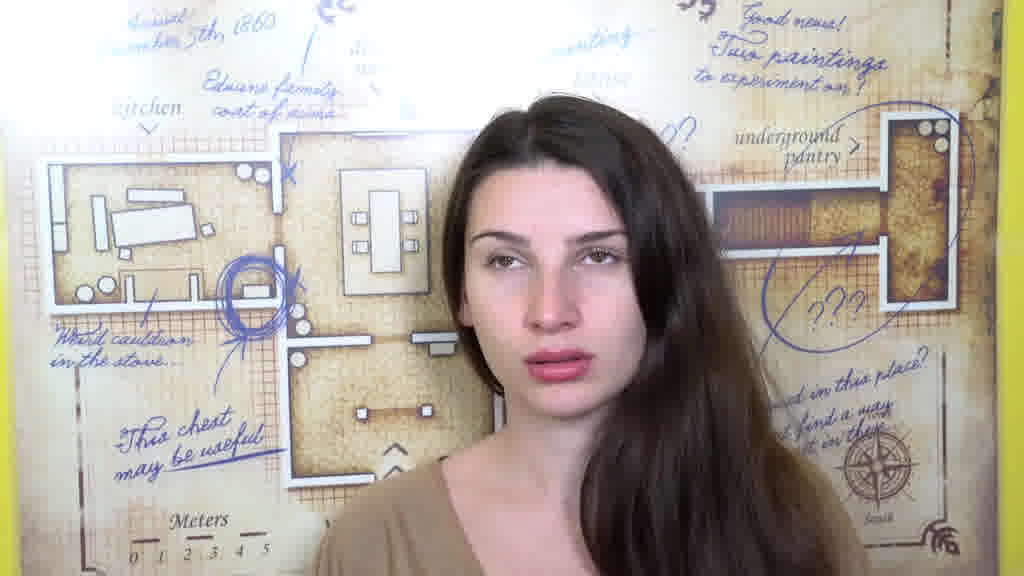}} & \raisebox{-.5\height}{\includegraphics[width=0.09\textwidth]{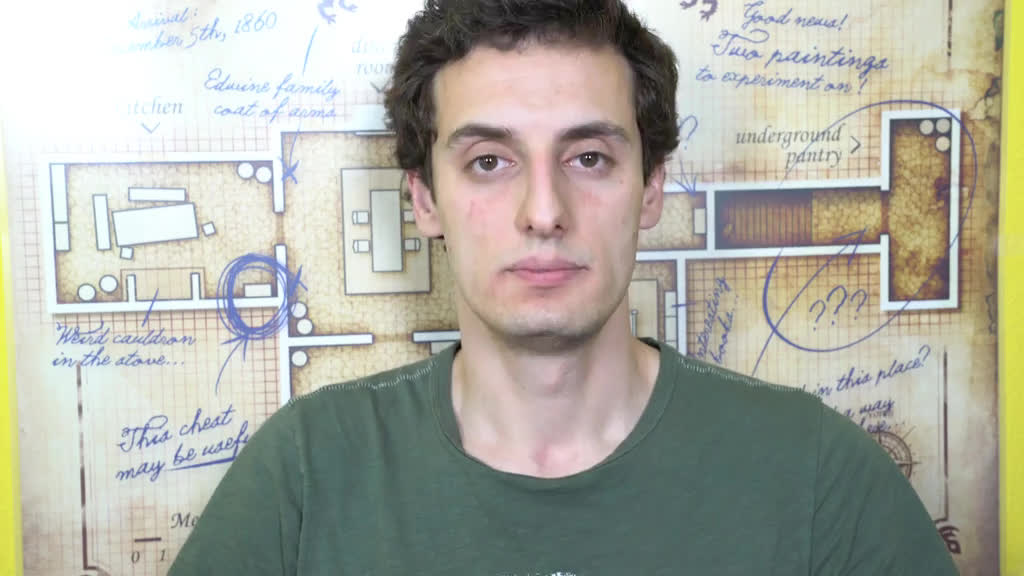}} & \raisebox{-.5\height}{\includegraphics[width=0.09\textwidth]{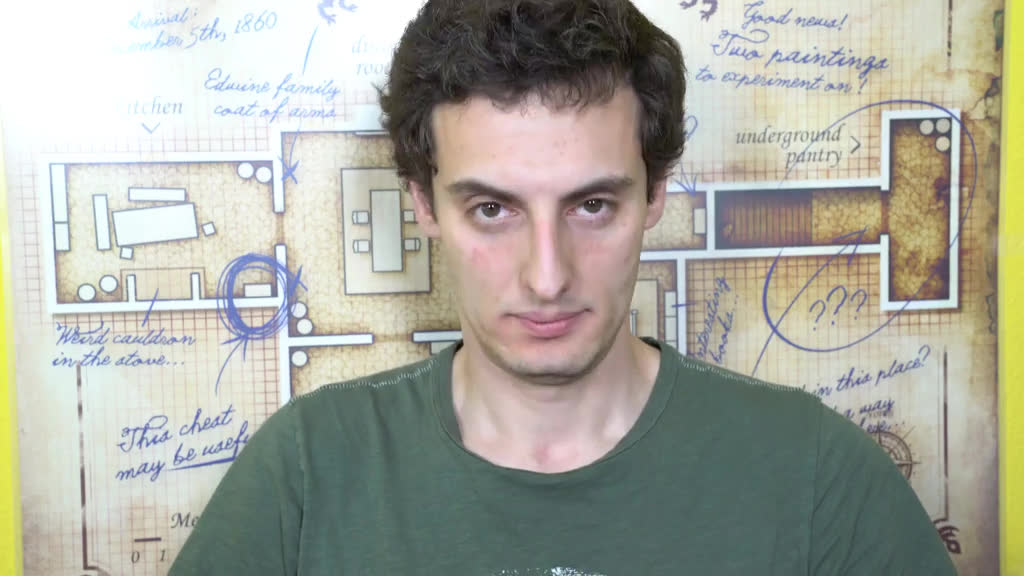}} \raisebox{-.5\height}{\includegraphics[width=0.09\textwidth]{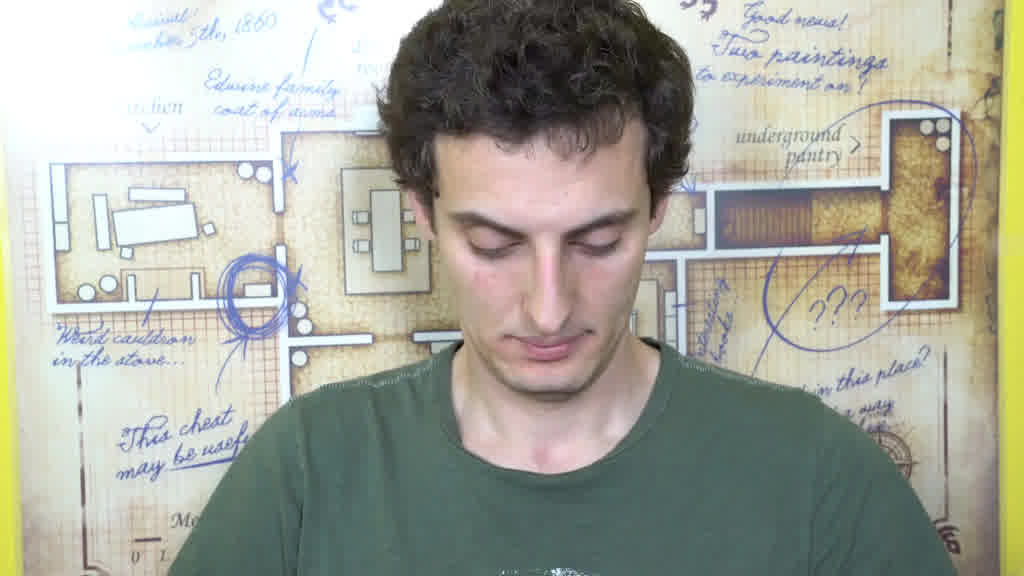}} \raisebox{-.5\height}{\includegraphics[width=0.09\textwidth]{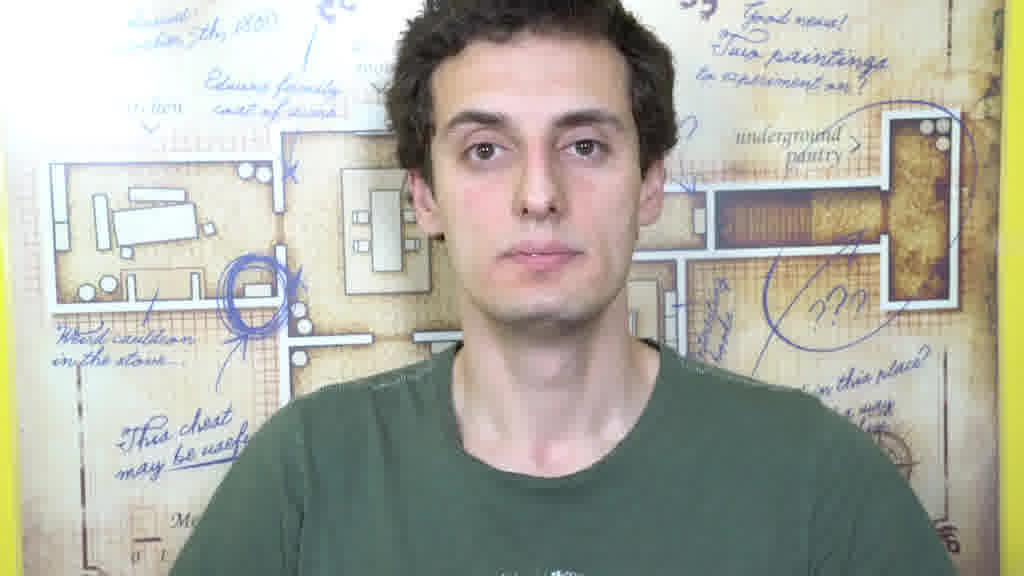}} \\
		{Gen. 1}  & \raisebox{-.5\height}{\includegraphics[width=0.09\textwidth]{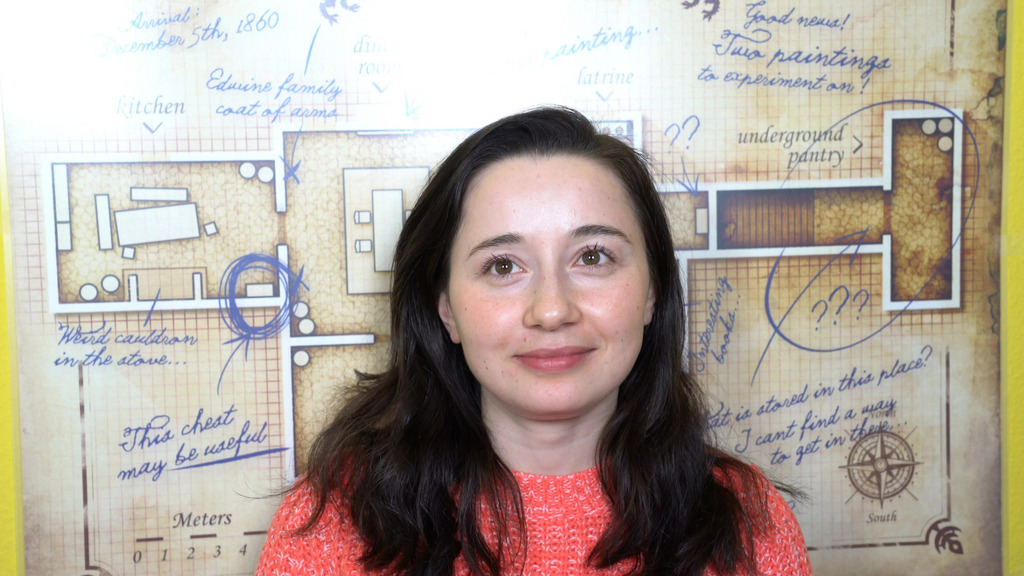}} & \raisebox{-.5\height}{\includegraphics[width=0.09\textwidth]{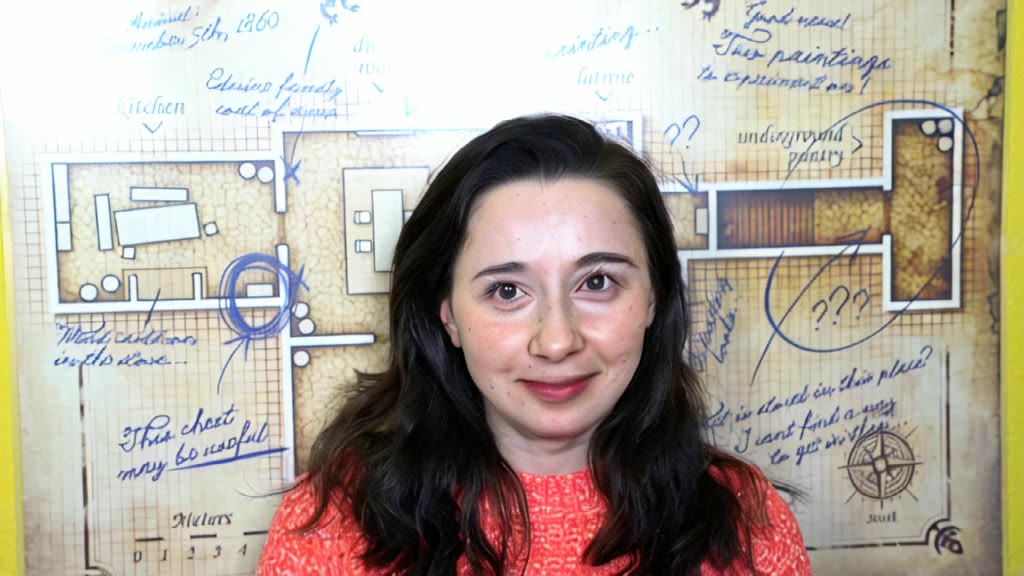}} \raisebox{-.5\height}{\includegraphics[width=0.09\textwidth]{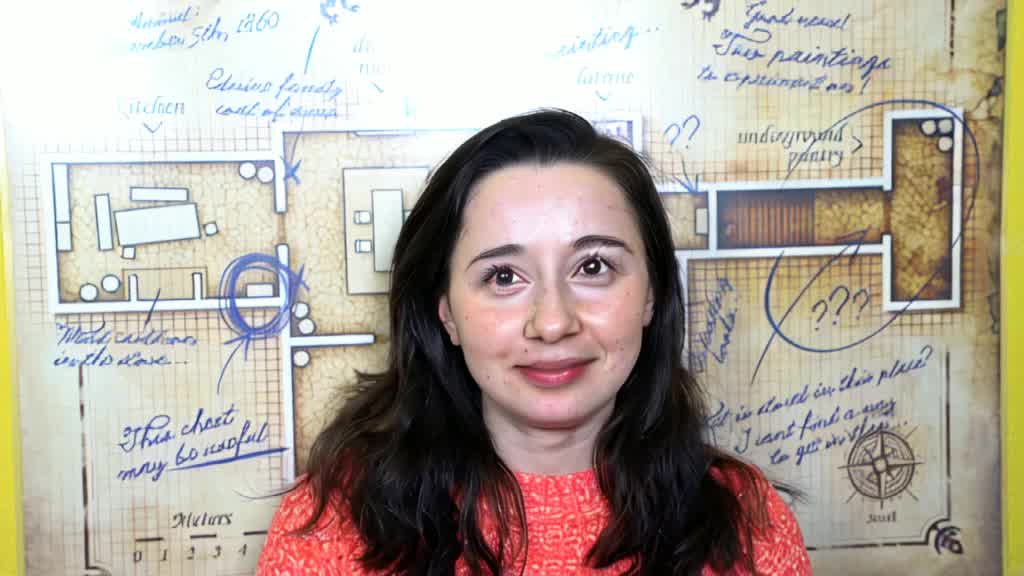}} \raisebox{-.5\height}{\includegraphics[width=0.09\textwidth]{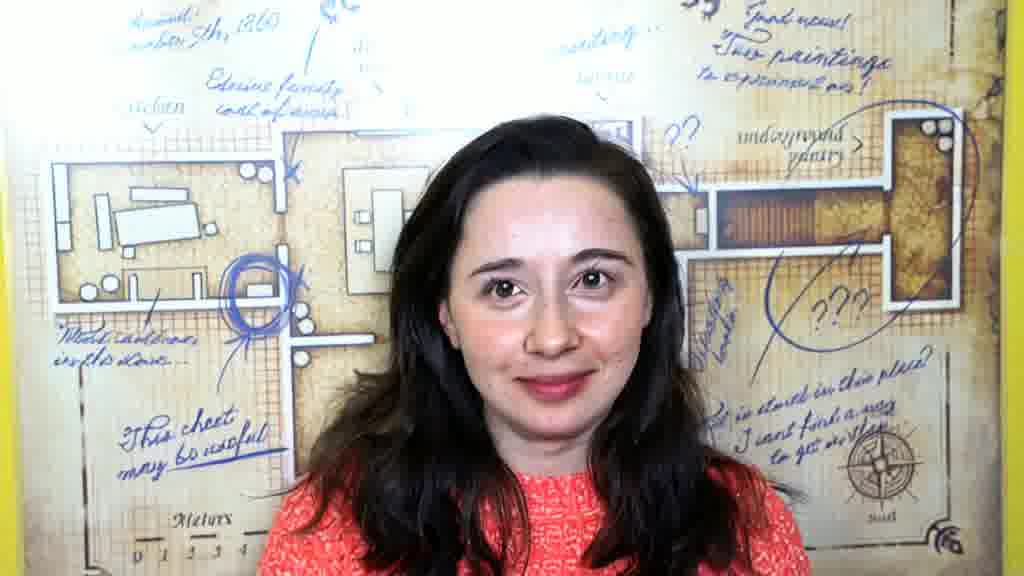}} & \raisebox{-.5\height}{\includegraphics[width=0.09\textwidth]{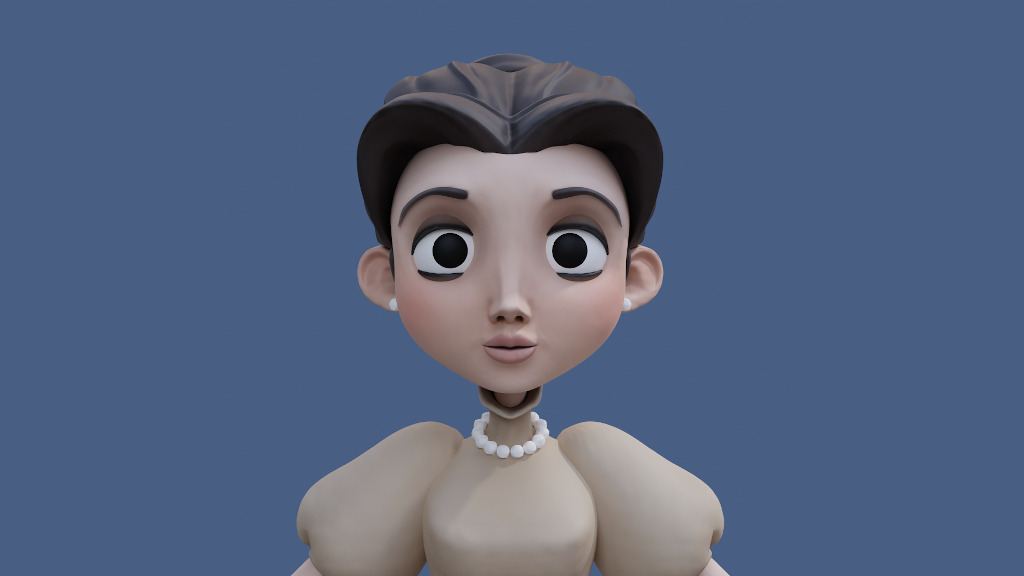}} & \raisebox{-.5\height}{\includegraphics[width=0.09\textwidth]{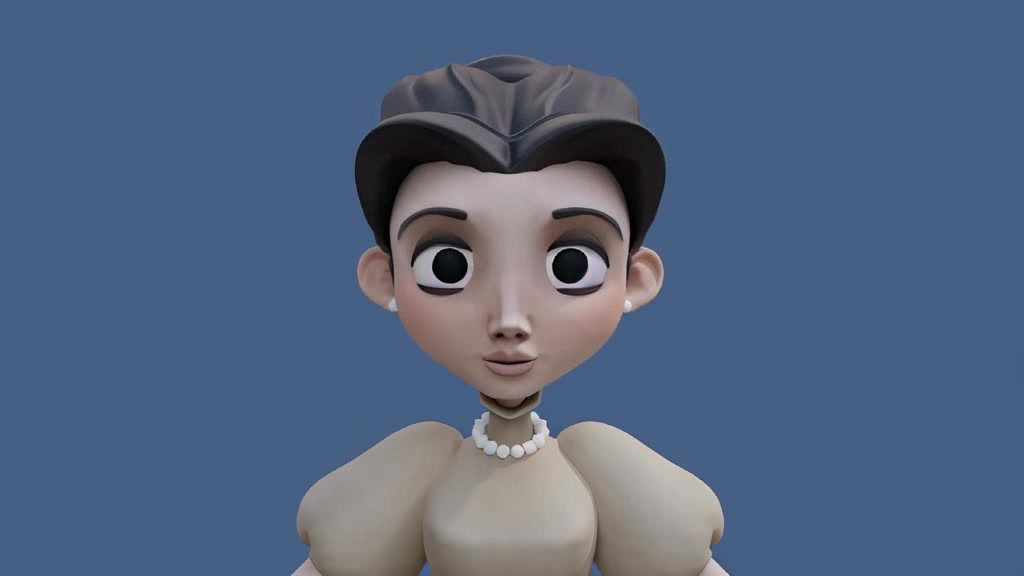}} \raisebox{-.5\height}{\includegraphics[width=0.09\textwidth]{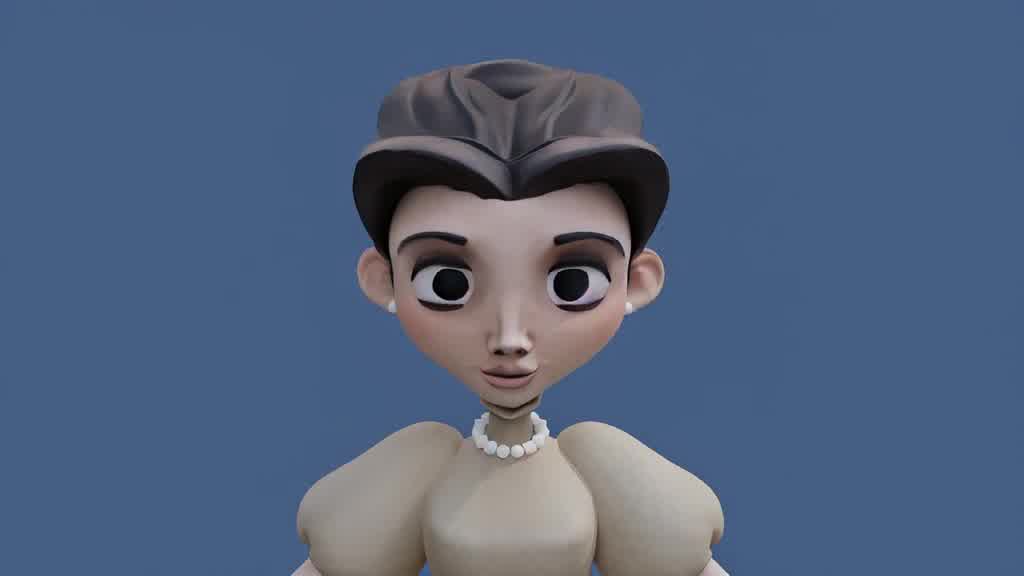}} \raisebox{-.5\height}{\includegraphics[width=0.09\textwidth]{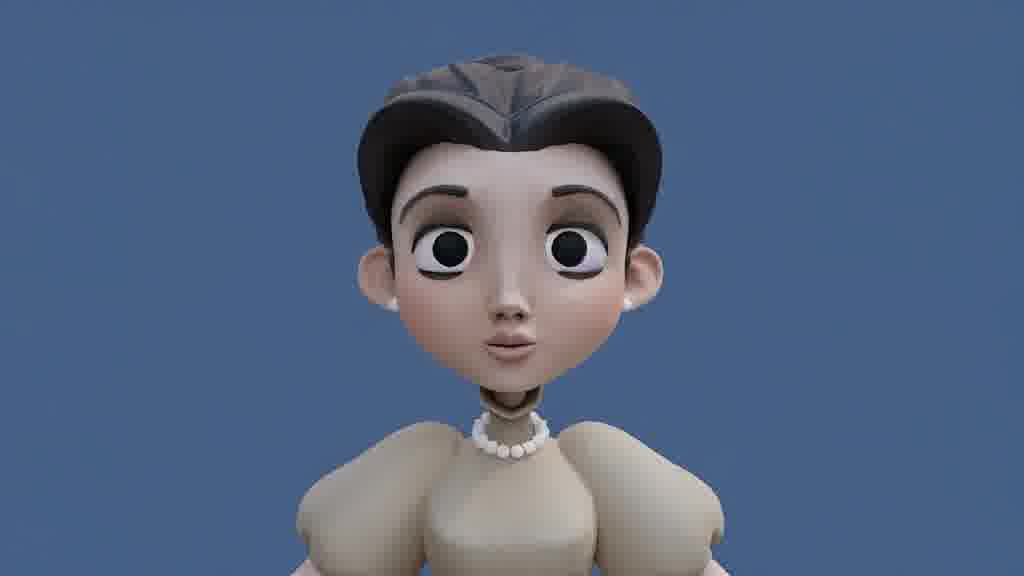}} \\
		{Gen. 2}  & \raisebox{-.5\height}{\includegraphics[width=0.09\textwidth]{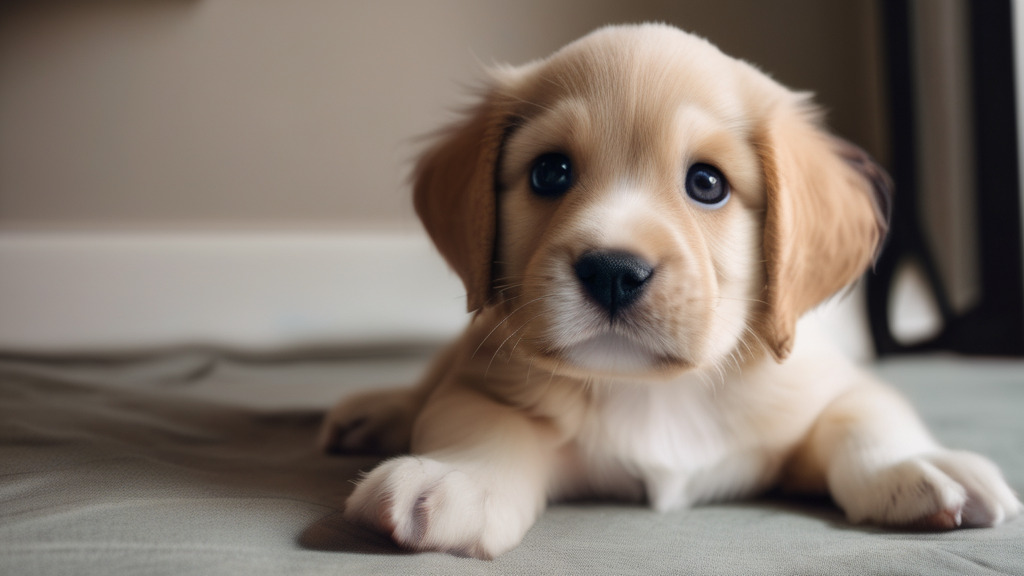}} & \raisebox{-.5\height}{\includegraphics[width=0.09\textwidth]{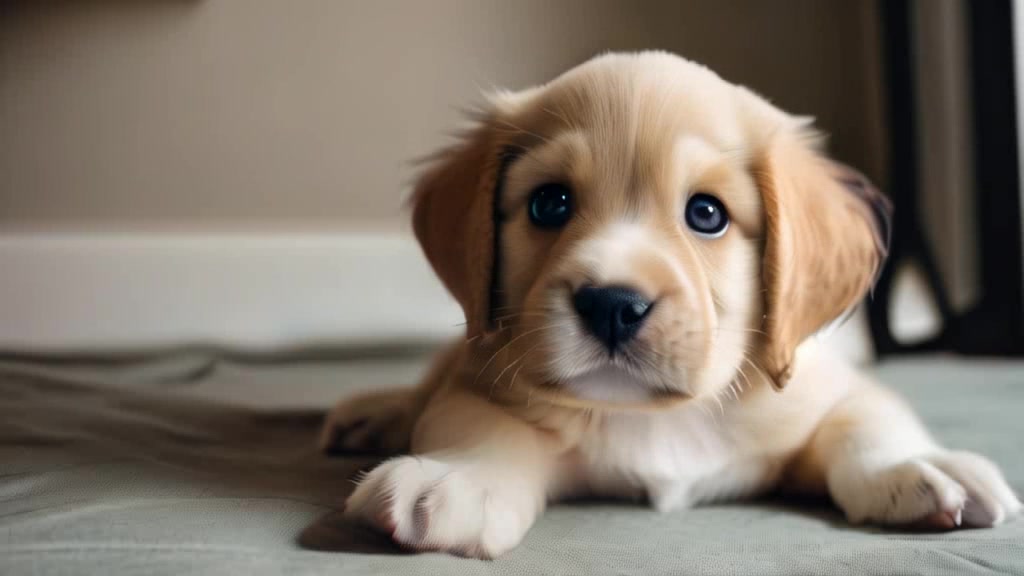}} \raisebox{-.5\height}{\includegraphics[width=0.09\textwidth]{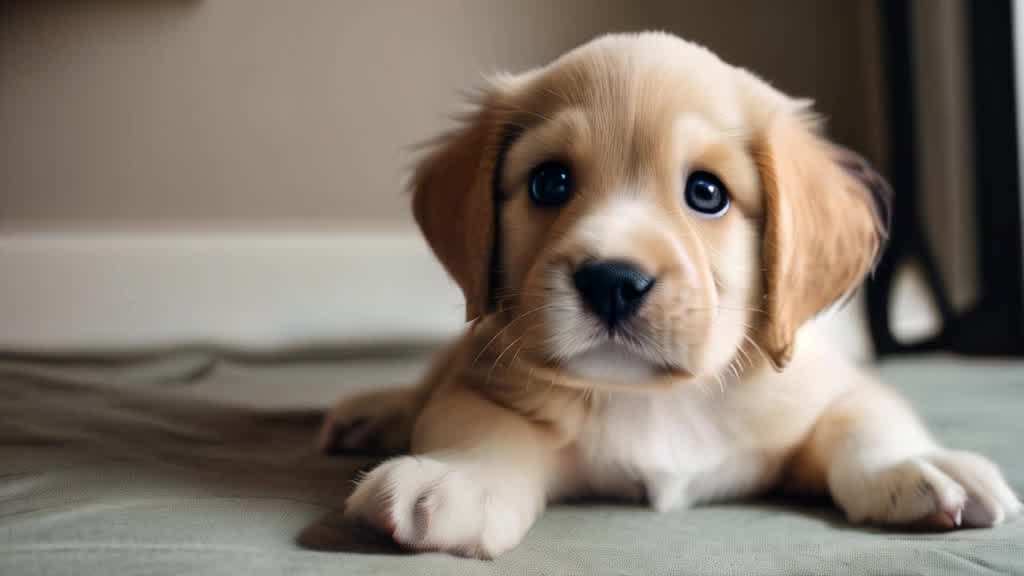}} \raisebox{-.5\height}{\includegraphics[width=0.09\textwidth]{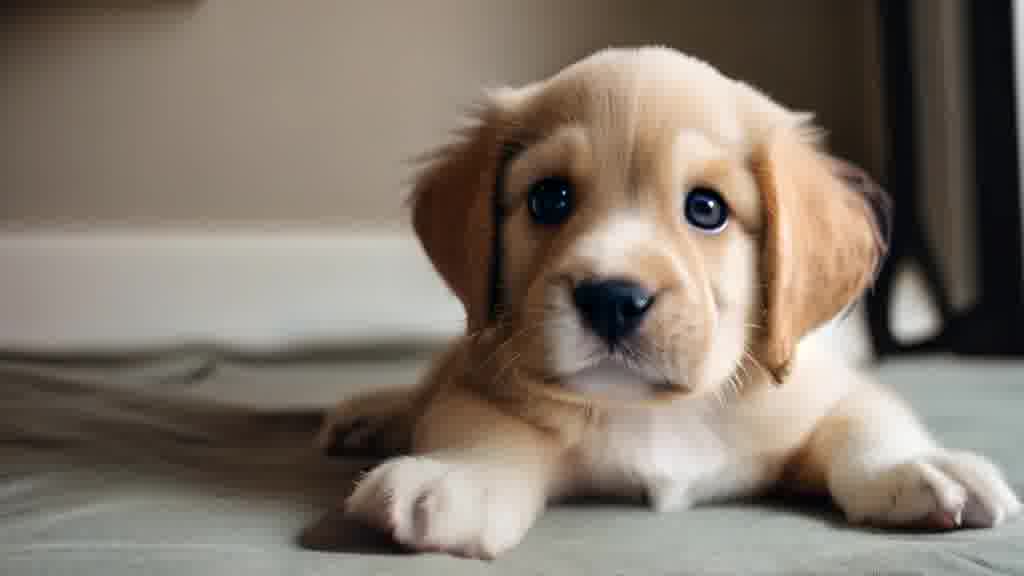}} & \raisebox{-.5\height}{\includegraphics[width=0.09\textwidth]{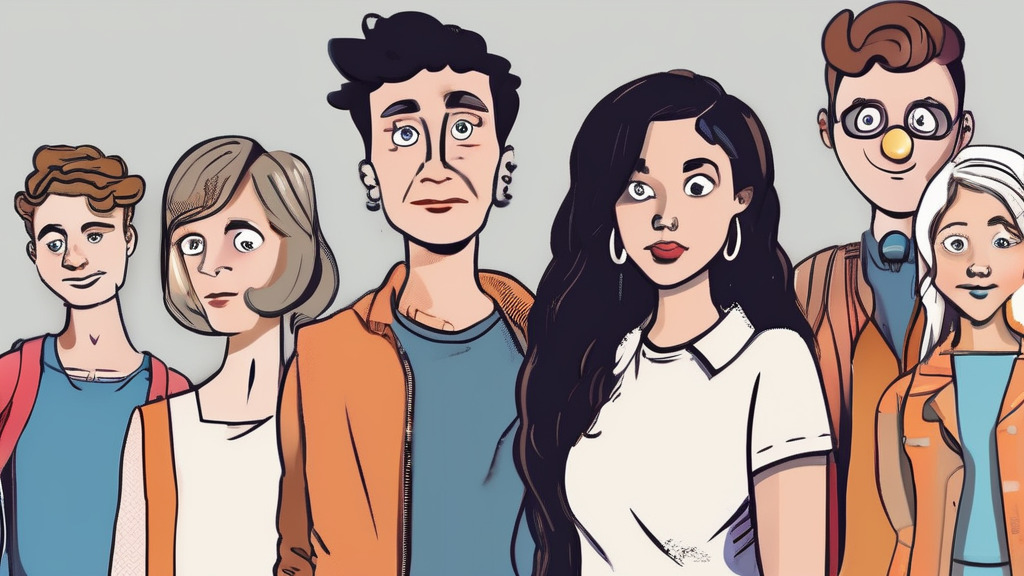}} & \raisebox{-.5\height}{\includegraphics[width=0.09\textwidth]{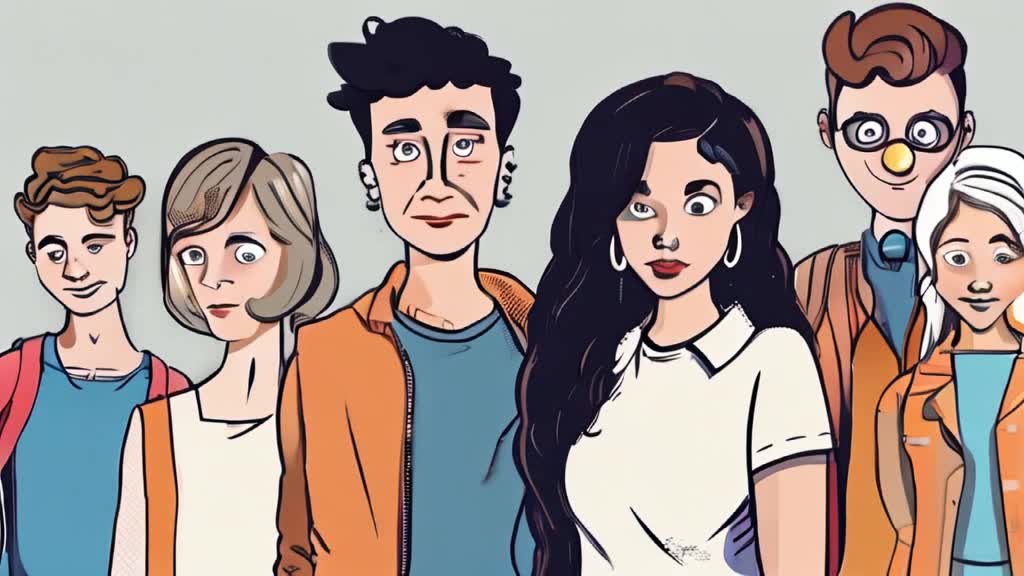}} \raisebox{-.5\height}{\includegraphics[width=0.09\textwidth]{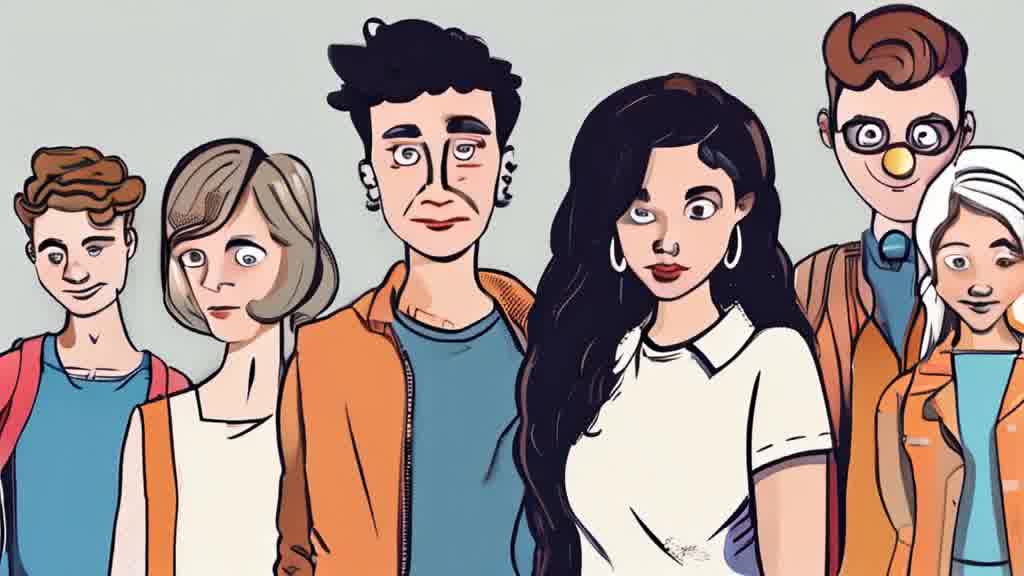}} \raisebox{-.5\height}{\includegraphics[width=0.09\textwidth]{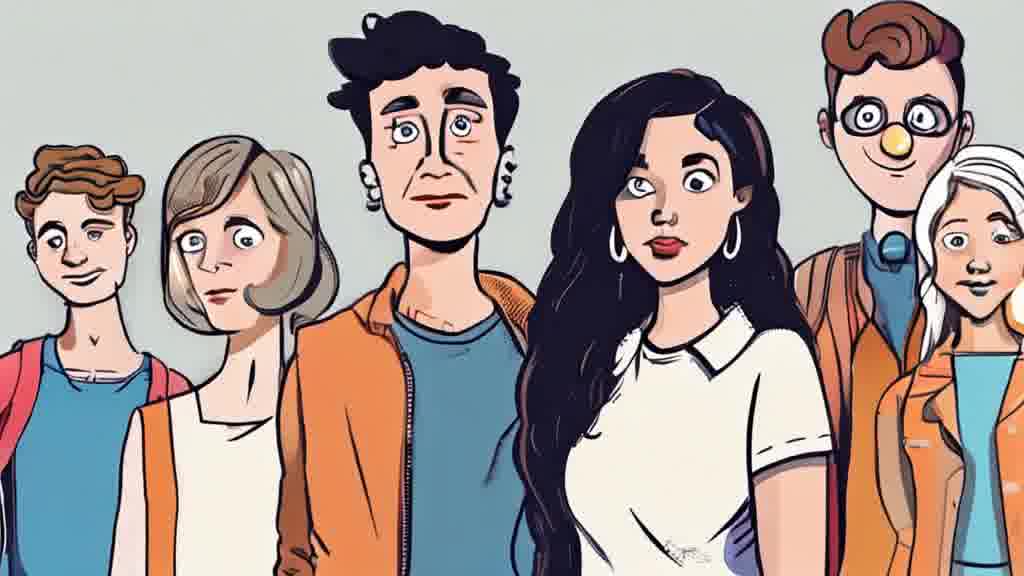}} \\
		{Ref.} & \raisebox{-.5\height}{\includegraphics[width=0.09\textwidth]{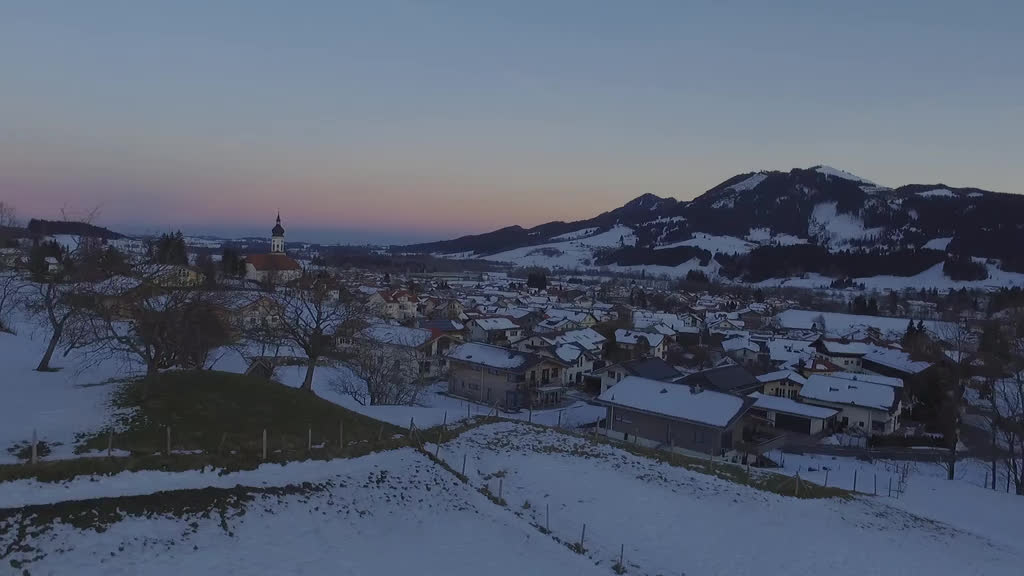}} & \raisebox{-.5\height}{\includegraphics[width=0.09\textwidth]{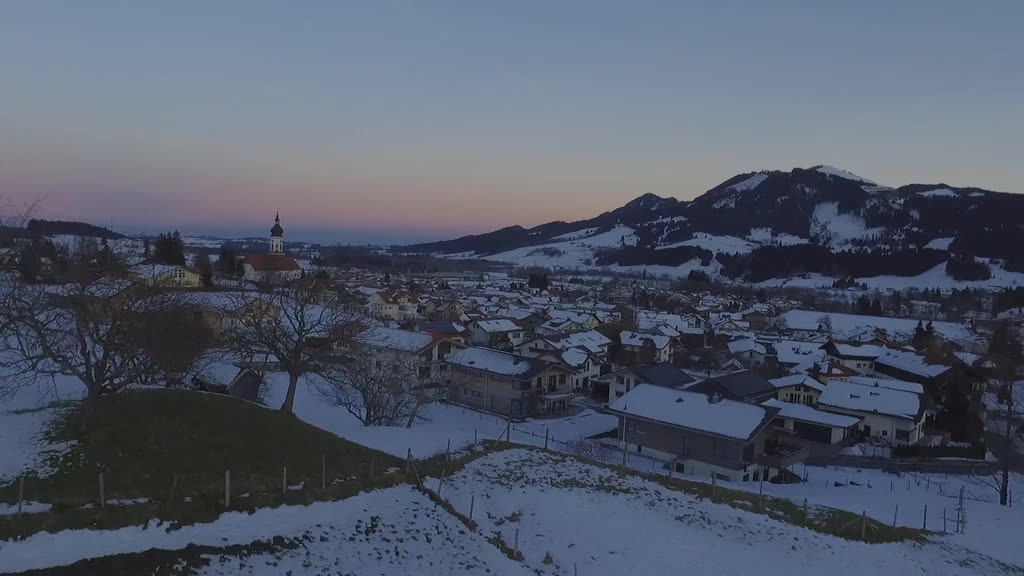}} \raisebox{-.5\height}{\includegraphics[width=0.09\textwidth]{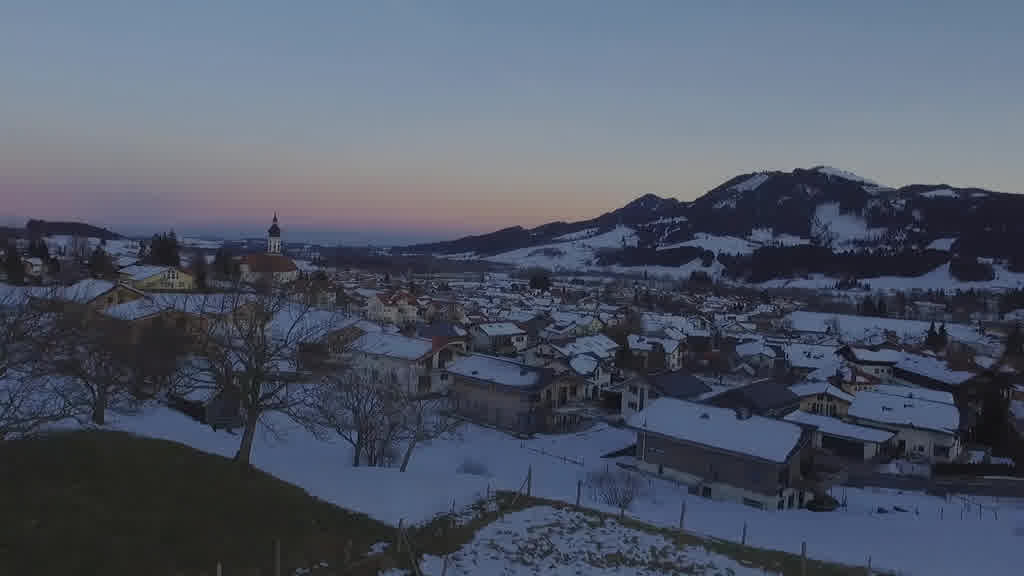}} \raisebox{-.5\height}{\includegraphics[width=0.09\textwidth]{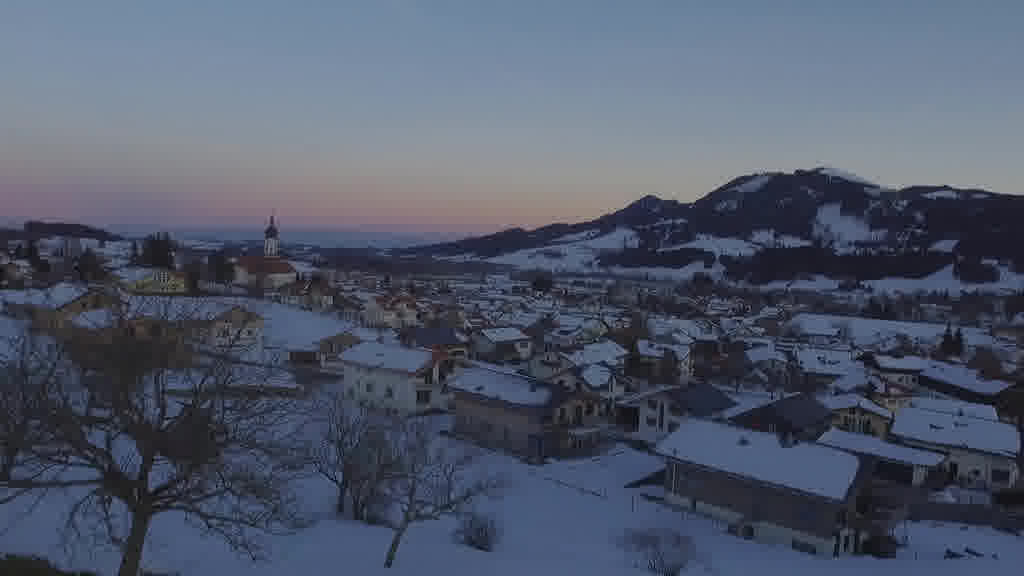}} & \raisebox{-.5\height}{\includegraphics[width=0.09\textwidth]{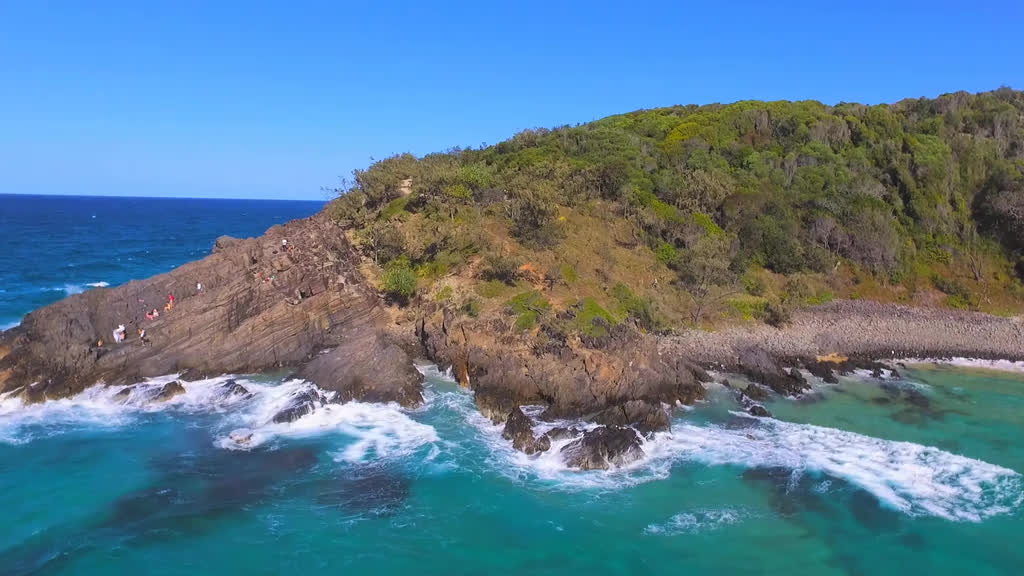}} & \raisebox{-.5\height}{\includegraphics[width=0.09\textwidth]{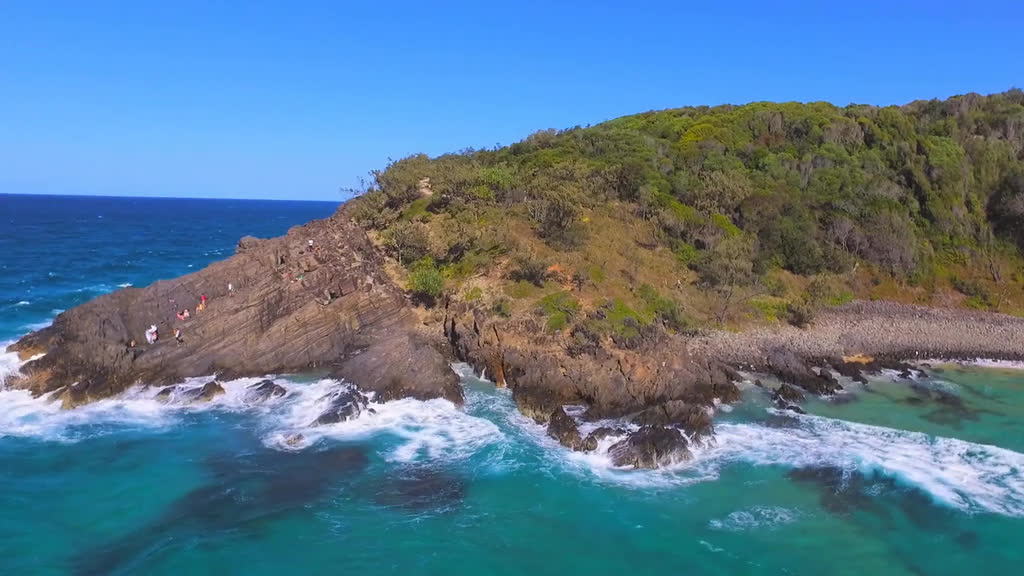}} \raisebox{-.5\height}{\includegraphics[width=0.09\textwidth]{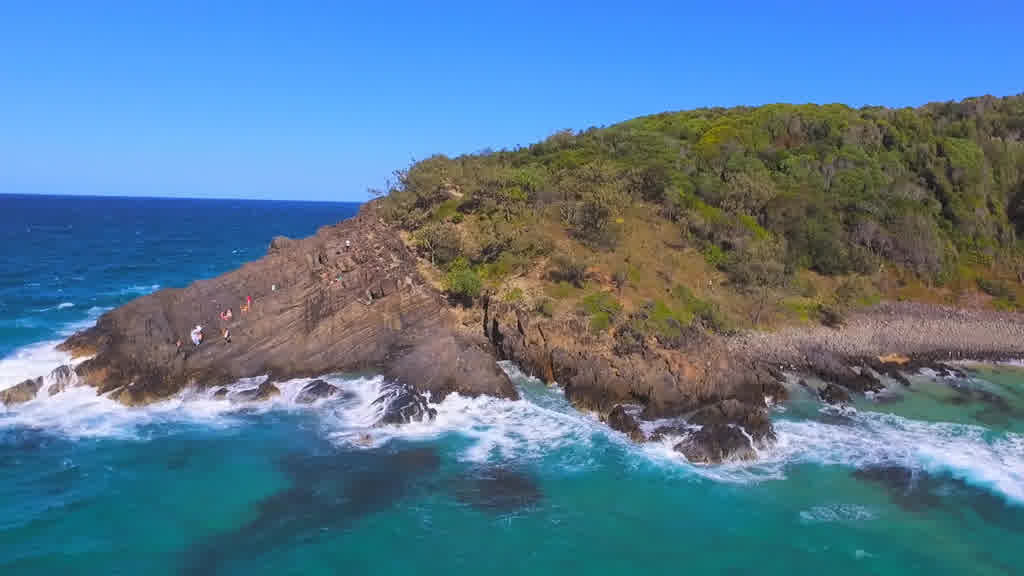}} \raisebox{-.5\height}{\includegraphics[width=0.09\textwidth]{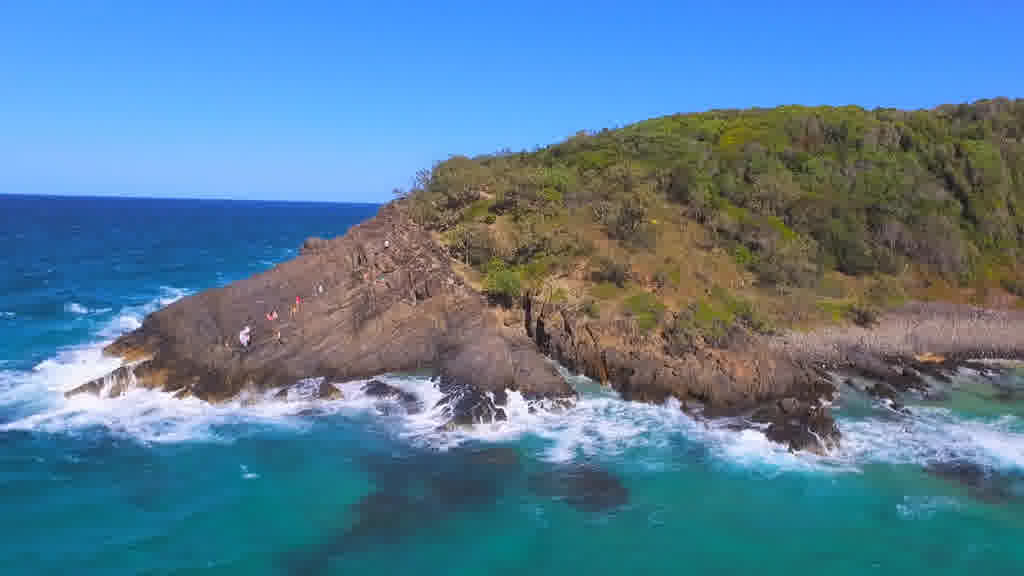}} \\
		{Gen. 1}  & \raisebox{-.5\height}{\includegraphics[width=0.09\textwidth]{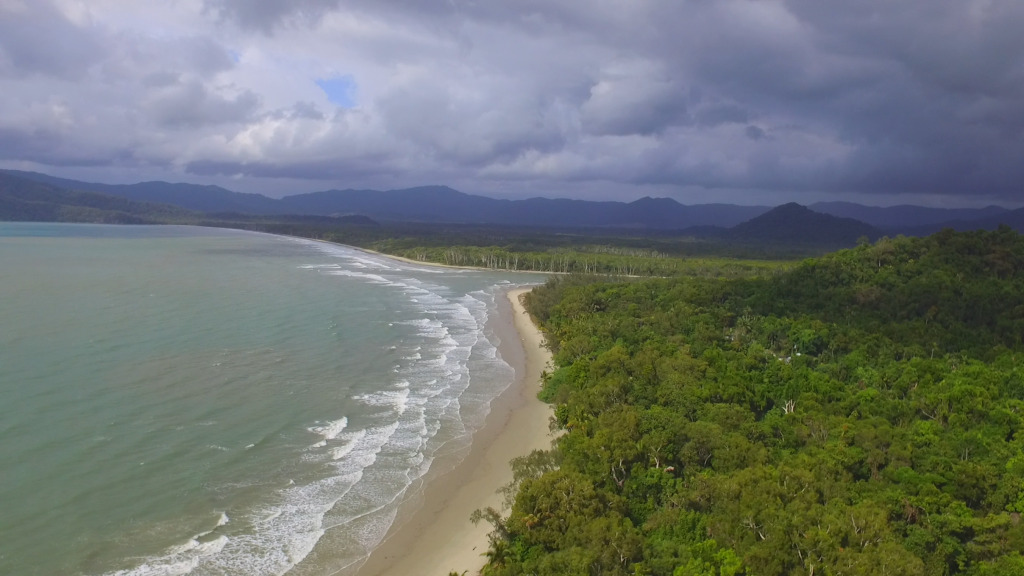}} & \raisebox{-.5\height}{\includegraphics[width=0.09\textwidth]{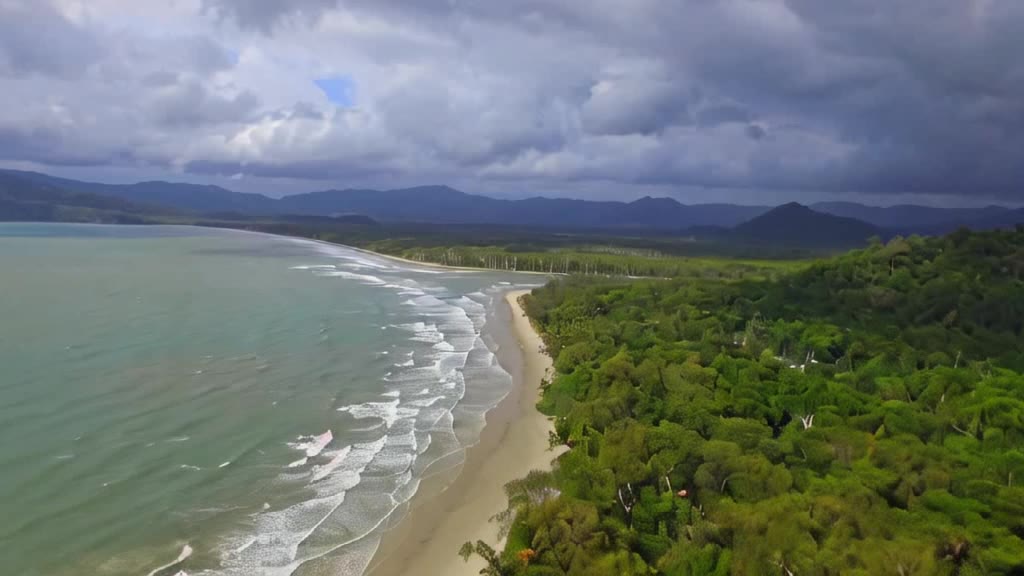}} \raisebox{-.5\height}{\includegraphics[width=0.09\textwidth]{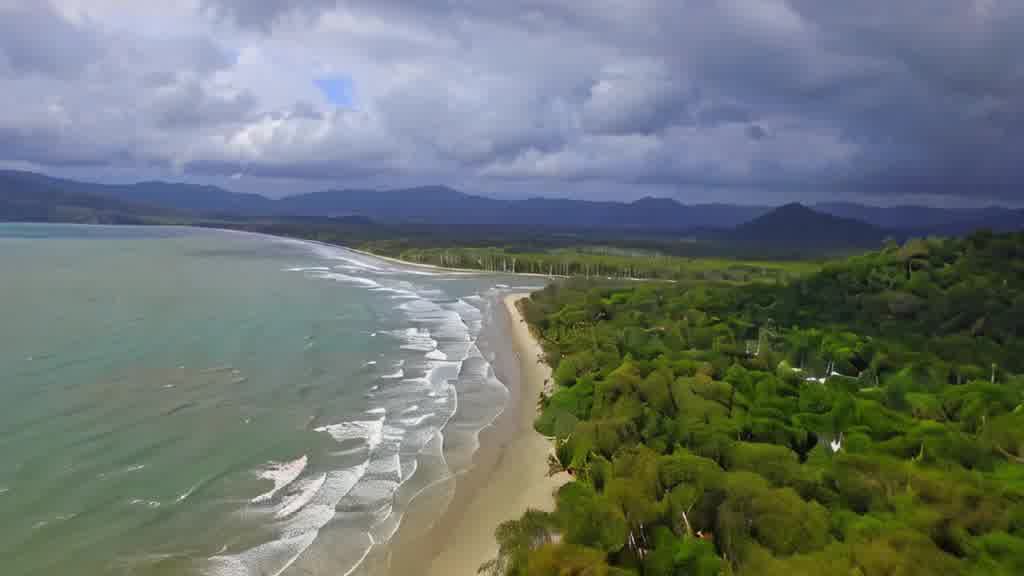}} \raisebox{-.5\height}{\includegraphics[width=0.09\textwidth]{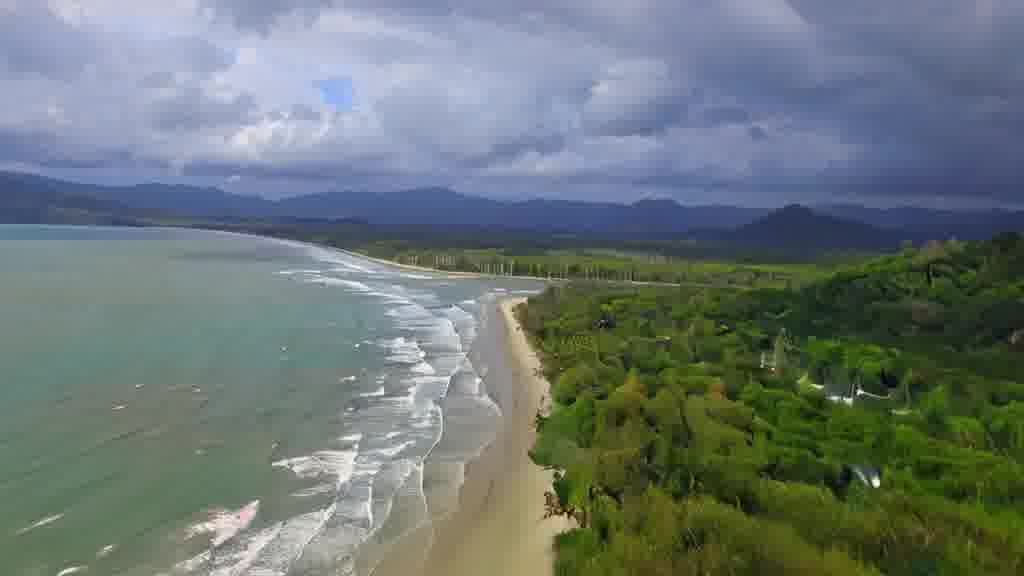}} & \raisebox{-.5\height}{\includegraphics[width=0.09\textwidth]{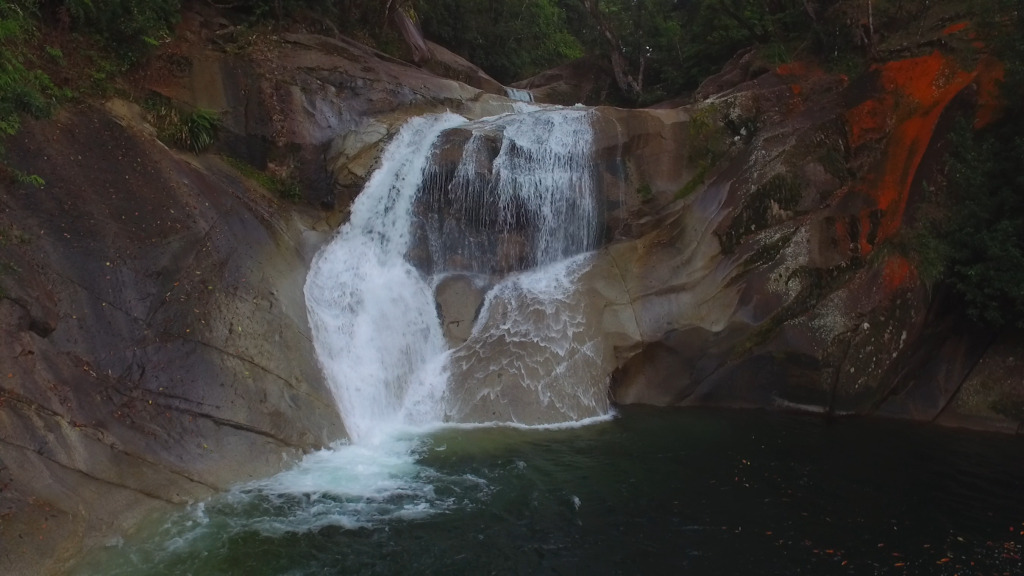}} & \raisebox{-.5\height}{\includegraphics[width=0.09\textwidth]{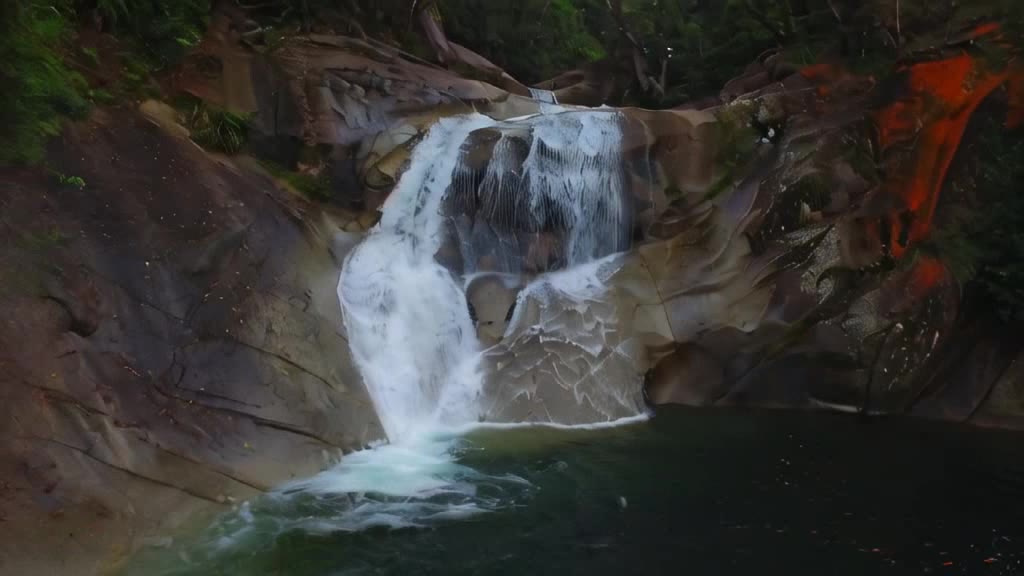}} \raisebox{-.5\height}{\includegraphics[width=0.09\textwidth]{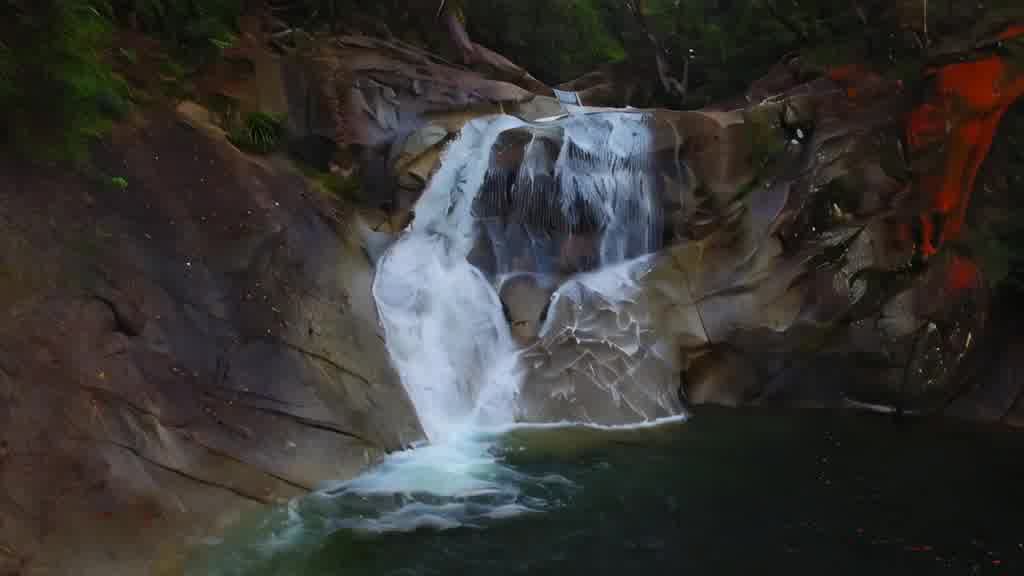}} \raisebox{-.5\height}{\includegraphics[width=0.09\textwidth]{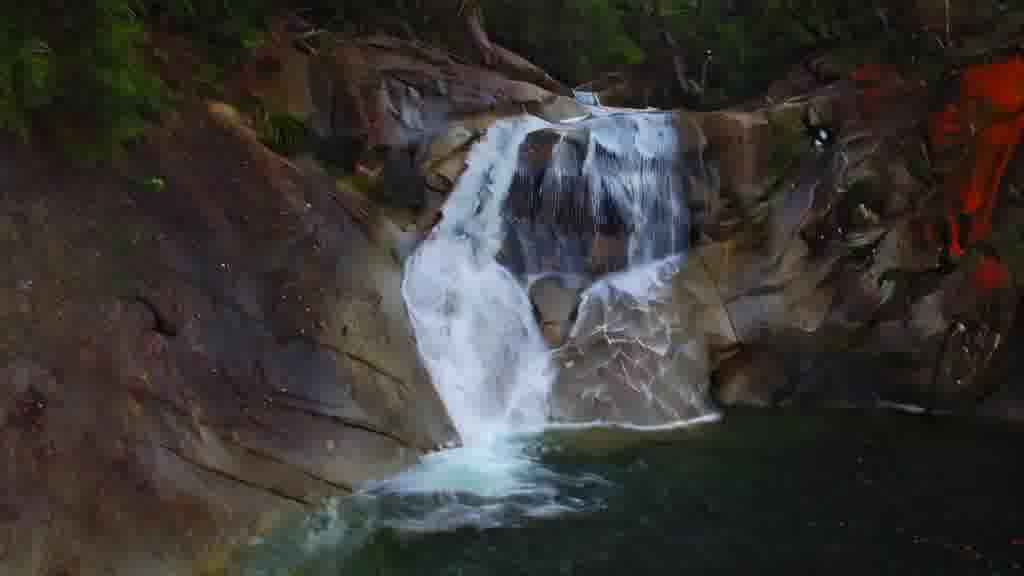}} \\
		{Gen. 2}  & \raisebox{-.5\height}{\includegraphics[width=0.09\textwidth]{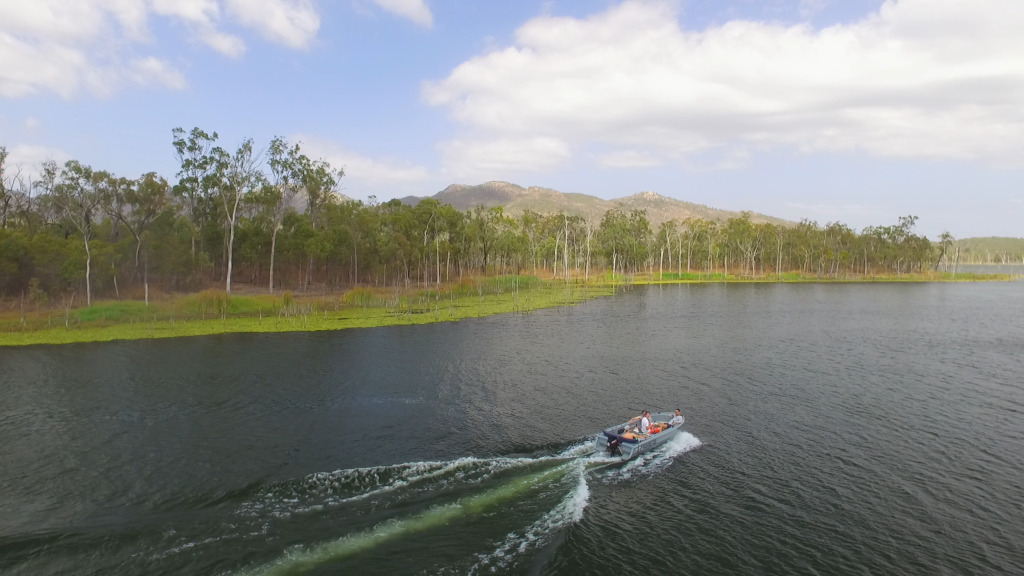}} & \raisebox{-.5\height}{\includegraphics[width=0.09\textwidth]{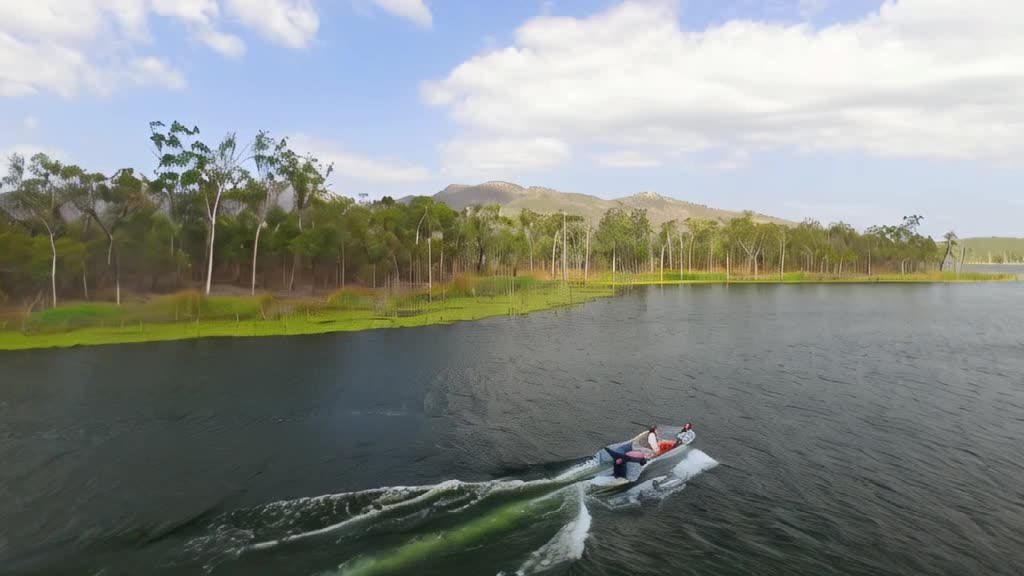}} \raisebox{-.5\height}{\includegraphics[width=0.09\textwidth]{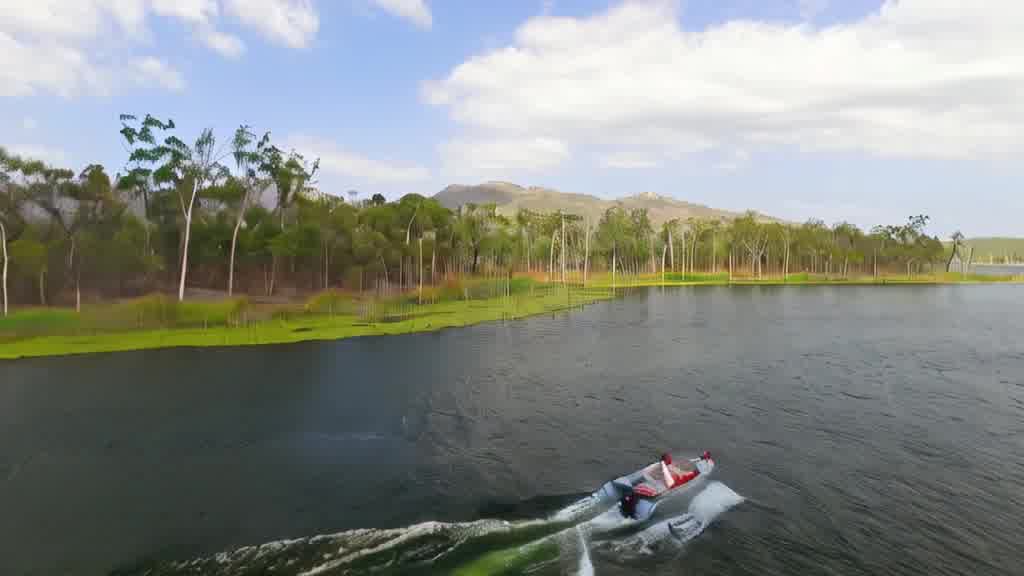}} \raisebox{-.5\height}{\includegraphics[width=0.09\textwidth]{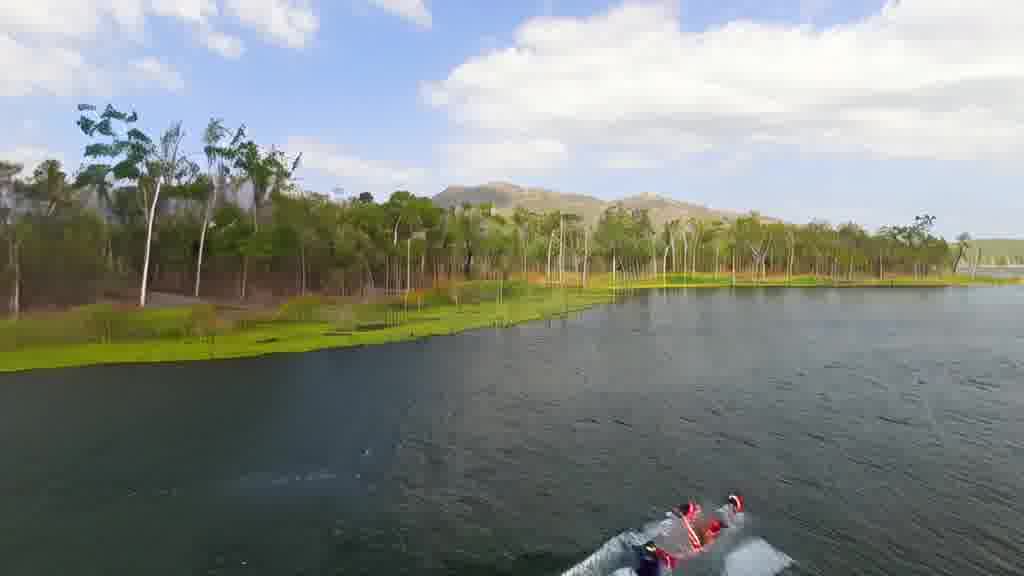}} & \raisebox{-.5\height}{\includegraphics[width=0.09\textwidth]{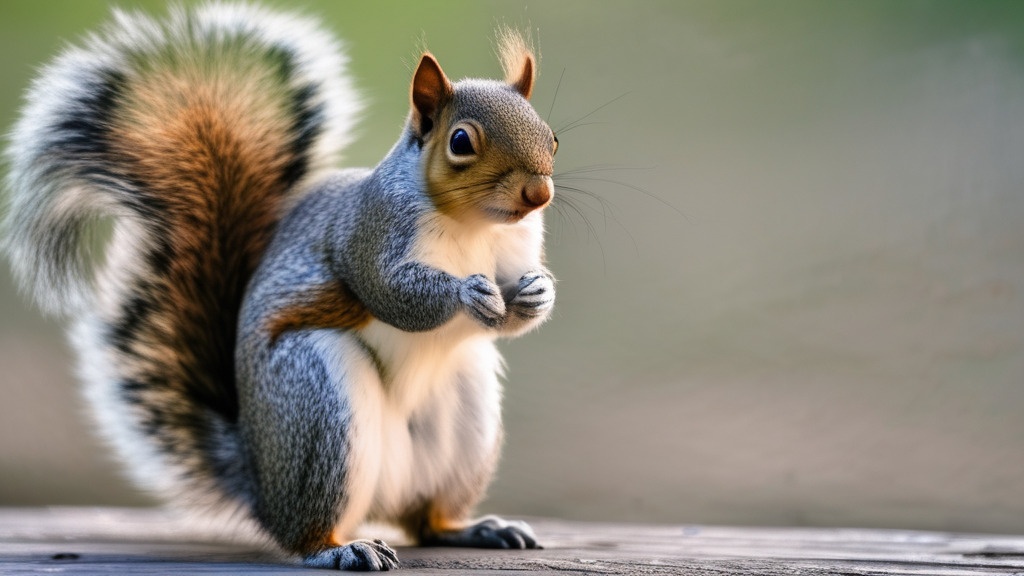}} & \raisebox{-.5\height}{\includegraphics[width=0.09\textwidth]{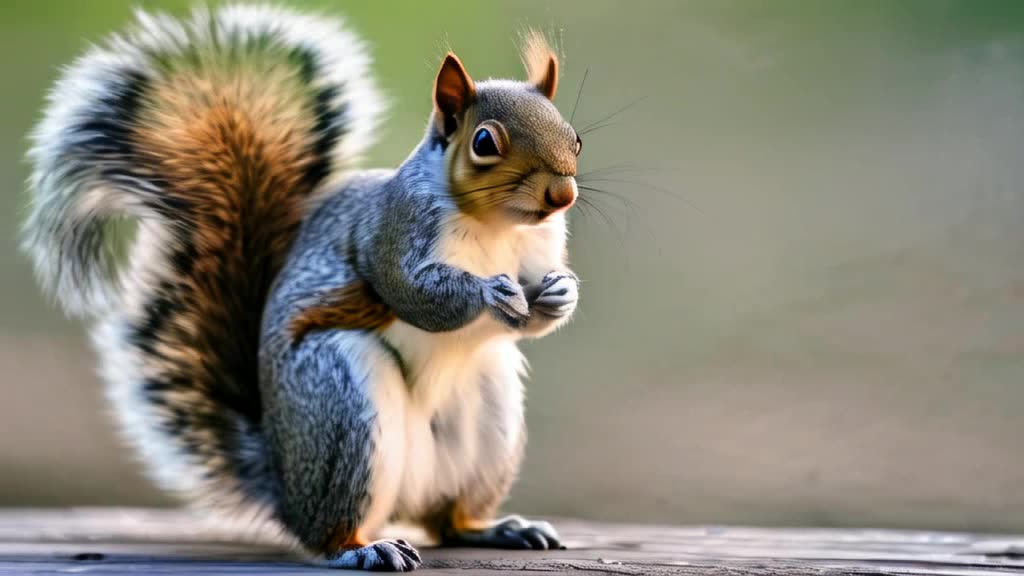}} \raisebox{-.5\height}{\includegraphics[width=0.09\textwidth]{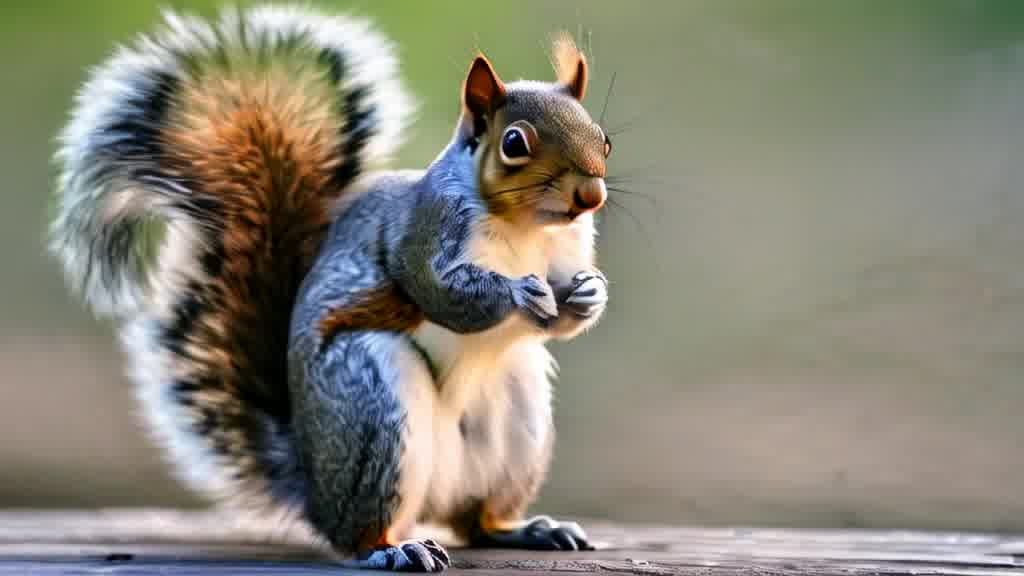}} \raisebox{-.5\height}{\includegraphics[width=0.09\textwidth]{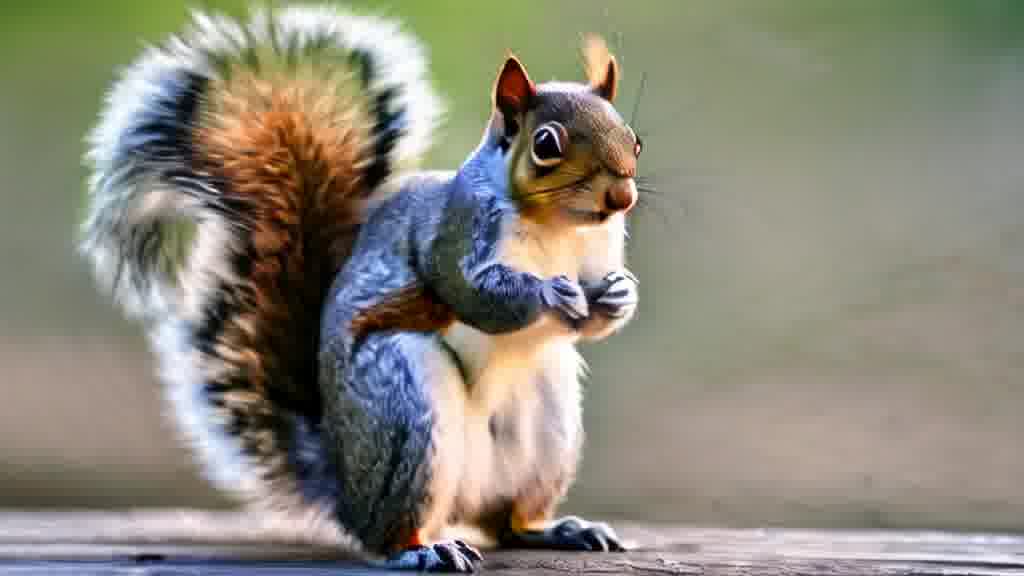}} \\
		{Ref.} & \raisebox{-.5\height}{\includegraphics[width=0.09\textwidth]{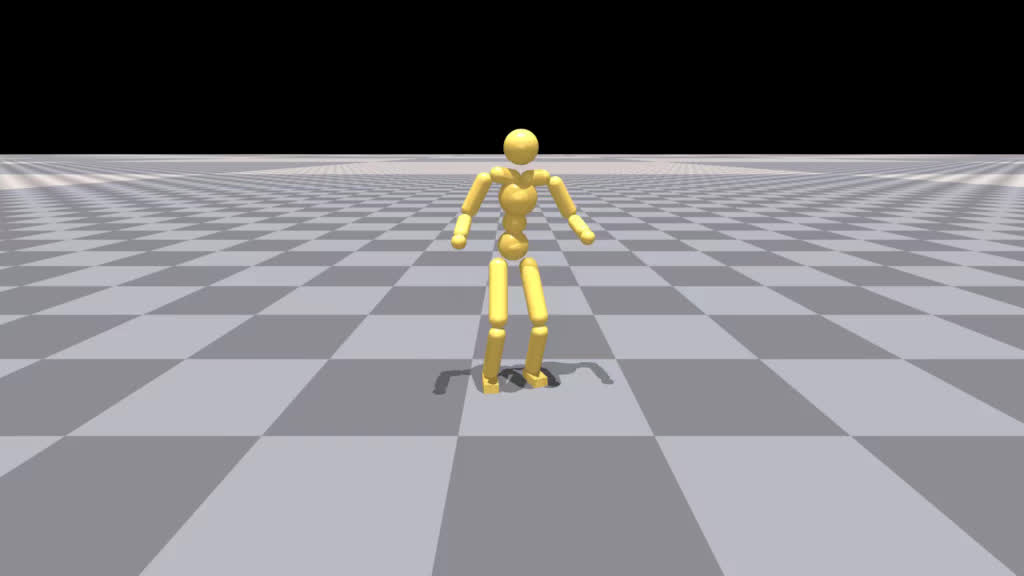}} & \raisebox{-.5\height}{\includegraphics[width=0.09\textwidth]{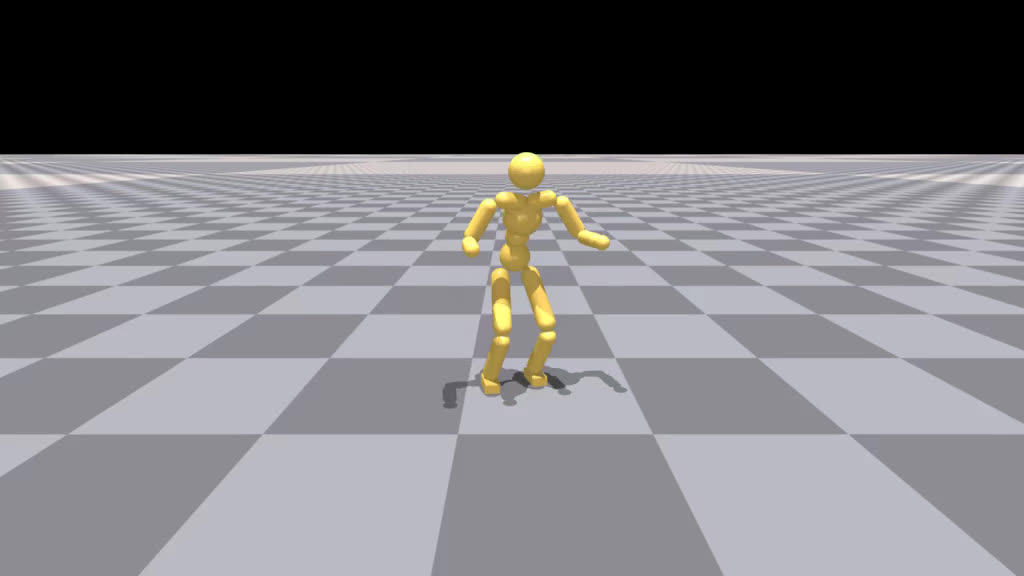}} \raisebox{-.5\height}{\includegraphics[width=0.09\textwidth]{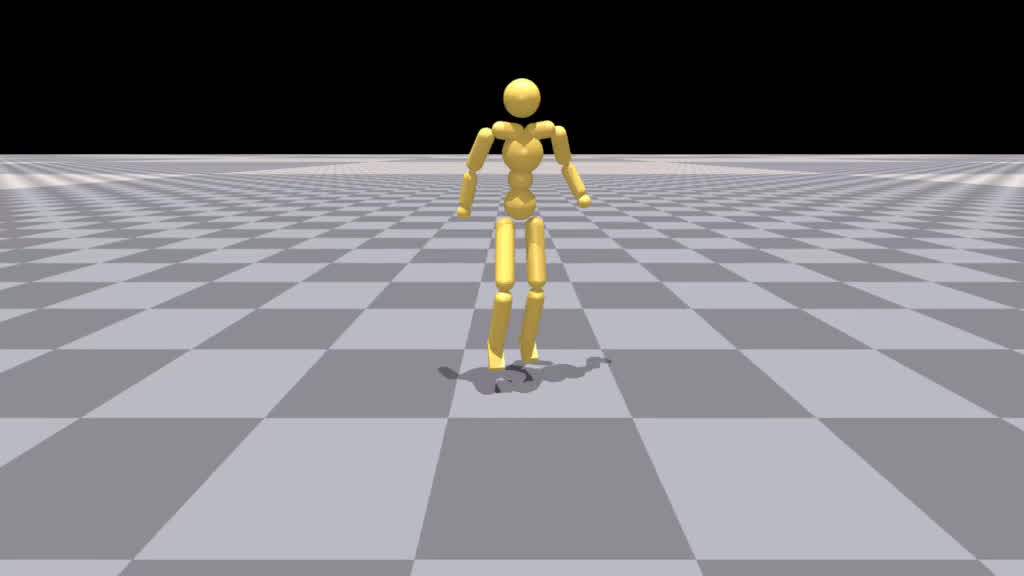}} \raisebox{-.5\height}{\includegraphics[width=0.09\textwidth]{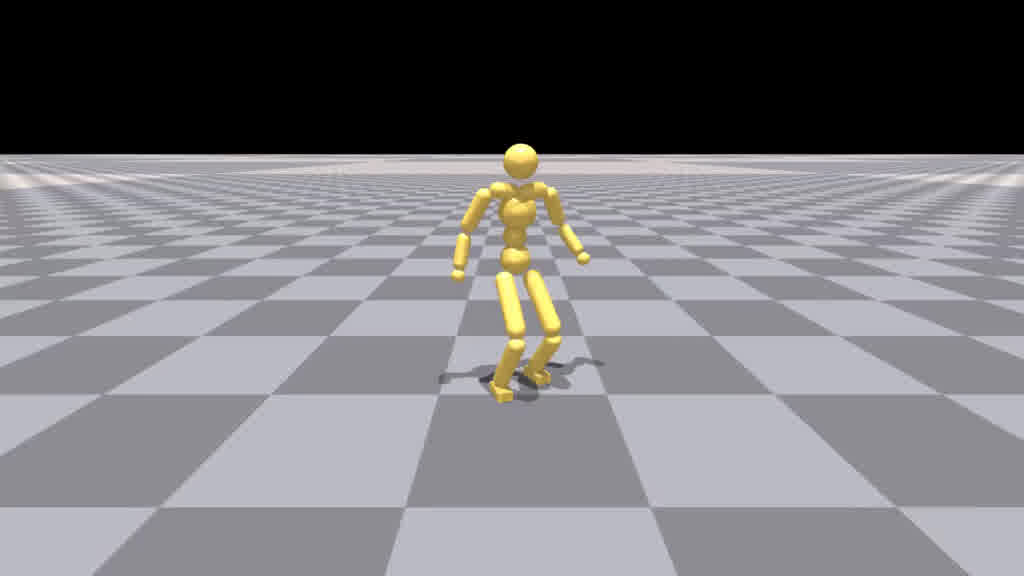}} & \raisebox{-.5\height}{\includegraphics[width=0.09\textwidth]{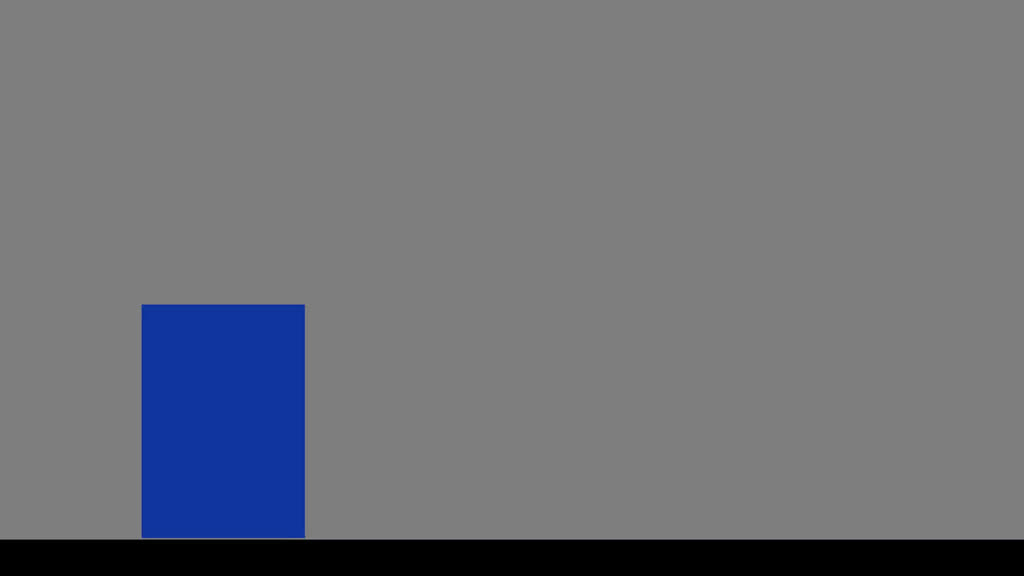}} & \raisebox{-.5\height}{\includegraphics[width=0.09\textwidth]{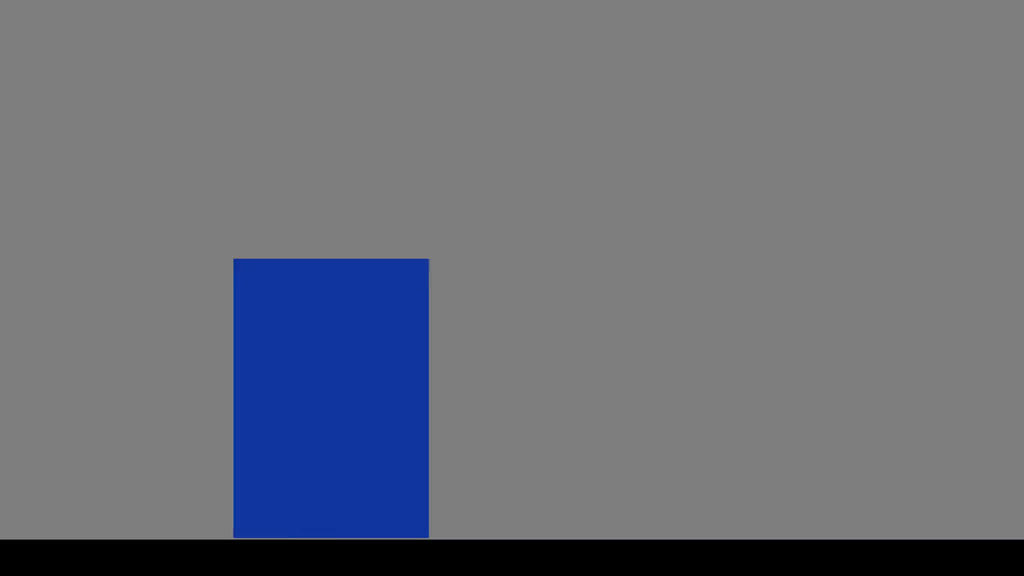}} \raisebox{-.5\height}{\includegraphics[width=0.09\textwidth]{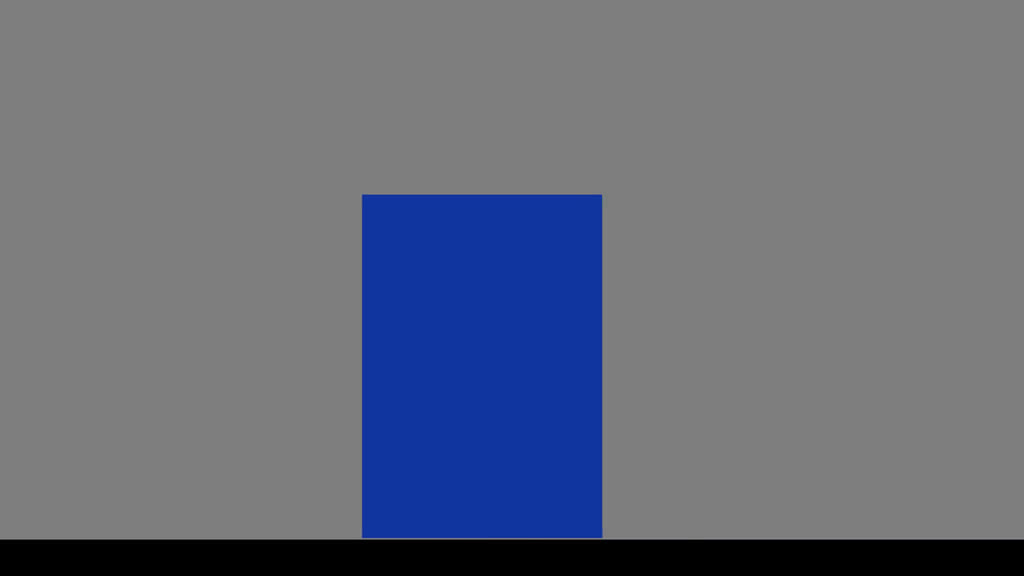}} \raisebox{-.5\height}{\includegraphics[width=0.09\textwidth]{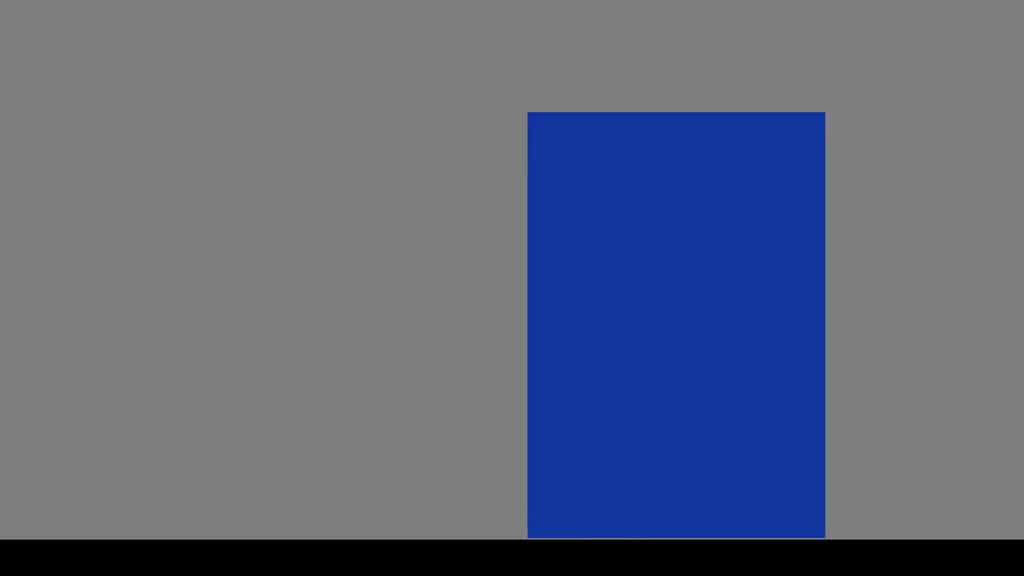}} \\
		{Gen. 1}  & \raisebox{-.5\height}{\includegraphics[width=0.09\textwidth]{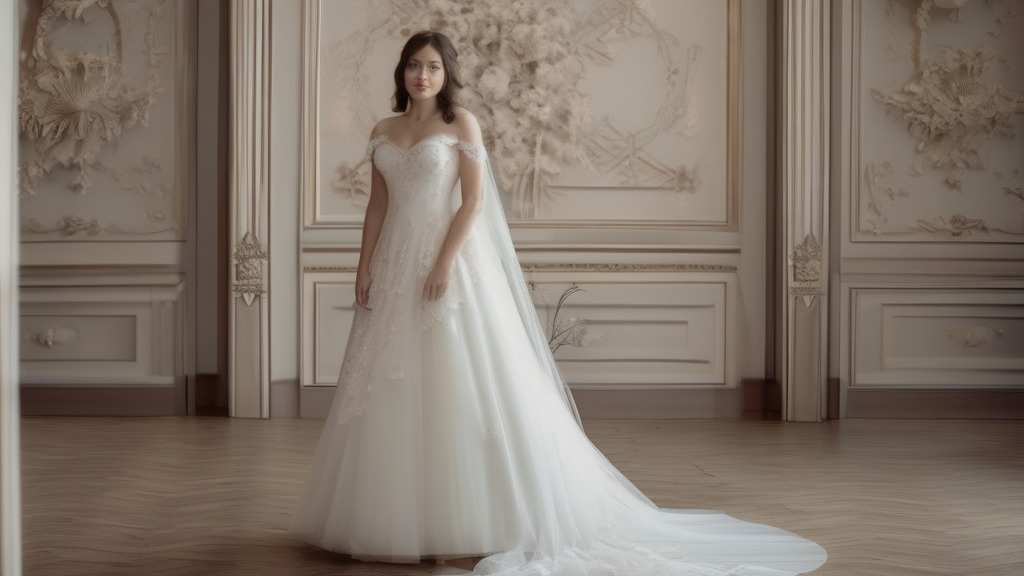}} & \raisebox{-.5\height}{\includegraphics[width=0.09\textwidth]{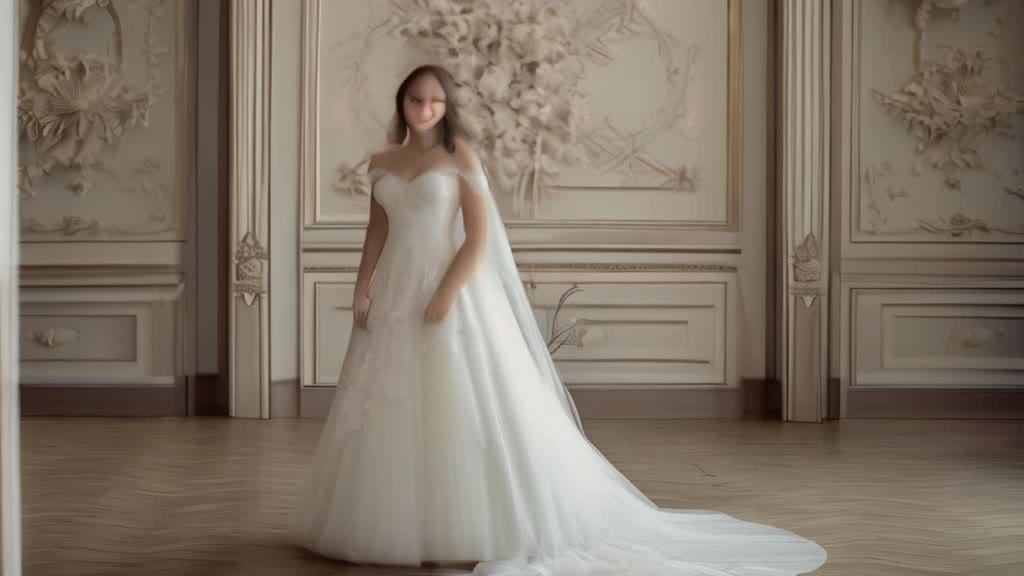}} \raisebox{-.5\height}{\includegraphics[width=0.09\textwidth]{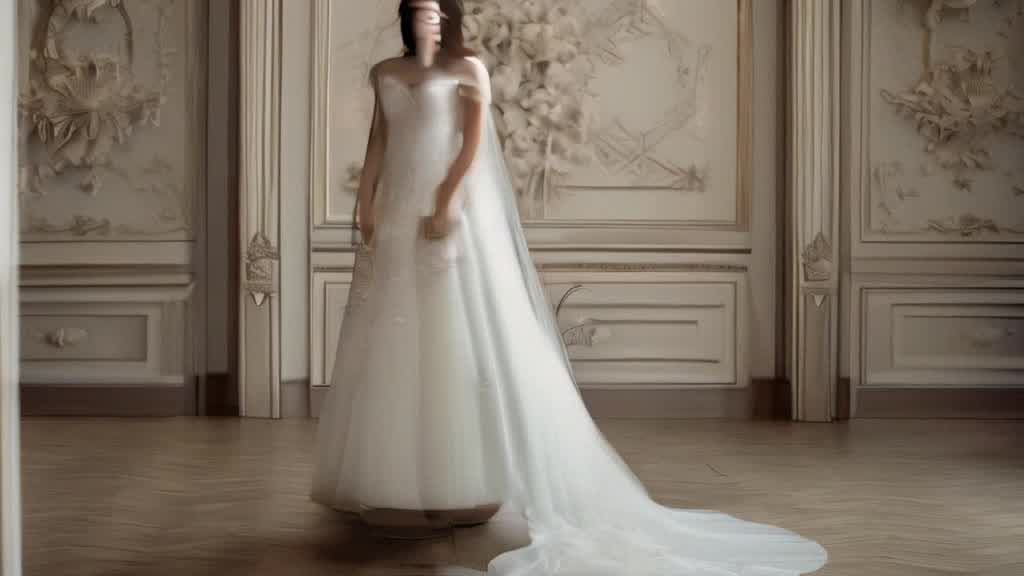}} \raisebox{-.5\height}{\includegraphics[width=0.09\textwidth]{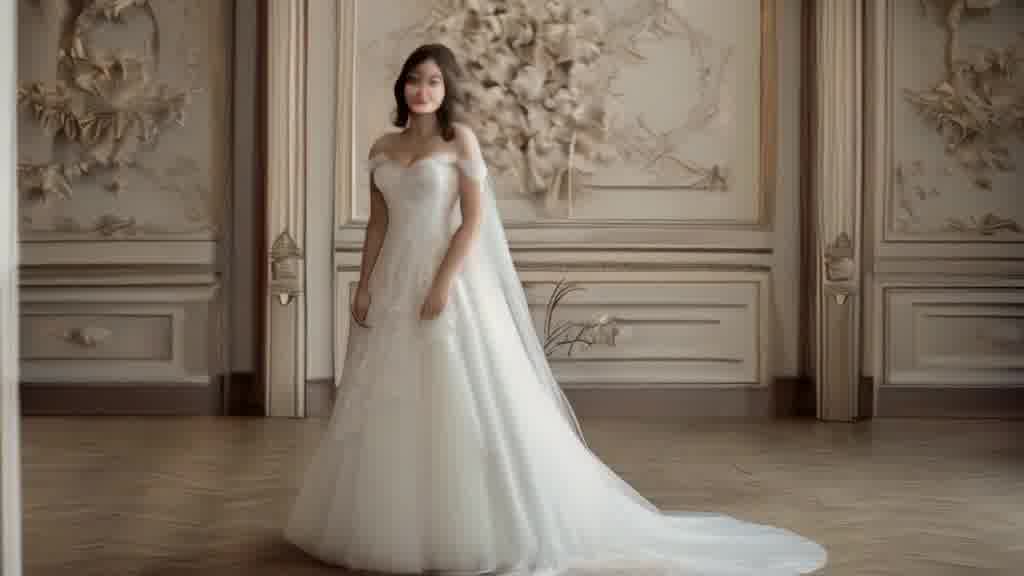}} & \raisebox{-.5\height}{\includegraphics[width=0.09\textwidth]{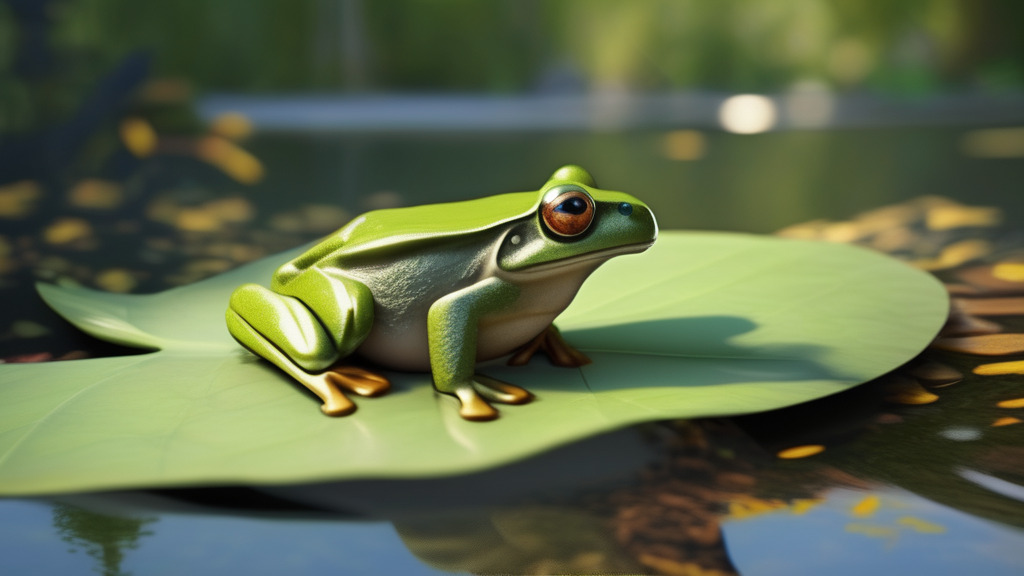}} & \raisebox{-.5\height}{\includegraphics[width=0.09\textwidth]{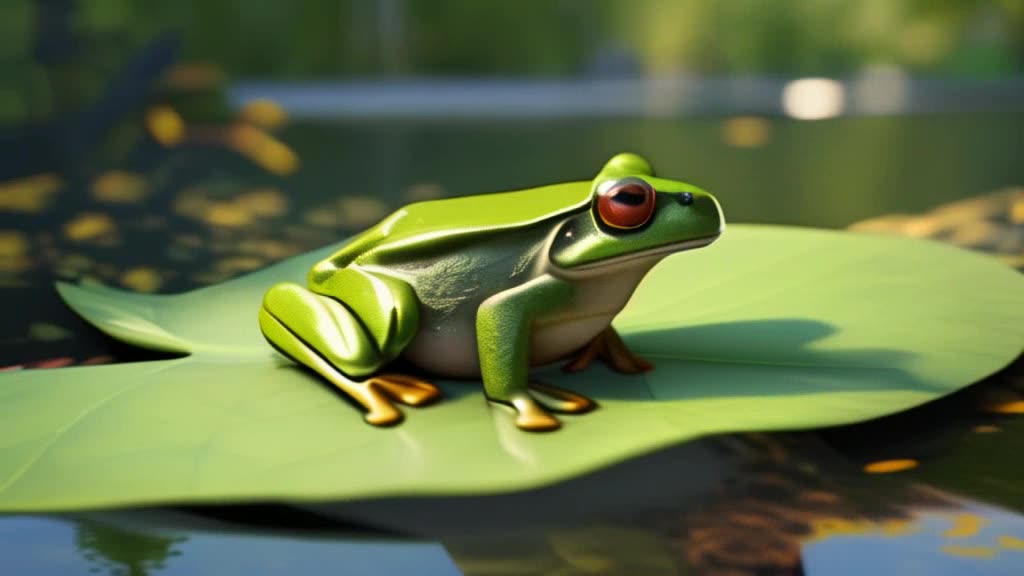}} \raisebox{-.5\height}{\includegraphics[width=0.09\textwidth]{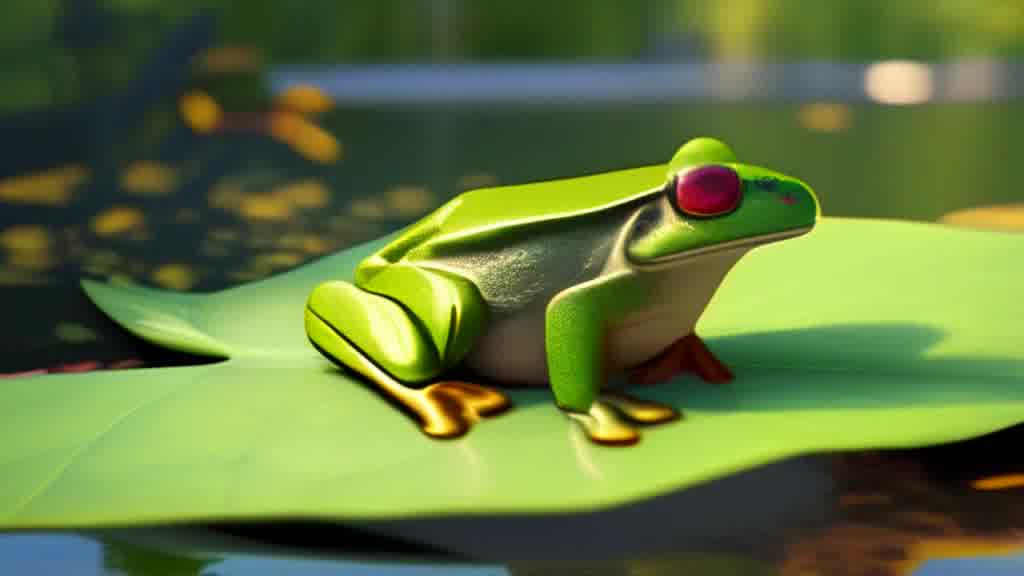}} \raisebox{-.5\height}{\includegraphics[width=0.09\textwidth]{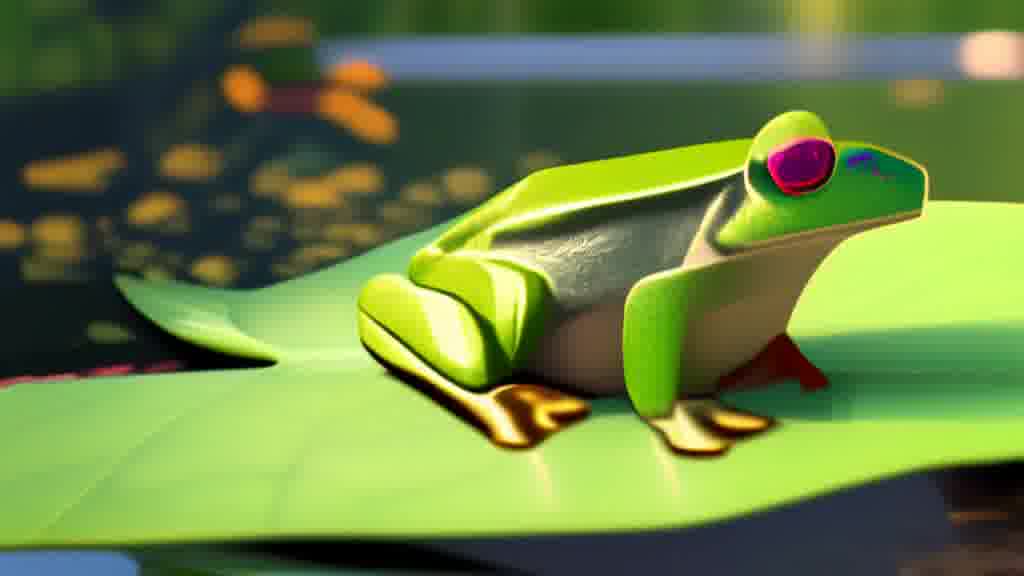}} \\
		{Gen. 2}  & \raisebox{-.5\height}{\includegraphics[width=0.09\textwidth]{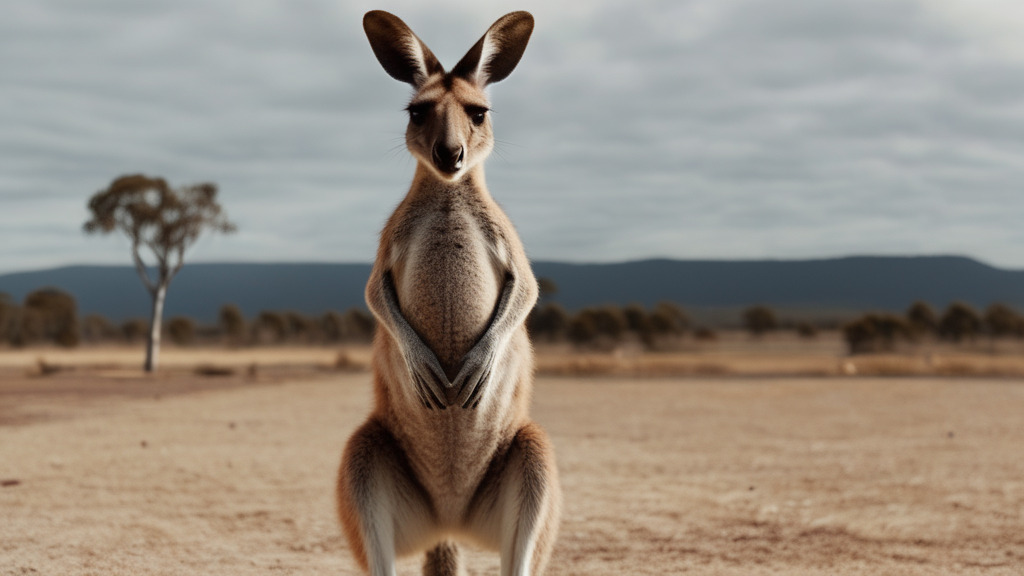}} & \raisebox{-.5\height}{\includegraphics[width=0.09\textwidth]{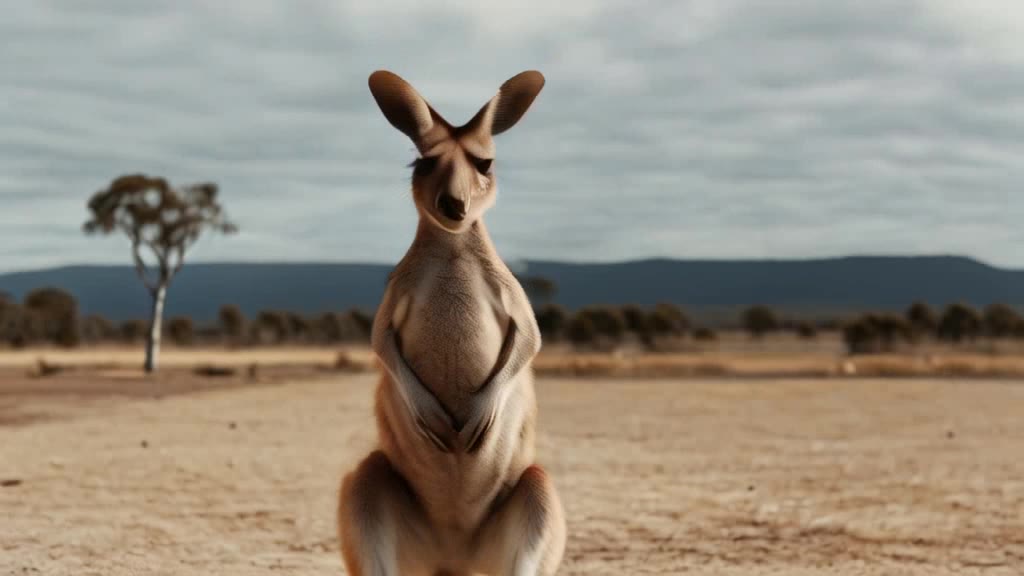}} \raisebox{-.5\height}{\includegraphics[width=0.09\textwidth]{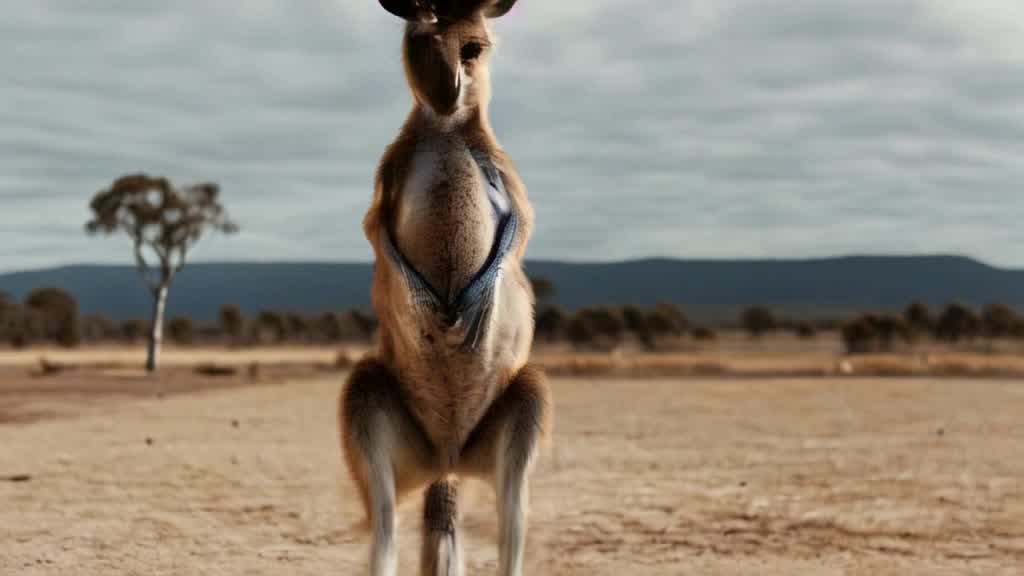}} \raisebox{-.5\height}{\includegraphics[width=0.09\textwidth]{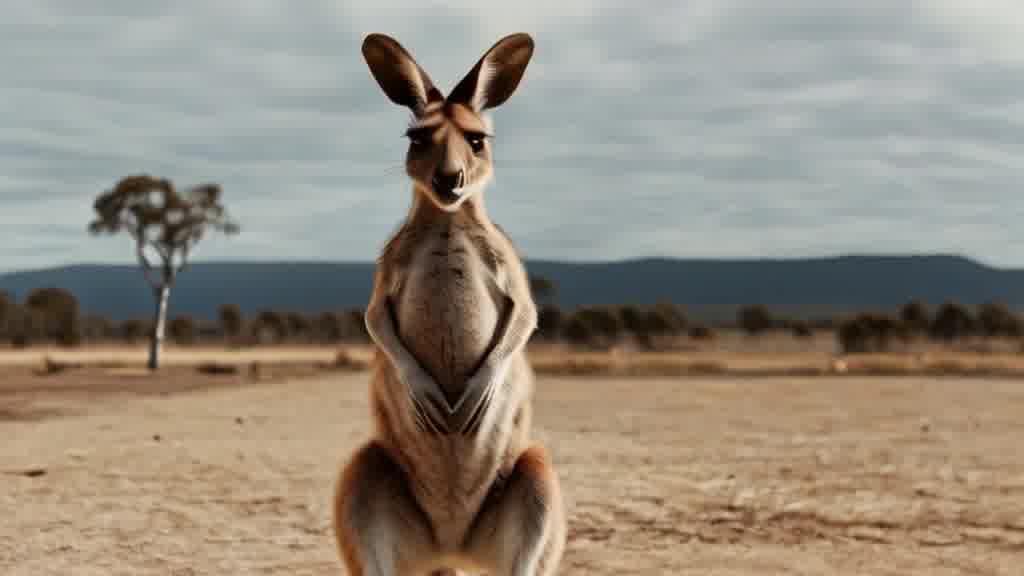}} & \raisebox{-.5\height}{\includegraphics[width=0.09\textwidth]{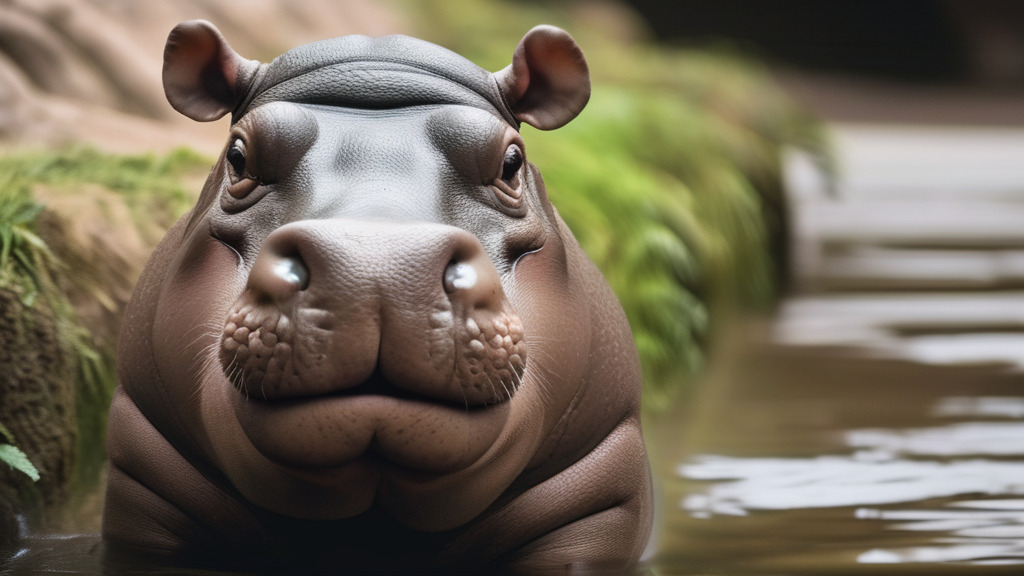}} & \raisebox{-.5\height}{\includegraphics[width=0.09\textwidth]{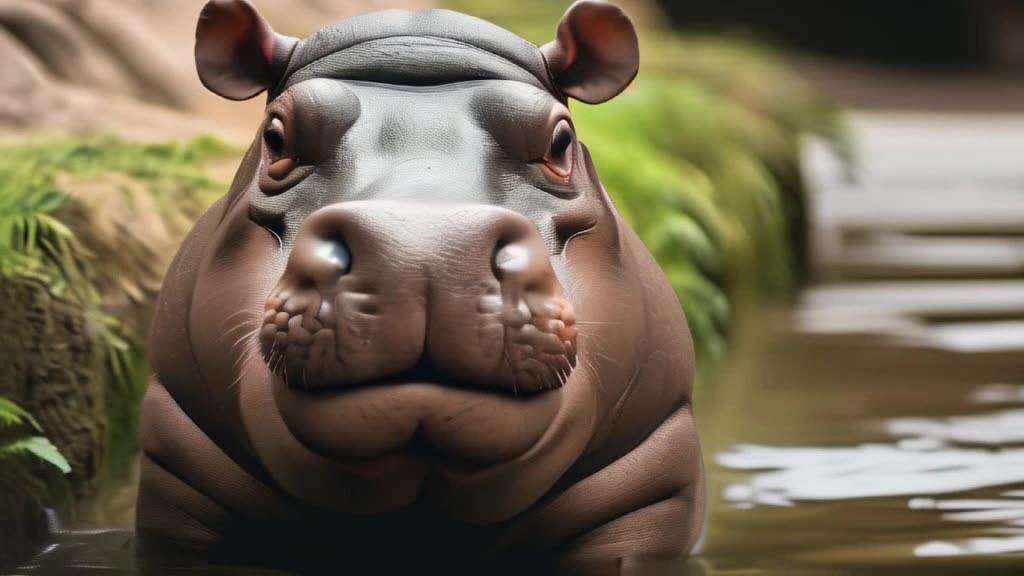}} \raisebox{-.5\height}{\includegraphics[width=0.09\textwidth]{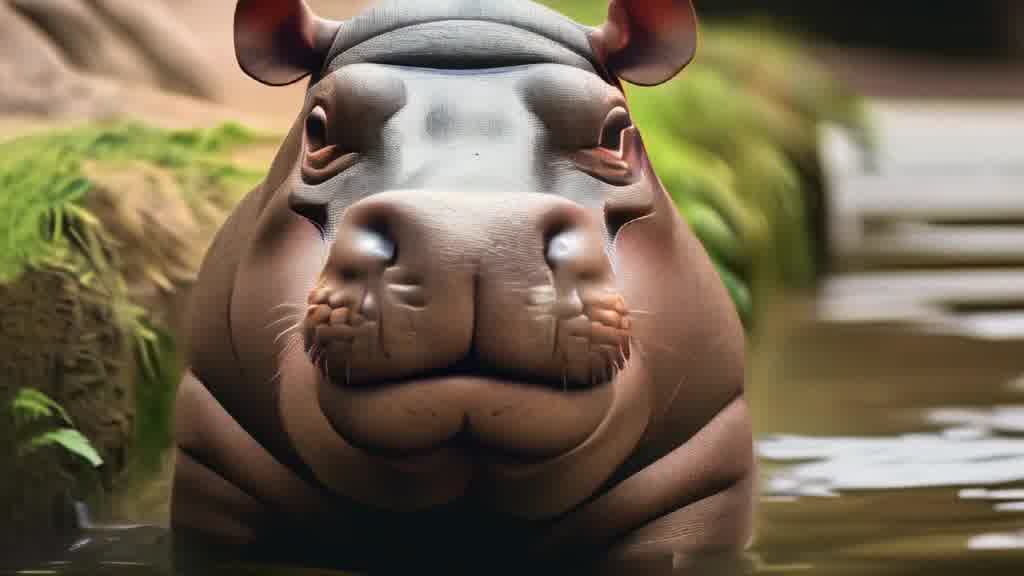}} \raisebox{-.5\height}{\includegraphics[width=0.09\textwidth]{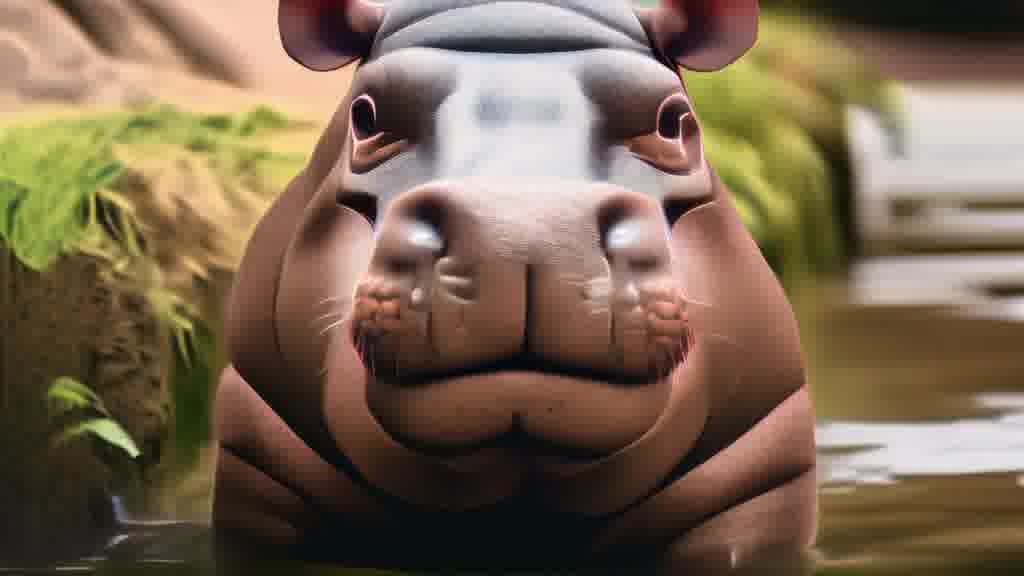}}
	\end{tblr}
	\caption{Additional results. Our learned motion-text embeddings can be applied to multiple target images, resulting in semantically similar motions.
	}
	\Description{Grid showing eight different motion transfer examples arranged in a grid with four rows and two columns. From top to bottom and left to right: jumping jacks from human to human and figurine, chewing from alpaca to vari and rhinoceros, nodding from human to human and puppy, tilting head down from human to digital character and multiple cartoon people, flying forward camera from landscape to landscape and landscape with a boat in the foreground, left-moving camera from landscape to landscape and squirrel, simple animation of stick figure jumping to human and kangaroo, rectangle growing as it moves from left to right to frog and hippopotamus.}
	\label{fig:results_extra}
\end{figure*}

\begin{figure*}[htbp]
	\centering
	\begin{tblr}{
			vline{3} = {2,3,4,5,7,8,9,10}{dashed},
			vline{4} = {1-10}{},
			vline{5} = {2,3,4,5,7,8,9,10}{dashed},
			hline{6} = {1-5}{},
		}
		{Ref.} & \raisebox{-.5\height}{\includegraphics[width=0.09\textwidth]{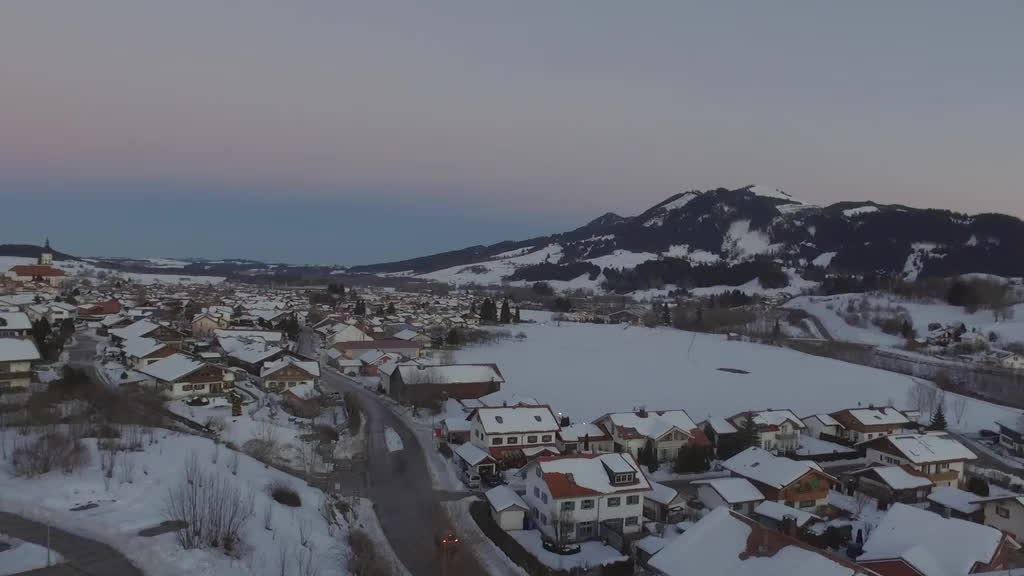}} & \raisebox{-.5\height}{\includegraphics[width=0.09\textwidth]{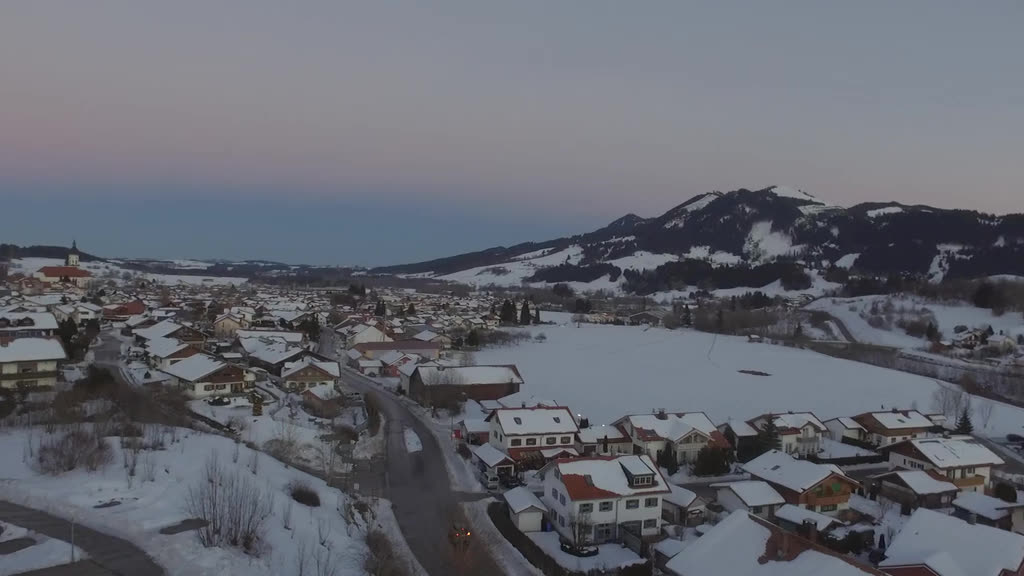}} \raisebox{-.5\height}{\includegraphics[width=0.09\textwidth]{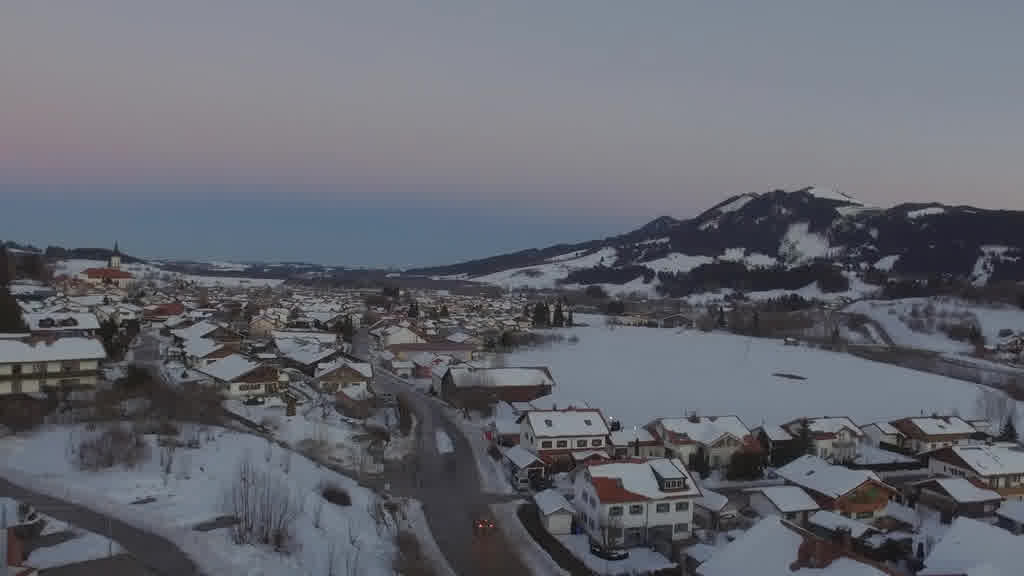}} \raisebox{-.5\height}{\includegraphics[width=0.09\textwidth]{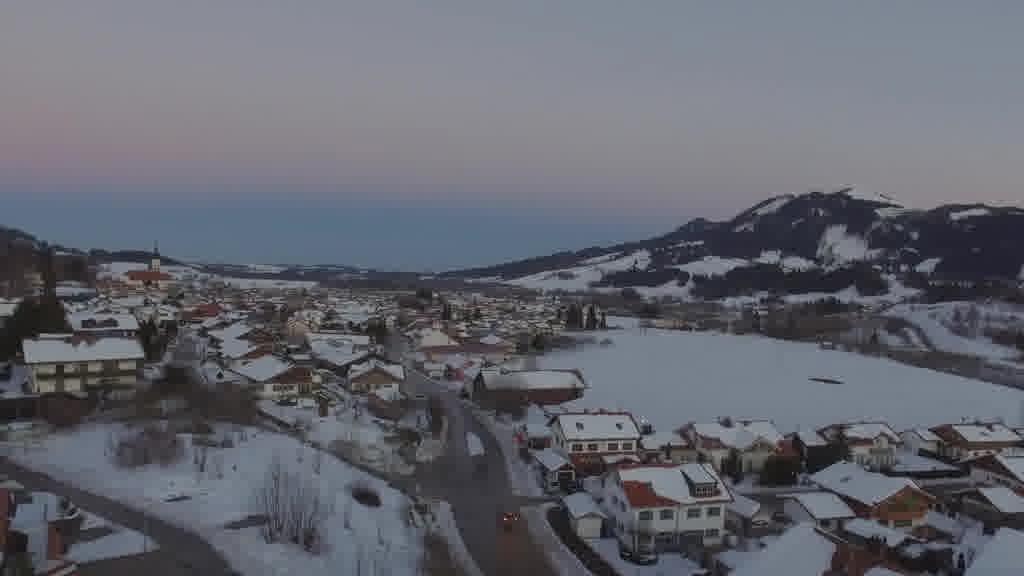}} & \raisebox{-.5\height}{\includegraphics[width=0.09\textwidth]{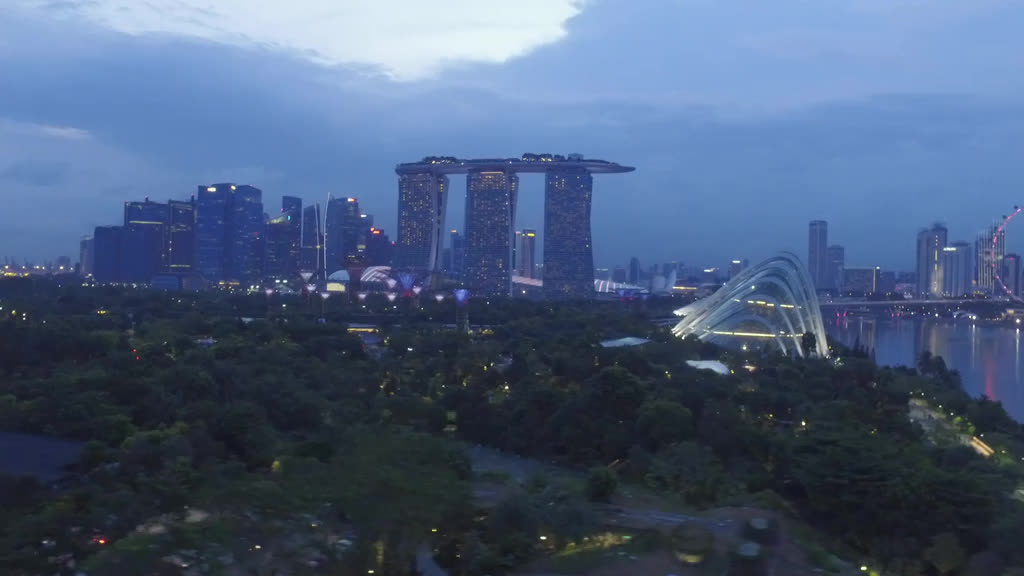}} & \raisebox{-.5\height}{\includegraphics[width=0.09\textwidth]{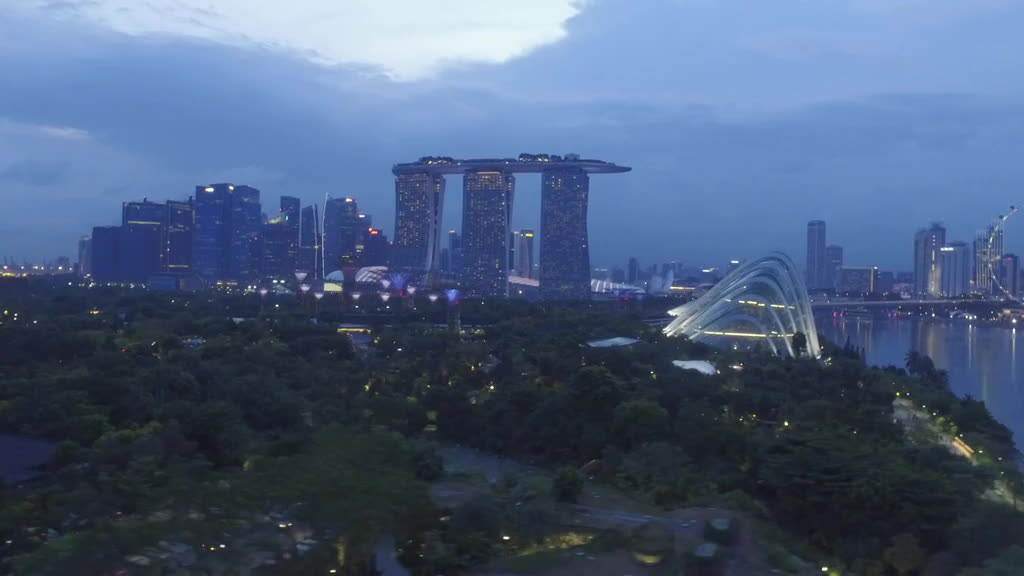}} \raisebox{-.5\height}{\includegraphics[width=0.09\textwidth]{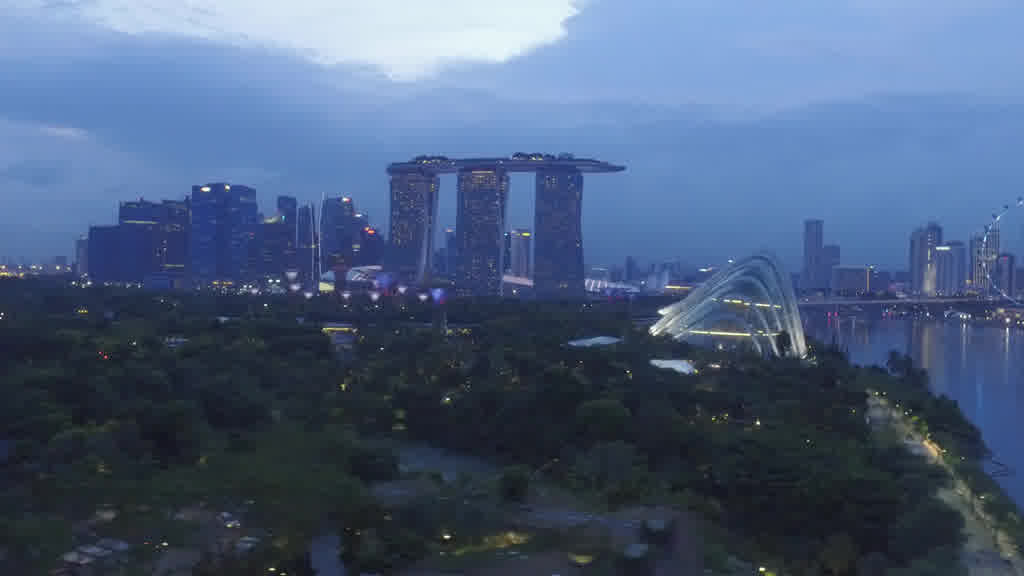}} \raisebox{-.5\height}{\includegraphics[width=0.09\textwidth]{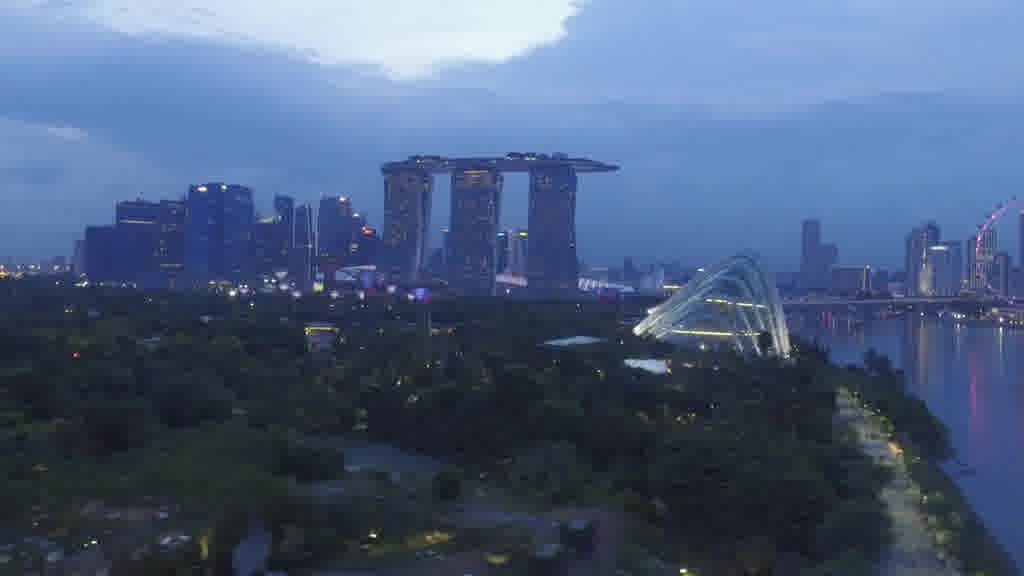}} \\
		{Gen. 1}  & \raisebox{-.5\height}{\includegraphics[width=0.09\textwidth]{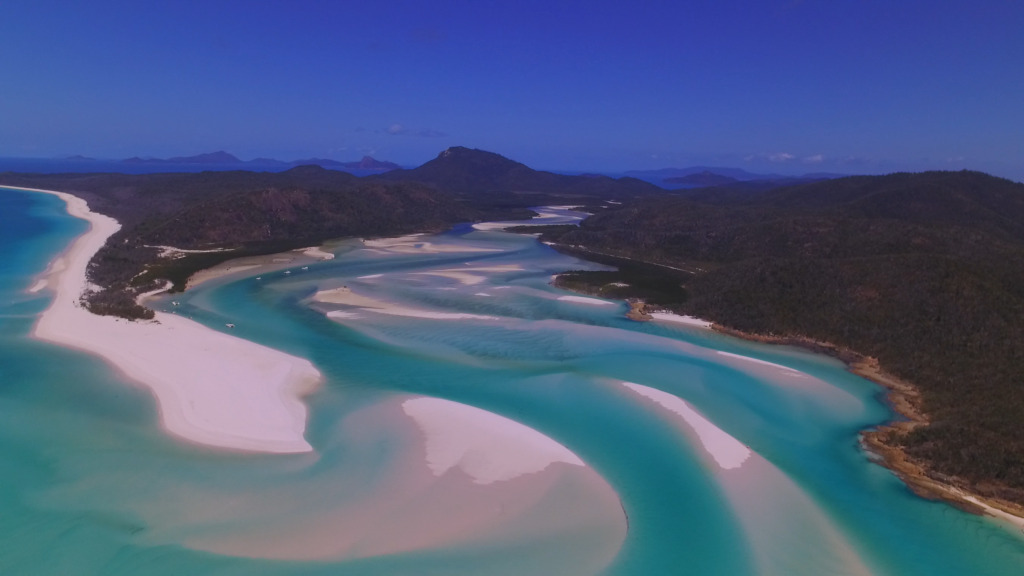}} & \raisebox{-.5\height}{\includegraphics[width=0.09\textwidth]{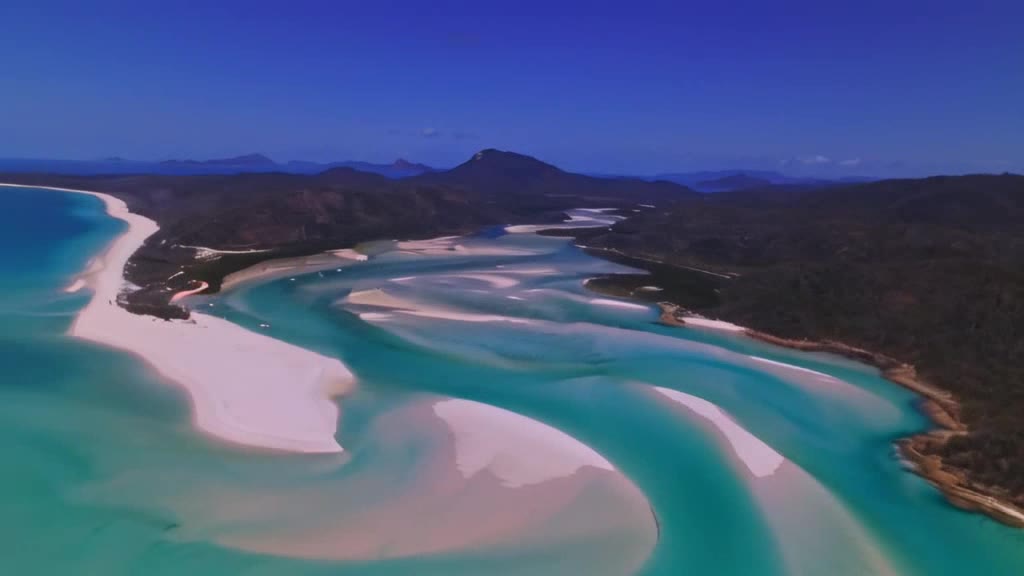}} \raisebox{-.5\height}{\includegraphics[width=0.09\textwidth]{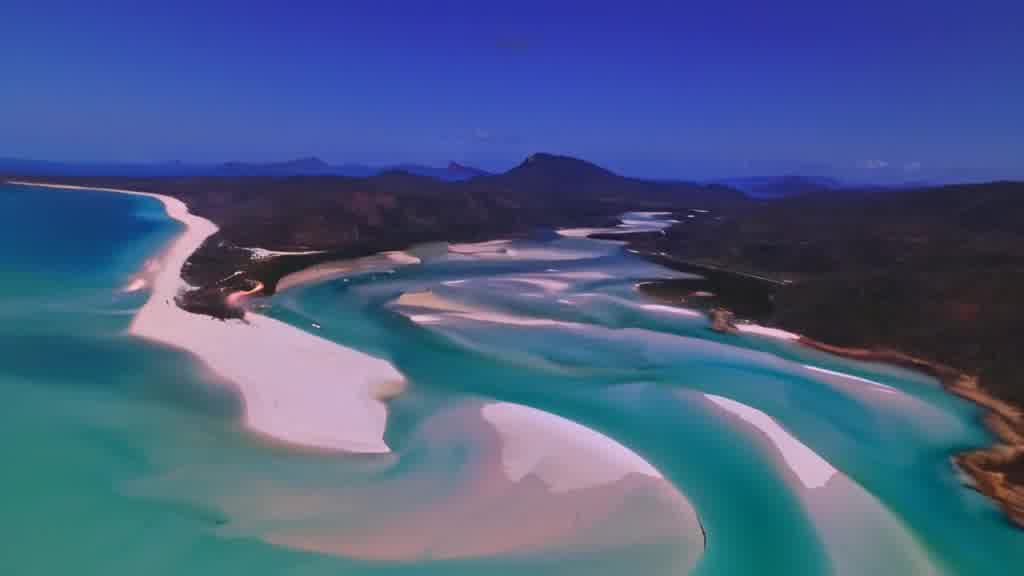}} \raisebox{-.5\height}{\includegraphics[width=0.09\textwidth]{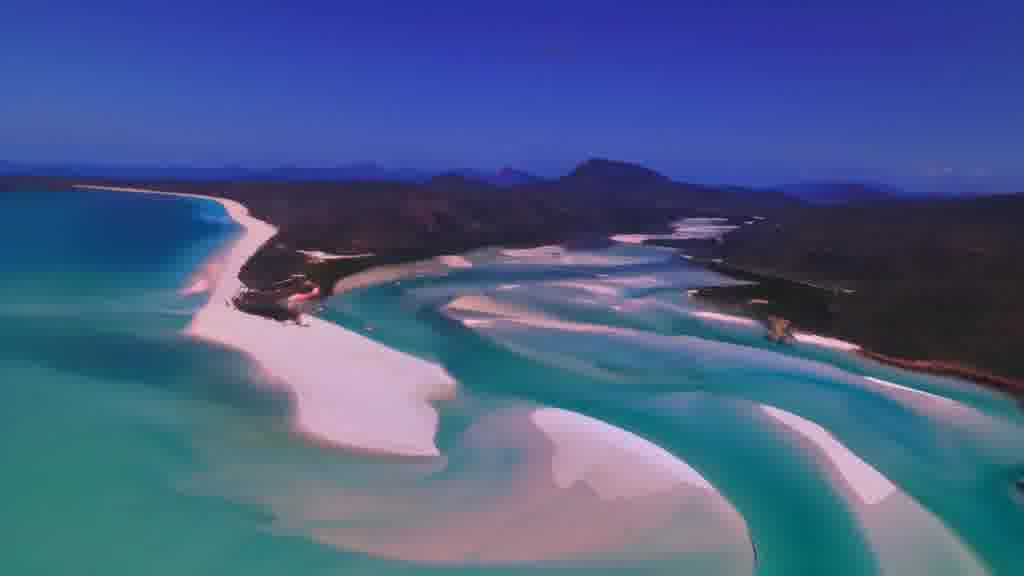}} & \raisebox{-.5\height}{\includegraphics[width=0.09\textwidth]{figures/cam_grid/first_frame.jpg}} & \raisebox{-.5\height}{\includegraphics[width=0.09\textwidth]{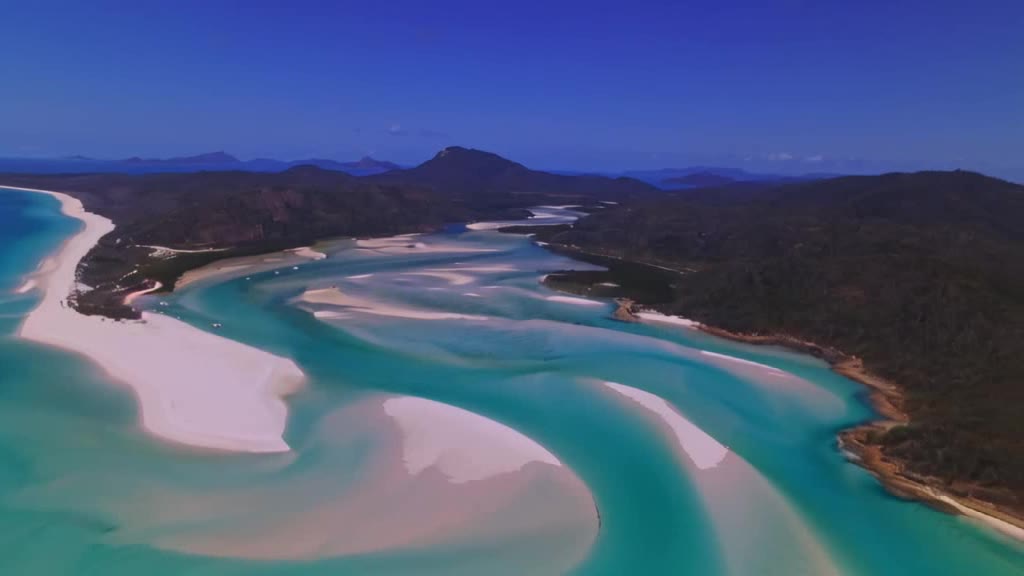}} \raisebox{-.5\height}{\includegraphics[width=0.09\textwidth]{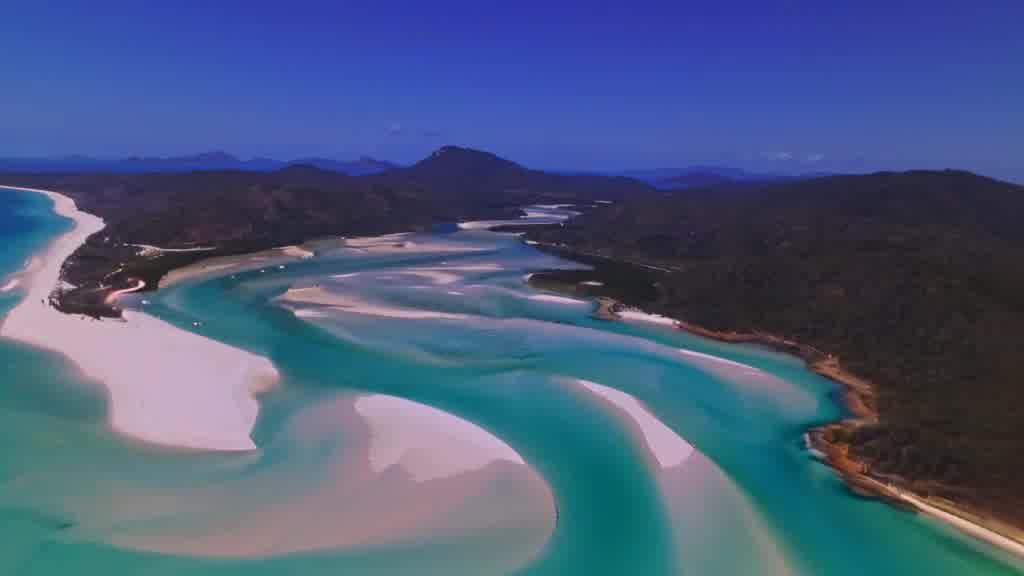}} \raisebox{-.5\height}{\includegraphics[width=0.09\textwidth]{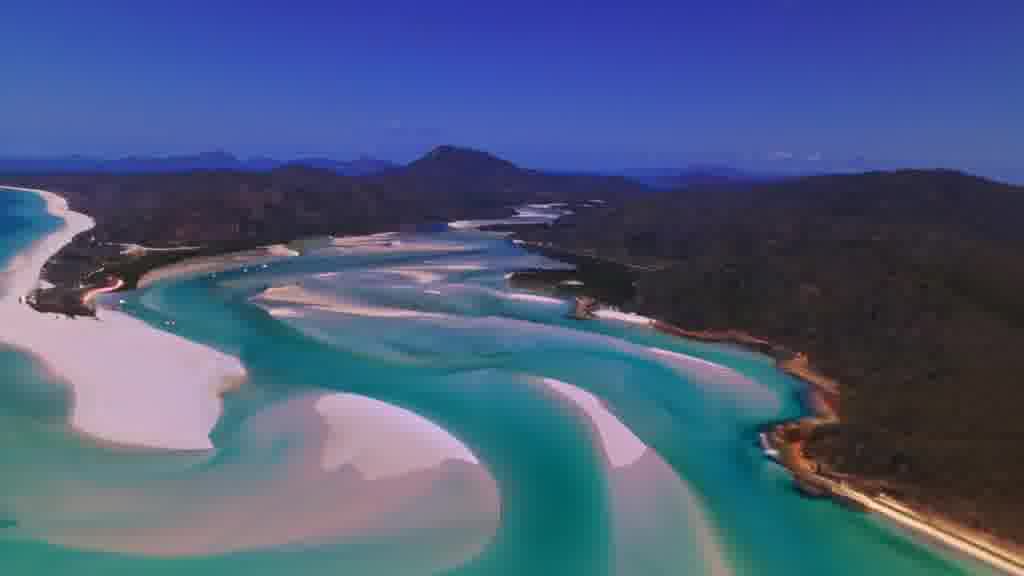}} \\
		{Gen. 2}  & \raisebox{-.5\height}{\includegraphics[width=0.09\textwidth]{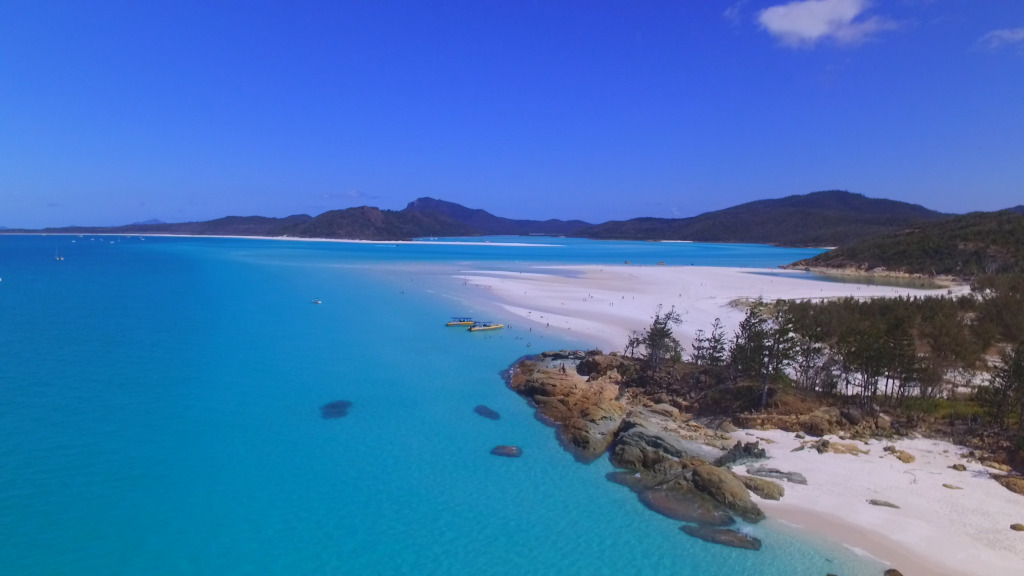}} & \raisebox{-.5\height}{\includegraphics[width=0.09\textwidth]{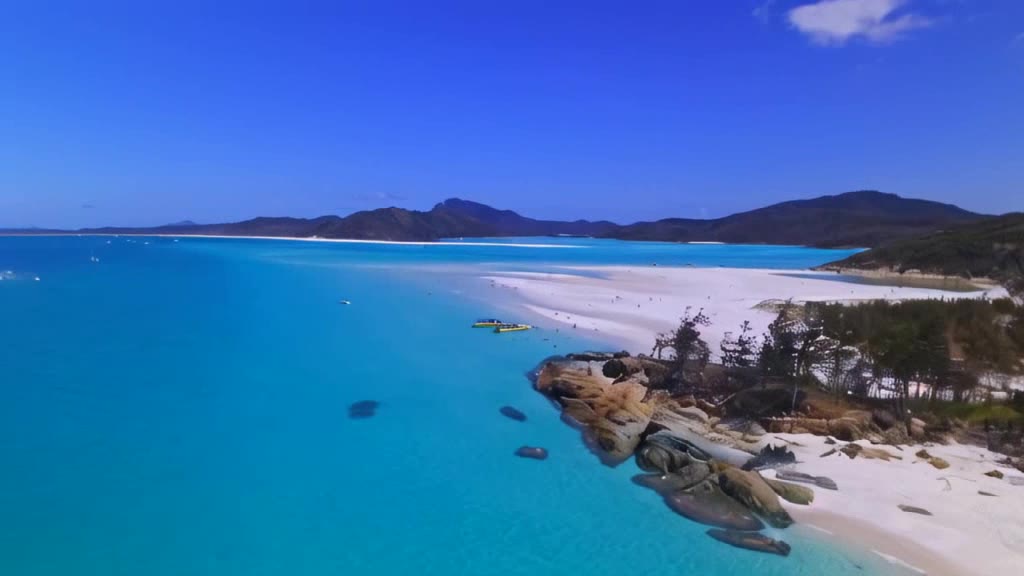}} \raisebox{-.5\height}{\includegraphics[width=0.09\textwidth]{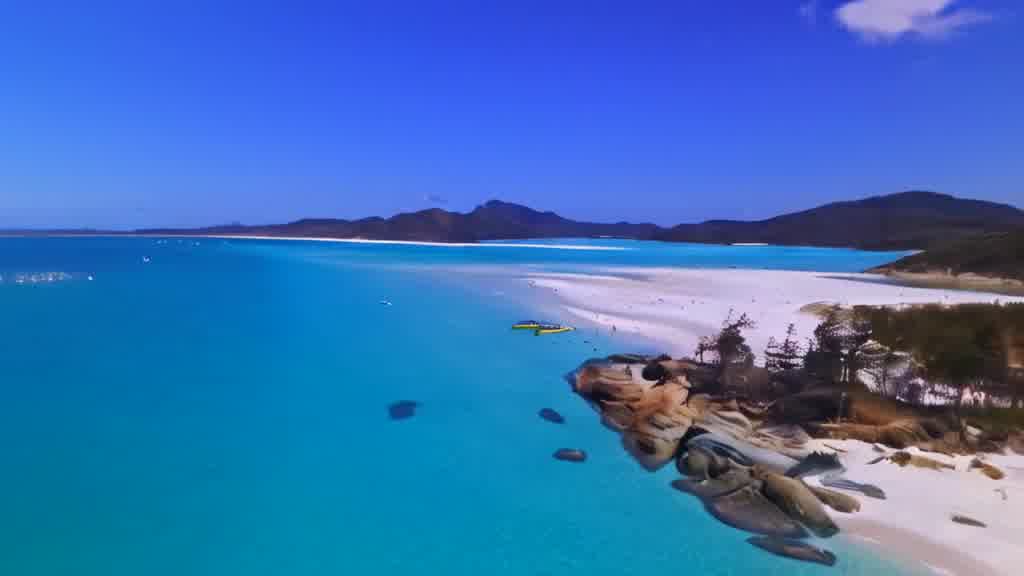}} \raisebox{-.5\height}{\includegraphics[width=0.09\textwidth]{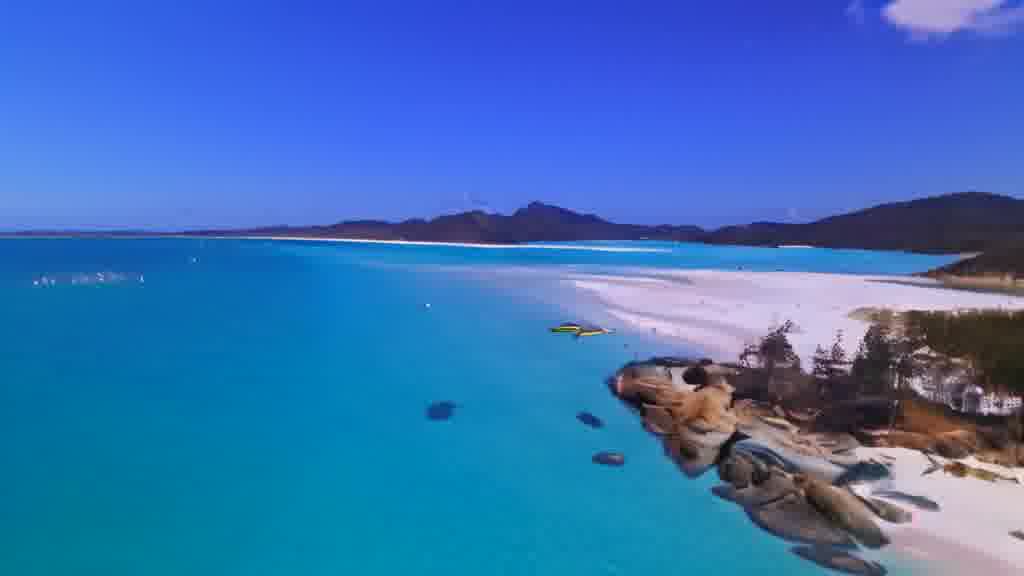}} & \raisebox{-.5\height}{\includegraphics[width=0.09\textwidth]{figures/cam_grid/first_frame_2.jpg}} & \raisebox{-.5\height}{\includegraphics[width=0.09\textwidth]{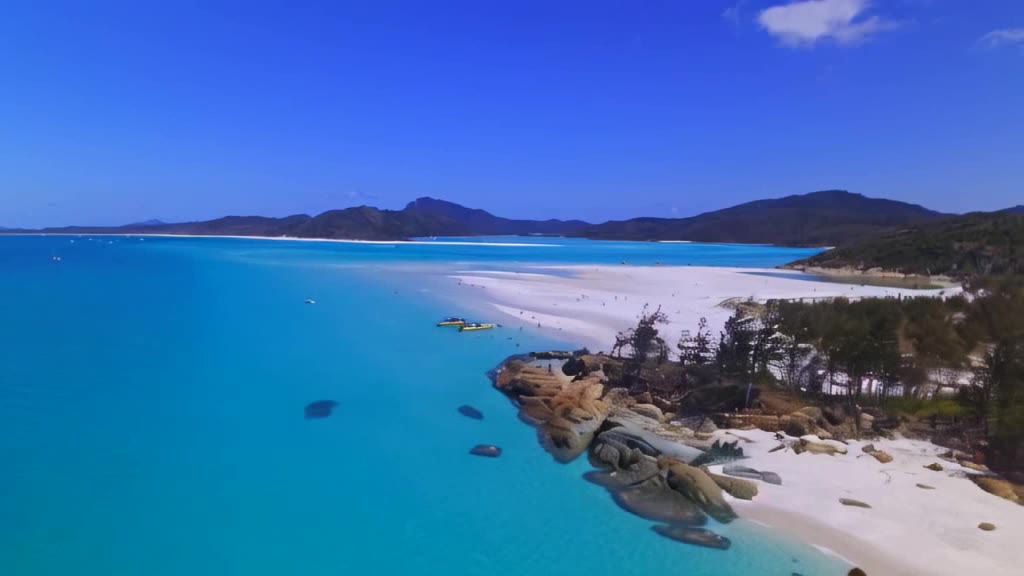}} \raisebox{-.5\height}{\includegraphics[width=0.09\textwidth]{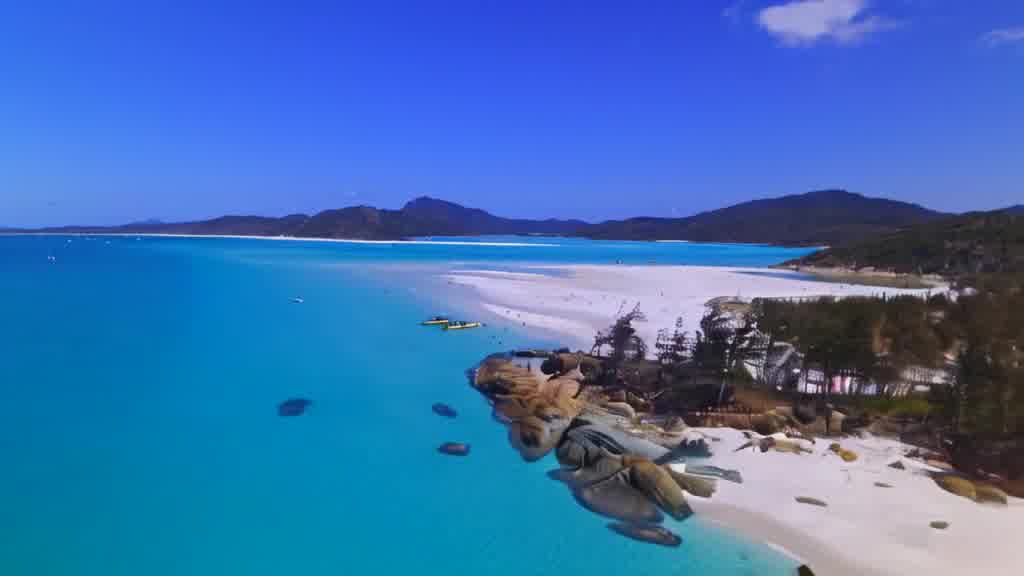}} \raisebox{-.5\height}{\includegraphics[width=0.09\textwidth]{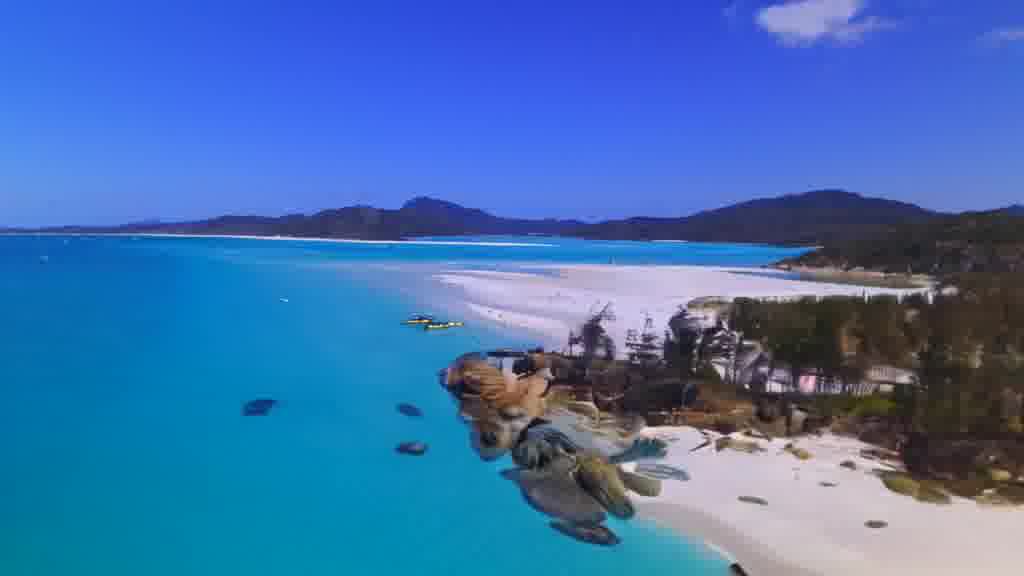}} \\
		{Gen. 3}  & \raisebox{-.5\height}{\includegraphics[width=0.09\textwidth]{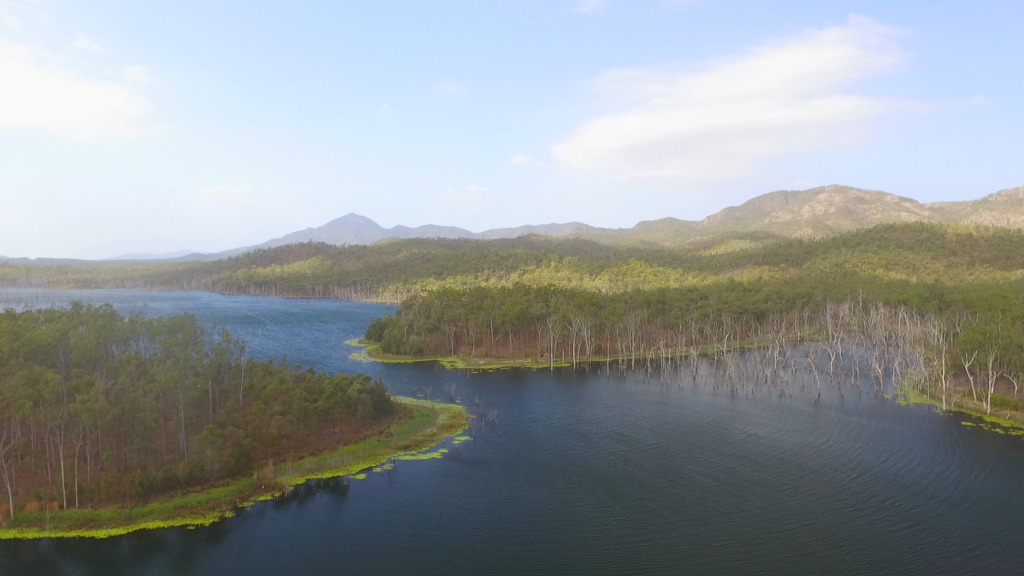}} & \raisebox{-.5\height}{\includegraphics[width=0.09\textwidth]{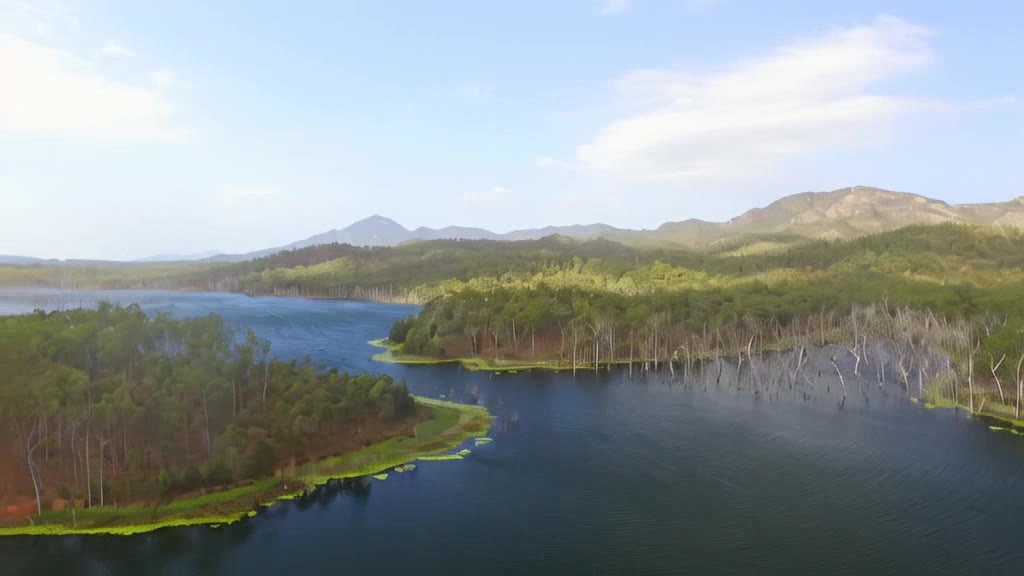}} \raisebox{-.5\height}{\includegraphics[width=0.09\textwidth]{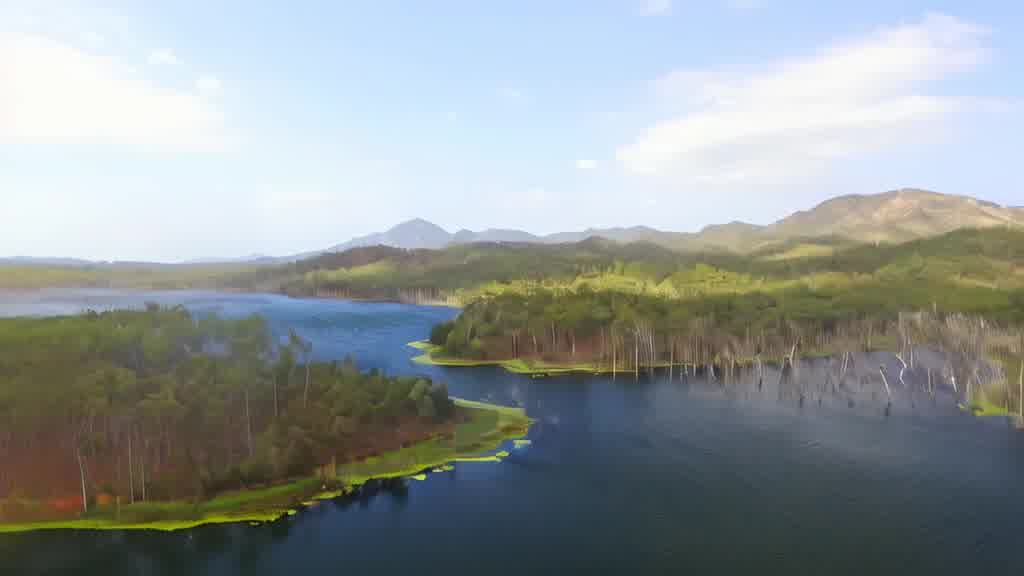}} \raisebox{-.5\height}{\includegraphics[width=0.09\textwidth]{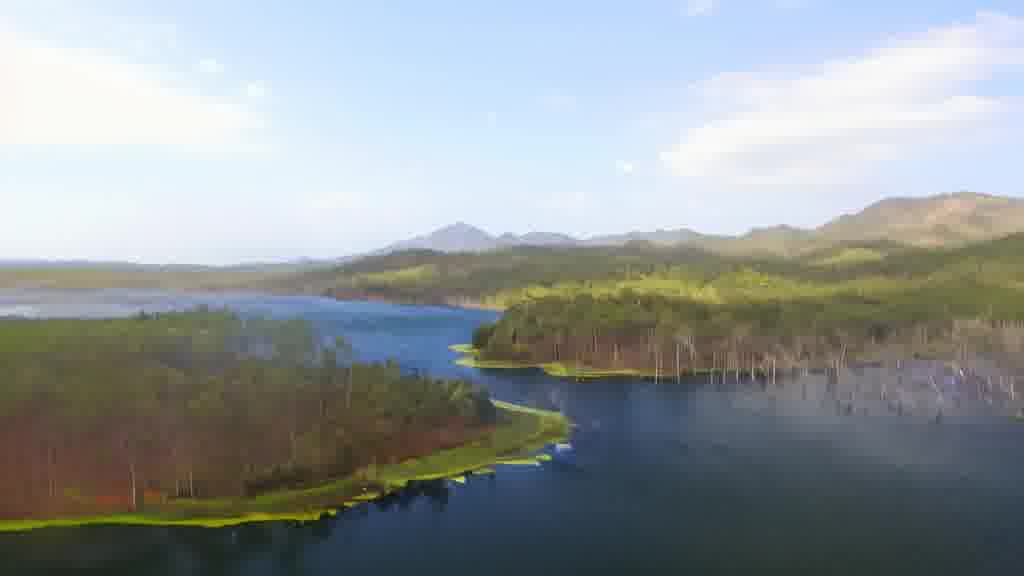}} & \raisebox{-.5\height}{\includegraphics[width=0.09\textwidth]{figures/cam_grid/first_frame_3.jpg}} & \raisebox{-.5\height}{\includegraphics[width=0.09\textwidth]{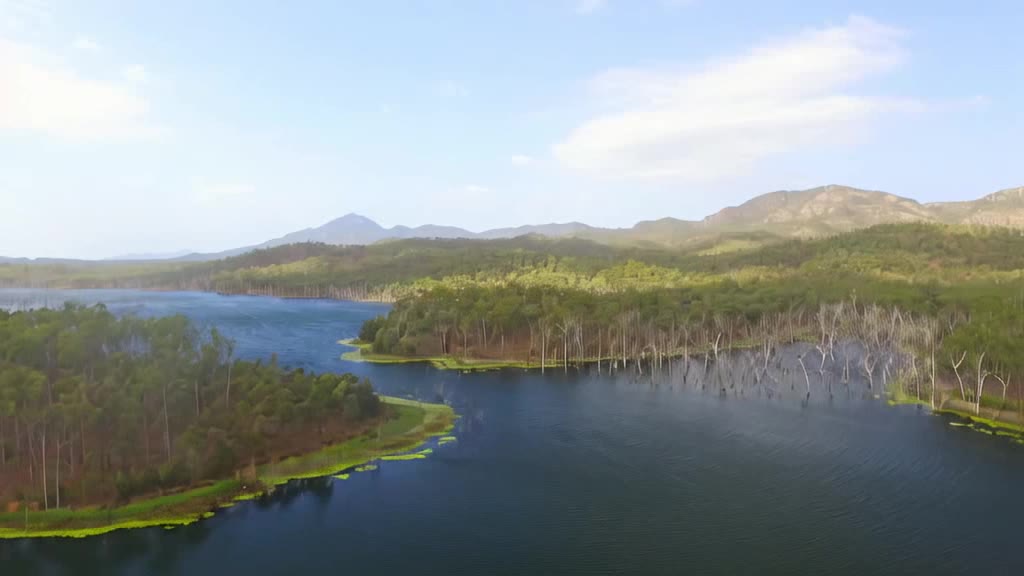}} \raisebox{-.5\height}{\includegraphics[width=0.09\textwidth]{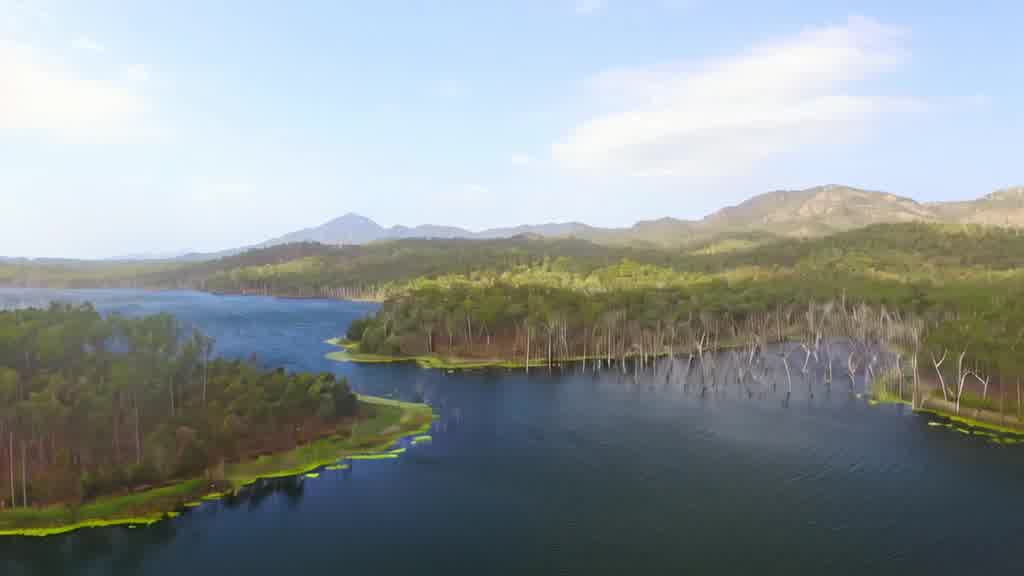}} \raisebox{-.5\height}{\includegraphics[width=0.09\textwidth]{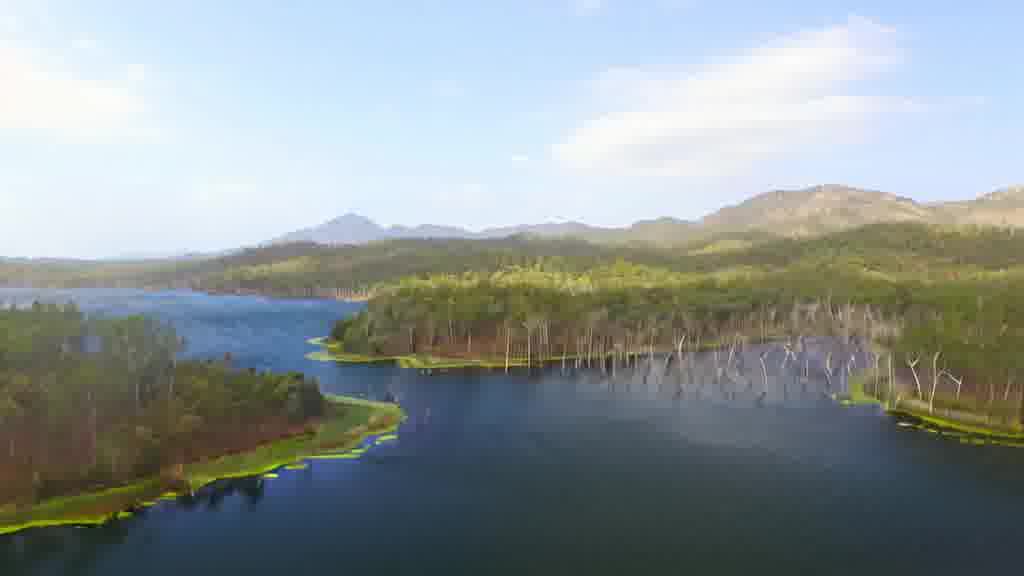}} \\
		{Gen. 4}  & \raisebox{-.5\height}{\includegraphics[width=0.09\textwidth]{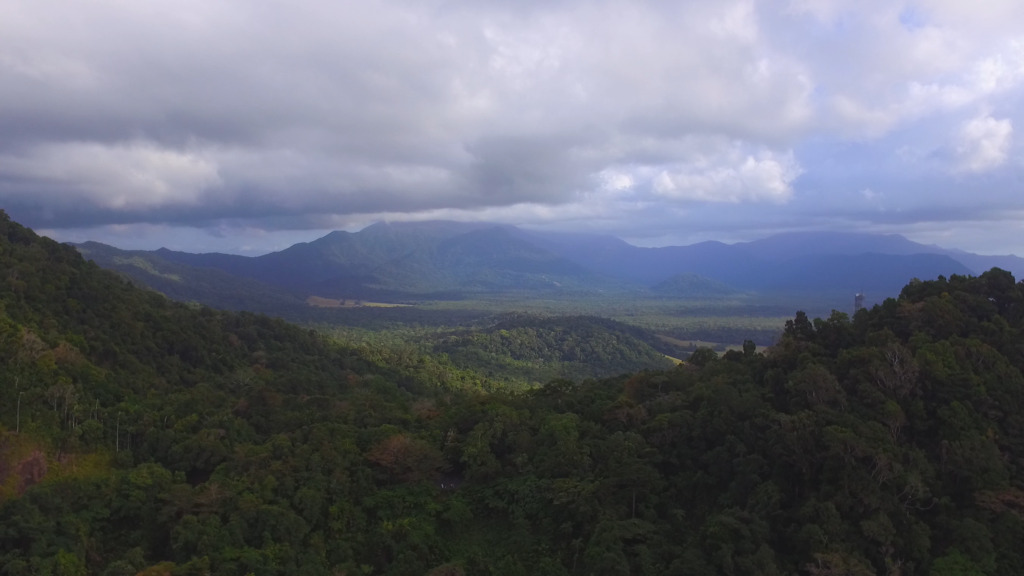}} & \raisebox{-.5\height}{\includegraphics[width=0.09\textwidth]{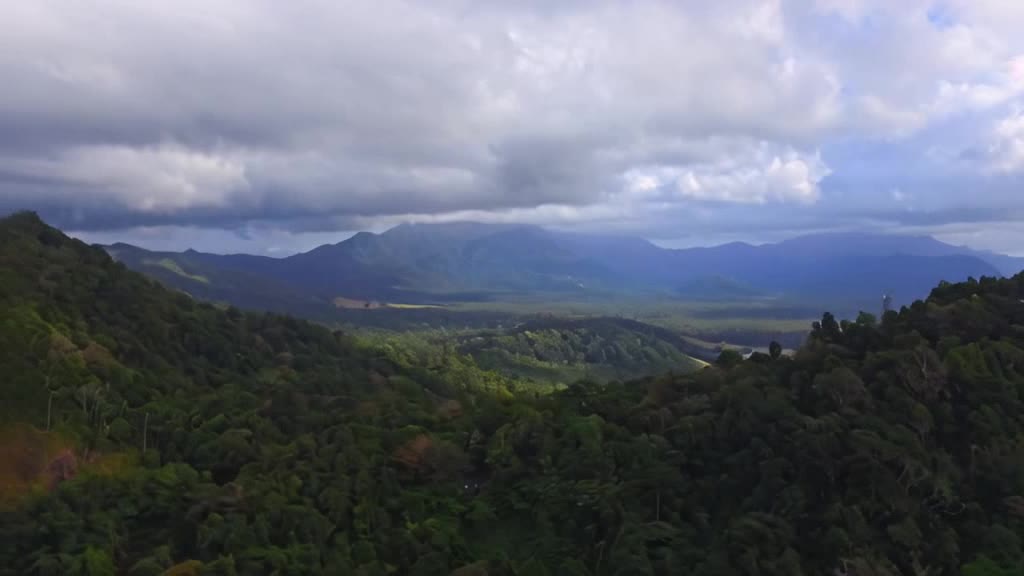}} \raisebox{-.5\height}{\includegraphics[width=0.09\textwidth]{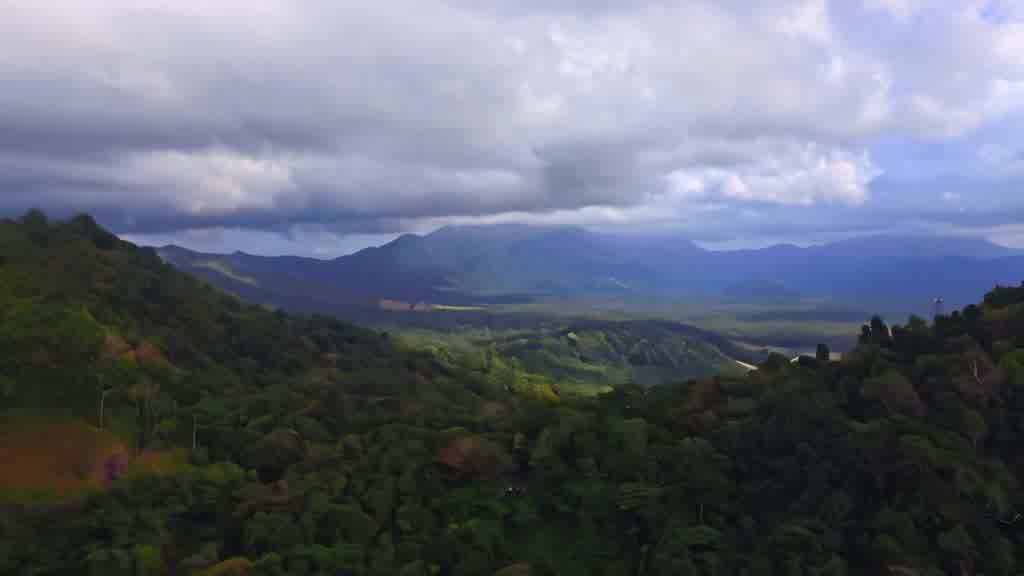}} \raisebox{-.5\height}{\includegraphics[width=0.09\textwidth]{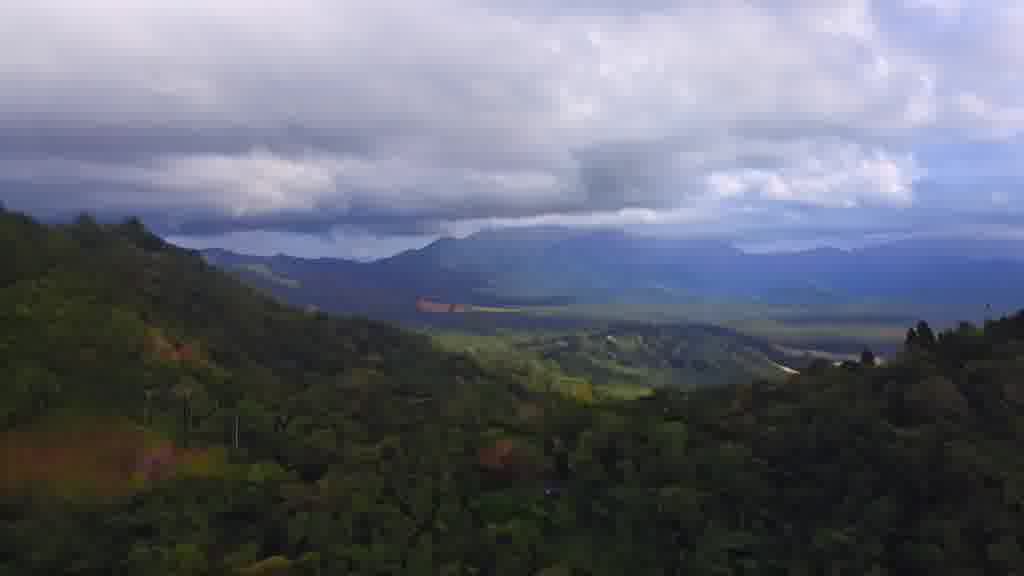}} & \raisebox{-.5\height}{\includegraphics[width=0.09\textwidth]{figures/cam_grid/first_frame_4.jpg}} & \raisebox{-.5\height}{\includegraphics[width=0.09\textwidth]{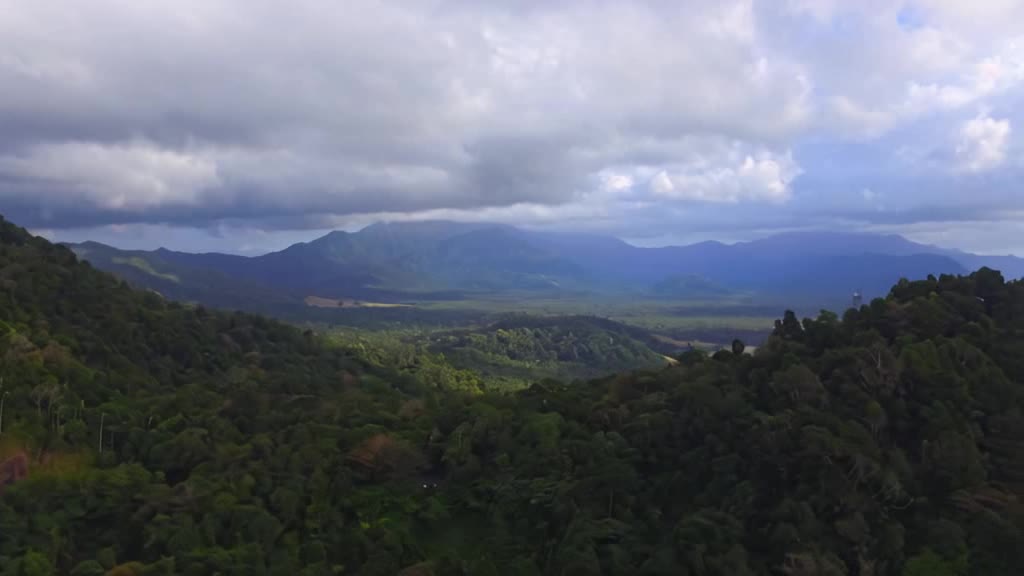}} \raisebox{-.5\height}{\includegraphics[width=0.09\textwidth]{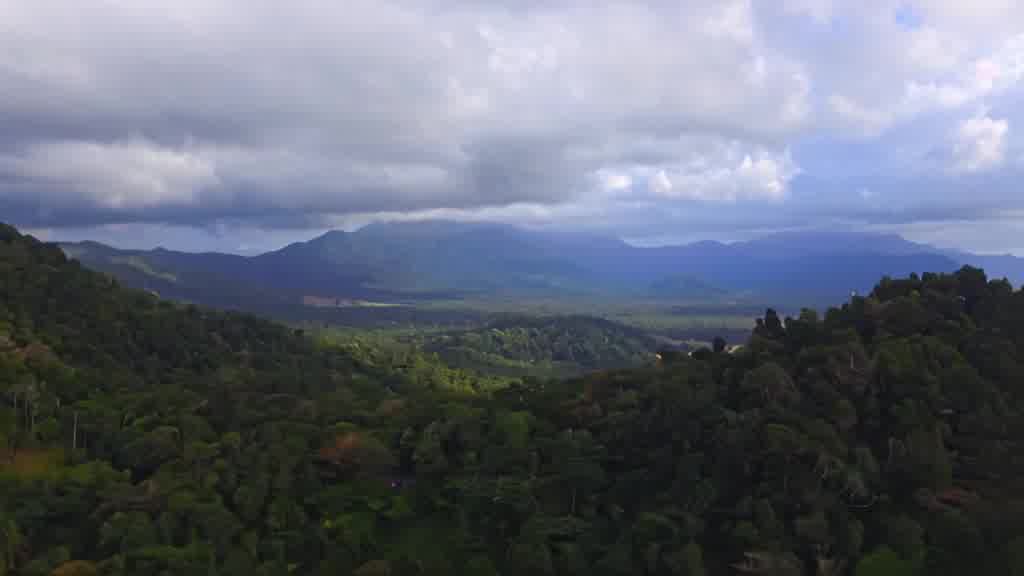}} \raisebox{-.5\height}{\includegraphics[width=0.09\textwidth]{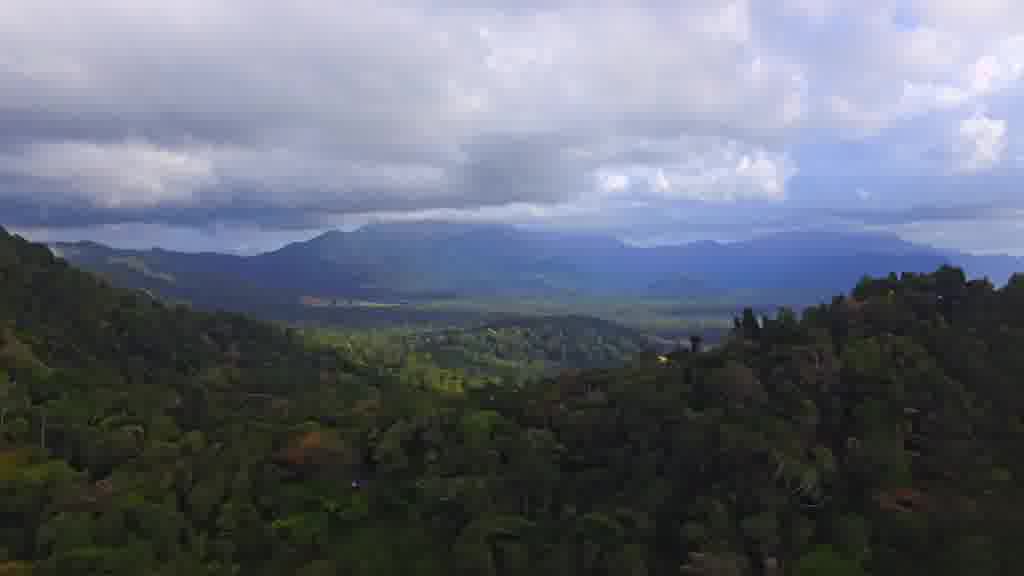}} \\
		{Ref.} & \raisebox{-.5\height}{\includegraphics[width=0.09\textwidth]{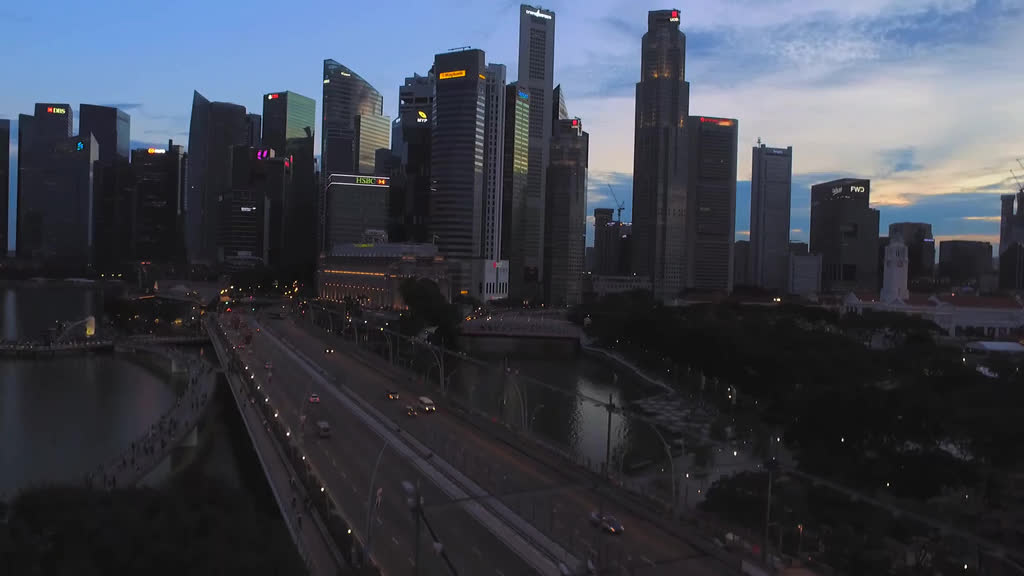}} & \raisebox{-.5\height}{\includegraphics[width=0.09\textwidth]{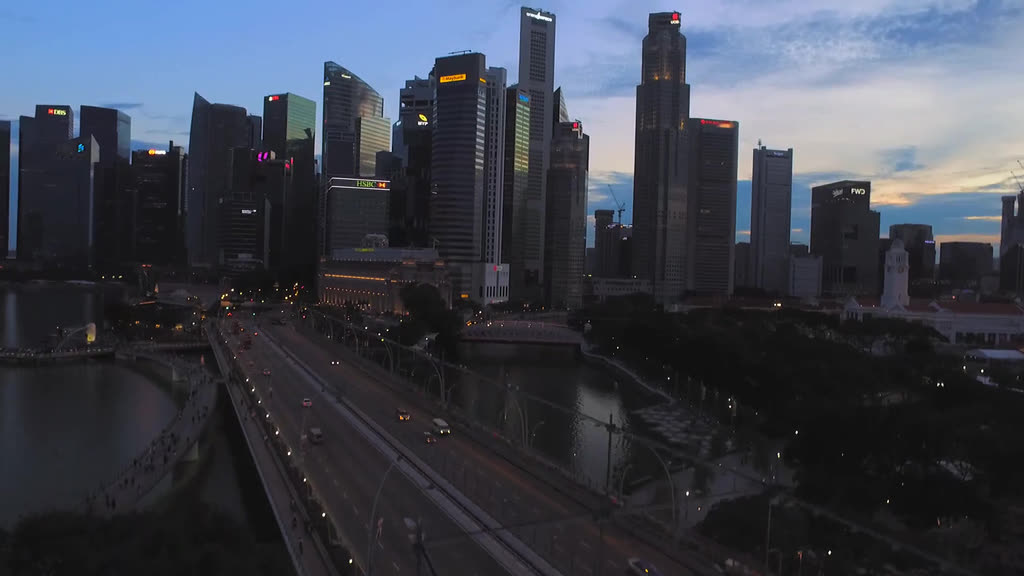}} \raisebox{-.5\height}{\includegraphics[width=0.09\textwidth]{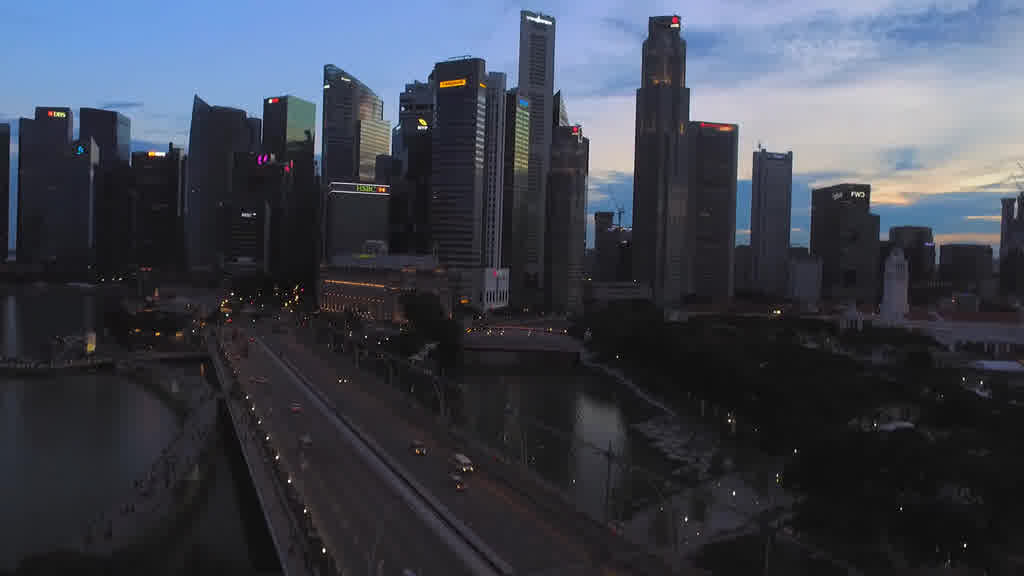}} \raisebox{-.5\height}{\includegraphics[width=0.09\textwidth]{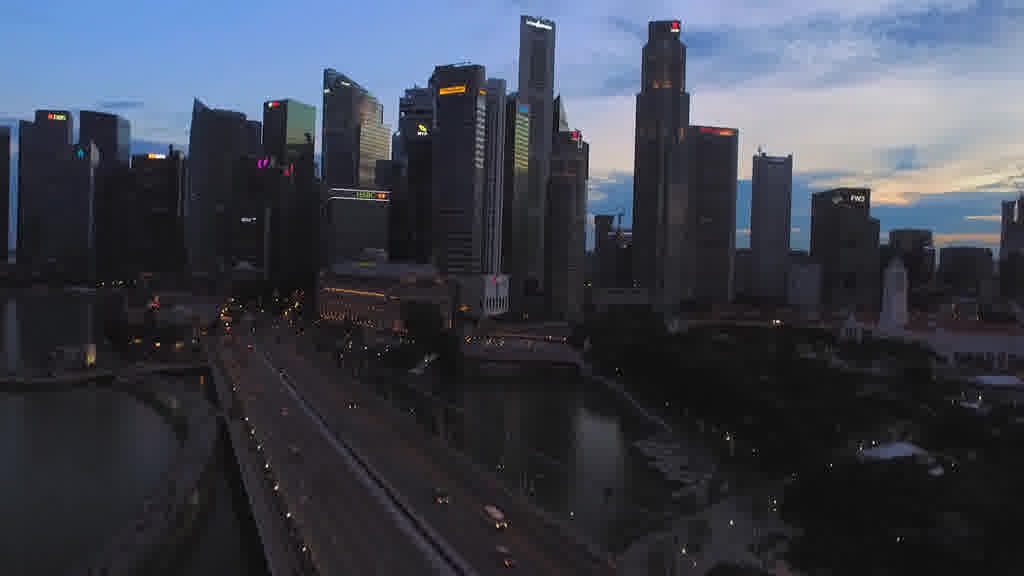}} & \raisebox{-.5\height}{\includegraphics[width=0.09\textwidth]{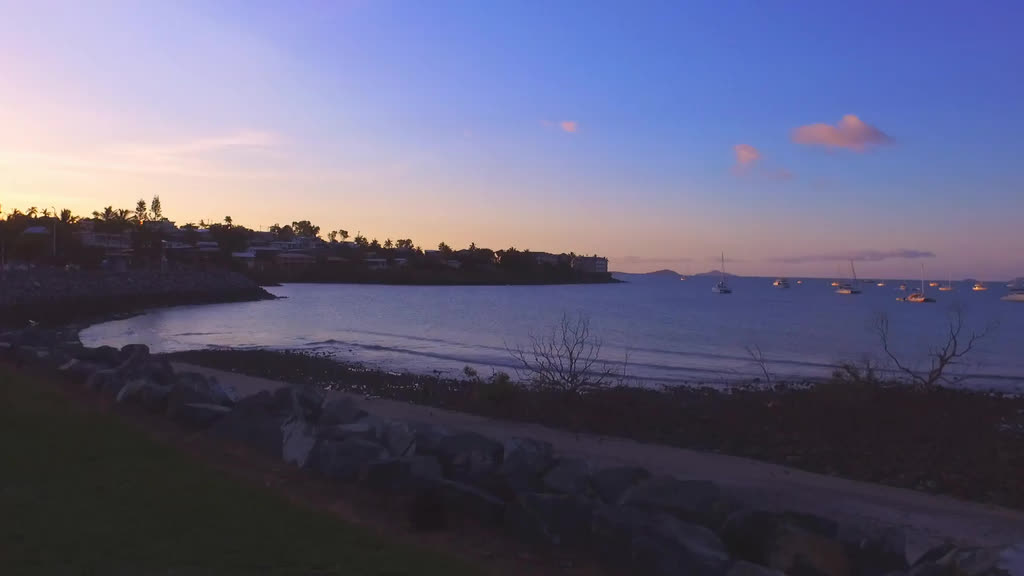}} & \raisebox{-.5\height}{\includegraphics[width=0.09\textwidth]{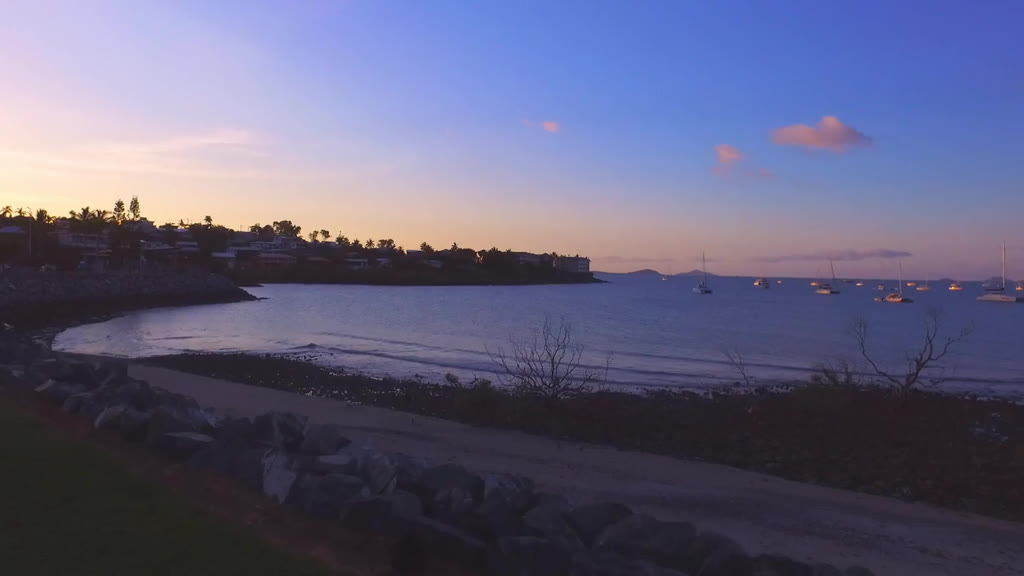}} \raisebox{-.5\height}{\includegraphics[width=0.09\textwidth]{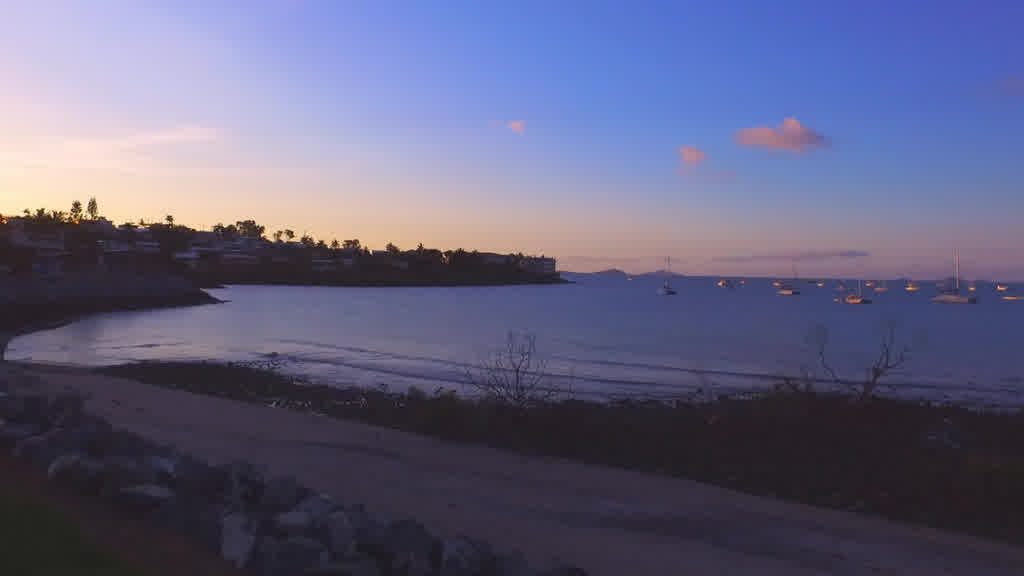}} \raisebox{-.5\height}{\includegraphics[width=0.09\textwidth]{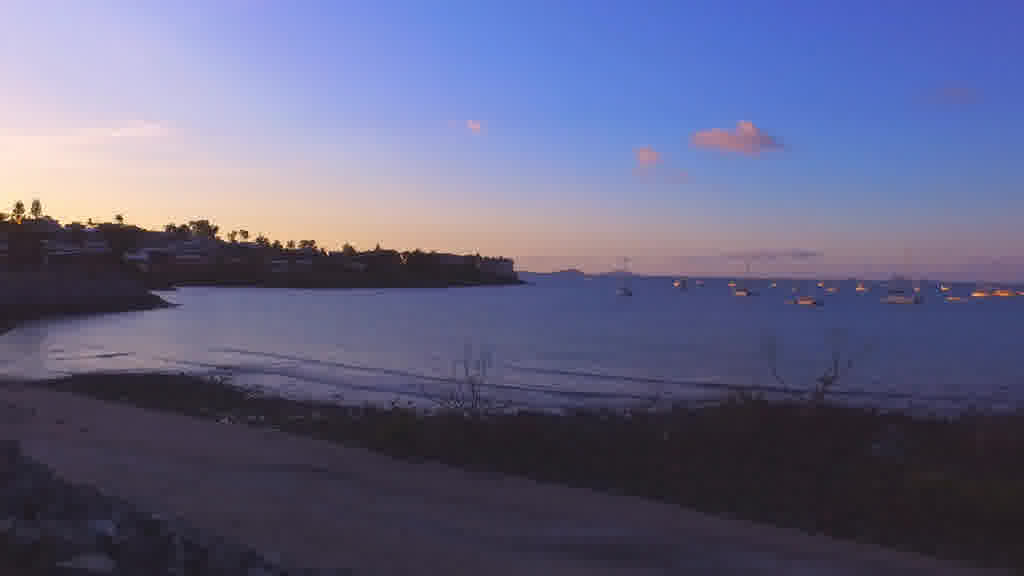}} \\
		{Gen. 1}  & \raisebox{-.5\height}{\includegraphics[width=0.09\textwidth]{figures/cam_grid/first_frame.jpg}} & \raisebox{-.5\height}{\includegraphics[width=0.09\textwidth]{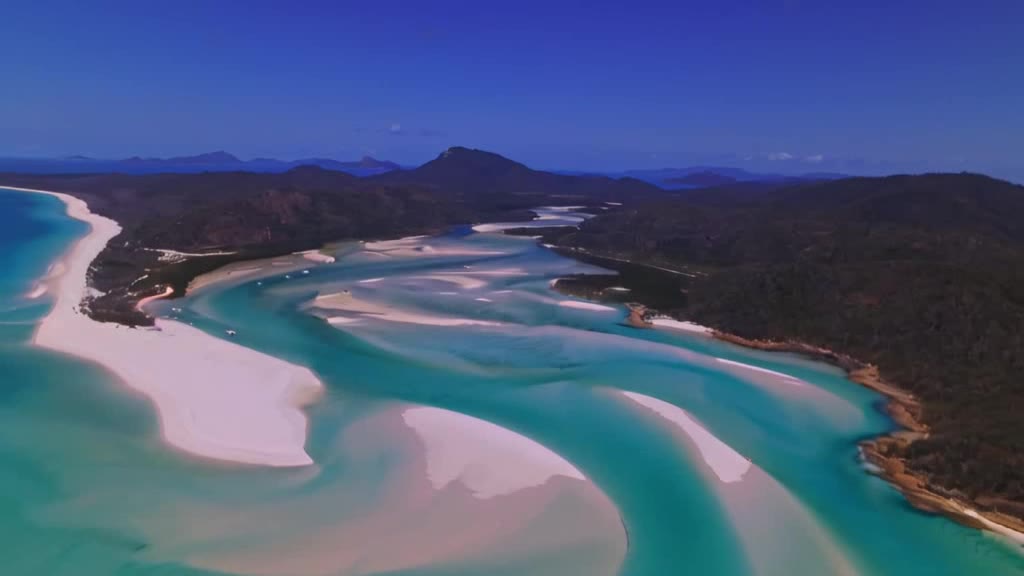}} \raisebox{-.5\height}{\includegraphics[width=0.09\textwidth]{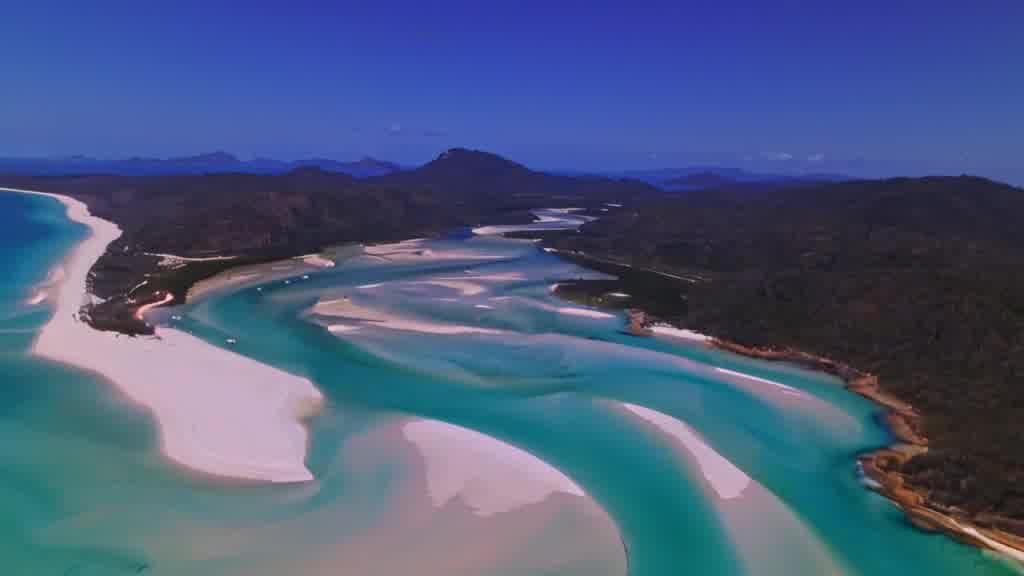}} \raisebox{-.5\height}{\includegraphics[width=0.09\textwidth]{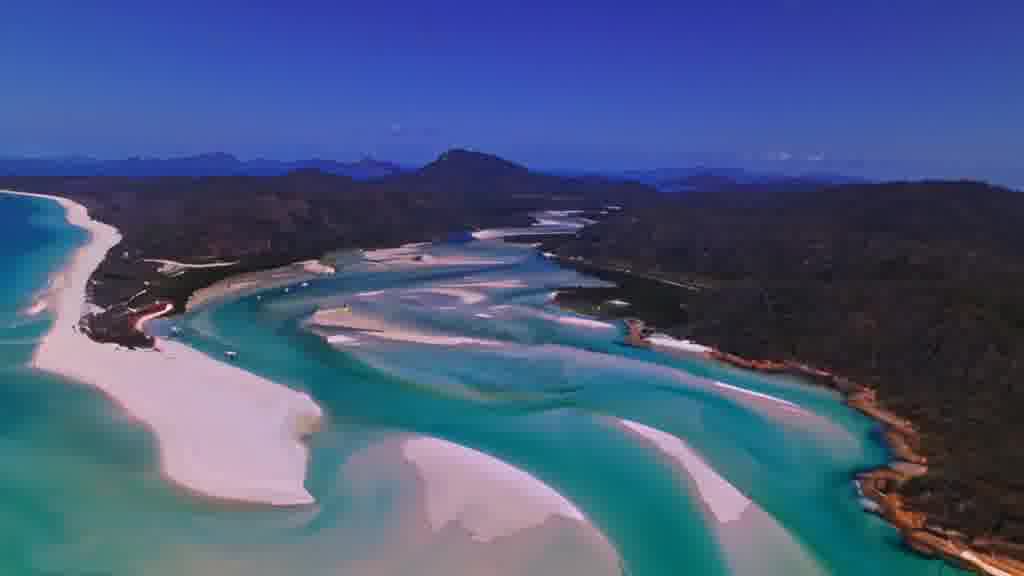}} & \raisebox{-.5\height}{\includegraphics[width=0.09\textwidth]{figures/cam_grid/first_frame.jpg}} & \raisebox{-.5\height}{\includegraphics[width=0.09\textwidth]{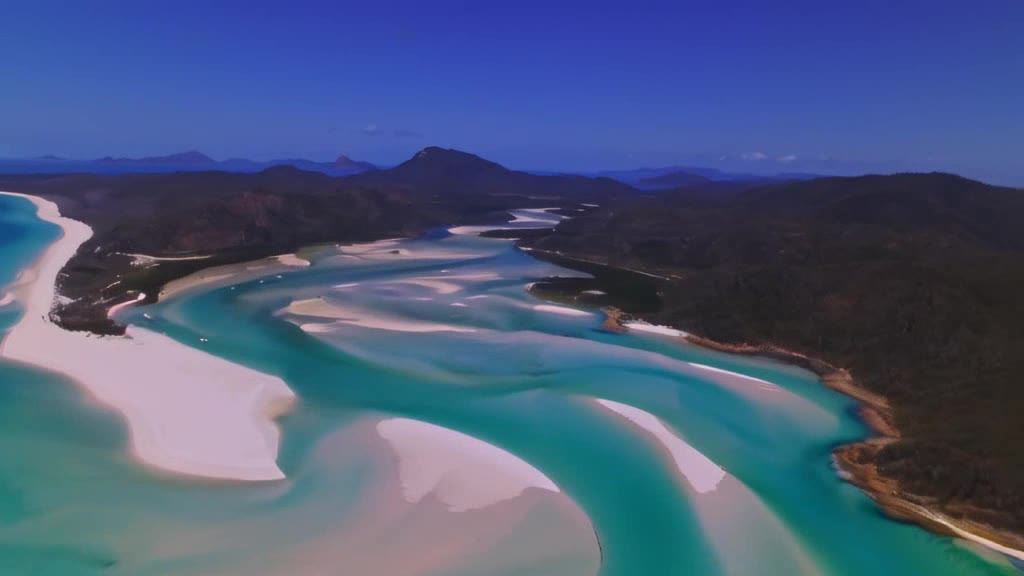}} \raisebox{-.5\height}{\includegraphics[width=0.09\textwidth]{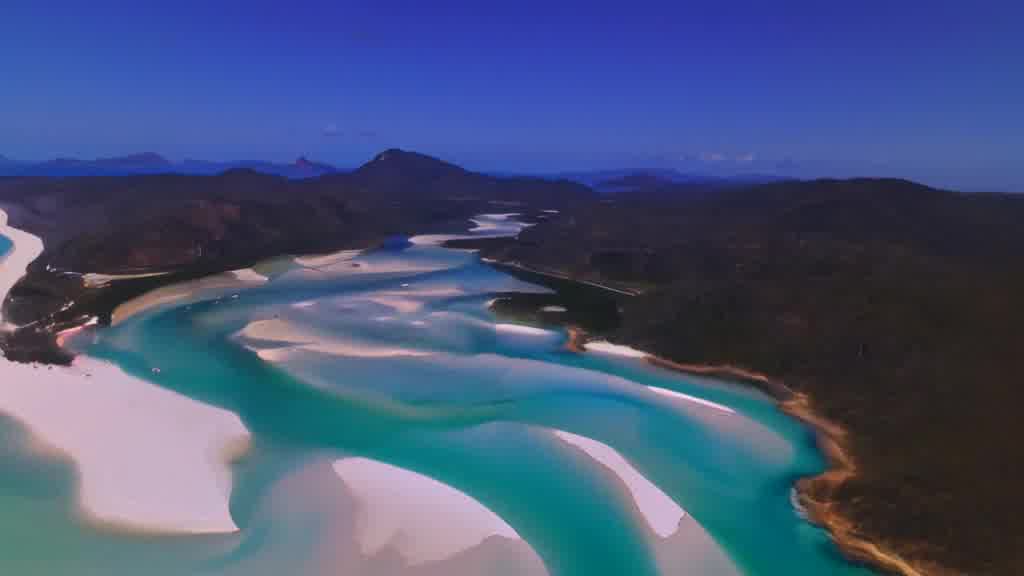}} \raisebox{-.5\height}{\includegraphics[width=0.09\textwidth]{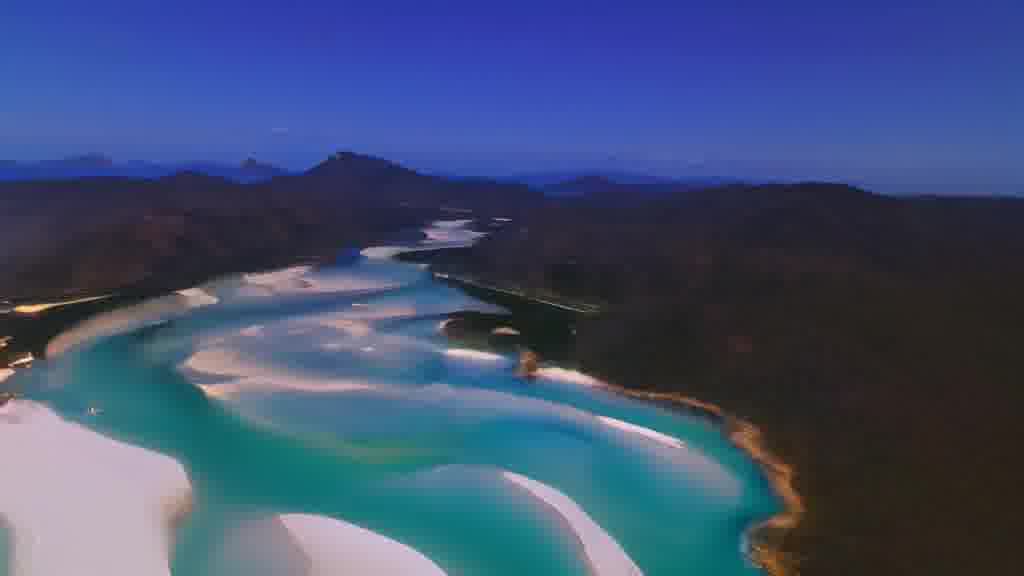}} \\
		{Gen. 2}  & \raisebox{-.5\height}{\includegraphics[width=0.09\textwidth]{figures/cam_grid/first_frame_2.jpg}} & \raisebox{-.5\height}{\includegraphics[width=0.09\textwidth]{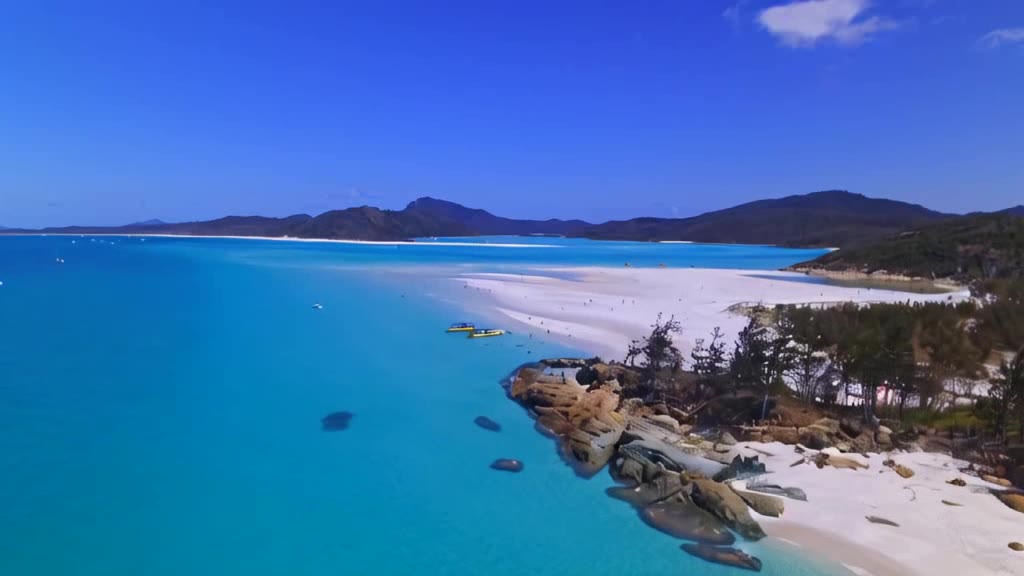}} \raisebox{-.5\height}{\includegraphics[width=0.09\textwidth]{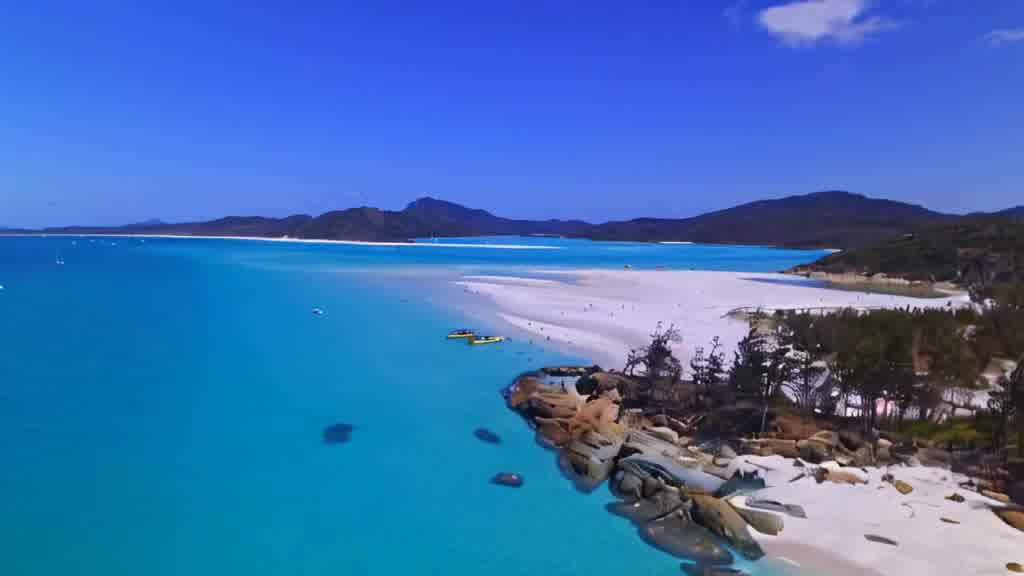}} \raisebox{-.5\height}{\includegraphics[width=0.09\textwidth]{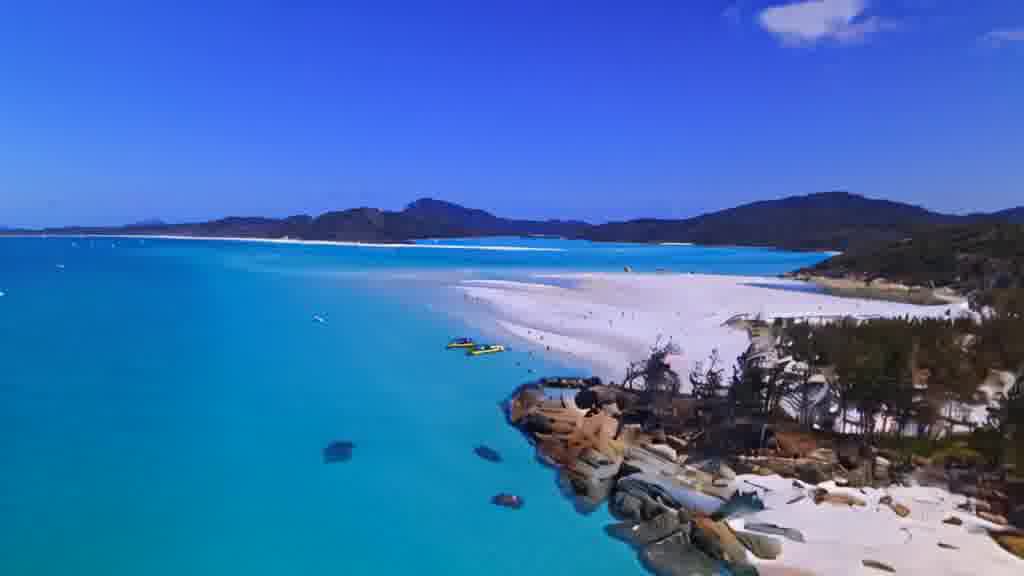}} & \raisebox{-.5\height}{\includegraphics[width=0.09\textwidth]{figures/cam_grid/first_frame_2.jpg}} & \raisebox{-.5\height}{\includegraphics[width=0.09\textwidth]{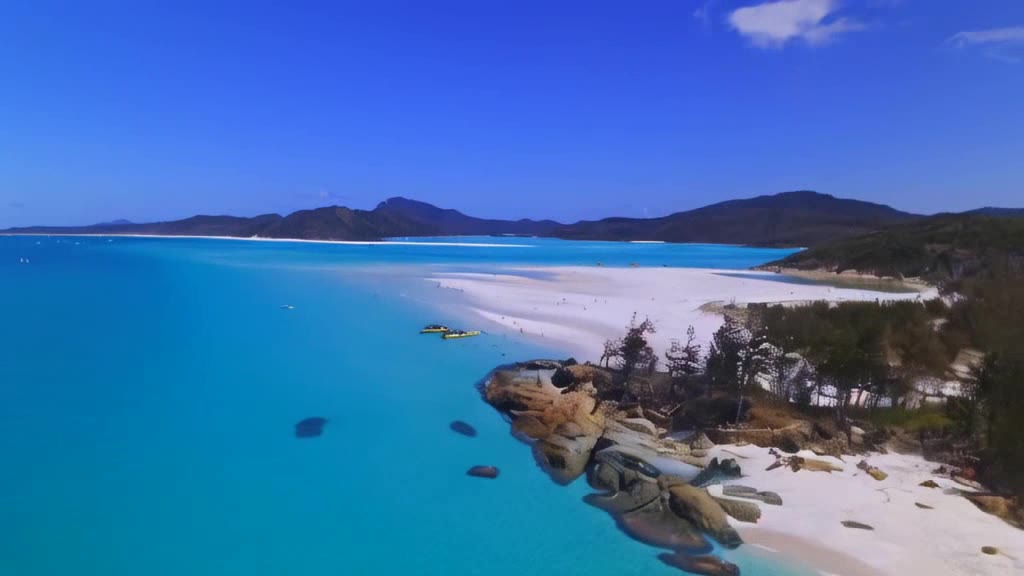}} \raisebox{-.5\height}{\includegraphics[width=0.09\textwidth]{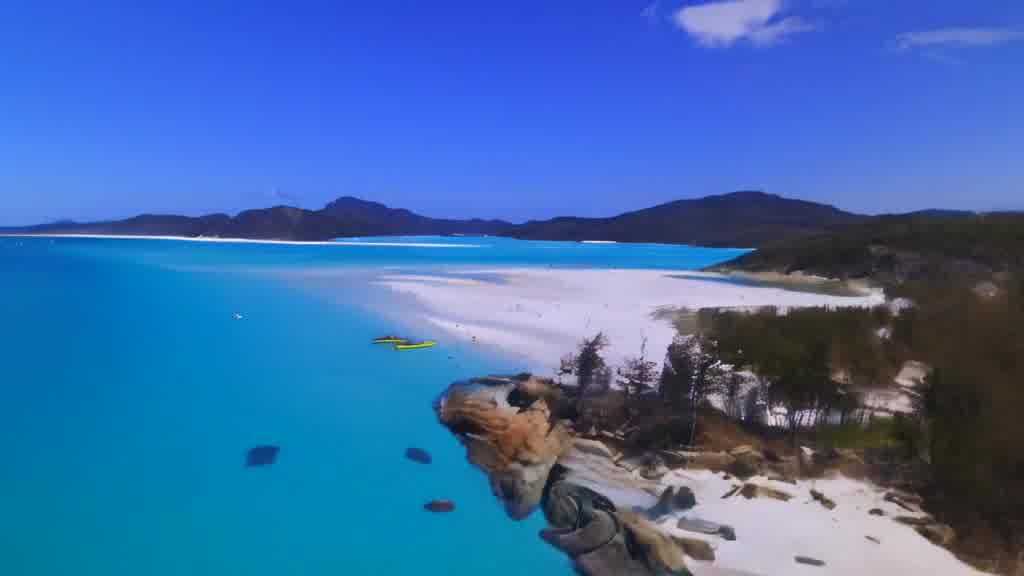}} \raisebox{-.5\height}{\includegraphics[width=0.09\textwidth]{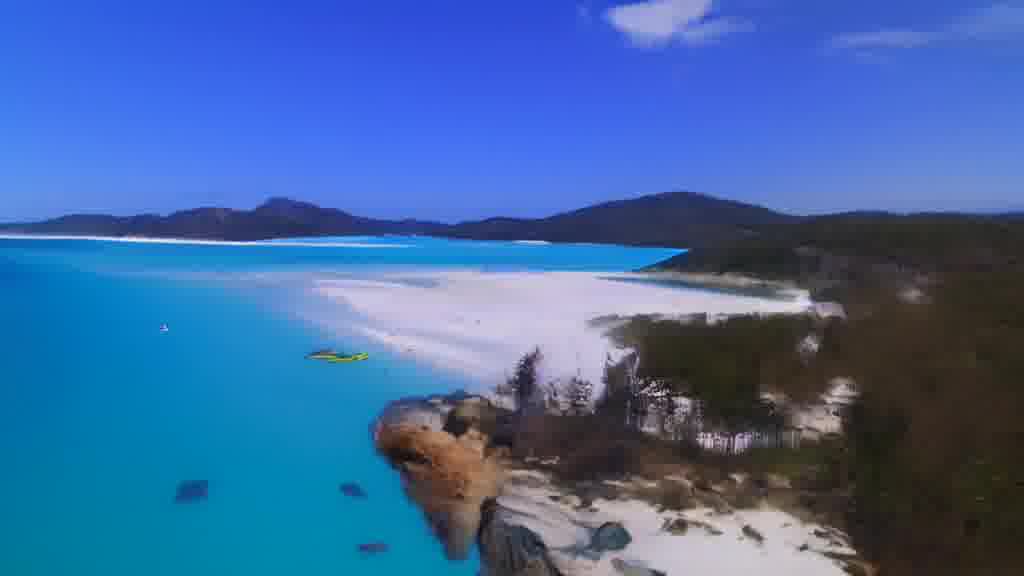}} \\
		{Gen. 3}  & \raisebox{-.5\height}{\includegraphics[width=0.09\textwidth]{figures/cam_grid/first_frame_3.jpg}} & \raisebox{-.5\height}{\includegraphics[width=0.09\textwidth]{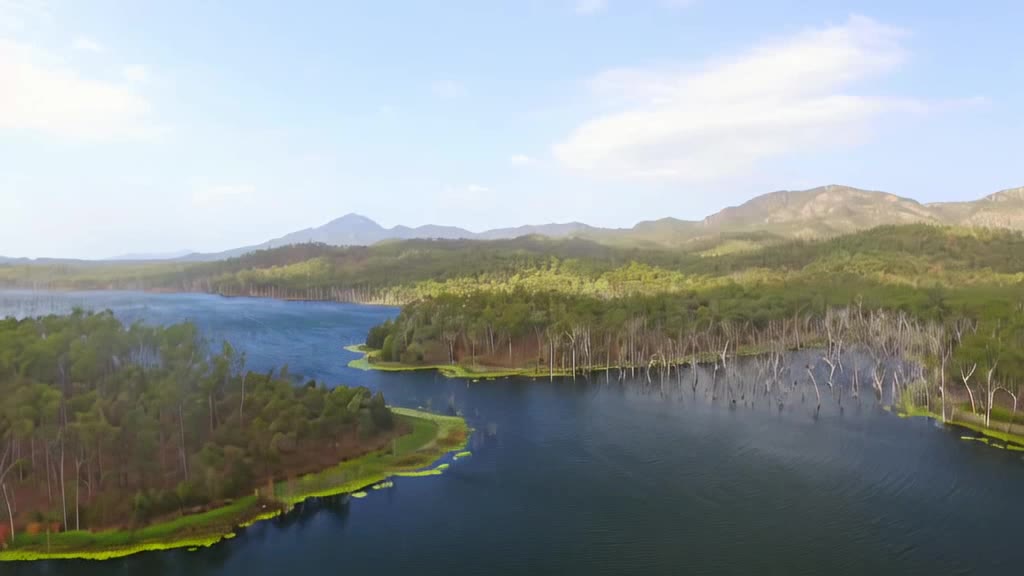}} \raisebox{-.5\height}{\includegraphics[width=0.09\textwidth]{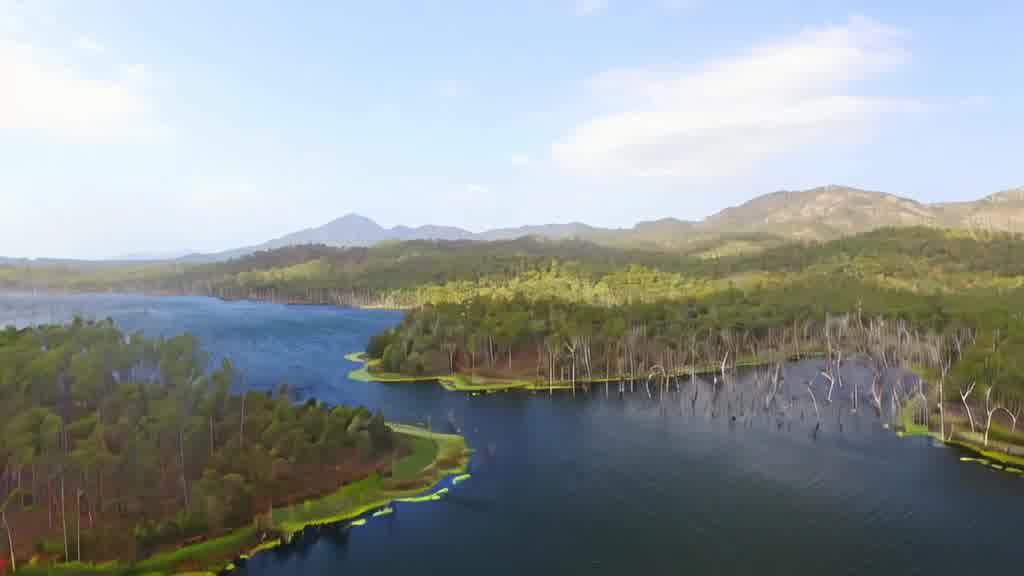}} \raisebox{-.5\height}{\includegraphics[width=0.09\textwidth]{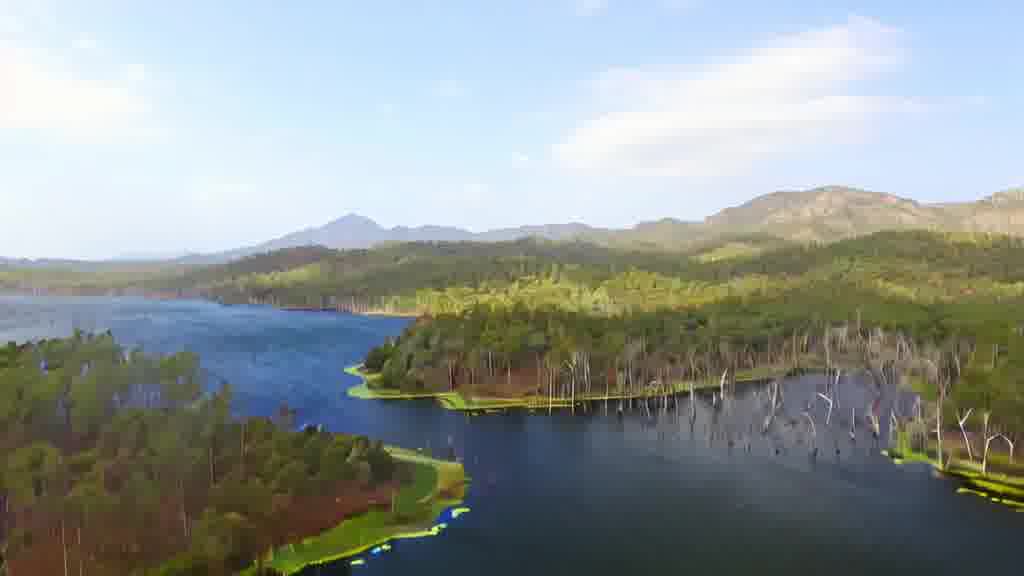}} & \raisebox{-.5\height}{\includegraphics[width=0.09\textwidth]{figures/cam_grid/first_frame_3.jpg}} & \raisebox{-.5\height}{\includegraphics[width=0.09\textwidth]{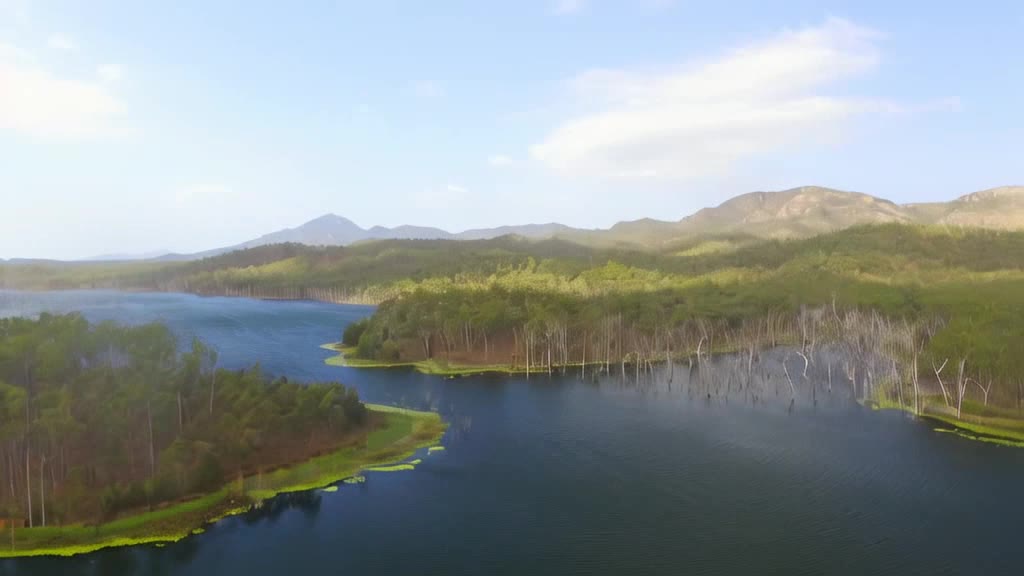}} \raisebox{-.5\height}{\includegraphics[width=0.09\textwidth]{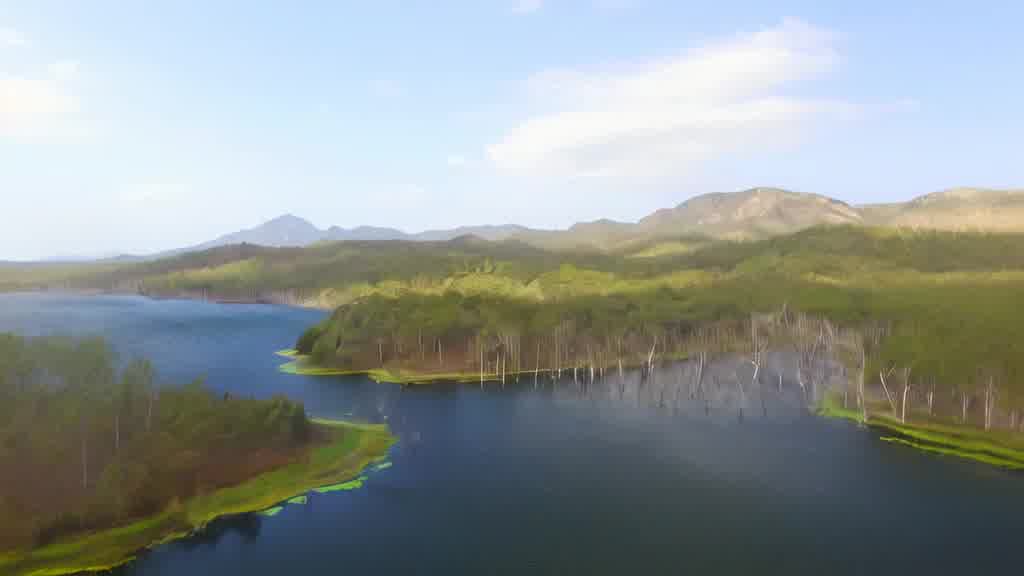}} \raisebox{-.5\height}{\includegraphics[width=0.09\textwidth]{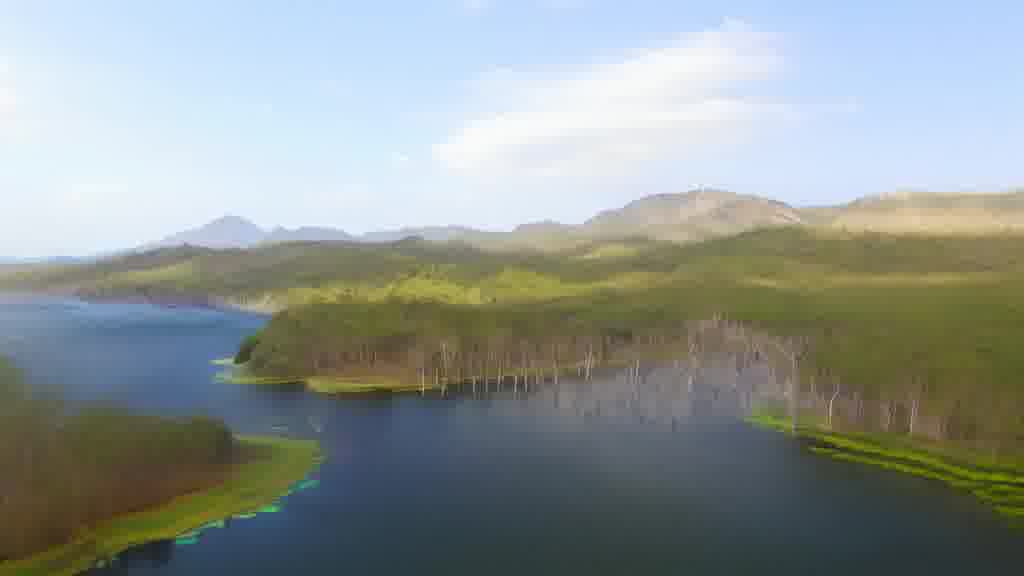}} \\
		{Gen. 4}  & \raisebox{-.5\height}{\includegraphics[width=0.09\textwidth]{figures/cam_grid/first_frame_4.jpg}} & \raisebox{-.5\height}{\includegraphics[width=0.09\textwidth]{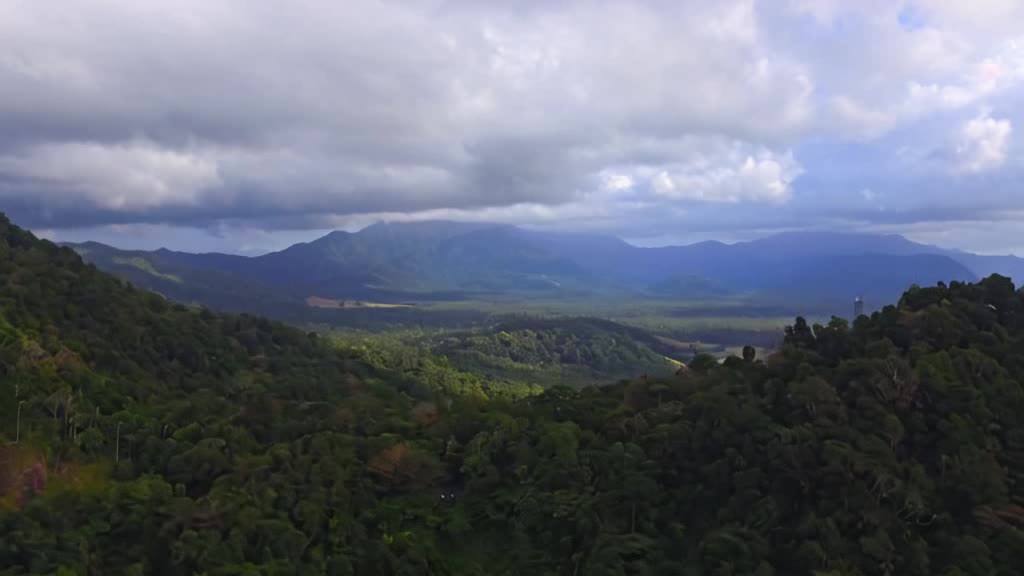}} \raisebox{-.5\height}{\includegraphics[width=0.09\textwidth]{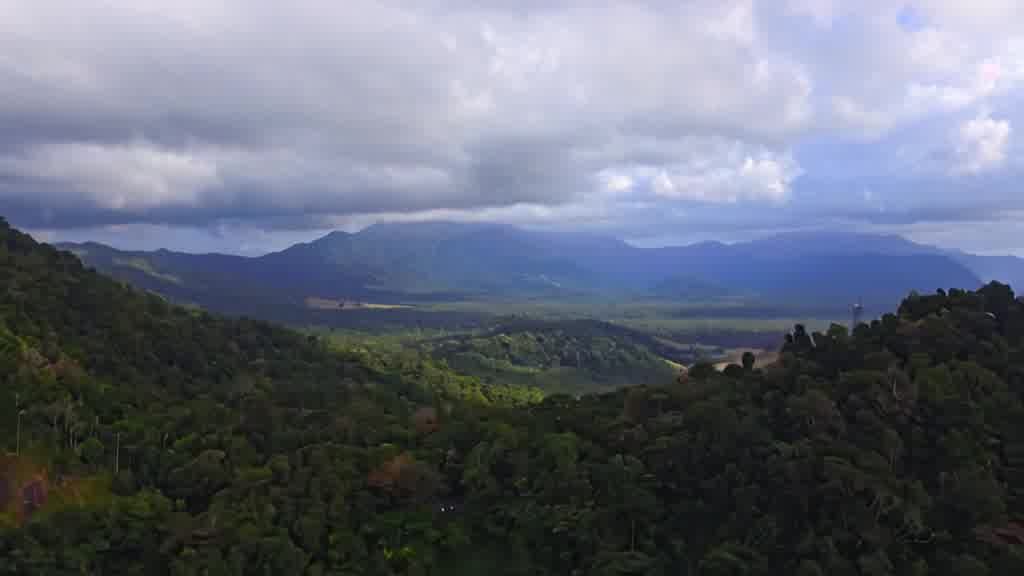}} \raisebox{-.5\height}{\includegraphics[width=0.09\textwidth]{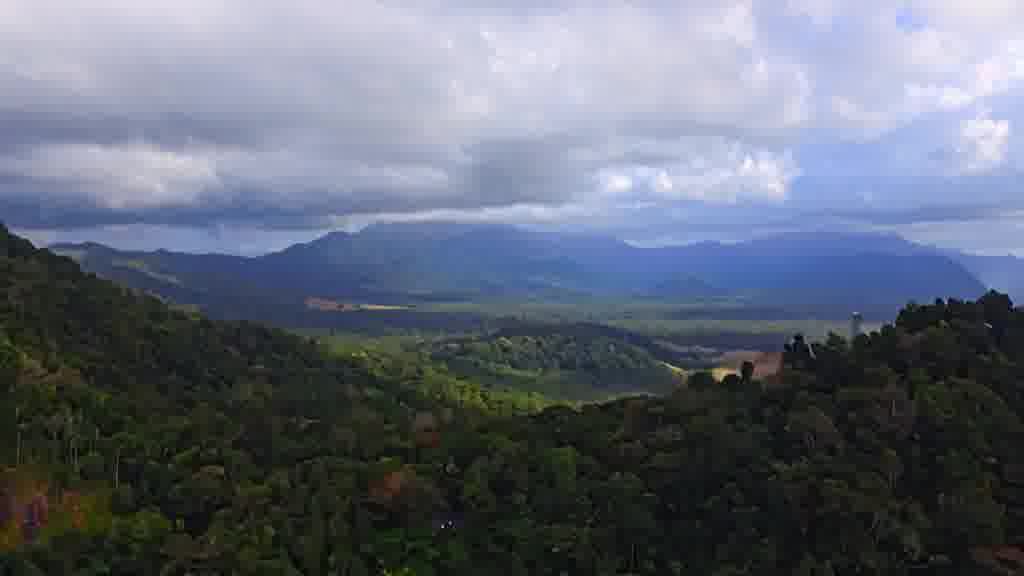}} & \raisebox{-.5\height}{\includegraphics[width=0.09\textwidth]{figures/cam_grid/first_frame_4.jpg}} & \raisebox{-.5\height}{\includegraphics[width=0.09\textwidth]{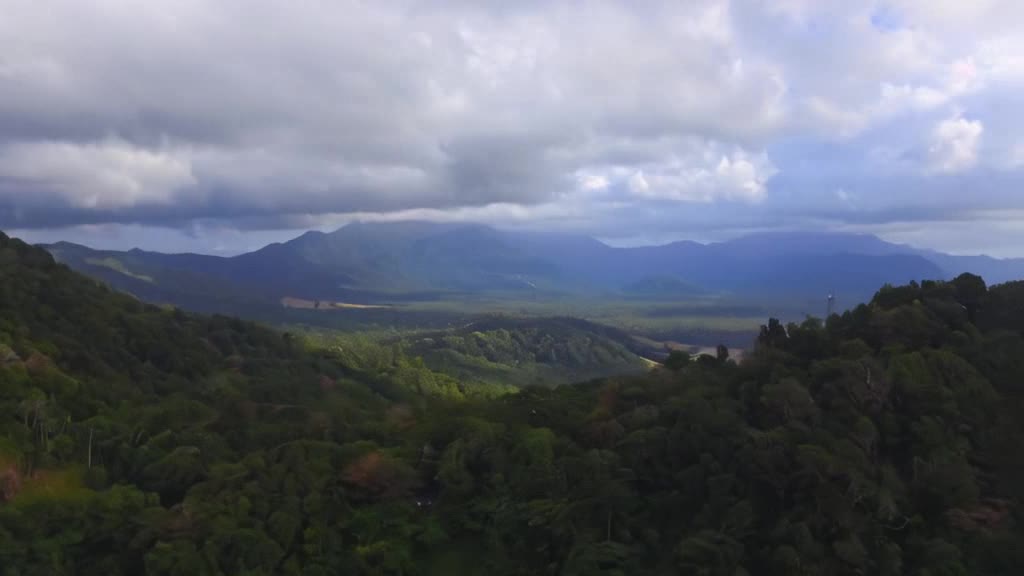}} \raisebox{-.5\height}{\includegraphics[width=0.09\textwidth]{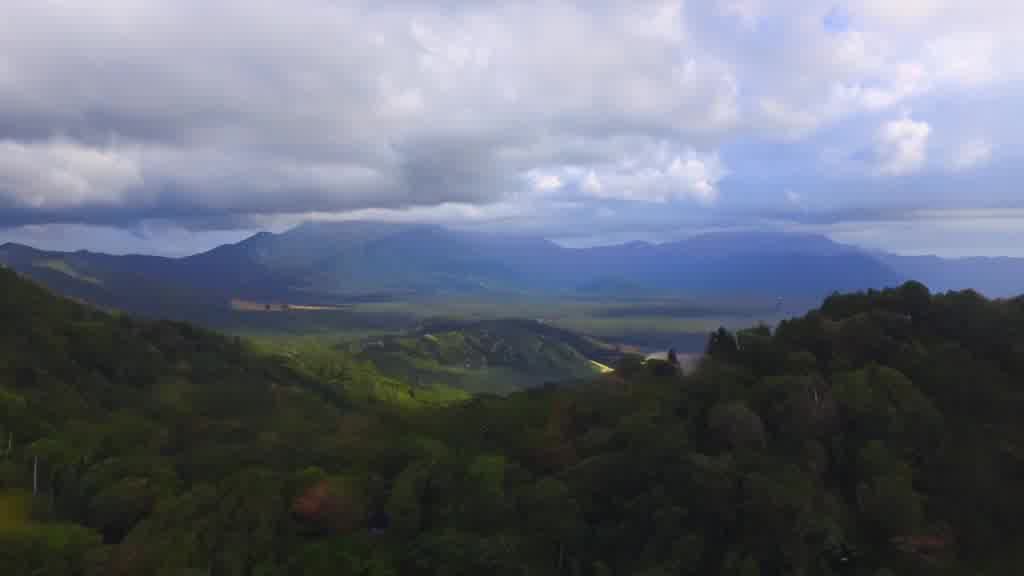}} \raisebox{-.5\height}{\includegraphics[width=0.09\textwidth]{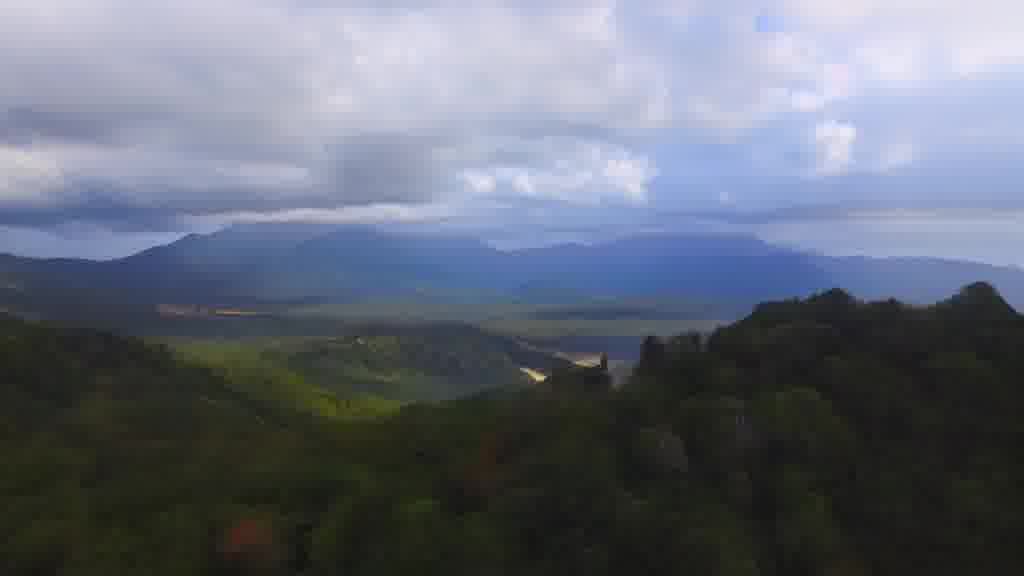}}
	\end{tblr}
	\caption{Camera motion grid. Our learned motion-text embeddings handle camera motions robustly, enabling us to apply a given motion to various target images and various motions to a given target image. The results are best seen in the project website.
	}
	\Description{Grid showing four different camera motion transfer examples arranged in a grid with two rows and two columns. From top to bottom and left to right: turning left, flying right, flying up, and turning right as the camera is flying up. For each grid element, the camera motion is transferred to four different starting images. The four starting images are the same for all motions.}
	\label{fig:results_cam_grid}
\end{figure*}

Fig.~\ref{fig:results_style} shows that our method does not only apply the rough motion category but also its style, even in difficult cases where the domains differ vastly, e.g., transferring the motion of a horse to a cereal box. Furthermore, these examples demonstrate that our method can transfer joint subject and camera motion. Fig.~\ref{fig:results_no_alignment} demonstrates that our method transfers the same semantic rather than spatial motion by applying the same learned motion to a flipped target image. Fig.~\ref{fig:results_extra} shows additional results of our method, where we apply the same optimized motion to different target images to showcase our method's impressive cross-domain capabilities and temporal alignment. Lastly, Fig.~\ref{fig:results_cam_grid} transfers the same four camera motions to four different target images in a grid, demonstrating the robustness of our method for camera motions.

\clearpage

\section{Failure Rate Analysis} \label{sec:failure-rate-analysis}

As is common practice in diffusion-based video generation, we sampled multiple outputs per input and selected the best for display. Quantifying failure rates is difficult, as success can be subjective and depends heavily on the complexity of the motion. Table~\ref{table:quant_eval_by_motion} shows metrics broken down by motion category. The Acc-Top-1 metric reports the percentage of videos correctly classified by an action recognition model~\cite{videomae} and can loosely be interpreted as a success rate for the semantic motion transfer (independent of visual artifacts). Our method achieves much higher accuracy for camera motions (72\%) than for object motions (36\%). It is worth noting that the main challenge in the quantitative evaluation on Something-Something V2~\cite{something_something} stems from the domain gap between the motion reference video and the target image---e.g., transferring a toy car rolling down a book to a pen rolling down a rock---rather than the motion complexity itself. In contrast, our qualitative experiments explored more complex motions to better test the limits of our method, and thus had higher failure rates: approximately 1 in 10 motions resulted in good motion transfers for more than half of the tested target images. To give a more intuitive sense of when our method succeeds or fails, we list motion categories based on how reliably they could typically be transferred in Table~\ref{table:motion_success_rates}.

\begin{table*}[htbp]
	\centering
	\caption{Summary of motion types by performance.}
	\label{table:motion_success_rates}
	\small{
		\begin{tabular}{p{2.5cm}p{13.5cm}}
			\toprule
			Performance & Motion Types \\
			\midrule
			\rowcolor{lightgreen}Motions good \newline Quality good &
			\textbf{Camera motions:} bird’s-eye panning/zooming/rotation, panoramas, smooth drone flights, object tracking \newline
			\textbf{Common head motions:} nodding, facial expressions (surprise, yawning, opening mouth) \newline			
			\textbf{Some full-body motions:} walking (human to human, four-legged to four-legged), jumping jacks \newline
			\textbf{Handcrafted motions with small domain gap:} colliding/passing circles of similar shapes/colors \\
			\midrule
			\rowcolor{lightyellow}Motions good/okay \newline Quality bad &
			\textbf{Fast motions:} boxing, fast running animals (left/right limb confusion) \newline
			\textbf{Head motions with drastic appearance changes:} frontal-to-profile rotations, extremely wide mouth openings, revealing teeth from closed mouth \newline
			\textbf{Some full-body motions:} jumping forward far, walking into jump, karate kicks \newline
			\textbf{Handcrafted motions where target object has many details:} texture-free bouncing ball transferred to soccer ball with many patches, stick figure to detailed human / two-legged animal \\
			\midrule
			\rowcolor{lightred}Motions bad &
			\textbf{Fine-grained motions:} tongue movement, eyebrow raises, small/distant actions \newline
			\textbf{Emerging objects:} hand entering frame \newline
			\textbf{Large domain gap:} human face motions to minimalistic cartoon or ostrich, human to kangaroo, bouncing ball to landscape scene with sun \newline
			\textbf{Complex full-body motions:} running into forward roll, handstands, swinging arm punch, yoga/stretching \\
			\bottomrule
			\\
		\end{tabular}
	}
\end{table*}

\begin{figure*}[htbp]
	\centering
	\begin{tblr}{
			vline{3} = {2-3}{dashed},
		}
		{Reference} & \raisebox{-.5\height}{\includegraphics[width=0.15\textwidth]{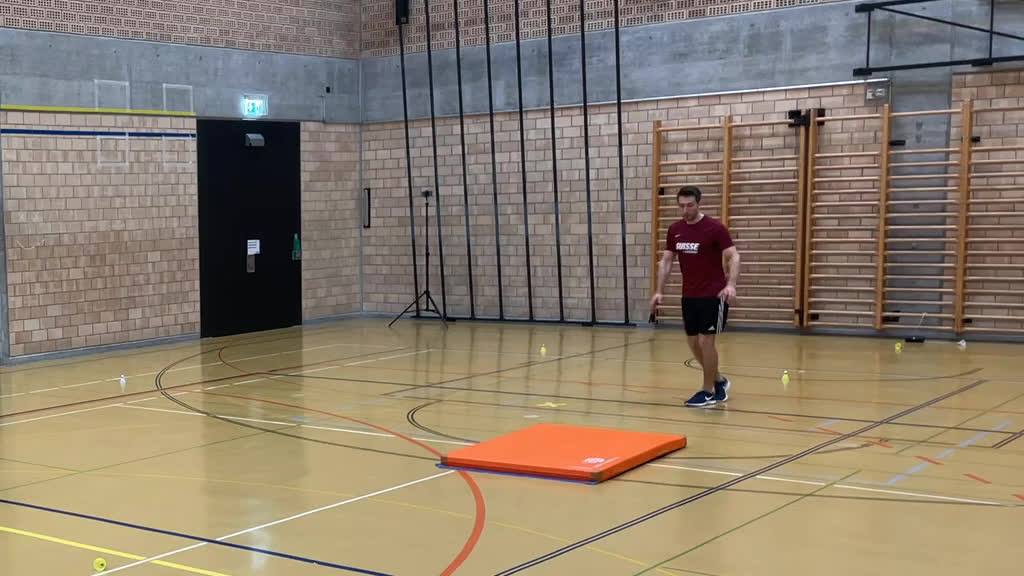}} & \raisebox{-.5\height}{\includegraphics[width=0.15\textwidth]{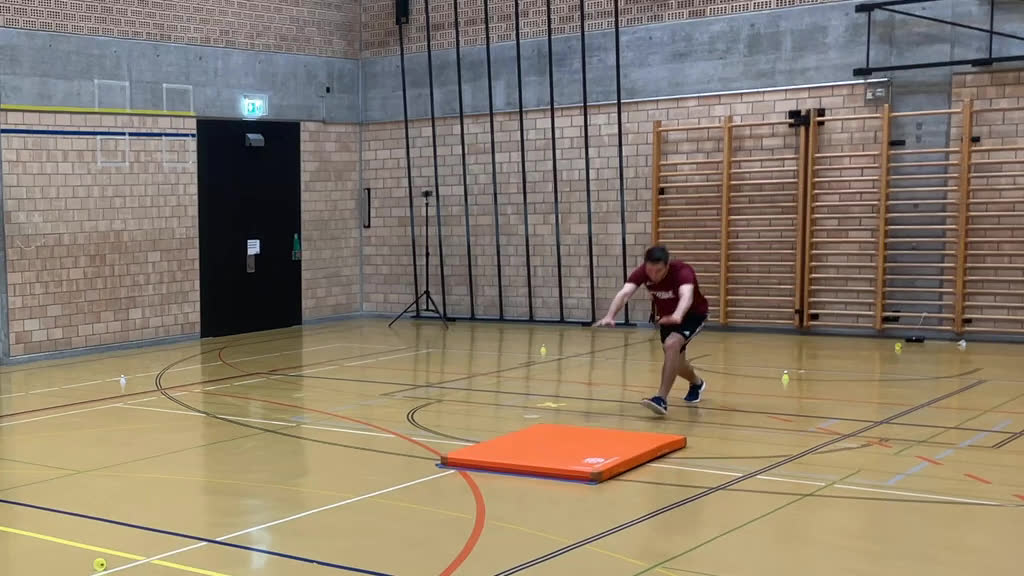}} \raisebox{-.5\height}{\includegraphics[width=0.15\textwidth]{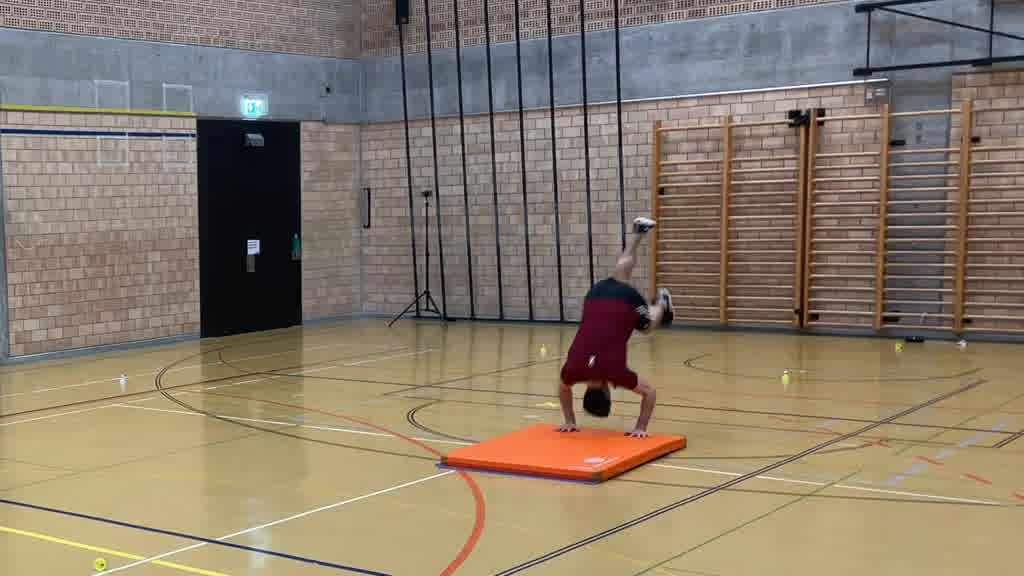}} \raisebox{-.5\height}{\includegraphics[width=0.15\textwidth]{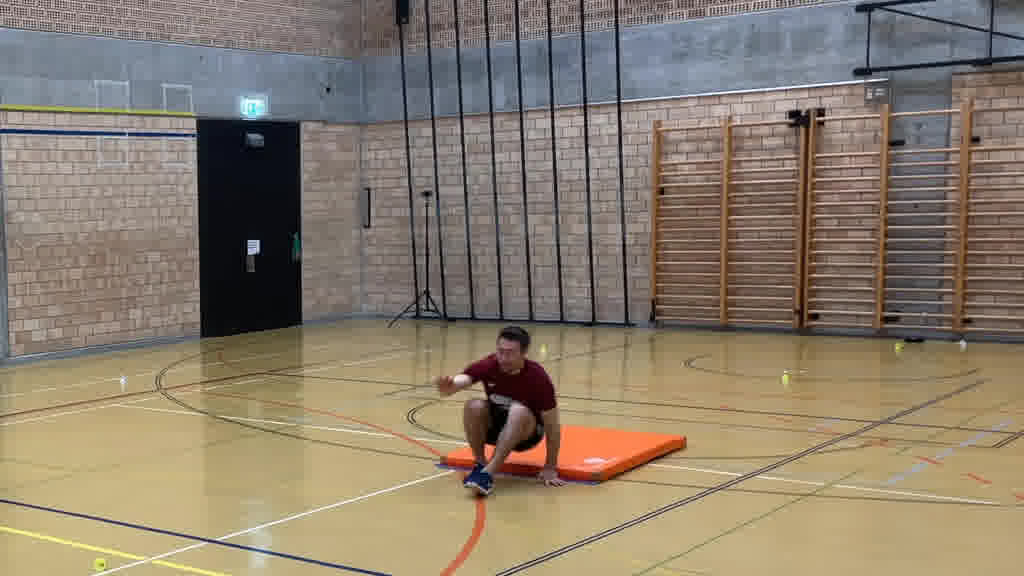}} \\
		{Reconstruction}  & \raisebox{-.5\height}{\includegraphics[width=0.15\textwidth]{figures/failure_recon/frames_input/001.jpg}} & \raisebox{-.5\height}{\includegraphics[width=0.15\textwidth]{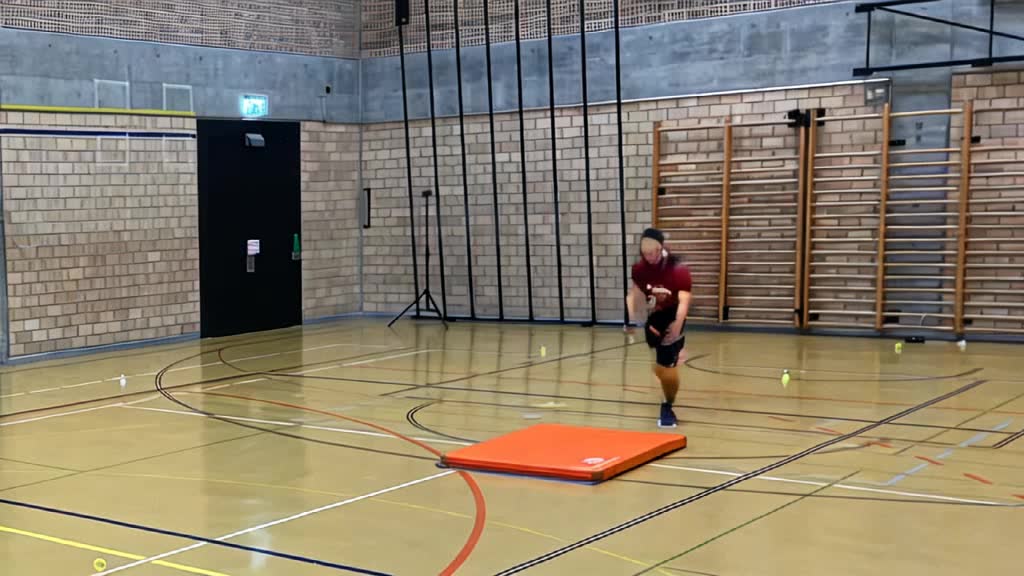}} \raisebox{-.5\height}{\includegraphics[width=0.15\textwidth]{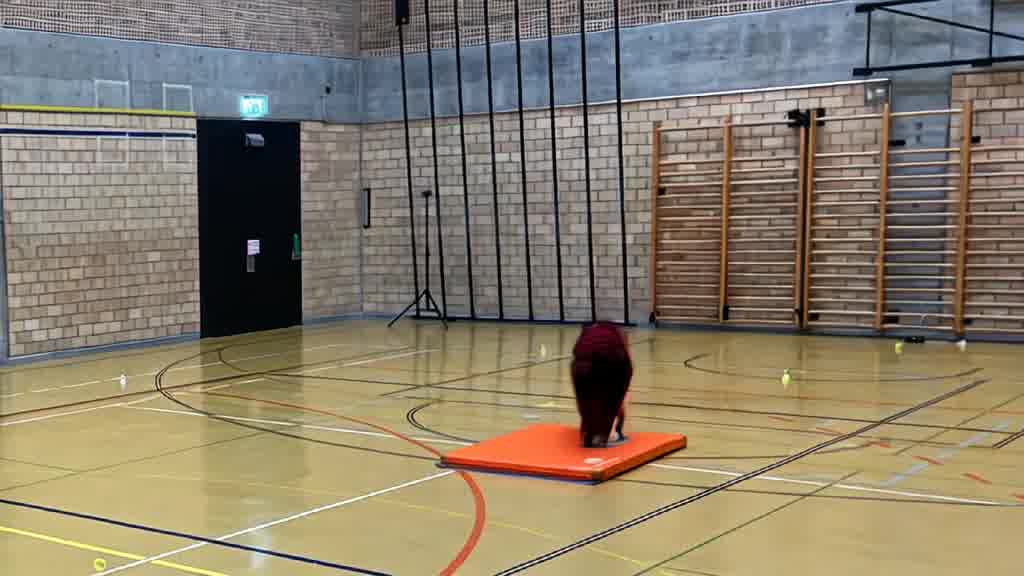}} \raisebox{-.5\height}{\includegraphics[width=0.15\textwidth]{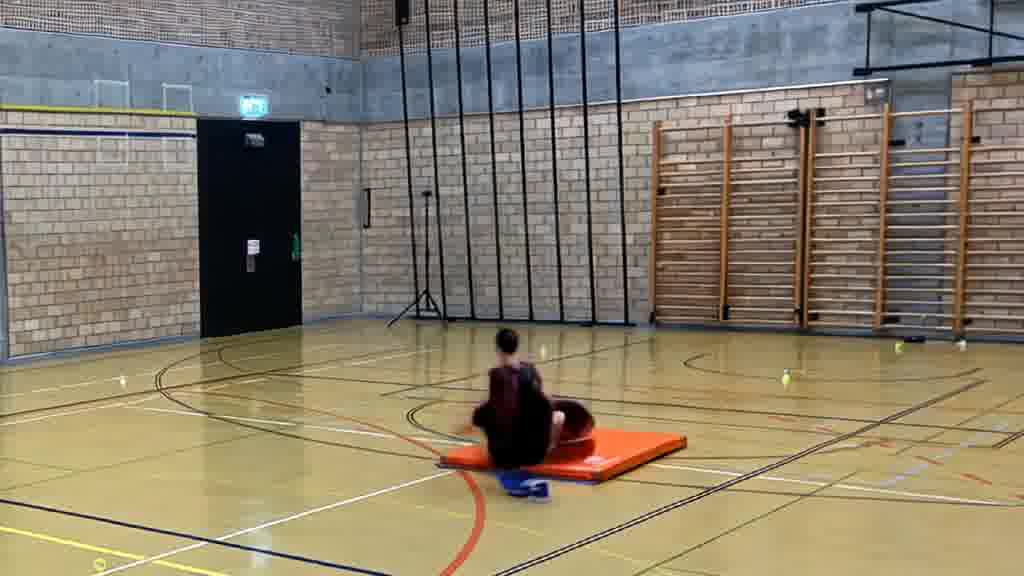}} \\
		{Transfer}  & \raisebox{-.5\height}{\includegraphics[width=0.15\textwidth]{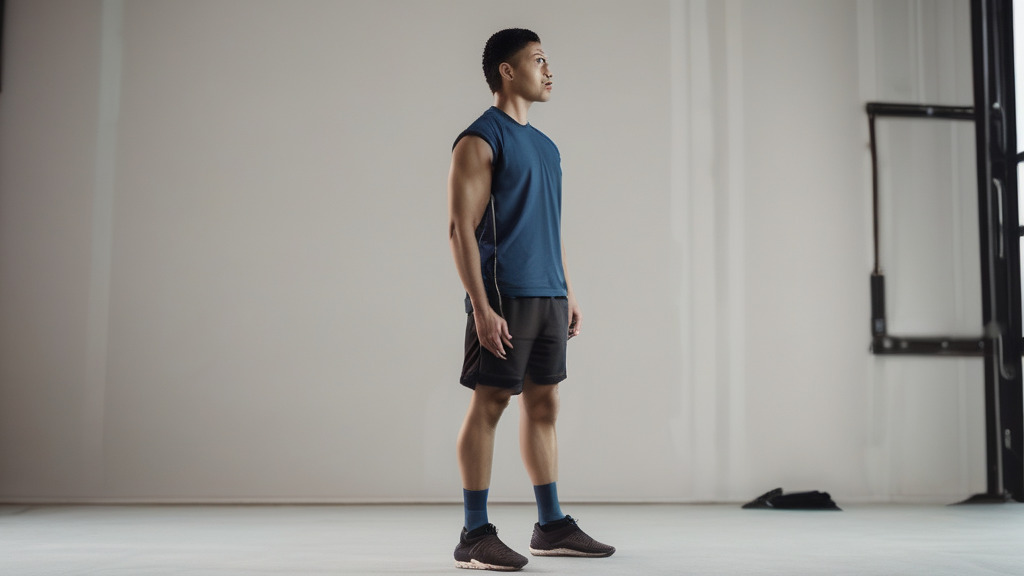}} & \raisebox{-.5\height}{\includegraphics[width=0.15\textwidth]{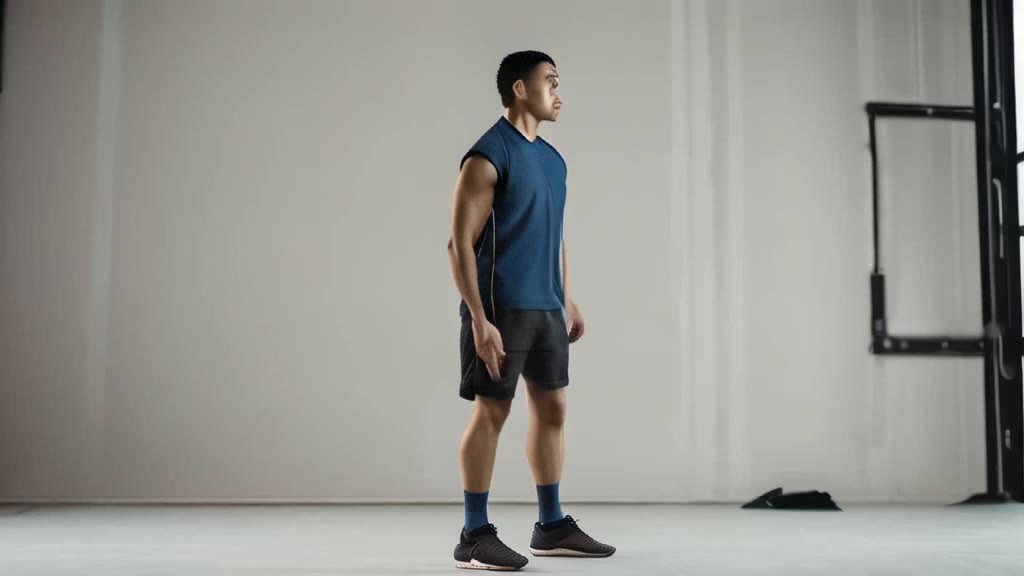}} \raisebox{-.5\height}{\includegraphics[width=0.15\textwidth]{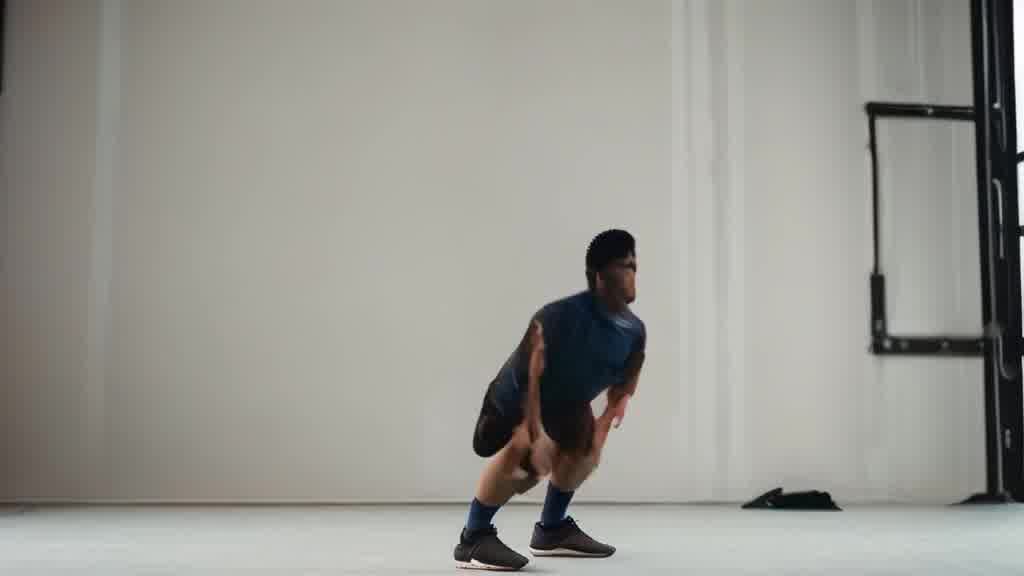}} \raisebox{-.5\height}{\includegraphics[width=0.15\textwidth]{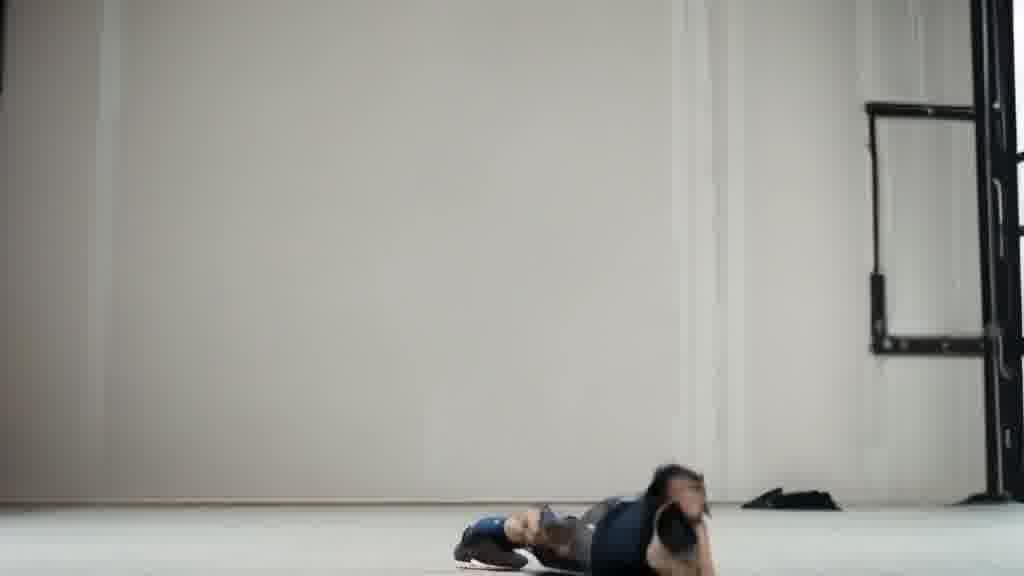}}
	\end{tblr}
	\caption{Failure case with poor reconstruction. When the optimized motion-text embedding fails to accurately reconstruct the reference motion, the subsequent transfer to a new target typically fails as well.}
	\Description{In the top row, i.e., the reference video, a person is running from right to left and then performs a forward roll on a mat. In the second row, this motion is applied to the first frame of the reference video, but the reconstruction fails: rather than capturing the correct semantic motion, the person collapses into a vague, blob-like shape. In the third row, the same motion is transferred to a different target person, who also collapses into a blob instead of performing the intended forward roll.}
	\label{fig:failure-recon-transfer}
\end{figure*}

In our experiments, we observed that the reconstruction quality of the motion reference video, i.e., applying the optimized motion-text embedding to the first frame of the motion reference video, is a strong indicator of the final motion transfer performance. If the model fails to reconstruct the reference video accurately, it suggests that the optimized motion-text embedding does not effectively capture the semantics of the motion. In such cases, applying the same embedding to a different target image typically also fails. This issue is illustrated in Fig.~\ref{fig:failure-recon-transfer}, where the reconstructed video collapses the person into a blob-like shape rather than depicting a realistic forward roll. The same collapse occurs when transferring the motion to a different target image. One contributing factor may be the use of a simple mean-squared error loss, which can lead to pixels being placed in roughly the correct spatial positions, even if the resulting motion does not semantically match the reference. Another potential reason for failure is that some motions may be out-of-domain for the pre-trained Stable Video Diffusion~\cite{svd}. Since our approach optimizes only the input motion-text embedding without fine-tuning the model itself, it is challenging to capture entirely novel or complex motion types that the model has not seen during training. To mitigate these issues, we encourage future work to explore more semantically meaningful loss functions, regularize the embedding to remain closer to the original CLIP~\cite{clip} space, or adopt recent video diffusion models with stronger motion understanding, such as VideoJAM~\cite{videojam}.

\end{document}